\documentclass[10pt, letterpaper]{article}

\usepackage[margin=1in]{geometry}

\usepackage{booktabs}
\usepackage{booktabs,multirow,array}
\usepackage{nicematrix}
\usepackage{tabularx}

\usepackage[round]{natbib}

\usepackage{graphicx} 
\usepackage{microtype}

\usepackage{amsfonts,amsmath,amssymb,amsthm}       
\usepackage{natbib}
\usepackage{array}
\usepackage{todonotes}
\usepackage[dvipsnames]{xcolor}
\usepackage{subcaption,float,graphicx}
\usepackage{thmtools,thm-restate}
\usepackage{url}            
\usepackage{booktabs}       
\usepackage{nicefrac}       
\usepackage{microtype}      
\usepackage{mathrsfs}
\usepackage[
    colorlinks=true,
    linkcolor=blue,
    filecolor=magenta,      
    urlcolor=cyan,
    citecolor=blue,
]{hyperref}

\usepackage{amsmath,amsfonts,bm}

\def\1{\bm{1}}

\DeclareMathAlphabet{\mathsfit}{\encodingdefault}{\sfdefault}{m}{sl}
\SetMathAlphabet{\mathsfit}{bold}{\encodingdefault}{\sfdefault}{bx}{n}

\newcommand{\beq}{\begin{equation}}
\newcommand{\eeq}{\end{equation}}

\newcommand\zeros{\mathbf{0}}

\renewcommand{\v}{\mathbf{v}}

\newcommand{\x}{\mathbf{x}}
\newcommand{\y}{\mathbf{y}}

\newcommand{\cN}{{\cal N}}

\newcommand{\cD}{{\cal D}}

\theoremstyle{definition}

\newcommand {\commentout}[1] {}

\def\ints{{{\rm Z} \kern -.35em {\rm Z} }}  
\def\smallints{{{\rm Z} \kern -.3em {\rm Z} }}  
\def\pints{{{\rm I} \kern -.15em {\rm N} }}      
\newcommand{\reals}{\mathbb R}

\def\cplx{{{\rm I} \kern -.45em {\rm C} }}       
\def\l2{\rm {\mathcal L}^{2}(\reals)}            

\newcommand{\be}{\begin{eqnarray}}
\newcommand{\ee}{\end{eqnarray}}
\newcommand{\bea}{\begin{eqnarray}}
\newcommand{\eea}{\end{eqnarray}}
\newcommand{\beaa}{\begin{eqnarray*}}
\newcommand{\eeaa}{\end{eqnarray*}}
\newcommand{\bnad}{\begin{nad}}
\newcommand{\enad}{\end{nad}}

\usepackage{enumerate,color,multirow}
\usepackage{xpatch}

\usepackage{xcolor}
\newcommand{\pc}[1]{{\color{blue} #1}}

\usepackage{algorithm}
\usepackage{algorithmic}

\usepackage{amsmath}
\usepackage{amssymb}
\usepackage{mathtools}
\usepackage{amsthm}

\usepackage{comment}

\title{It Takes Two to Tango: Two Parallel Samplers Improve Quality in Diffusion Models for Limited Steps}
\author{Pedro Cisneros-Velarde\\VMware Research\\\texttt{pacisne@gmail.com}}
\date{}

\begin{document}

\maketitle

\begin{abstract}
We consider the situation where we have a limited number of denoising steps, i.e., of evaluations of a diffusion model. 
We show that two parallel processors or samplers under such limitation can improve the quality of the sampled image. 
Particularly, the two samplers make denoising steps at successive times, and their information is appropriately integrated in the latent image. 
Remarkably, our method is simple both conceptually and to implement---it is plug-\&-play, model agnostic, and does not require any additional fine-tuning or external models.
We test our method with both automated and human evaluations 
for different diffusion models. 
We also show that a naive integration of the information from the two samplers lowers sample quality.
Finally, we find that adding more parallel samplers does not necessarily improve sample quality. 
%
%
\end{abstract}

\section{Introduction}
\label{sec:intro}

Denoising diffusion models~\citep{sohl-2015-noneqthermo,song-2019-genmodelestigr,Ho-2020-ddpm} are a family of generative models that have been widely used due to their success in generating high-quality images~\citep{dhariwal-2021-diffusion}---becoming the standard of image generation---and 
data of diverse modalities, such as 3D objects~\citep{shi-2024-mvdream} and videos~\citep{gupta-2025-photorealisticvideo}. Their extensive use 
has 
impacted 
various fields such as privacy~\citep{carlini-2023-exttrain}, arts~\citep{jiang-2023-artai}, robotics~\citep{carvalho-2025-motionpl}, forecasting~\citep{meijer2024risedifftimeseries}, medical imaging~\citep{kazerouni-2023-dmmedical}, 
inverse problems~\citep{chung-2022-conditionalinverseproblems}, etc.

The inference process of diffusion models requires the repeated denoising of 
%
a latent image, which was initialized with Gaussian noise, until it becomes the sampled data~\citep{Ho-2020-ddpm}. The larger the number of \emph{denoising steps}, the better the quality of the generated sample. However, this denoising process 
%
requires the repeated and computationally expensive evaluation of a 
neural model---thus, there is a trade-off between sample quality and inference efficiency. 
Consequently, considerable efforts have been 
made 
to speed up inference without sacrificing excessive image quality---not necessarily by greatly reducing the number of evaluation steps, but the total 
\emph{time} it takes to run the inference.    
%
Alternatively, other generative models 
have been formulated for faster inference such as consistency models~\citep{song-2023-consistency} and flow matching~\citep{lipman-2023-flowmatching}. Nonetheless, diffusion models are more widely adopted 
and are still at the forefront of 
image generation: they can produce 
higher quality samples 
and have a relatively easier training/fine-tuning process~\citep{heek-2024-multistepconsist,schusterbauer-2025-diff2flow}, despite the slower sampling. 
%
%
Regarding approaches to speed up the inference of diffusion models, we have:
%
removing stochasticity from the sampling trajectory~\citep{song-2021-denoisingimplicit,song-2021-scorebased,lu-2022-dpmsolver,shih-2023-parallel}; altering the scheduling of the denoising steps~\citep{watson-2021-learningefficientlysample,ye-2025-scheduleonthefly}; alternating the use of multiple models of different capabilities~\citep{liu-2023-omsdpmmodelschedule,li-2025-morse}; distilling models that require less steps~\citep{salimans-2022-progressive}; and 
leveraging parallel processing~\citep{shih-2023-parallel,hu-2025-diffusion}.
%
%
Some of these approaches
%
%
can 
sacrifice a degree of image quality, while others  
need a large number of steps to achieve good quality or 
need to solve additional complex problems (e.g., optimization or training of models). 
Yet, arguably, greatly reducing the number of denoising steps 
is \emph{in practice} the 
most efficient 
way to sample an image, albeit with image quality loss. Then the question is: \emph{Can we somehow improve image quality during the sampling process in the domain of low number of denoising steps?}
%

Concretely, we explore this question in the context of parallel computing:
%
we 
assume access to parallel processors.
%
%
Then, the research question we focus is: \emph{Is it possible to use parallel processors to 
improve image quality during the sampling process 
while still running a low number of denoising steps per processor?} 
In the interest of practical relevance,
a solution method to this question must be
as \emph{simple} 
as possible. Particularly, we want a solution method to (i) be \emph{plug-\&-play} (which excludes solving additional optimization problems or training other models); 
(ii) be \emph{agnostic} to the diffusion model being employed (which excludes any fine-tuning of diffusion models or using white-box knowledge); 
(iii) not require additional external neural models (which otherwise increase the complexity of the solution and introduce extra computation); and (iv) use the least amount of parallel processors---just two. Thus, 
we look for a 
minimalist solution to enhance image quality for a low number of denoising steps. To the best of our knowledge, this problem setting has not been explored in the literature.
We remark that 
(iii) and (iv) exclude the use of \emph{ensemble methods} (more information in Appendix~\ref{app:ensemble}). 

In this work, we respond to our research question with an affirmative answer and propose a simple method, both in concept and implementation, that satisfies every desiderata (i) to (iv).
Our method, by continuously integrating information from two 
processors across the denoising process, is able to modify different image attributes of the sampled image, such as contrast, brightness, and sharpness of features---sometimes leading to semantic changes---in order to improve image quality. 

We emphasize that our problem setting is not about \emph{accelerating} the inference process of diffusion models, 
but about 
enhancing sample quality on 
the restricted environment of low 
number of denoising steps. 
Indeed, 
our problem setting 
is orthogonal to 
the one of improving inference latency, since both can mutually benefit each other. 
Finally, we mention that 
%
our work contributes to the literature on techniques that seek to \emph{drive} or \emph{control} the sampling process to produce images of certain desirable attributes. These techniques could potentially be used 
in tandem with our work. 
For example, 
%
we have the steering 
of the sampling process 
to: balance sample fidelity and diversity within the image manifold~\citep{dhariwal-2021-diffusion,ho-2021-classifierfree, nichol-2022-glide}; 
sample from low-density regions~\citep{Sehwag-2022-lowregion}; 
or to sample according to 
semantic modalities~\citep{bansal-2024-universal}. We also have the addition of conditioning controls to affect image semantics~\citep{zhang-2023-addingcontroltexttoimage}, as well as constraints on the image generation~\citep{graikos-2022-diffusion}. Enhancing image quality can also benefit 
super-resolution 
applications~\citep{Moser-2025-DMSuper-resolution}. 
%

%

\subsection*{Contributions}
We propose SE2P, a method to improve the quality of sampled images under a low number of denoising steps by simply using two parallel processors. 
Our method is plug-\&-play
and does not require any fine-tuning or the evaluation of external neural models. 
    
SE2P is simple in both concept and implementation. Conceptually, it is based on the integration of latent predictions of one processor into the latent values of the other processor. 
The implementation is at the scheduler level, which makes it model agnostic.
    
We test the effectiveness of SE2P on models that have different backbones (U-Net and transformers), that are conditional and unconditional, and that operate on both pixel and latent spaces. We perform a qualitative study showing the visual changes done by our method, as well as limitations. We also perform two quantitative studies: a human evaluation one and an automated one using 
different image quality assessment 
metrics. 
SE2P overall shows better performance than the baseline.
%
    
Finally, we show that: (i) removing the prediction from the integration process and directly integrating the latent values of both processors leads to image quality loss; and (ii) adding more parallel processors does not necessarily lead to better sample quality.
    %

\section{Model Setting}
\label{sec:model-setting}
We briefly introduce denoising probabilistic models~\citep{sohl-2015-noneqthermo, Ho-2020-ddpm}.
Given 
a sampled 
vector 
$\x_0\sim \cD$ from data distribution $\cD$, the forward diffusion process iteratively   
adds Gaussian noise 
in $T$ steps:
\begin{equation}
\label{eq:forward}
\x_t = \sqrt{1-\beta_t}\x_{t-1} + \sqrt{\beta_t}\epsilon  
\end{equation}
with $\epsilon\sim\cN(\zeros,I)$ and $t=1,\dots,T$; and where $(\beta_t)_{t=1}^T$ is the variance schedule, $\zeros$ the vector of all zeros, and $I$ the identity matrix. 
Now, the goal 
is to train a 
\emph{diffusion model} 
such that given a Gaussian random variable $\x_{T}\sim\cN(\zeros,I)$, 
it aims to generate 
a sample $\x_0\sim\cD$ after following some reverse-time \emph{denoising process}.
%
Taking the \emph{implementation} in \citep{Ho-2020-ddpm}:
\begin{equation}
\label{eq:denoise}
\x_{t-1} = \frac{1}{\sqrt{\alpha_t}}\left(\x_t-\frac{1-\alpha_t}{\sqrt{1-\bar{\alpha}_t}}\varepsilon_{\theta}(\x_t,t)\right) + \sqrt{\tilde{\beta}_t} \epsilon,
\end{equation}
with $\epsilon\sim\cN(\zeros_n,I_n)$ and $t=T,\dots,1$; and where $\alpha_t:=1-\beta_t$, $\bar{\alpha}_t:=\prod^{t}_{s=1}\alpha_s$, $\tilde{\beta}_t:=\frac{1-\bar{\alpha}_{t-1}}{1-\bar{\alpha}_t}\beta_t$, and
$\epsilon_\theta$ is the (pretrained) diffusion model. $T$ is now the number of denoising steps. 
$\epsilon_\theta(\x_t,t)$ estimates the level of noise present in 
the value of the latent $\x_t$ at the denoising step $t$. %
This parameterization of the diffusion model in~\eqref{eq:denoise} defines the standard \emph{DDPM scheduler}.
%
%
%
%
Notice that both the reverse~\eqref{eq:denoise} and forward~\eqref{eq:forward} process are  
stochastic.
%
%
Informally, 
the denoising process seeks to do the ``opposite'' as the forward process---
more details 
in~\citep{Ho-2020-ddpm}. 
Since the denoising process is the \emph{inference} process for generating samples, 
it follows that 
the diffusion model is trained using samples from $\cD$. 

The DDPM scheduler typically uses $T=1000$ for sample generation. In practice, a sample $\x_0$ resulting from a pretrained model does not \emph{exactly} correspond to a sample from $\cD$. This is 
more noticeable when the denoising process 
has 
\emph{jump sampling}, i.e., we iterate~\eqref{eq:denoise} over a subsequence of steps $t=t_{N-1}, t_{N-2}, \cdots, t_{0}$ with $N<T$ instead of the standard $t=T,\dots,1$.
The practical advantage of jump sampling is to generate 
samples with less model evaluations, which are computationally the most expensive part of the inference process.  
%
%
%
%
A \emph{low iteration regime} has jump sampling with $N\ll1000$
%
using the following 
convention: since $N<500$, the $N$ denoising steps are approximately equally spaced across the $1000$ standard steps, ending with time-step $1$. Examples are found in Appendix~\ref{app:exp}.

Finally, the particular diffusion 
model $\epsilon_\theta$ in~\eqref{eq:denoise} is \emph{unconditional}, i.e., it does not take any additional input 
besides the 
(noisy) latent 
and the time-step.
A \emph{conditional model}, instead, has additional inputs 
%
%
to somehow condition the content of the final sampled image $\x_0$; 
%
e.g., 
an indicator of image classes~\citep{dit-peebles-20230-DiT}, text embeddings~\citep{LD_Rombach_2022_CVPR}, etc.
%

\textbf{Notation:} For simplicity, we denote the right-hand side of equation~\eqref{eq:denoise} by the notation $\operatorname{denoise}(t,\x_{t},\epsilon_\theta)$. 

\section{The SE2P Algorithm}
\label{sec:algorithm}
We propose the algorithm ``Sample Enhancement Using Two Processors'' or SE2P, described in Algorithm~\ref{alg:main_SE2P} 
(where ``Proc $j$'', $j\in\{0,1\}$, at the end of an instruction comment indicates that ``Processor $j$'' is running such instruction).
%
%
Consider a jump sampling regime $(t_k)_{k=0}^{N-1}$. At step $t_k$, one processor obtains a noisy latent and, \emph{in parallel}, another processor obtains one evaluated at the consecutive step $t_k+1$; see lines 13 to 17.
%
%
%
%
The main idea of SE2P is to \emph{integrate} the information of both processors' latents 
throughout the 
denoising process with the goal of improving the quality of the sampled image at the end of it.
%
The question is how to perform this integration. 
One naive way 
is 
to directly combine both processors' latents through a convex combination. Instead, what we first do is to take the latent 
from 
the processor at step $t_k+1$ and use it to produce a \emph{predictor} of the latent at step $t_k$---resulting in $\hat{\x}_{\operatorname{pred}}$ from line 8. This predictor is computed from an estimate of the fully-denoised image $\hat{\x}_0$ found in line 6, 
based on~\citep[Equations~(7),(15)]{Ho-2020-ddpm} and~\citep[Remark~1]{chung-2023-diffusion}.
Particular to SE2P,  
we scale the variance of the 
noise to be injected in the predictor 
by some constant $\rho>0$ in line 8 (technically, the variance is \emph{scaled} by $\rho^2$), which we call the \emph{variance scaling} parameter. Setting $\rho=1$ provides the standard predictor. 
Now that we have 
$\hat{\x}_{\operatorname{pred}}$, we \emph{integrate} its information with the latent from step $t_k$ through a simple \emph{convex combination} in line 9---the \emph{mixing parameter} $\gamma$ weights the importance 
of the 
latent. 
%
%
%
Finally, we note that computing $\hat{\x}_{\operatorname{pred}}$ does not need an extra evaluation of the diffusion model because of the previously stored variable $\mathbf{v}^{(0)}$ in line 6. 
We summarize SE2P in Fig.~\ref{fig:Se2P-algorithm} 
More technical notes are in Appendix~\ref{app:technicalnotesS2EP}.

\begin{algorithm}[t!]
  \caption{Sample Enhancement Using Two Processors (SE2P)}
  \label{alg:main_SE2P}
\begin{algorithmic}[1]
    \STATE {\bfseries Input:} $(t_{N-1},\dots,t_{0})$, $(\beta_t)_{t=0}^T$, pretrained model $\epsilon_\theta$, parameters $\rho>0$, $\gamma\in(0,1)$ 
    \STATE $\x^{(0)}_{N-1}\sim\cN(\zeros,I)$, $\x^{(1)}_{N-1}=\x^{(0)}_{N-1}$ 
%
%
    \FOR{$k=N-1,\cdots,0$}
        \IF{$k\neq N-1$}
            \STATE $\hat{t}_{k} = t_k + 1$ \# Proc $0$
            \STATE $\hat{\x}_0 = (\x_{k}^{(0)}-\sqrt{1-\bar{\alpha}_{\hat{t}_{k}}}\mathbf{v}^{(0)})/\sqrt{\bar{\alpha}_{\hat{t}_{k}}}$ \# Proc $0$ 
            \STATE $\tilde{\mu}_{\operatorname{pred}}=(\sqrt{\bar{\alpha}_{\hat{t}_{k}-1}}\beta_{\hat{t}_{k}}\hat{\x}_{0}+\sqrt{\alpha_{\hat{t}_{k}}}(1-\bar{\alpha}_{{\hat{t}_{k}}-1})\x_{k}^{(0)})/\sqrt{1-\bar{\alpha}_{\hat{t}_{k}}}$ \# Proc $0$
            \STATE $\hat{\x}_{\operatorname{pred}}=\tilde{\mu}_{\operatorname{pred}}+\rho\cdot\sqrt{\tilde{\beta}_{\hat{t}_{k}}}\epsilon,~\epsilon\sim\cN(\zeros_n,I_n)$ \# Proc $0$
            \STATE $\x^{(1)}_{k} = \gamma\cdot\x^{(1)}_k + (1-\gamma)\cdot\hat{\x}_{\operatorname{pred}}$ \# Integration of information. Proc $0$ sends $\hat{\x}_{\operatorname{pred}}$ to Proc $1$. Proc $1$ 
            \STATE $\x^{(0)}_{k} = \x^{(1)}_{k}$ \# Proc $1$ sends $\hat{\x}^{(1)}_k$ to Proc $0$. Proc $0$
        \ENDIF
        \STATE Proc $0$ and Proc $1$ fix the same random seed.
        \STATE {\bfseries PARALLEL} denoising for each Proc $j\in\{0,1\}$
        %
        \begin{ALC@g}
        \STATE $t^{(j)}_{k} = t_k + 1 - j$ 
        \STATE $\mathbf{v}^{(j)}=\epsilon_{\theta}(\x^{(j)}_k,t^{(j)}_{k})$
        \STATE $\x^{(j)}_{k-1} = \operatorname{denoise}(t^{(j)}_{k},\x^{(j)}_k,\mathbf{v}^{(j)})$ \# Using previously defined seed 
        \end{ALC@g}
        \STATE {\bfseries end PARALLEL}
        %
        %
    \ENDFOR
    \STATE {\bfseries Return:} $\x_{0}^{(0)}$
%
\end{algorithmic}
\end{algorithm}

\begin{figure}[t!]
\centering
  \includegraphics[width=0.78\linewidth]{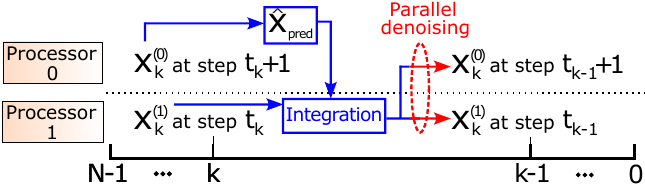}
  \caption{
  \textbf{Sample Enhancement Using Two Processors (SE2P).} Considering Algorithm~\ref{alg:main_SE2P}, we represent the three main computations: (i) the predictor value ($\hat{\x}_{\operatorname{pred}}$ corresponding to line~8), (ii) the integration of information (``integration'' block corresponding to line~9), and (ii) the parallel sampling (lines~13 to 17).
  }
  \label{fig:Se2P-algorithm}
\end{figure}

\section{Qualitative Study}
\label{sec:qualitative}
We discuss the appearance and quality of sampled images from SE2P.
%
We test four different models: DDPM~\citep{Ho-2020-ddpm}, Latend Diffusion (LD)~\citep{LD_Rombach_2022_CVPR}, Diffusion Transformers (DiT)~\citep{dit-peebles-20230-DiT}, and Stable Diffusion (SD)~\citep{LD_Rombach_2022_CVPR}. Thus, we test our method across models that are unconditional and conditional (by class or text), U-net and transformed based, on pixel and latent spaces.\footnote{We mostly use the term ``latent'' to refer to the intermediate images generated by the denoising process, whether it is in the \emph{pixel} or \emph{latent} space.} 
The \emph{baseline method} is simply running the jump sampling on one processor. 
For our analysis, 
due to the stochastic scheduler, we fix the random seed 
whenever
we present a comparison of 
sampled images.
All experiments have the same fixed value for the mixing parameter. 
Additional figures are in Appendix~\ref{app:exp}. 
%

\subsection{DDPM Model}

%
The DDPM model is a pixel-space unconditional model based on U-Nets. We consider one pretrained on the CelebA-HQ dataset~\citep{karras-2018-celebahq}. 
%
%

\begin{figure*}[t!]
    \centering
\includegraphics[width=0.075\columnwidth]{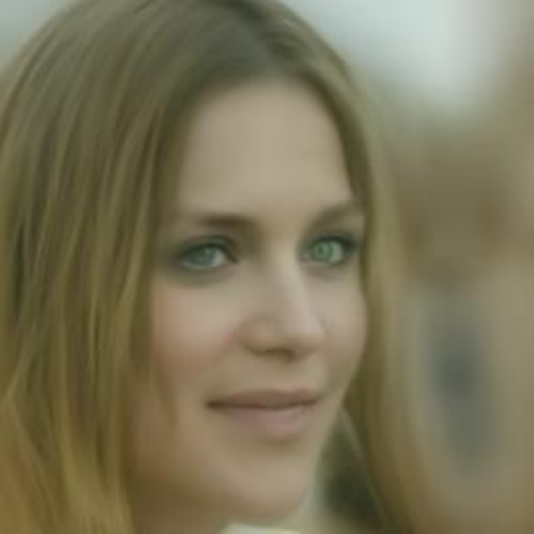}
\includegraphics[width=0.075\columnwidth]{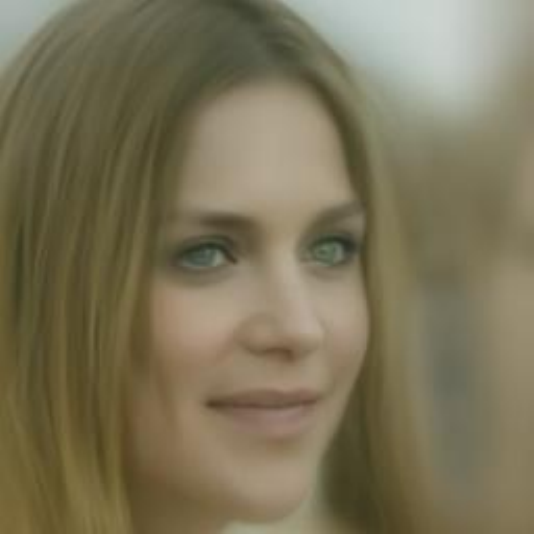}
\includegraphics[width=0.075\columnwidth]{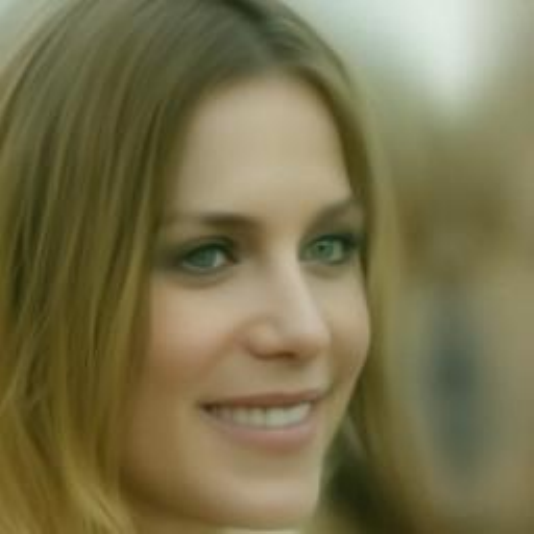}
\includegraphics[width=0.075\columnwidth]{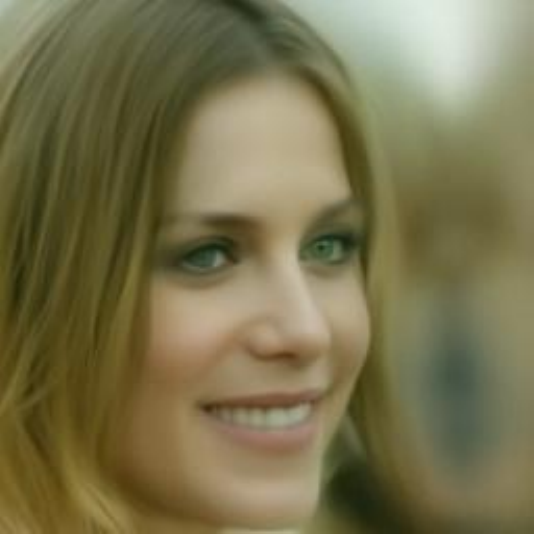}
\hspace{1ex}
\includegraphics[width=0.075\columnwidth]{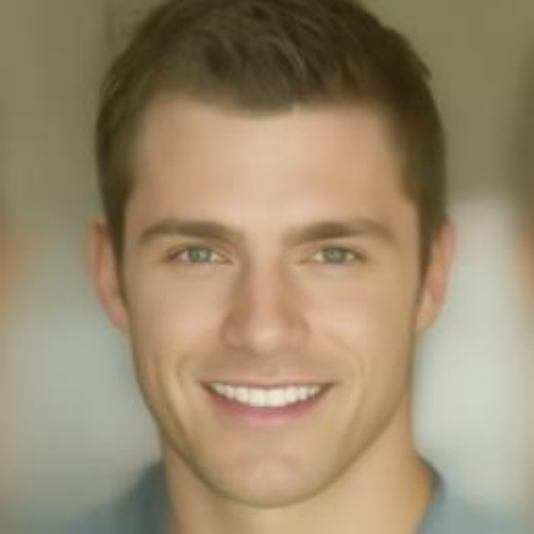}
\includegraphics[width=0.075\columnwidth]{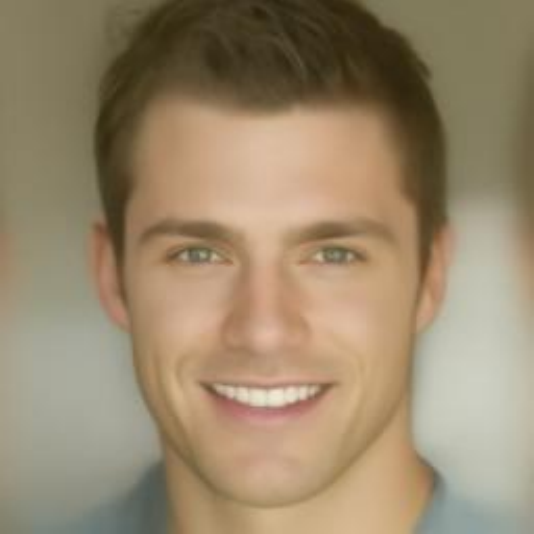}
\includegraphics[width=0.075\columnwidth]{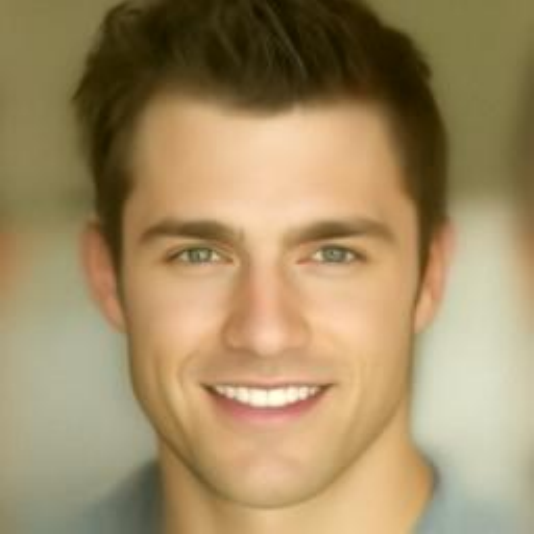}
\includegraphics[width=0.075\columnwidth]{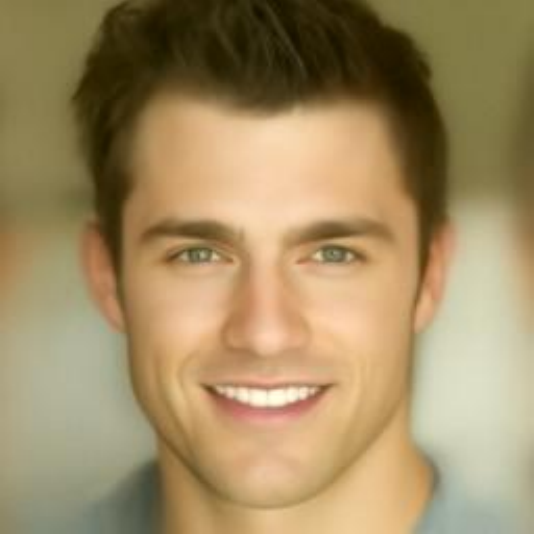}
\hspace{1ex}
\includegraphics[width=0.075\columnwidth]{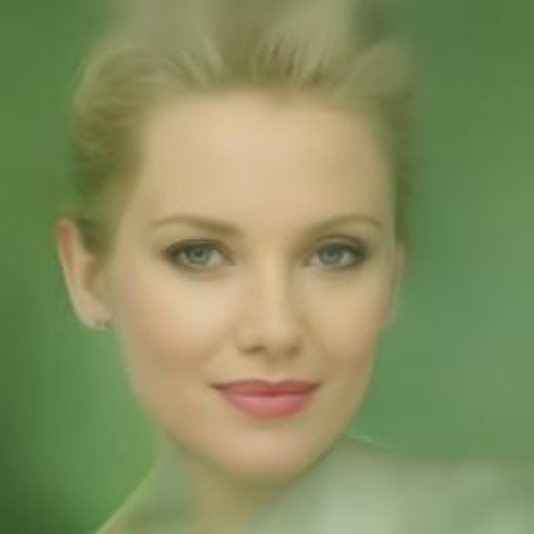}
\includegraphics[width=0.075\columnwidth]{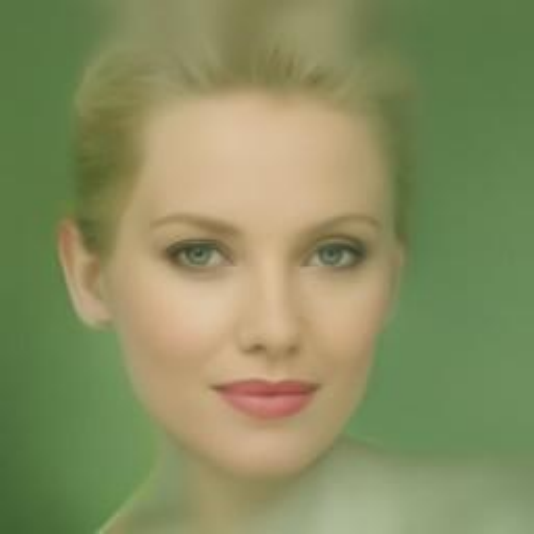}
\includegraphics[width=0.075\columnwidth]{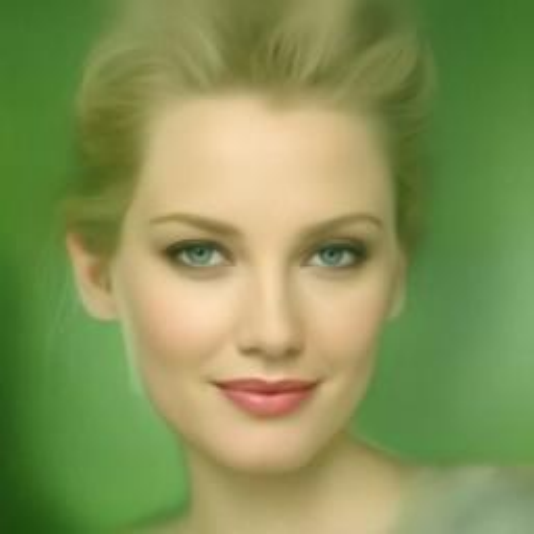}
\includegraphics[width=0.075\columnwidth]{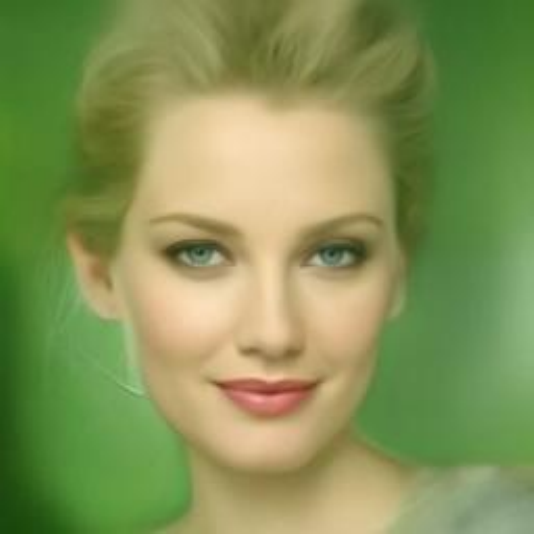}
\\
\vspace{-8pt}
\rule{0.9\textwidth}{0.4pt}
\\
\vspace{3pt}
\includegraphics[width=0.075\columnwidth]{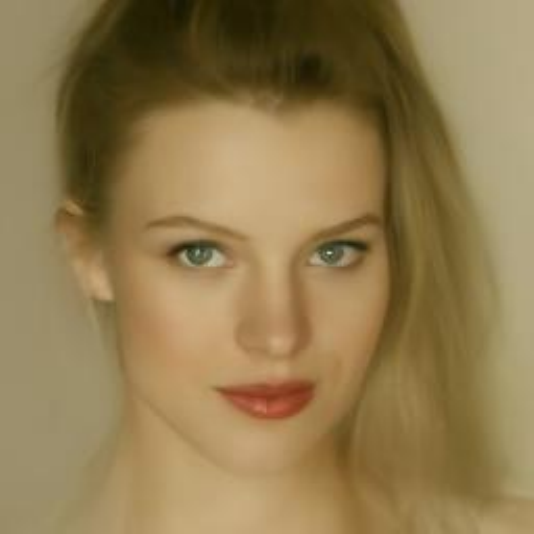}
\includegraphics[width=0.075\columnwidth]{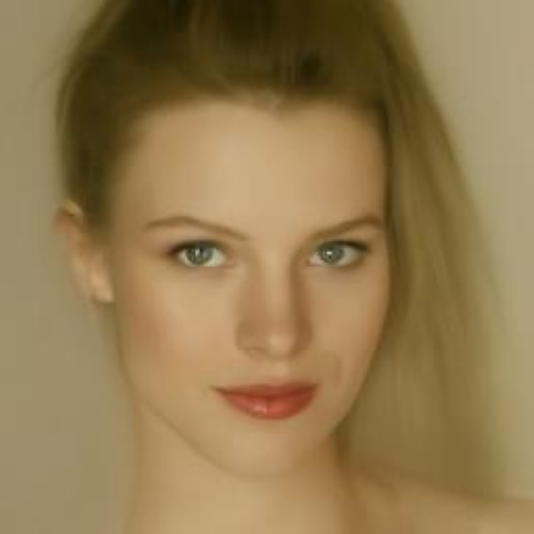}
\includegraphics[width=0.075\columnwidth]{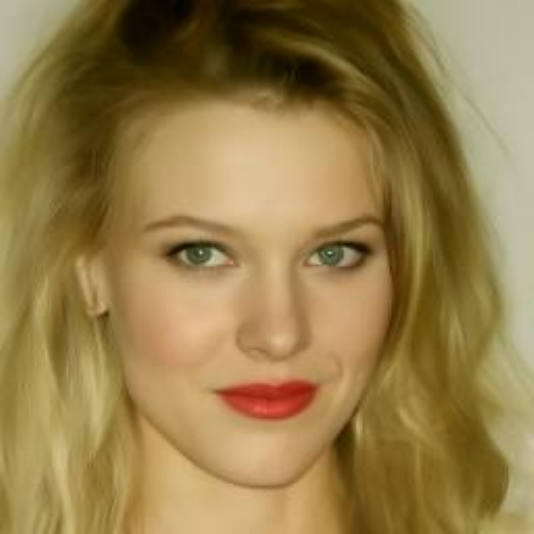}
\includegraphics[width=0.075\columnwidth]{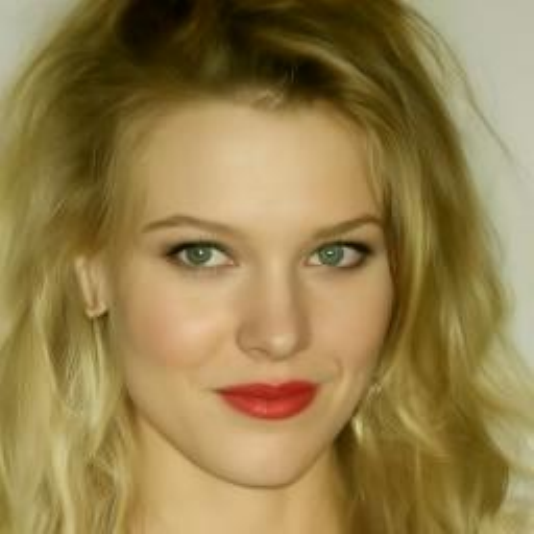}
\hspace{1ex}
\includegraphics[width=0.075\columnwidth]{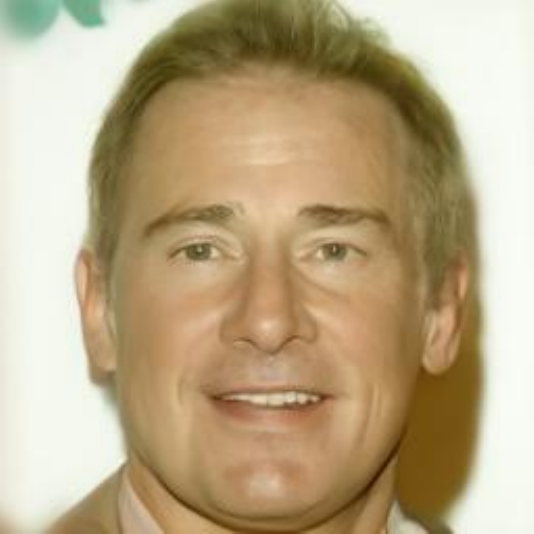}
\includegraphics[width=0.075\columnwidth]{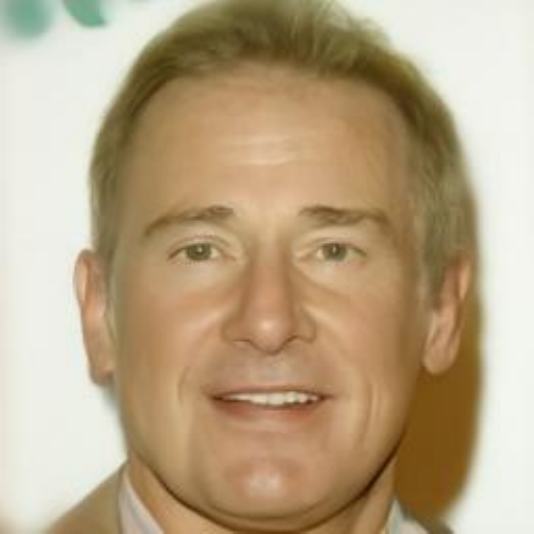}
\includegraphics[width=0.075\columnwidth]{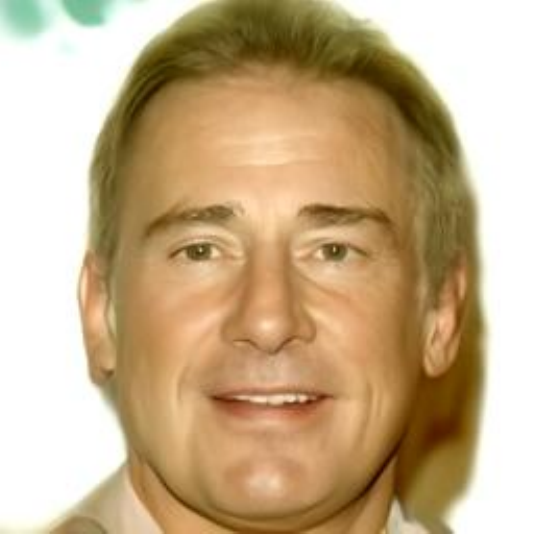}
\includegraphics[width=0.075\columnwidth]{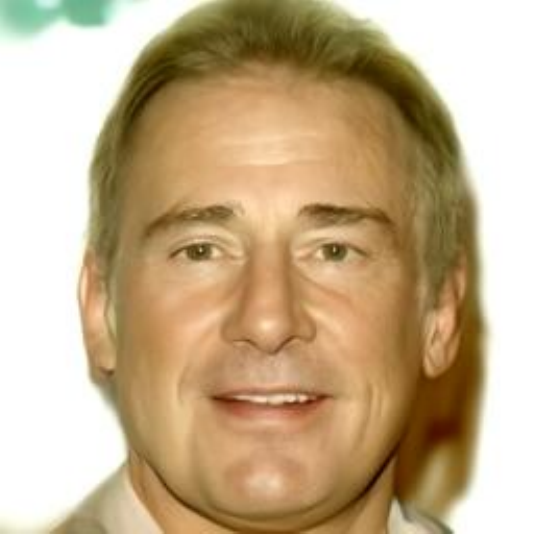}
\hspace{1ex}
\includegraphics[width=0.075\columnwidth]{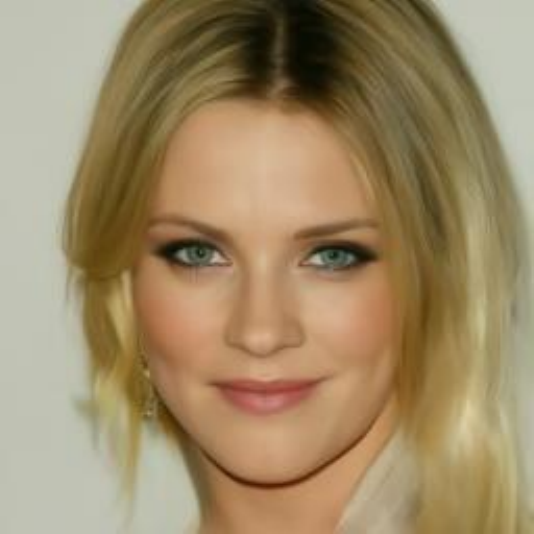}
\includegraphics[width=0.075\columnwidth]{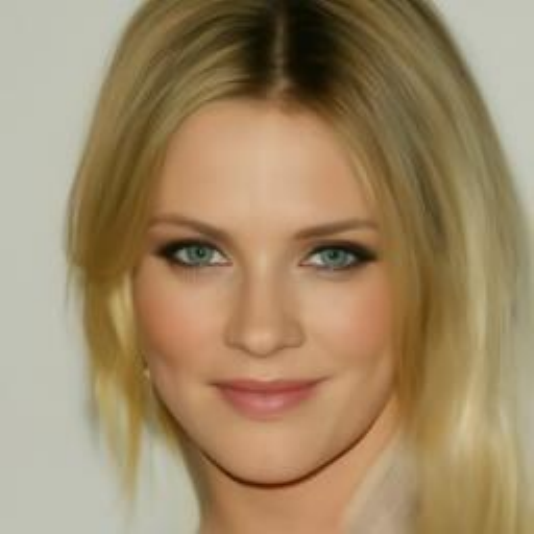}
\includegraphics[width=0.075\columnwidth]{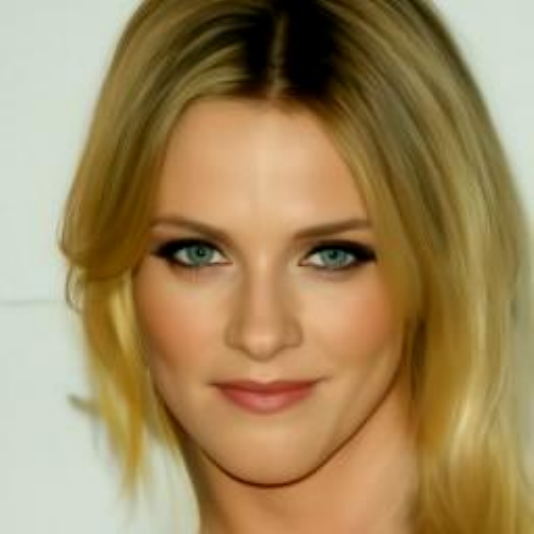}
\includegraphics[width=0.075\columnwidth]{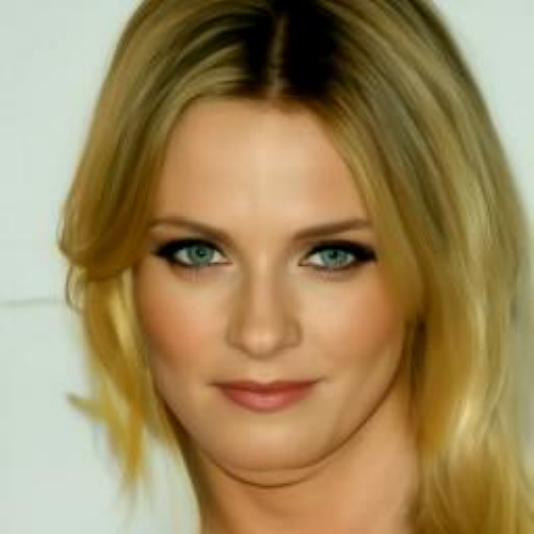}
\\
\vspace{-8pt}
\rule{0.9\textwidth}{0.4pt}
\\
\vspace{3pt}
\includegraphics[width=0.075\columnwidth]{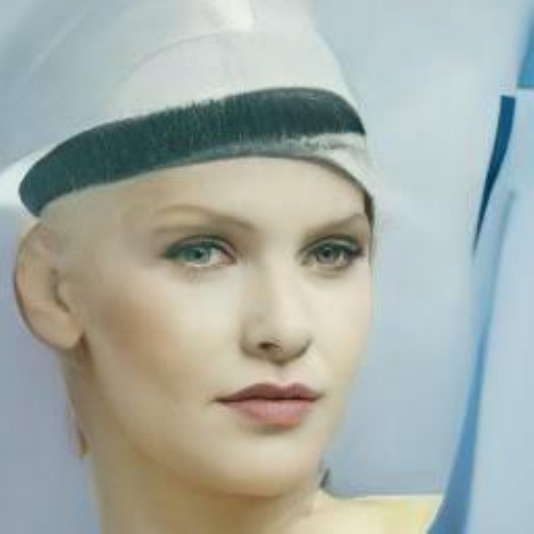}
\includegraphics[width=0.075\columnwidth]{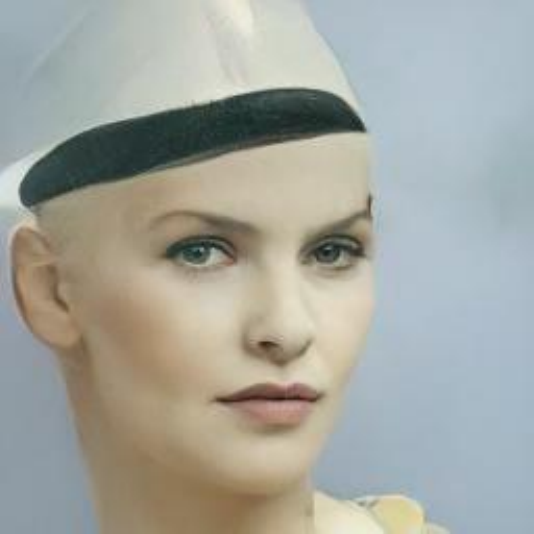}
\includegraphics[width=0.075\columnwidth]{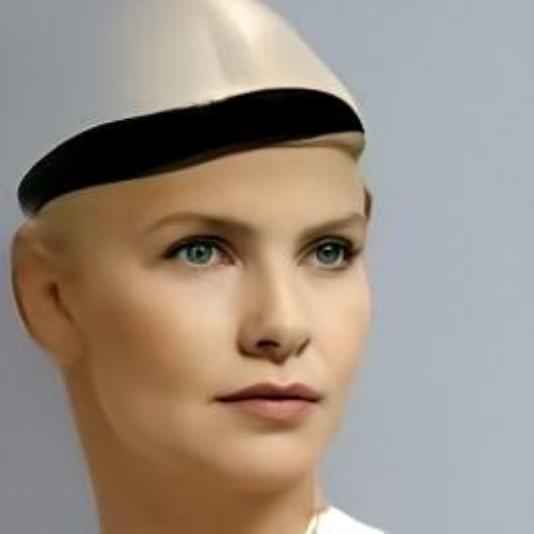}
\includegraphics[width=0.075\columnwidth]{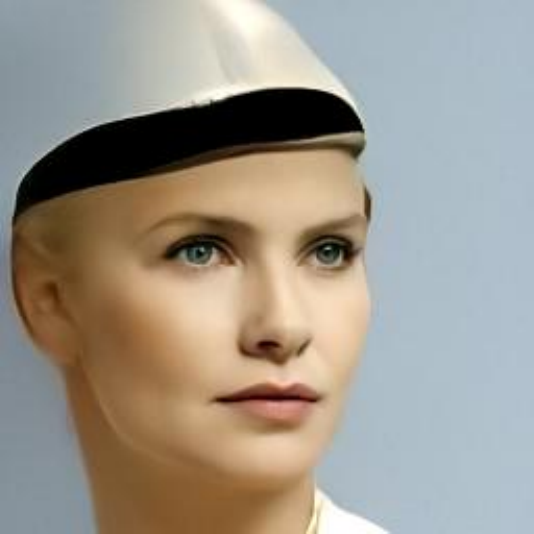}
\hspace{1ex}
\includegraphics[width=0.075\columnwidth]{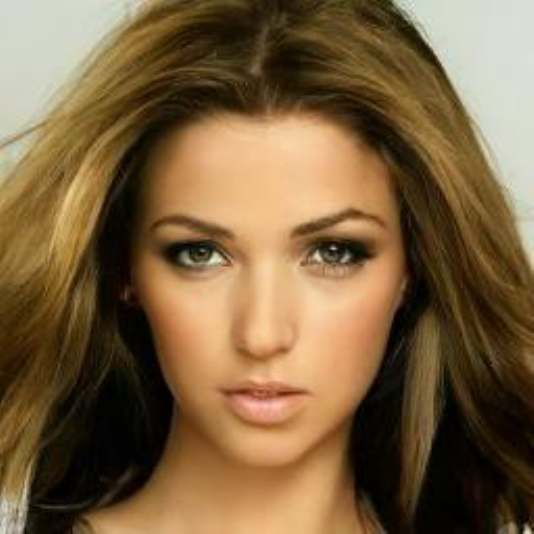}
\includegraphics[width=0.075\columnwidth]{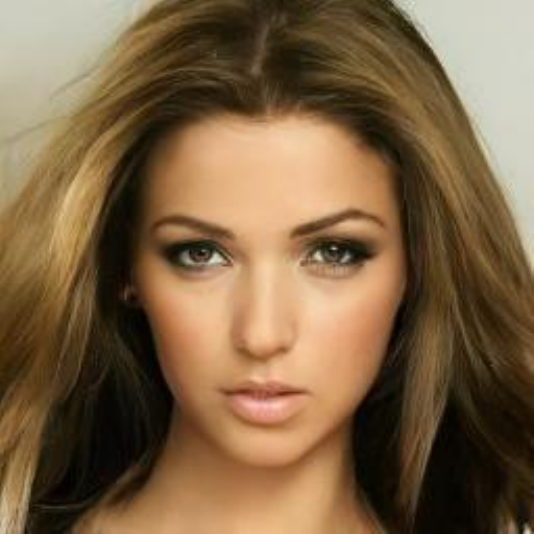}
\includegraphics[width=0.075\columnwidth]{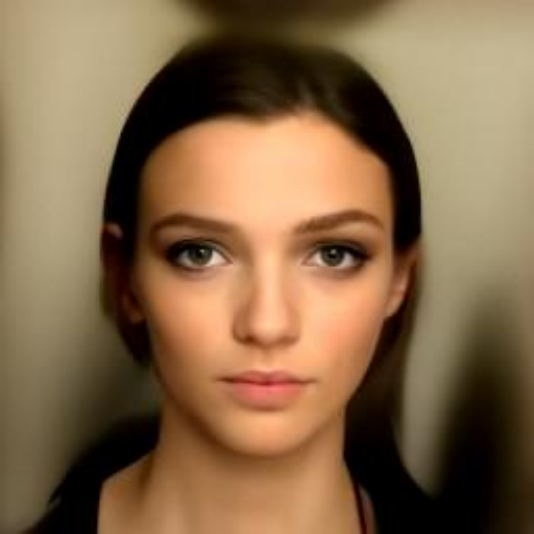}
\includegraphics[width=0.075\columnwidth]{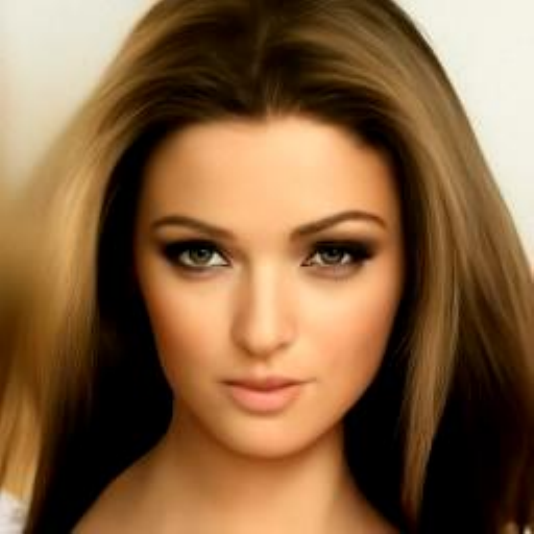}
\hspace{1ex}
\includegraphics[width=0.075\columnwidth]{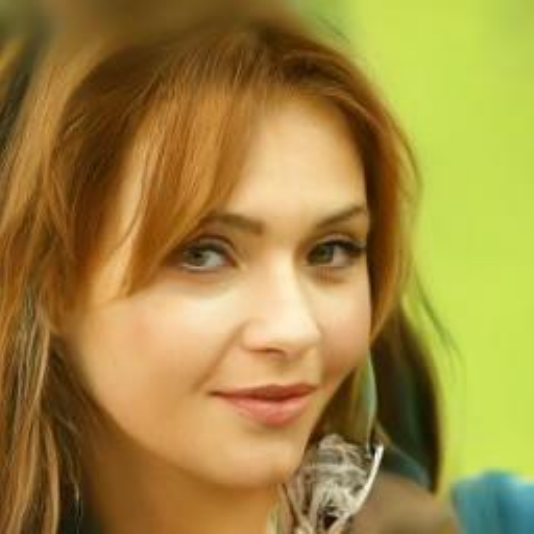}
\includegraphics[width=0.075\columnwidth]{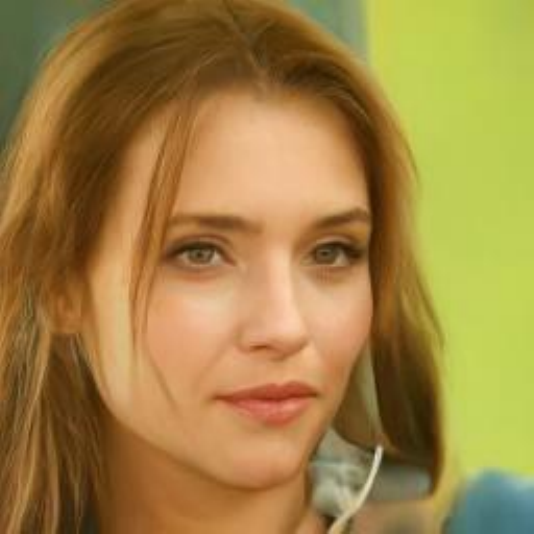}
\includegraphics[width=0.075\columnwidth]{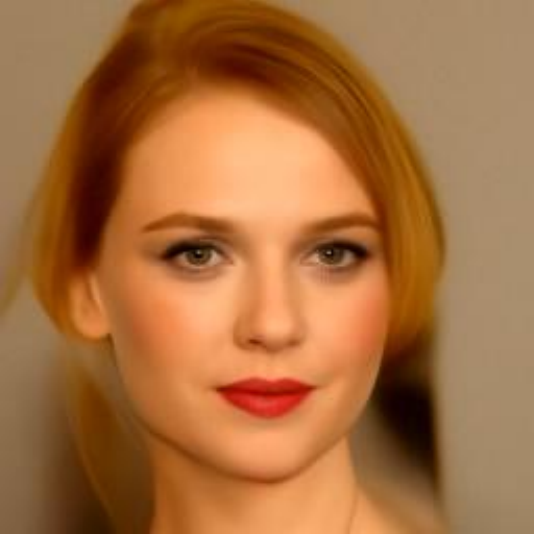}
\includegraphics[width=0.075\columnwidth]{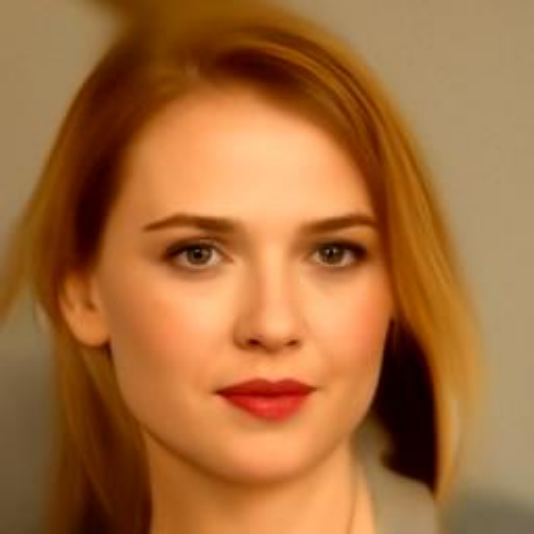}
%
%
%
\caption{\textbf{DDPM with $10$ (Above), $20$ (Middle), \& $100$ (Below) Steps.}  
For each group of four images, from left to right: baseline, ablation, and generated images with noise scalings $1.35$ and $1.55$ for $10$ and $20$ steps, or $0.25$ and $0.50$ for $100$ steps. 
%
%
}
    \label{fig:ddpm-o}
    \end{figure*}

We start
our analysis 
with $10$ denoising steps ($1\%$ of the typical $1000$ steps). Sampled images are shown in Fig.~\ref{fig:ddpm-o} and Fig.~\ref{fig:ddpm-10} from Appendix~\ref{app:exp}. 
Our method generally leads to images with more contrast, brightness and vivid colors, which can improve overall visual quality. Baseline images 
in general present low contrast, often look ``wash-out'', and may contain blurry features; conversely, SE2P 
can improve 
image quality by reversing such undesirable attributes. 
A shortcoming with our method: 
if the baseline image has long exposure or brightness,
SE2P may not be able to reverse it; 
see Fig.~\ref{fig:shortcoming} in Appendix~\ref{app:exp}. Interestingly, even in these negative cases, SE2P can still improve the contrast among image features. 
In the case of $20$ steps, similar observations hold with the obvious difference that baseline images have better quality than before. 
Samples are shown in Fig.~\ref{fig:ddpm-o} and Fig.~\ref{fig:ddpm-20} from Appendix~\ref{app:exp}. 
In the case of $100$ steps, 
our method still manages to often improve the contrast, brightness, and even exposure; however, 
%
the improvement appears weaker 
due to the improved baseline.
Samples are shown in Fig.~\ref{fig:ddpm-o} and Fig.~\ref{fig:DDPM-100-semantic} from Appendix~\ref{app:exp}.
%
Indeed, 
while overexposure can still occur (Fig.~\ref{fig:shortcoming}), S2EP can also reverse it from the baseline; see Fig.~\ref{fig:DDPM-100-contrast} in Appendix~\ref{app:exp}.

We also find that SE2P is able to change the \emph{semantic content} of the sampled image more often and more strongly as the number of steps increases. 
%
%
%
%
It is possible that these strong semantic changes by S2EP is what produce images with less exposure and brightness than the baseline; e.g., for $100$ steps in Fig.~\ref{fig:DDPM-100-contrast}.

%

Finally, 
compared to the case of $10$ and $20$ steps, it was necessary to considerably reduce the variance scaling for $100$ steps in order to avoid image quality degradation. 
For a fixed value of the mixing parameter, more denoising steps require a lower variance scaling. 
%
%

%
%
%
%


\subsection{Latent Diffusion (LD) Model}

The LD model is a latent-space unconditional model based on U-Nets. 
For comparison purposes, we also consider 
%
%
a 
pretrained one on the CelebA-HQ dataset.
%
Remarkably, our 
observations about the positive effects of S2EP on the sampled images are similar to the DDPM case, showing 
our method's effectiveness 
on both pixel and latent space models.

\begin{figure*}[h!t]
    \centering
\includegraphics[width=0.075\columnwidth]{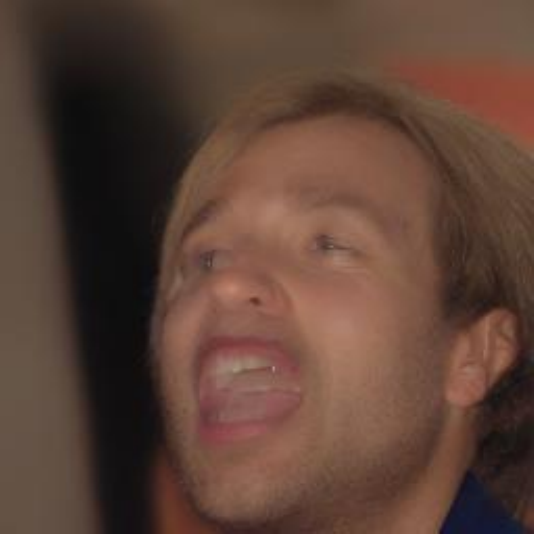}
\includegraphics[width=0.075\columnwidth]{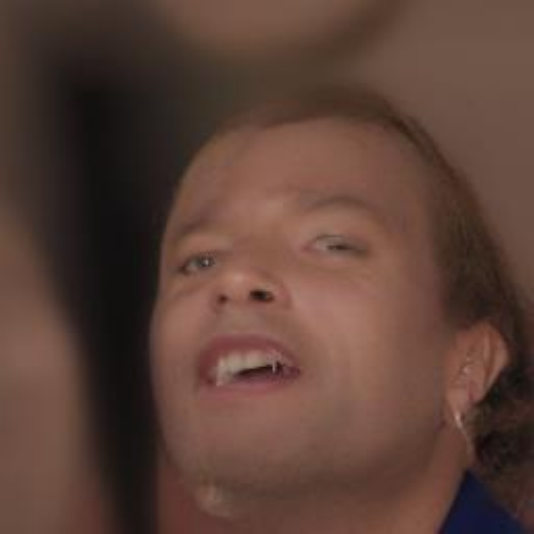}
\includegraphics[width=0.075\columnwidth]{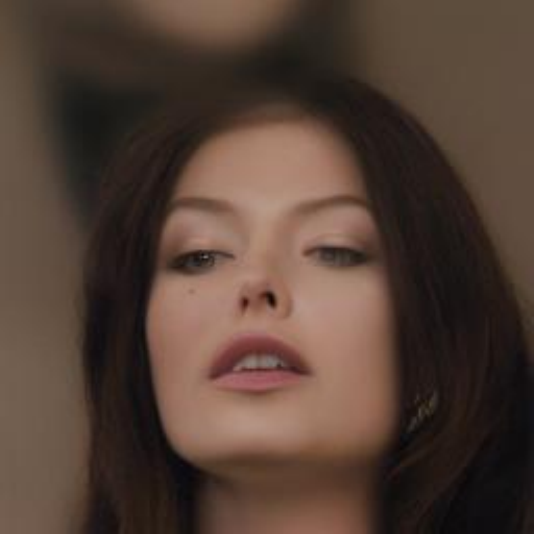}
\includegraphics[width=0.075\columnwidth]{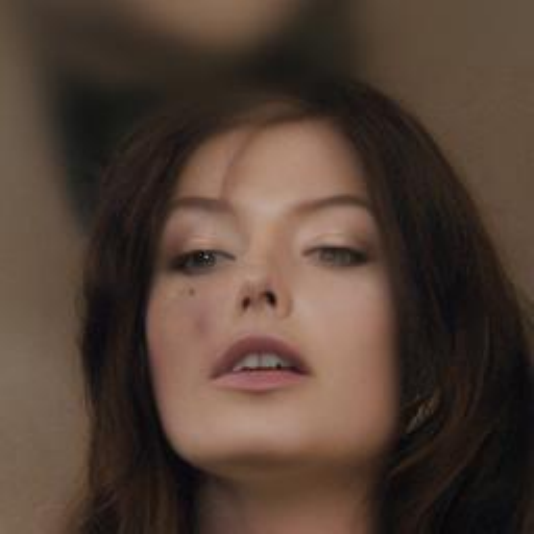}
\hspace{1ex}
\includegraphics[width=0.075\columnwidth]{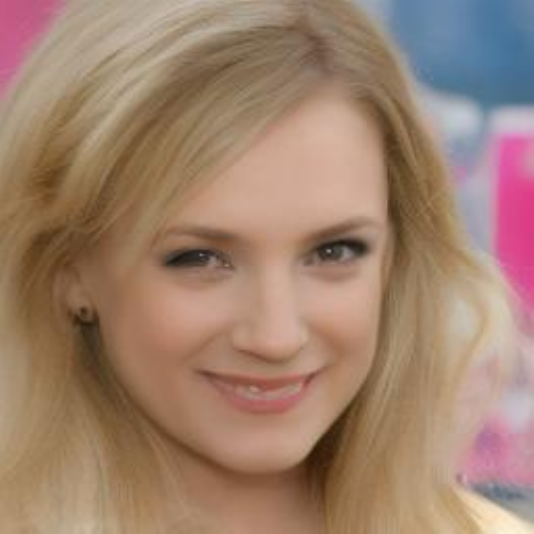}
\includegraphics[width=0.075\columnwidth]{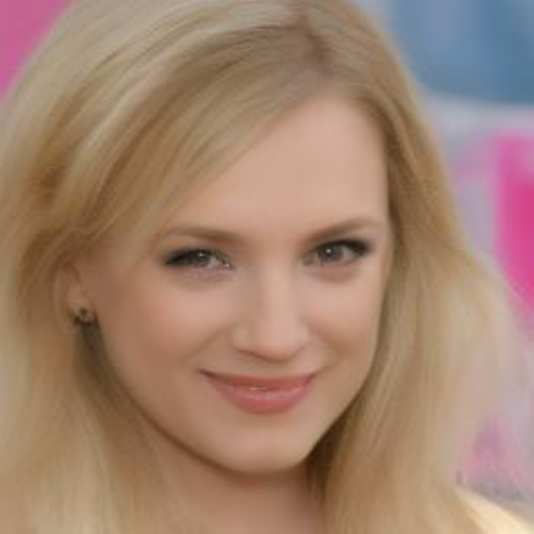}
\includegraphics[width=0.075\columnwidth]{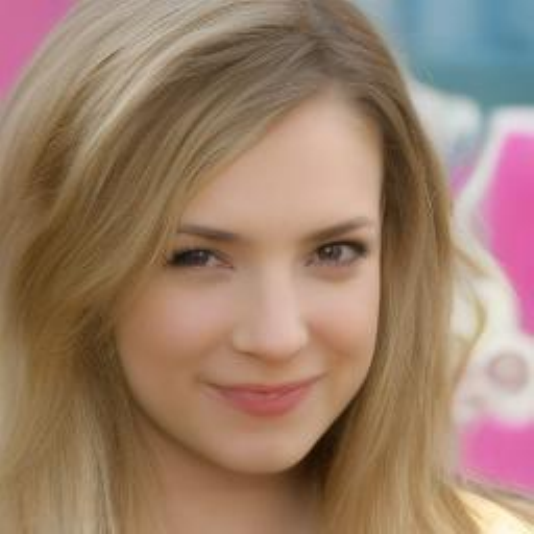}
\includegraphics[width=0.075\columnwidth]{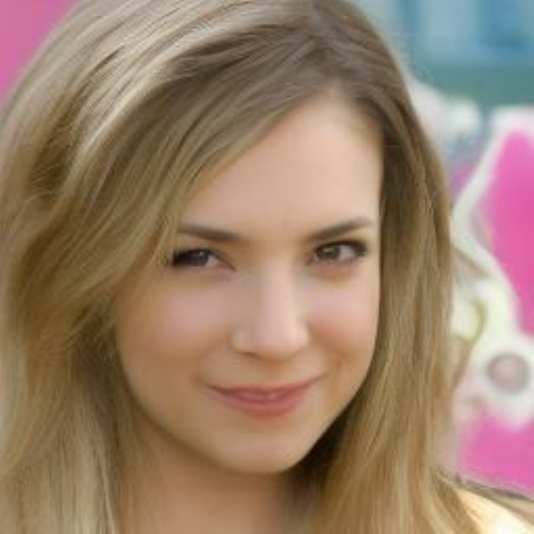}
\hspace{1ex}
\includegraphics[width=0.075\columnwidth]{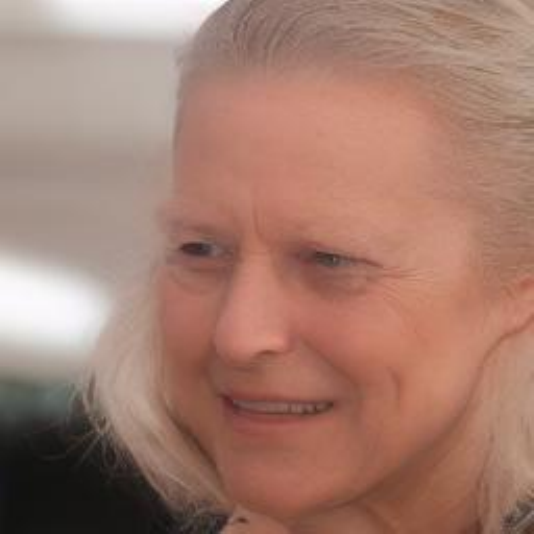}
\includegraphics[width=0.075\columnwidth]{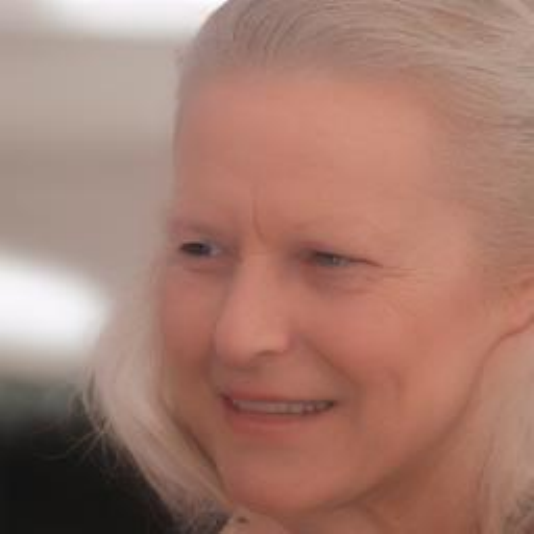}
\includegraphics[width=0.075\columnwidth]{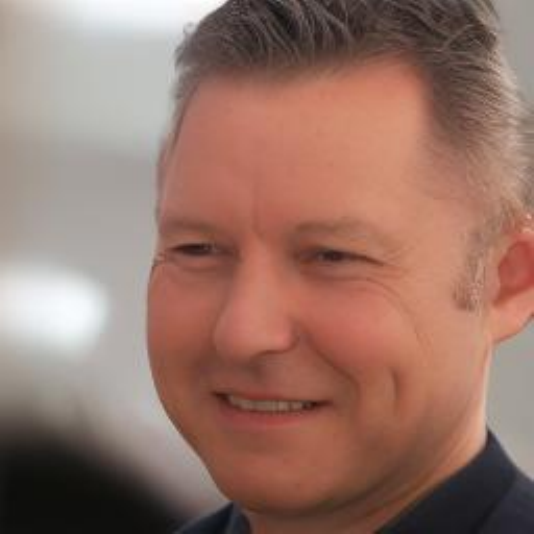}
\includegraphics[width=0.075\columnwidth]{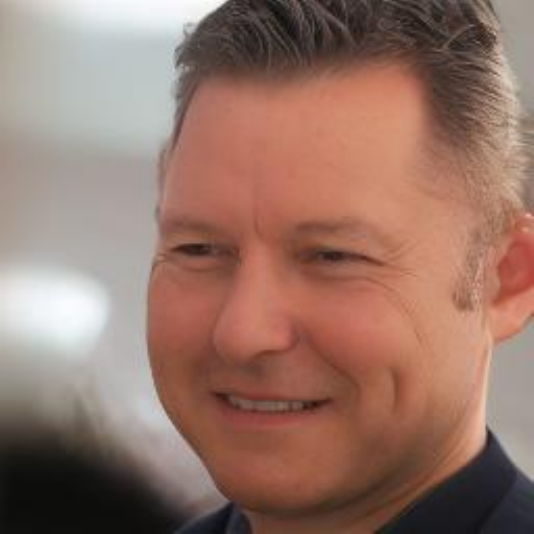}
\\
\vspace{-8pt}
\rule{0.9\textwidth}{0.4pt}
\\
\vspace{3pt}
\includegraphics[width=0.075\columnwidth]{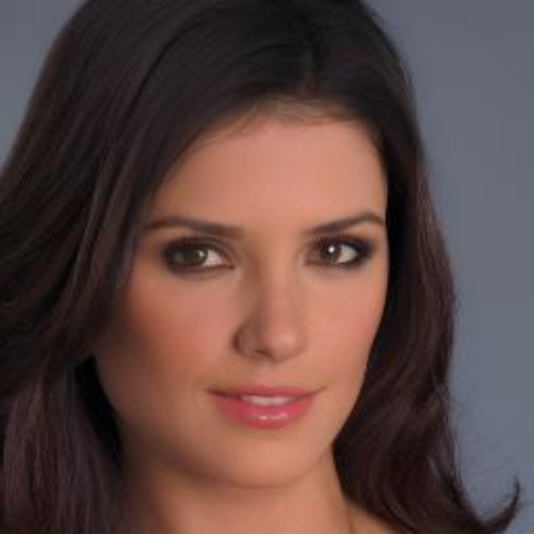}
\includegraphics[width=0.075\columnwidth]{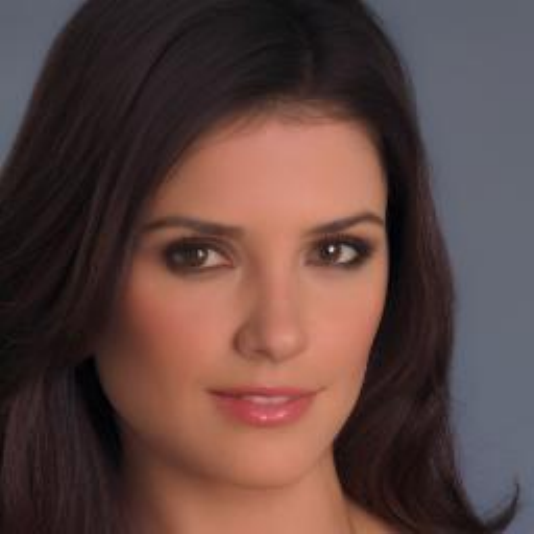}
\includegraphics[width=0.075\columnwidth]{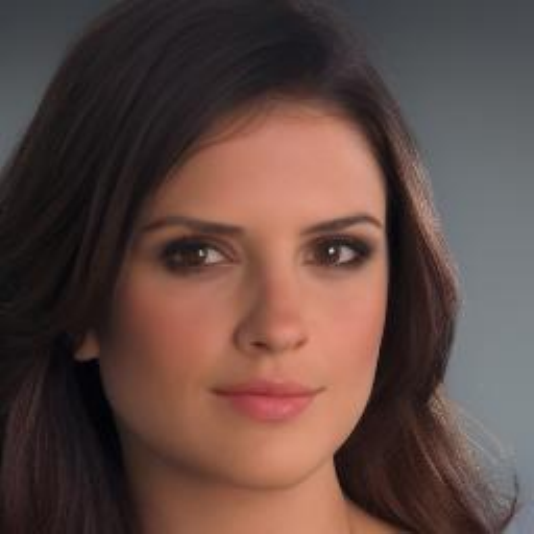}
\includegraphics[width=0.075\columnwidth]{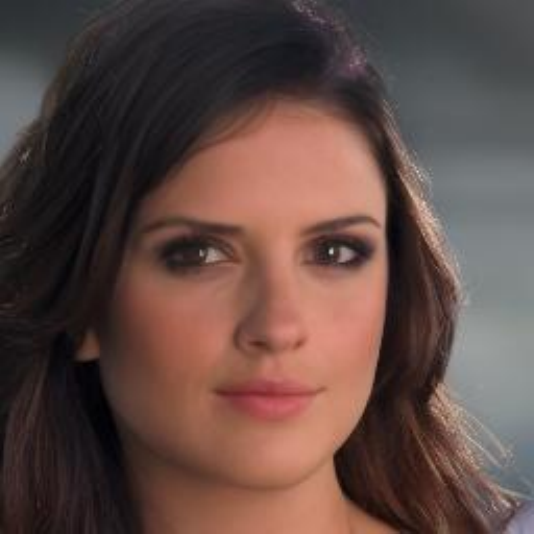}
\hspace{1ex}
\includegraphics[width=0.075\columnwidth]{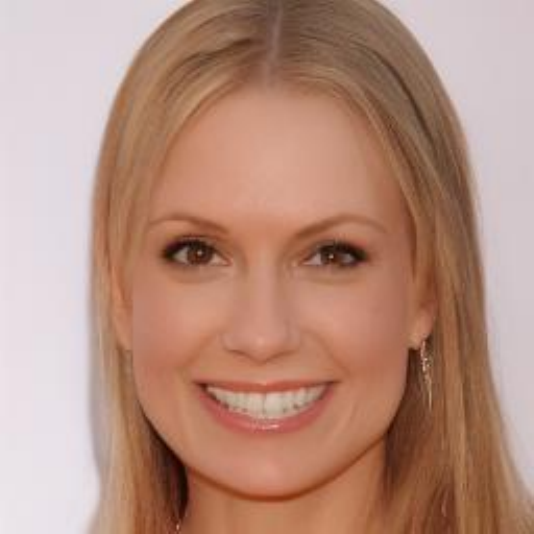}
\includegraphics[width=0.075\columnwidth]{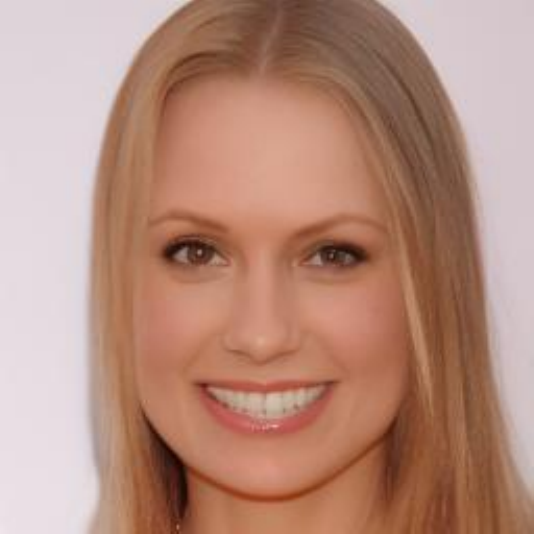}
\includegraphics[width=0.075\columnwidth]{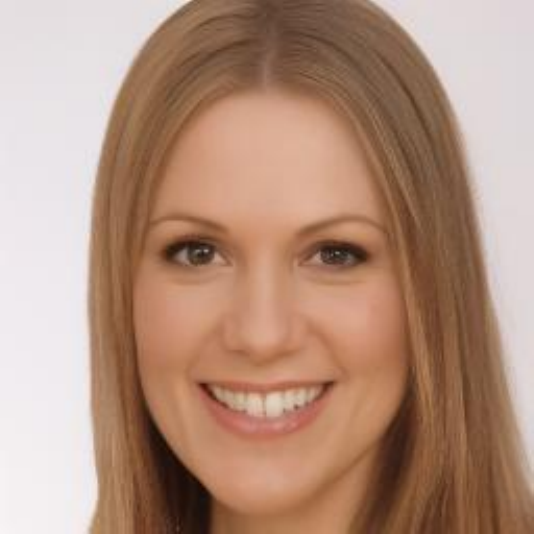}
\includegraphics[width=0.075\columnwidth]{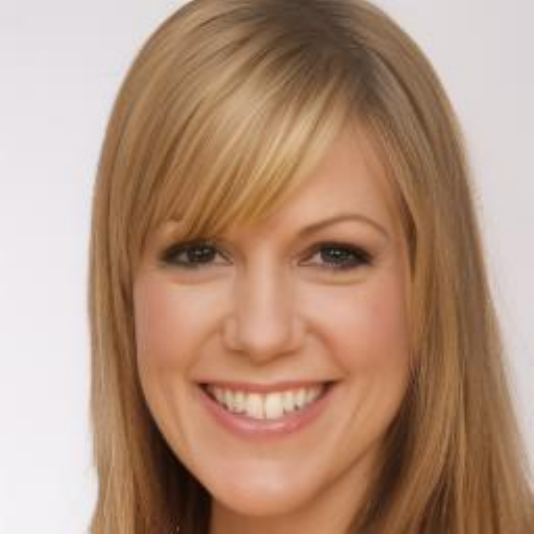}
\hspace{1ex}
\includegraphics[width=0.075\columnwidth]{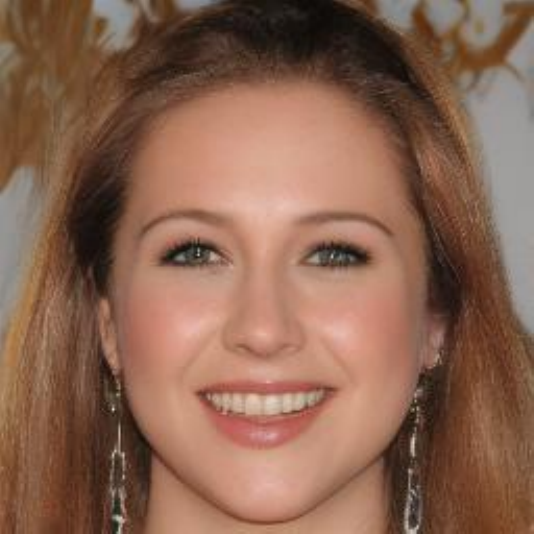}
\includegraphics[width=0.075\columnwidth]{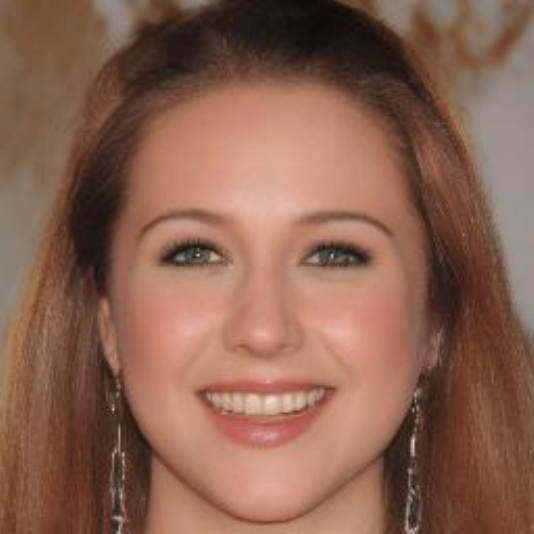}
\includegraphics[width=0.075\columnwidth]{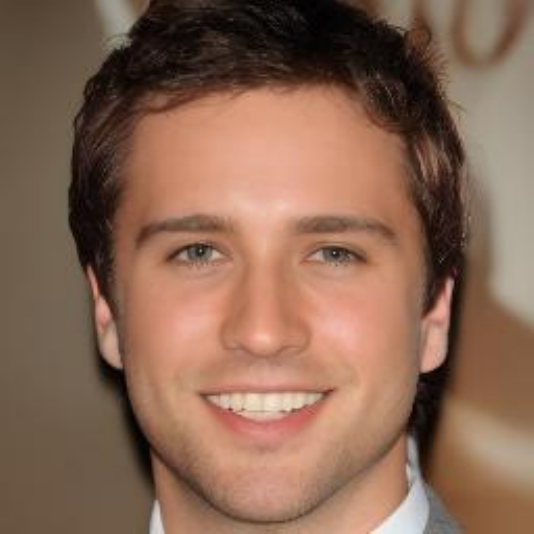}
\includegraphics[width=0.075\columnwidth]{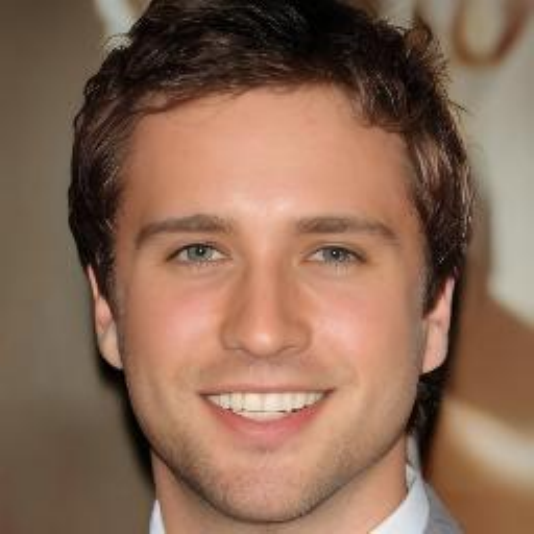}
\\
\vspace{-8pt}
\rule{0.9\textwidth}{0.4pt}
\\
\vspace{3pt}
\includegraphics[width=0.075\columnwidth]{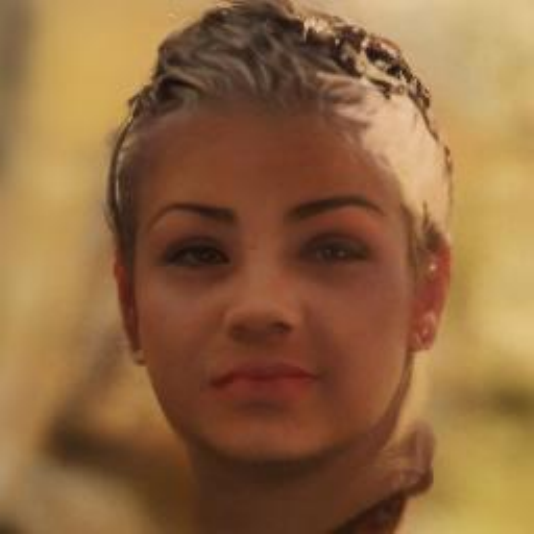}
\includegraphics[width=0.075\columnwidth]{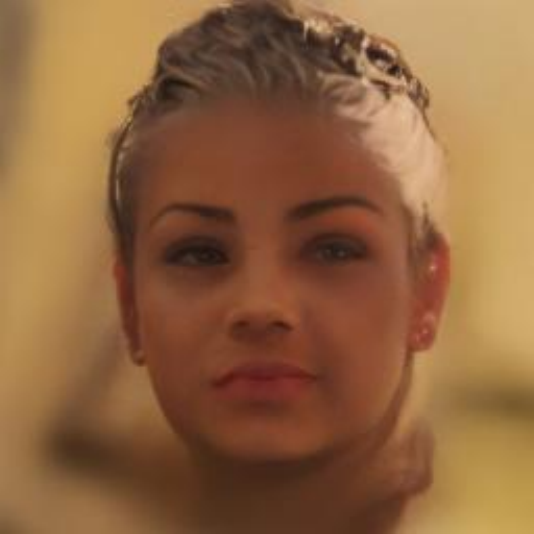}
\includegraphics[width=0.075\columnwidth]{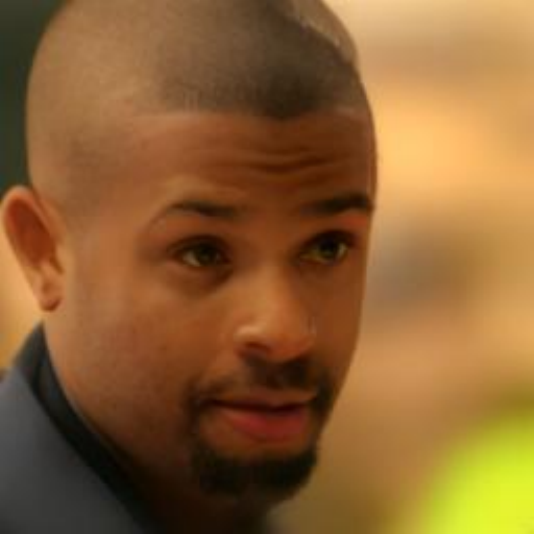}
\includegraphics[width=0.075\columnwidth]{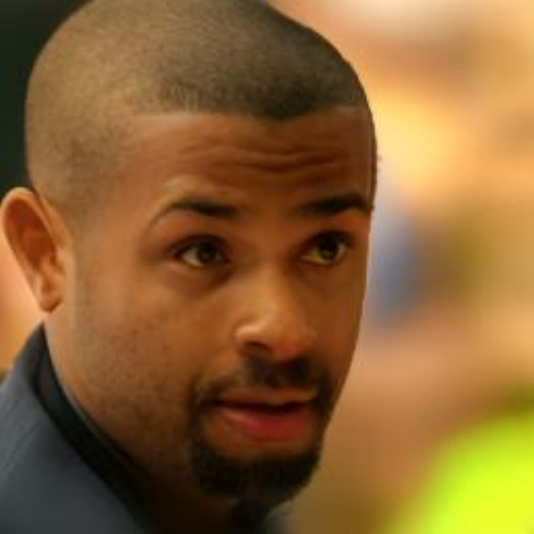}
\hspace{1ex}
\includegraphics[width=0.075\columnwidth]{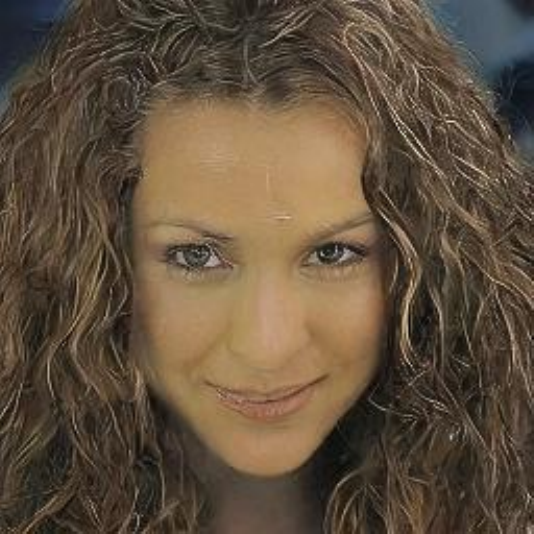}
\includegraphics[width=0.075\columnwidth]{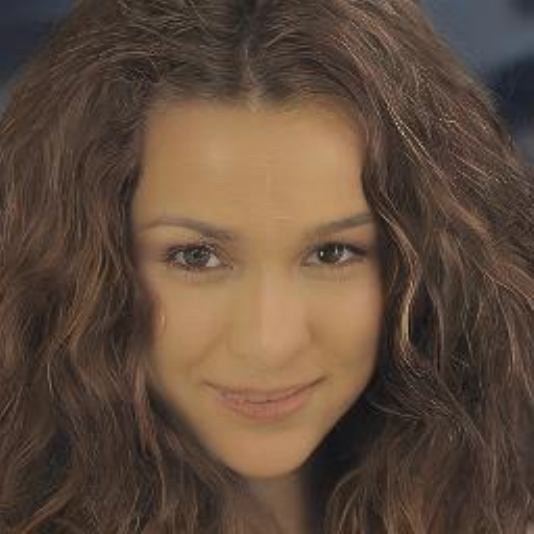}
\includegraphics[width=0.075\columnwidth]{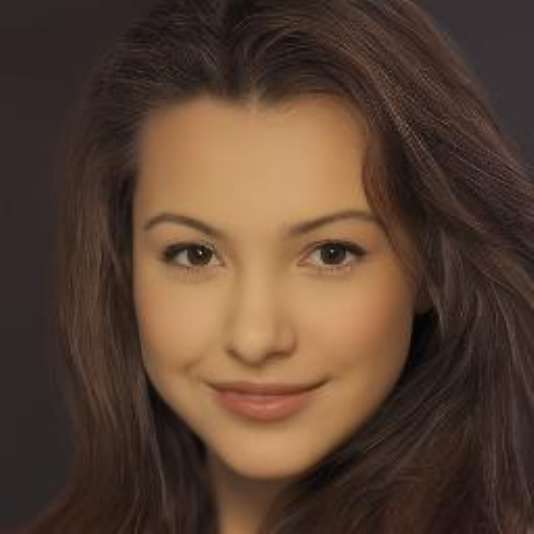}
\includegraphics[width=0.075\columnwidth]{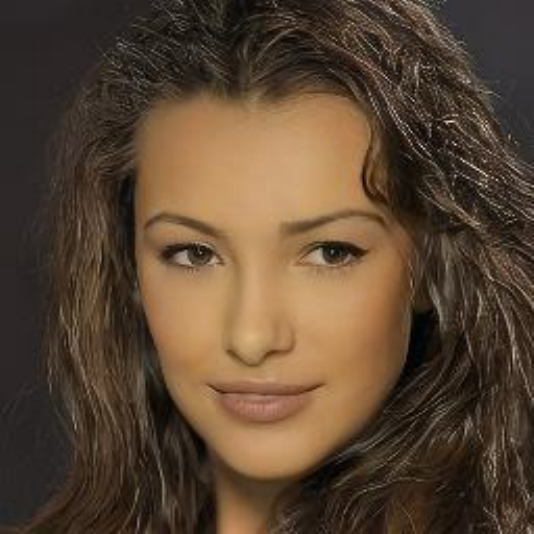}
\hspace{1ex}
\includegraphics[width=0.075\columnwidth]{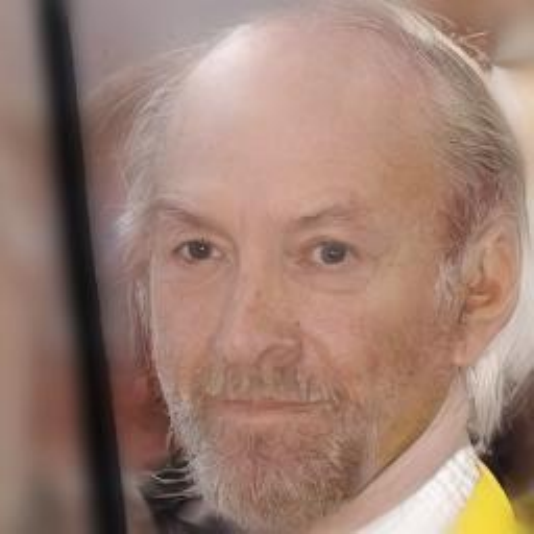}
\includegraphics[width=0.075\columnwidth]{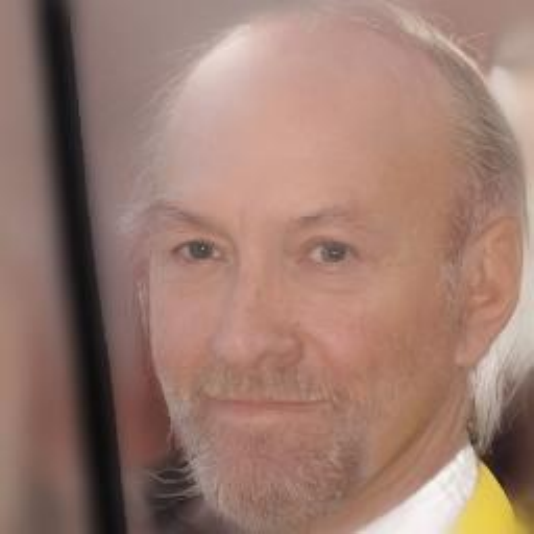}
\includegraphics[width=0.075\columnwidth]{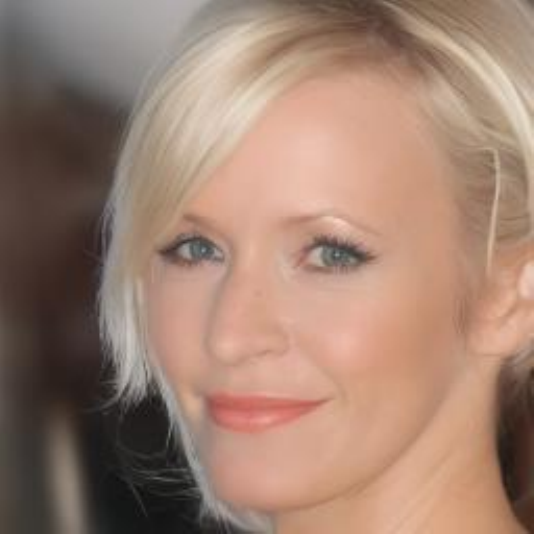}
\includegraphics[width=0.075\columnwidth]{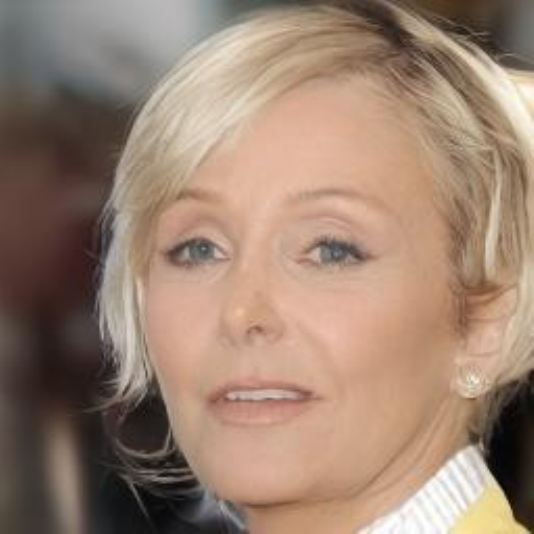}
\caption{\textbf{LD with $20$ (Above), $40$ (Middle), and $80$ (Below) Steps.} 
For each group of four images, from left to right: baseline, ablation, and generated images with noise scalings $1.35$ and $1.55$ for $20$ steps, $1.20$ and $1.35$ for $40$ steps, or $0.90$ and $1.10$ for $80$ steps.
%
%
%
}
    \label{fig:ld-o}
    \end{figure*}

We start with $20$ denoising steps.
Samples are shown in Fig.~\ref{fig:ld-o} and Fig.~\ref{fig:ld-20} from Appendix~\ref{app:exp} . 
%
%
Like DDPM, S2EP leads to an overall increase in contrast, often accompanied with more brightness and vivid colors, which can improve image quality. 
Overexposure can still occur, though at a seemingly lower rate than DDPM; see Fig~\ref{fig:shortcoming}.
We also observe S2EP is able to sample images that do not drastically increase exposure, despite an overexposed baseline.
For $40$ steps, we lower the variance scaling
to avoid 
image quality degradation (see Section~\ref{subsec:var-higher}).
Sampled images are in Fig.~\ref{fig:ld-o} and Fig.~\ref{fig:ld-40} from Appendix~\ref{app:exp}. Qualitative observations are similar to before.
%
%
%
%

For a larger number of steps, we observe that SE2P 
leads to more image degradation than with DDPM across diverse values of variance scaling.
%
%
Thus, we consider $80$ steps instead of $100$ steps as in DDPM.
Sampled images are in Fig.~\ref{fig:ld-o} and Fig.~\ref{fig:ld-80} from Appendix~\ref{app:exp}. Like DDPM, LD now has more frequent and stronger 
changes on image semantics than before. 
Likewise, this can lead to images with less exposure and brightness; see Fig.~\ref{fig:LD-80-contrast} in Appendix~\ref{app:exp}. 

%
It is important to highlight that, in the case of LD, the information that SE2P integrates 
\emph{does not} map directly to the pixel space, but through a highly nonlinear decoder.
Despite this nonlinearity, 
it is remarkable that the means by which our method improves image quality---better contrast, intensity, and sharper features---are shared by bot h latent and pixel space models.

\subsection{Diffusion Models with Transformers (DiTs)}

The DiT model is a latent-space and \emph{class-conditional} model with transformers as backbone (instead of U-Nets). 
We consider one pretrained on $1000$ classes from the ImageNet dataset~\citep{krizhevsky-2012-imagenet}
%

\begin{figure*}[t]
    \centering
\includegraphics[width=0.075\columnwidth]{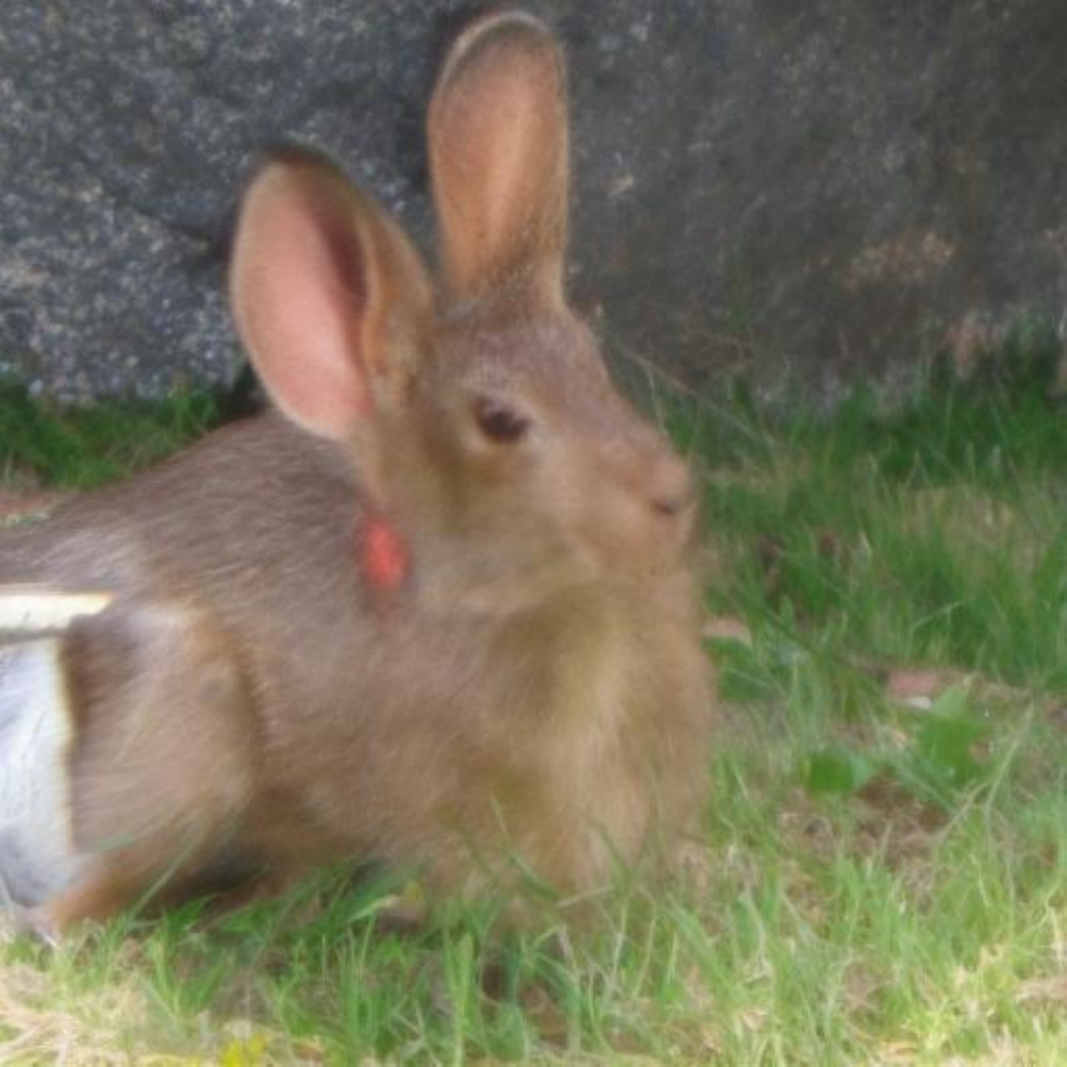}
\includegraphics[width=0.075\columnwidth]{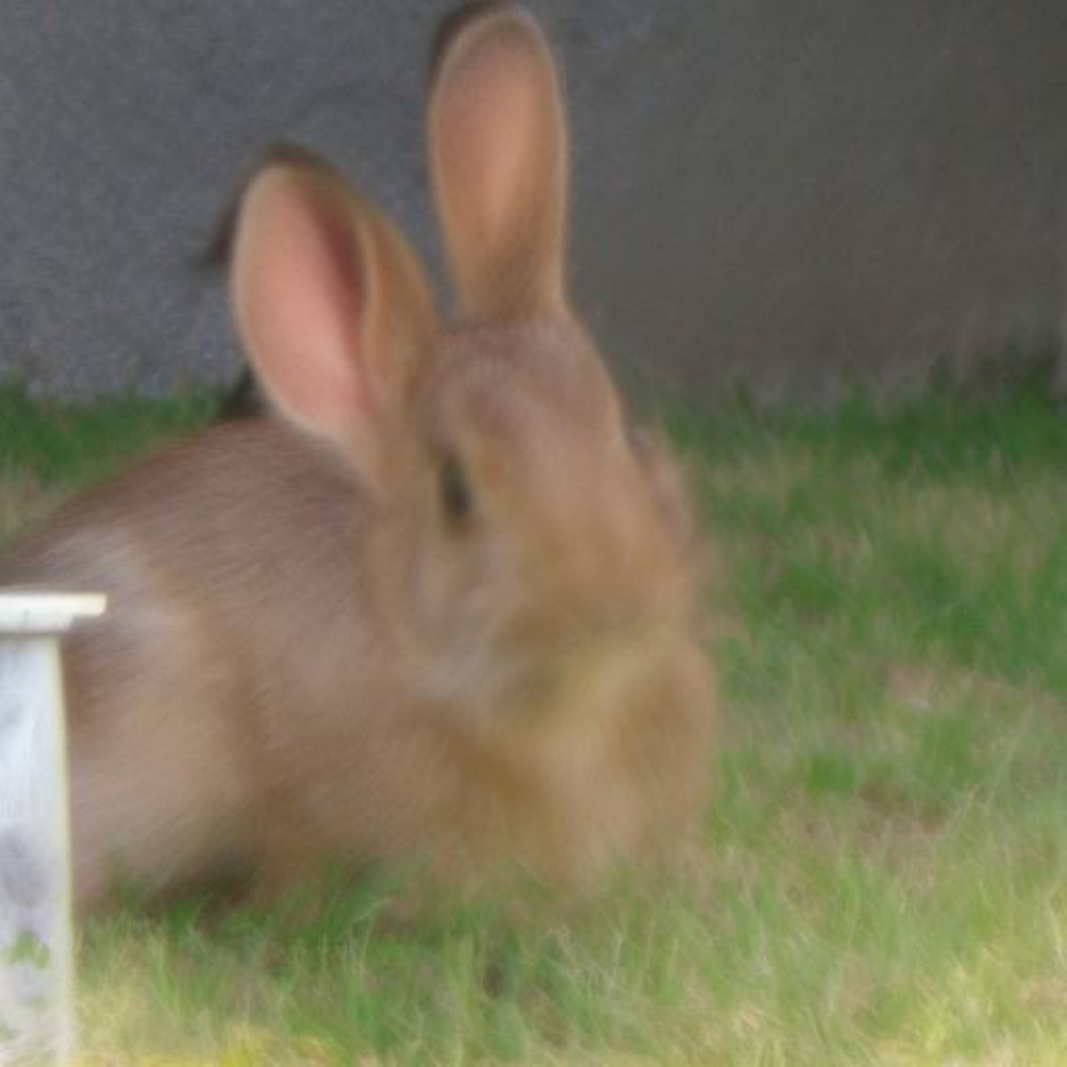}
\includegraphics[width=0.075\columnwidth]{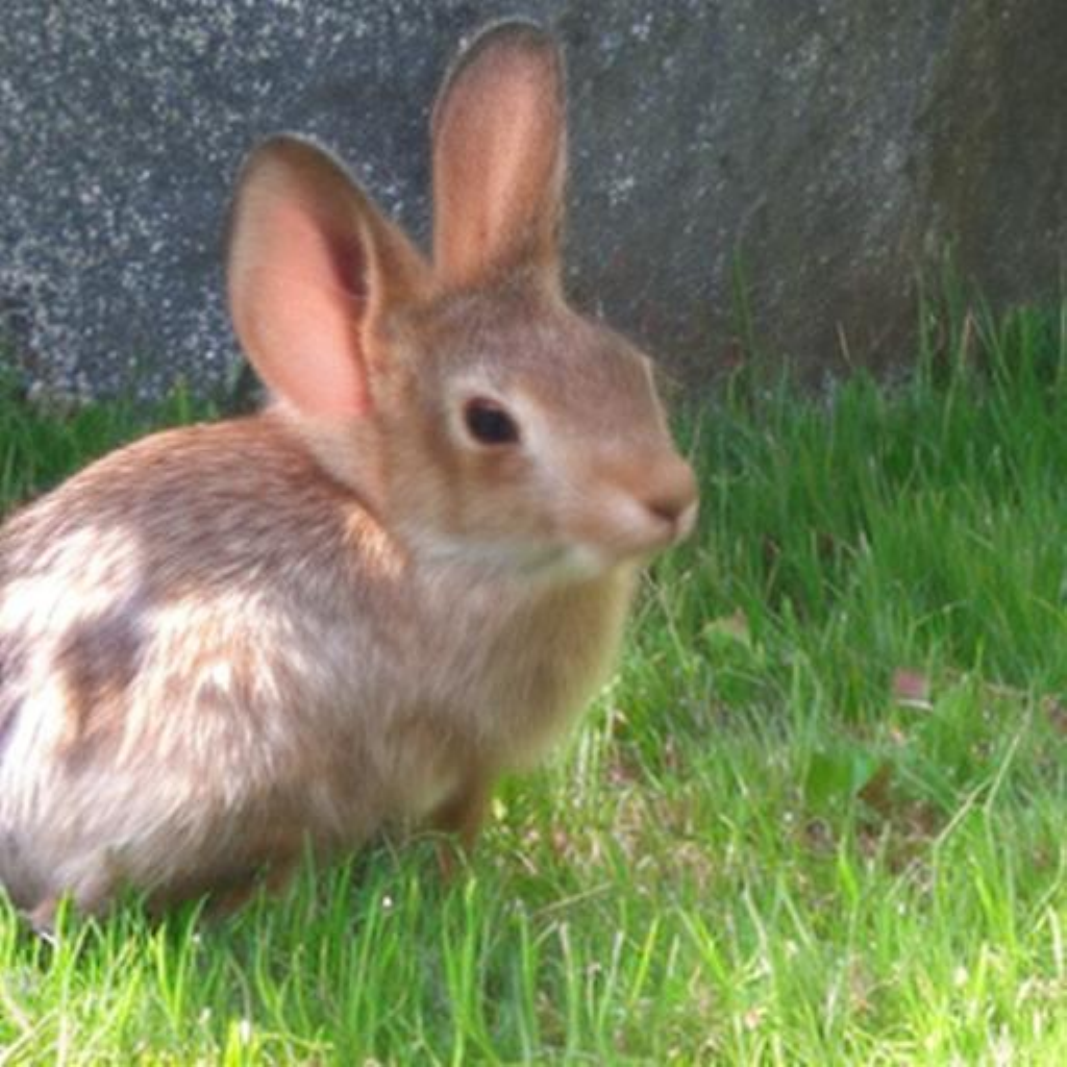}
\includegraphics[width=0.075\columnwidth]{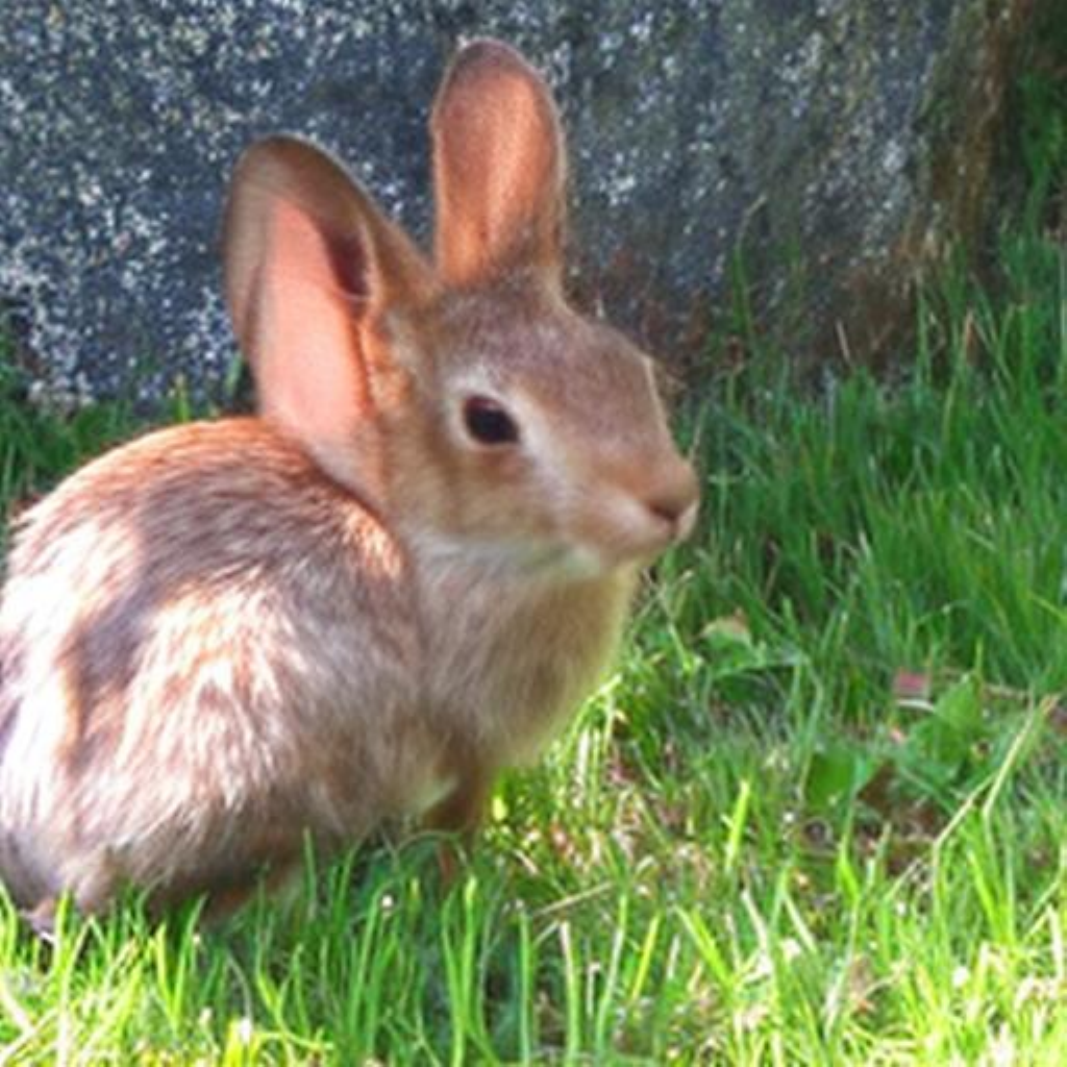}
\hspace{1ex}
\includegraphics[width=0.075\columnwidth]{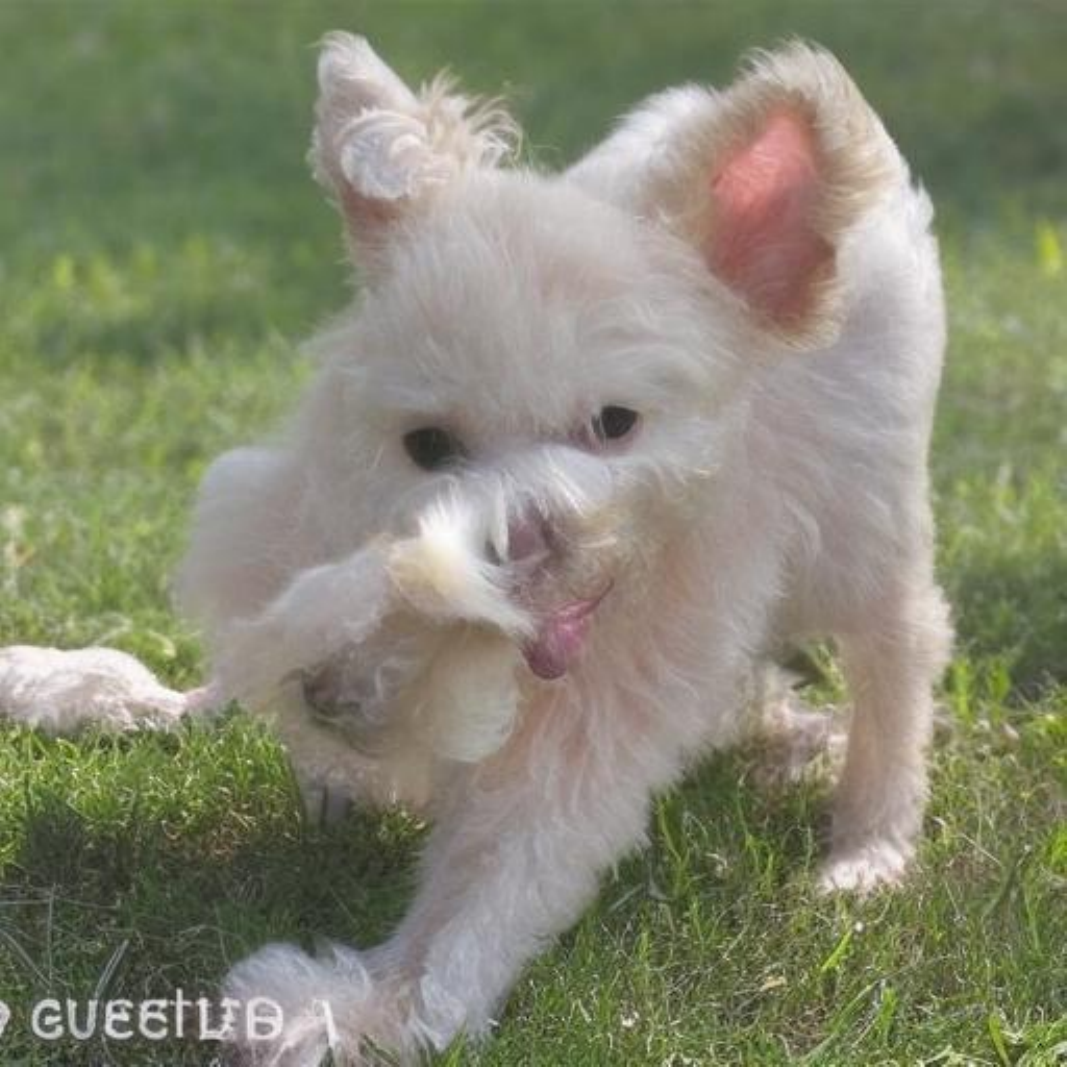}
\includegraphics[width=0.075\columnwidth]{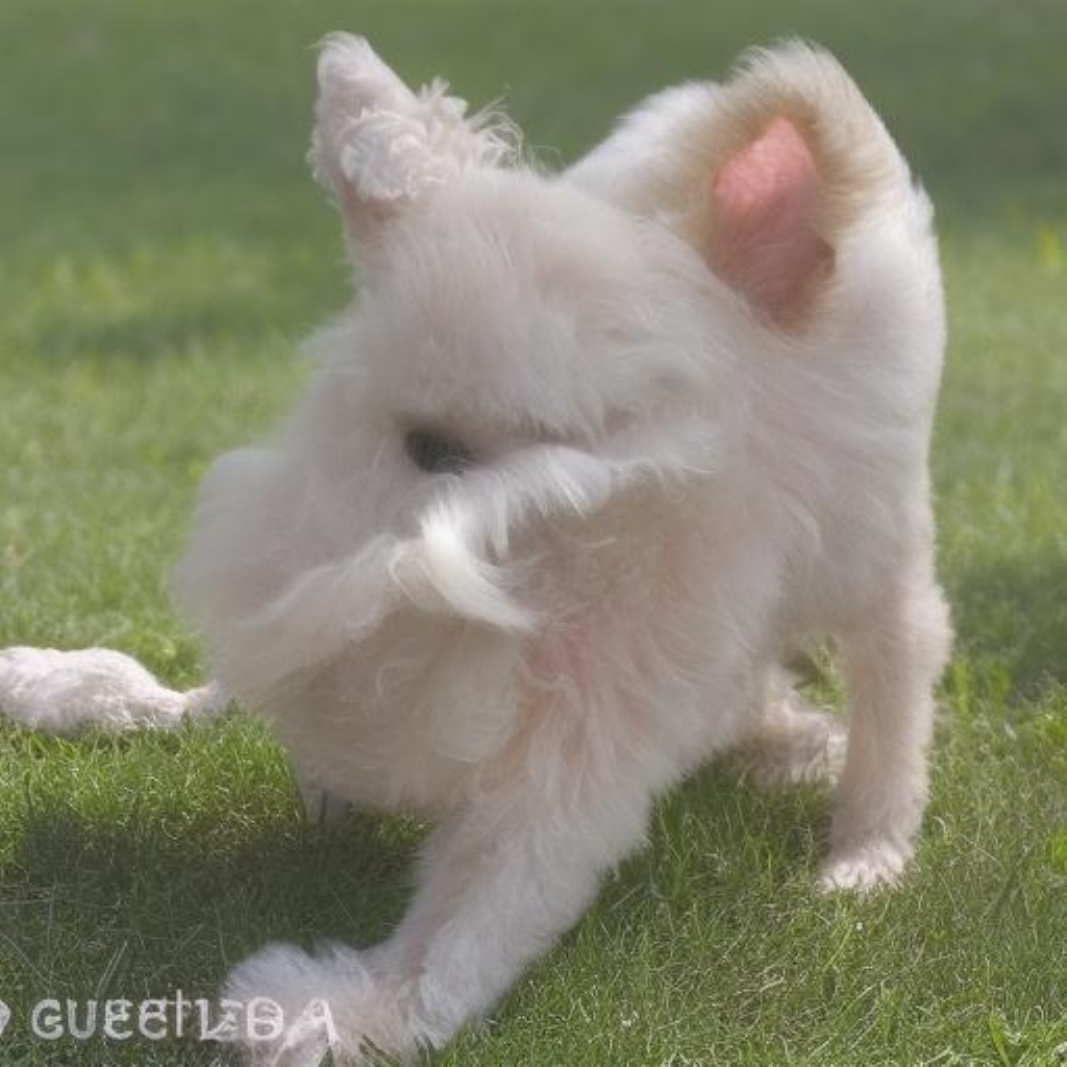}
\includegraphics[width=0.075\columnwidth]{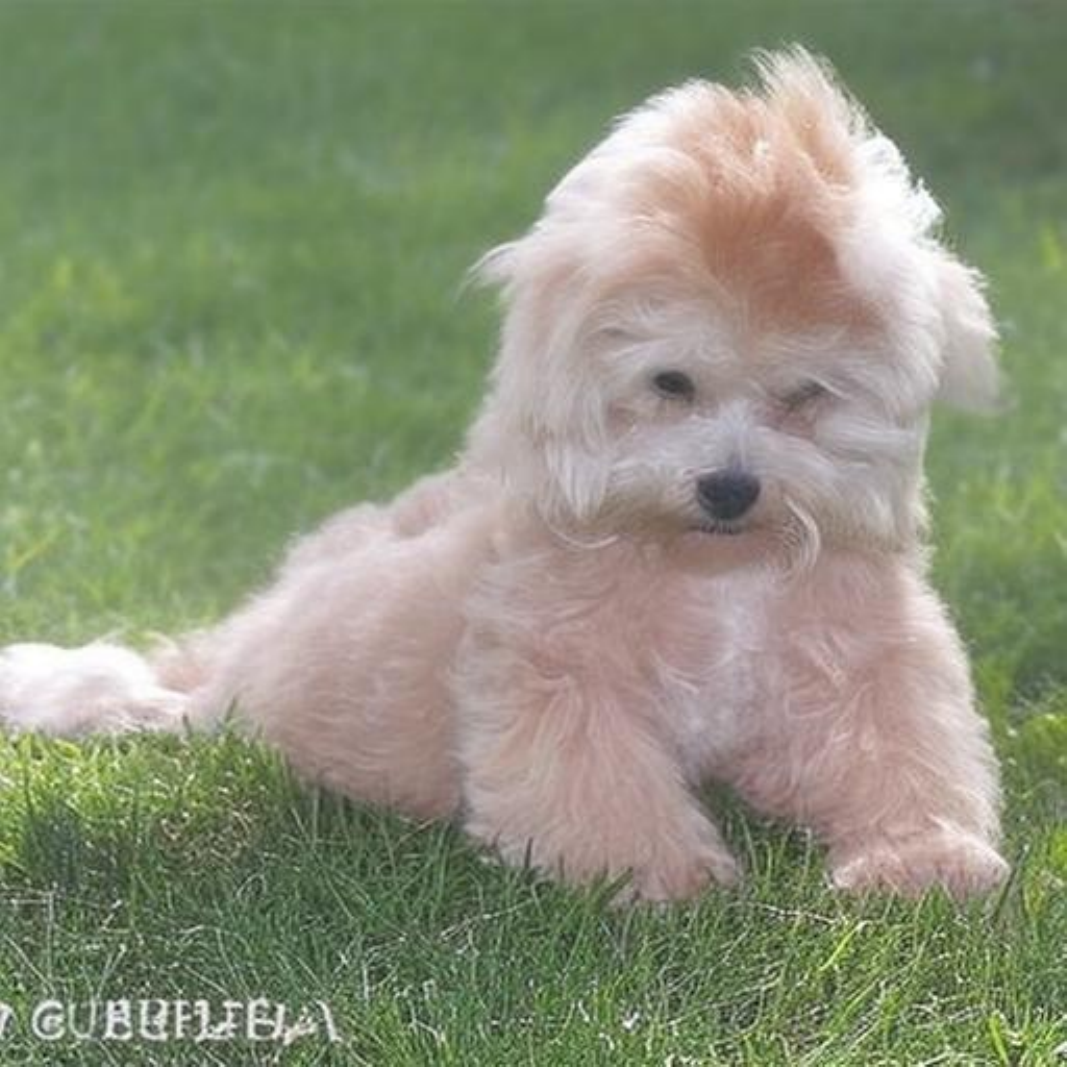}
\includegraphics[width=0.075\columnwidth]{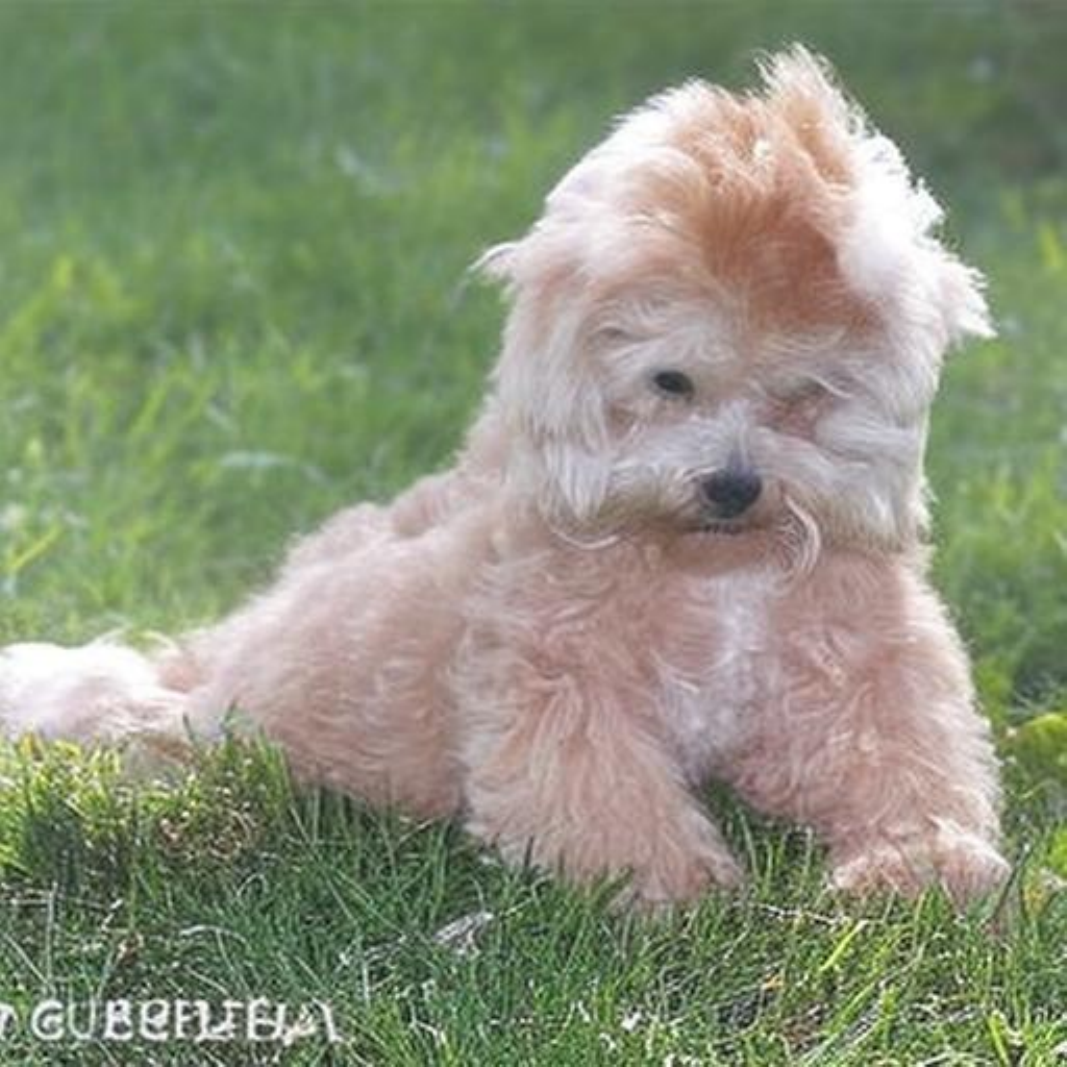}
\hspace{1ex}
\includegraphics[width=0.075\columnwidth]{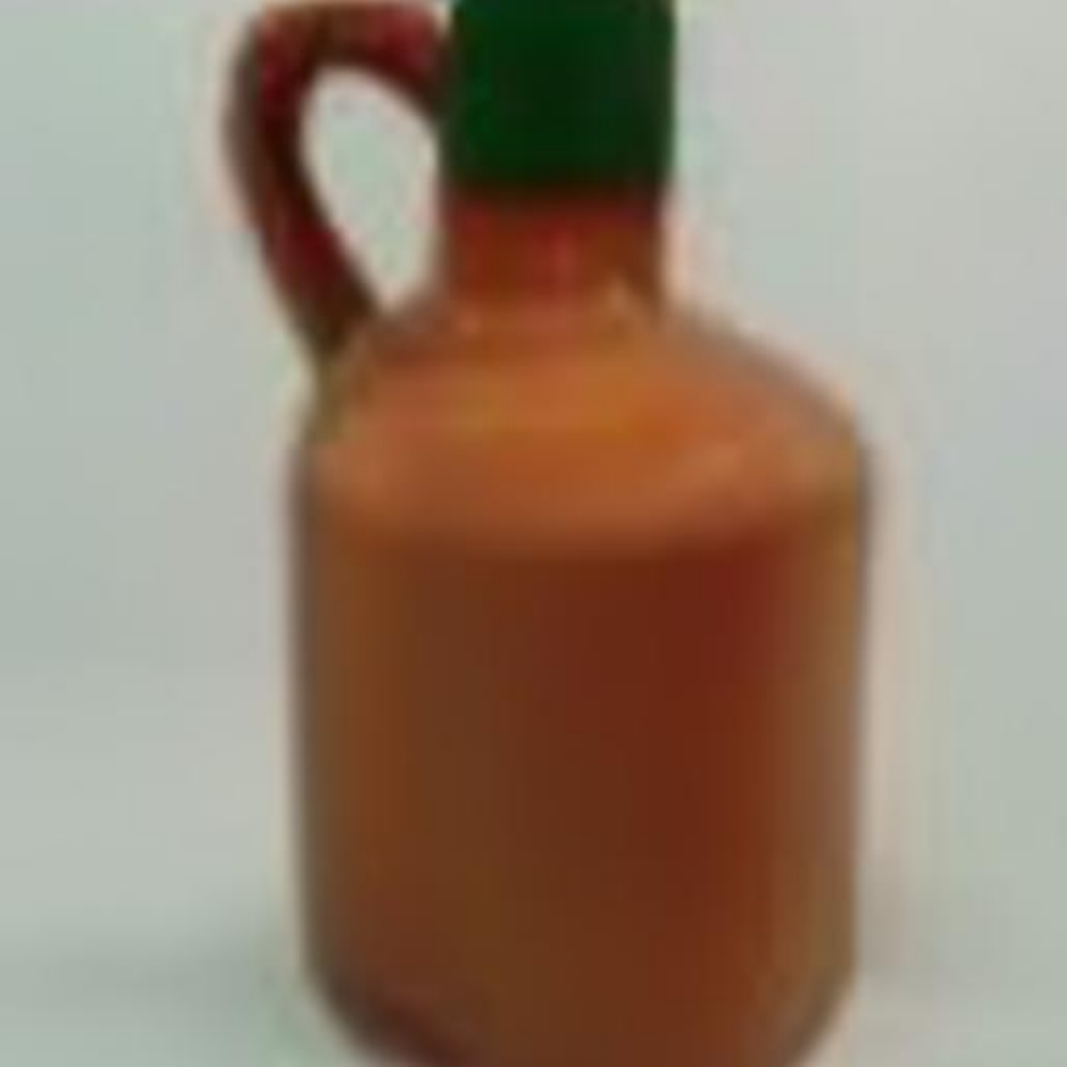}
\includegraphics[width=0.075\columnwidth]{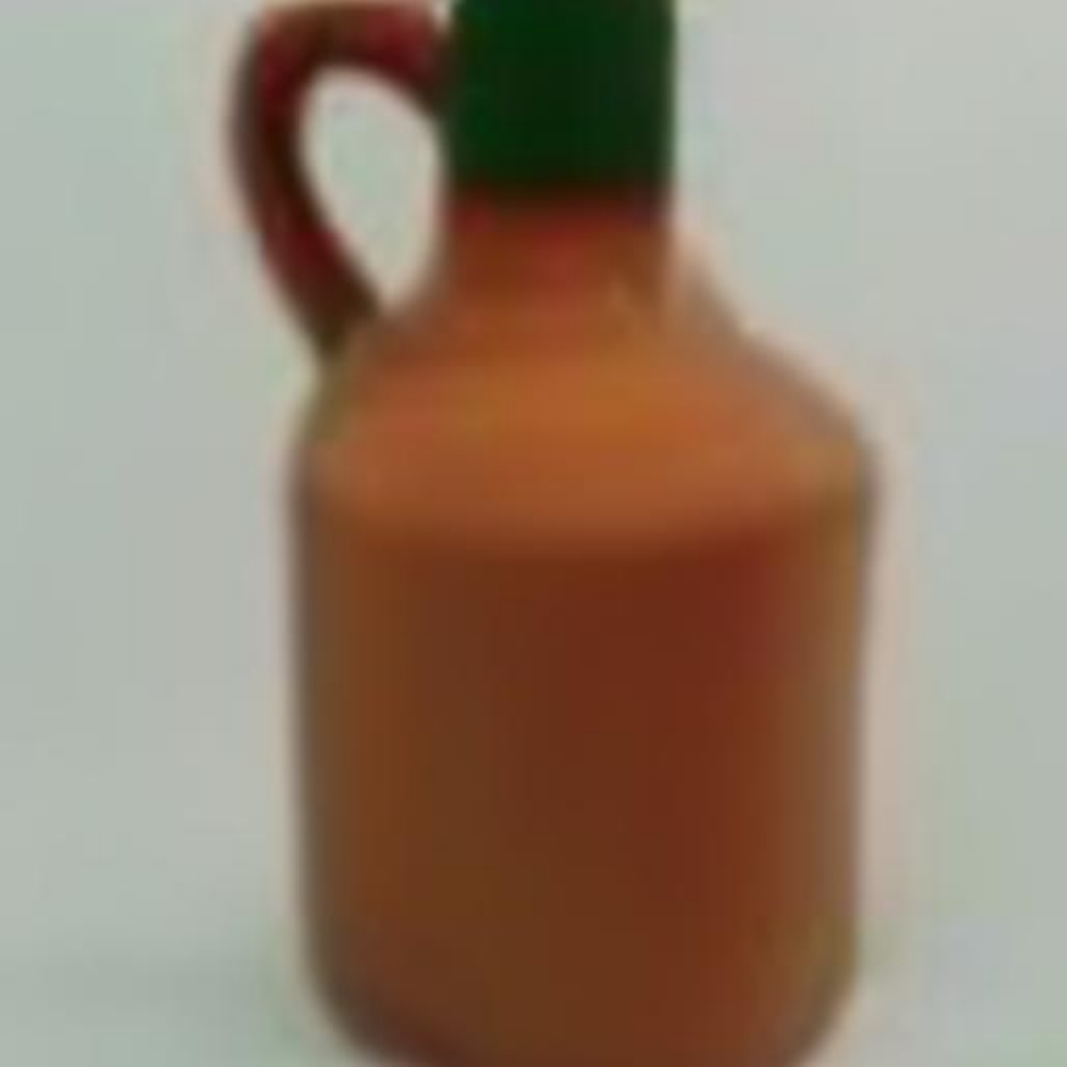}
\includegraphics[width=0.075\columnwidth]{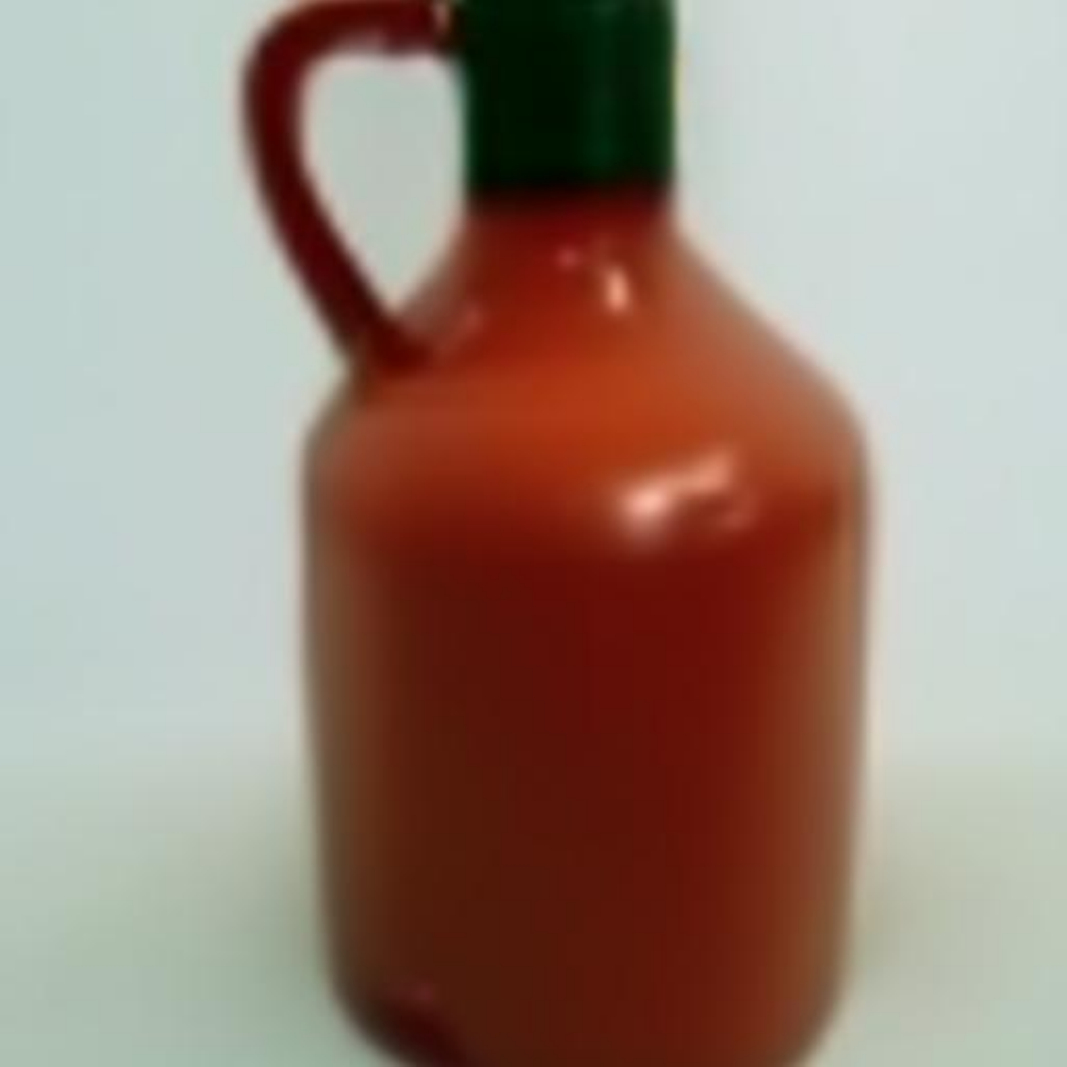}
\includegraphics[width=0.075\columnwidth]{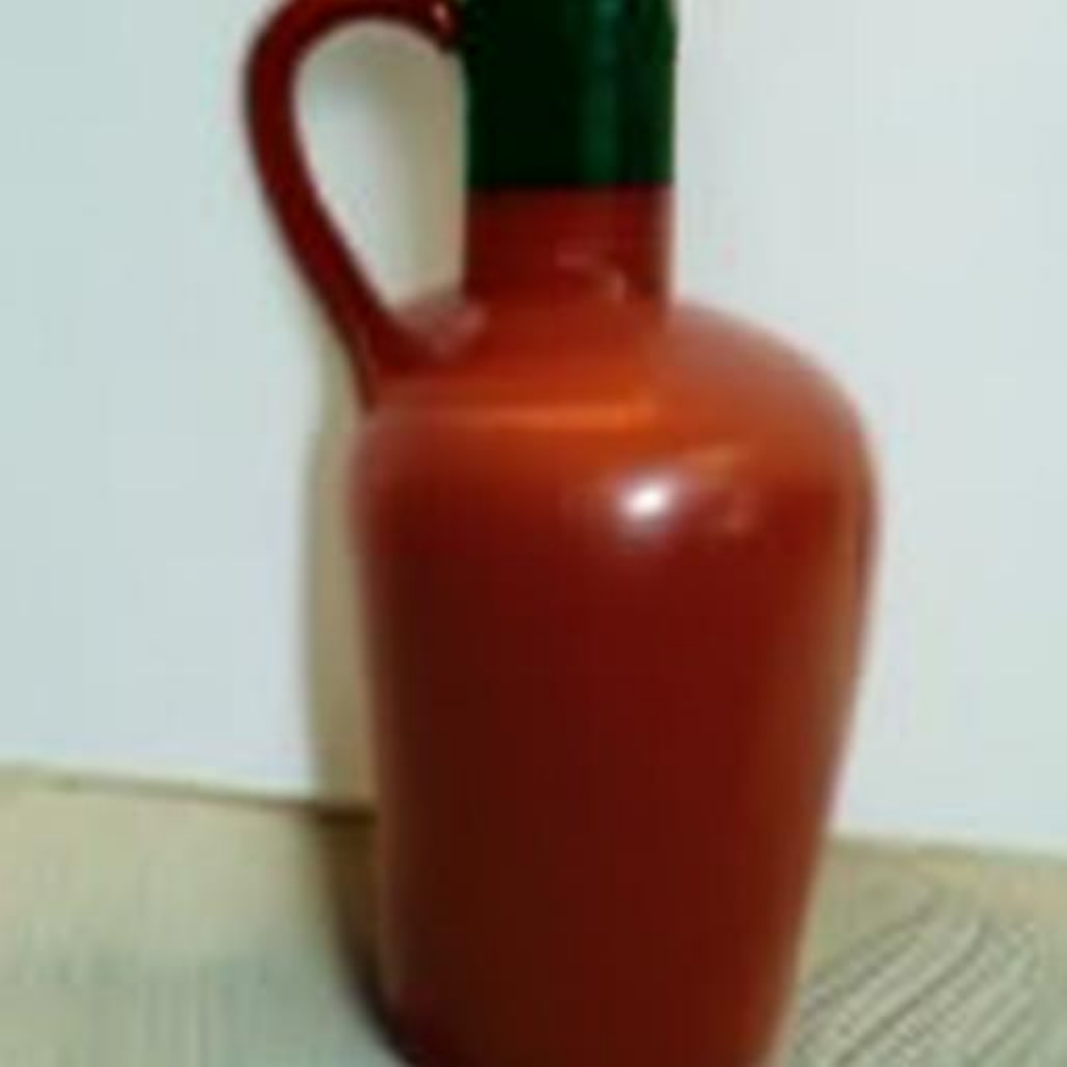}
\caption{\textbf{DiT with $40$ Steps.} For each group of four images, from left to right: baseline, ablation, and generated images with variance scalings $1.15$ and $1.35$.
%
The classes are: \emph{cottontail bunny}, \emph{miniature poodle}, and \emph{whisky jug}.
}
    \label{fig:dit-o}
    \end{figure*}

We start our analysis with $40$ steps. Samples are shown in Fig.~\ref{fig:dit-o} and Fig.~\ref{fig:dit-40} from Appendix~\ref{app:exp}. As in previous diffusion models, S2EP can enhance image quality by presenting images with more contrast and more vivid 
colors. However, it can also produce images with less color intensity; 
see the first two rows of Fig.~\ref{fig:dit-40-things} in Appendix~\ref{app:exp}. Moreover, across diverse classes, S2EP  
often leads to considerable semantic changes compared to the baseline (Fig.~\ref{fig:dit-40}), often improving the overall appearance corresponding to the class subject of the sampled image.
%
An observed shortcoming of S2EP 
is that 
it 
can 
lead to the appearance of spurious artifacts on some images, 
as well as oversaturation; see the last two rows in Fig.~\ref{fig:dit-40-things}. 
%
For larger numbers of denoising steps, it is more difficult to calibrate the variance scaling without 
noticeable 
image quality degradation. We thus consider only $80$ steps. Samples are shown in Fig~\ref{fig:dit-80} from Appendix~\ref{app:exp}. We observe a larger tendency towards more contrast and brighter colors than 
before---%
which can enhance image quality but also lead to oversaturated images.
%
%

Our observations across DDPM, LD and DiT reinforce the 
idea that a lower number of steps is more appropriate for the effectiveness of our method. 
Moreover, 
the baseline improves with more steps, and so 
possible enhancements by our method 
are also visually less effective. 

\subsection{Stable Diffusion (SD) Model}

\begin{figure*}[t]
    \centering
\includegraphics[width=0.075\columnwidth]{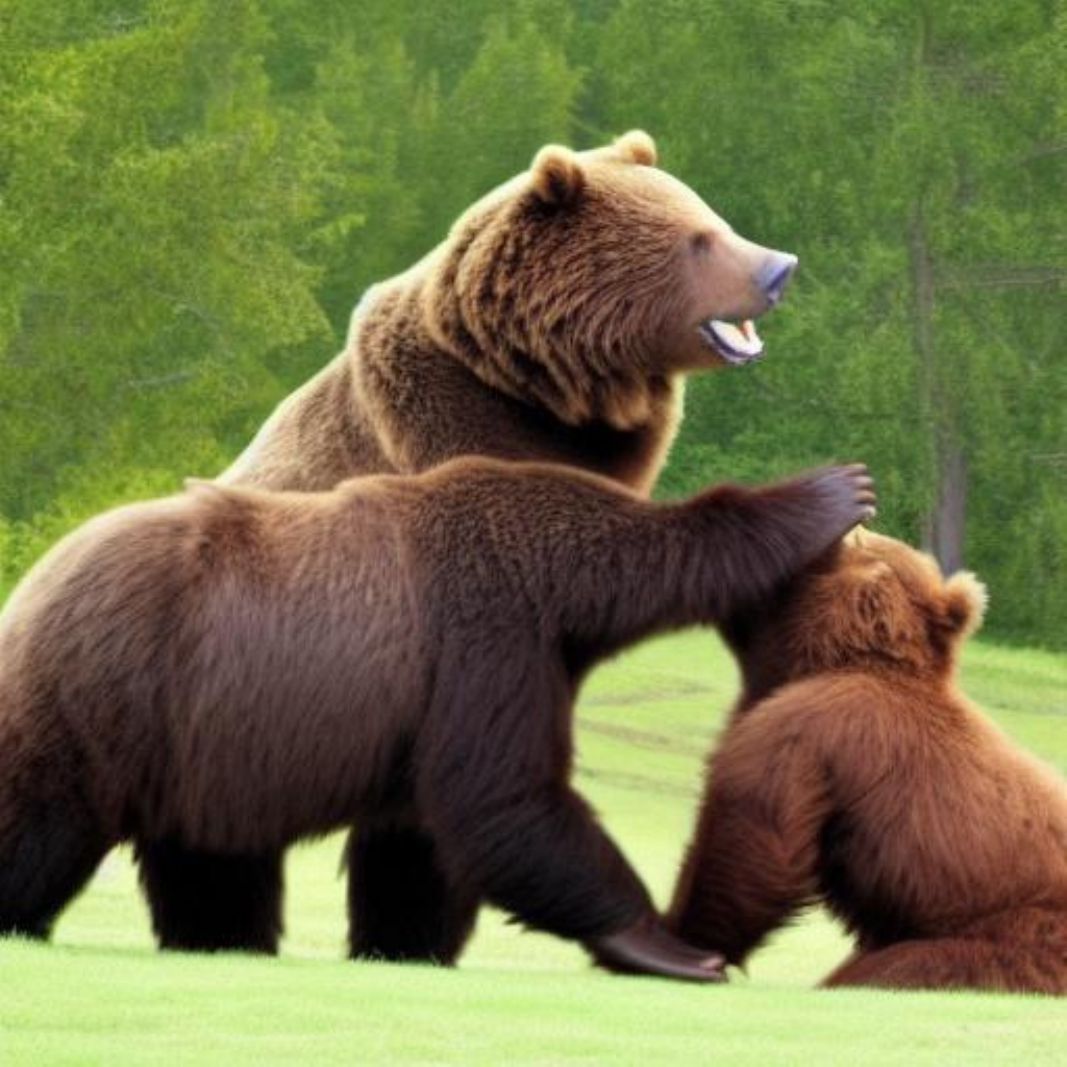}
\includegraphics[width=0.075\columnwidth]{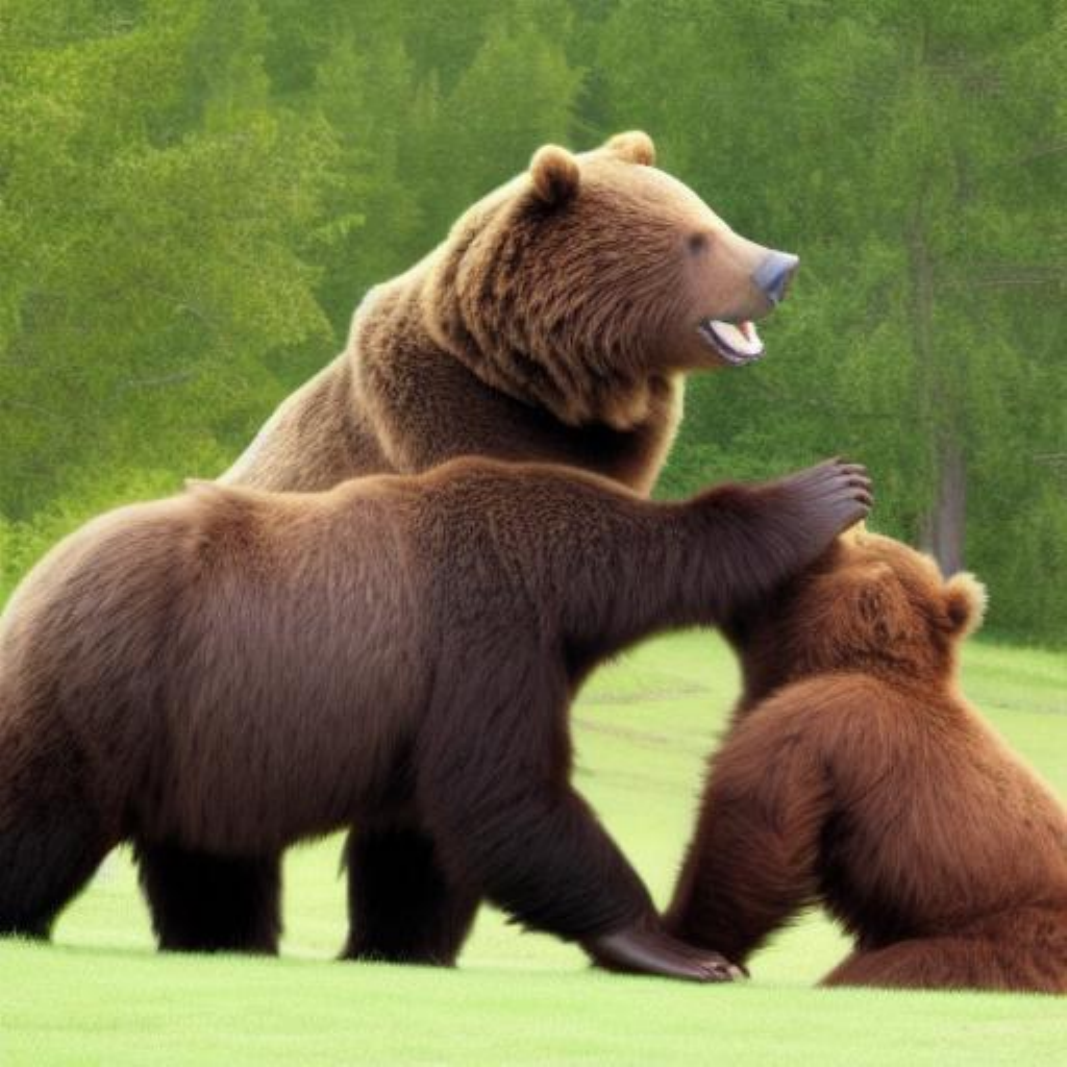}
\includegraphics[width=0.075\columnwidth]{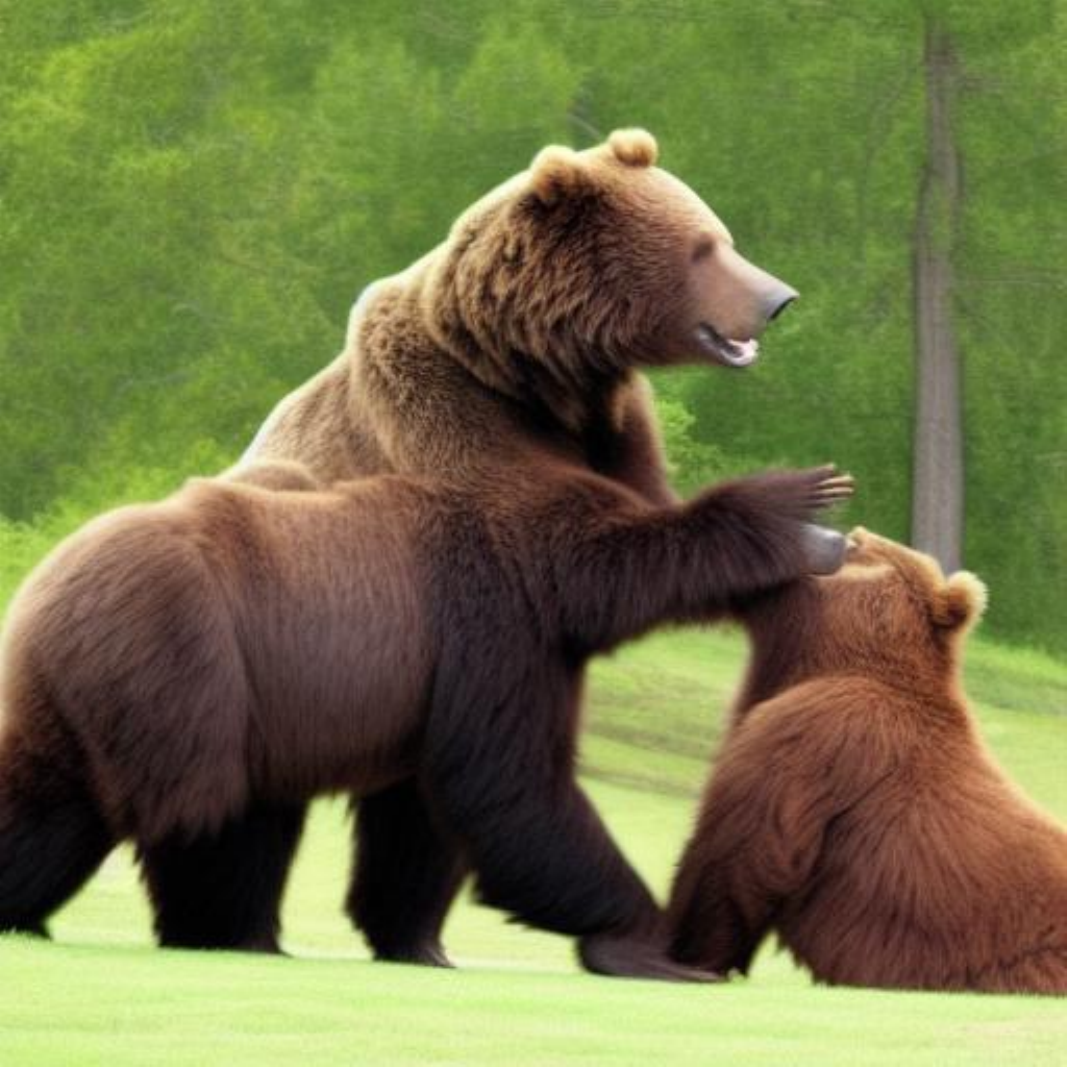}
\includegraphics[width=0.075\columnwidth]{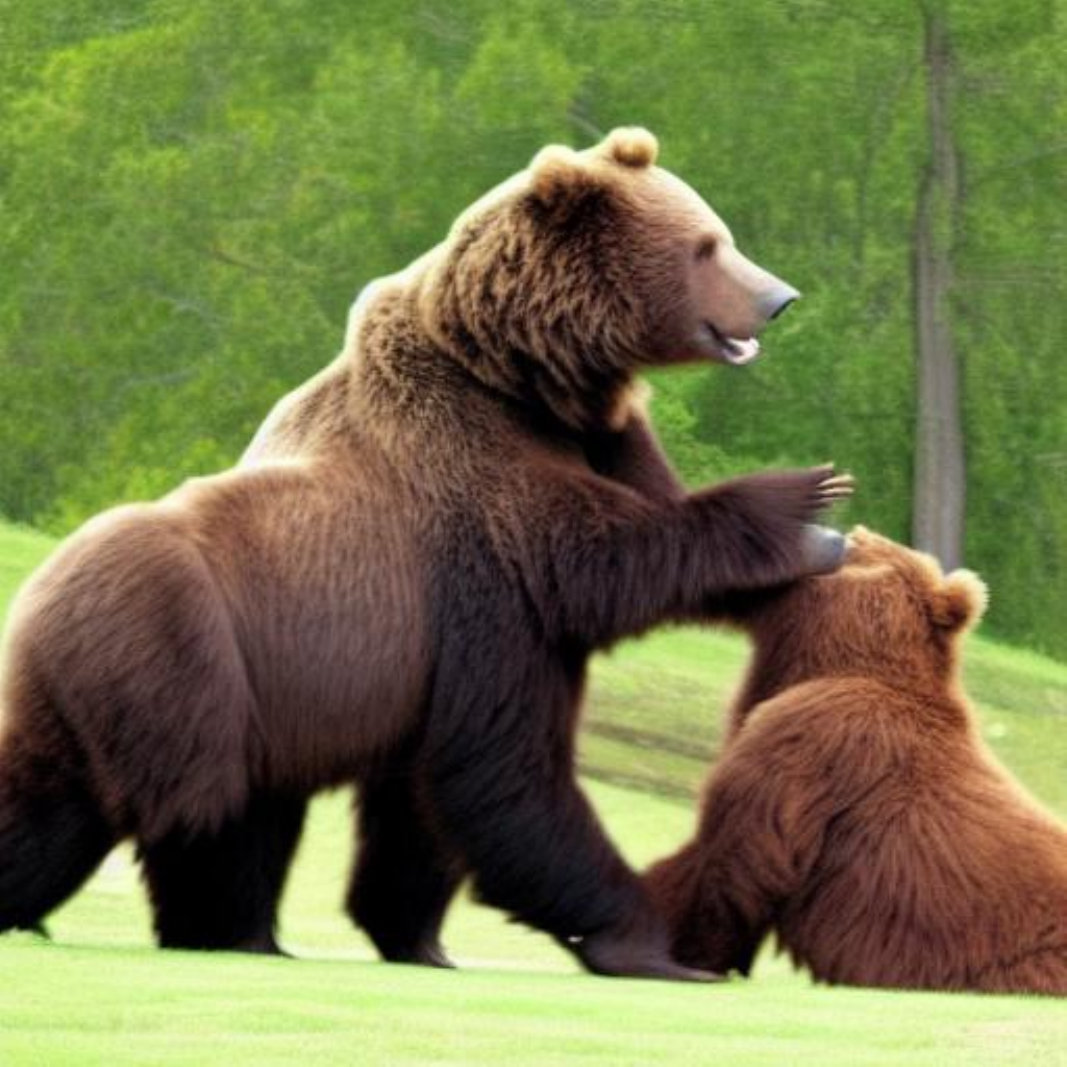}
\hspace{1ex}
\includegraphics[width=0.075\columnwidth]{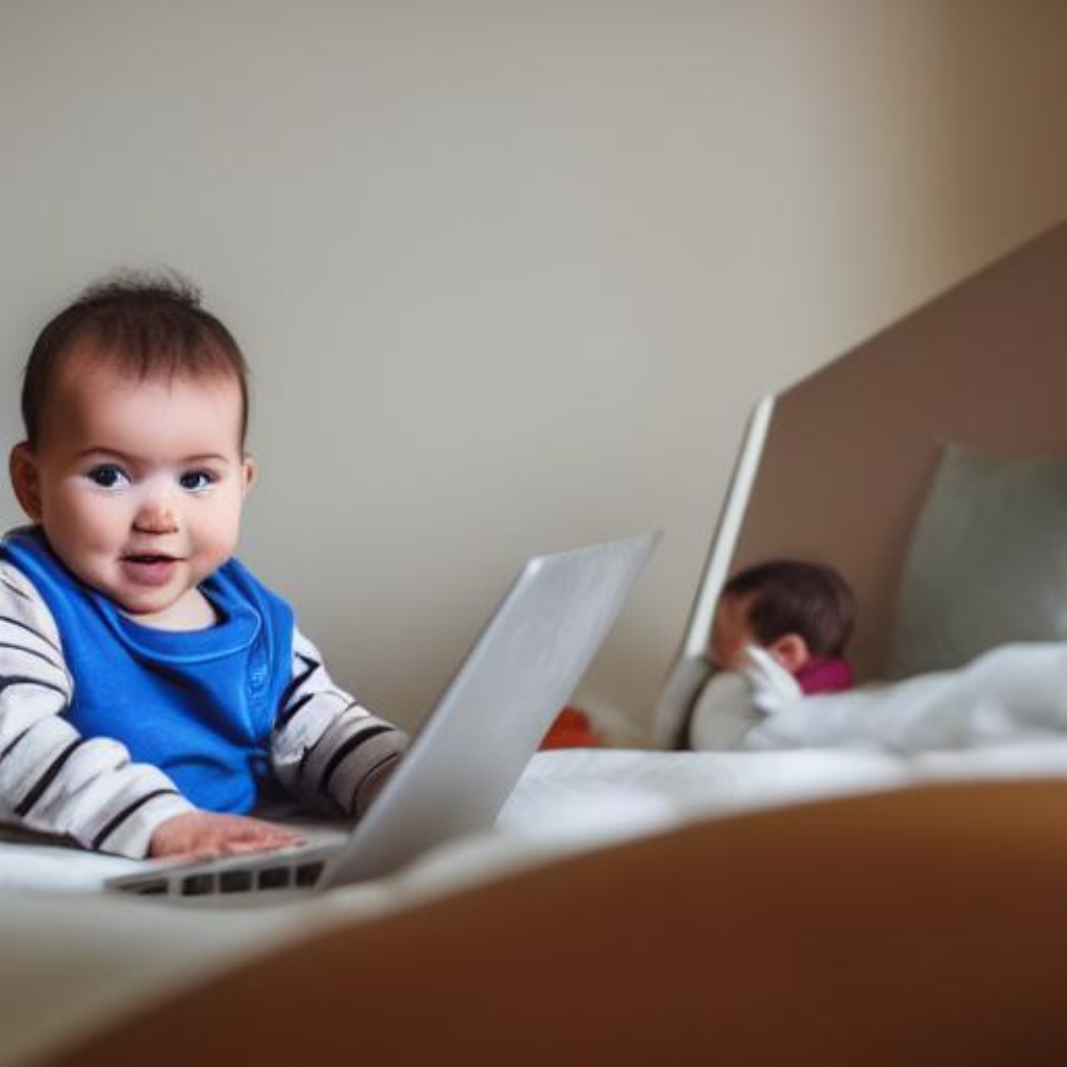}
\includegraphics[width=0.075\columnwidth]{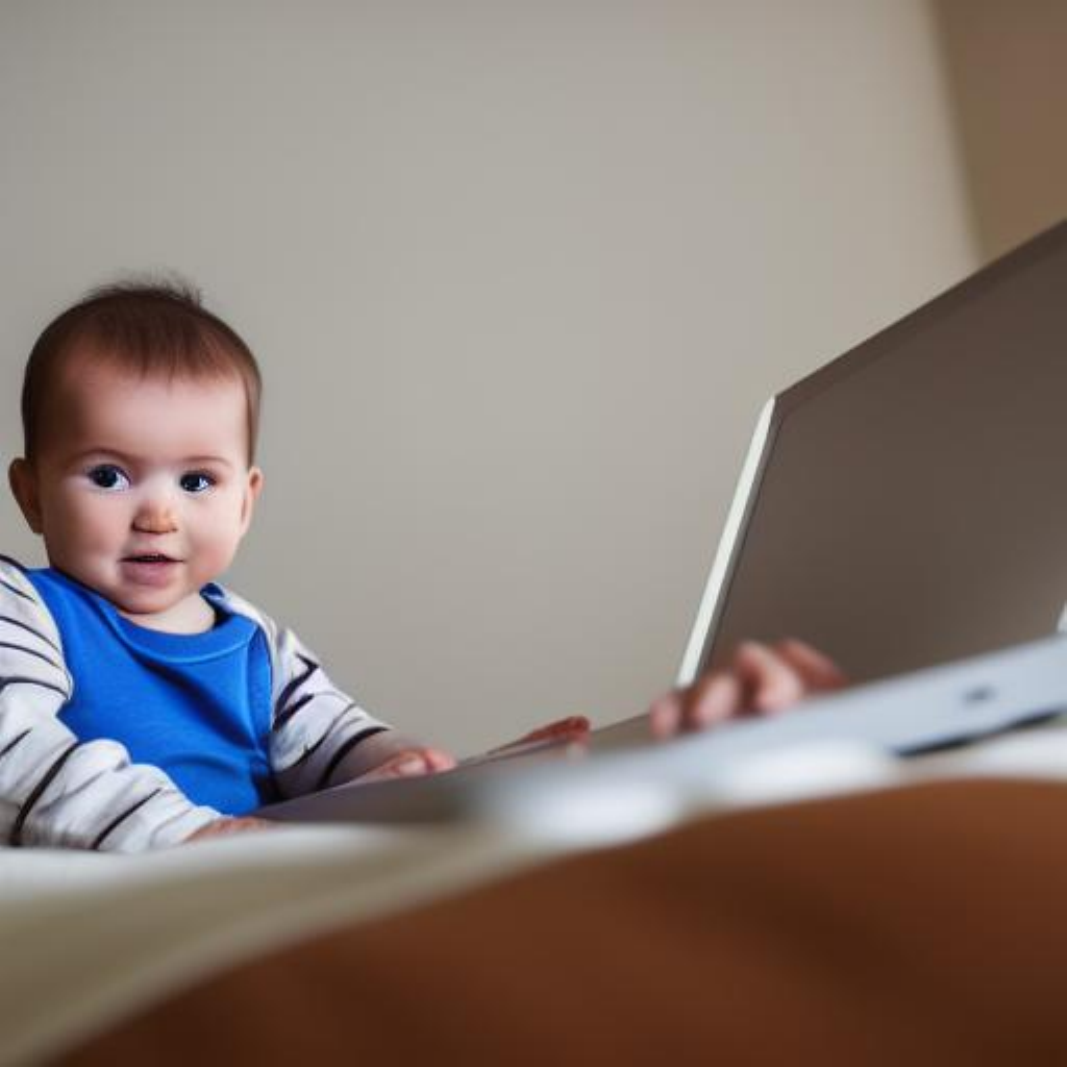}
\includegraphics[width=0.075\columnwidth]{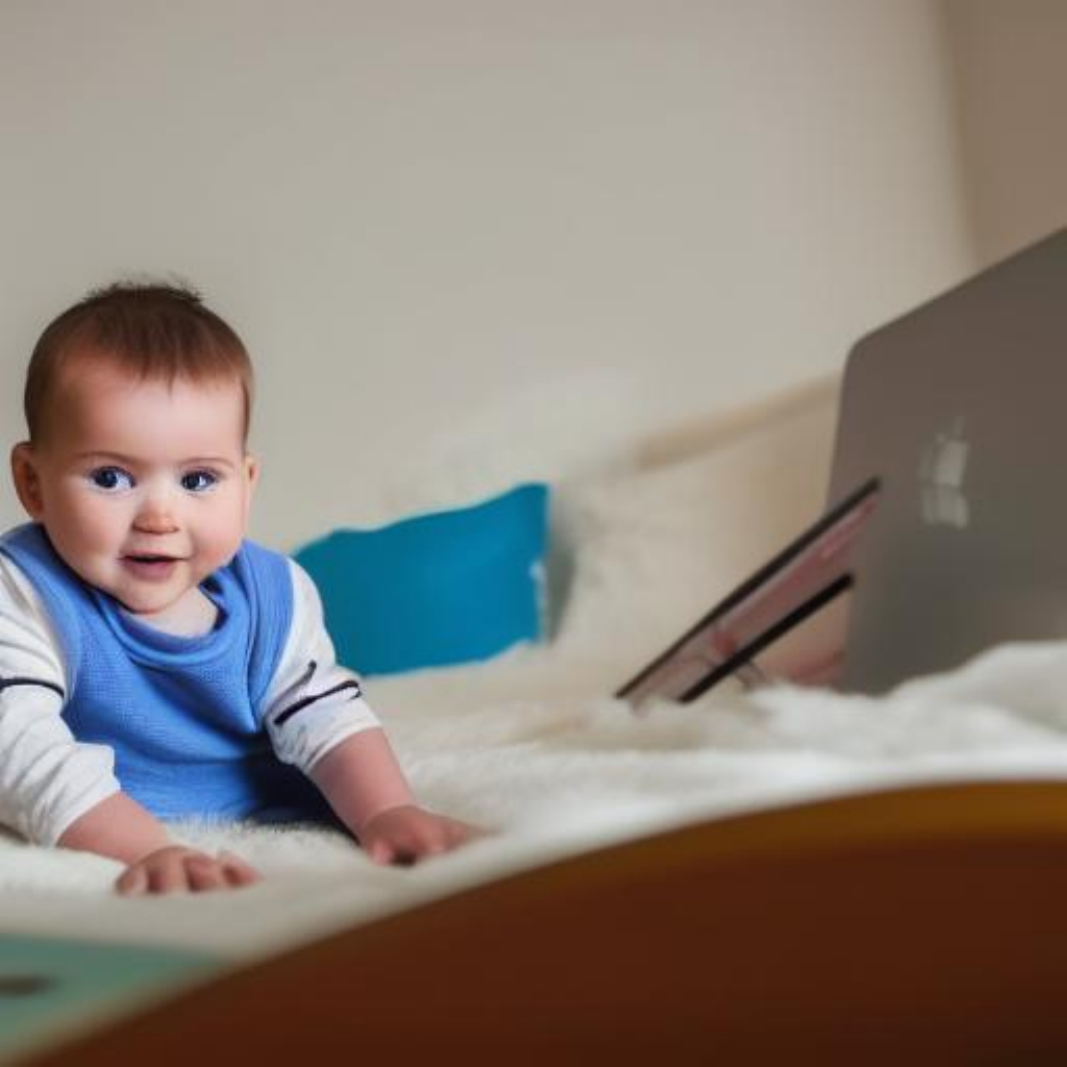}
\includegraphics[width=0.075\columnwidth]{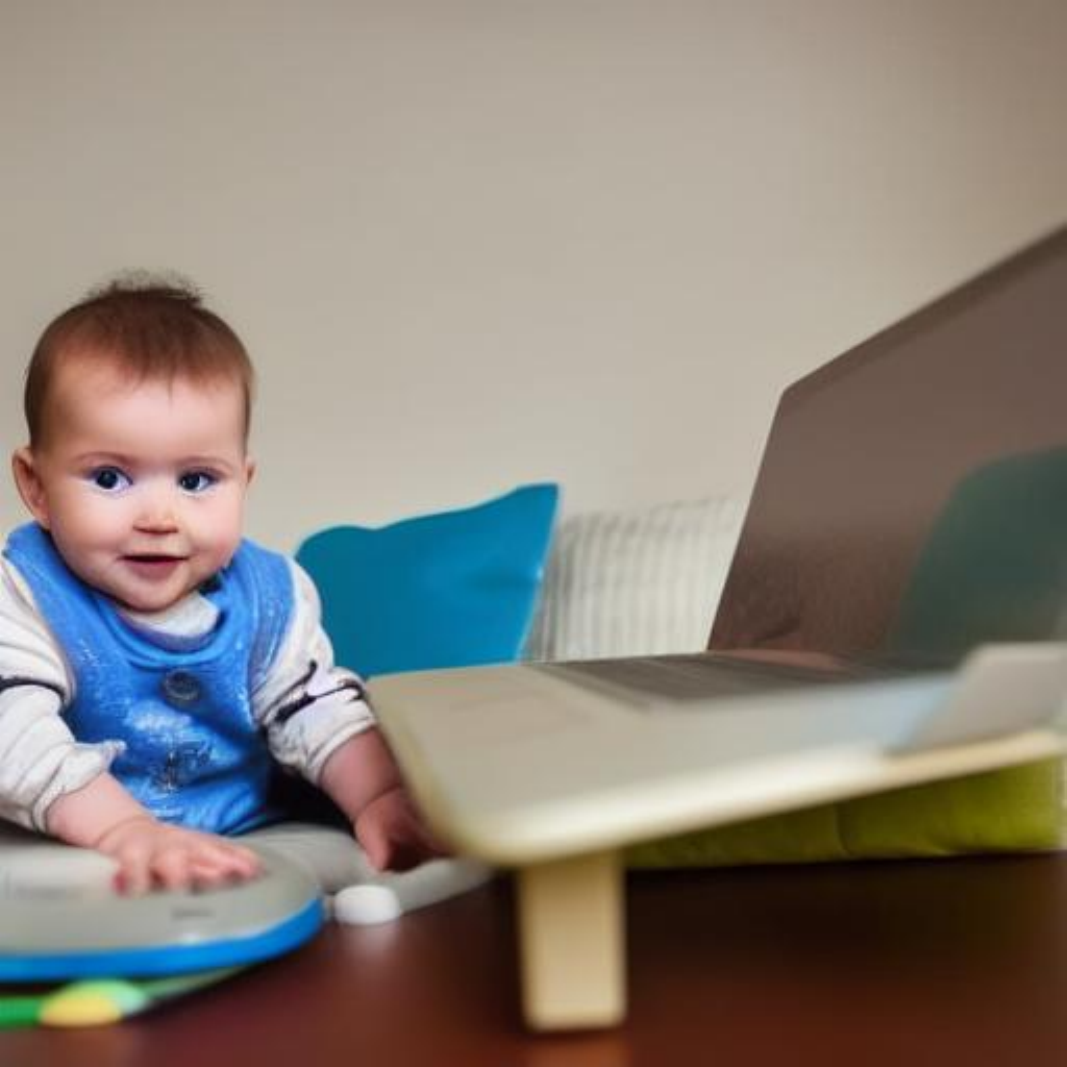}
\hspace{1ex}
\includegraphics[width=0.075\columnwidth]{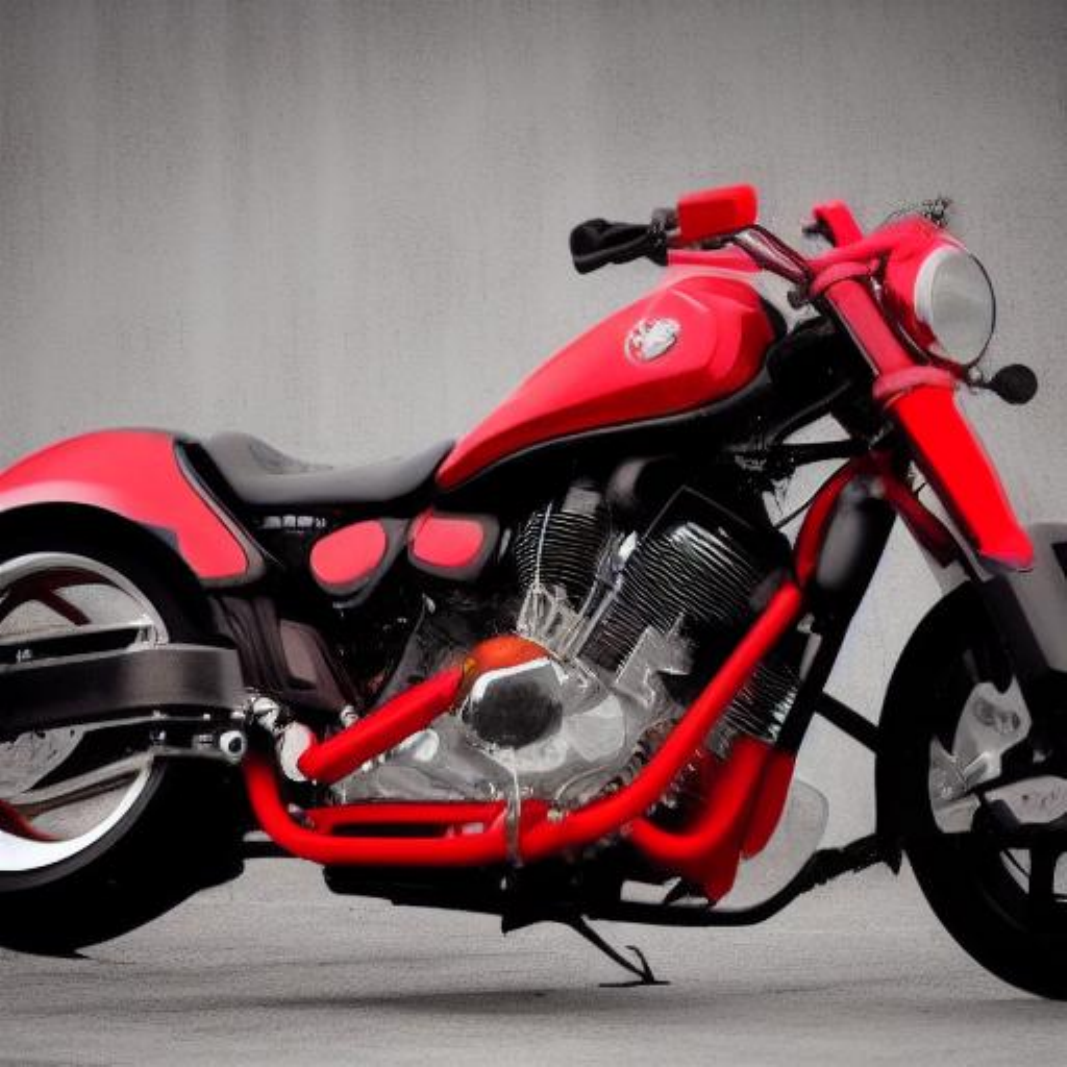}
\includegraphics[width=0.075\columnwidth]{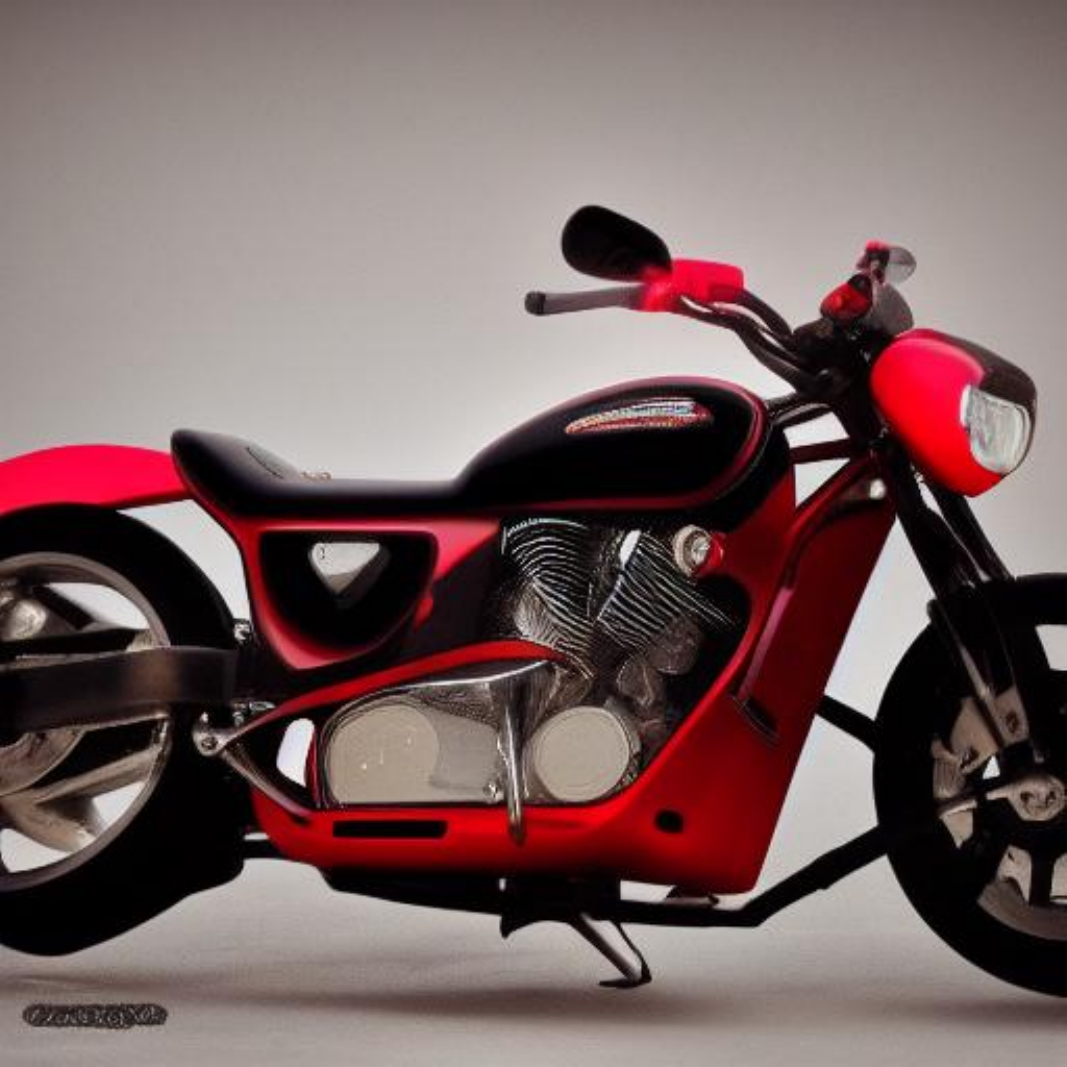}
\includegraphics[width=0.075\columnwidth]{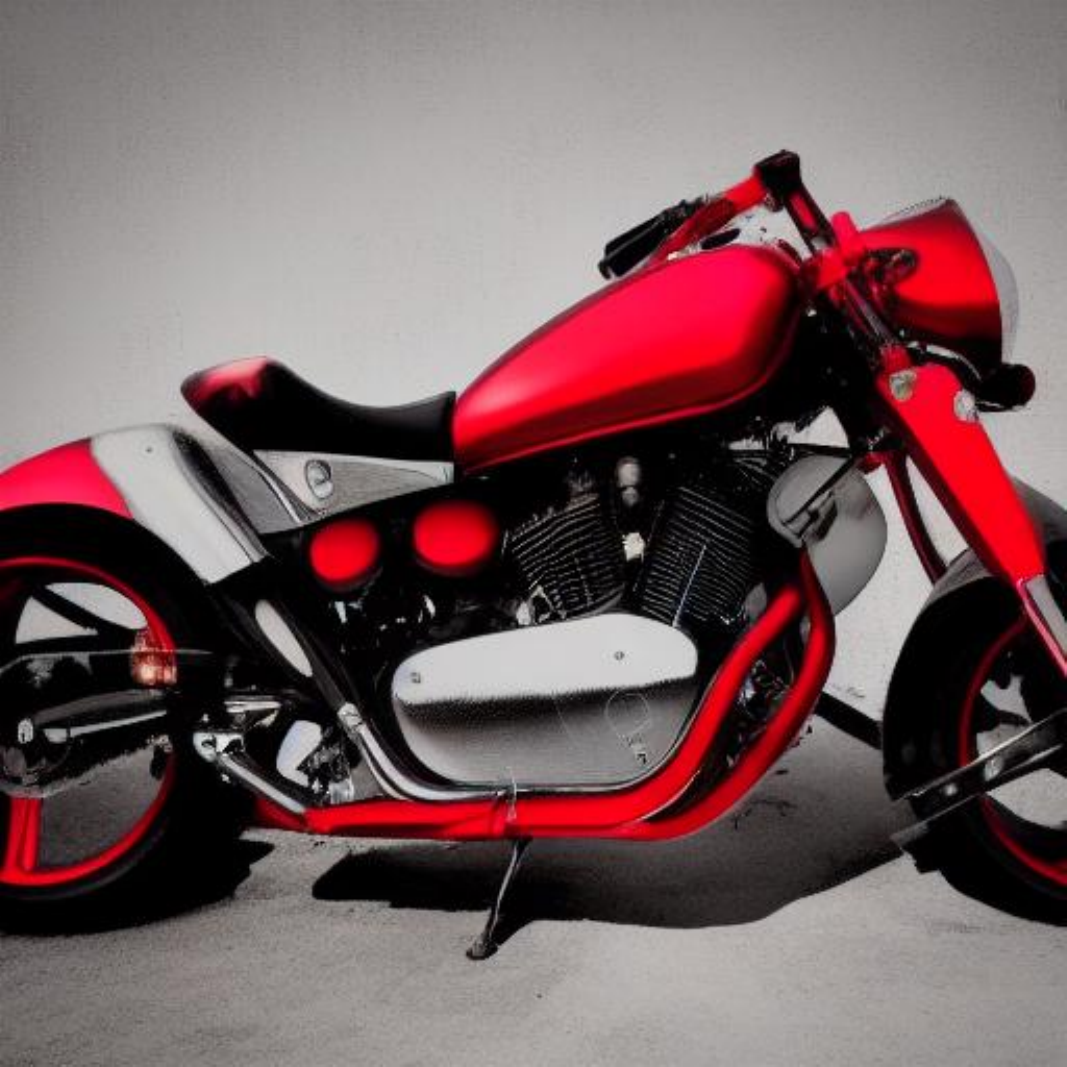}
\includegraphics[width=0.075\columnwidth]{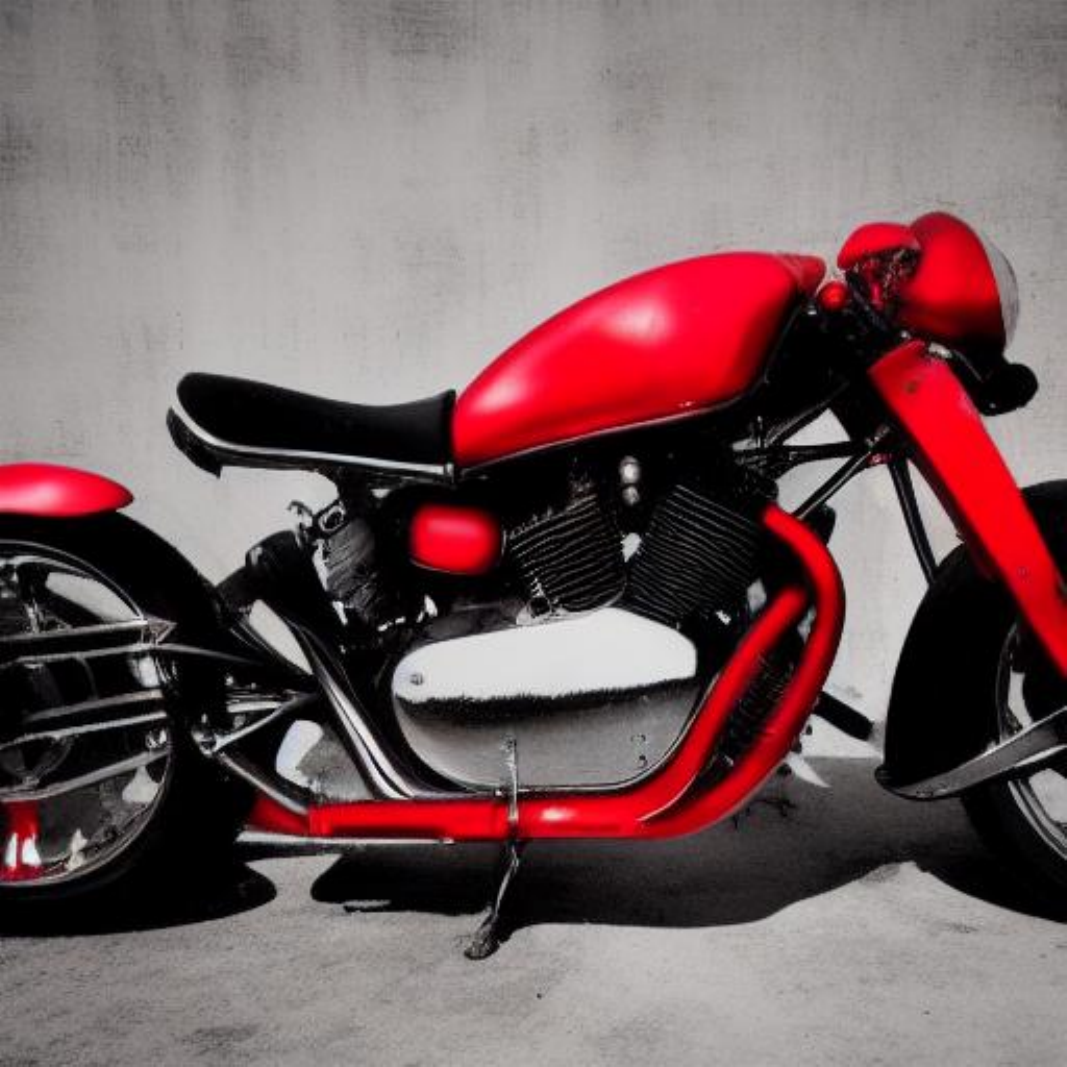}
\\
\vspace{-8pt}
\rule{0.9\textwidth}{0.4pt}
\\
\vspace{3pt}
\includegraphics[width=0.075\columnwidth]{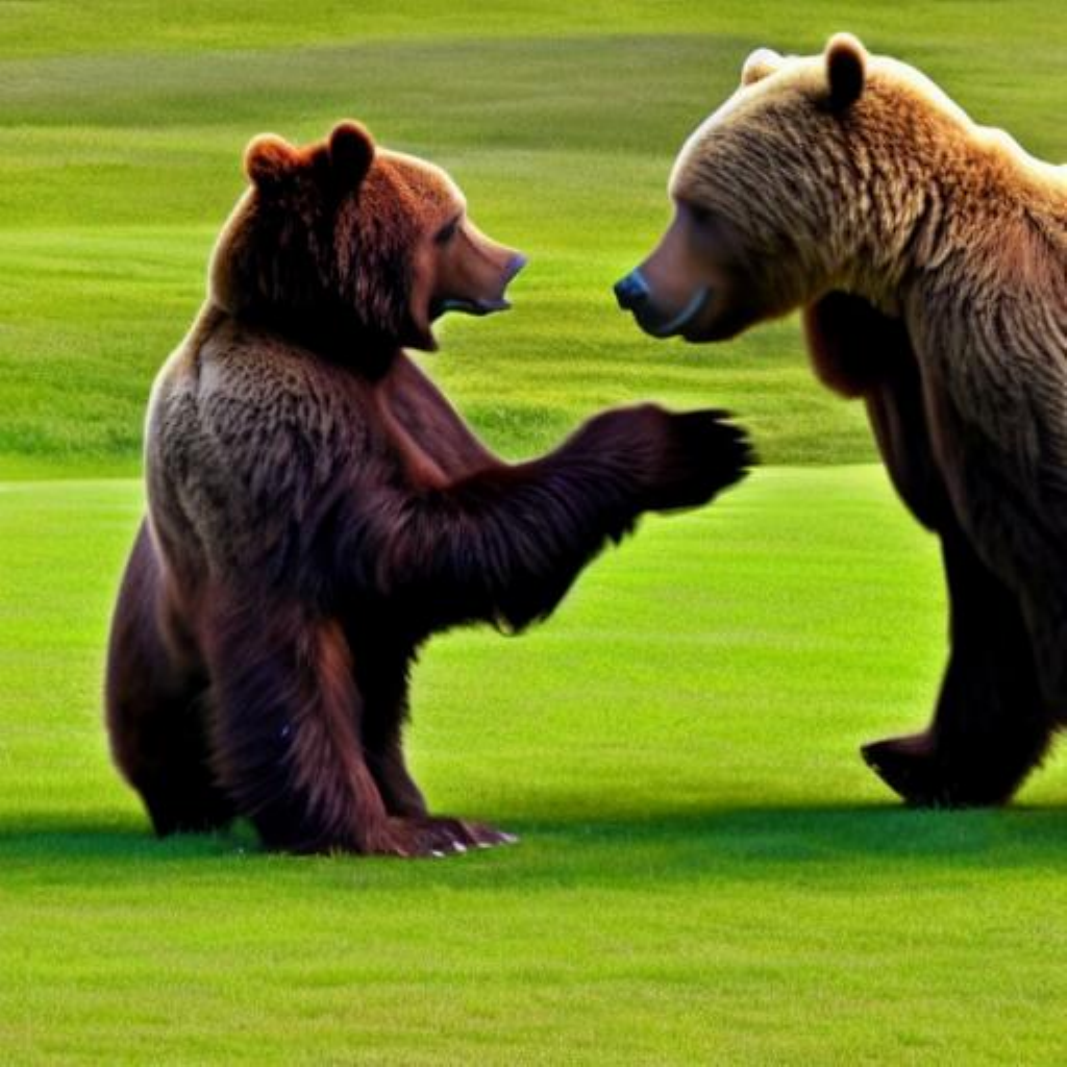}
\includegraphics[width=0.075\columnwidth]{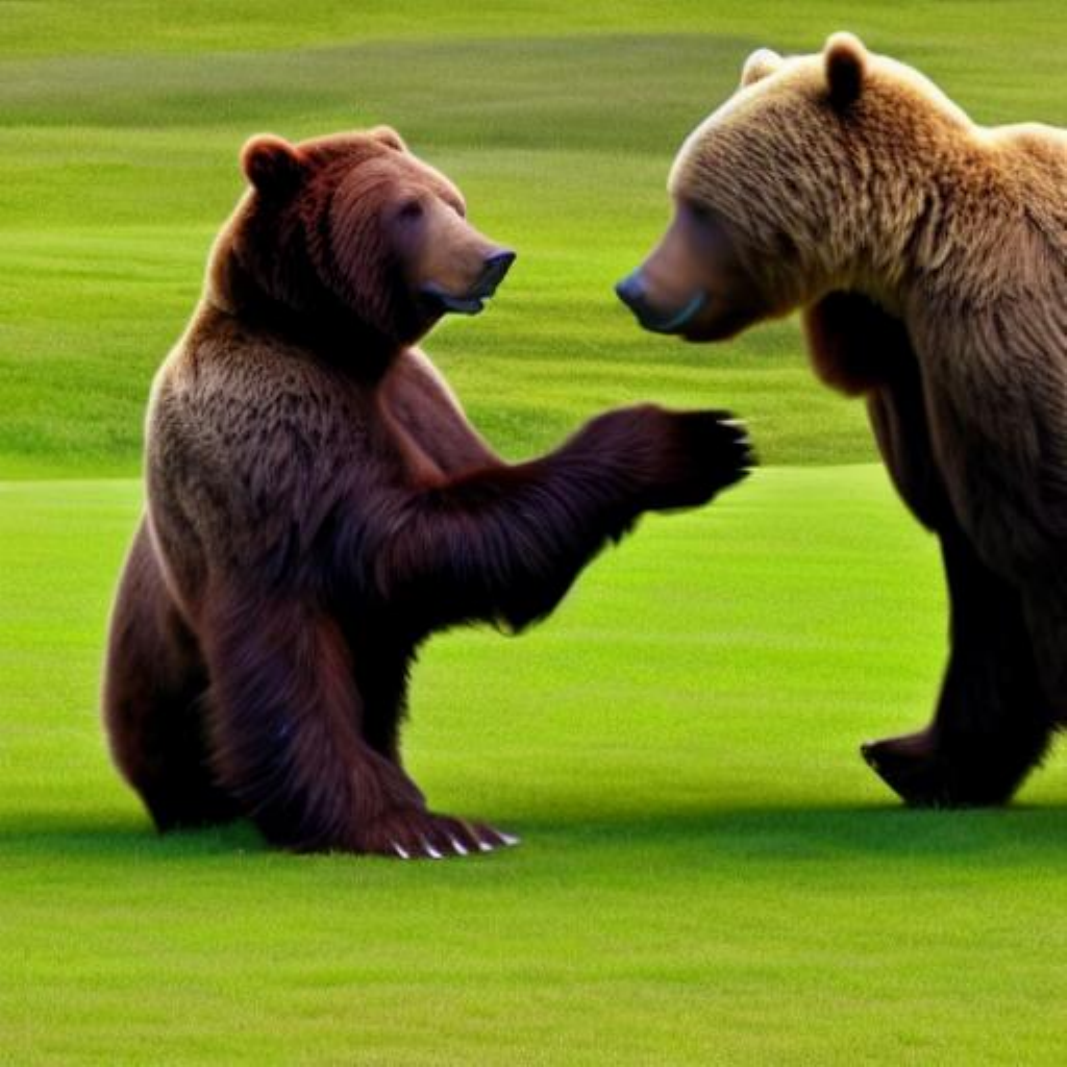}
\includegraphics[width=0.075\columnwidth]{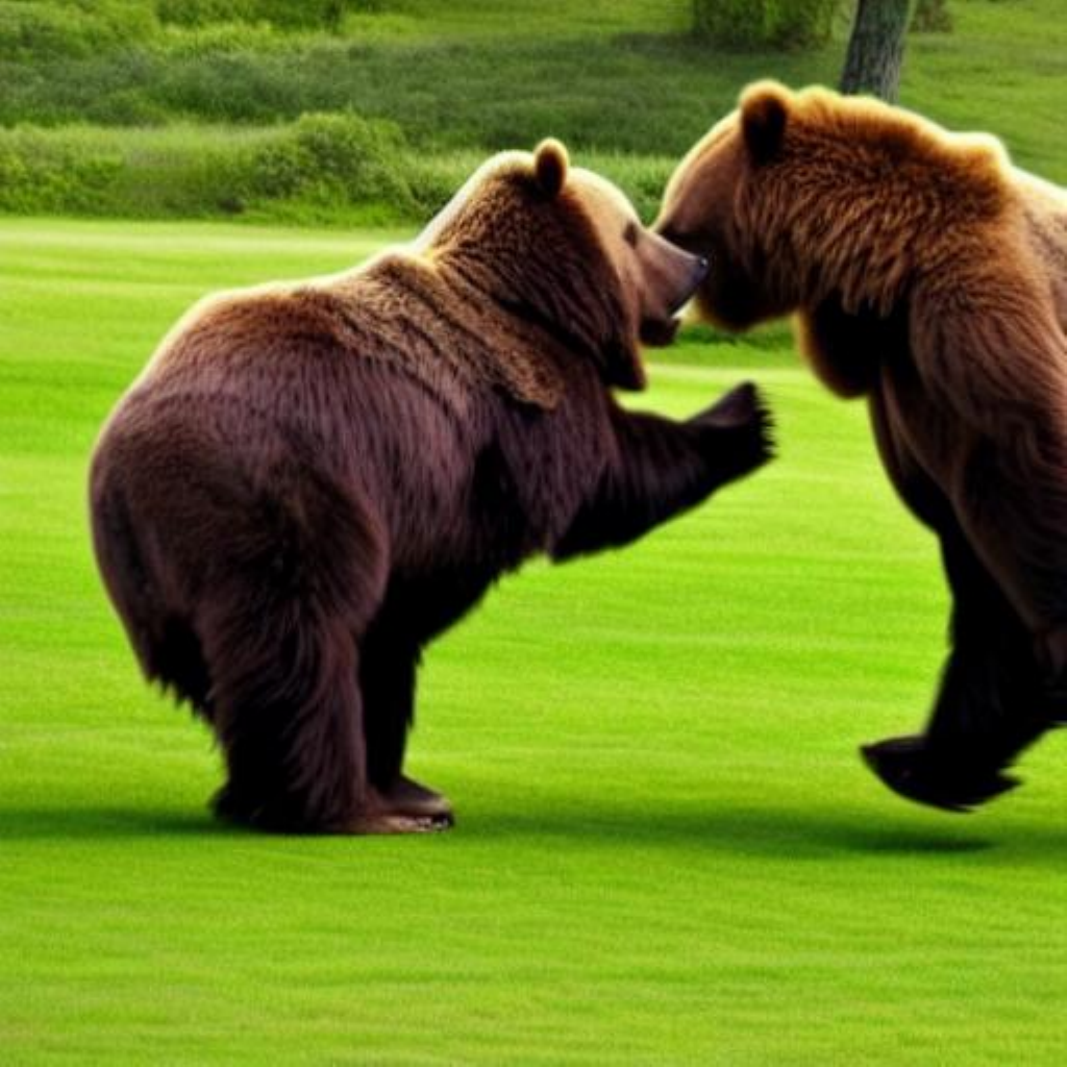}
\includegraphics[width=0.075\columnwidth]{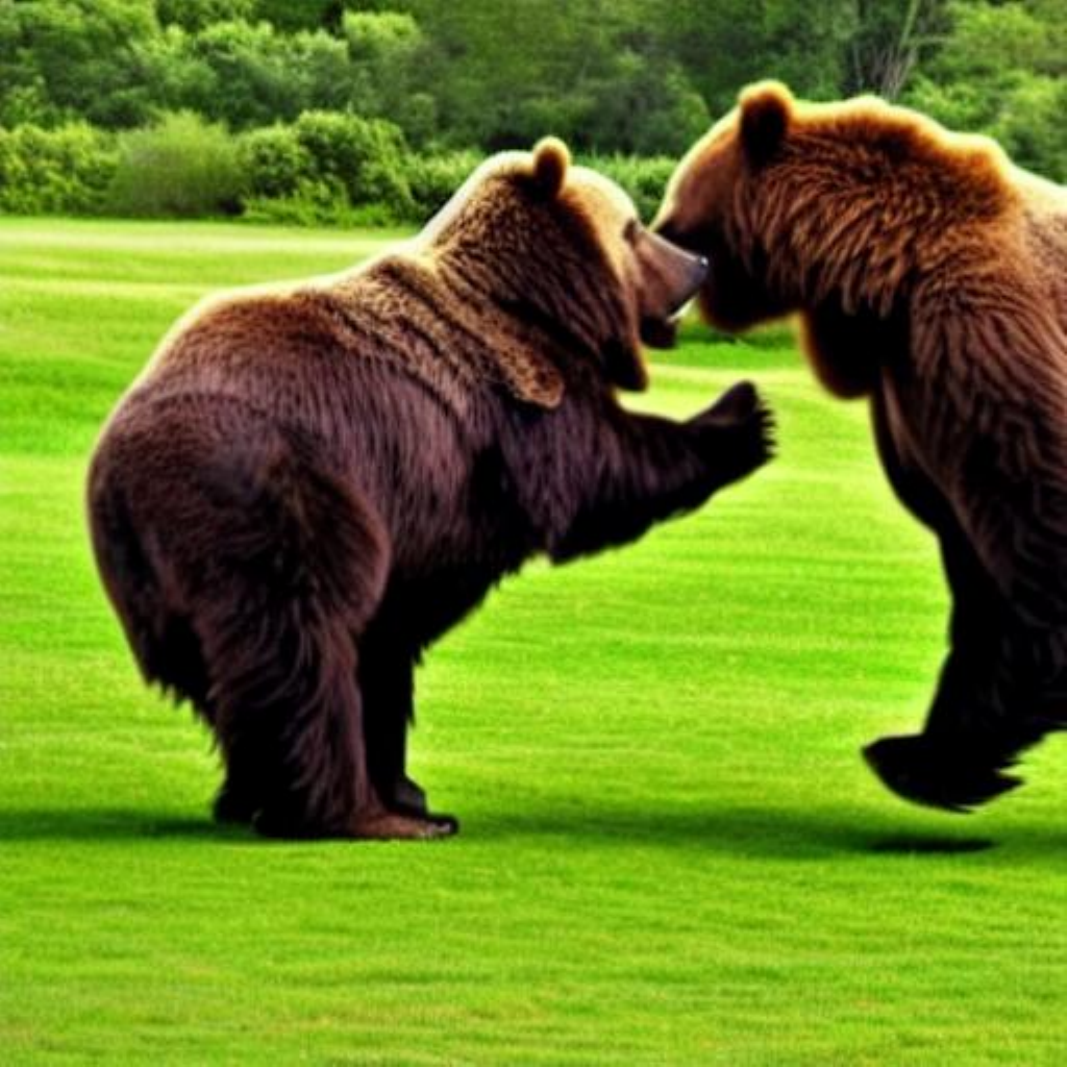}
\hspace{1ex}
\includegraphics[width=0.075\columnwidth]{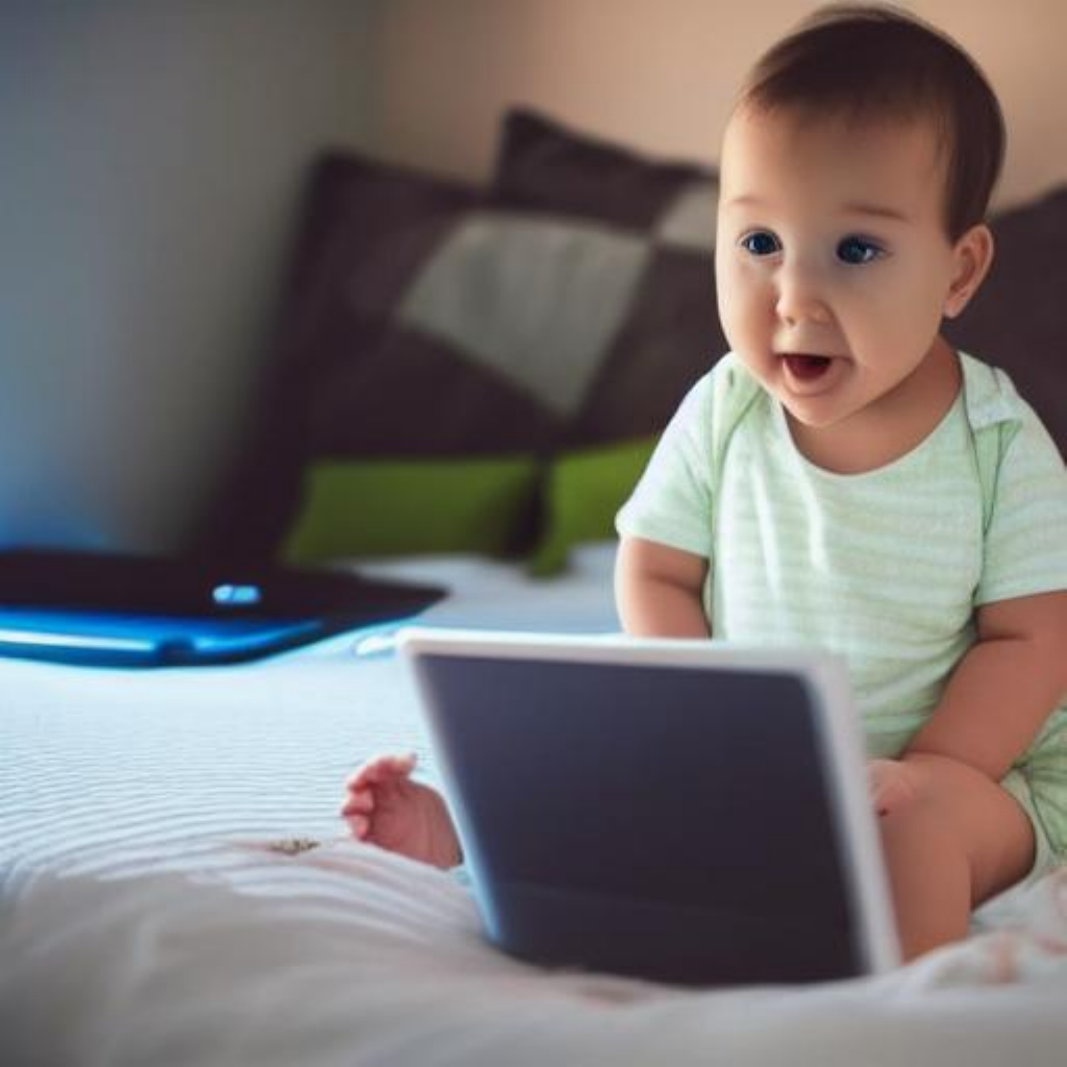}
\includegraphics[width=0.075\columnwidth]{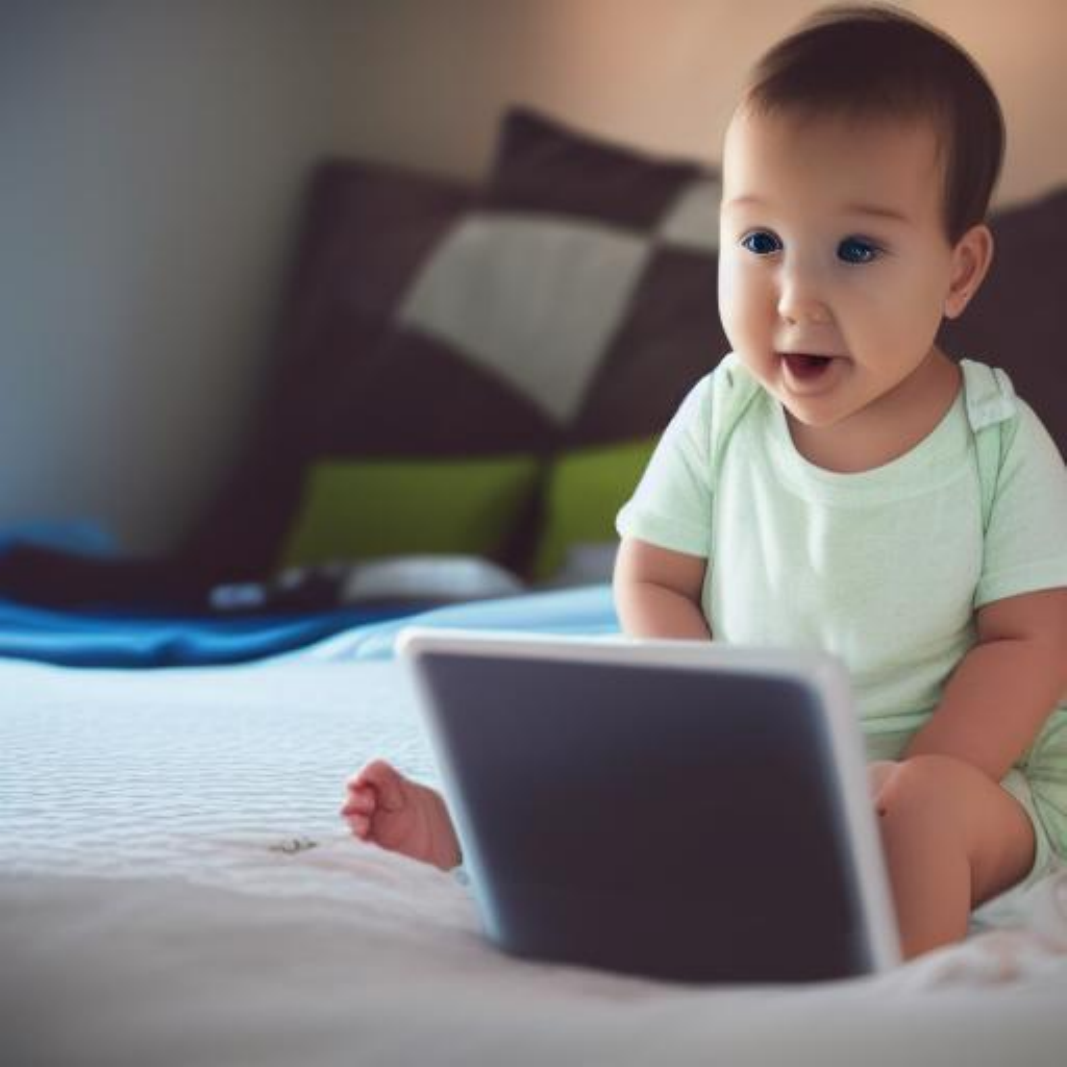}
\includegraphics[width=0.075\columnwidth]{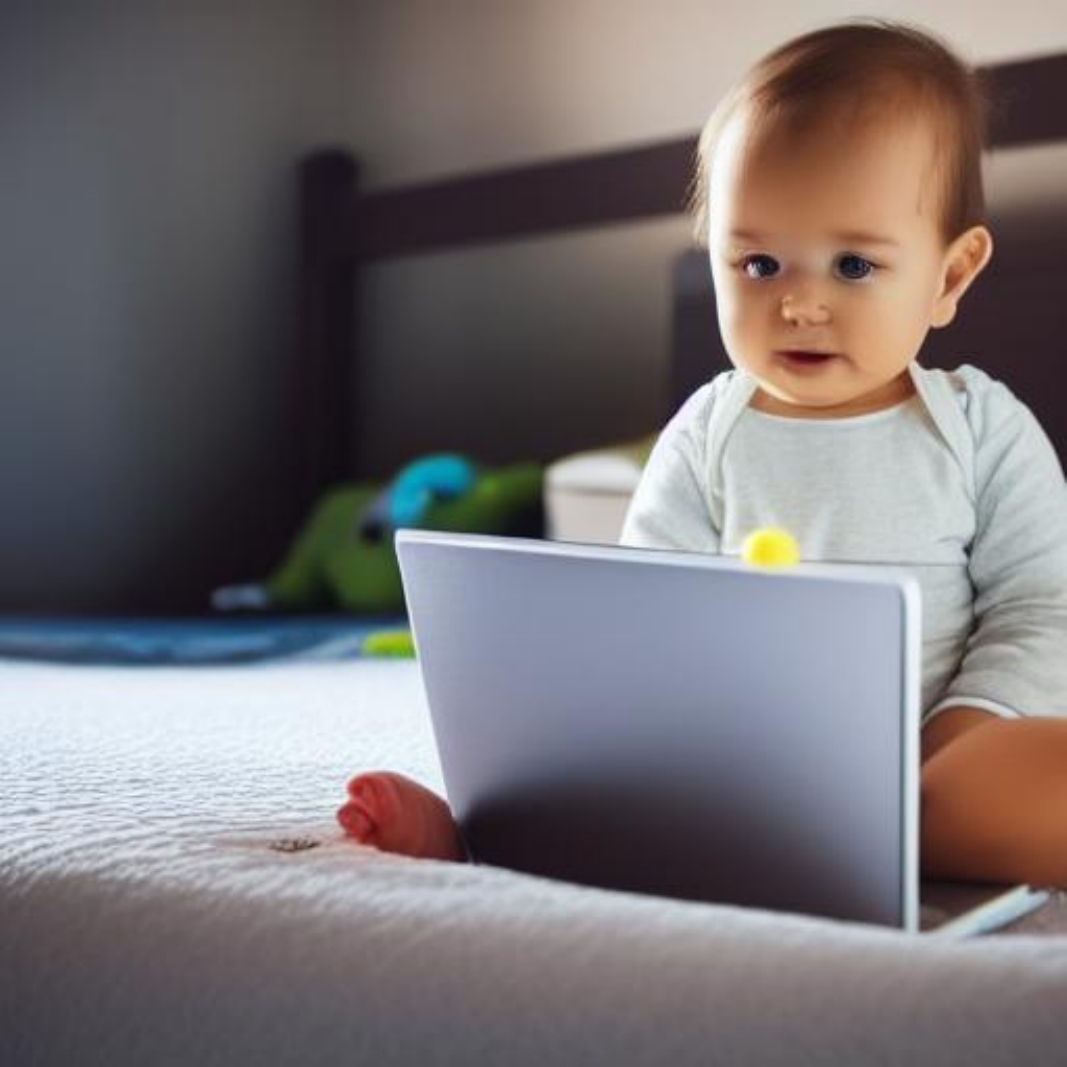}
\includegraphics[width=0.075\columnwidth]{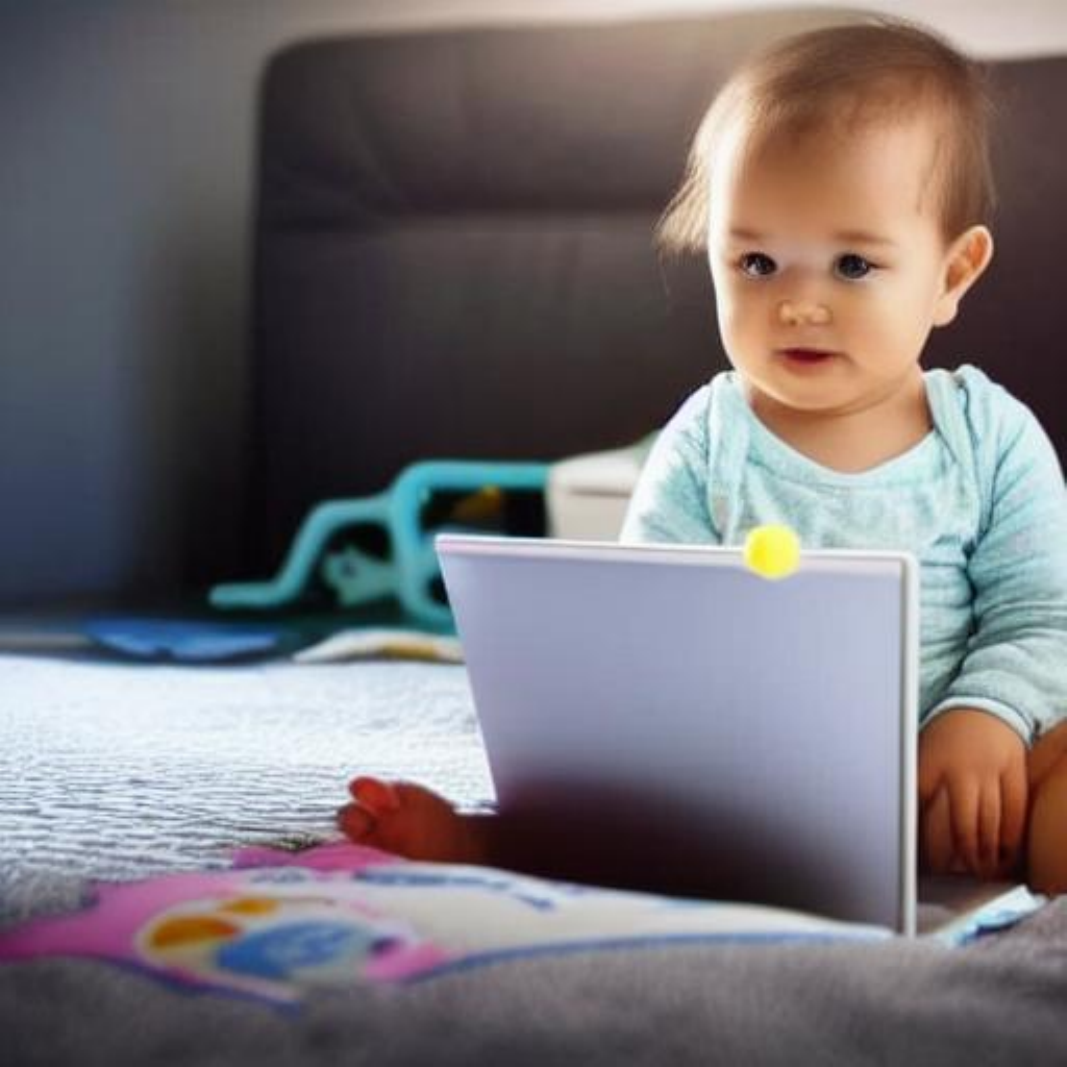}
\hspace{1ex}
\includegraphics[width=0.075\columnwidth]{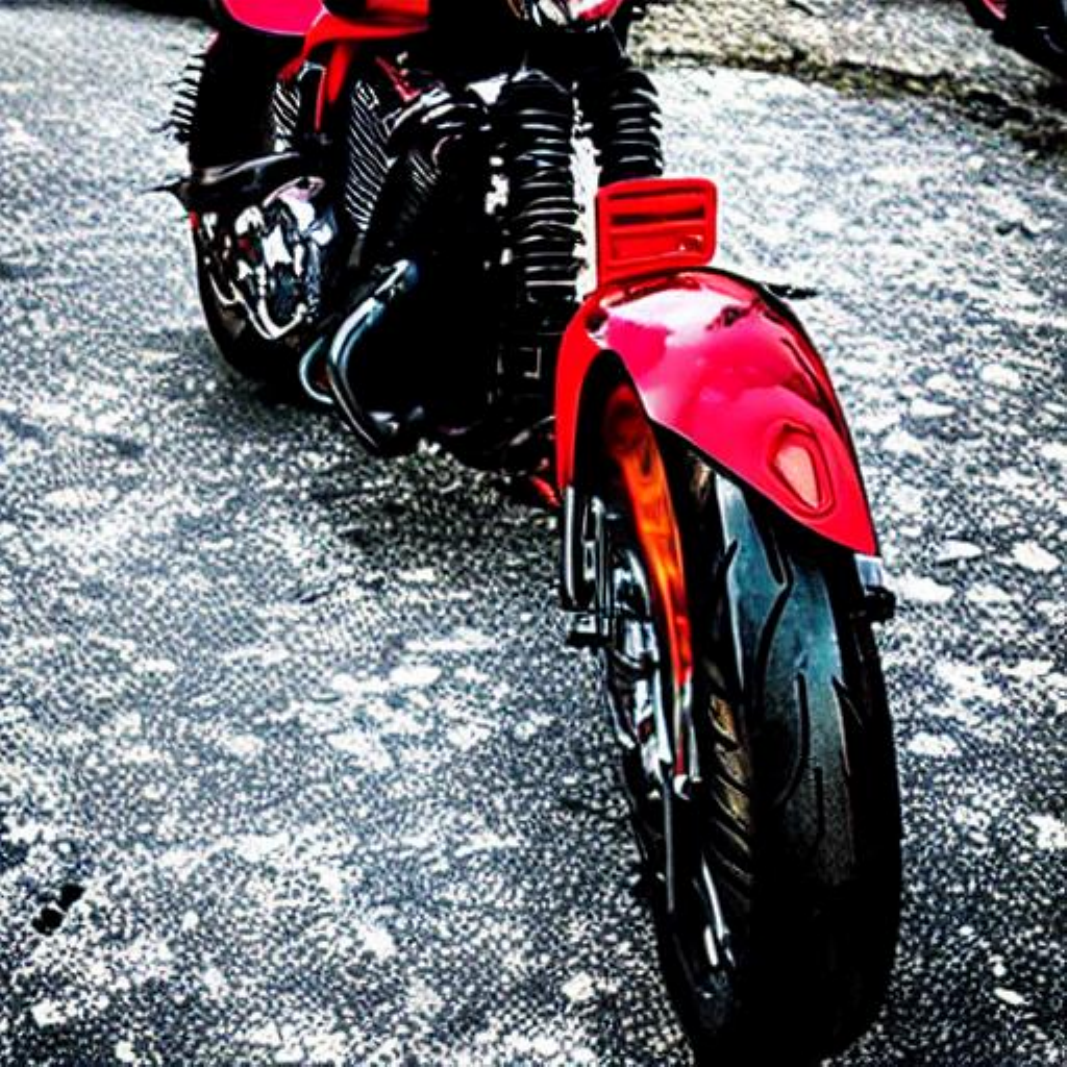}
\includegraphics[width=0.075\columnwidth]{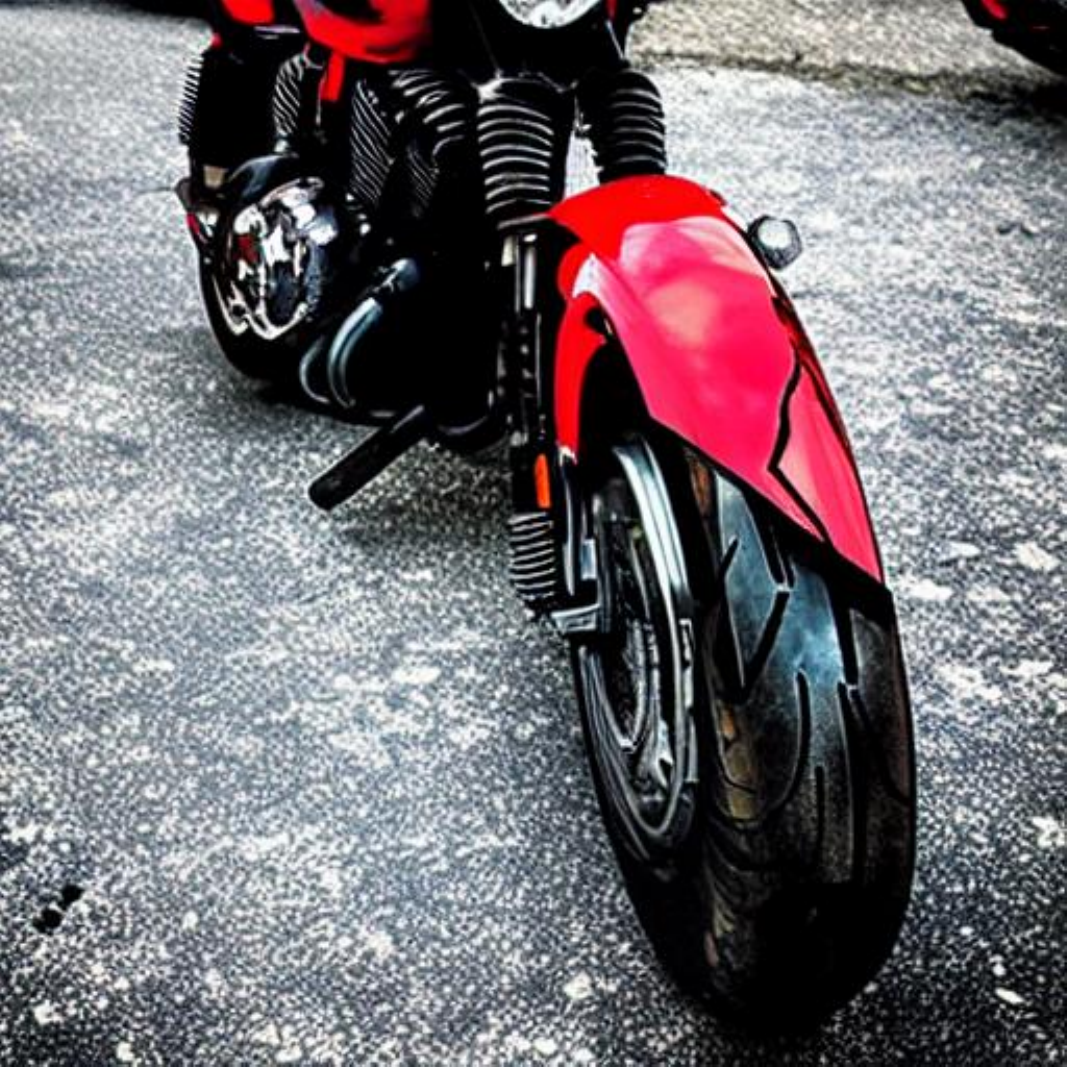}
\includegraphics[width=0.075\columnwidth]{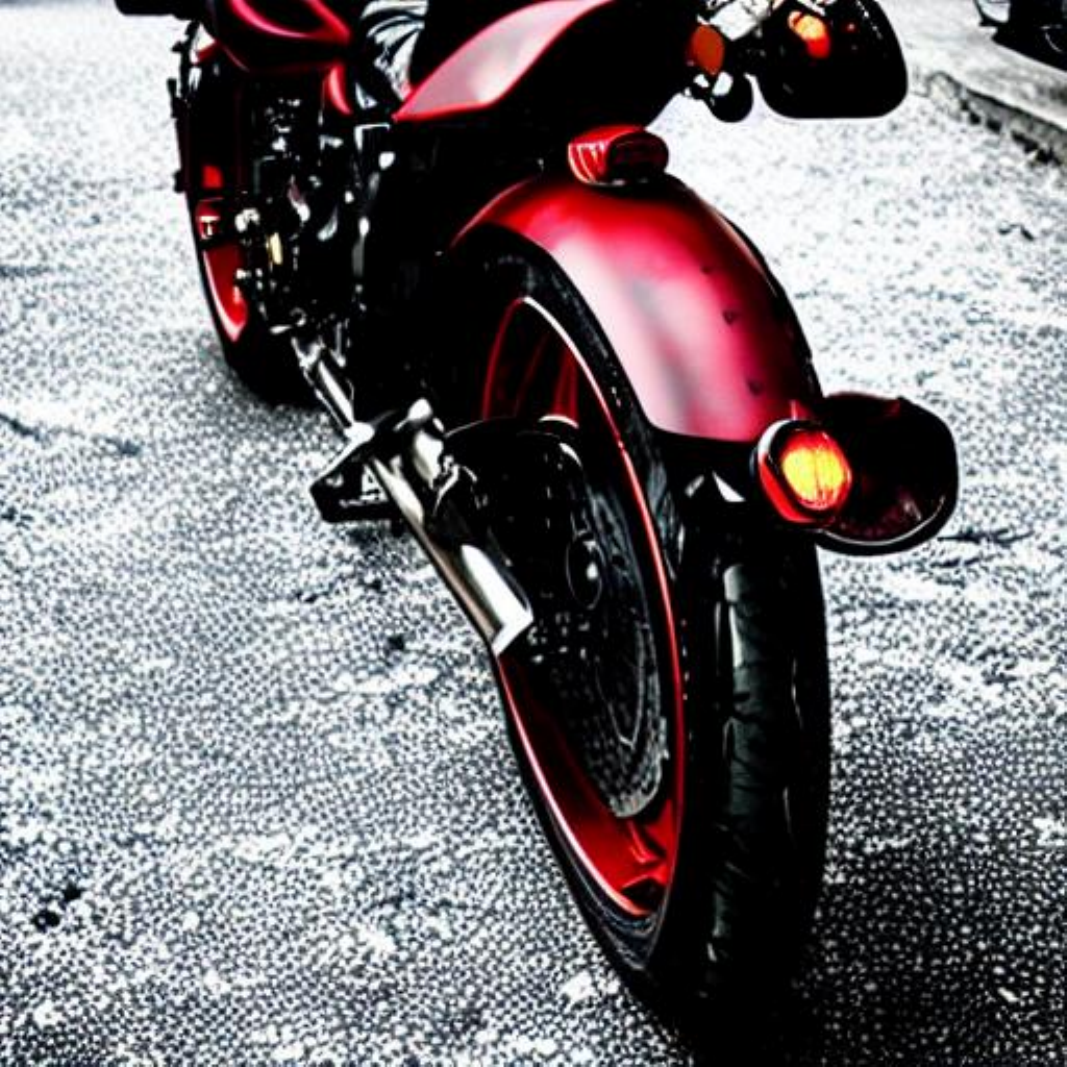}
\includegraphics[width=0.075\columnwidth]{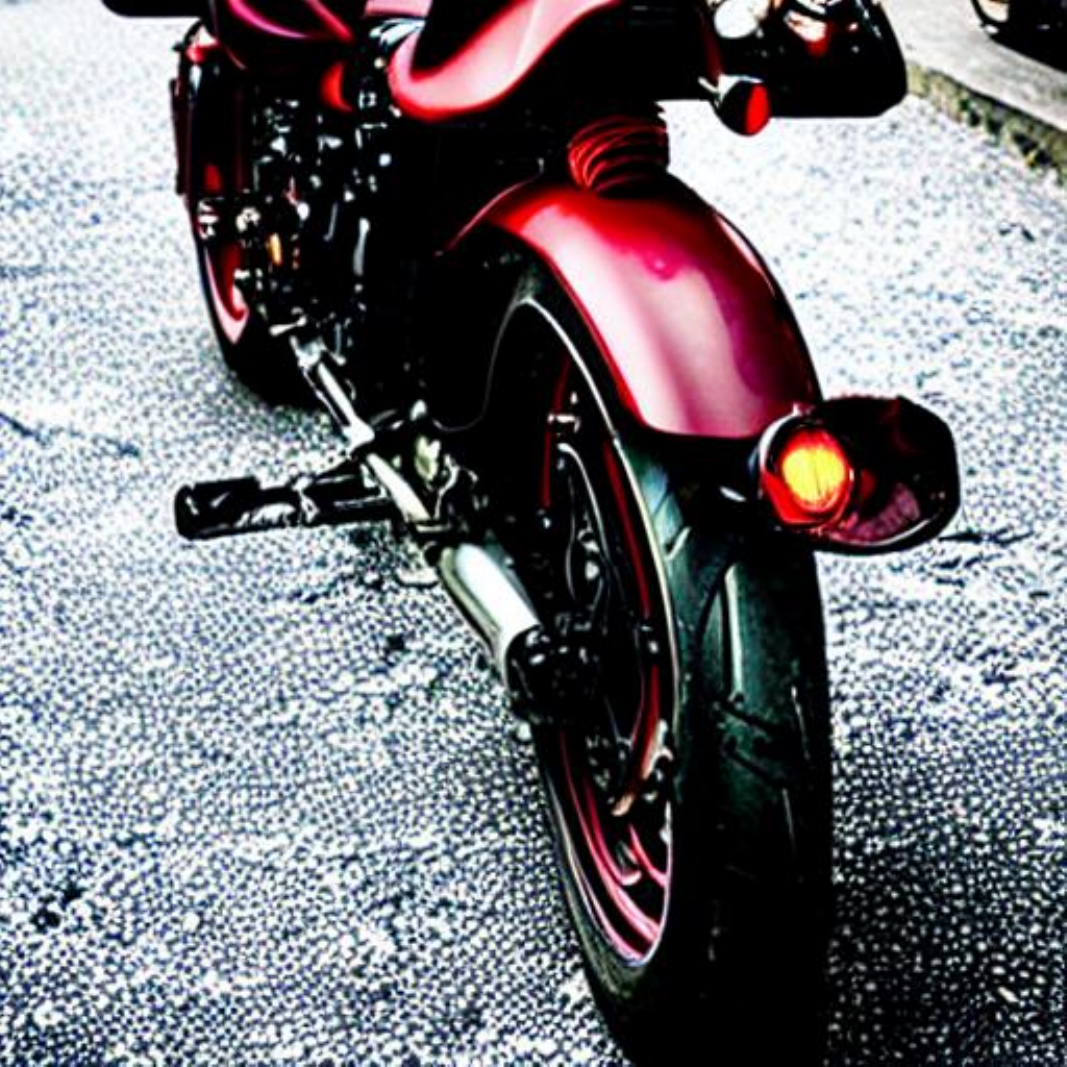}
\\
\vspace{-8pt}
\rule{0.9\textwidth}{0.4pt}
\\
\vspace{3pt}
\includegraphics[width=0.075\columnwidth]{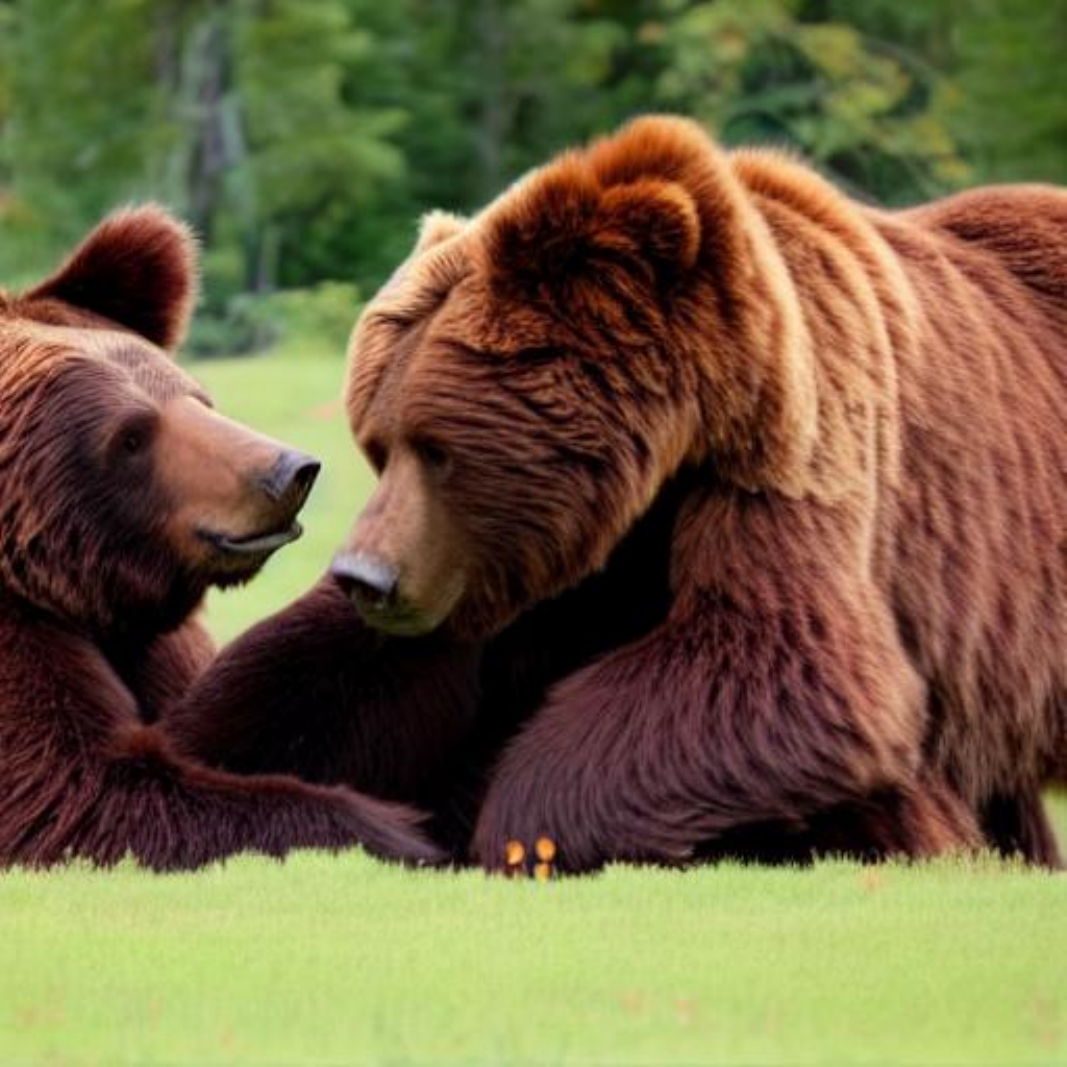}
\includegraphics[width=0.075\columnwidth]{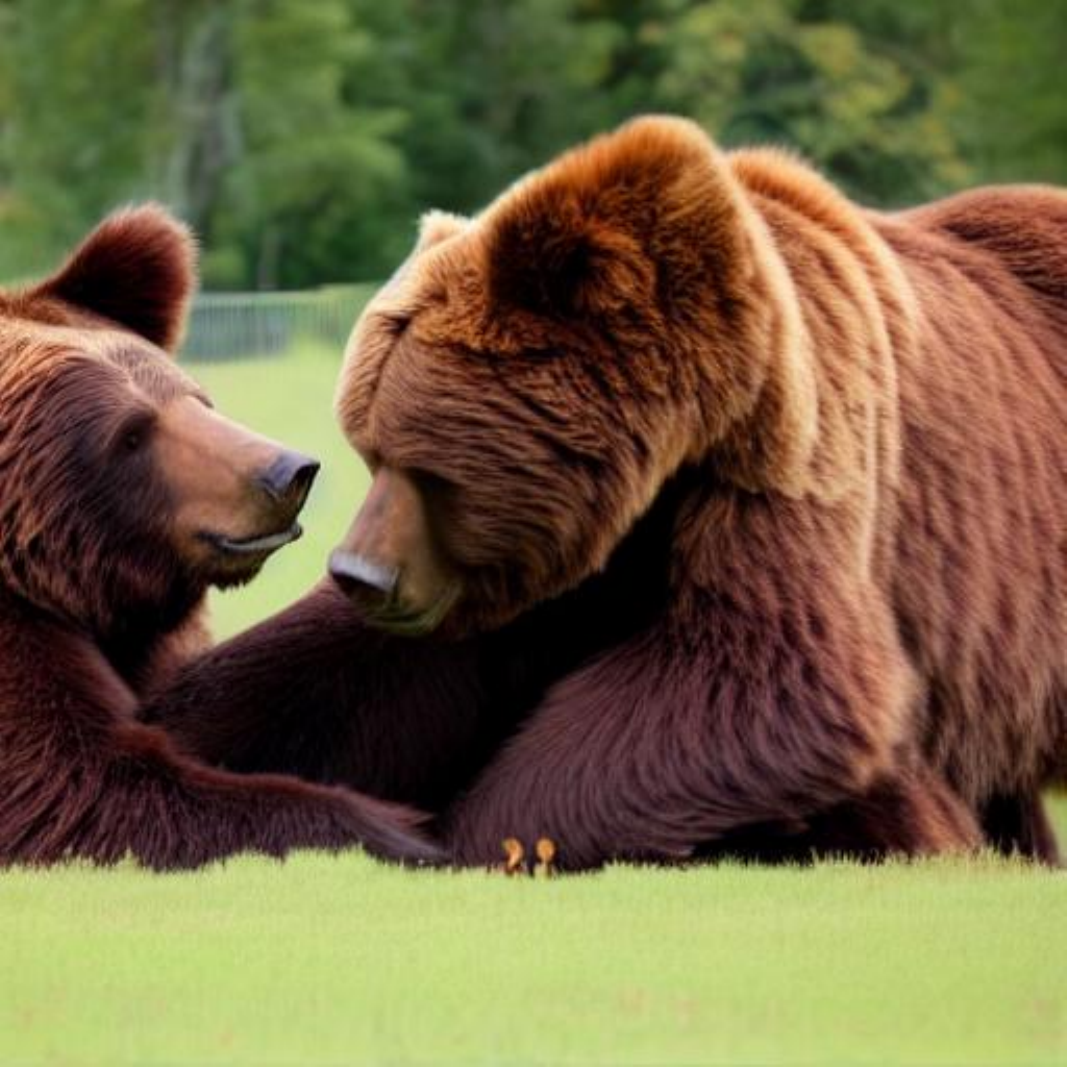}
\includegraphics[width=0.075\columnwidth]{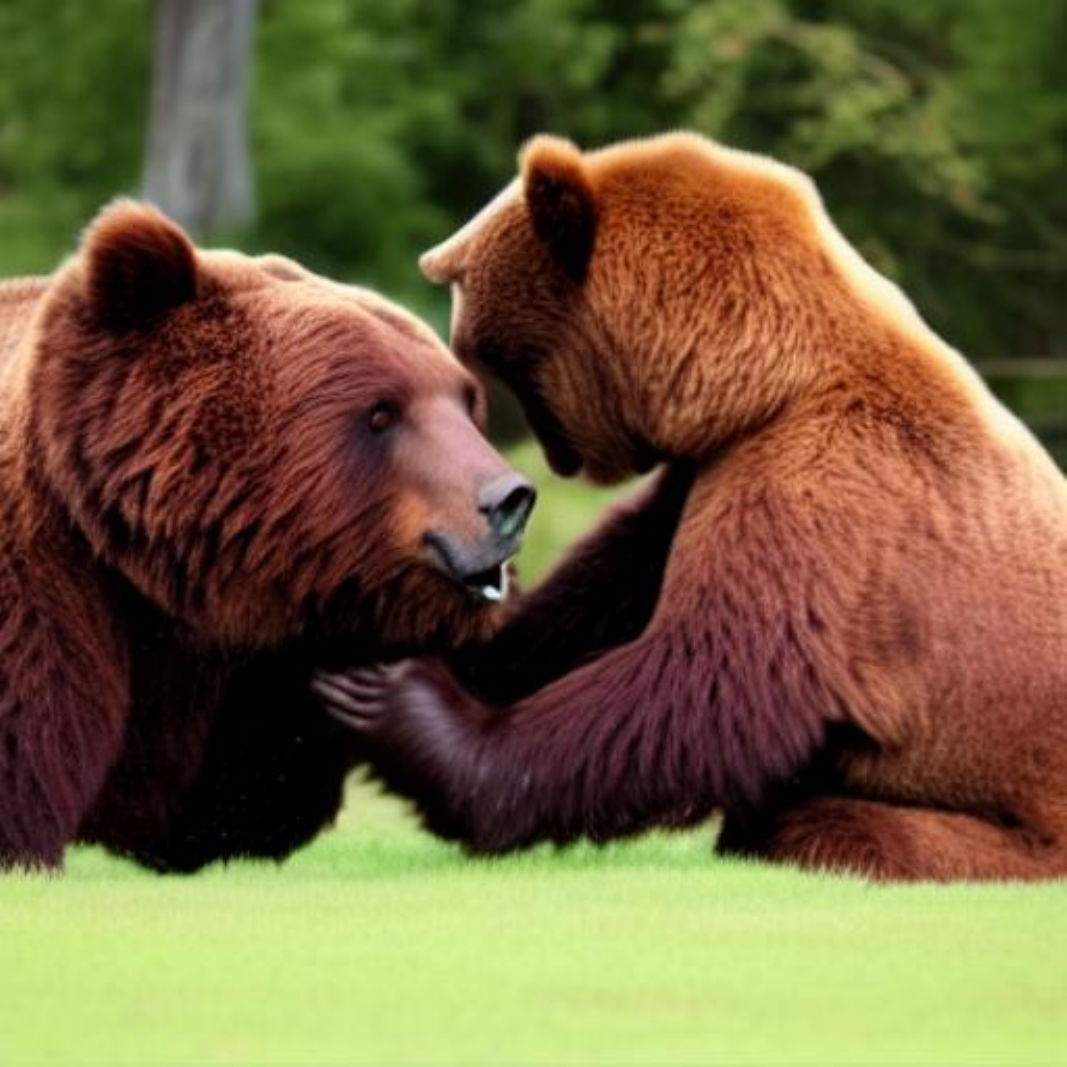}
\includegraphics[width=0.075\columnwidth]{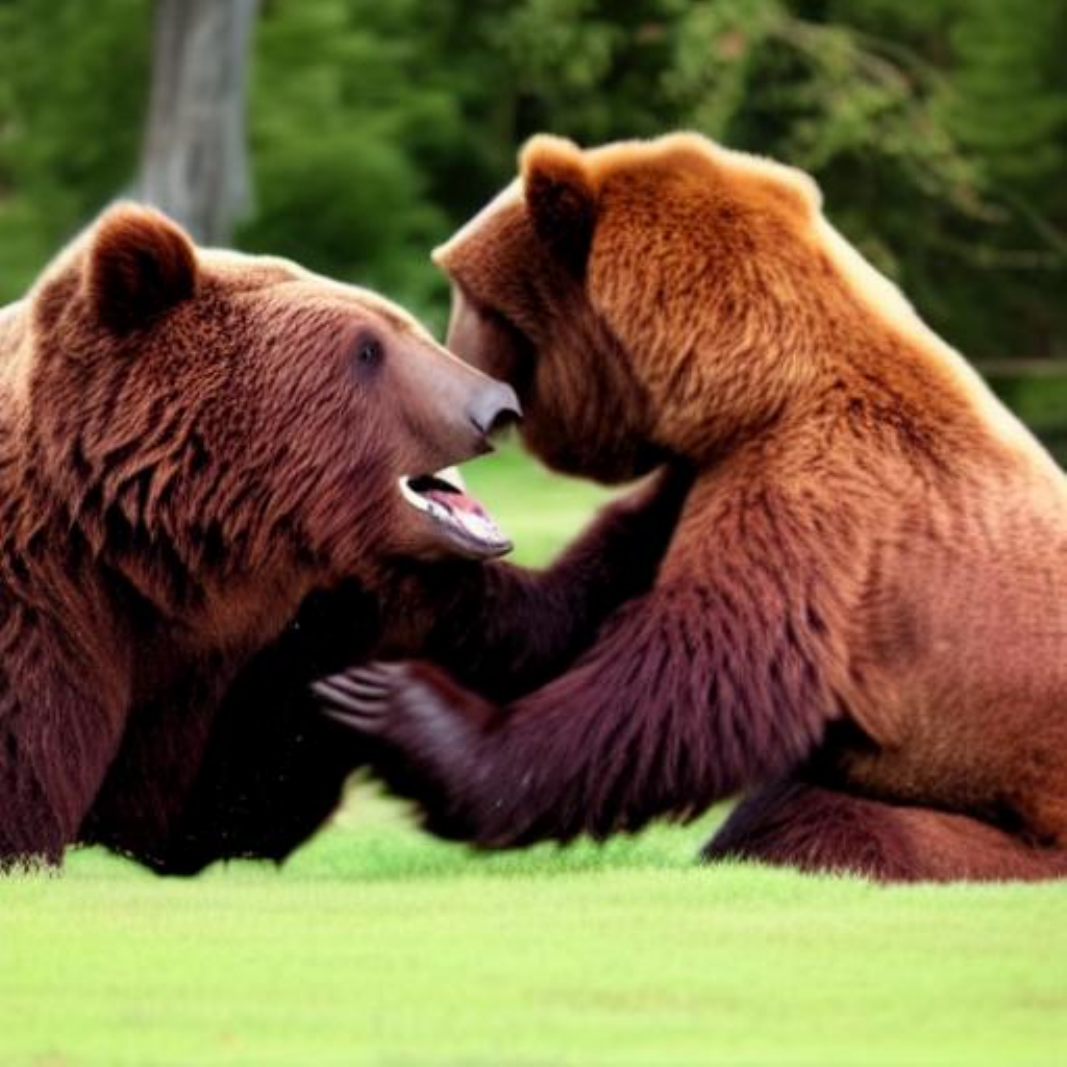}
\hspace{1ex}
\includegraphics[width=0.075\columnwidth]{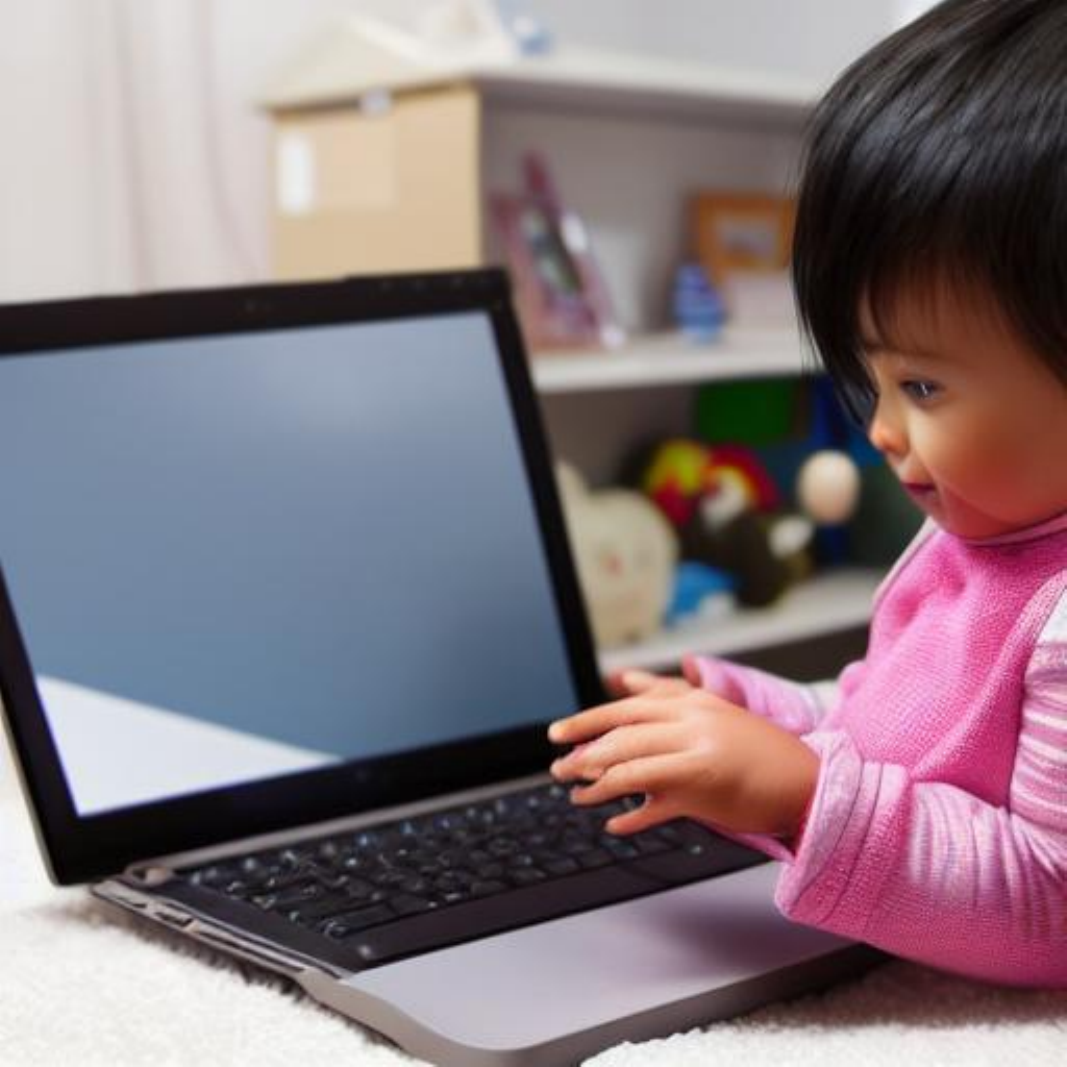}
\includegraphics[width=0.075\columnwidth]{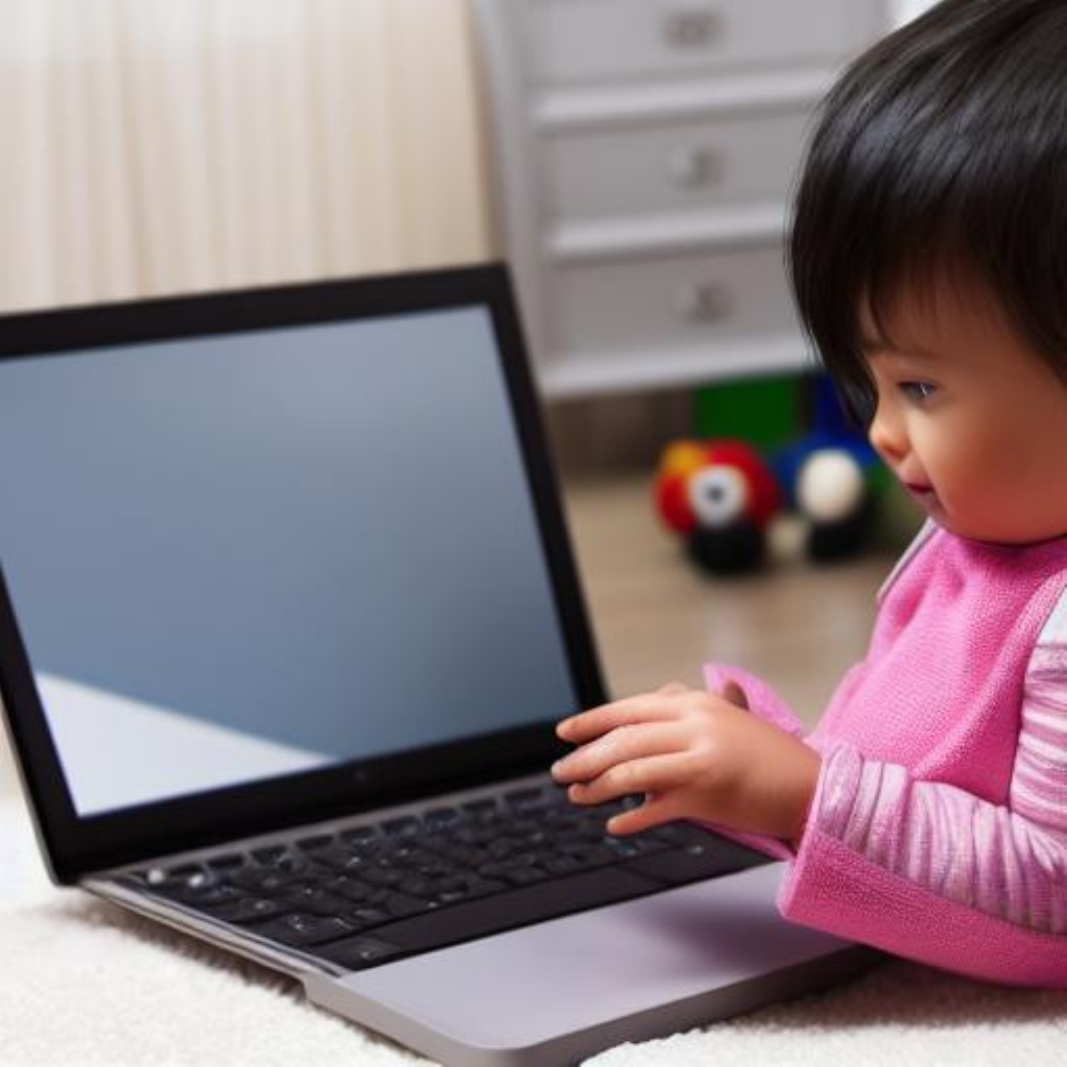}
\includegraphics[width=0.075\columnwidth]{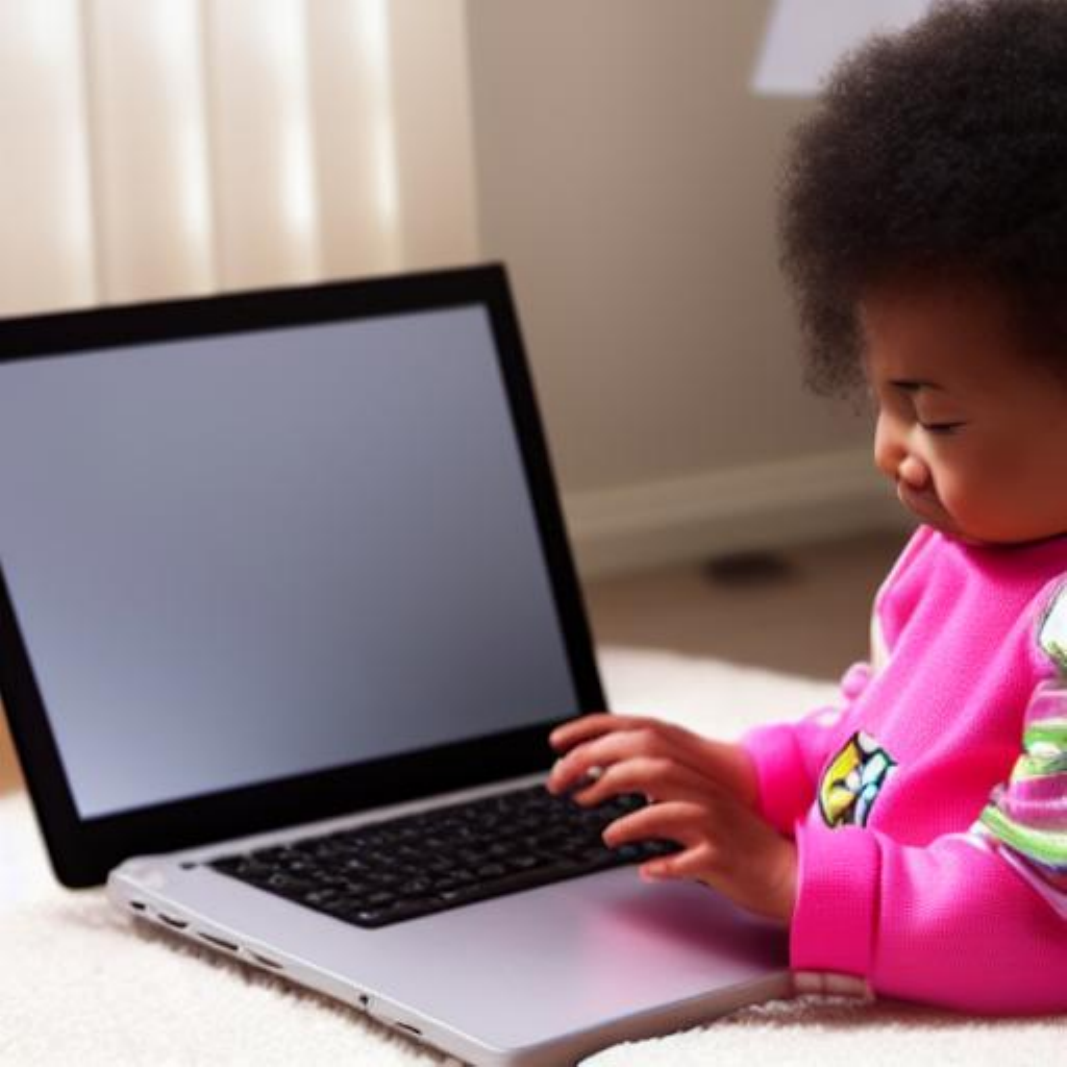}
\includegraphics[width=0.075\columnwidth]{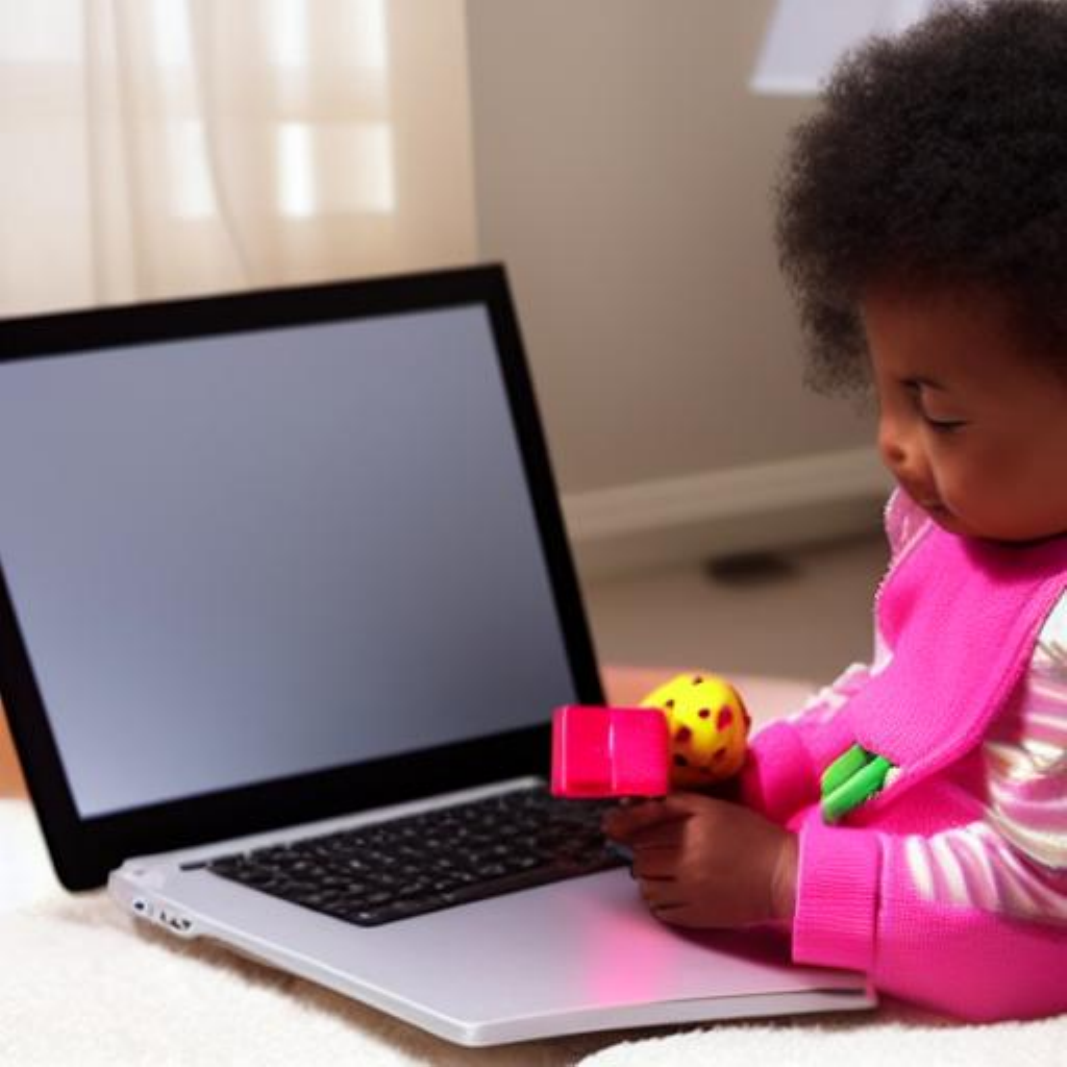}
\hspace{1ex}
\includegraphics[width=0.075\columnwidth]{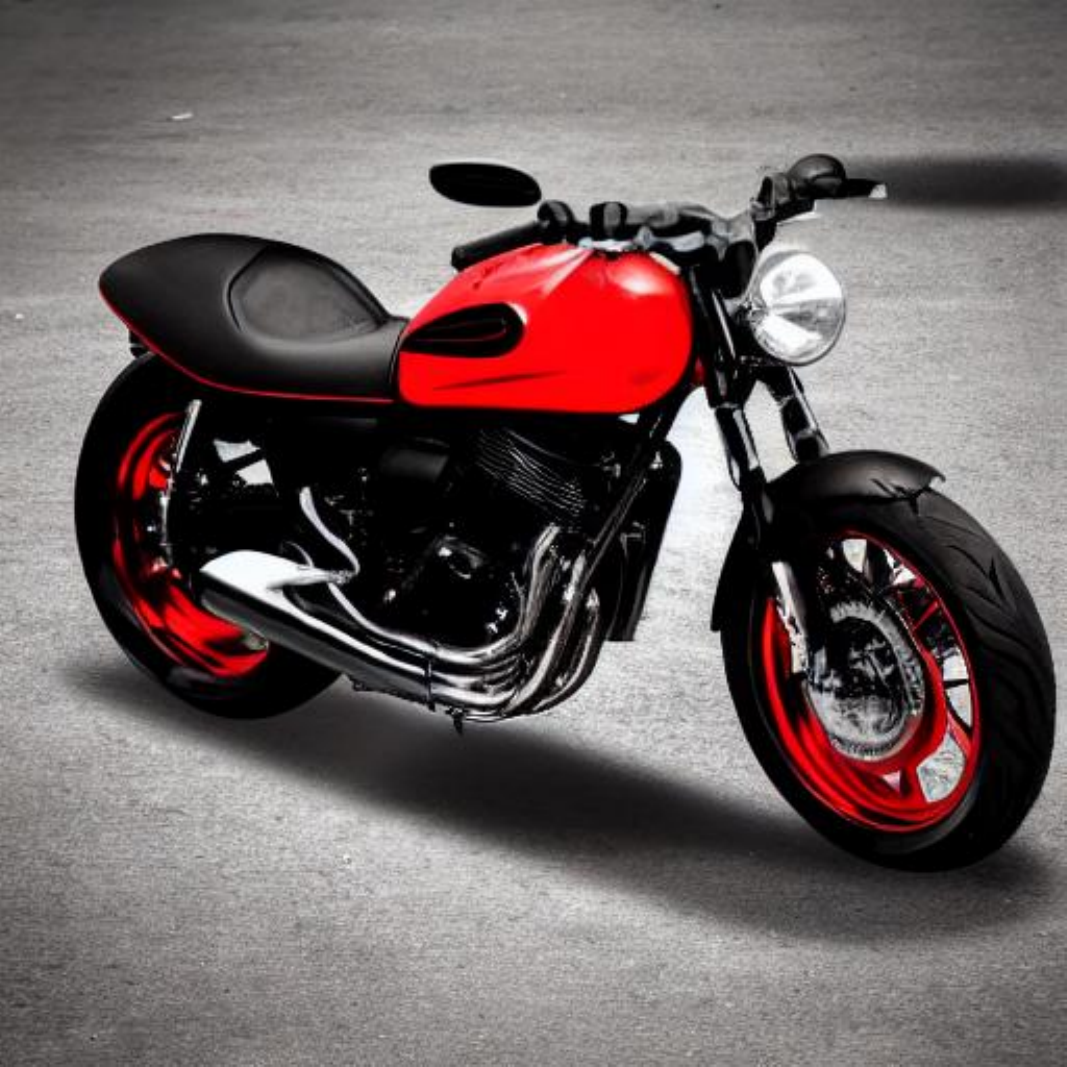}
\includegraphics[width=0.075\columnwidth]{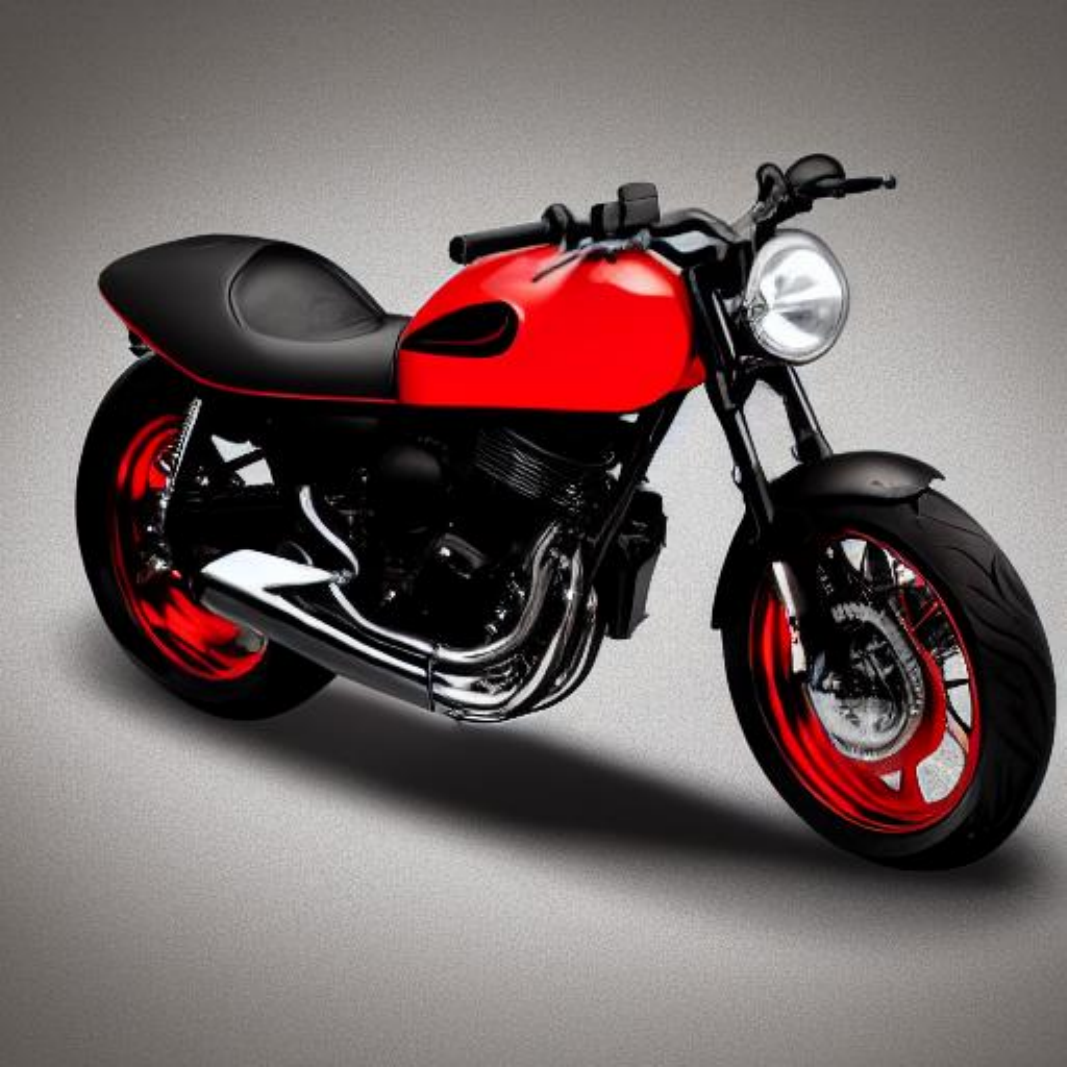}
\includegraphics[width=0.075\columnwidth]{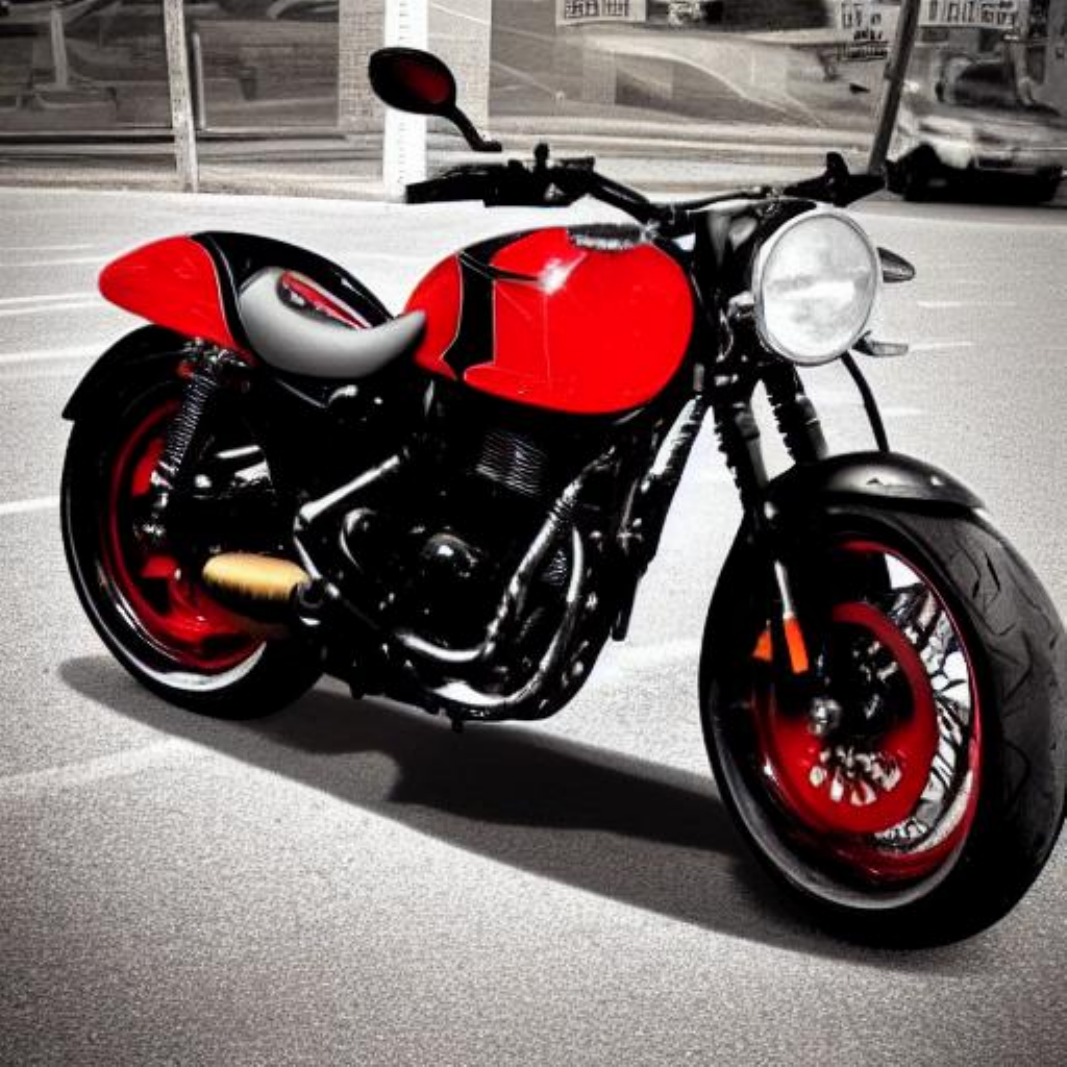}
\includegraphics[width=0.075\columnwidth]{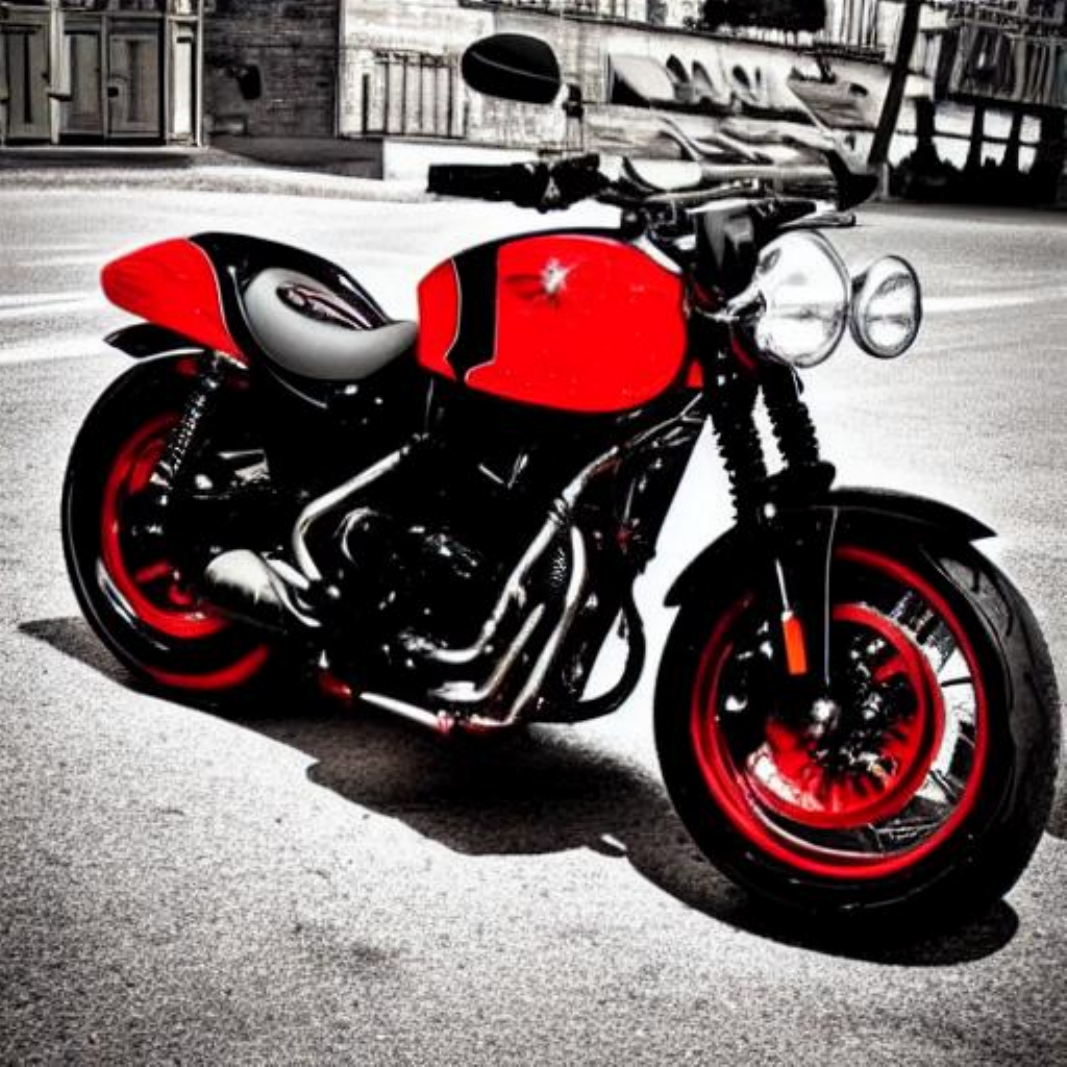}
\caption{\textbf{Stable Diffusion with $20$ (Above), $40$ (Middle), and $80$ (Below) Steps.} 
For each group of four images, from left to right: 
baseline, ablation, and generated images with variance scalings $1.25$ and $1.55$ for $20$ and $40$ steps, or $1.10$ and $1.20$ for $80$ steps. 
%
%
}
    \label{fig:sd-o}
    \end{figure*}

The SD model is a \emph{text-conditional} model in the latent-space. 
Thus, it can produce images with more diverse semantic content than the previous models. 
We analyze the case of $20$, $40$ and $80$ steps, examples given in Fig.~\ref{fig:sd-o} and Figs.~\ref{fig:sd-20}, \ref{fig:sd-40}, and \ref{fig:sd-80} from Appendix~\ref{app:exp}. The prompts for the images are randomly extracted from the COCO dataset~\citep{lin-2014-COCO}.

For a fixed prompt, S2EP is able to introduce substantial semantic change (relative to the baseline) across all number of steps, and---as in previous models---the frequency and intensity of such change increase as the number of steps do. 
%
%
Contrast and brightness  
can improve with S2EP; however, it is often hard to 
notice 
any clear quality difference between
the baseline and S2EP. 
Unlike all previous models, 
%
there does not seem to be a clear overall trend on the type of changes S2EP does, except perhaps for a tendency towards more vivid colors as the number of steps increases (similar to DiT).
%
%
Moreover, baseline images very often already have oversaturated colors that S2EP also translates to 
the sampled image.
%
%
%
As in previous models, 
very large values of variance scaling can negatively affect image quality.
%
%
%

In conclusion, both large semantic changes and the lack of a noticeable trend of image enhancements make it difficult to qualitatively assess the effect of SE2P on SD.\footnote{We are only concerned with 
image quality, 
and not with the
correspondence between image \emph{content} and the \emph{prompt}.}
%
%

\subsection{Ablation}
\label{sec:ablation}
%
Our \emph{ablation} 
directly integrates the latent at time $t_k$ with the one at $t_k+1$, i.e., it is 
%
the ``naive'' integration method described in Section~\ref{sec:algorithm}. 
%
%
Sampled images from our ablation are shown in every figure referenced in Section~\ref{sec:qualitative}. 
For DDPM and LD, the ablation 
%
%
mostly smooths out the baseline image, without improving contrast or brightness, and eliminating sharp edges that are important for image quality. 
%
%
%
For DiT, the ablation degrades the baseline 
by 
dimming and muting its colors, 
as well as blurring distinctive details or features corresponding to the image class subject. For SD, the ablation can also blur image details and textures, but very often produces 
semantic changes that make it difficult to assess how much degradation is actually introduces overall.  
%

In conclusion, the direct integration of the two consecutive steps does not help improve sample quality.

\section{Quantitative Study}
\label{sec:quantitative}
Having done a qualitative analysis of the generated images, we now quantify their quality.

\subsection{Automated Evaluation}
\label{sec:metric}
\begin{figure*}[t!]
    \centering
\includegraphics[width=0.075\columnwidth]{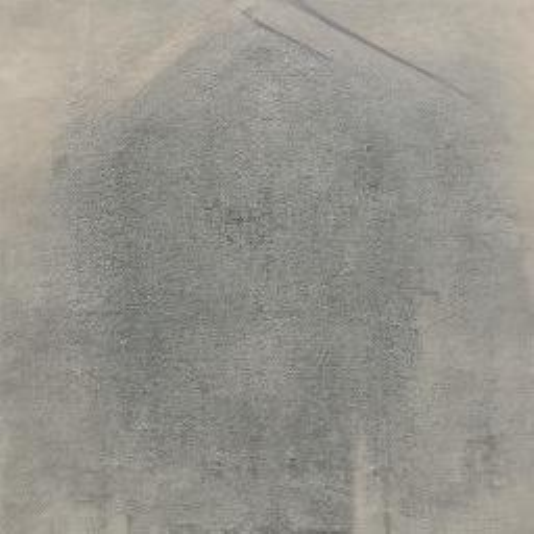}
\includegraphics[width=0.075\columnwidth]{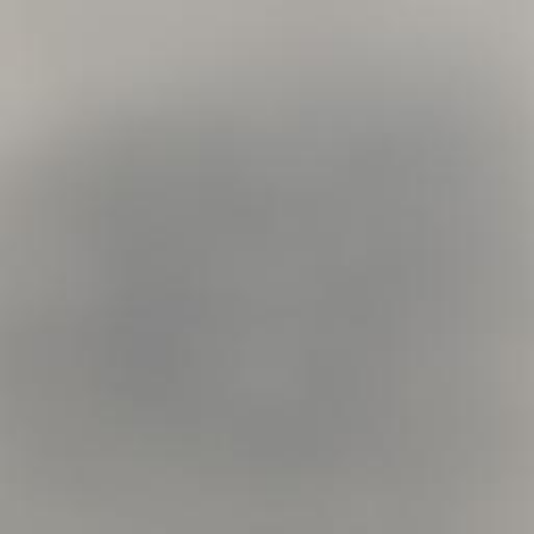}
\includegraphics[width=0.075\columnwidth]{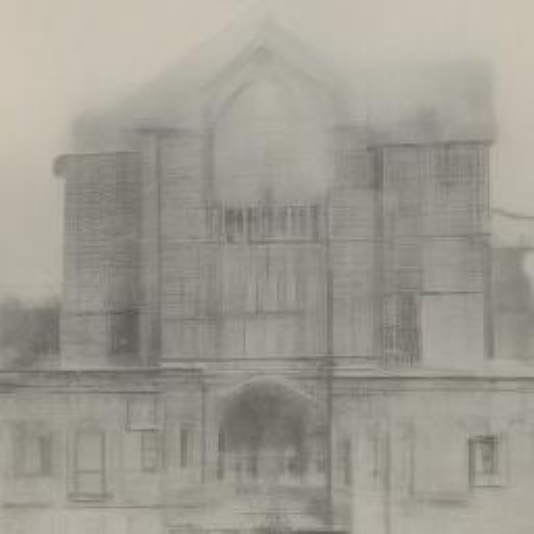}
\includegraphics[width=0.075\columnwidth]{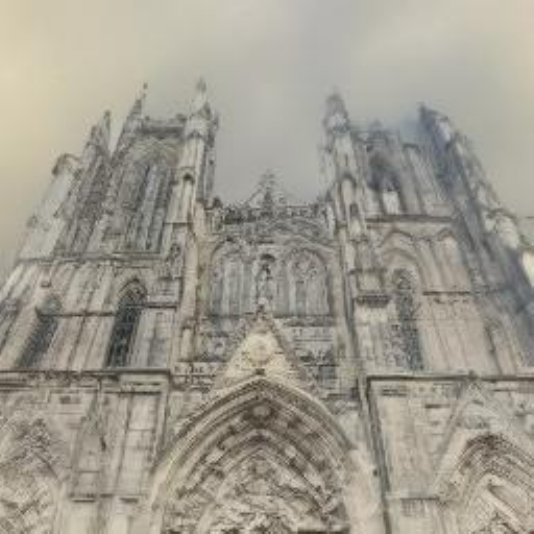}
\hspace{1ex}
\includegraphics[width=0.075\columnwidth]{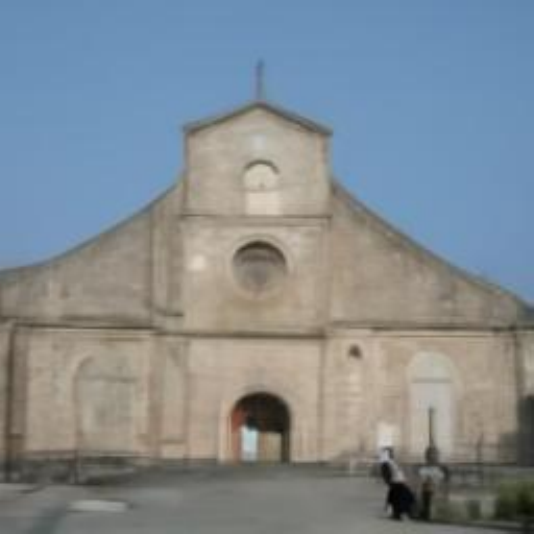}
\includegraphics[width=0.075\columnwidth]{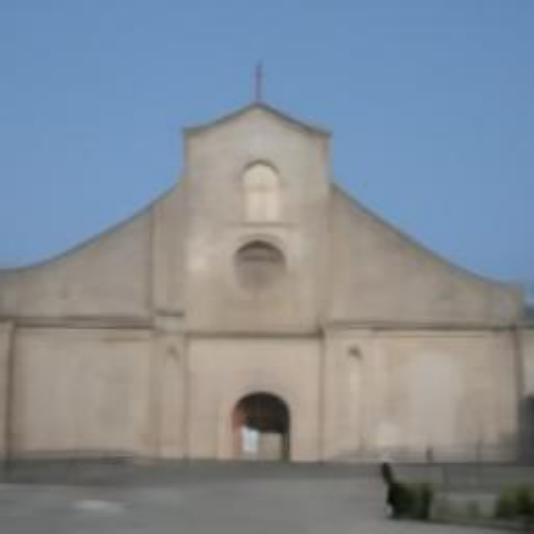}
\includegraphics[width=0.075\columnwidth]{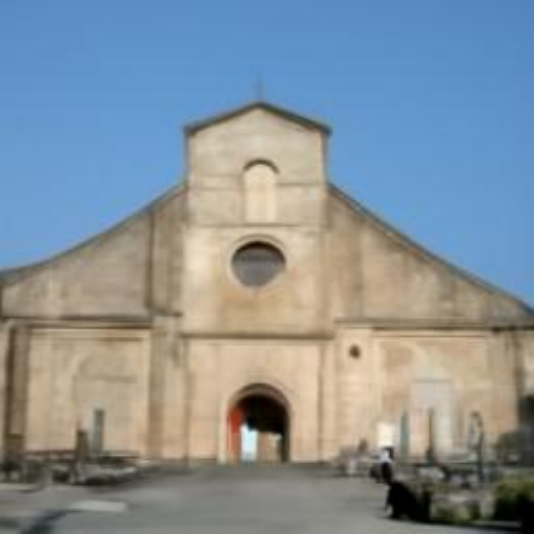}
\includegraphics[width=0.075\columnwidth]{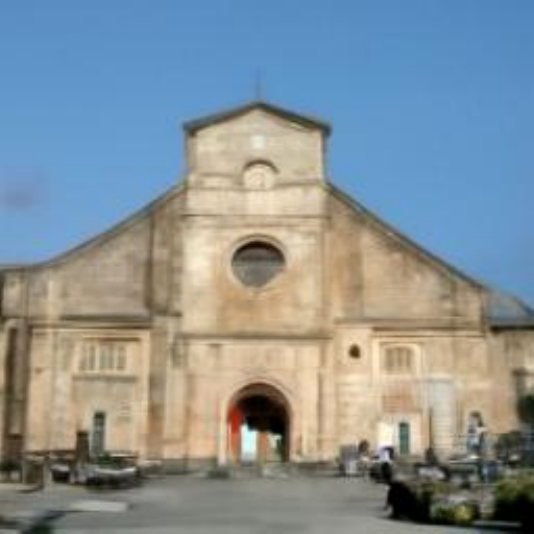}
\hspace{1ex}
\includegraphics[width=0.075\columnwidth]{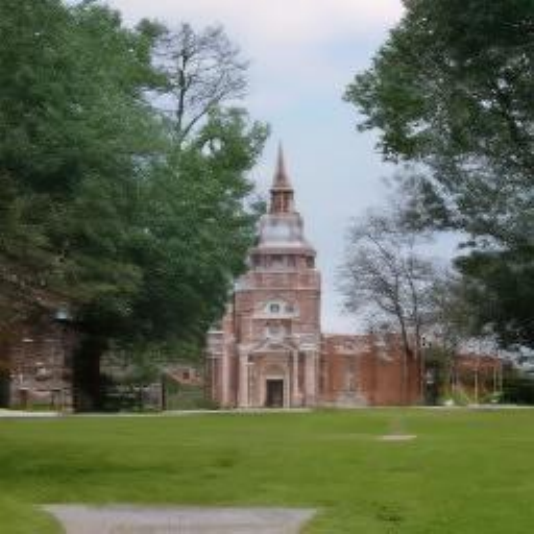}
\includegraphics[width=0.075\columnwidth]{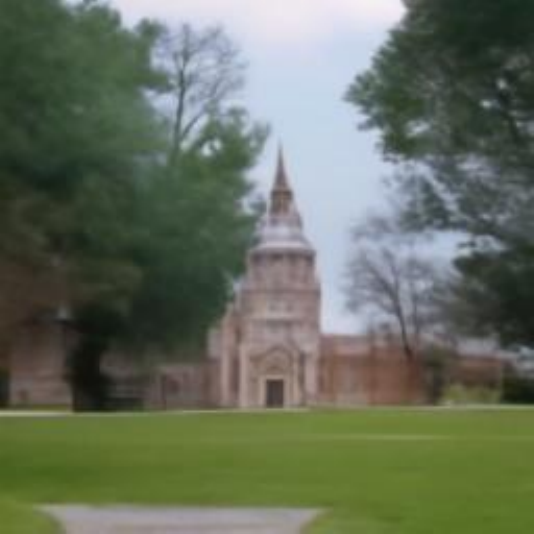}
\includegraphics[width=0.075\columnwidth]{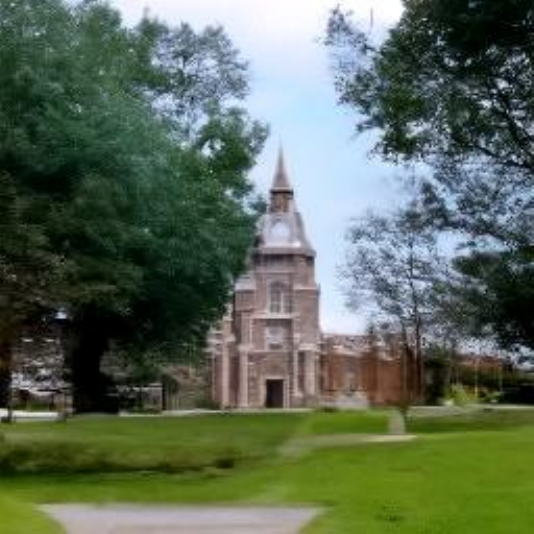}
\includegraphics[width=0.075\columnwidth]{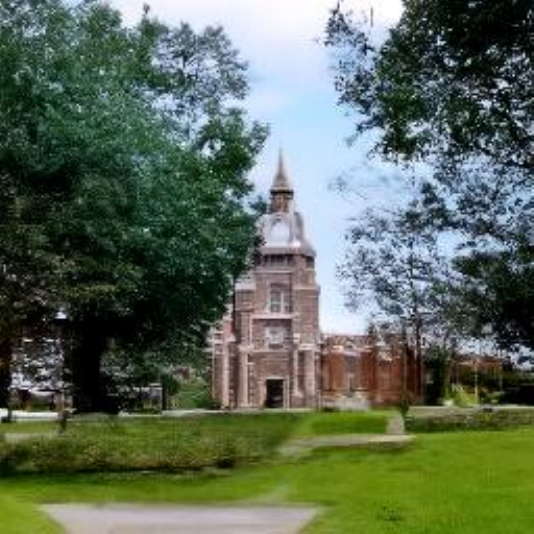}
\\
\includegraphics[width=0.075\columnwidth]{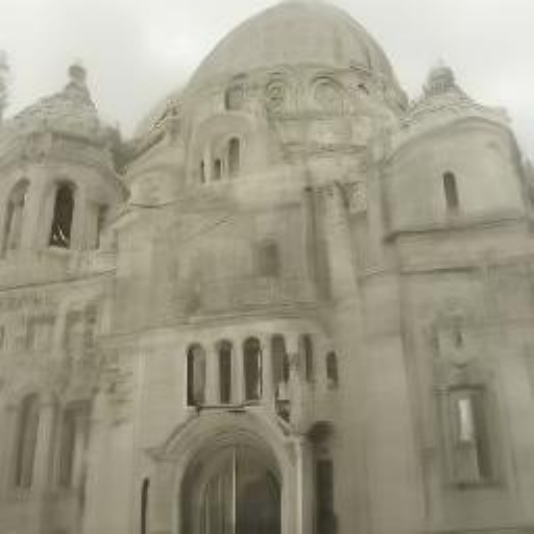}
\includegraphics[width=0.075\columnwidth]{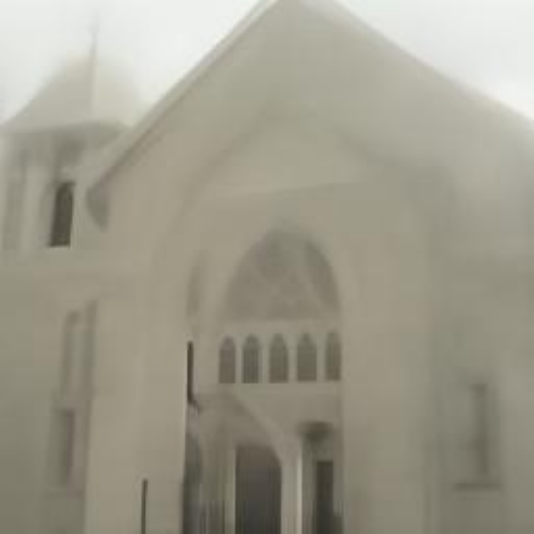}
\includegraphics[width=0.075\columnwidth]{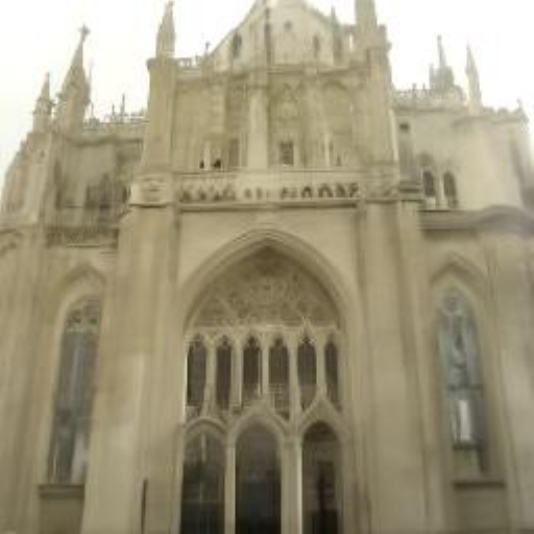}
\includegraphics[width=0.075\columnwidth]{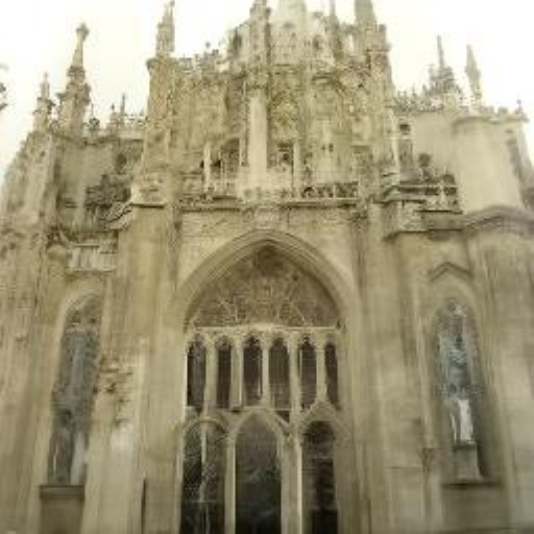}
\hspace{1ex}
\includegraphics[width=0.075\columnwidth]{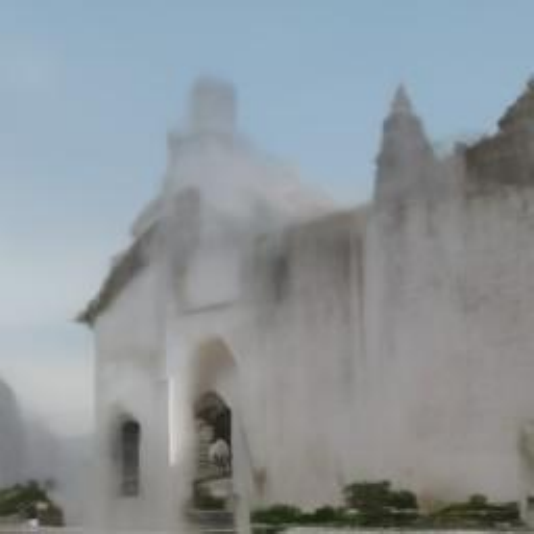}
\includegraphics[width=0.075\columnwidth]{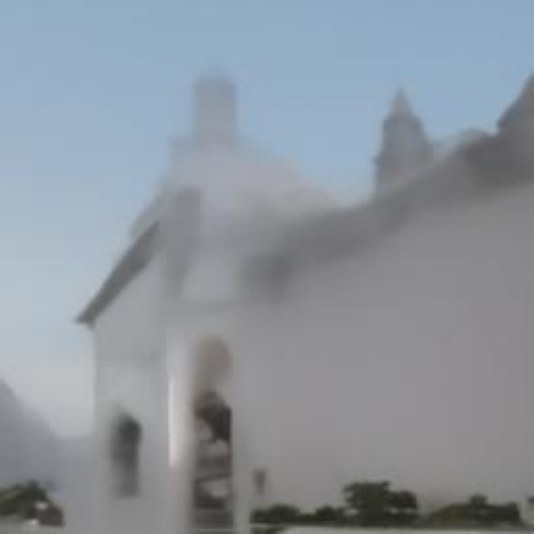}
\includegraphics[width=0.075\columnwidth]{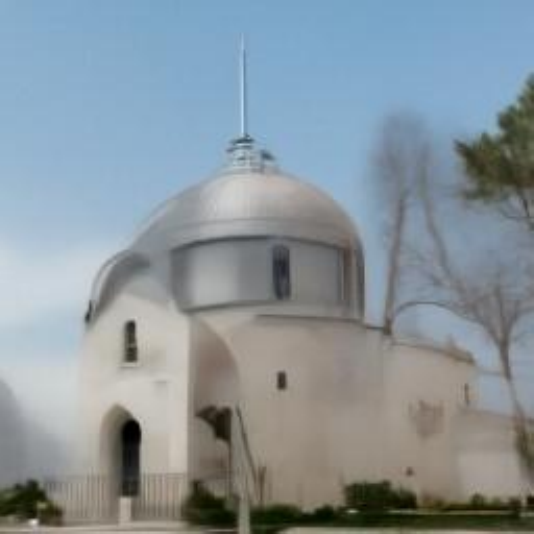}
\includegraphics[width=0.075\columnwidth]{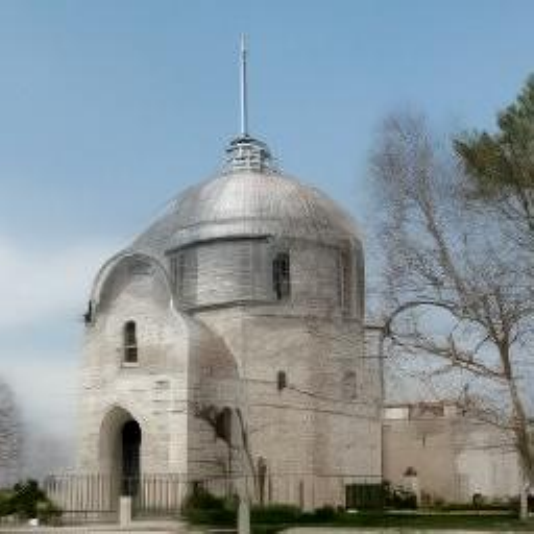}
\hspace{1ex}
\includegraphics[width=0.075\columnwidth]{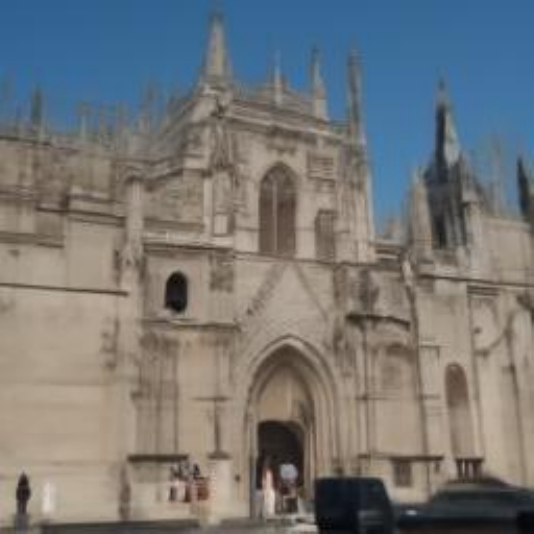}
\includegraphics[width=0.075\columnwidth]{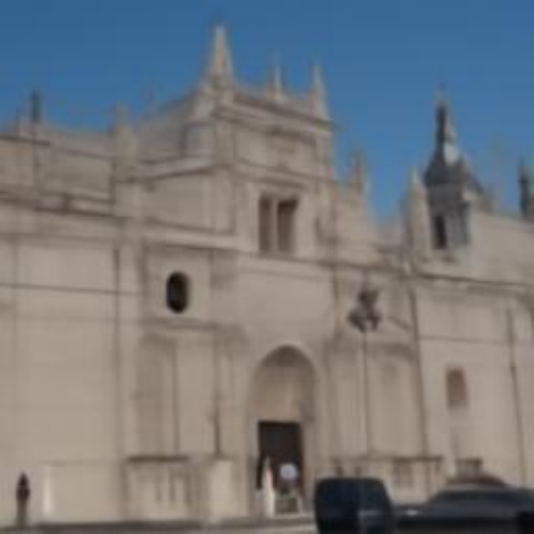}
\includegraphics[width=0.075\columnwidth]{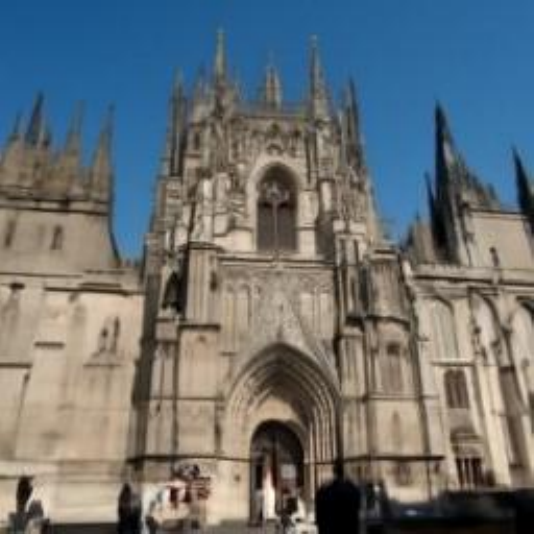}
\includegraphics[width=0.075\columnwidth]{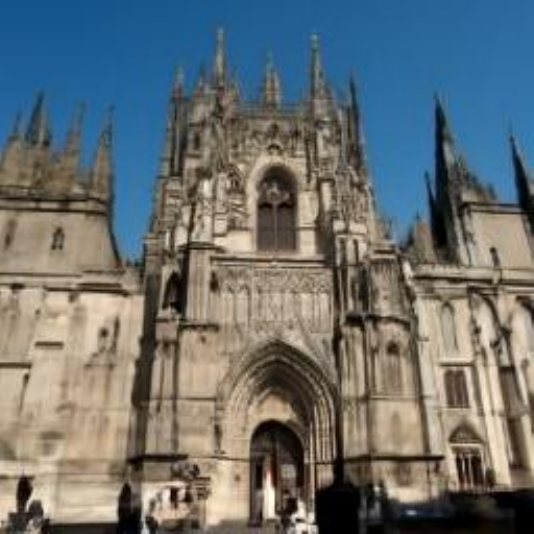}
\\
\vspace{-8pt}
\rule{0.9\textwidth}{0.4pt}
\\
\vspace{3pt}
\includegraphics[width=0.075\columnwidth]{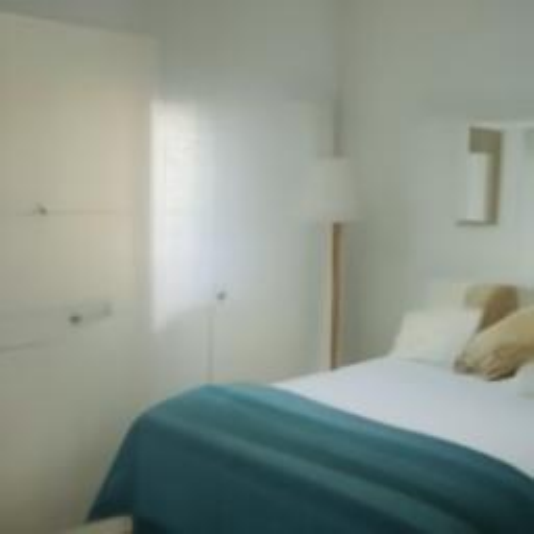}
\includegraphics[width=0.075\columnwidth]{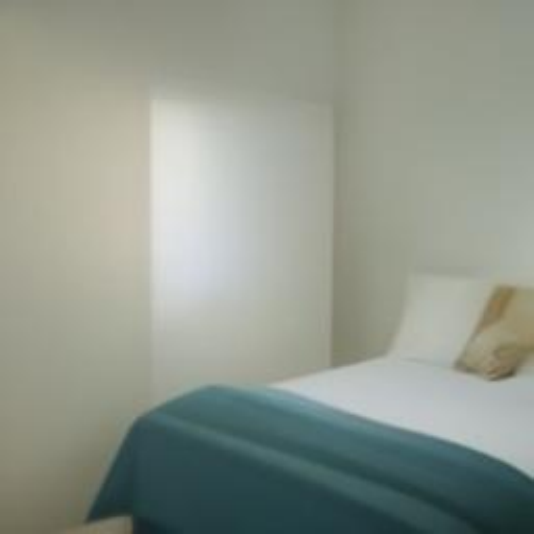}
\includegraphics[width=0.075\columnwidth]{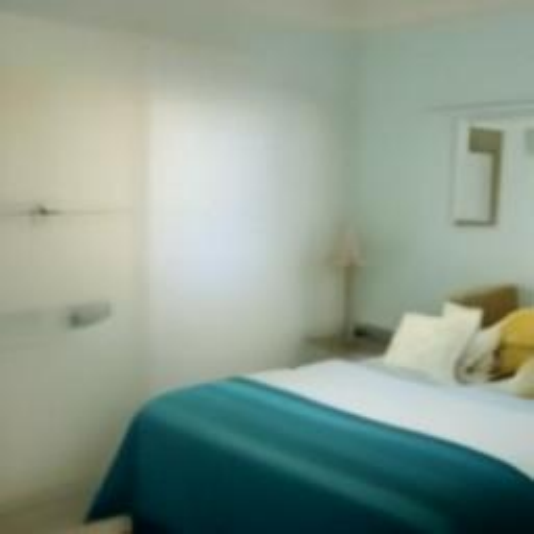}
\includegraphics[width=0.075\columnwidth]{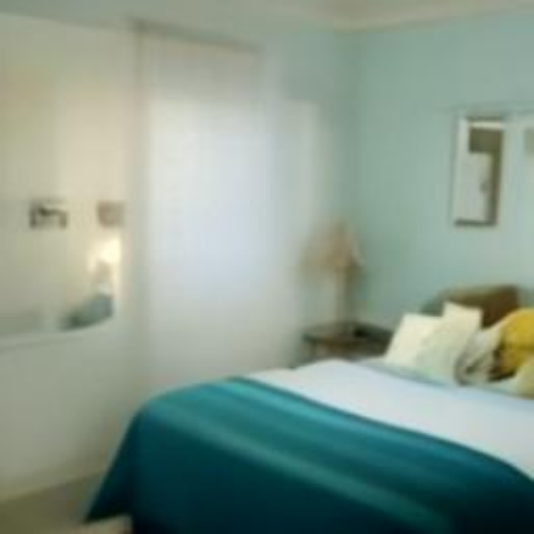}
\hspace{1ex}
\includegraphics[width=0.075\columnwidth]{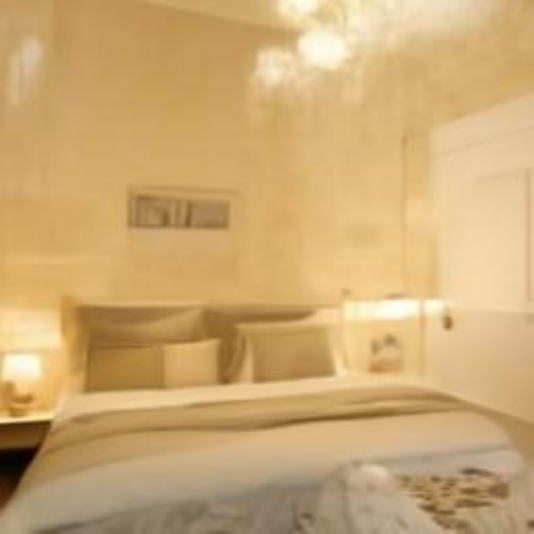}
\includegraphics[width=0.075\columnwidth]{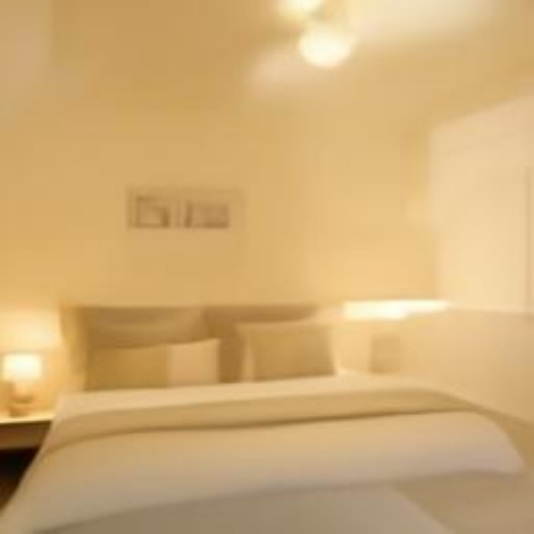}
\includegraphics[width=0.075\columnwidth]{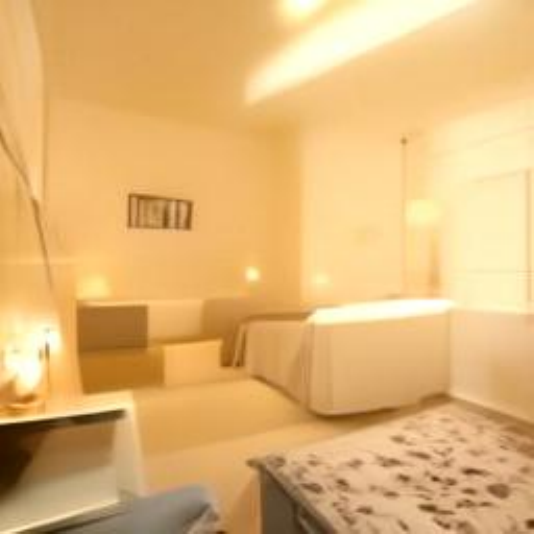}
\includegraphics[width=0.075\columnwidth]{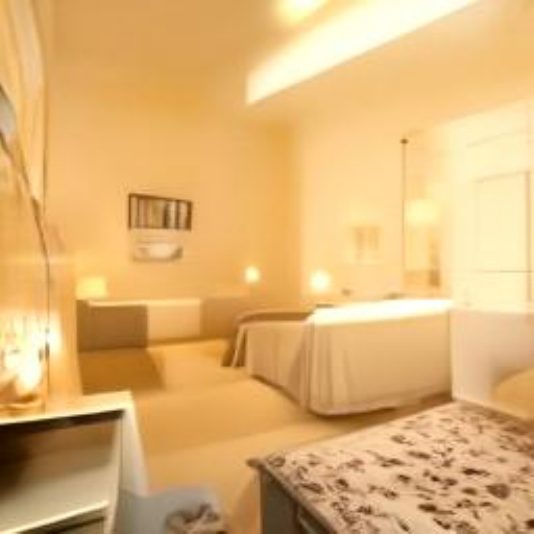}
\hspace{1ex}
\includegraphics[width=0.075\columnwidth]{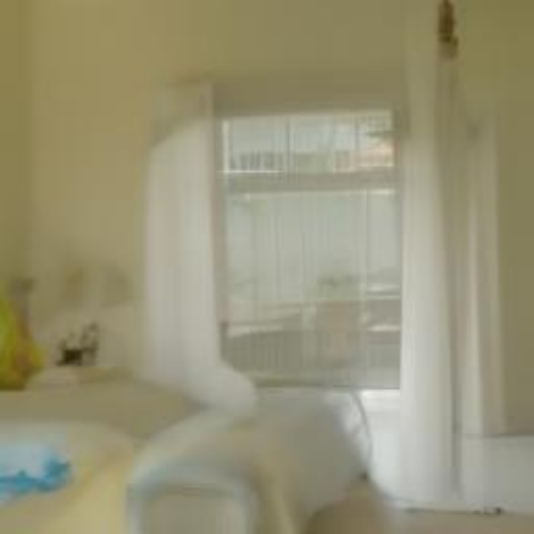}
\includegraphics[width=0.075\columnwidth]{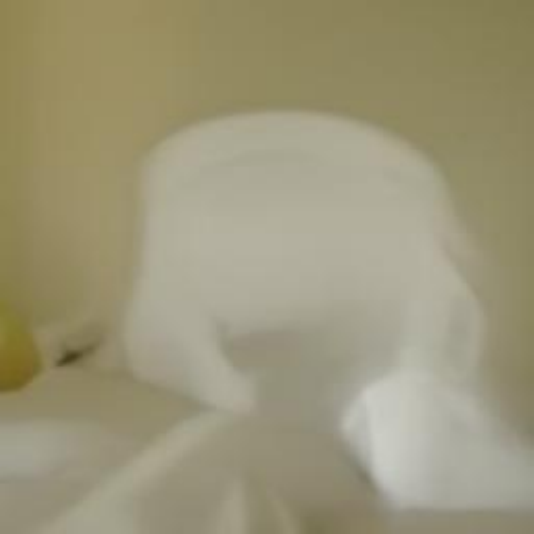}
\includegraphics[width=0.075\columnwidth]{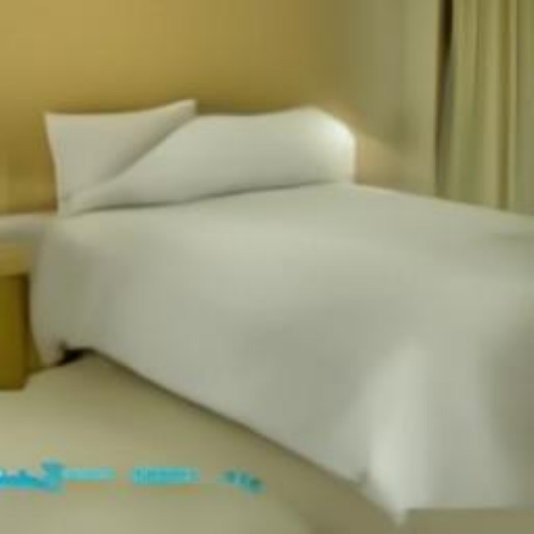}
\includegraphics[width=0.075\columnwidth]{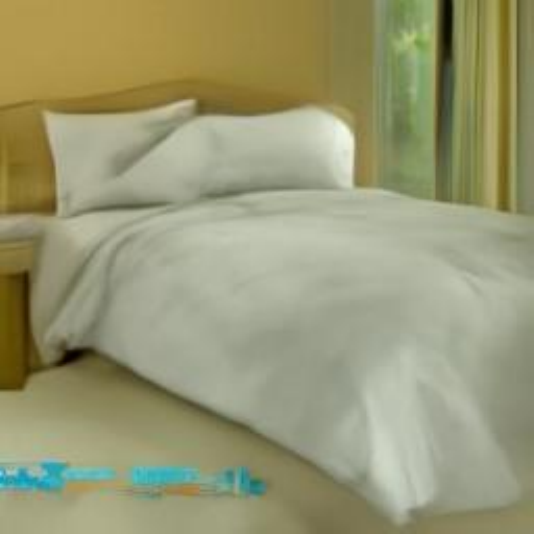}
\\
\includegraphics[width=0.075\columnwidth]{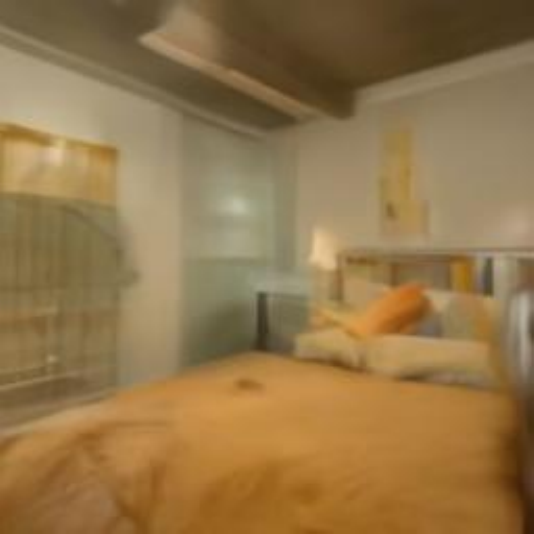}
\includegraphics[width=0.075\columnwidth]{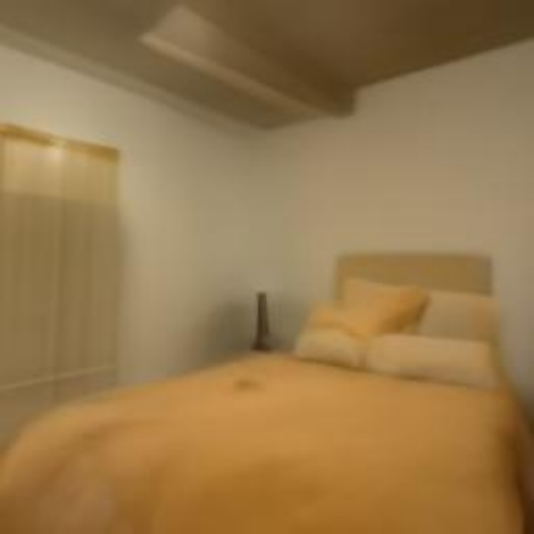}
\includegraphics[width=0.075\columnwidth]{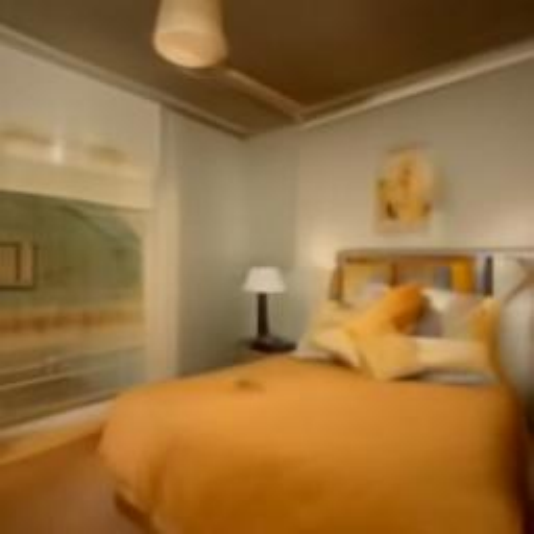}
\includegraphics[width=0.075\columnwidth]{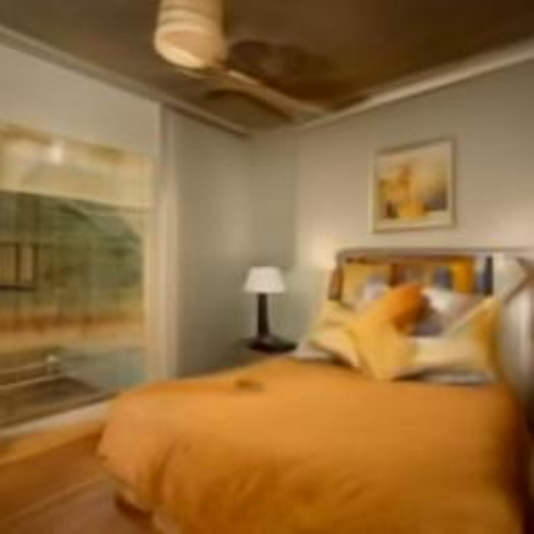}
\hspace{1ex}
\includegraphics[width=0.075\columnwidth]{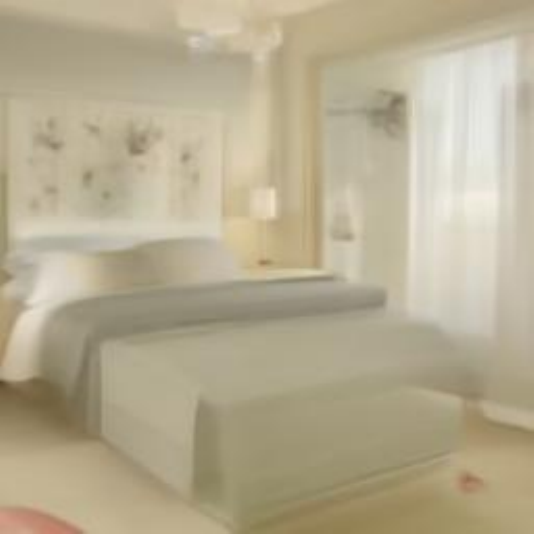}
\includegraphics[width=0.075\columnwidth]{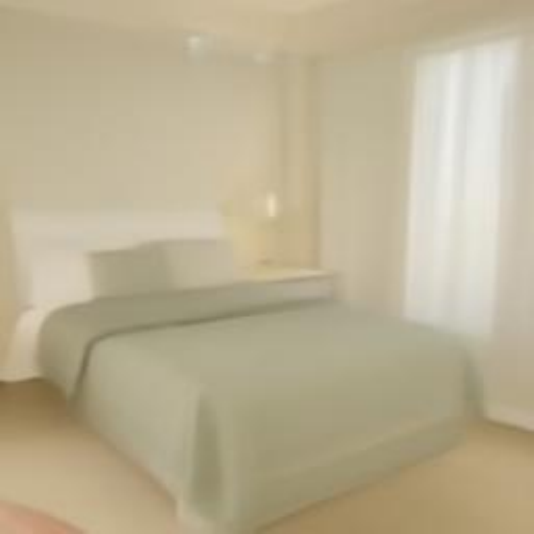}
\includegraphics[width=0.075\columnwidth]{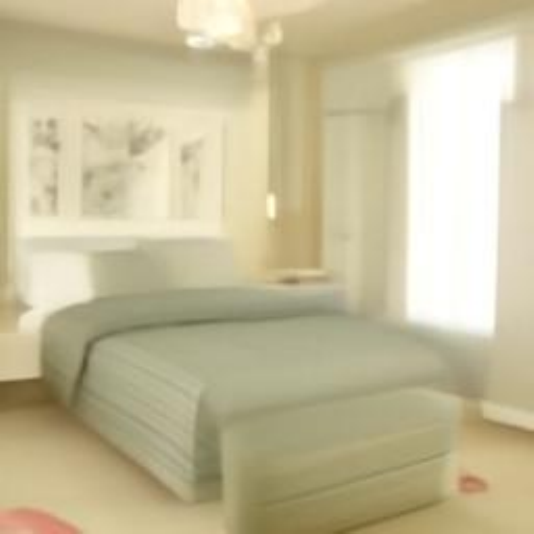}
\includegraphics[width=0.075\columnwidth]{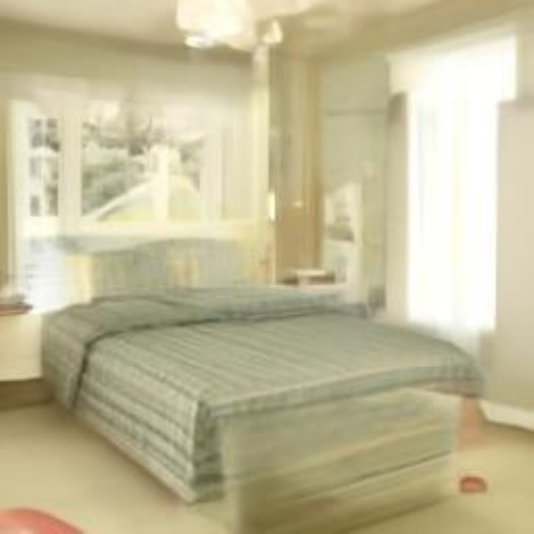}
\hspace{1ex}
\includegraphics[width=0.075\columnwidth]{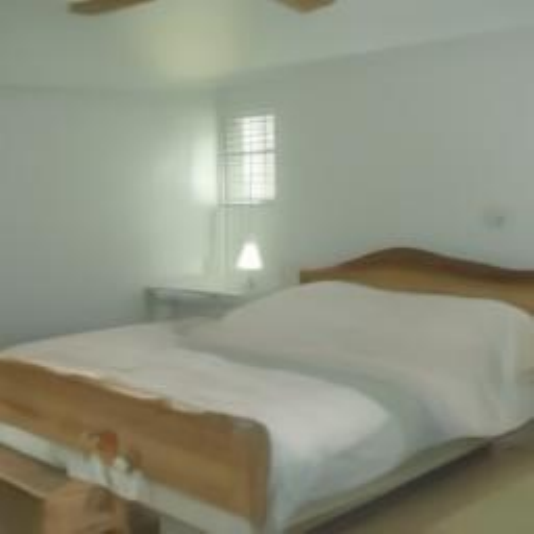}
\includegraphics[width=0.075\columnwidth]{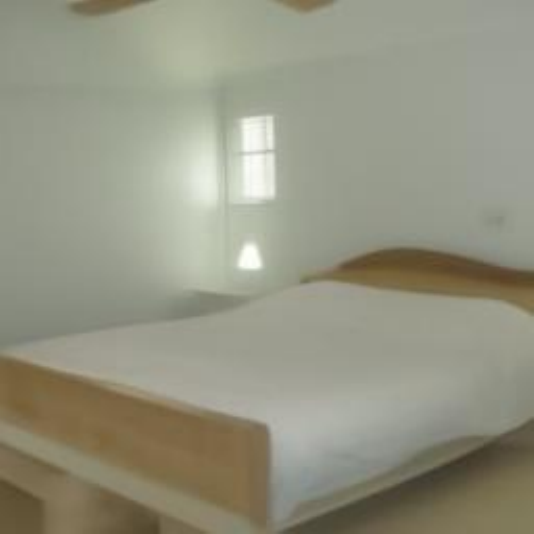}
\includegraphics[width=0.075\columnwidth]{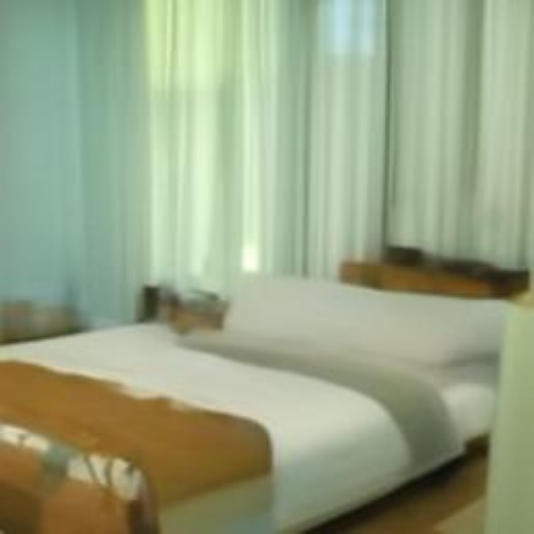}
\includegraphics[width=0.075\columnwidth]{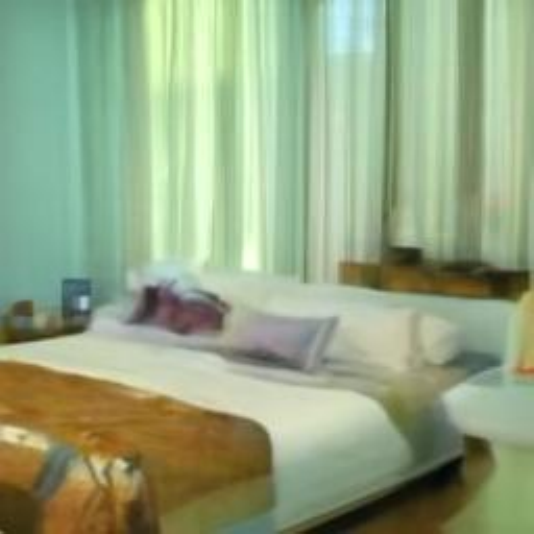}
\caption{\textbf{DDPM with $20$ Steps for Churches (Above) and Bedrooms (Below).}
For each group of four images, from left to right: baseline, ablation, and generated images with noise scalings $1.35$ and $1.55$.
}
    \label{fig:ddpm-20-lsun}
    \end{figure*}

%
%
%
%
%
Since our goal is to improve sampled image quality, 
we compare SE2P with the baseline and ablation on multiple metrics from the \emph{image quality assessment} (IQA) literature. 
In particular, we consider: three 
variants of MUSIQ trained on different IQA datasets~\citep{musiq-2021-junjie}; CLIP-IQA and CLIP-IQA$^+$~\citep{CLIPIQA-2023-Wang}; and the two variants of NIMA~\citep{NIMA-2018-talebi}. 
%
We evaluate the same setting 
as in Section~\ref{sec:qualitative} 
and focus on the lowest number of steps 
in Table~\ref{tab:metric-results}. 
%
%
We also evaluate SE2P with two pretrained DDPMs on the ``church outdoor'' and ``bedroom'' categories, respectively, from the LSUN dataset~\citep{yu-2015-lsun}. Their results are in Table~\ref{tab:metric-results-lsun}, and sample images are shown in Fig.~\ref{fig:ddpm-20-lsun} and Figs.~\ref{fig:ddpm-20-churches} and~\ref{fig:ddpm-20-bed} from Appendix~\ref{app:exp}.
%
%
%
All tables report the average value of each metric across multiple sampled images. 
Each row is described by  
the model name, number of steps, and whether it is the baseline (B), ablation (Abl), or SE2P with some numerical value for the variance scaling. More experimental details and tables are in Appendix~\ref{app:exp}.
%
%
%

For DDPM, considering both CelebA-HQ and LSUN altogether, SE2P obtains the highest score across the vast majority of metrics for some value of the variance scaling (the only exceptions were in sampling bedrooms).
For LD and DiT, SE2P always obtains the highest score across all metrics. 
The results for SD, considering both $20$ and $40$ steps (in Appendix~\ref{app:exp}), provide a more diverse distribution of scores across the baseline, ablation, and SE2P. This possibly reflects our qualitative observation that it is difficult to assess strong visual quality differences for SD (Section~\ref{sec:qualitative}). 
Nonetheless, our results suggest that SE2P \emph{does not reduce} sampled image quality with SD.  
%
%

We highlight that it is possible that our results could further improve for other values of variance scaling. 
Nonetheless, our results are enough to 
show 
that SE2P can improve sampled image quality.



In conclusion, our results overall match our qualitative study: across most models and metrics, SE2P consistently surpasses the baseline on quality, and the ablation 
performs the worst. 
%
%
%
%
%
%
%
%
\begin{table*}[t!]
    \caption{
    \textbf{IQA Scores -- Setting as in Section~\ref{sec:qualitative}.}}
    \label{tab:metric-results}
    \centering
    \def\arraystretch{1}
    
     \small
    \resizebox{0.9\textwidth}{!}{
    \begin{tabularx}{\textwidth}{c*{7}{>{\centering\arraybackslash}X}}
    \toprule
    & \textbf{MUSIQ-koniq} & \textbf{MUSIQ-ava} & \textbf{MUSIQ-paq2piq} & \textbf{CLIP-IQA} & \textbf{CLIP-IQA$^+$} & \textbf{NIMA-inceptv2} & \textbf{NIMA-vgg16}\\ 
    \cmidrule{2-8}

\textbf{DDPM--10--B}  & \underline{45.376} & 4.322 & 74.062 & 0.523 & 0.559 & 4.161 & 4.524\\ 
\textbf{DDPM--10--Abl}  & 44.303 & 4.353 & 73.876 & 0.519 & 0.554 & 4.184 & 4.498\\ 
\textbf{DDPM--10--1.35}  & 45.129 & \textbf{4.393} & \underline{74.262} & \underline{0.531} & \underline{0.567} & \textbf{4.302} & \underline{4.710}\\ 
\textbf{DDPM--10--1.55}  & \textbf{45.529} & \underline{4.381} & \textbf{74.342} & \textbf{0.533} & \textbf{0.569} & \underline{4.295} & \textbf{4.722}\\ 
\midrule
\textbf{LD--20--B}  & 54.939 & 4.327 & 77.007 & 0.605 & 0.634 & 4.114 & 4.653\\
\textbf{LD--20--Abl}  & 51.785 & \underline{4.362} & 76.395 & 0.571 & 0.614 & 4.107 & 4.604\\
\textbf{LD--20--1.35}  & \underline{55.877} & \textbf{4.381} & \underline{77.270} & \underline{0.614} & \underline{0.646} & \underline{4.216} & \underline{4.738}\\
\textbf{LD--20--1.55}  & \textbf{57.948} & 4.350 & \textbf{77.596} & \textbf{0.635} & \textbf{0.655} & \textbf{4.223} & \textbf{4.767} \\
\midrule
\textbf{DiT--40--B}  & 47.943 & 3.833 & 73.215 & 0.499 & 0.571 & 3.870 & 4.061\\
\textbf{DiT--40--Abl}  & 45.056 & 3.866 & 71.199 & 0.478 & 0.548 & 3.951 & 4.080\\
\textbf{DiT--40--1.15}  & \underline{49.436} & \textbf{3.928} & \underline{73.737} & \underline{0.530} & \underline{0.589} & \textbf{4.059} & \textbf{4.296}\\
\textbf{DiT--40--1.35}  & \textbf{50.158} & \underline{3.876} & \textbf{74.697} & \textbf{0.539} & \textbf{0.592} & \underline{3.964} & \underline{4.243} \\
\midrule
\textbf{SD--20--B}  & \underline{57.096} & 4.327 & \underline{77.072} & \underline{0.604} & \underline{0.679} & 4.456 & 4.587 \\
\textbf{SD--20--Abl}  & 56.767 & \textbf{4.358} & 76.776 & 0.599 & 0.673 & \underline{4.494} & 4.593 \\
\textbf{SD--20--1.25}  & 56.928 & \underline{4.356} & 77.029 & \underline{0.604} & 0.677 & \textbf{4.503} & \textbf{4.617} \\
\textbf{SD--20--1.55}  & \textbf{57.214} & 4.326 & \textbf{77.270} & \textbf{0.607} & \textbf{0.681} & 4.466 & \underline{4.609} \\
    \bottomrule
    \end{tabularx}}
    %
    %
\end{table*}

\begin{table*}[t!]
    \caption{
    \textbf{IQA Scores -- ``Outdoor Church'' (Above) \& ``Bedroom'' (Below) from LSUN Dataset.}}
    \label{tab:metric-results-lsun}
    \centering
    \def\arraystretch{1}
    
     \small
    \resizebox{0.9\textwidth}{!}{
    \begin{tabularx}{\textwidth}{c*{7}{>{\centering\arraybackslash}X}}
    \toprule
    & \textbf{MUSIQ-koniq} & \textbf{MUSIQ-ava} & \textbf{MUSIQ-paq2piq} & \textbf{CLIP-IQA} & \textbf{CLIP-IQA$^+$} & \textbf{NIMA-inceptv2} & \textbf{NIMA-vgg16}\\ 
    \cmidrule{2-8}

\textbf{DDPM--20--B}  & \underline{40.772} & \underline{3.820} & \underline{69.450} & \underline{0.348} & 0.550 & 4.199 & 4.279 \\
\textbf{DDPM--20--Abl}  & 30.478 & 3.630 & 64.337 & 0.315 & 0.468 & 4.158 & 4.128 \\
\textbf{DDPM--20--1.35}  & 39.053 & 3.762 & 69.160 & 0.343 & \underline{0.552} & \underline{4.219} & \underline{4.416} \\
\textbf{DDPM--20--1.55}  & \textbf{50.159} & \textbf{4.079} & \textbf{73.623} & \textbf{0.435} & \textbf{0.620} & \textbf{4.512} & \textbf{4.758} \\
\midrule
\textbf{DDPM--20--B}  & \underline{30.394} & \underline{3.384} & 64.444 & \underline{0.201} & \underline{0.484} & \underline{3.623} & 3.727 \\
\textbf{DDPM--20--Abl}  & 28.006 & \textbf{3.438} & 64.196 & \textbf{0.228} & \textbf{0.507} & \textbf{3.716} & 3.783 \\
\textbf{DDPM--20--1.35}  & 29.444 & 3.347 & \underline{64.570} & 0.194 & 0.478 & 3.597 & \textbf{3.798} \\
\textbf{DDPM--20--1.55}  & \textbf{32.530} & 3.357 & \textbf{65.619} & 0.195 & 0.458 & 3.591 & \underline{3.795} \\
%
    \bottomrule
    \end{tabularx}}
    %
    %
\end{table*}

\subsection{Human Evaluation}
\label{sec:human}
We conducted a human evaluation study on image quality assessment. 
For each DDPM (on CelebA-HQ), LD, and DiT models, we present to $33$ people $25$ pairs of randomly sampled images---%
each pair has an image generated by the baseline and one by SE2P sharing the same random seed.  
We did not consider SD in light of our previous discussion.
%
For each pair, the human evaluator is asked to choose which image has better quality. 
To avoid positional bias, we randomly shuffle the order of the images generated by SE2P at each pair. 
Further motivation and details on our 
setup and a comparison to other human evaluation studies are in Appendix~\ref{app:humaneval}.
%
%
%
%
%
%
%
%
%

Our results are in Table~\ref{tab:human-eval}. 
The first row describes the percentage of human evaluators who chose SE2P in the \emph{majority} of pairs: 
remarkably, this is no less than $76\%$ across all models. 
%
%
%
%
Now, let us consider the percentage of pairs (over the total of $25$) in which an evaluator chose SE2P over the baseline. Per model, we obtained both the \emph{mean} and \emph{median} of such percentage across all evaluators, and reported them in the second and third rows of Table~\ref{tab:human-eval}, respectively. 
%
%
Remarkably, this mean rate of choosing SE2P is no less than $60\%$ across all models, and the median no less than $64\%$. 
%
Since the median is larger than the mean in every model,
a large concentration of people 
chose SE2P at a high rate. 

\begin{table}[t!]
\caption{
    \textbf{Human Evaluation Study of SE2P.}
    %
    }
    \label{tab:human-eval}
    \centering
    \def\arraystretch{1}
    
    \resizebox{0.45\columnwidth}{!}{
    \begin{tabular}{@{}cccc@{}}
    \toprule
    & \textbf{DDPM} & \textbf{LD} & \textbf{DiT} \\ 
    \cmidrule{2-4}

\textbf{\% of evaluators choosing S2EP}  & 87.88 & 91.18 & 76.47 \\
%
%
\textbf{mean \% of pairs with S2EP}  & 65.21  & 63.64  & 60.61 \\
\textbf{median \% of pairs with S2EP}  & 68.00 & 64.00 & 64.00 \\
    \bottomrule
    \end{tabular}}
    %
    %
\end{table}

\section{Parameter Changes}
\label{sec:param-change}
We explore the effects of changing SE2P's parameters (mixing parameter and variance scaling) and the number of parallel processors. 
%
%
We particularly focus on exemplifying when parameter change can be detrimental to sampled image quality. 

%
%
%

\subsection{Changing the Variance Scaling} 
\label{subsec:var-higher}

In the domain of low number of steps, we find that noticeable image deterioration can start to appear as we vary the variance scaling. 
This is illustrated for LD for $20$ and $40$ steps in Fig.~\ref{fig:sweep-var} from Appendix~\ref{app:param-changes}. 
Indeed, values of variance scaling that worked for 
a low number of steps can have a negative effect on a large number of steps 
across DDPM, LD, DiT and SD; see 
Fig.~\ref{fig:sc-var-wr} in Appendix~\ref{app:param-changes}. 

\subsection{Changing the Mixing Parameter}
\label{sec:mixparam}
Our results thus far have the mixing parameter value fixed. 
Interestingly, reducing the mixing parameter can lead to image degradation, while increasing it can lead to image loss---see the case of DDPM in Fig.~\ref{fig:sweep-mix} from Appendix~\ref{app:param-changes}. 
Remarkably, we find a similar image degradation when integrating the predictive value $\hat{\x}_{\operatorname{pred}}$ from Algorithm~\ref{alg:main_SE2P} with the latent at step $t_k+1$ (instead of the usual latent at step $t_k$)---see the case of DDPM in Fig.~\ref{fig:back-int} from Appendix~\ref{app:param-changes}.
%

\subsection{Increasing the Number of Parallel Processors}

Adding more parallel processors increases computation costs and undermines 
the motivation of our paper, nonetheless, it is fair to ask whether this is a price to pay to further improve sampled image quality. 
%
Certainly, adding \emph{more} parallel processors means integrating \emph{more} information from different steps, and one could hypothesize this ultimately can result in \emph{more} sampled image quality. 
%
Thus, to test this hypothesis, we easily expand 
SE2P 
to more than two parallel processors---see Algorithm~\ref{alg:main_SE2P_more} in Appendix~\ref{app:param-changes}---and generate samples from it. 
%
%
%
We actually find that increasing the number of processors does not imply a monotonic increase in image quality---instead, larger numbers of parallel processors constantly lead to image quality degradation. 
This is illustrated for DDPM and LD in Fig.~\ref{fig:num-proc} from Appendix~\ref{app:param-changes}.
Therefore, we believe that two processors are enough for practical applications. 

\section{Related Literature}
\label{sec:lit-review}

Although it tackles a different problem, the literature on inference acceleration of diffusion models is perhaps the most related to our work. 
Example approaches are to: construct deterministic sample paths with 
all randomness coming from the initial Gaussian latent~\citep{song-2021-denoisingimplicit,song-2021-scorebased,lu-2022-dpmsolver}, 
find time schedules that reduce denoising steps for a pre-trained model through optimization problems~\citep{watson-2021-learningefficientlysample}, 
solve an optimization problem to schedule 
diverse pre-trained models in order to optimize both quality and latency~\citep{liu-2023-omsdpmmodelschedule}, 
distill models that require less steps~\citep{salimans-2022-progressive}, use external components to adaptively adjust diffusion time and number of denoising steps on the fly~\citep{ye-2025-scheduleonthefly}, switch across a tailored small model to lower overall latency~\citep{li-2025-morse}, etc. 
Some other works 
leverage parallel processing to improve inference latency 
while aiming to not sacrifice image quality.
\cite{shih-2023-parallel} 
use parallel processors to solve a fixed-point problem that jointly estimates the full sample trajectory and iteratively refine such estimations in parallel. 
The refinement process stops through some heuristic
which leads 
to less iterations 
than the number of denoising steps 
it would take otherwise, but it 
could
also 
lead to image quality loss. 
%
The number of parallel processors 
is related to the discretization 
of the 
sample trajectory, and reported results use 
as far as $80$ processors.
We refer to~\citep{pokle-2022-deepequivnetdiffmod,cao-2024-deepequivdiffrestor,tang-2024-accparasampl} for other works that also use similar fixed-point approaches for full trajectory estimation using parallel processing---in particular,~\citet{tang-2024-accparasampl} generalize the approach by~\citet{shih-2023-parallel}. 
%
\cite{hu-2025-diffusion} 
adapt speculative decoding to diffusion models and make repetitive parallel model calls to make predictions of future mean values of the latent image. Remarkably, they are able to reduce inference latency without image quality loss. The number of parallel processors 
is proportional to the amount of speculation 
and is what reduces the inference latency
---noticeable 
speed-up is reported 
for at least four processors. 
\citet{bortoli-2025-accelerated} propose another speculative sampling approach that, instead, accelerates inference by making parallel evaluations of a more efficient draft model that matches the distribution of generated latents of the original pretrained diffusion model. 
Lastly,~\citet{li-2024-distriinf} take a completely different approach to inference acceleration: it splits the latent images across different parallel processors, while ensuring consistency among the partitions. Since each processor now deals with smaller latent images, the denoising evaluations are more efficient. 

Finally, we mention that another 
machine learning field
where parallel computation is used to \emph{improve} some outcome under a \emph{restriction} on the number of algorithmic iterations is 
reinforcement learning (RL). For example, RL training can use parallel processors to improve sample efficiency and reduce the length of policy exploration
~\citep{dima-2018-coordexplorconcur,zhang-2020-modelfree,cineros-velarde-2023-onepolicyRL,li-2023-parallelqlearning}.   

\section{Conclusion}
We present SE2P, a simple algorithm to improve sampled image quality 
using two parallel processors given a low number of denoising steps. Our algorithm is easy to implement and has overall outperformed the baseline in both automated and human evaluations, across different diffusion models. 

A future direction is to study how SE2P performs (perhaps under some adaptation) when used with diffusion models of other modalities. 
Another direction is to find theoretical underpinnings to (i) how the integration of a latent and its prediction can enhance sample quality; and (ii) 
%
%
why larger numbers of denoising steps can lead to more semantic change with SE2P---perhaps making a connection with the sampling guidance literature.    
%

%

\subsection*{Acknowledgments}
We thank the VMware Research Group. We thank the people at VMware who provided the resources to run the experiments. We also thank the participants in our evaluation study. Finally, we thank Jessica C. for the help in the experiments and writing of the paper.

\bibliographystyle{plainnat}
\bibliography{biblio}

\clearpage
\appendix
\thispagestyle{empty}


%
%

\section{About Ensemble Methods}
\label{app:ensemble}

As mentioned in Section~\ref{sec:intro}, the last two desiderata (iii) and (iv) exclude the use of \emph{ensemble methods}. 
Let us consider the practical 
``best-of-M runs'' method: running M parallel inference processes and finding a way to choose, out of the M sampled images, the one with best quality. The problem is that choosing such image implies the evaluation of image quality metrics over each of the M images, which implies the repeated use of external neural models for computing these metrics (e.g., see the citations of the IQA metrics in Section~\ref{sec:metric}), 
thus 
violating (iii). This also carries the additional problem of finding out the \emph{the most suitable} image quality metric to use. Moreover, the ensemble method's effectiveness 
benefits 
%
from a large M 
because this augments the diversity of the samples to choose from---thus 
using 
more than 
two parallel processors, 
violating (iv). 
Finally, we point out that our proposed method works by improving image quality over undesired image attributes that are typical of low iteration regimes---such as low contrast, blurred features, etc.---something 
that ensemble methods are not able to do. 
%
%
%
%


\section{Technical Notes About S2EP}
\label{app:technicalnotesS2EP}
We present four technical notes regarding SE2P from Algorithm~\ref{alg:main_SE2P}.
\begin{enumerate}
\item 
Computing $\hat{\x}_{\operatorname{pred}}$ (line 8) does not need an extra evaluation of the diffusion model 
because of 
the previously stored variable $\mathbf{v}^{(0)}$ (line 6) that was computed during the parallel denoising process (line 15). 
\item 
Algorithm~\ref{alg:main_SE2P} indicates that Processor 0 executes lines 5 to 8, however, this can be changed to Processor 1 if Processor 0 sends the variables $\v^{(0)}$ and $\x_k^{(0)}$ to Processor 1 before line 5 (and so $\x_{\operatorname{pred}}$ is computed locally in Processor 1). 
\item 
We initialize both processors equally (line 2) and fix the random seed to be equal for both processors during the parallel denoising (line 12): since we want to enhance image quality, the idea is to have the random stochastic paths of the denoising process not very different across both processors. Sharing the random seed can be accomplished by either (i) having one processor randomly sampling some seed and sending it to the other processor, or by (ii) pre-computing a unique list of random seeds which both processors store in memory during their initialization. 
Notice that only the instruction in line 16 uses the previously shared random seed---lines 2 and 8 do not use any predefined seed. 
\item 
The constants $\alpha_t$, $\bar{\alpha}_t$ and $\tilde{\beta}_t$ can also be pre-computed before running the algorithm and passed to the processors instead of being computed on the go.
\end{enumerate}
\section{Further Experimental Details \& Results}
\label{app:exp}

Our experiments were done using a mix of NVIDIA's RTX A1000 mobile GPU (found in commercial laptops), and the more powerful A100 and H100 GPUs.

We present a couple of examples of jump sampling following the notation of Section~\ref{sec:model-setting} (recalling 
that steps go in reverse order). For $N=10$, the denoising steps are $t = 901, 801, 701, 601, 501, 401, 301, 201, 101, 1$. For $N=20$, the denoising steps are $t= 951, 901, 851, 801, 751, 701, 651, 601, 551, 501, 451, 401, 351, 301, 251, 201, 151, 101,  51, 1$.

We remark that the mixing parameter is fixed to the value of $0.015$ throughout our qualitative (Section~\ref{sec:qualitative}) and quantitative (Section~\ref{sec:quantitative}) studies.

\subsection{Qualitative Study: More Figures}
\label{app:qual-res}

\begin{figure}[t]
    \centering
\includegraphics[width=0.15\columnwidth]{img_dif/DDPM_10_mix/b_63.pdf}
\includegraphics[width=0.15\columnwidth]{img_dif/DDPM_10_mix/a_63.pdf}
\includegraphics[width=0.15\columnwidth]{img_dif/DDPM_10_mix/m_135_63.pdf}
\includegraphics[width=0.15\columnwidth]{img_dif/DDPM_10_mix/m_155_63.pdf}
\\
\includegraphics[width=0.15\columnwidth]{img_dif/DDPM_10_mix/b_200.pdf}
\includegraphics[width=0.15\columnwidth]{img_dif/DDPM_10_mix/a_200.pdf}
\includegraphics[width=0.15\columnwidth]{img_dif/DDPM_10_mix/m_135_200.pdf}
\includegraphics[width=0.15\columnwidth]{img_dif/DDPM_10_mix/m_155_200.pdf}
\\
\includegraphics[width=0.15\columnwidth]{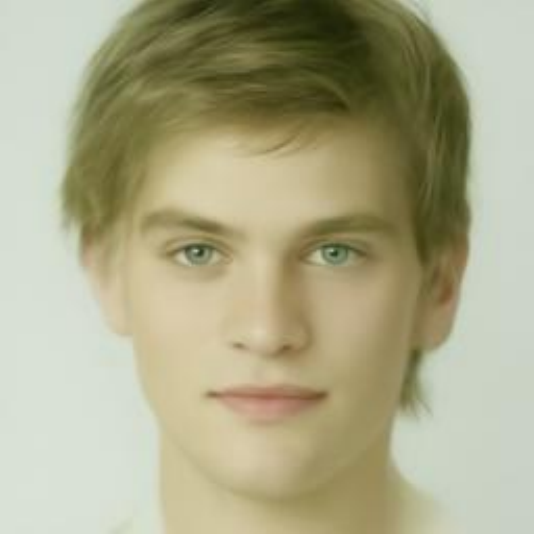}
\includegraphics[width=0.15\columnwidth]{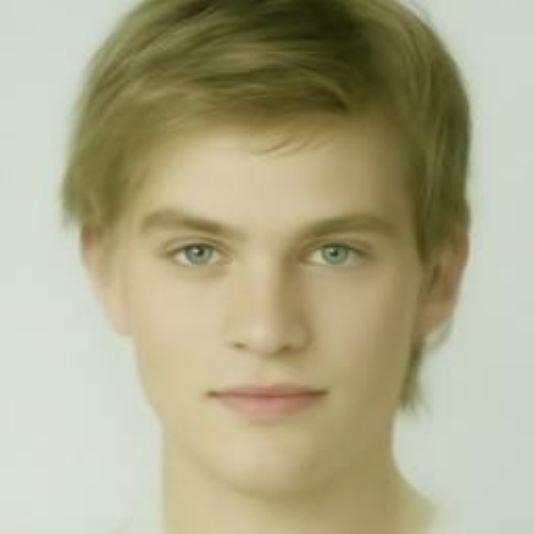}
\includegraphics[width=0.15\columnwidth]{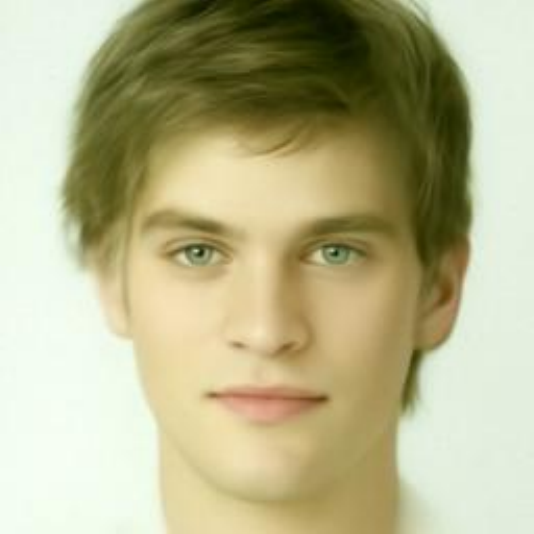}
\includegraphics[width=0.15\columnwidth]{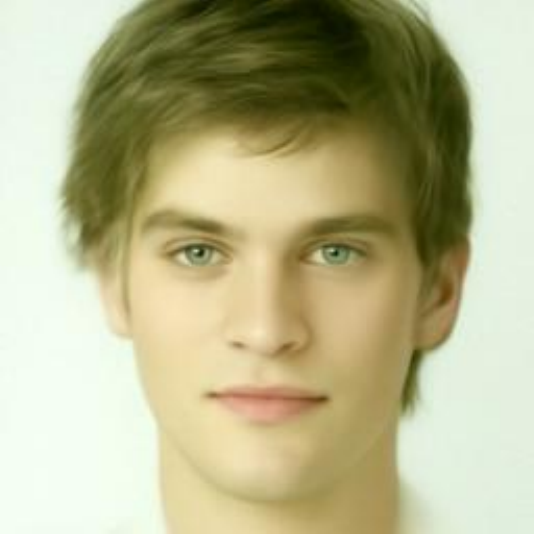}
\\
\includegraphics[width=0.15\columnwidth]{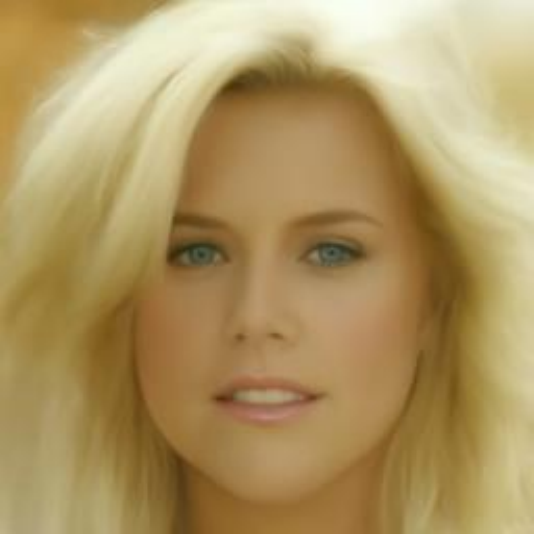}
\includegraphics[width=0.15\columnwidth]{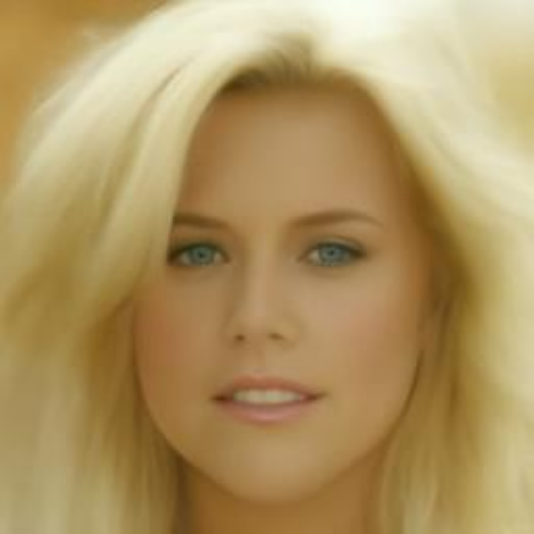}
\includegraphics[width=0.15\columnwidth]{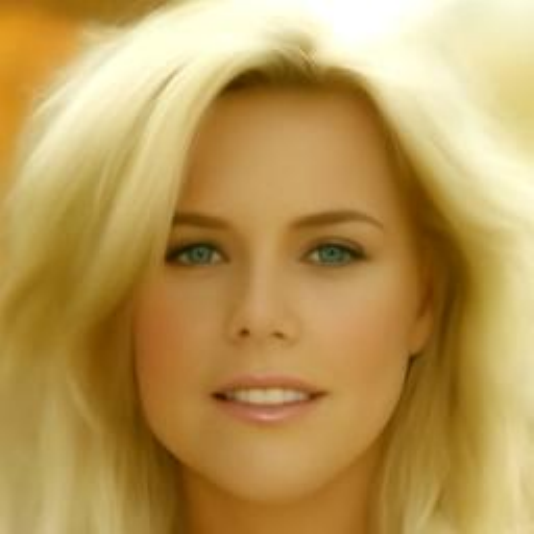}
\includegraphics[width=0.15\columnwidth]{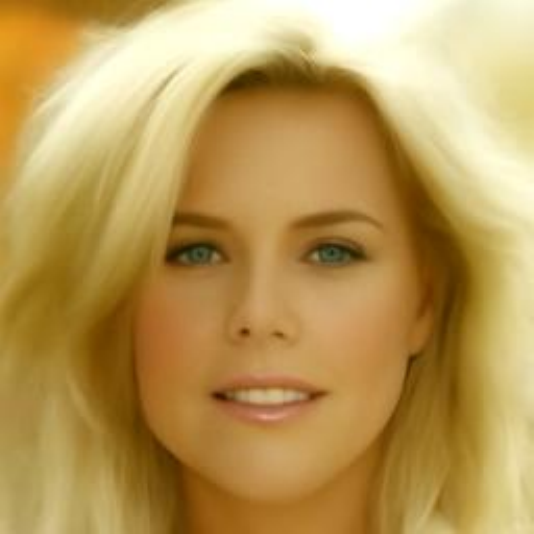}
\\
\includegraphics[width=0.15\columnwidth]{img_dif/DDPM_10_mix/b_254.pdf}
\includegraphics[width=0.15\columnwidth]{img_dif/DDPM_10_mix/a_254.pdf}
\includegraphics[width=0.15\columnwidth]{img_dif/DDPM_10_mix/m_135_254.pdf}
\includegraphics[width=0.15\columnwidth]{img_dif/DDPM_10_mix/m_155_254.pdf}
\\
\includegraphics[width=0.15\columnwidth]{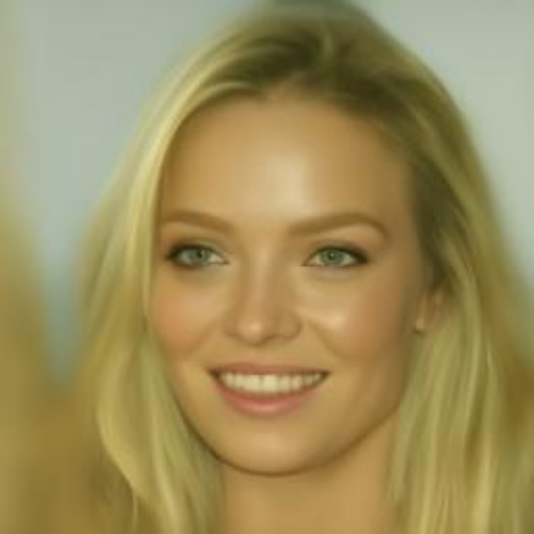}
\includegraphics[width=0.15\columnwidth]{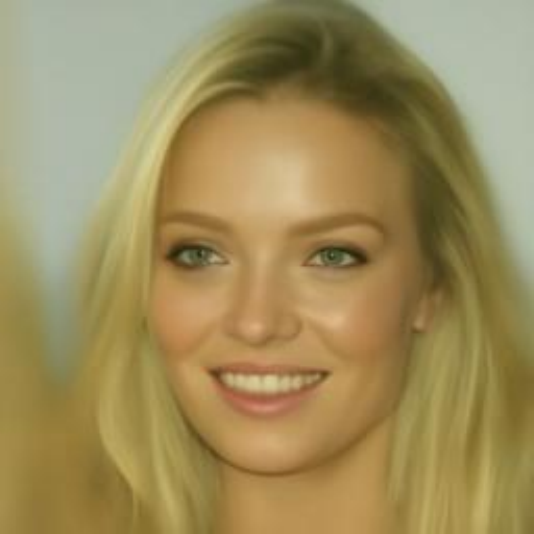}
\includegraphics[width=0.15\columnwidth]{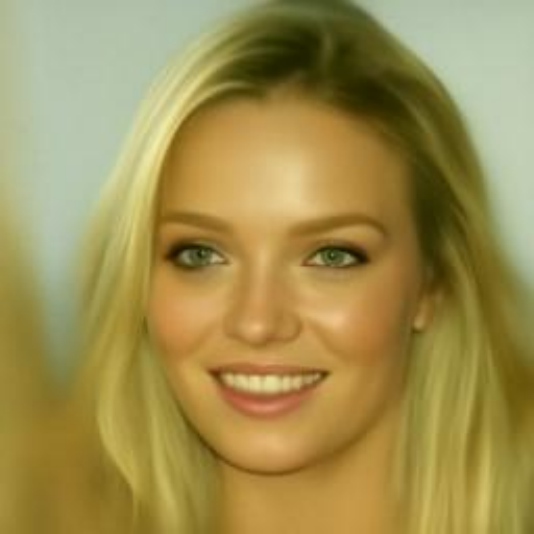}
\includegraphics[width=0.15\columnwidth]{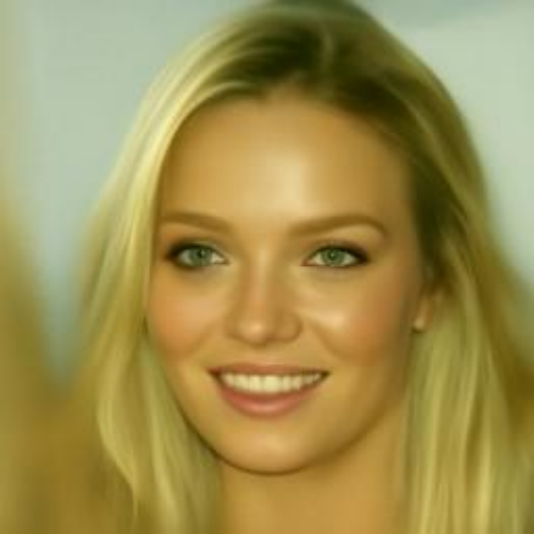}
\\
\includegraphics[width=0.15\columnwidth]{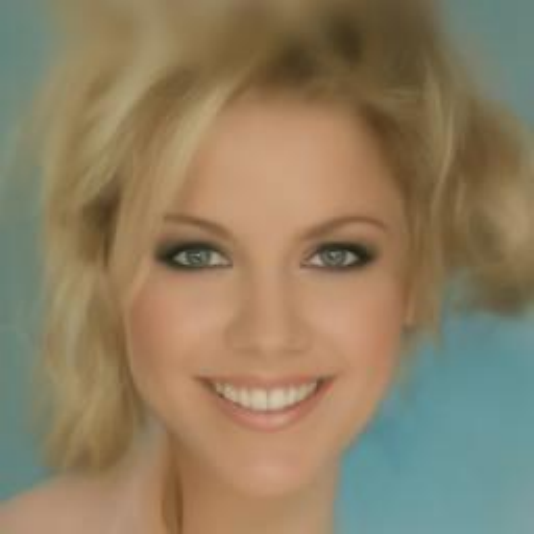}
\includegraphics[width=0.15\columnwidth]{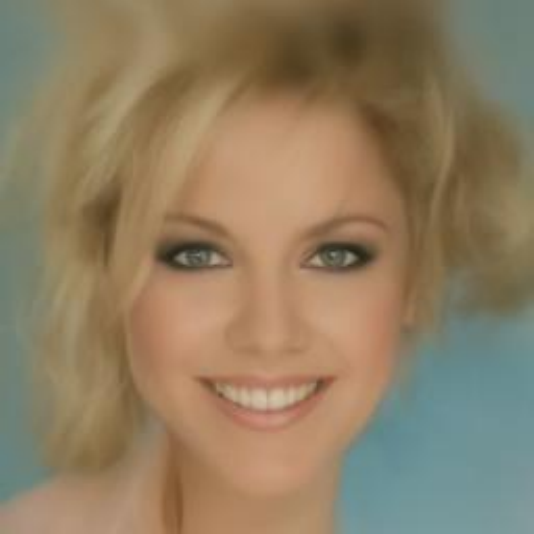}
\includegraphics[width=0.15\columnwidth]{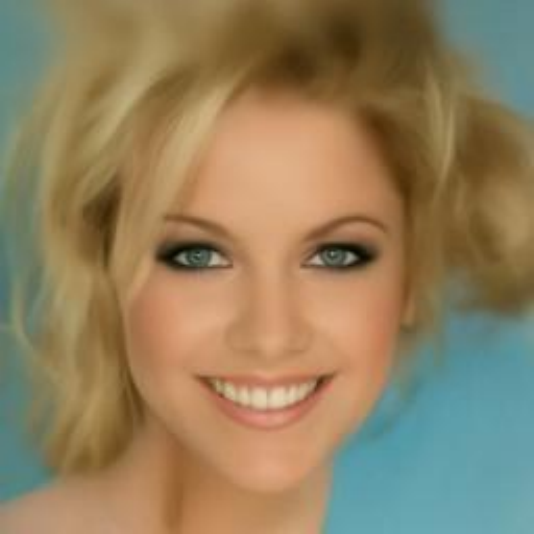}
\includegraphics[width=0.15\columnwidth]{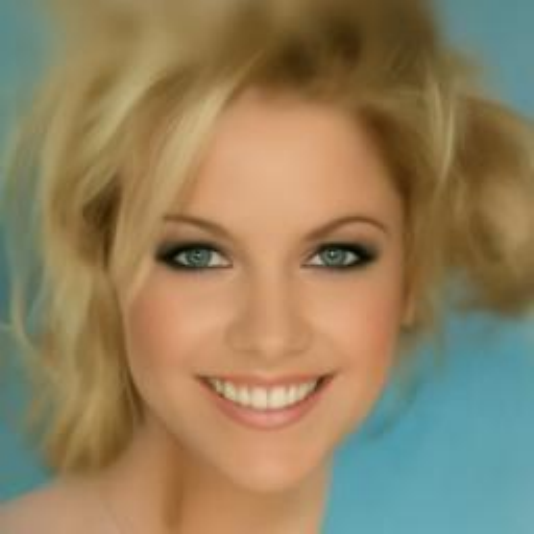}
\\
\includegraphics[width=0.15\columnwidth]{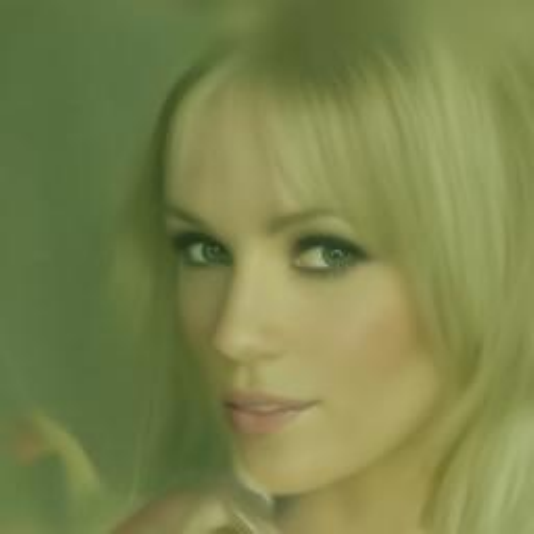}
\includegraphics[width=0.15\columnwidth]{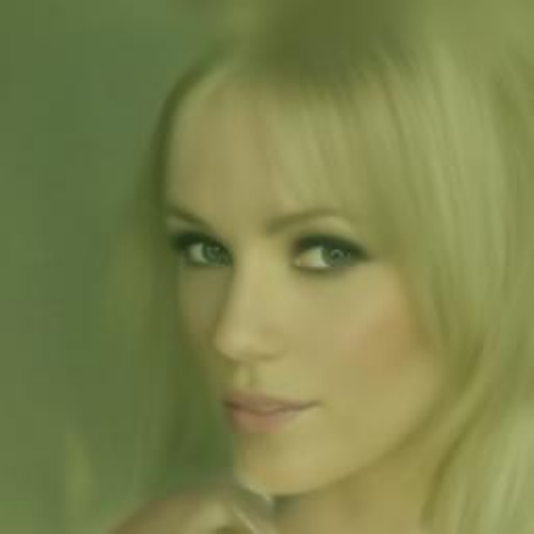}
\includegraphics[width=0.15\columnwidth]{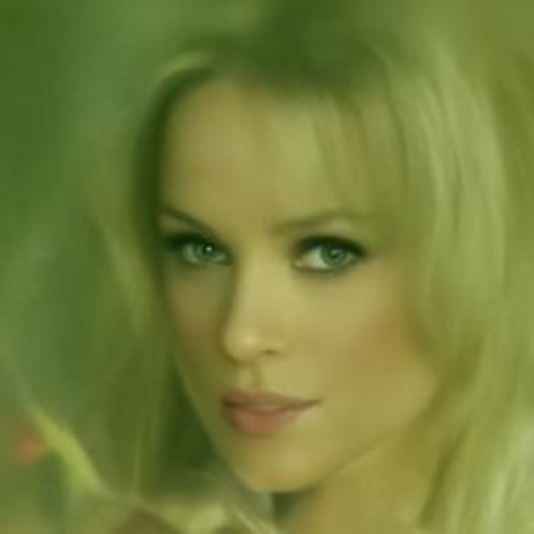}
\includegraphics[width=0.15\columnwidth]{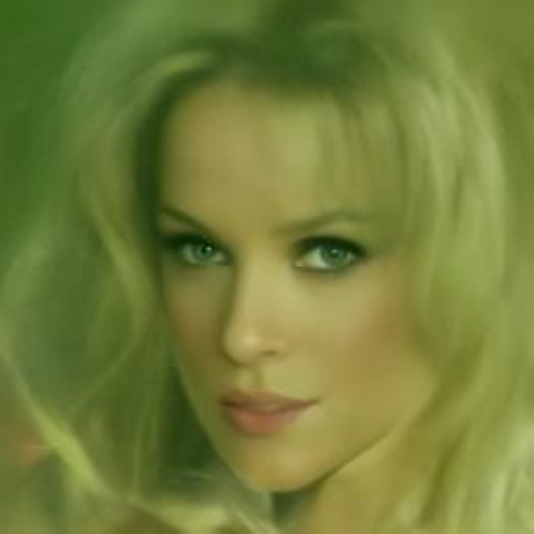}
\\
\caption{\textbf{DDPM with $10$ Steps.} 
For each row, from left to right: baseline, ablation, generated image with variance scaling $1.35$, and one with $1.55$. 
%
%
}
    \label{fig:ddpm-10}
    \end{figure}

\begin{figure}[t]
    \centering
\includegraphics[width=0.15\columnwidth]{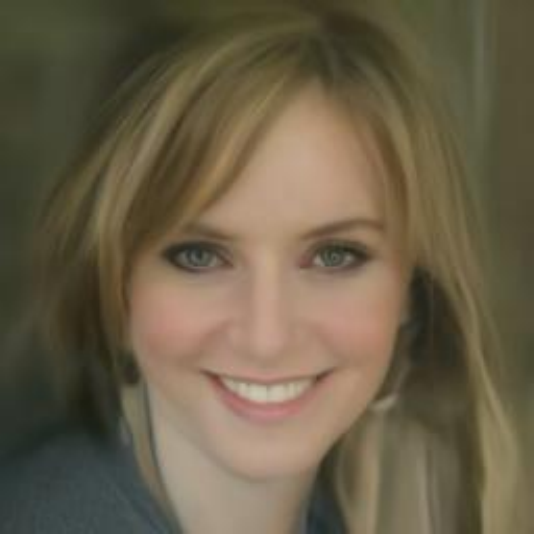}
\includegraphics[width=0.15\columnwidth]{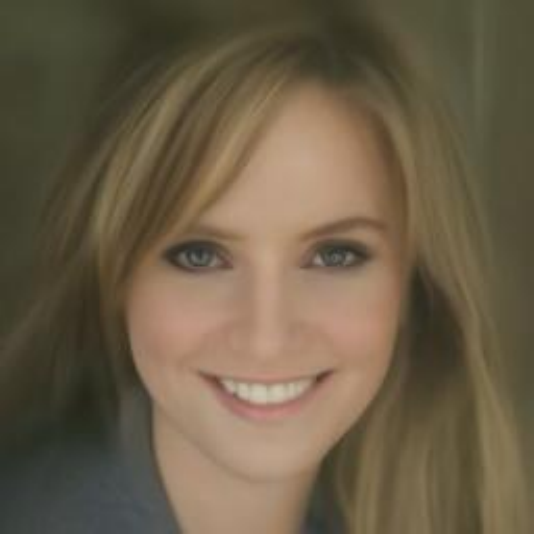}
\includegraphics[width=0.15\columnwidth]{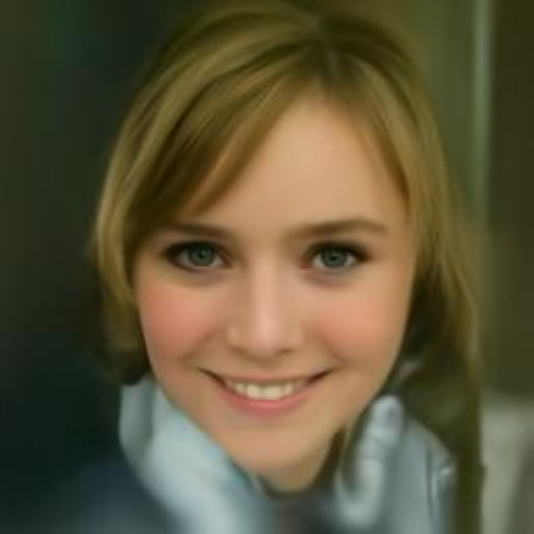}
\includegraphics[width=0.15\columnwidth]{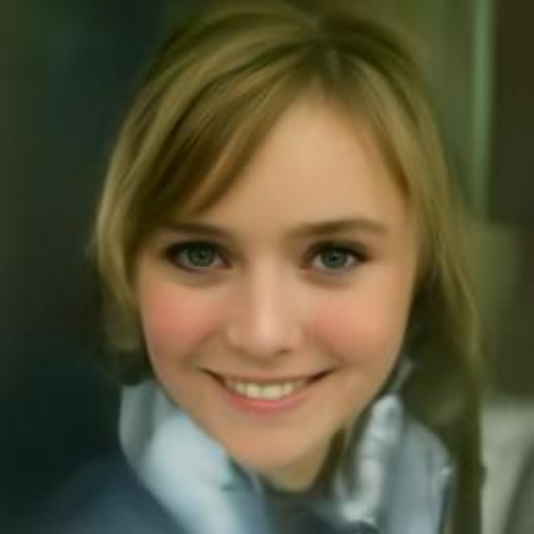}
\\
\includegraphics[width=0.15\columnwidth]{img_dif/DDPM_20_mix/b_50.pdf}
\includegraphics[width=0.15\columnwidth]{img_dif/DDPM_20_mix/a_50.pdf}
\includegraphics[width=0.15\columnwidth]{img_dif/DDPM_20_mix/m_135_50.pdf}
\includegraphics[width=0.15\columnwidth]{img_dif/DDPM_20_mix/m_155_50.pdf}
\\
\includegraphics[width=0.15\columnwidth]{img_dif/DDPM_20_mix/b_88.pdf}
\includegraphics[width=0.15\columnwidth]{img_dif/DDPM_20_mix/a_88.pdf}
\includegraphics[width=0.15\columnwidth]{img_dif/DDPM_20_mix/m_135_88.pdf}
\includegraphics[width=0.15\columnwidth]{img_dif/DDPM_20_mix/m_155_88.pdf}
\\
\includegraphics[width=0.15\columnwidth]{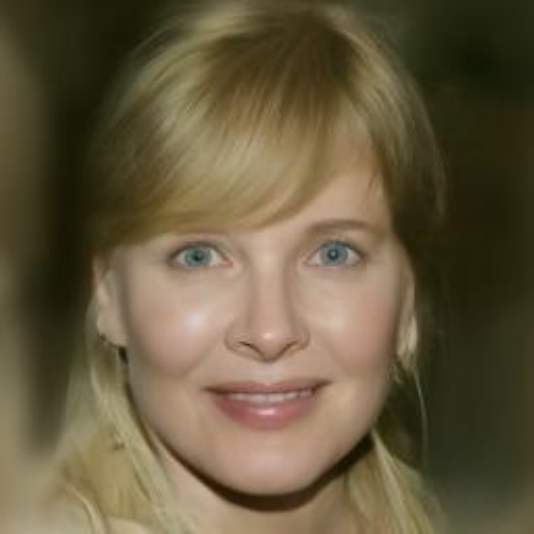}
\includegraphics[width=0.15\columnwidth]{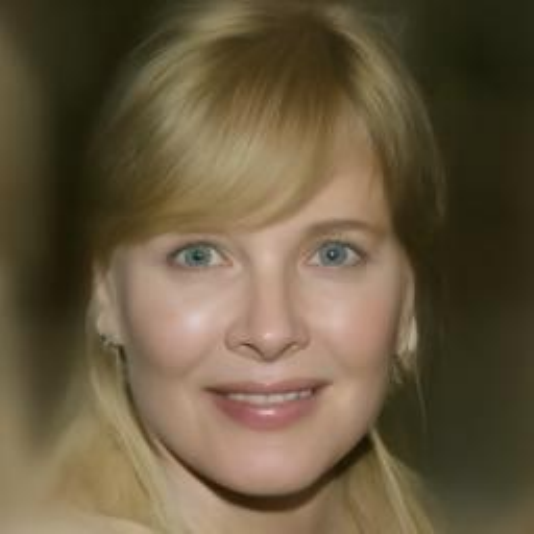}
\includegraphics[width=0.15\columnwidth]{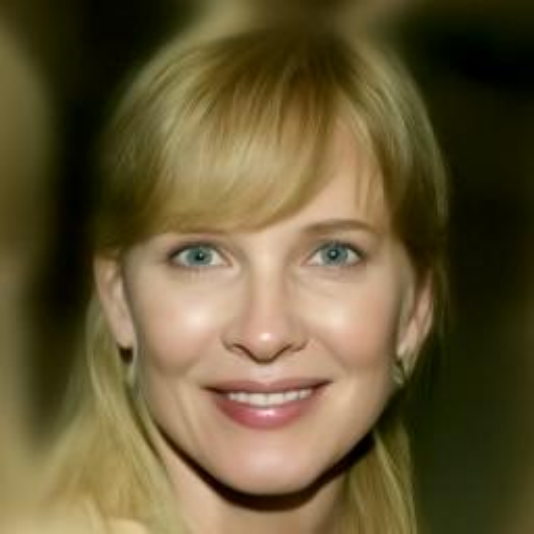}
\includegraphics[width=0.15\columnwidth]{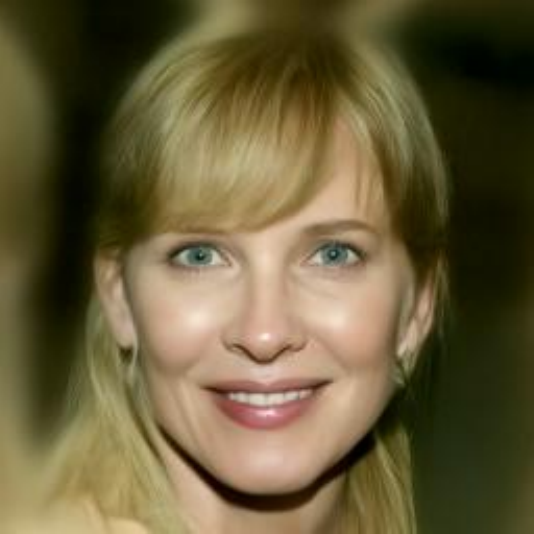}
\\
\includegraphics[width=0.15\columnwidth]{img_dif/DDPM_20_mix/b_453.pdf}
\includegraphics[width=0.15\columnwidth]{img_dif/DDPM_20_mix/a_453.pdf}
\includegraphics[width=0.15\columnwidth]{img_dif/DDPM_20_mix/m_135_453.pdf}
\includegraphics[width=0.15\columnwidth]{img_dif/DDPM_20_mix/m_155_453.pdf}
\\
\includegraphics[width=0.15\columnwidth]{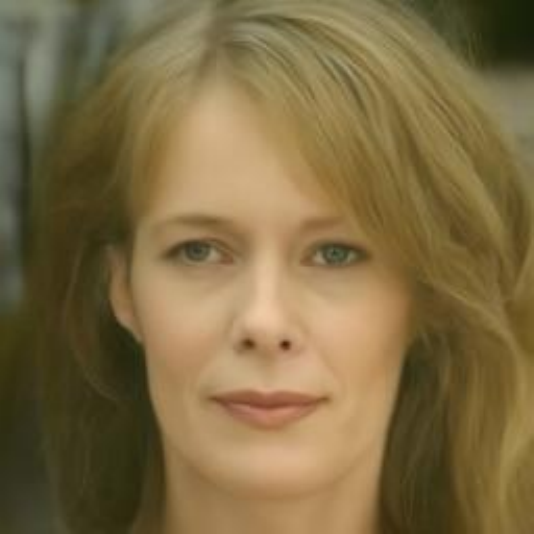}
\includegraphics[width=0.15\columnwidth]{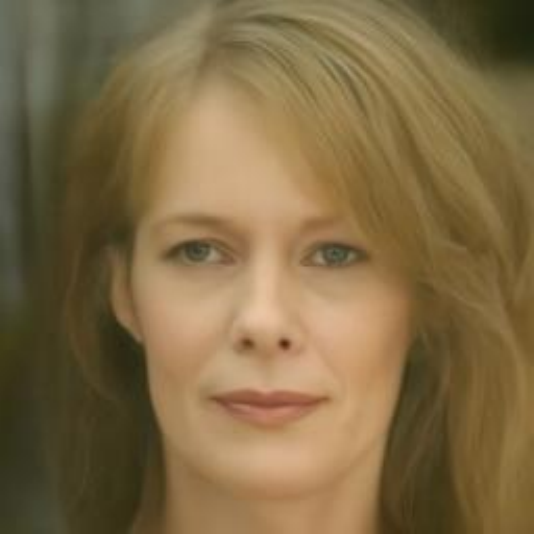}
\includegraphics[width=0.15\columnwidth]{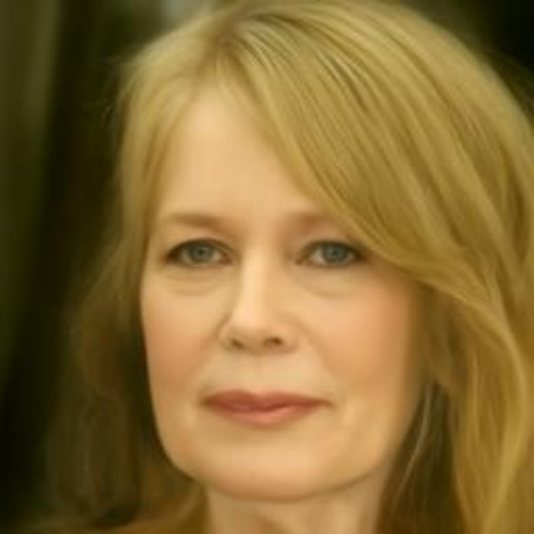}
\includegraphics[width=0.15\columnwidth]{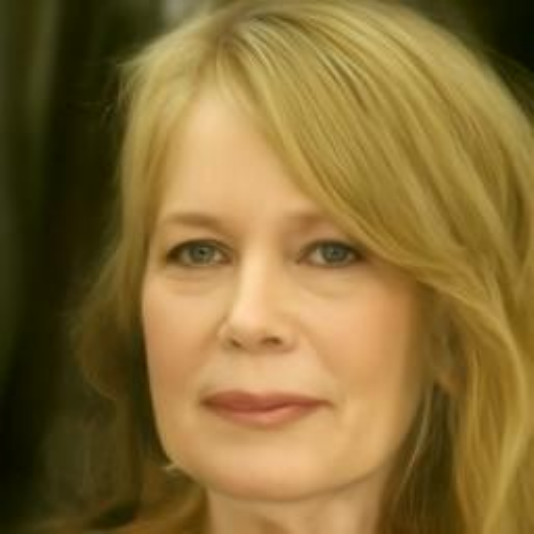}
\\
\includegraphics[width=0.15\columnwidth]{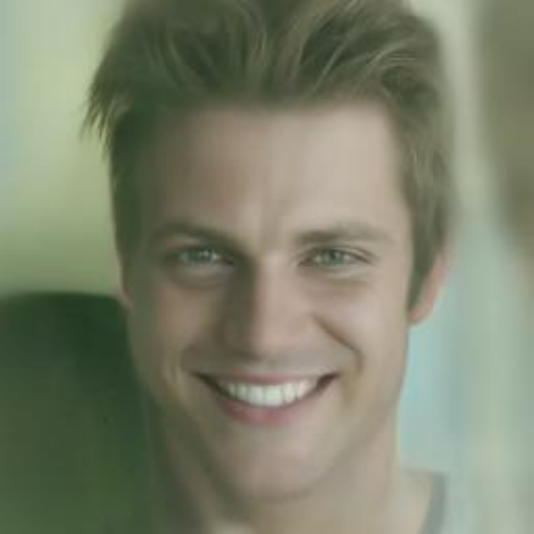}
\includegraphics[width=0.15\columnwidth]{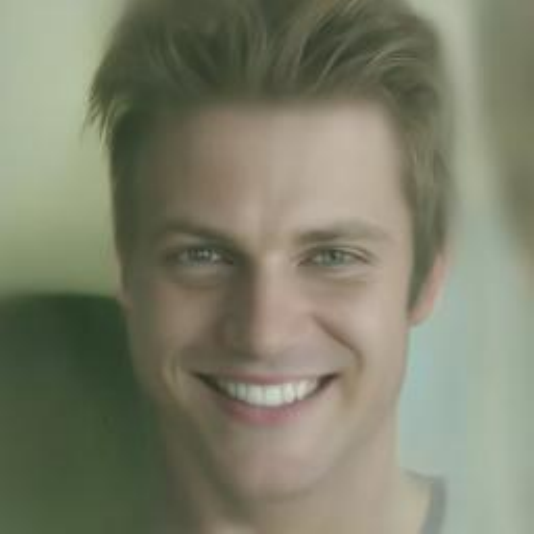}
\includegraphics[width=0.15\columnwidth]{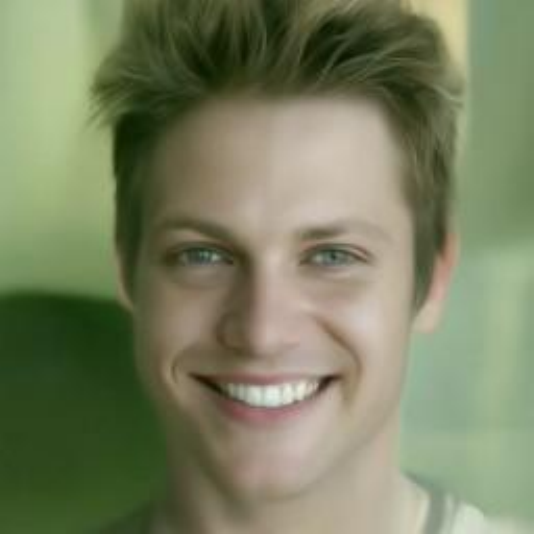}
\includegraphics[width=0.15\columnwidth]{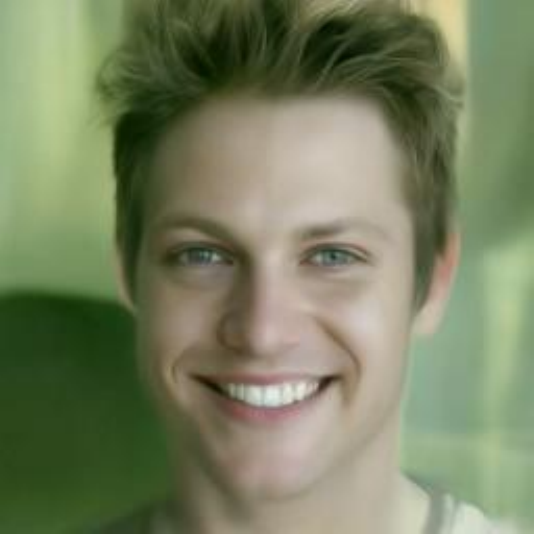}
\\
\includegraphics[width=0.15\columnwidth]{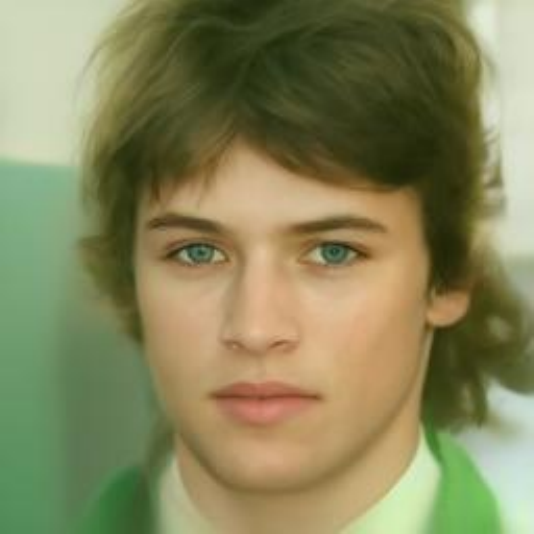}
\includegraphics[width=0.15\columnwidth]{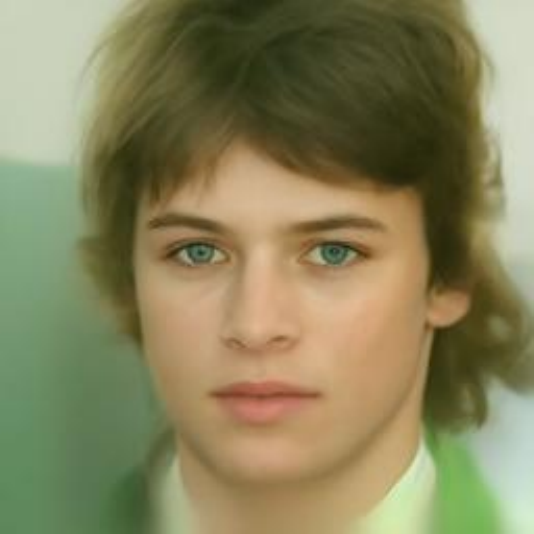}
\includegraphics[width=0.15\columnwidth]{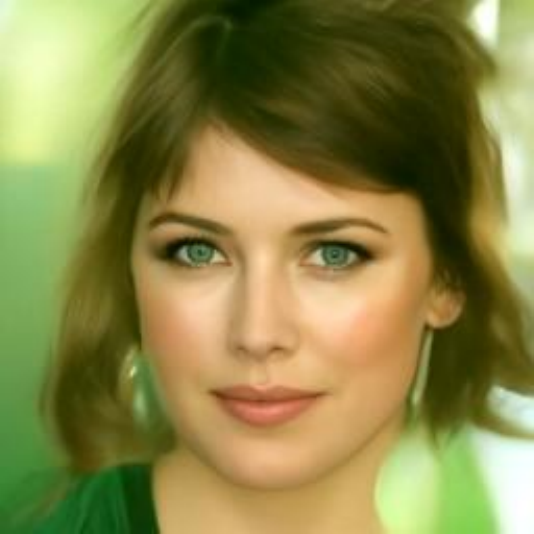}
\includegraphics[width=0.15\columnwidth]{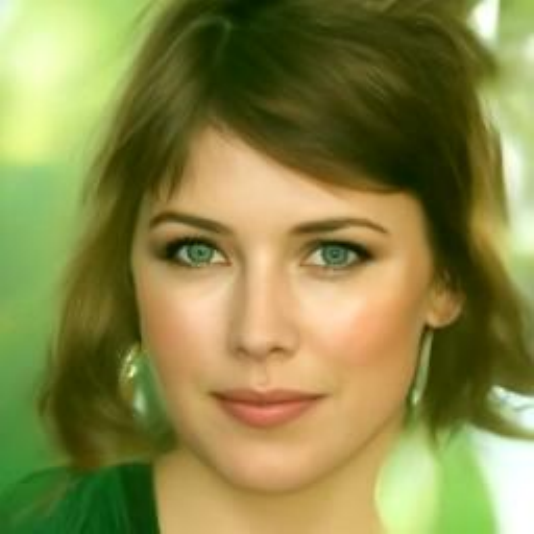}
\caption{\textbf{DDPM with $20$ Steps.} For each row, from left to right: baseline, ablation, generated image with variance scaling $1.35$, and one with $1.55$. 
%
%
%
}
    \label{fig:ddpm-20}
    \end{figure}

\begin{figure}[t]
    \centering
\includegraphics[width=0.15\columnwidth]{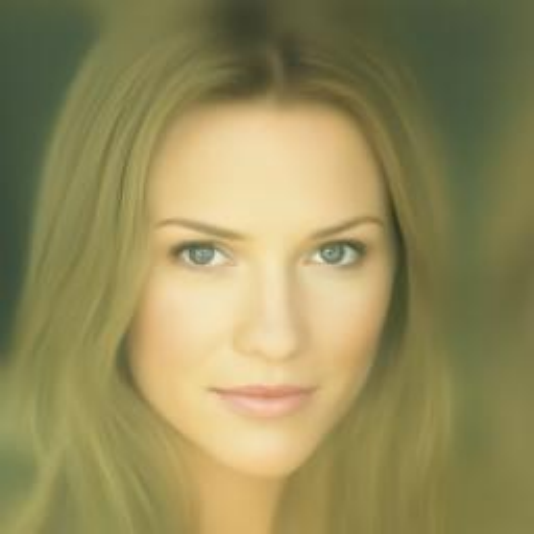}
\includegraphics[width=0.15\columnwidth]{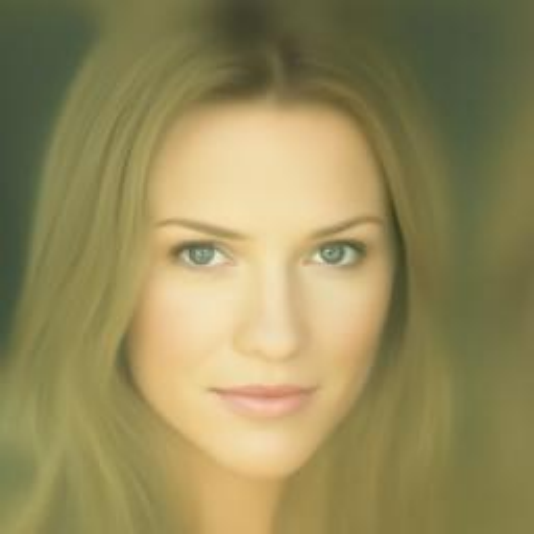}
\includegraphics[width=0.15\columnwidth]{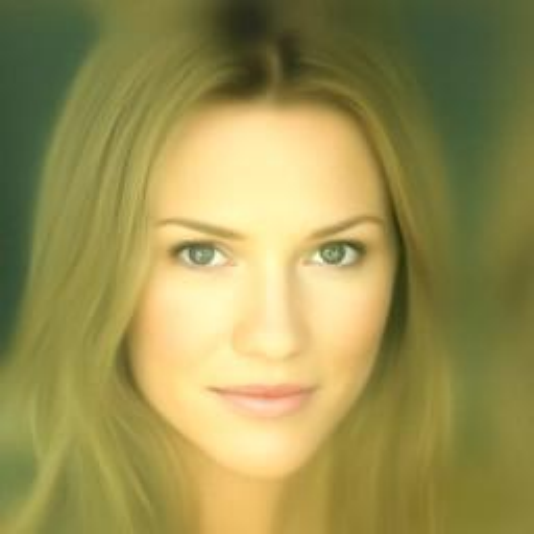}
\includegraphics[width=0.15\columnwidth]{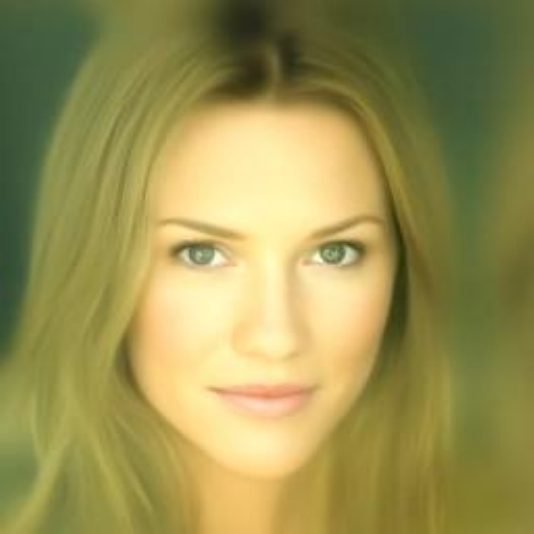}
\\
\includegraphics[width=0.15\columnwidth]{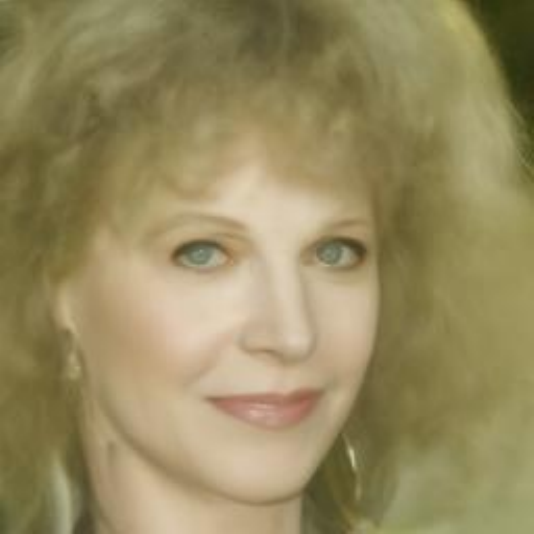}
\includegraphics[width=0.15\columnwidth]{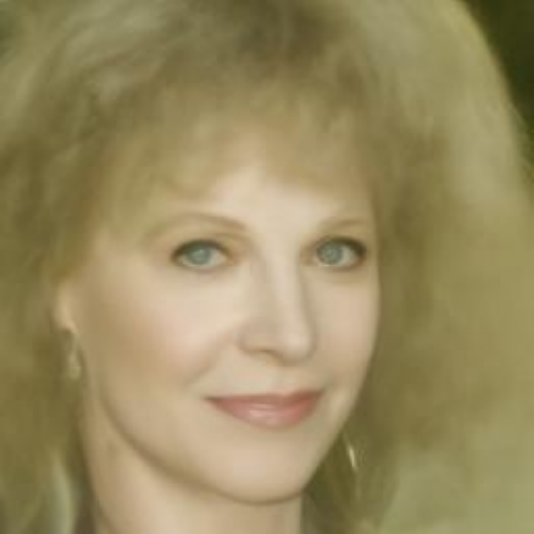}
\includegraphics[width=0.15\columnwidth]{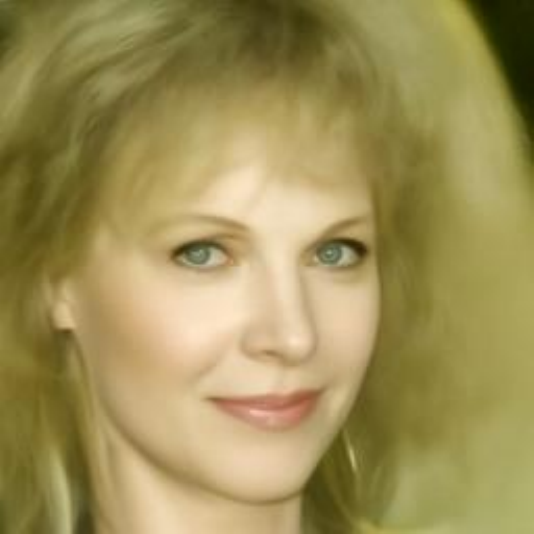}
\includegraphics[width=0.15\columnwidth]{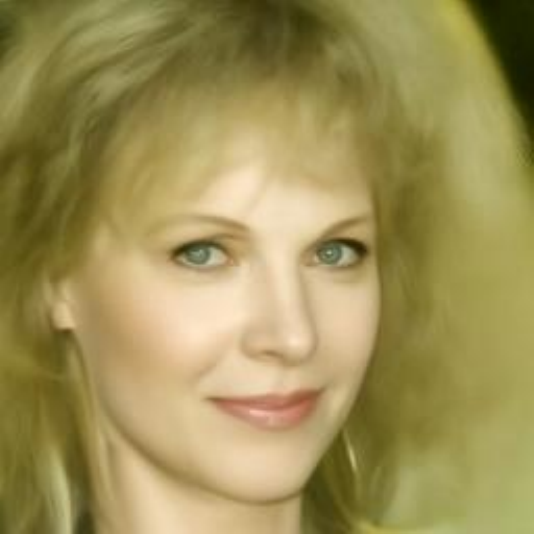}
\\
\includegraphics[width=0.15\columnwidth]{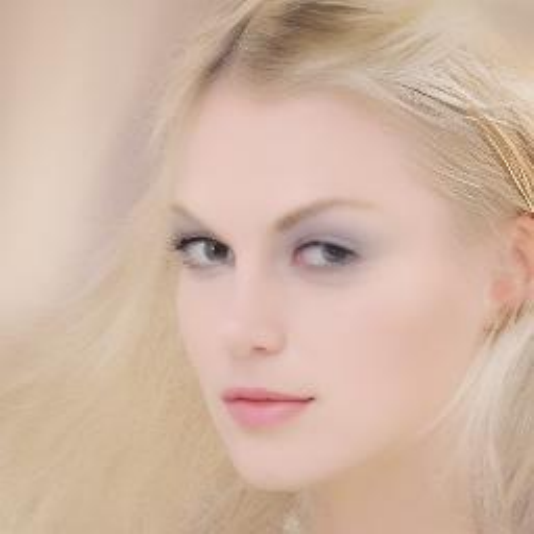}
\includegraphics[width=0.15\columnwidth]{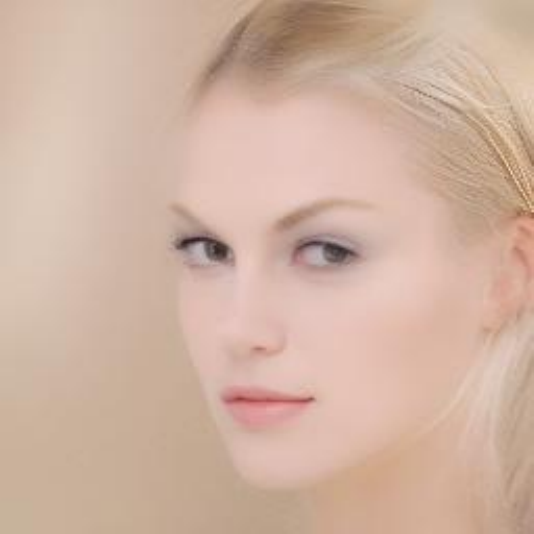}
\includegraphics[width=0.15\columnwidth]{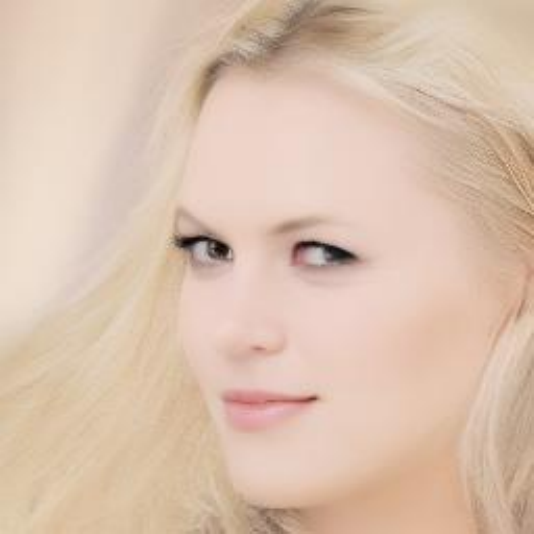}
\includegraphics[width=0.15\columnwidth]{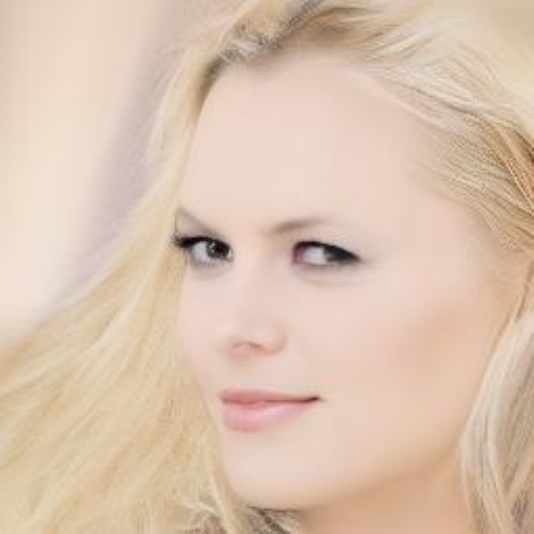}
\\
\includegraphics[width=0.15\columnwidth]{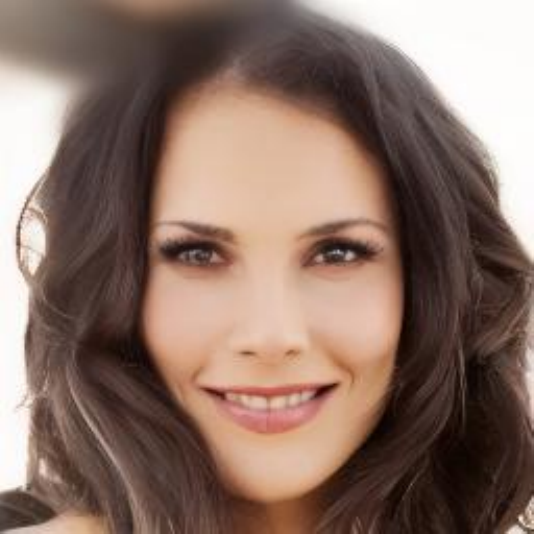}
\includegraphics[width=0.15\columnwidth]{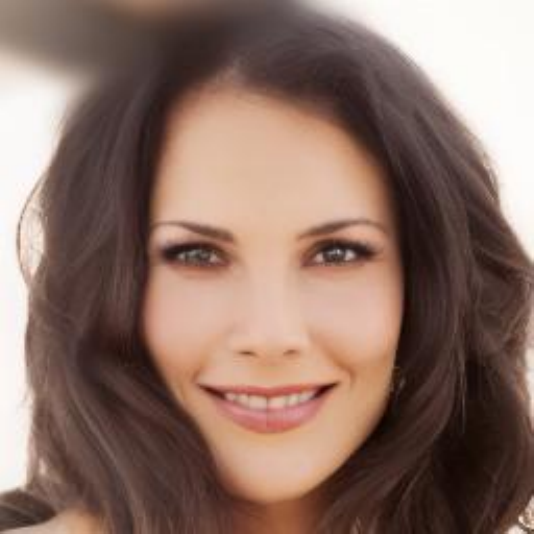}
\includegraphics[width=0.15\columnwidth]{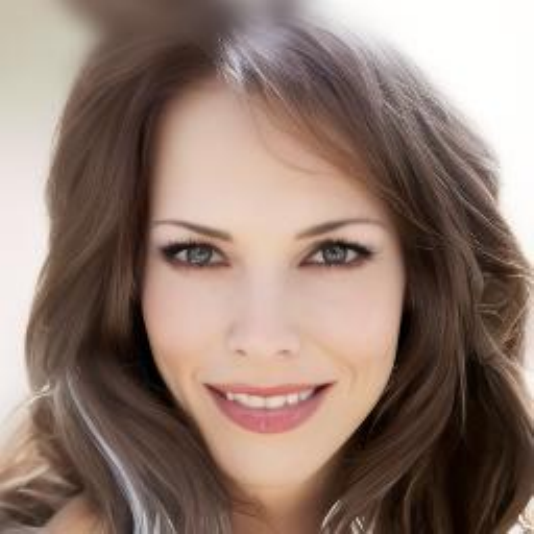}
\includegraphics[width=0.15\columnwidth]{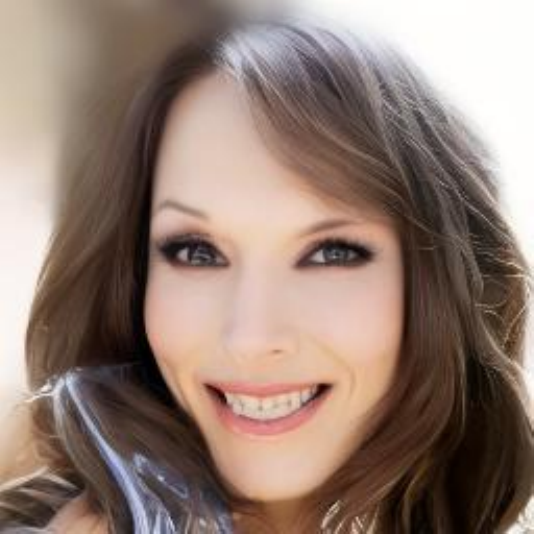}
\caption{\textbf{Examples of Shortcomings for DDPM and LD.} For each row, from left to right: baseline, ablation, generated image with variance scaling $\rho_1$, and one with variance scaling $\rho_2$. \textbf{Rows 1-2:} DDPM with $10$ and $20$ steps, respectively, with $\rho=1.35$ and $\rho_2=1.55$. 
\textbf{Row 3:} LD with $20$ steps with $\rho_1=1.35$ and $\rho_2=1.55$. \textbf{Row 4:} LD with $40$ steps with $\rho_1=1.20$ and $\rho_2=1.35$.
}
    \label{fig:shortcoming}
    \end{figure}

\begin{figure}[t]
    \centering
\includegraphics[width=0.15\columnwidth]{img_dif/DDPM_100_SEMAN/b_87.pdf}
\includegraphics[width=0.15\columnwidth]{img_dif/DDPM_100_SEMAN/a_87.pdf}
\includegraphics[width=0.15\columnwidth]{img_dif/DDPM_100_SEMAN/m_25_87.pdf}
\includegraphics[width=0.15\columnwidth]{img_dif/DDPM_100_SEMAN/m_50_87.pdf}
\\
\includegraphics[width=0.15\columnwidth]{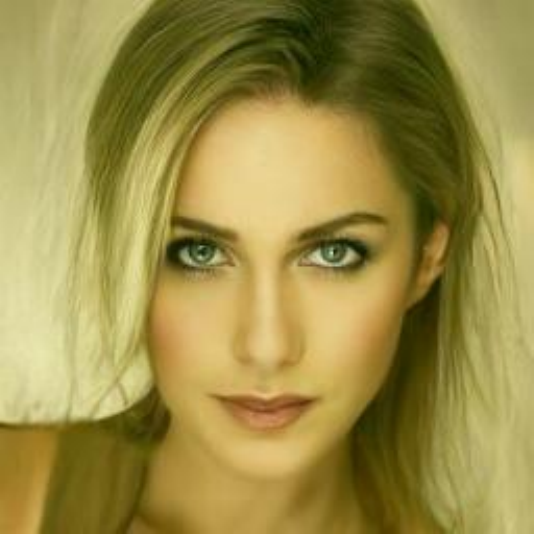}
\includegraphics[width=0.15\columnwidth]{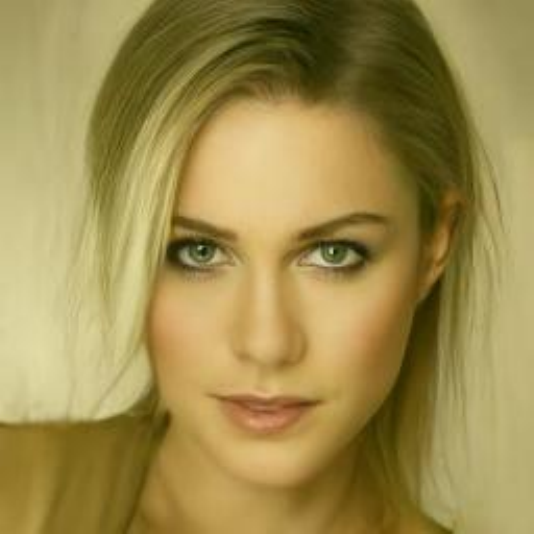}
\includegraphics[width=0.15\columnwidth]{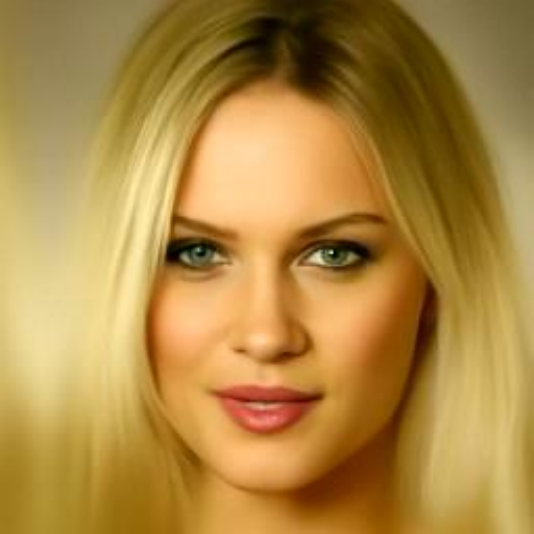}
\includegraphics[width=0.15\columnwidth]{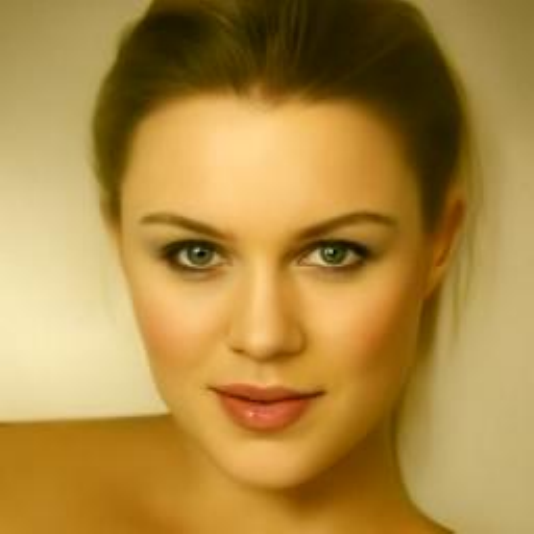}
\\
\includegraphics[width=0.15\columnwidth]{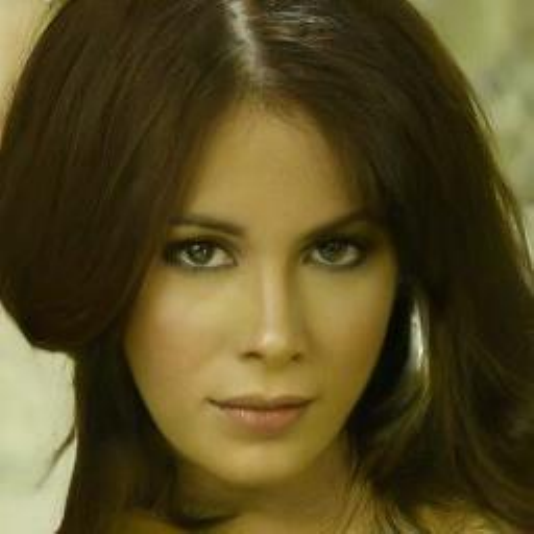}
\includegraphics[width=0.15\columnwidth]{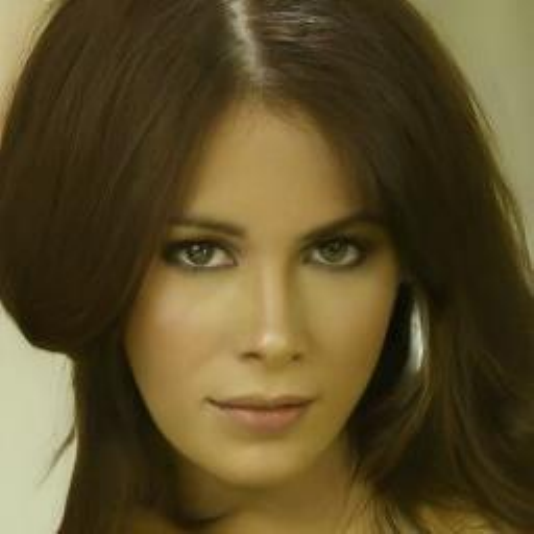}
\includegraphics[width=0.15\columnwidth]{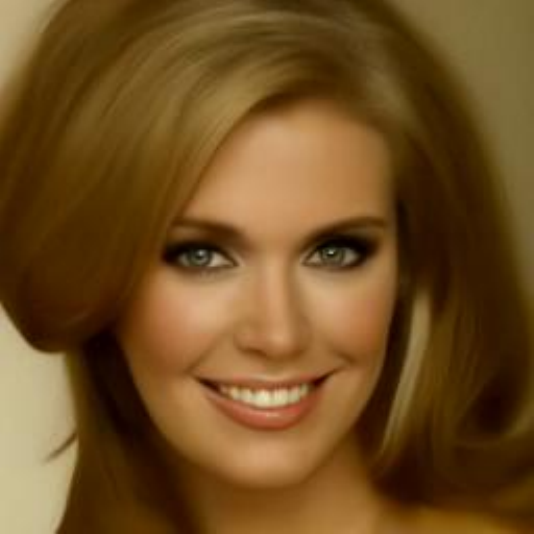}
\includegraphics[width=0.15\columnwidth]{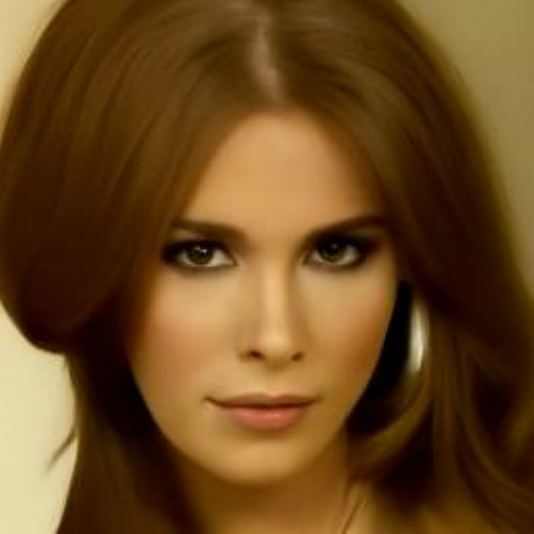}
\\
\includegraphics[width=0.15\columnwidth]{img_dif/DDPM_100_SEMAN/b_79.pdf}
\includegraphics[width=0.15\columnwidth]{img_dif/DDPM_100_SEMAN/a_79.pdf}
\includegraphics[width=0.15\columnwidth]{img_dif/DDPM_100_SEMAN/m_25_79.pdf}
\includegraphics[width=0.15\columnwidth]{img_dif/DDPM_100_SEMAN/m_50_79.pdf}
\\
\includegraphics[width=0.15\columnwidth]{img_dif/DDPM_100_SEMAN/b_208.pdf}
\includegraphics[width=0.15\columnwidth]{img_dif/DDPM_100_SEMAN/a_208.pdf}
\includegraphics[width=0.15\columnwidth]{img_dif/DDPM_100_SEMAN/m_25_208.pdf}
\includegraphics[width=0.15\columnwidth]{img_dif/DDPM_100_SEMAN/m_50_208.pdf}
\\
\includegraphics[width=0.15\columnwidth]{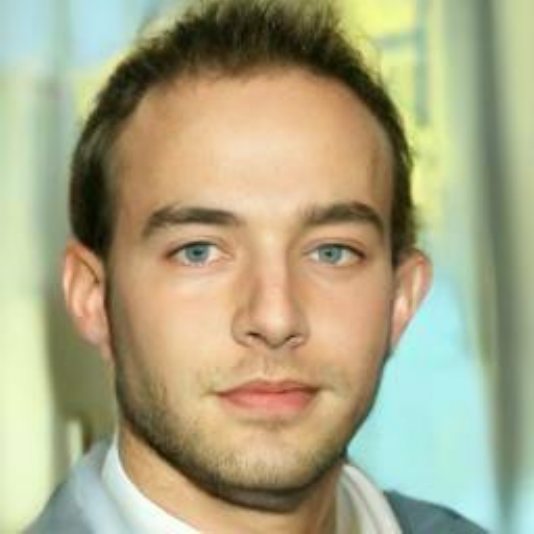}
\includegraphics[width=0.15\columnwidth]{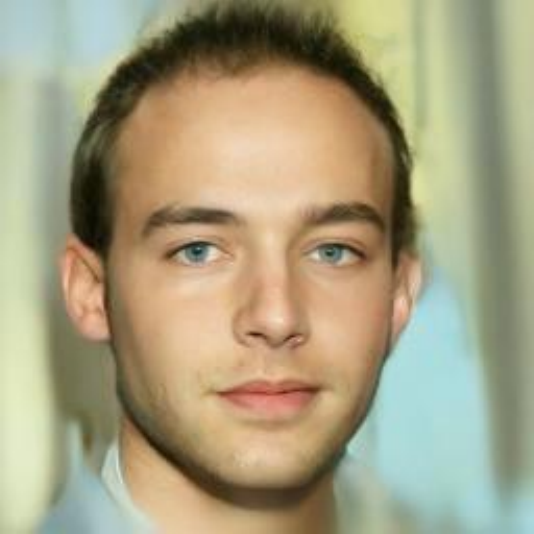}
\includegraphics[width=0.15\columnwidth]{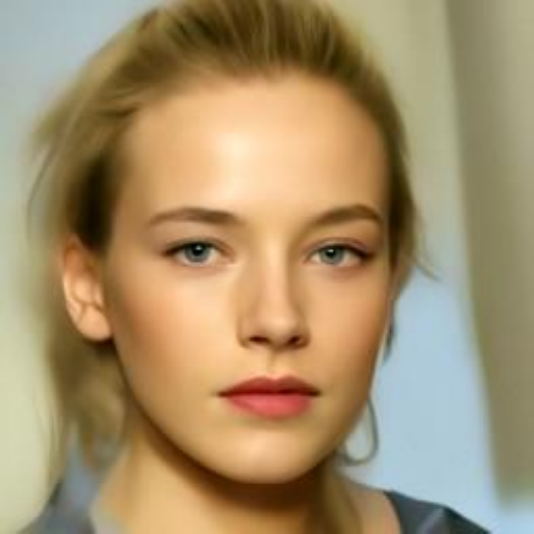}
\includegraphics[width=0.15\columnwidth]{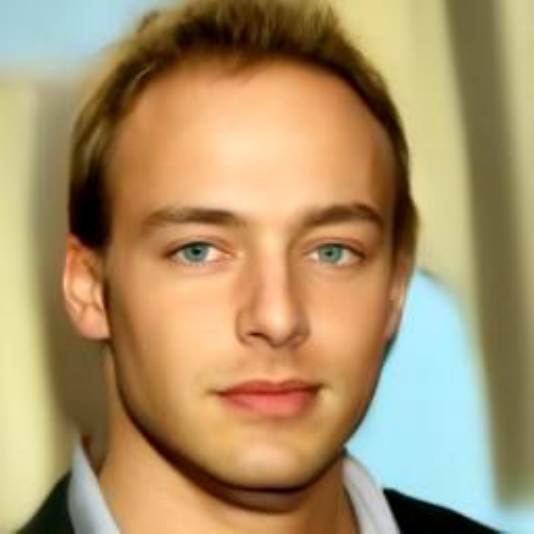}
\\
\includegraphics[width=0.15\columnwidth]{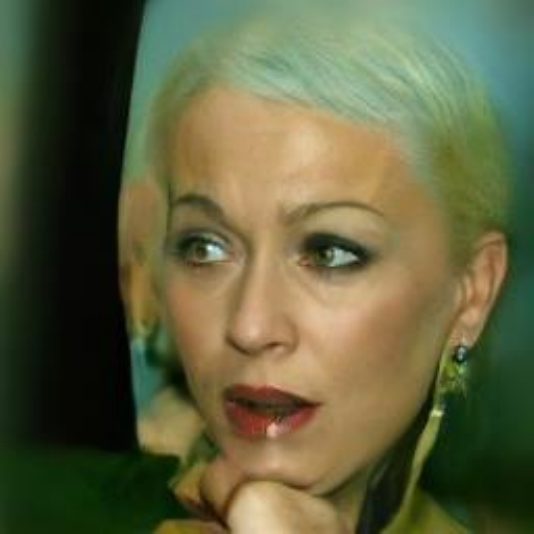}
\includegraphics[width=0.15\columnwidth]{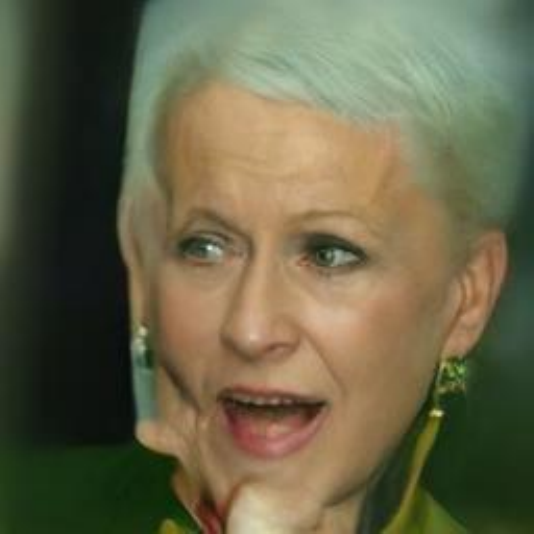}
\includegraphics[width=0.15\columnwidth]{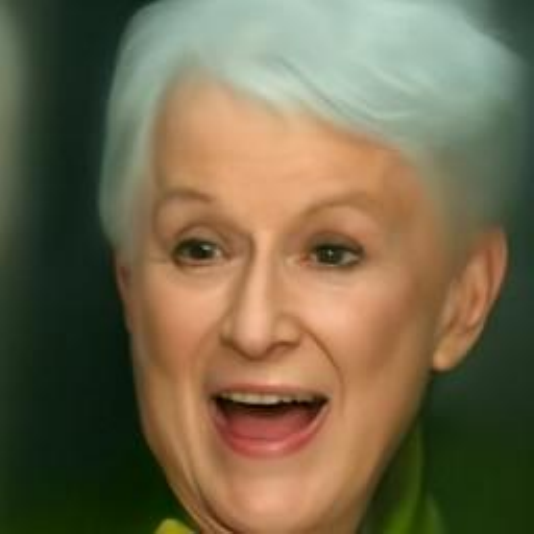}
\includegraphics[width=0.15\columnwidth]{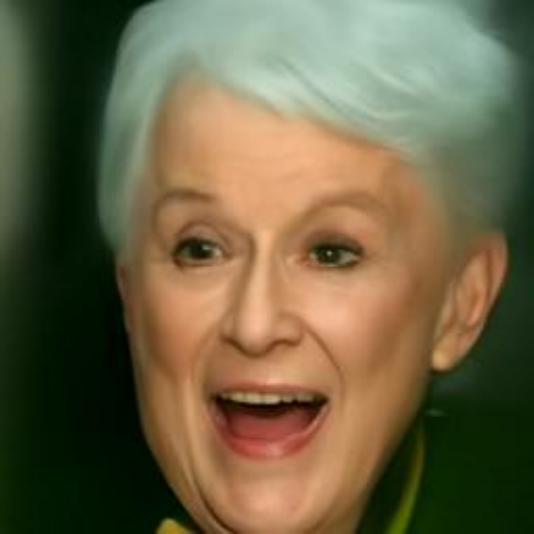}
\\
\includegraphics[width=0.15\columnwidth]{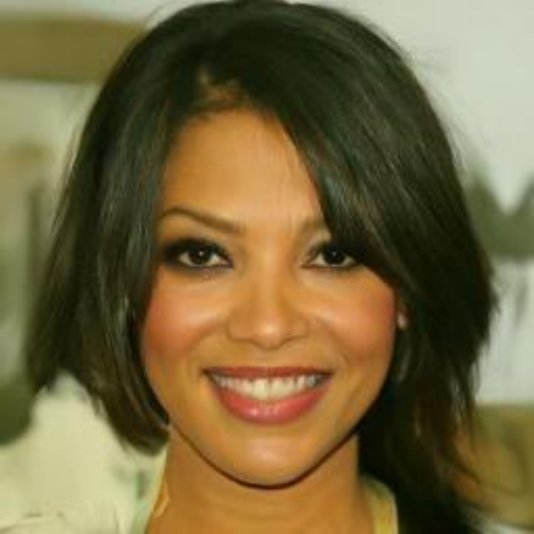}
\includegraphics[width=0.15\columnwidth]{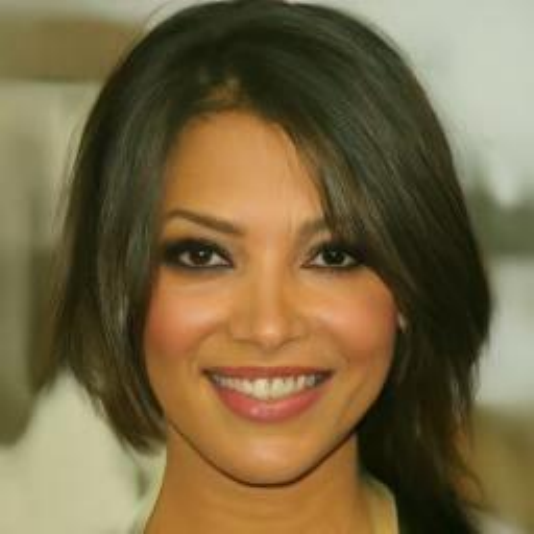}
\includegraphics[width=0.15\columnwidth]{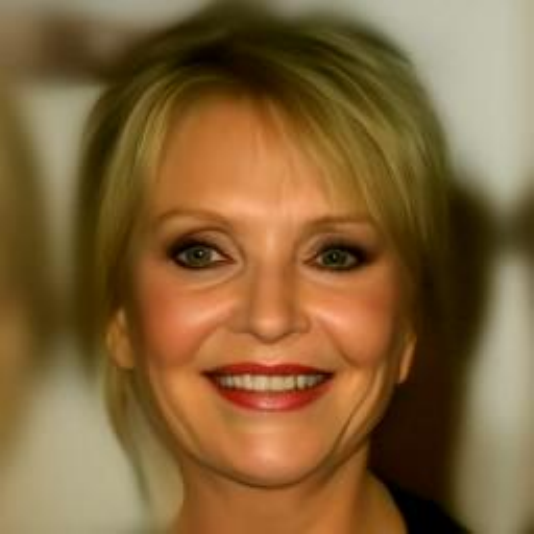}
\includegraphics[width=0.15\columnwidth]{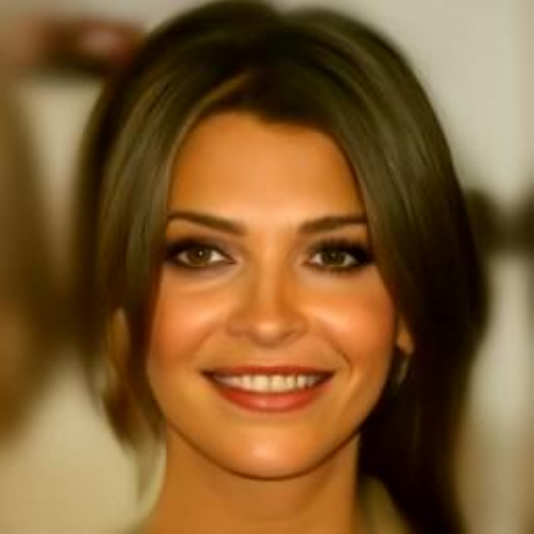}
\caption{\textbf{DDPM with $100$ Steps.} 
For each row, from left to right: baseline, ablation, generated image with variance scaling $0.25$, and one with $0.50$.
}
    \label{fig:DDPM-100-semantic}
    \end{figure}

\begin{figure}[t]
    \centering
\includegraphics[width=0.15\columnwidth]{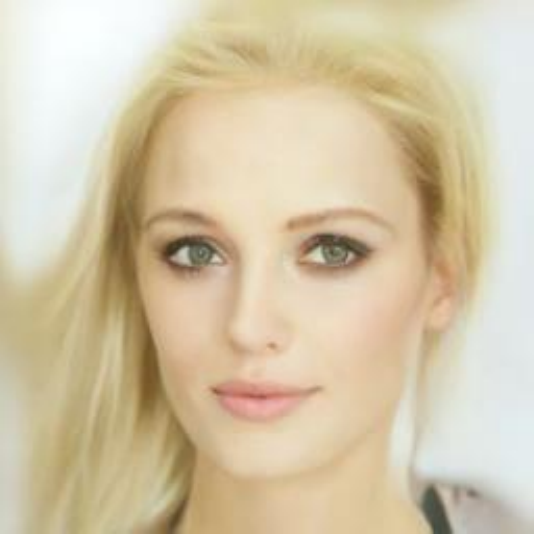}
\includegraphics[width=0.15\columnwidth]{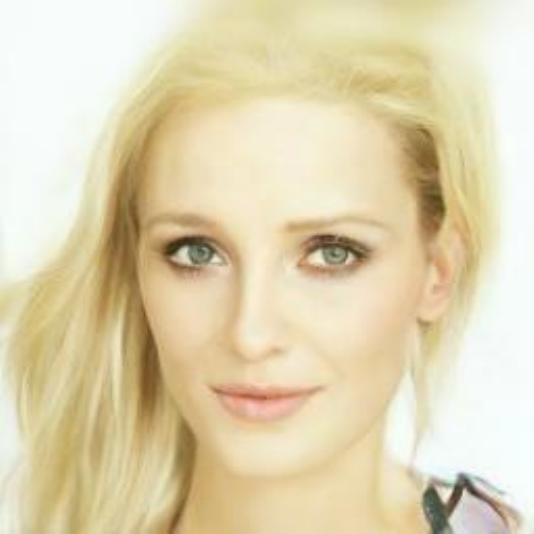}
\includegraphics[width=0.15\columnwidth]{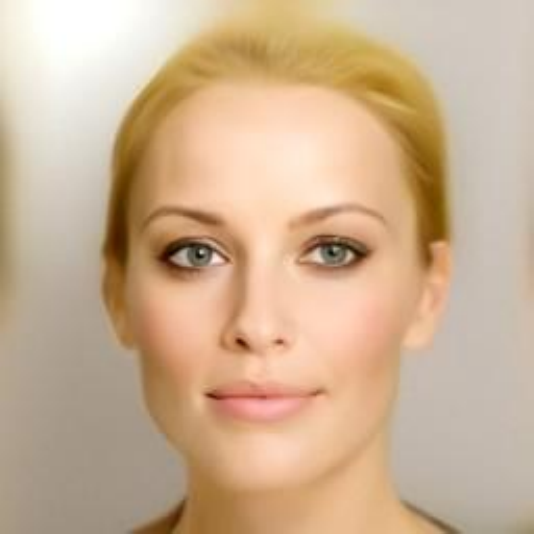}
\includegraphics[width=0.15\columnwidth]{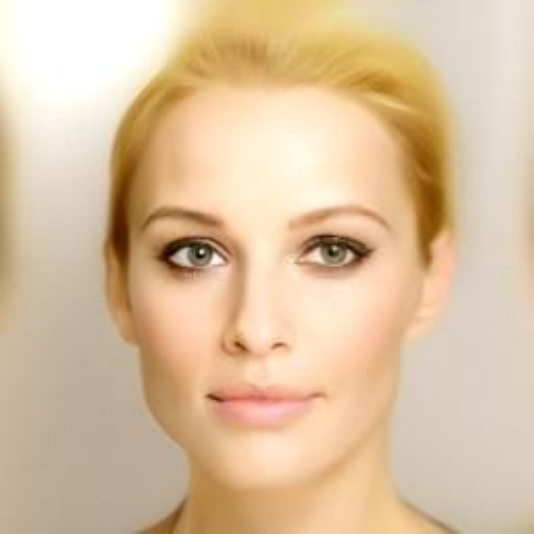}
\\
\includegraphics[width=0.15\columnwidth]{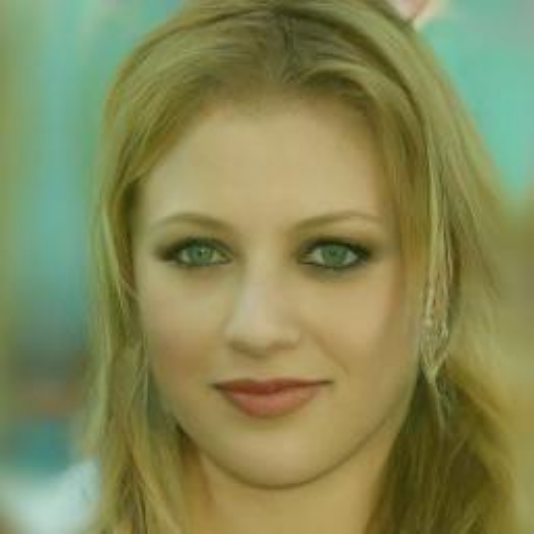}
\includegraphics[width=0.15\columnwidth]{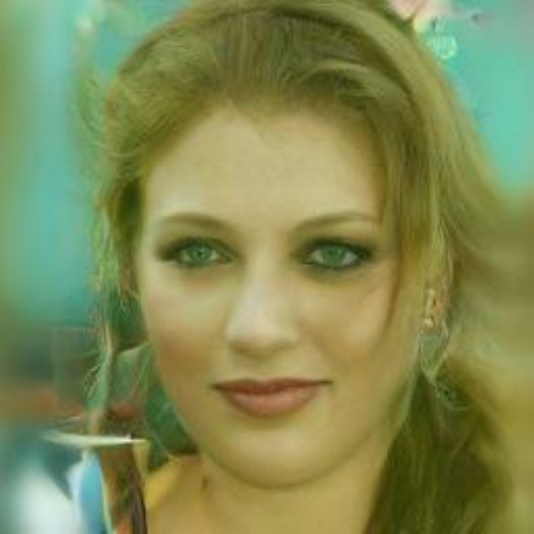}
\includegraphics[width=0.15\columnwidth]{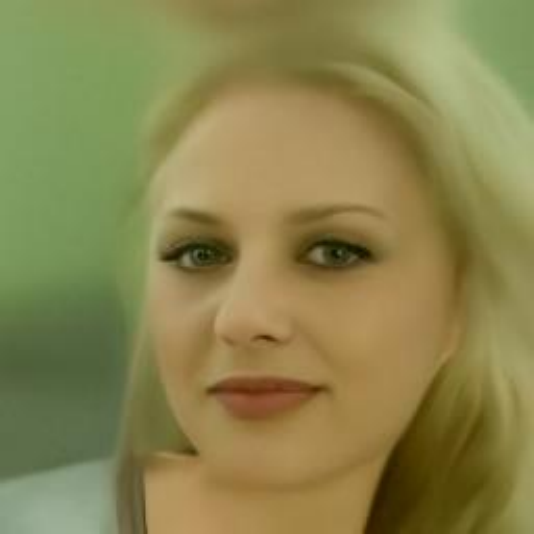}
\includegraphics[width=0.15\columnwidth]{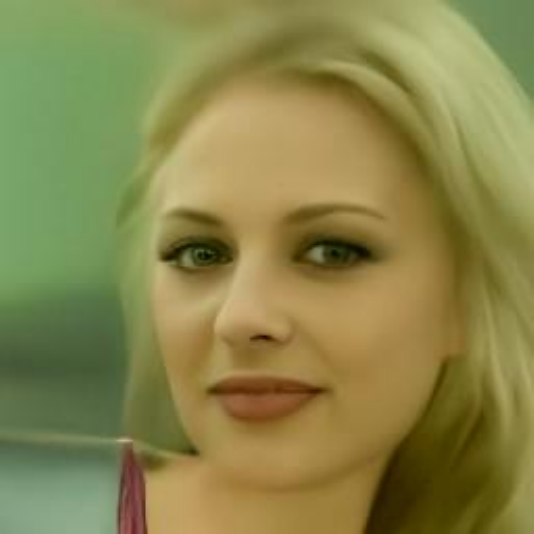}
\caption{\textbf{Examples of No Overexposure and Less Brightness in DDPM with $100$ Steps.} 
Left to right: baseline, ablation, generated image with variance scaling $0.25$, and one with $0.50$.
%
%
}
    \label{fig:DDPM-100-contrast}
    \end{figure}

\begin{figure}[t]
    \centering
\includegraphics[width=0.15\columnwidth]{img_dif/LD_20/b_1.pdf}
\includegraphics[width=0.15\columnwidth]{img_dif/LD_20/a_1.pdf}
\includegraphics[width=0.15\columnwidth]{img_dif/LD_20/m_135_1.pdf}
\includegraphics[width=0.15\columnwidth]{img_dif/LD_20/m_155_1.pdf}
\\
\includegraphics[width=0.15\columnwidth]{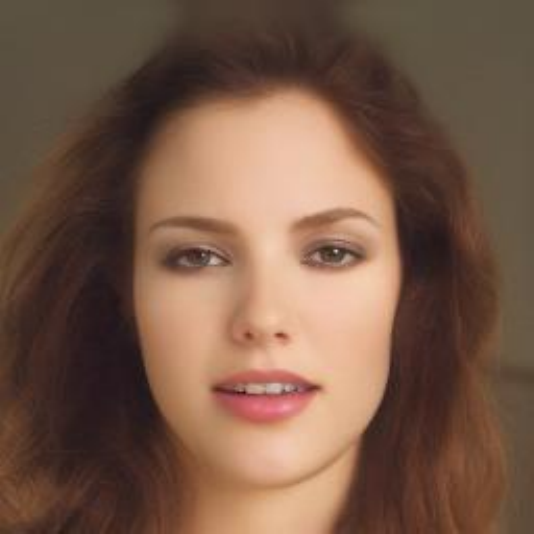}
\includegraphics[width=0.15\columnwidth]{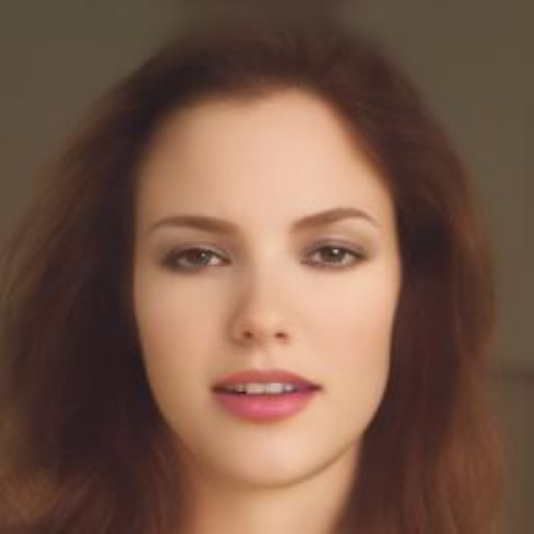}
\includegraphics[width=0.15\columnwidth]{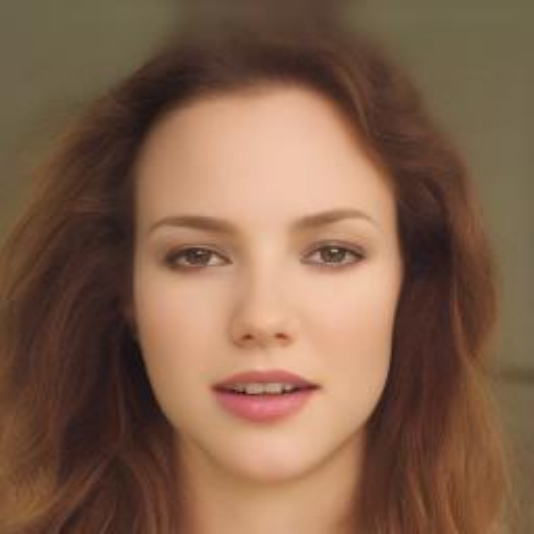}
\includegraphics[width=0.15\columnwidth]{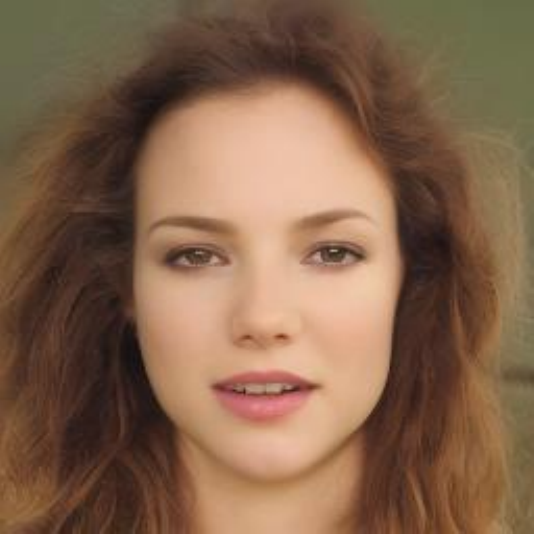}
\\
\includegraphics[width=0.15\columnwidth]{img_dif/LD_20/b_8.pdf}
\includegraphics[width=0.15\columnwidth]{img_dif/LD_20/a_8.pdf}
\includegraphics[width=0.15\columnwidth]{img_dif/LD_20/m_135_8.pdf}
\includegraphics[width=0.15\columnwidth]{img_dif/LD_20/m_155_8.pdf}
\\
\includegraphics[width=0.15\columnwidth]{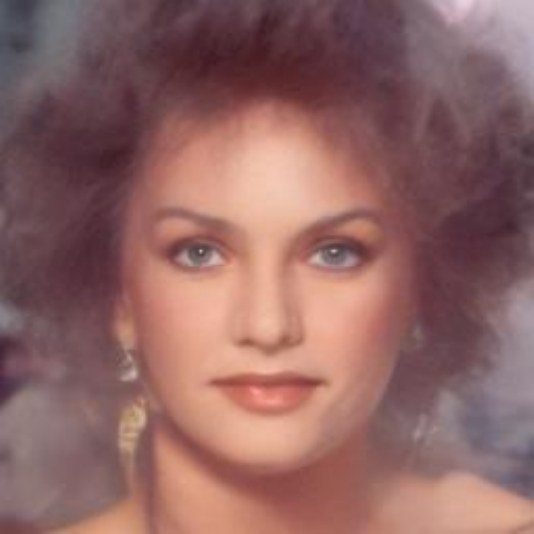}
\includegraphics[width=0.15\columnwidth]{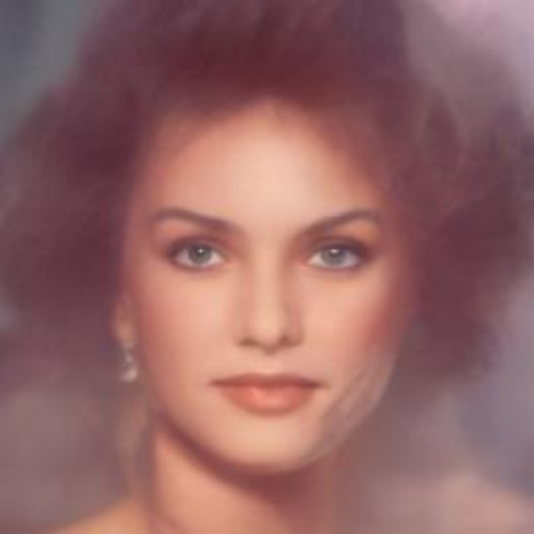}
\includegraphics[width=0.15\columnwidth]{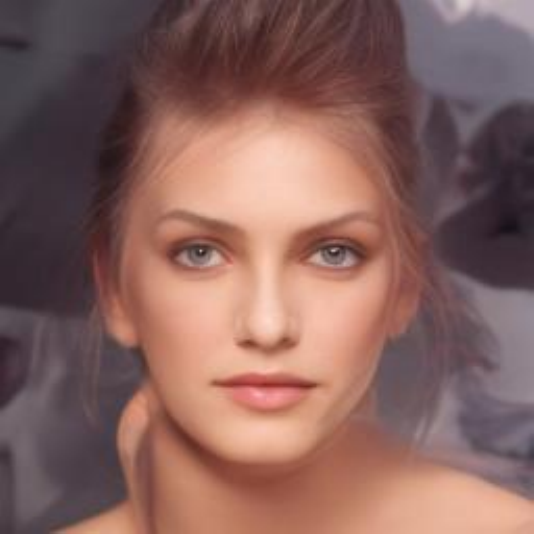}
\includegraphics[width=0.15\columnwidth]{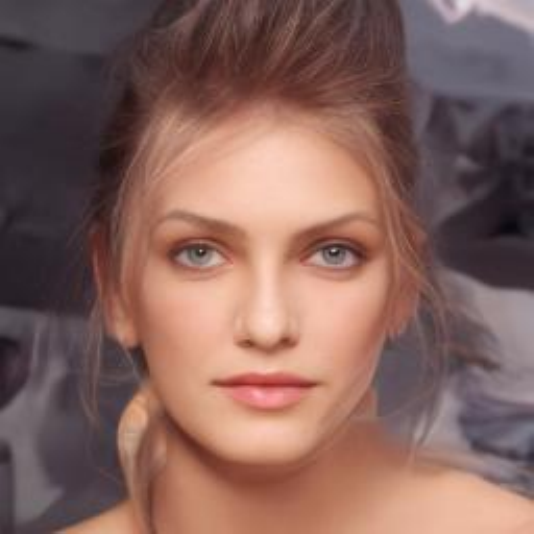}
\\
\includegraphics[width=0.15\columnwidth]{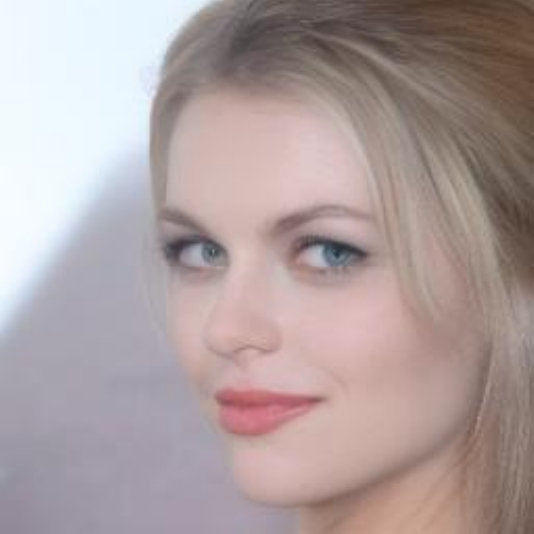}
\includegraphics[width=0.15\columnwidth]{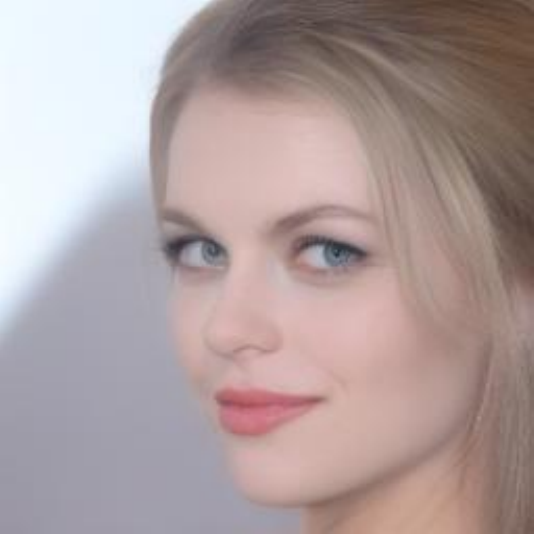}
\includegraphics[width=0.15\columnwidth]{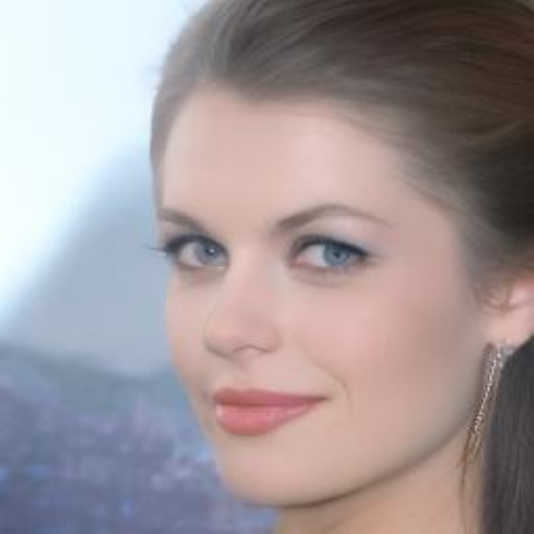}
\includegraphics[width=0.15\columnwidth]{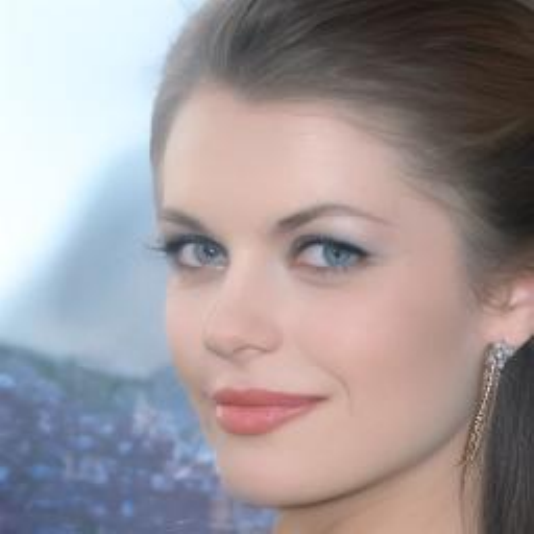}
\\
\includegraphics[width=0.15\columnwidth]{img_dif/LD_20/b_152.pdf}
\includegraphics[width=0.15\columnwidth]{img_dif/LD_20/a_152.pdf}
\includegraphics[width=0.15\columnwidth]{img_dif/LD_20/m_135_152.pdf}
\includegraphics[width=0.15\columnwidth]{img_dif/LD_20/m_155_152.pdf}
\\
\includegraphics[width=0.15\columnwidth]{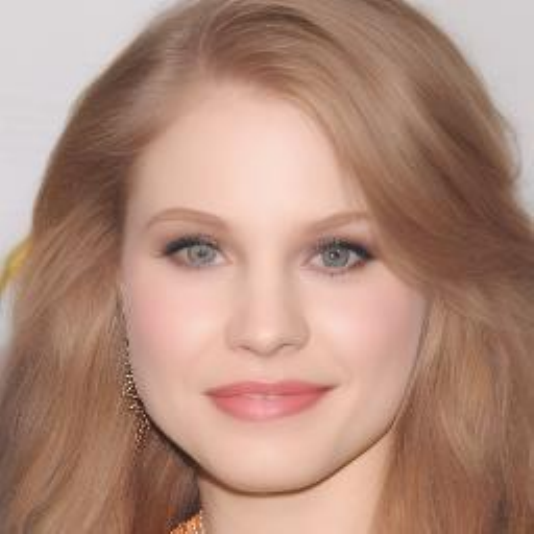}
\includegraphics[width=0.15\columnwidth]{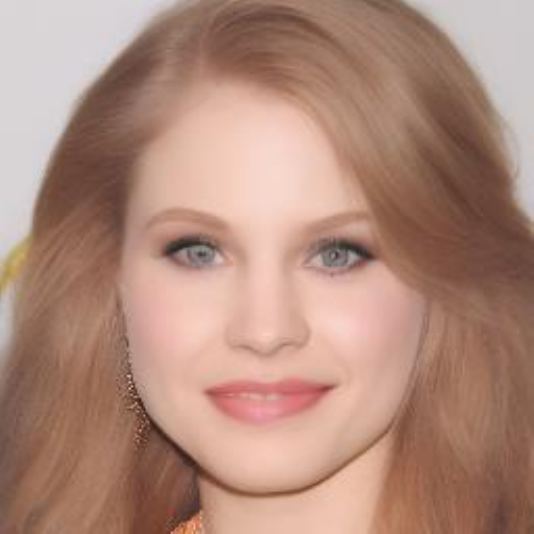}
\includegraphics[width=0.15\columnwidth]{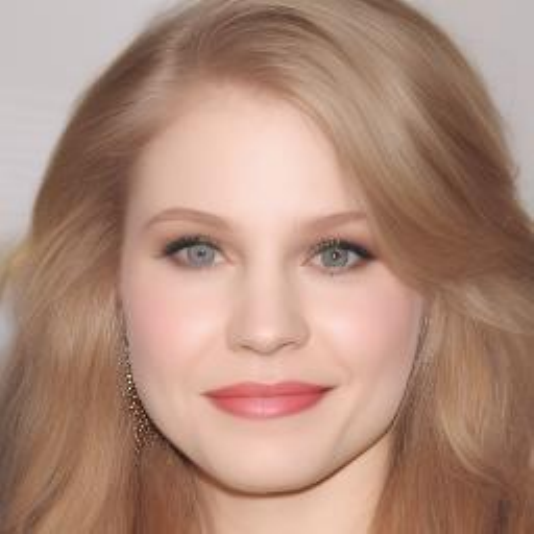}
\includegraphics[width=0.15\columnwidth]{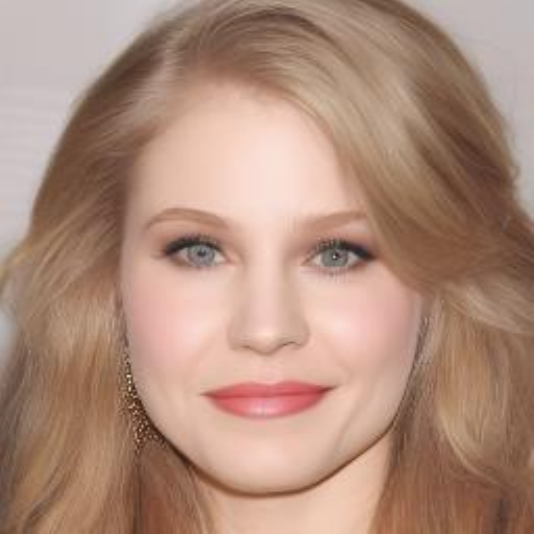}
\\
\includegraphics[width=0.15\columnwidth]{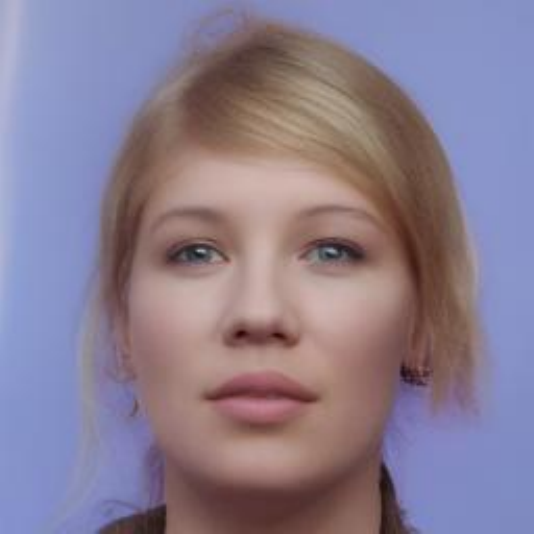}
\includegraphics[width=0.15\columnwidth]{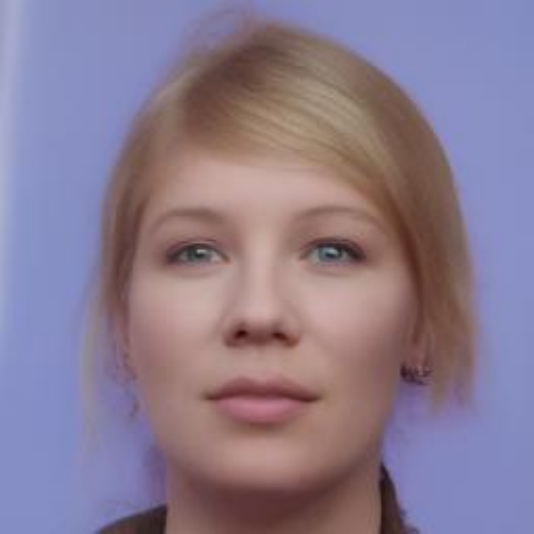}
\includegraphics[width=0.15\columnwidth]{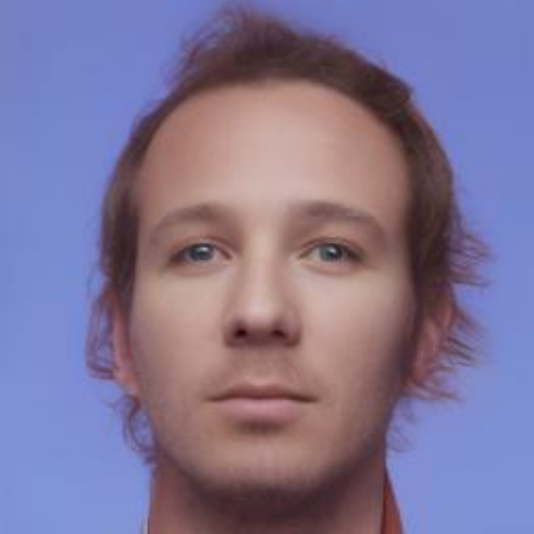}
\includegraphics[width=0.15\columnwidth]{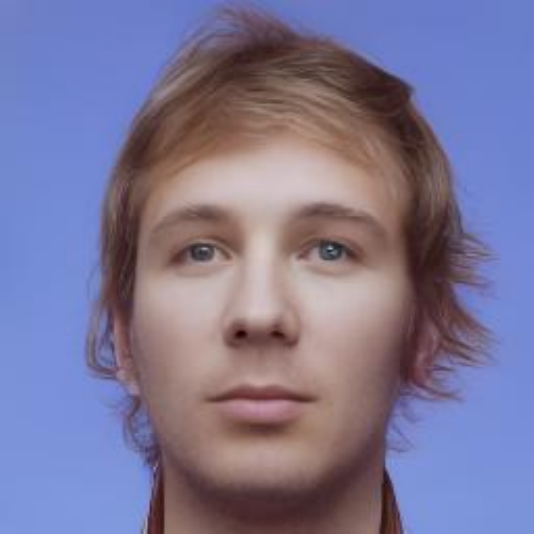}
\caption{\textbf{LD with $20$ Steps.} 
From left to right: baseline, ablation, generated image with variance scaling $1.35$, and one with $1.55$.
%
%
%
}
    \label{fig:ld-20}
    \end{figure}

\begin{figure}[t]
    \centering
\includegraphics[width=0.15\columnwidth]{img_dif/LD_40/b_11.pdf}
\includegraphics[width=0.15\columnwidth]{img_dif/LD_40/a_11.pdf}
\includegraphics[width=0.15\columnwidth]{img_dif/LD_40/m_120_11.pdf}
\includegraphics[width=0.15\columnwidth]{img_dif/LD_40/m_135_11.pdf}
\\
\includegraphics[width=0.15\columnwidth]{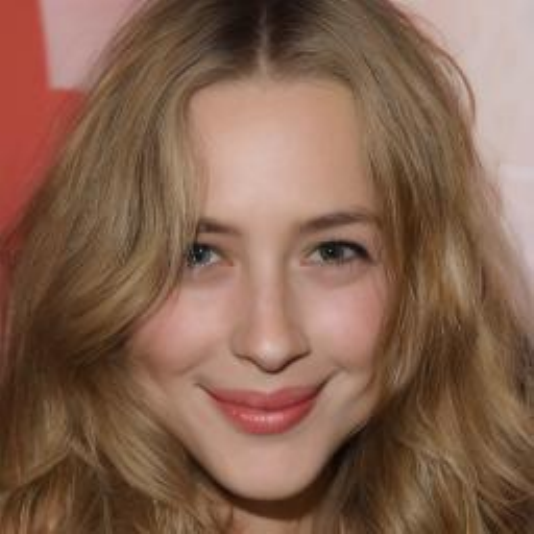}
\includegraphics[width=0.15\columnwidth]{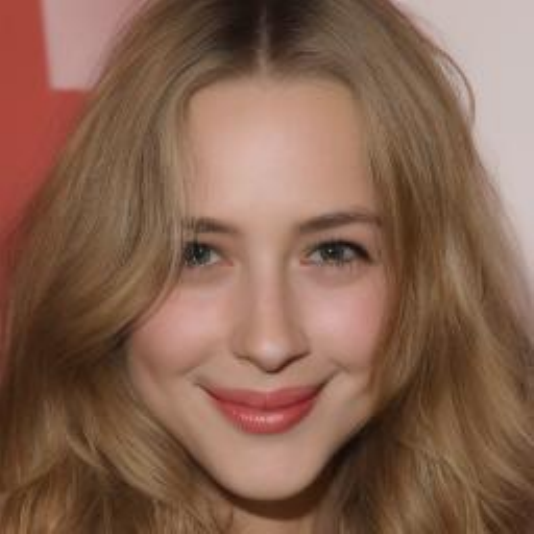}
\includegraphics[width=0.15\columnwidth]{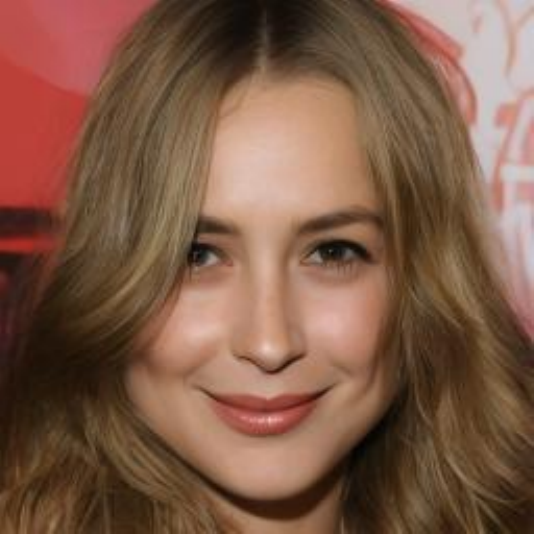}
\includegraphics[width=0.15\columnwidth]{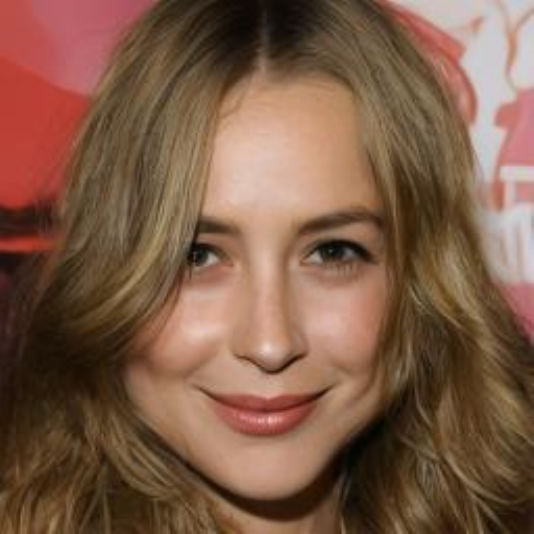}
\\
\includegraphics[width=0.15\columnwidth]{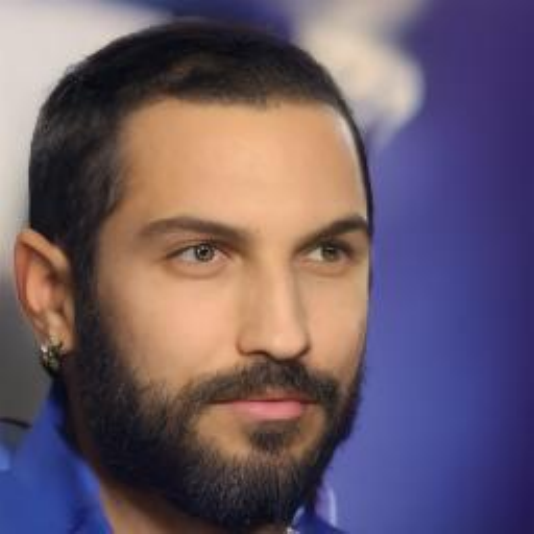}
\includegraphics[width=0.15\columnwidth]{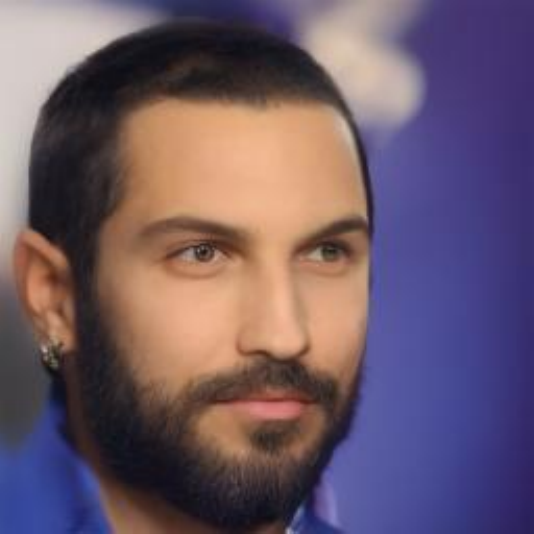}
\includegraphics[width=0.15\columnwidth]{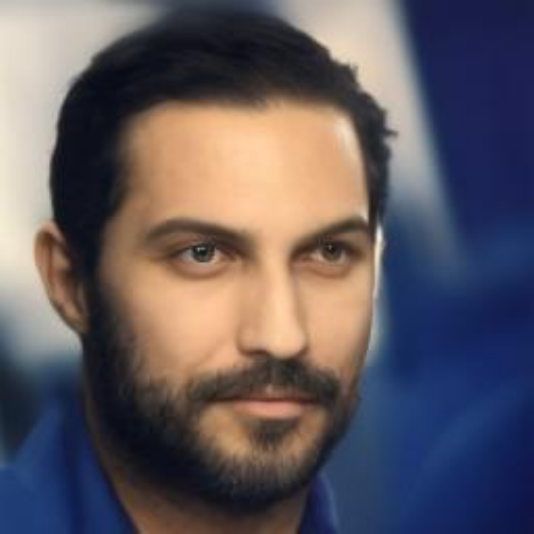}
\includegraphics[width=0.15\columnwidth]{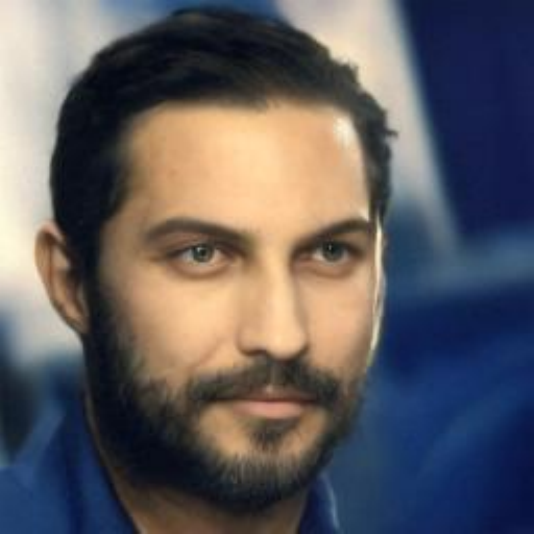}
\\
\includegraphics[width=0.15\columnwidth]{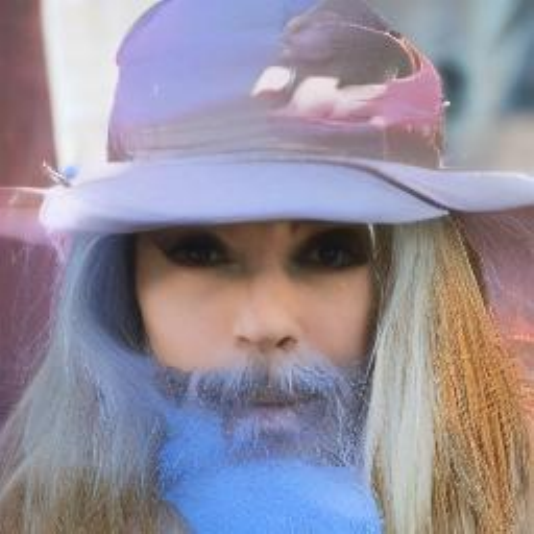}
\includegraphics[width=0.15\columnwidth]{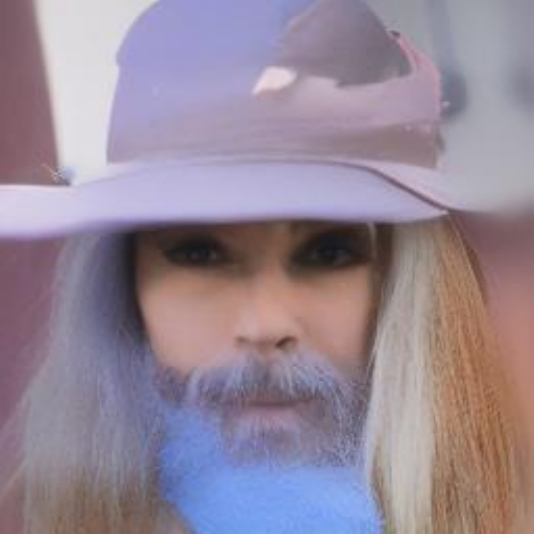}
\includegraphics[width=0.15\columnwidth]{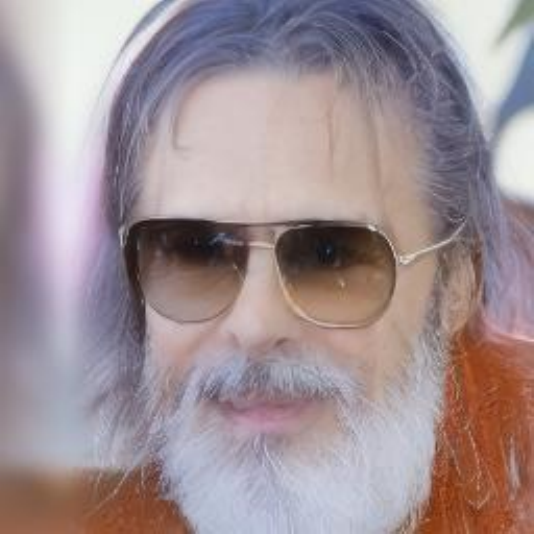}
\includegraphics[width=0.15\columnwidth]{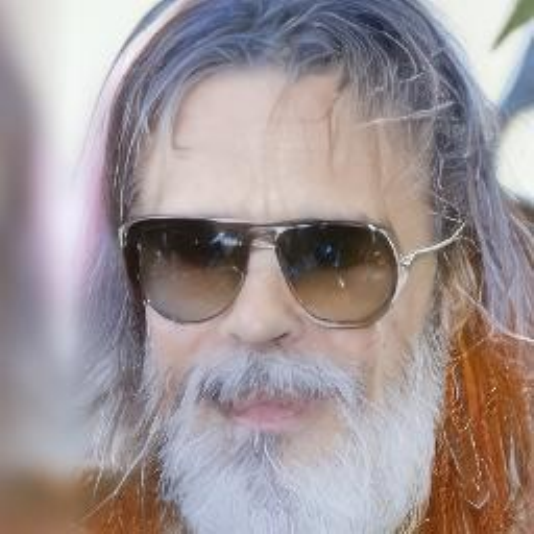}
\\
\includegraphics[width=0.15\columnwidth]{img_dif/LD_40/b_105.pdf}
\includegraphics[width=0.15\columnwidth]{img_dif/LD_40/a_105.pdf}
\includegraphics[width=0.15\columnwidth]{img_dif/LD_40/m_120_105.pdf}
\includegraphics[width=0.15\columnwidth]{img_dif/LD_40/m_135_105.pdf}
\\
\includegraphics[width=0.15\columnwidth]{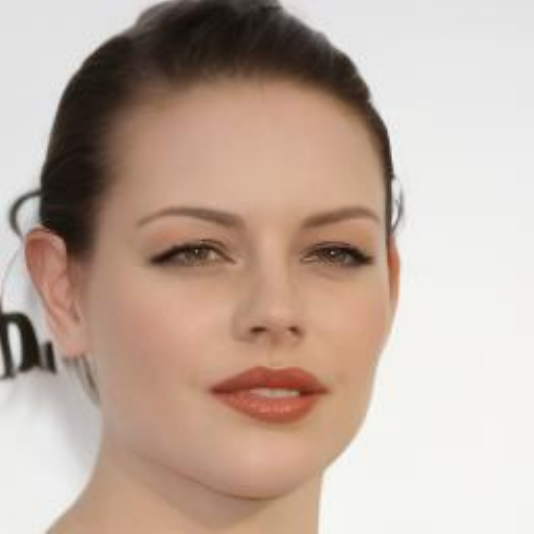}
\includegraphics[width=0.15\columnwidth]{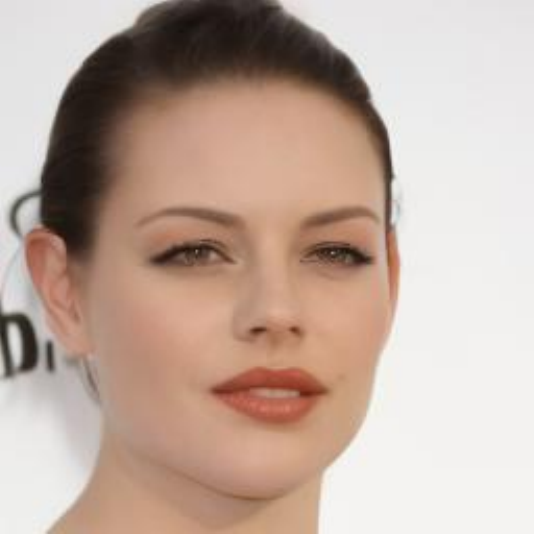}
\includegraphics[width=0.15\columnwidth]{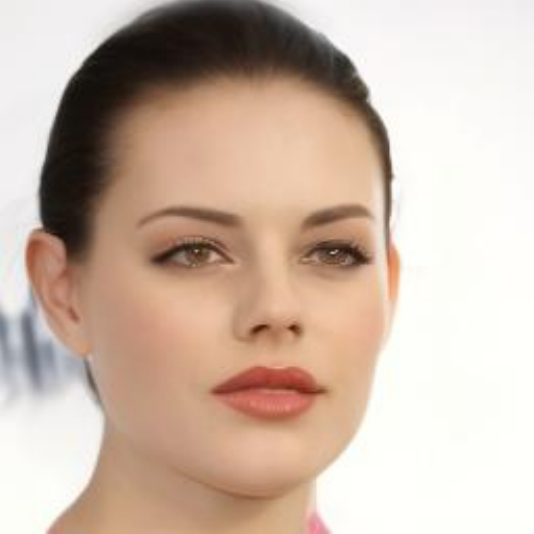}
\includegraphics[width=0.15\columnwidth]{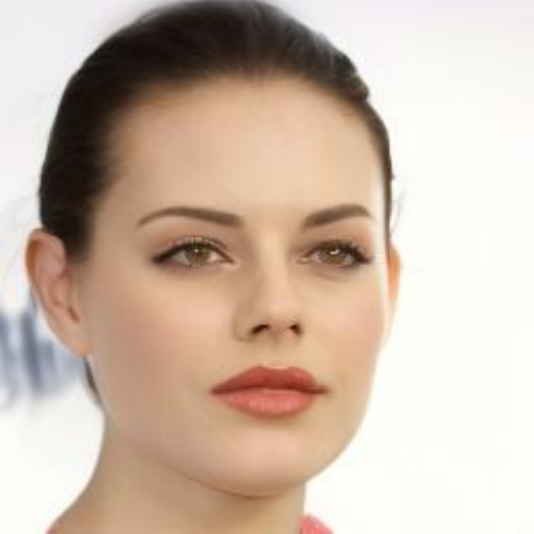}
\\
\includegraphics[width=0.15\columnwidth]{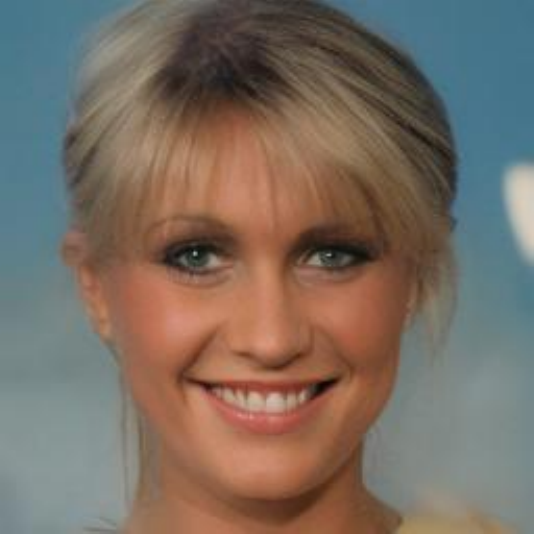}
\includegraphics[width=0.15\columnwidth]{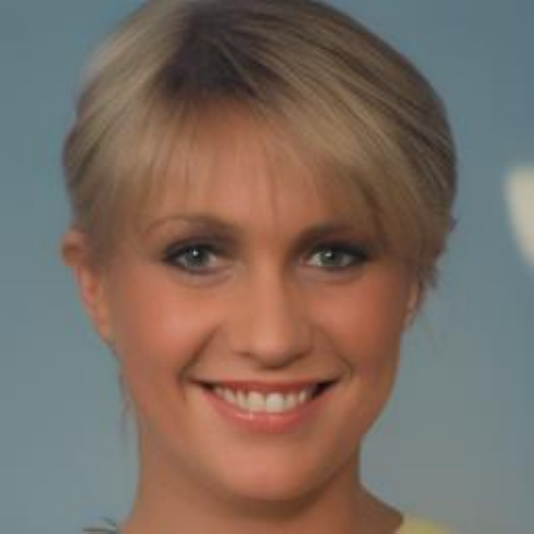}
\includegraphics[width=0.15\columnwidth]{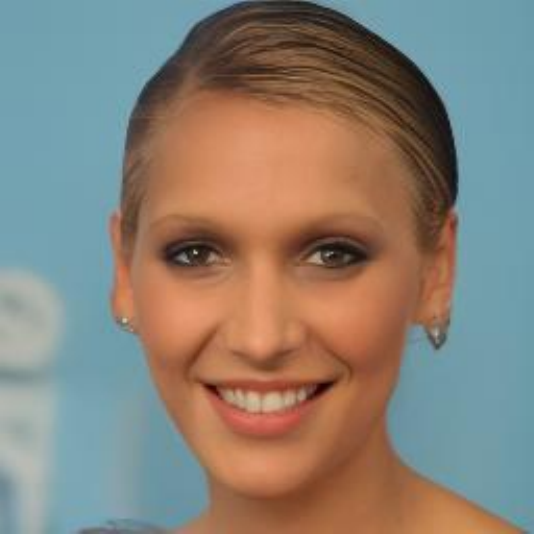}
\includegraphics[width=0.15\columnwidth]{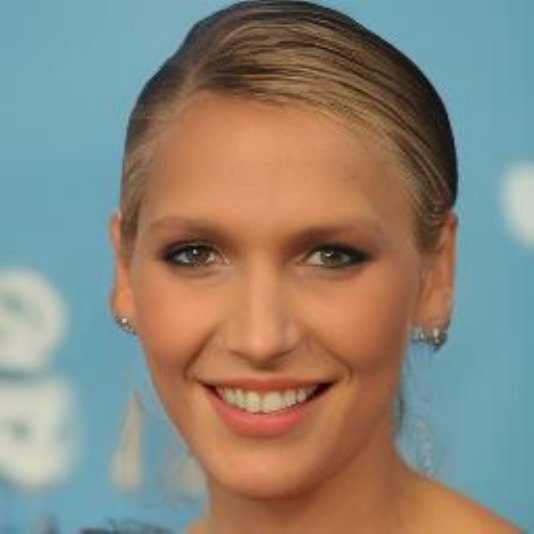}
\\
\includegraphics[width=0.15\columnwidth]{img_dif/LD_40/b_10.pdf}
\includegraphics[width=0.15\columnwidth]{img_dif/LD_40/a_10.pdf}
\includegraphics[width=0.15\columnwidth]{img_dif/LD_40/m_120_10.pdf}
\includegraphics[width=0.15\columnwidth]{img_dif/LD_40/m_135_10.pdf}
\caption{\textbf{LD with $40$ Steps.} 
From left to right: baseline, ablation, generated image with variance scaling $1.20$, and one with $1.35$. 
%
%
%
}
    \label{fig:ld-40}
    \end{figure}

\begin{figure}[t]
    \centering
\includegraphics[width=0.15\columnwidth]{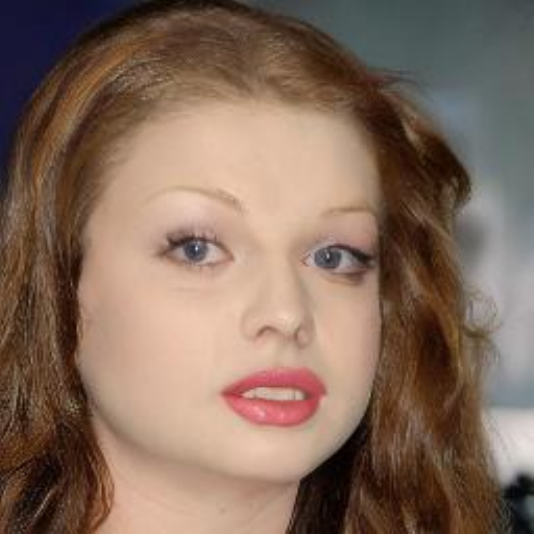}
\includegraphics[width=0.15\columnwidth]{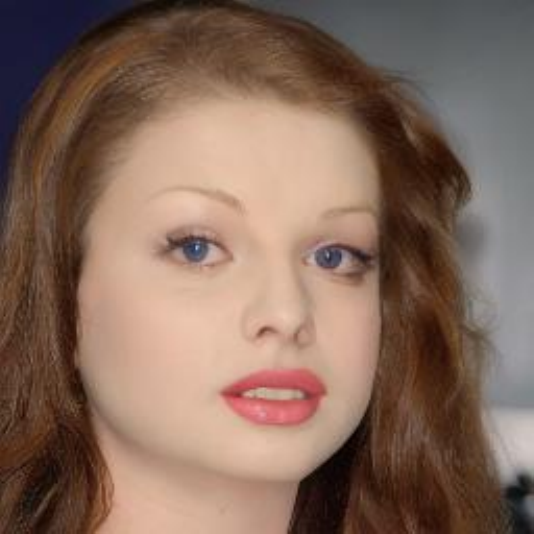}
\includegraphics[width=0.15\columnwidth]{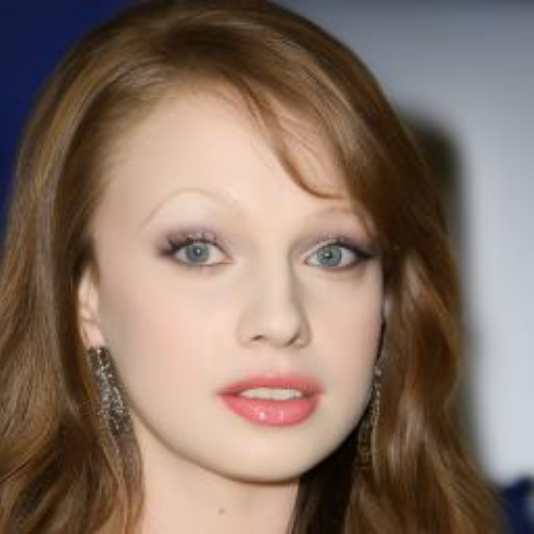}
\includegraphics[width=0.15\columnwidth]{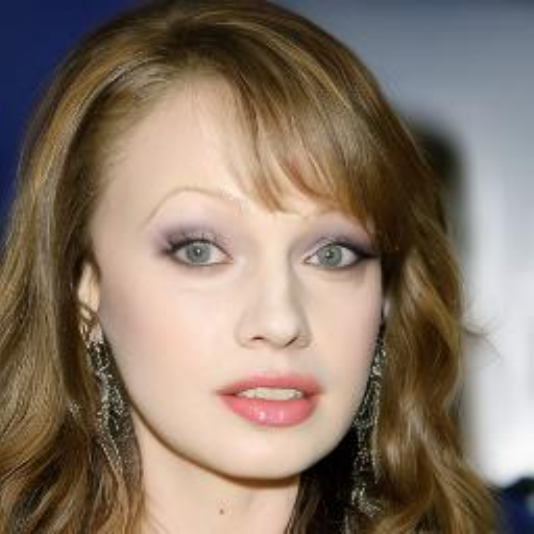}
\\
\includegraphics[width=0.15\columnwidth]{img_dif/LD_80/b_35.pdf}
\includegraphics[width=0.15\columnwidth]{img_dif/LD_80/a_35.pdf}
\includegraphics[width=0.15\columnwidth]{img_dif/LD_80/m_90_35.pdf}
\includegraphics[width=0.15\columnwidth]{img_dif/LD_80/m_110_35.pdf}
\\
\includegraphics[width=0.15\columnwidth]{img_dif/LD_80/b_22.pdf}
\includegraphics[width=0.15\columnwidth]{img_dif/LD_80/a_22.pdf}
\includegraphics[width=0.15\columnwidth]{img_dif/LD_80/m_90_22.pdf}
\includegraphics[width=0.15\columnwidth]{img_dif/LD_80/m_110_22.pdf}
\\
\includegraphics[width=0.15\columnwidth]{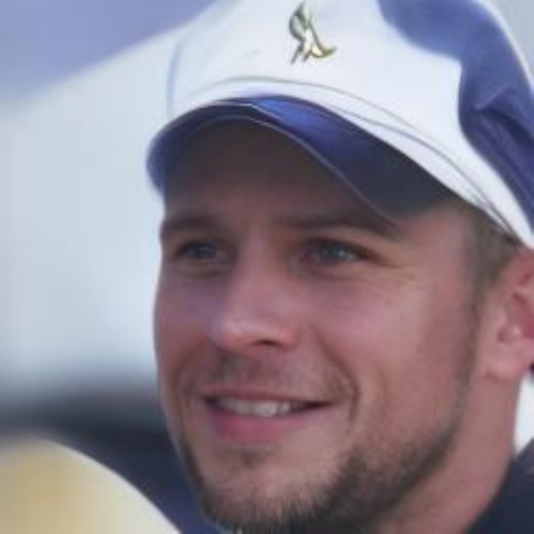}
\includegraphics[width=0.15\columnwidth]{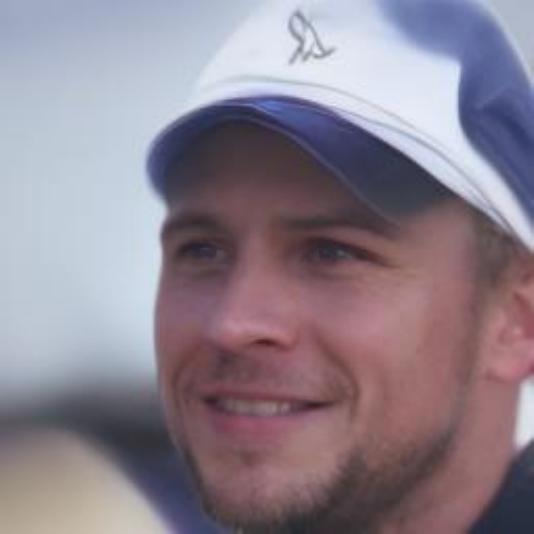}
\includegraphics[width=0.15\columnwidth]{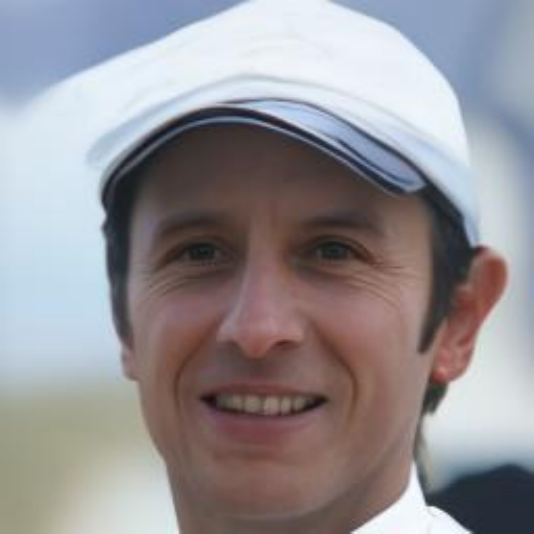}
\includegraphics[width=0.15\columnwidth]{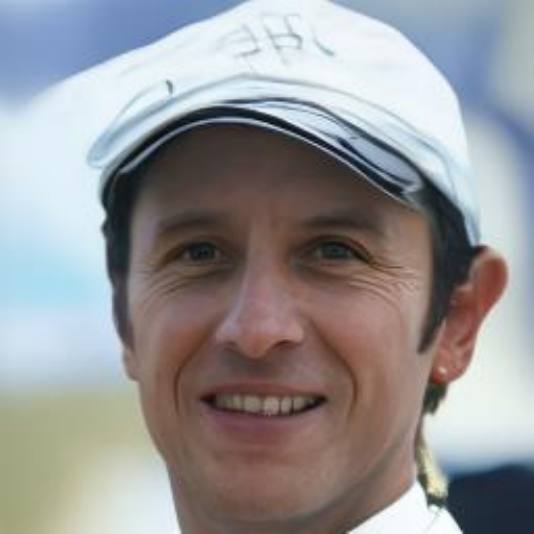}
\\
\includegraphics[width=0.15\columnwidth]{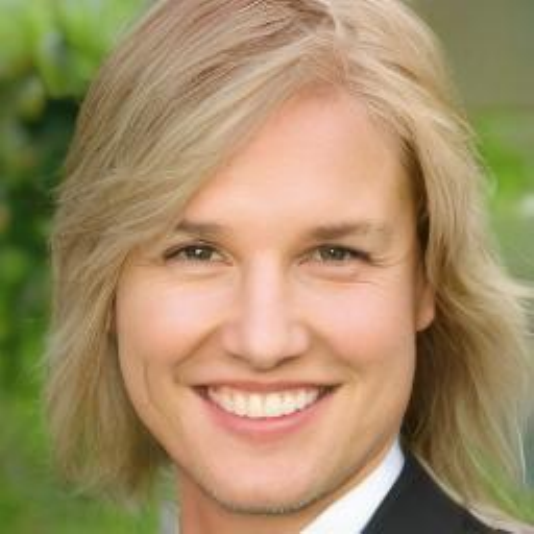}
\includegraphics[width=0.15\columnwidth]{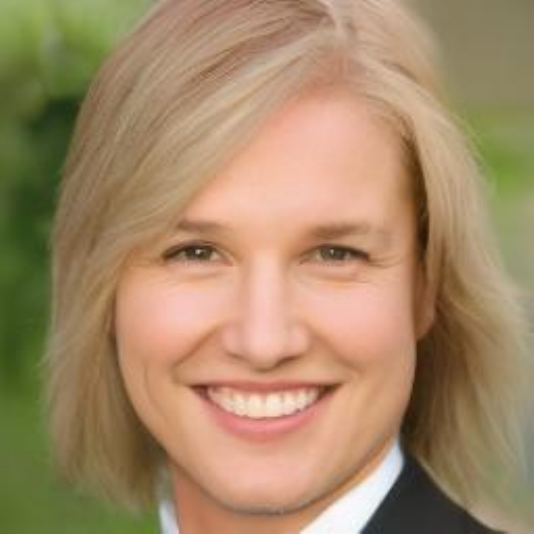}
\includegraphics[width=0.15\columnwidth]{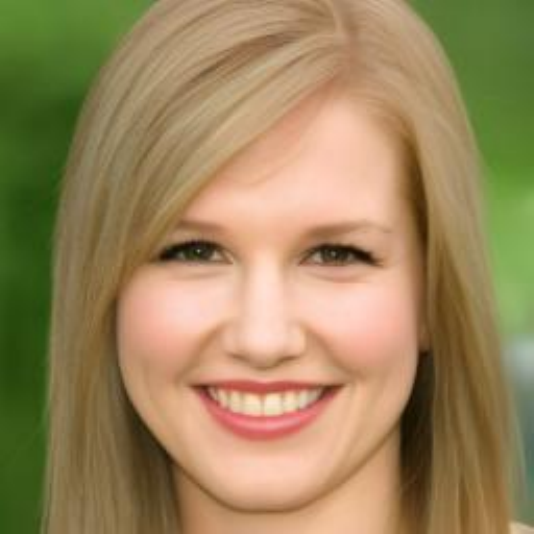}
\includegraphics[width=0.15\columnwidth]{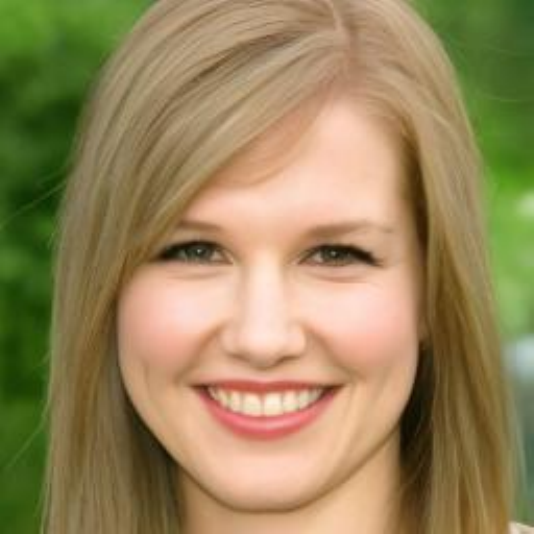}
\\
\includegraphics[width=0.15\columnwidth]{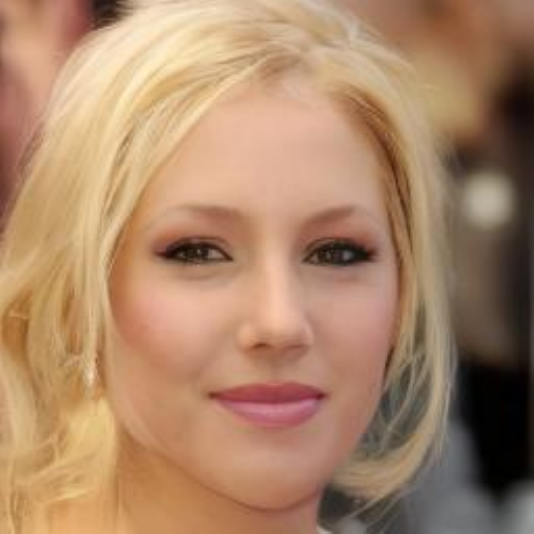}
\includegraphics[width=0.15\columnwidth]{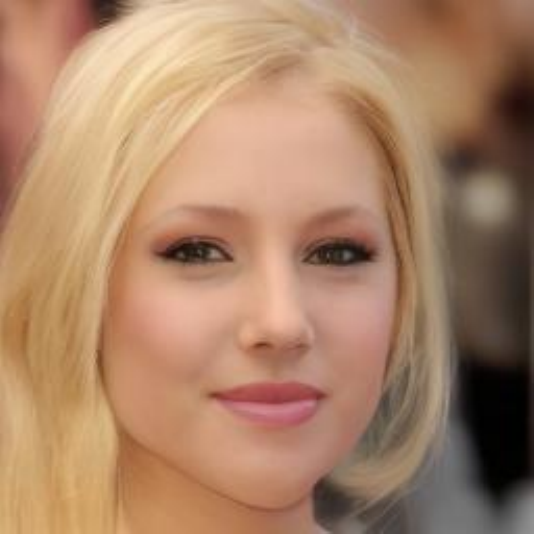}
\includegraphics[width=0.15\columnwidth]{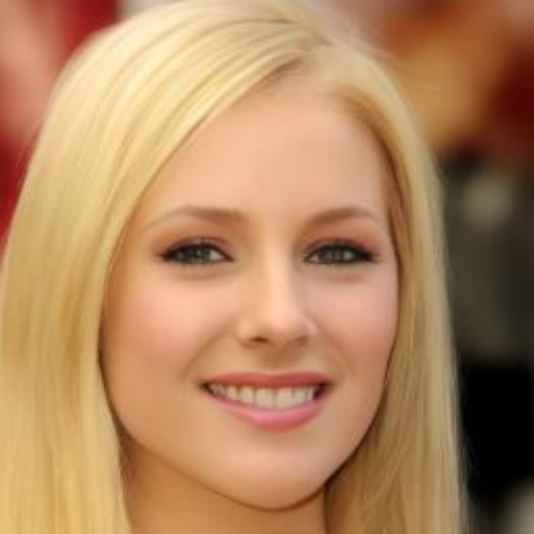}
\includegraphics[width=0.15\columnwidth]{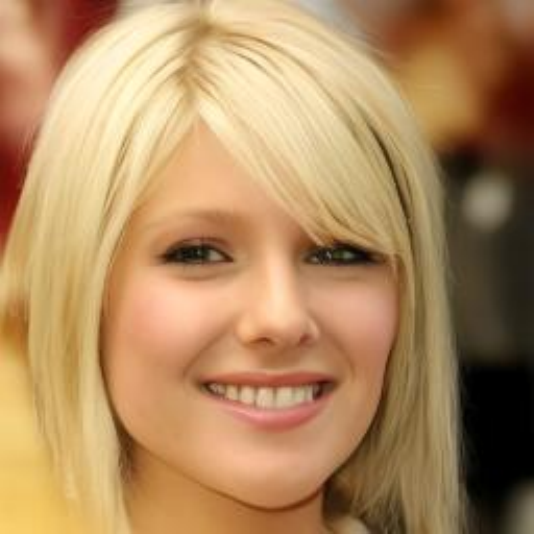}
\\
\includegraphics[width=0.15\columnwidth]{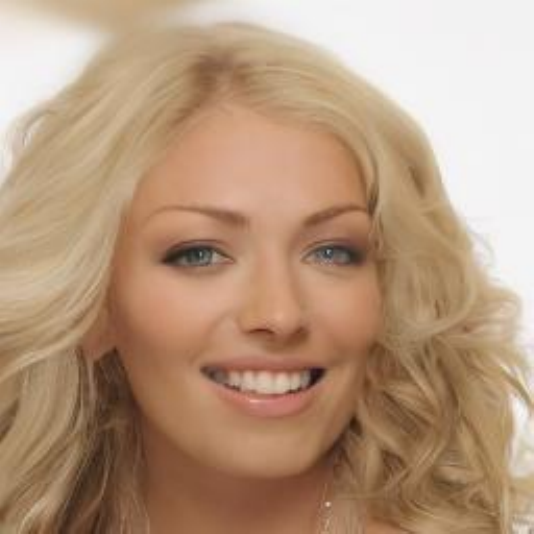}
\includegraphics[width=0.15\columnwidth]{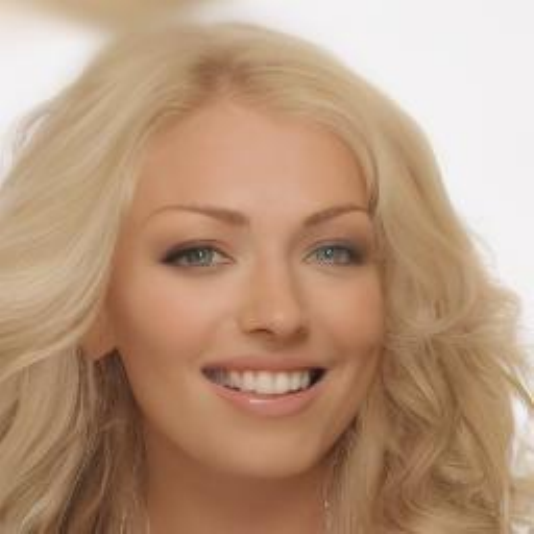}
\includegraphics[width=0.15\columnwidth]{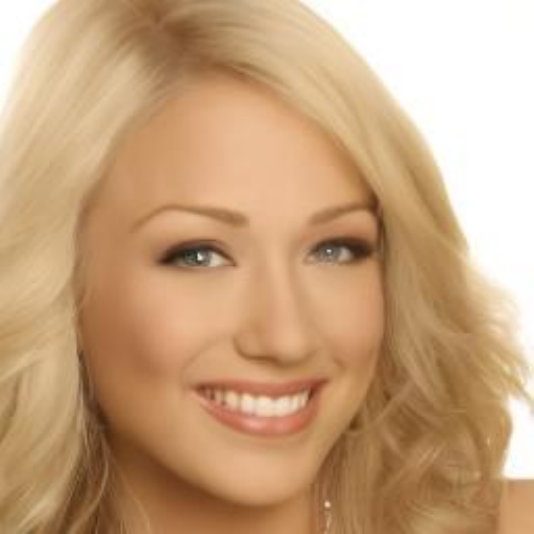}
\includegraphics[width=0.15\columnwidth]{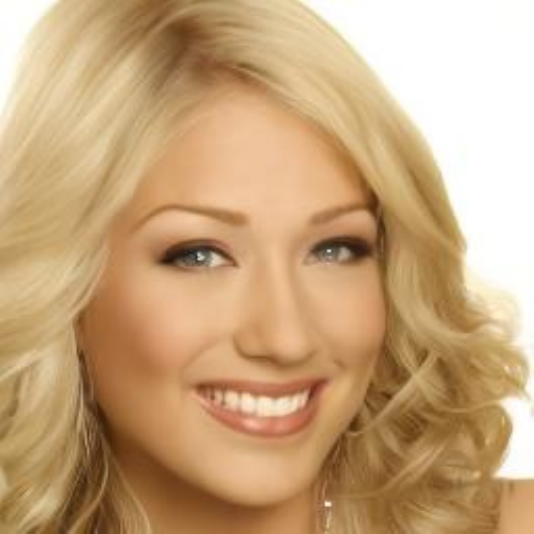}
\\
\includegraphics[width=0.15\columnwidth]{img_dif/LD_80/b_87.pdf}
\includegraphics[width=0.15\columnwidth]{img_dif/LD_80/a_87.pdf}
\includegraphics[width=0.15\columnwidth]{img_dif/LD_80/m_90_87.pdf}
\includegraphics[width=0.15\columnwidth]{img_dif/LD_80/m_110_87.pdf}
\caption{\textbf{LD with 80 Steps.} For each row, from left to right: baseline, ablation, generated image with variance scaling $0.90$, and one with $1.10$. 
%
%
}
    \label{fig:ld-80}
    \end{figure}

\begin{figure}[t]
    \centering
\includegraphics[width=0.15\columnwidth]{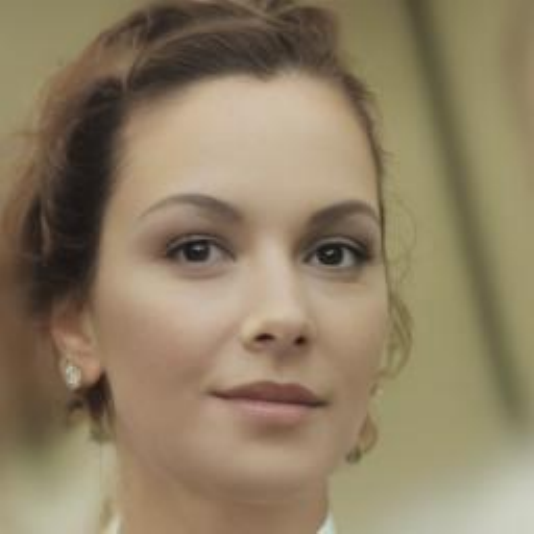}
\includegraphics[width=0.15\columnwidth]{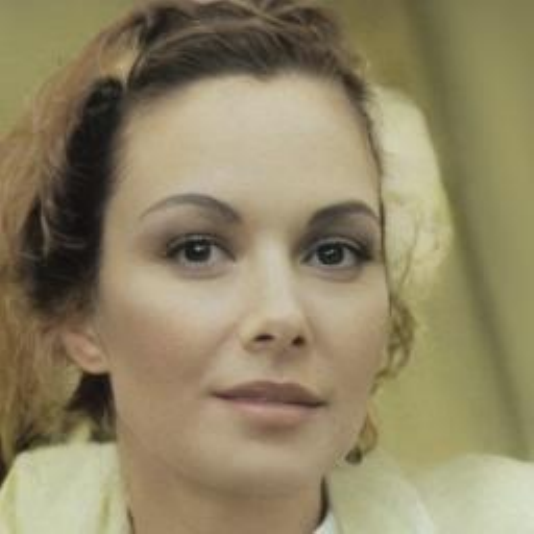}
\includegraphics[width=0.15\columnwidth]{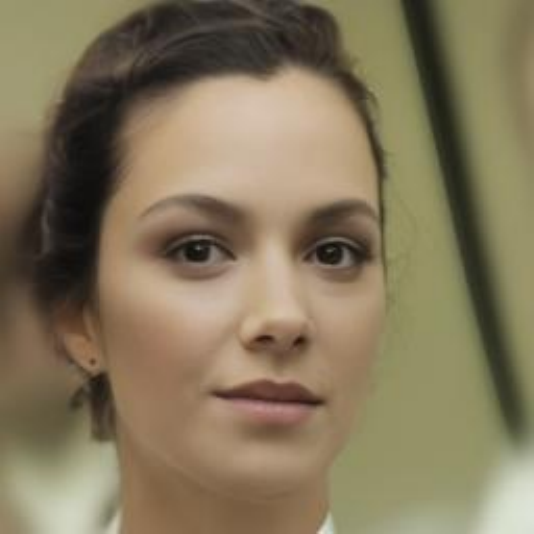}
\includegraphics[width=0.15\columnwidth]{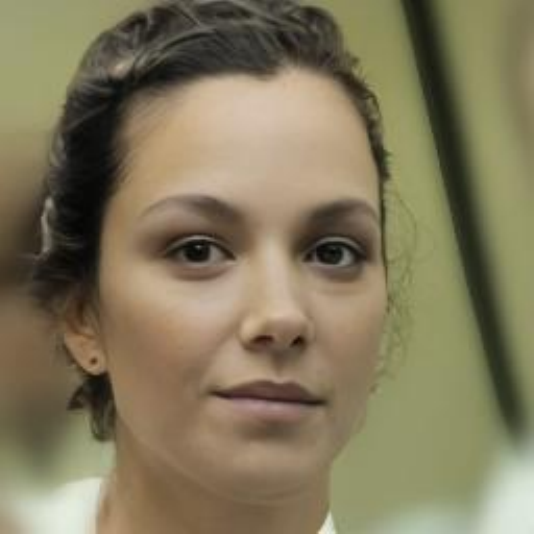}
\\
\includegraphics[width=0.15\columnwidth]{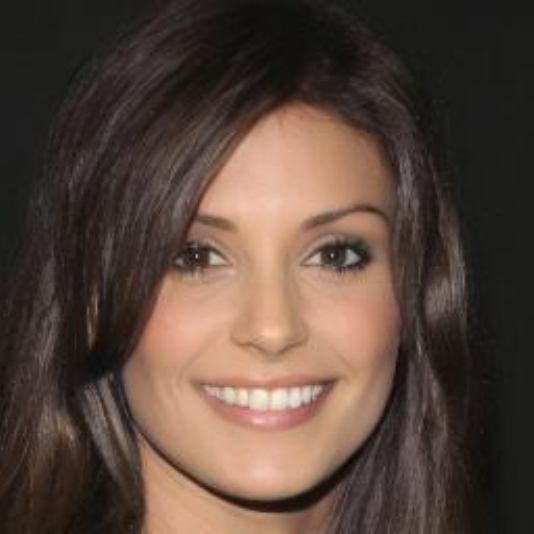}
\includegraphics[width=0.15\columnwidth]{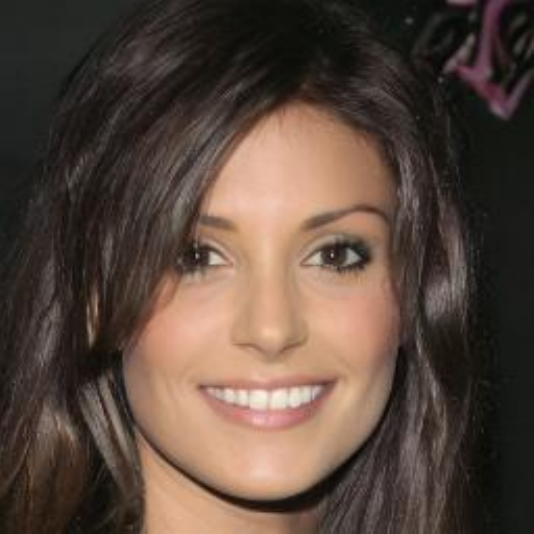}
\includegraphics[width=0.15\columnwidth]{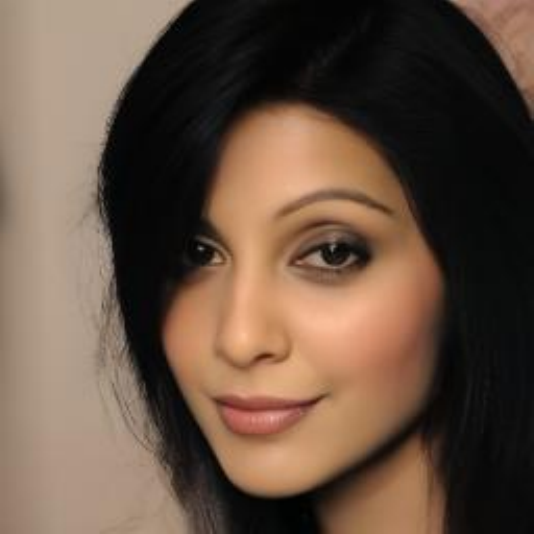}
\includegraphics[width=0.15\columnwidth]{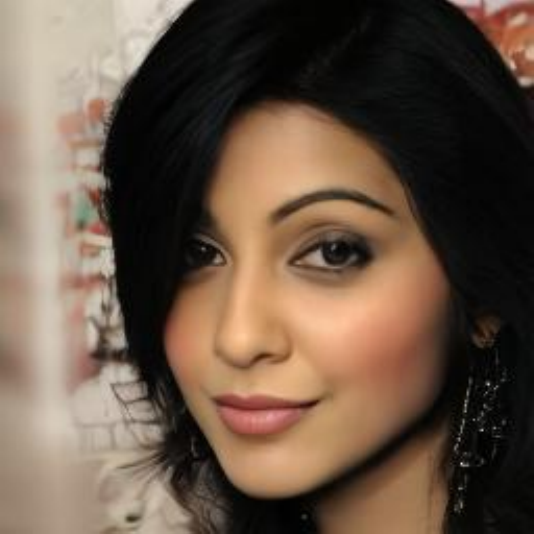}
\caption{\textbf{Examples of No Overexposure and Less Brightness in LD with $80$ Steps.} 
For each row, from left to right: baseline, ablation, generated image with variance scaling $0.90$, and one with $1.10$.
%
%
}
    \label{fig:LD-80-contrast}
    \end{figure}

\begin{figure}[t]
    \centering
\includegraphics[width=0.15\columnwidth]{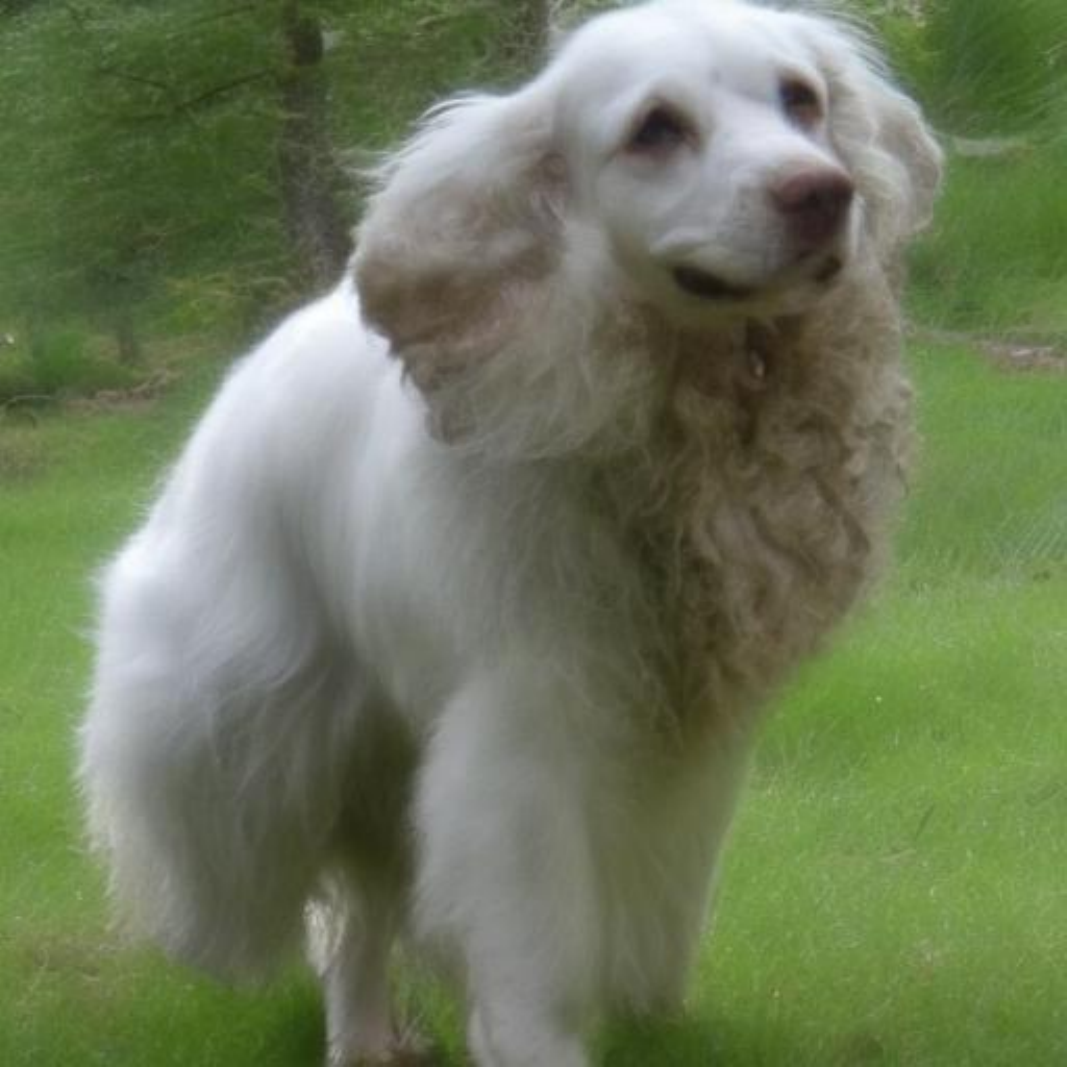}
\includegraphics[width=0.15\columnwidth]{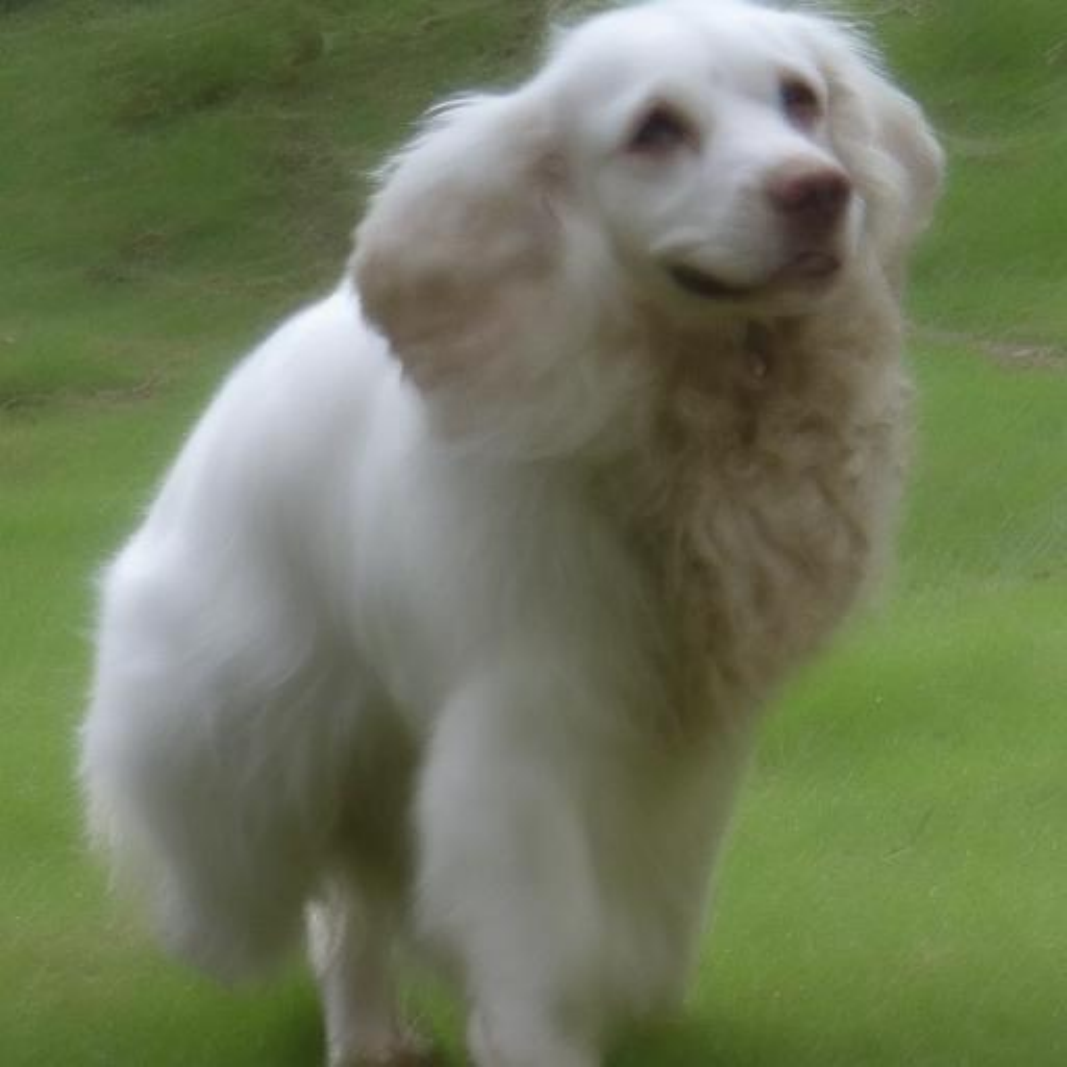}
\includegraphics[width=0.15\columnwidth]{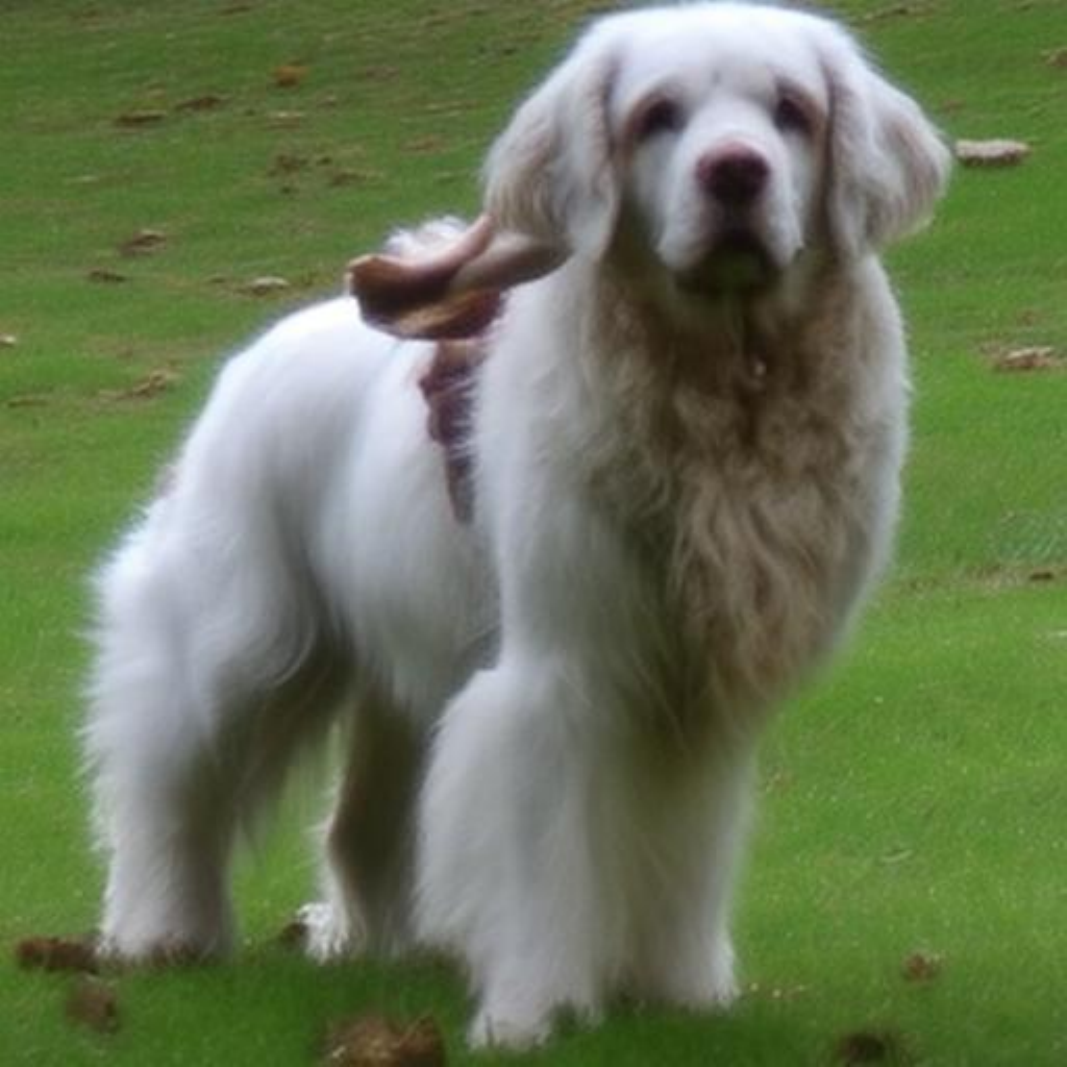}
\includegraphics[width=0.15\columnwidth]{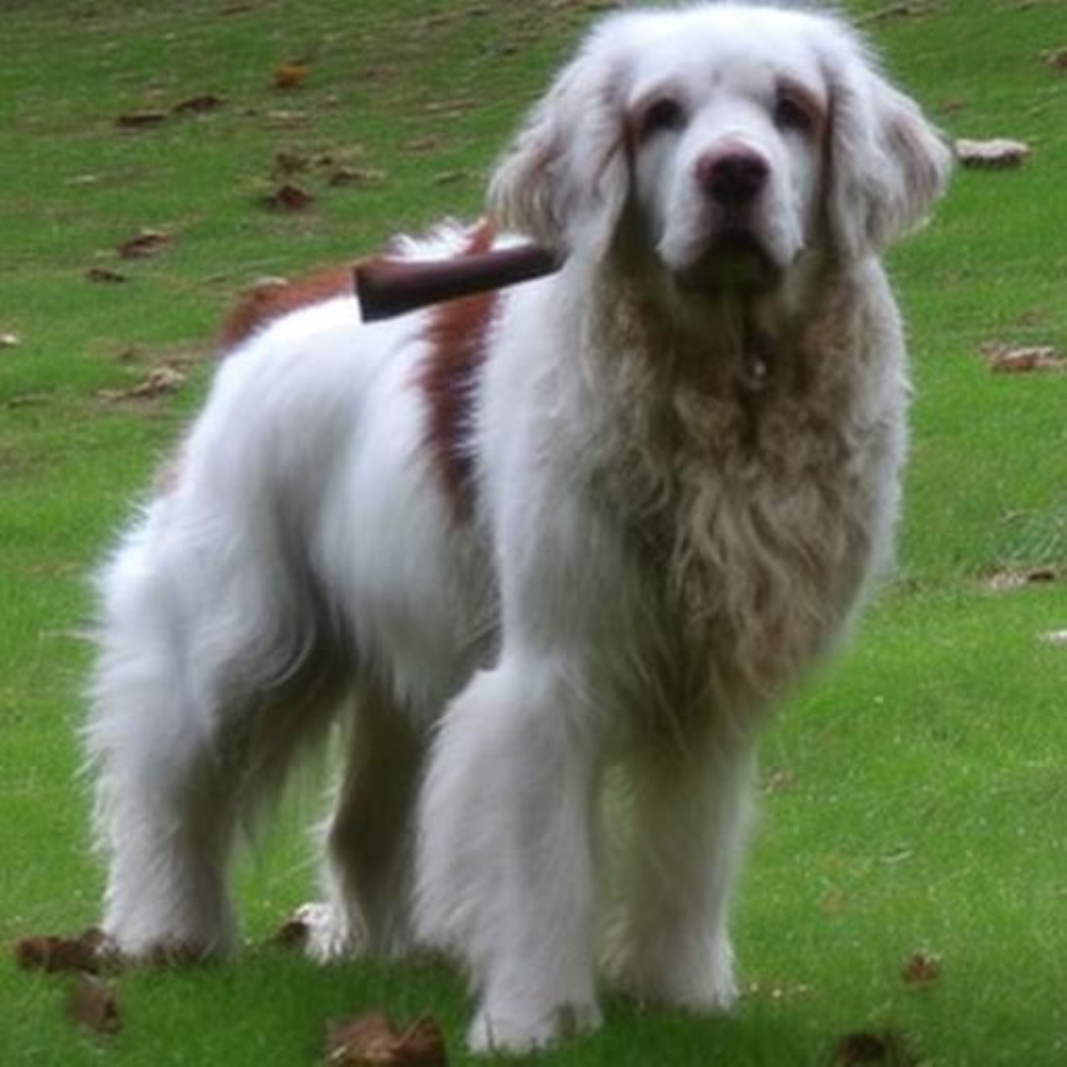}
\\
\includegraphics[width=0.15\columnwidth]{img_dif/DiT_40/b_225_330.pdf}
\includegraphics[width=0.15\columnwidth]{img_dif/DiT_40/a_225_330.pdf}
\includegraphics[width=0.15\columnwidth]{img_dif/DiT_40/m_114_225_330.pdf}
\includegraphics[width=0.15\columnwidth]{img_dif/DiT_40/m_135_225_330.pdf}
\\
\includegraphics[width=0.15\columnwidth]{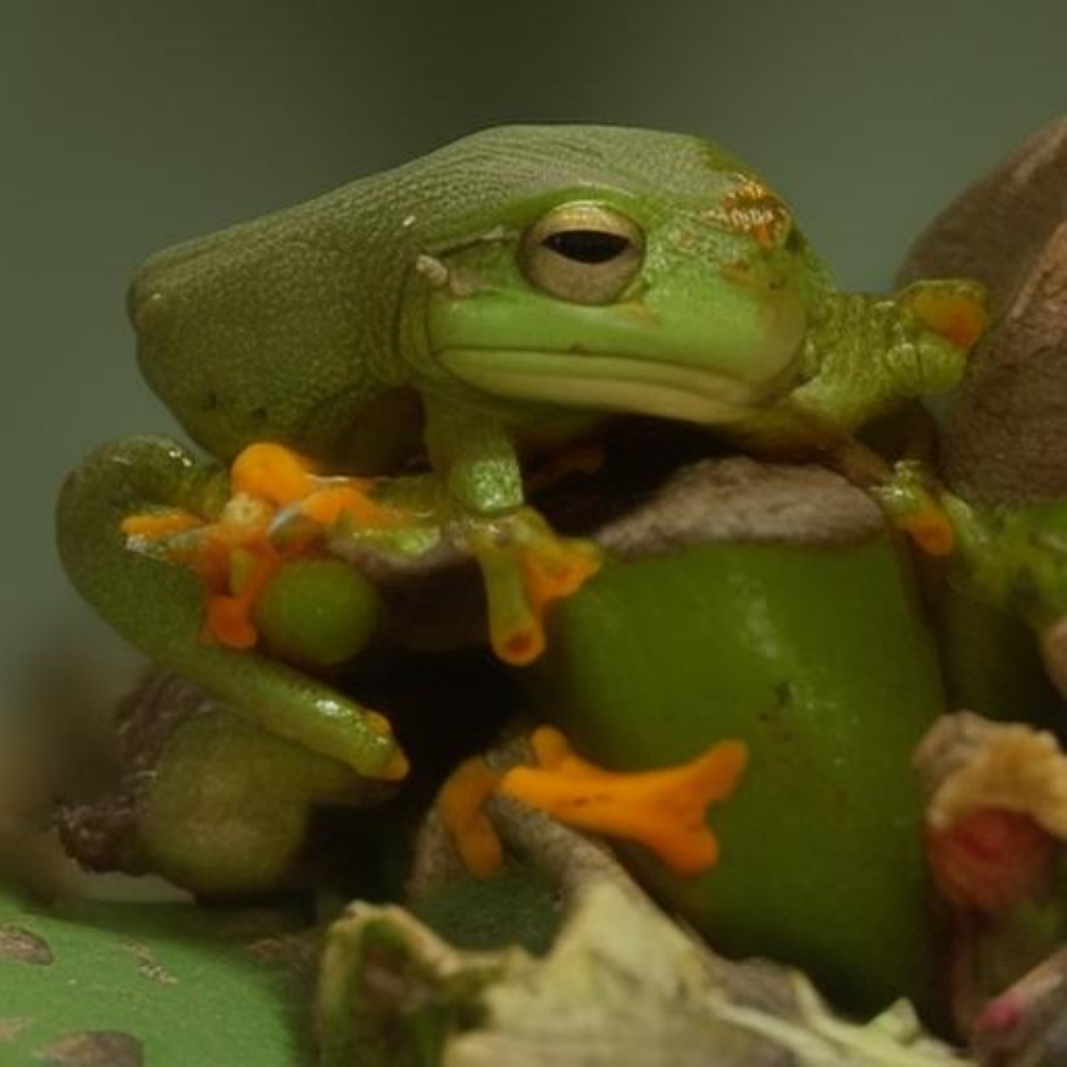}
\includegraphics[width=0.15\columnwidth]{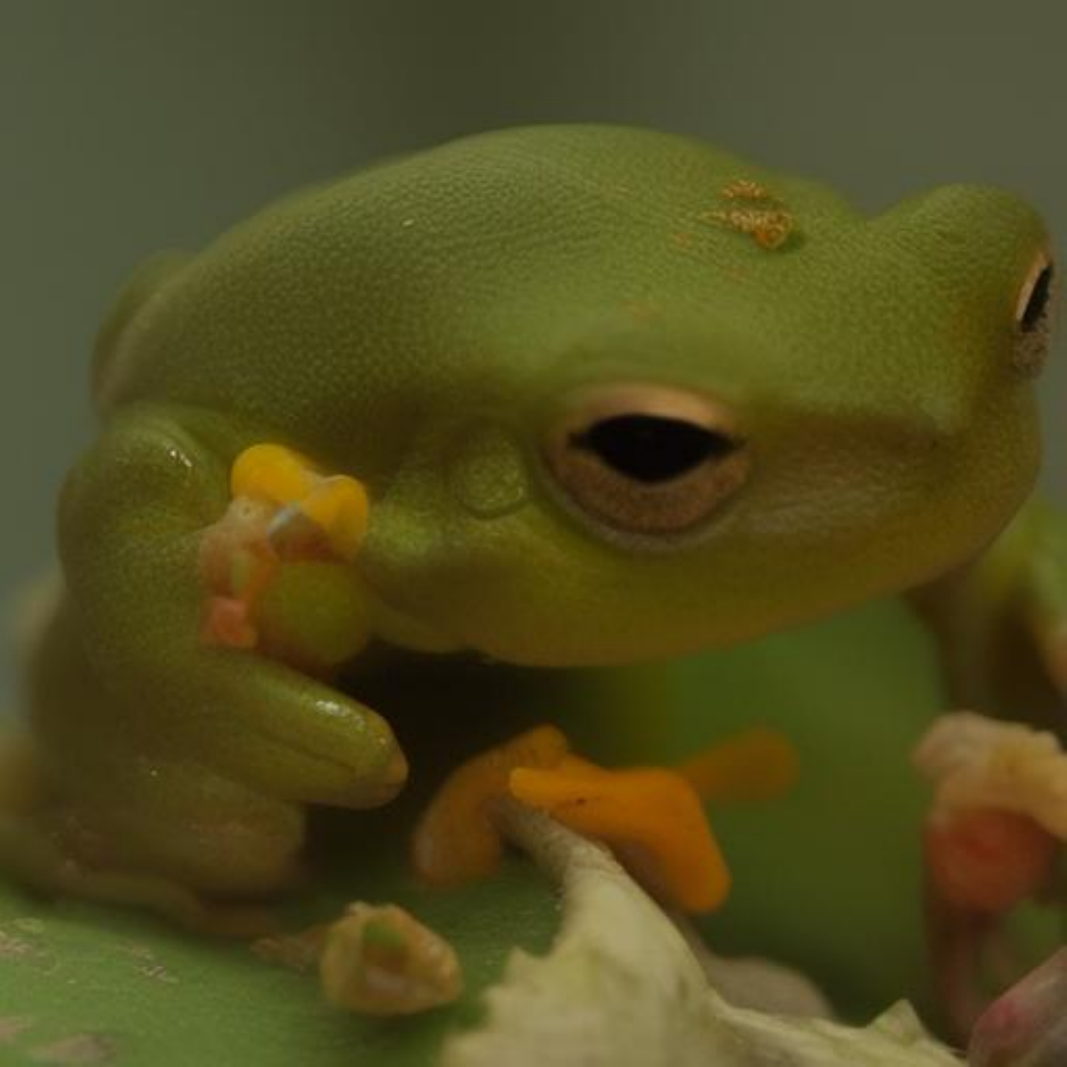}
\includegraphics[width=0.15\columnwidth]{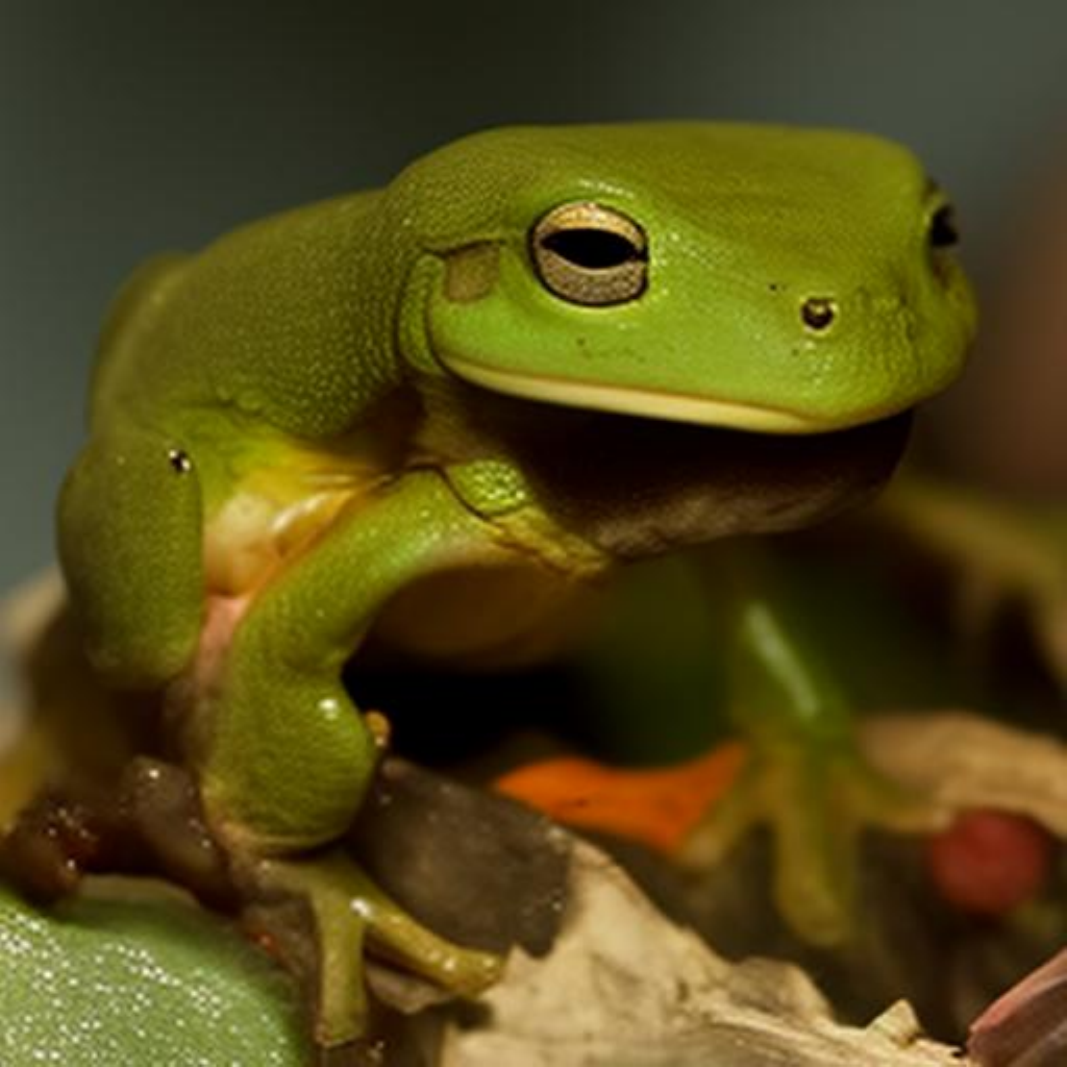}
\includegraphics[width=0.15\columnwidth]{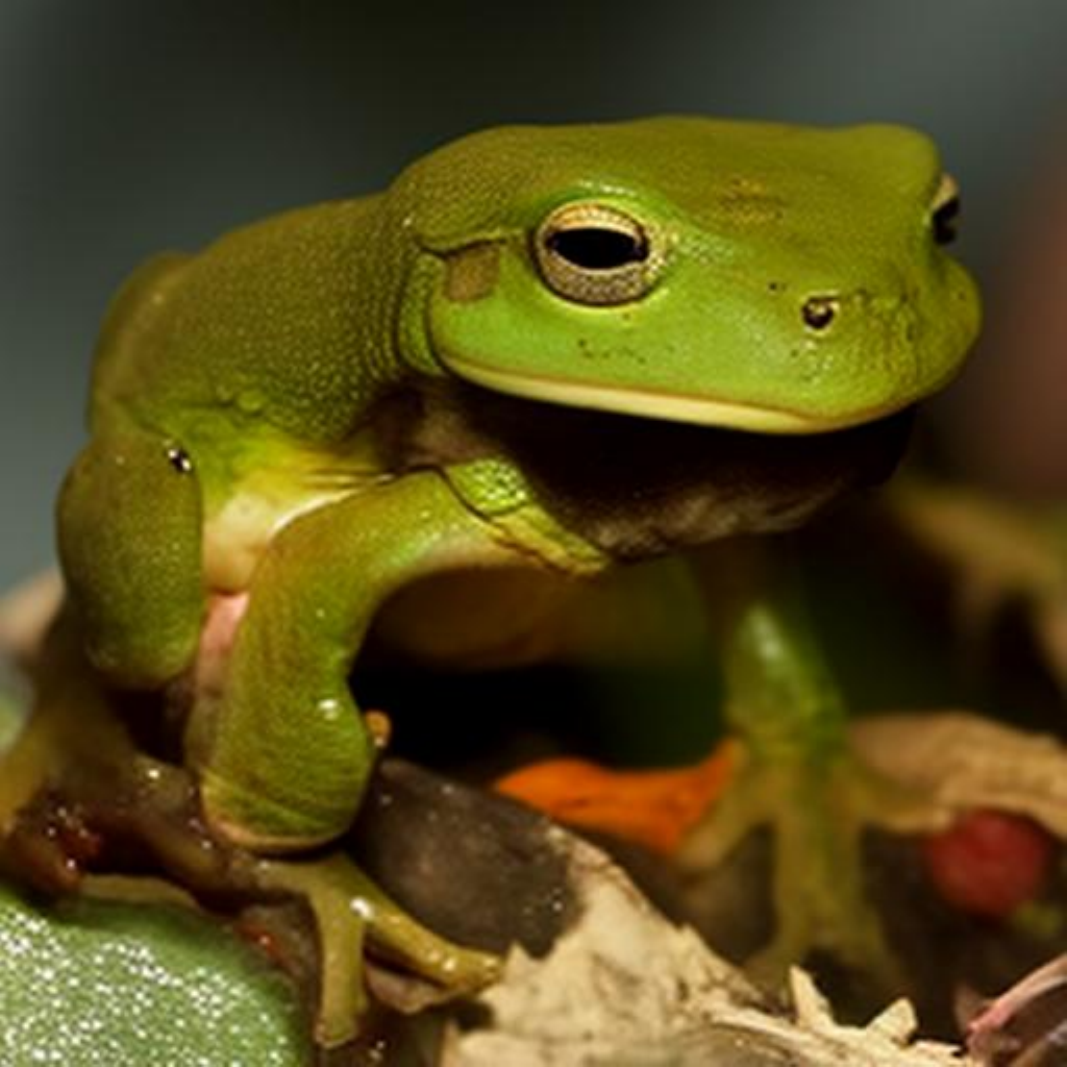}
\\
\includegraphics[width=0.15\columnwidth]{img_dif/DiT_40/b_209_266.pdf}
\includegraphics[width=0.15\columnwidth]{img_dif/DiT_40/a_209_266.pdf}
\includegraphics[width=0.15\columnwidth]{img_dif/DiT_40/m_114_209_266.pdf}
\includegraphics[width=0.15\columnwidth]{img_dif/DiT_40/m_135_209_266.pdf}
\\
\includegraphics[width=0.15\columnwidth]{img_dif/DiT_40/b_194_901.pdf}
\includegraphics[width=0.15\columnwidth]{img_dif/DiT_40/a_194_901.pdf}
\includegraphics[width=0.15\columnwidth]{img_dif/DiT_40/m_114_194_901.pdf}
\includegraphics[width=0.15\columnwidth]{img_dif/DiT_40/m_135_194_901.pdf}
\\
\includegraphics[width=0.15\columnwidth]{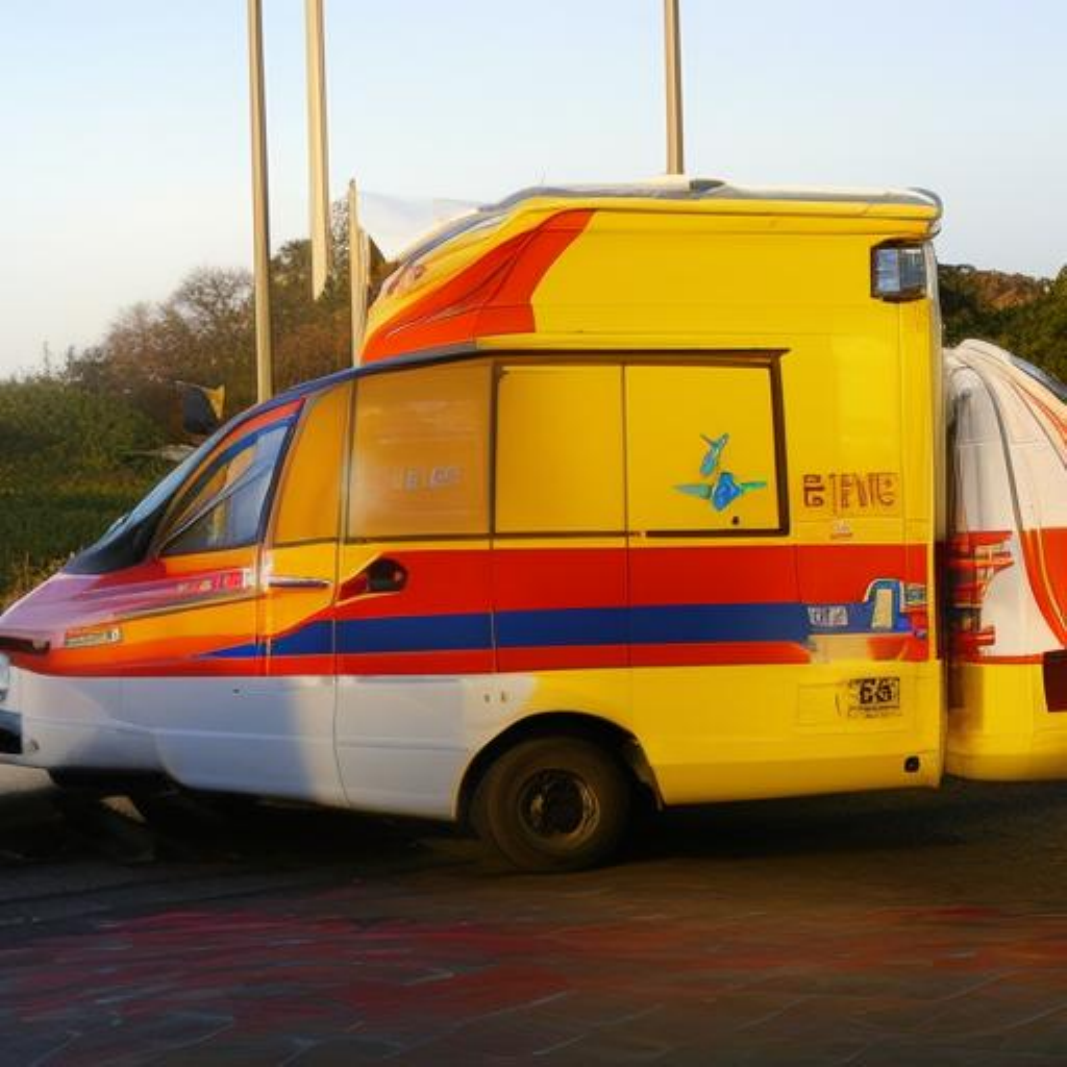}
\includegraphics[width=0.15\columnwidth]{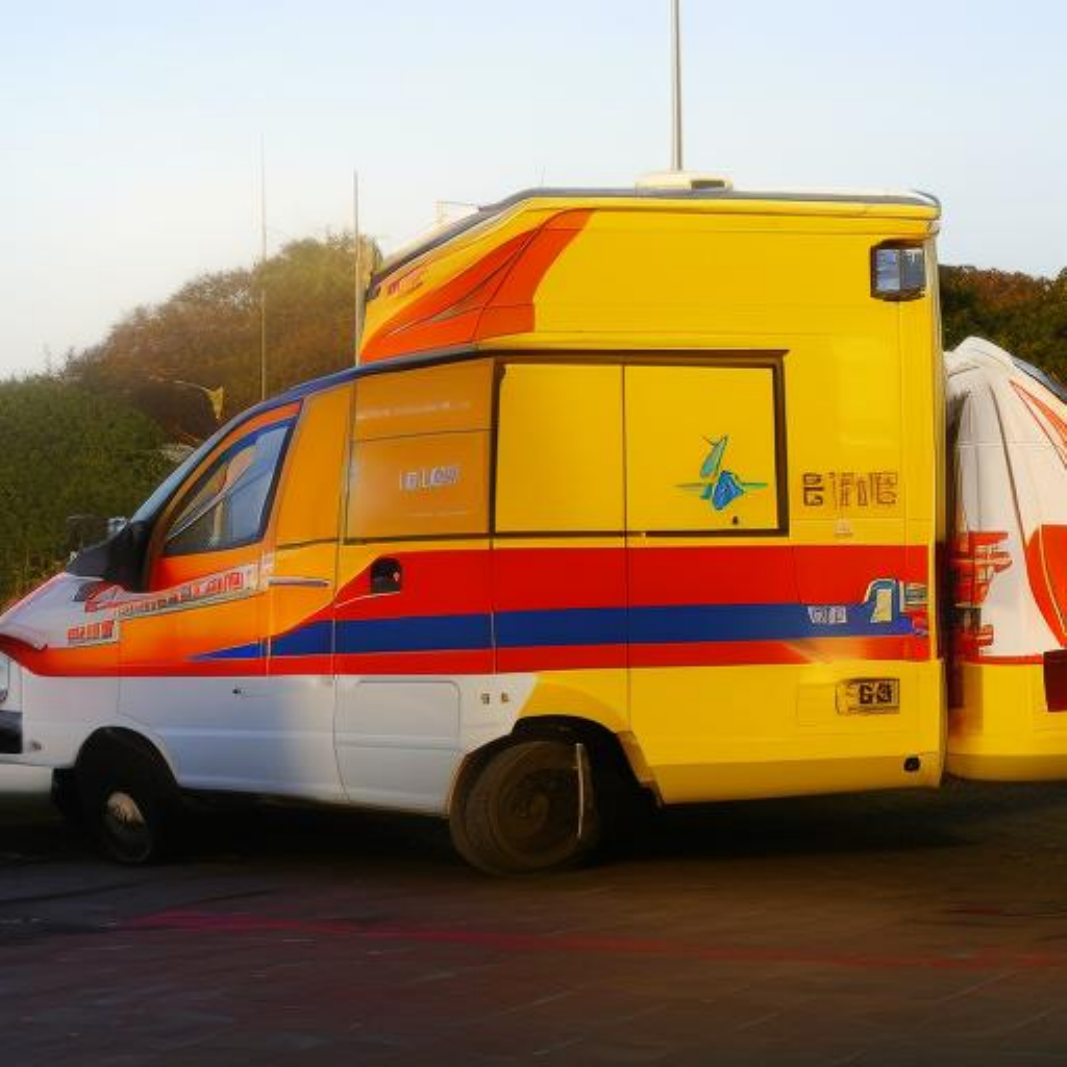}
\includegraphics[width=0.15\columnwidth]{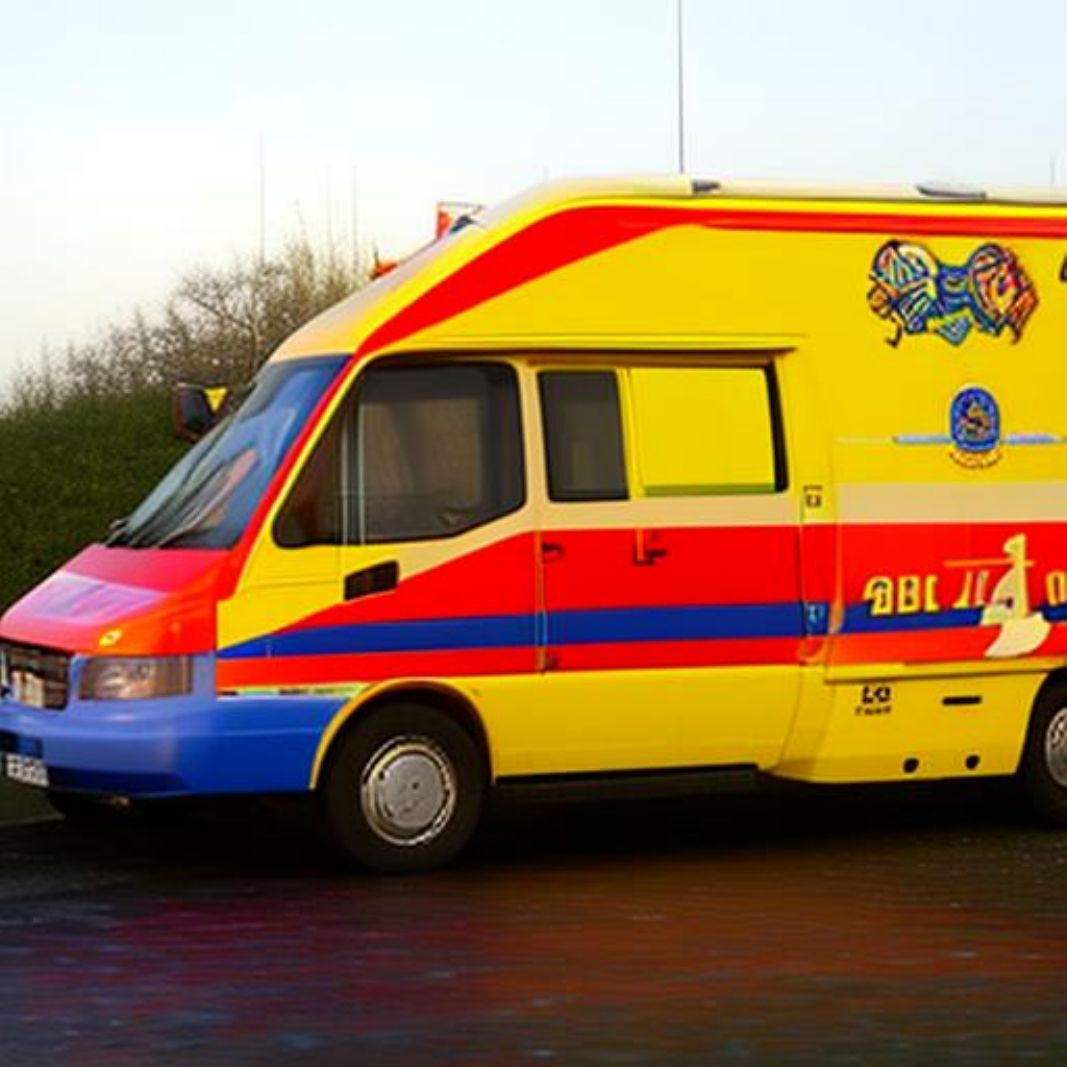}
\includegraphics[width=0.15\columnwidth]{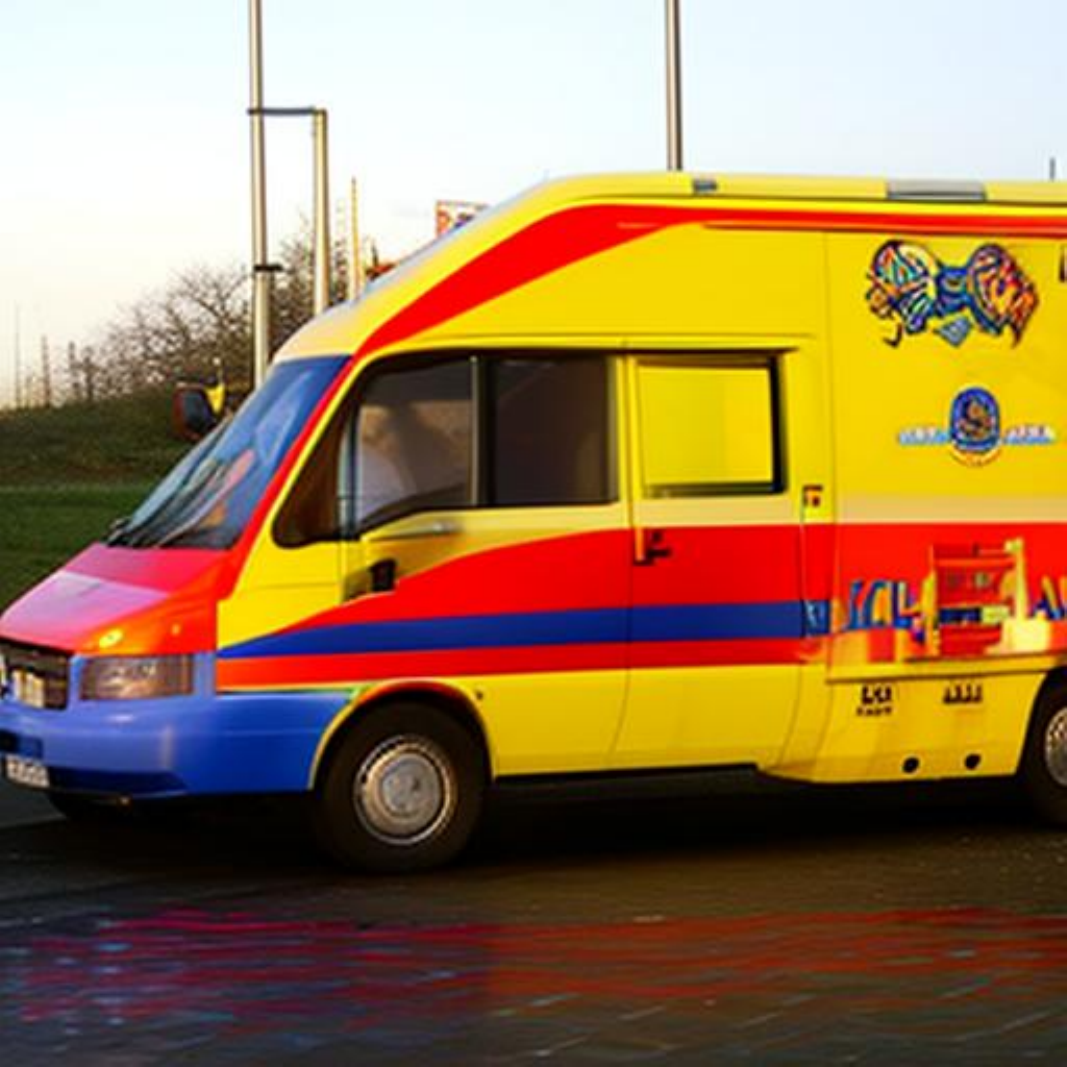}
\\
\includegraphics[width=0.15\columnwidth]{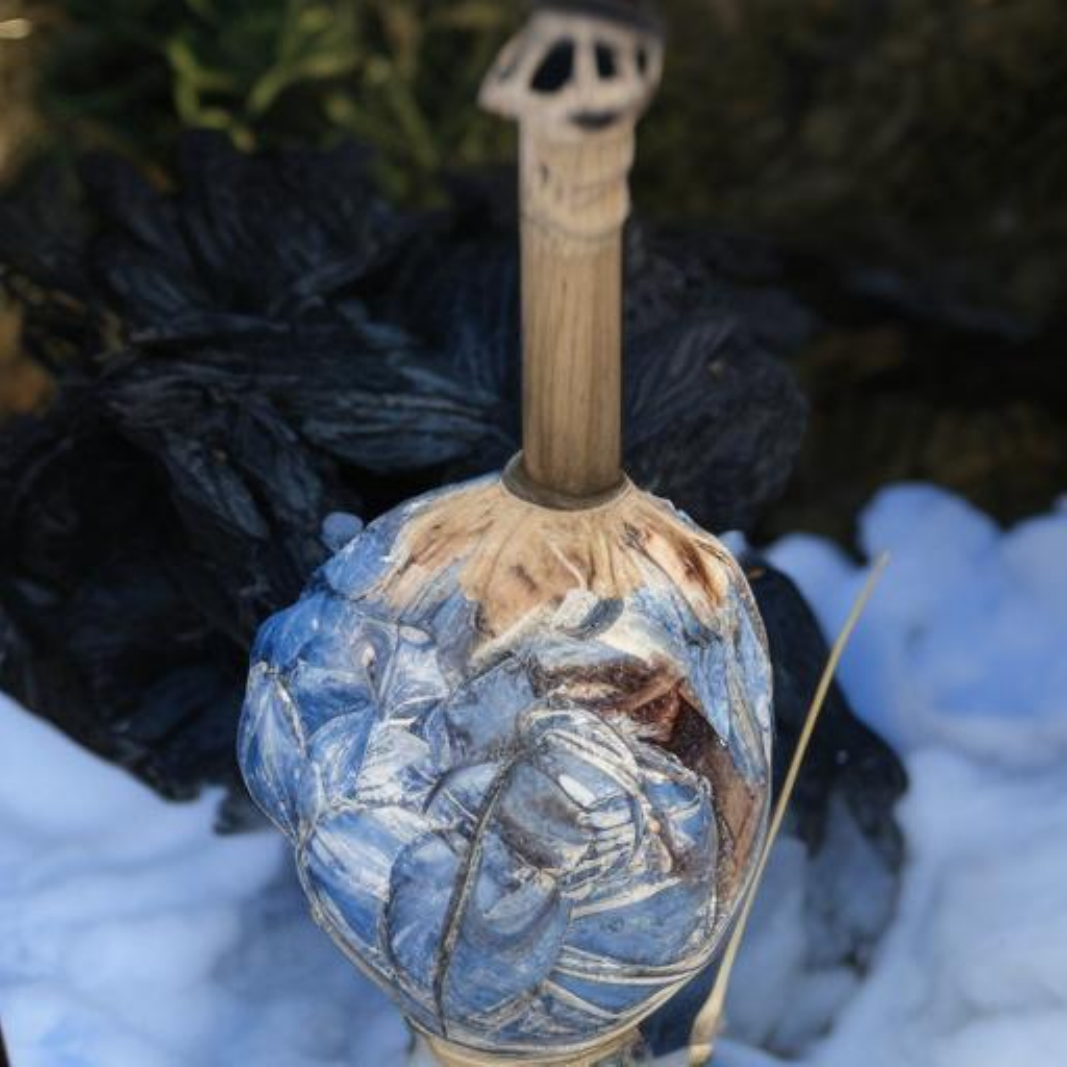}
\includegraphics[width=0.15\columnwidth]{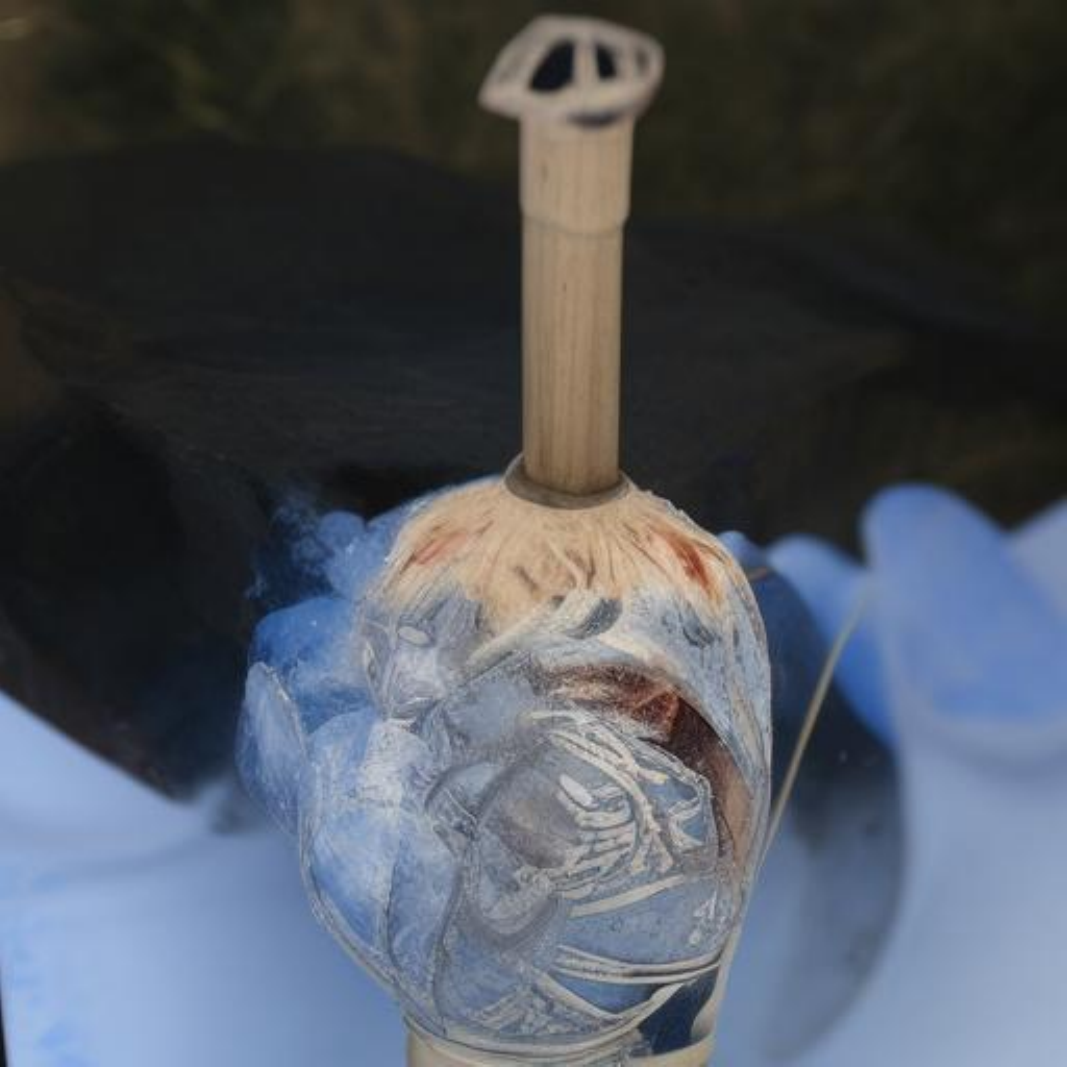}
\includegraphics[width=0.15\columnwidth]{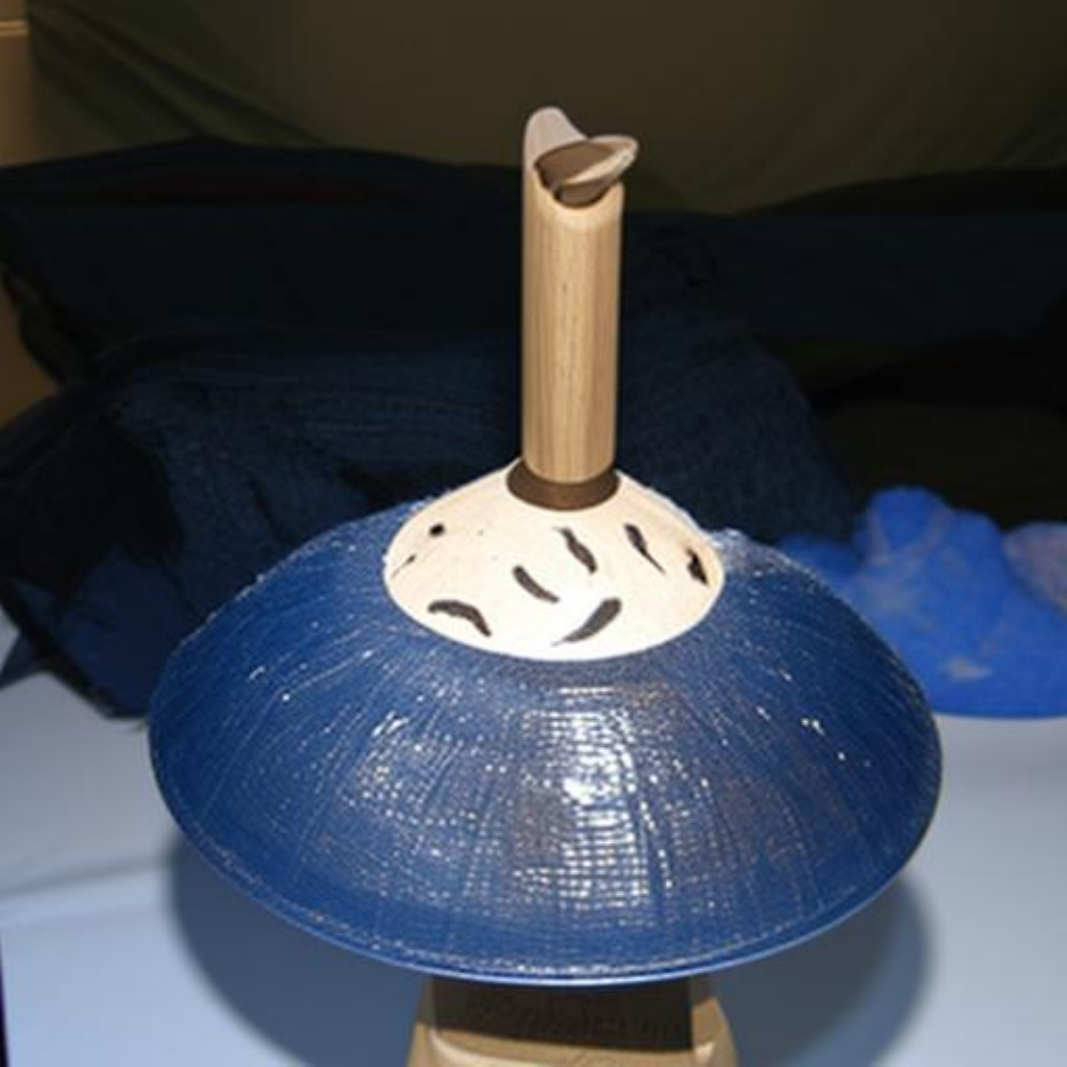}
\includegraphics[width=0.15\columnwidth]{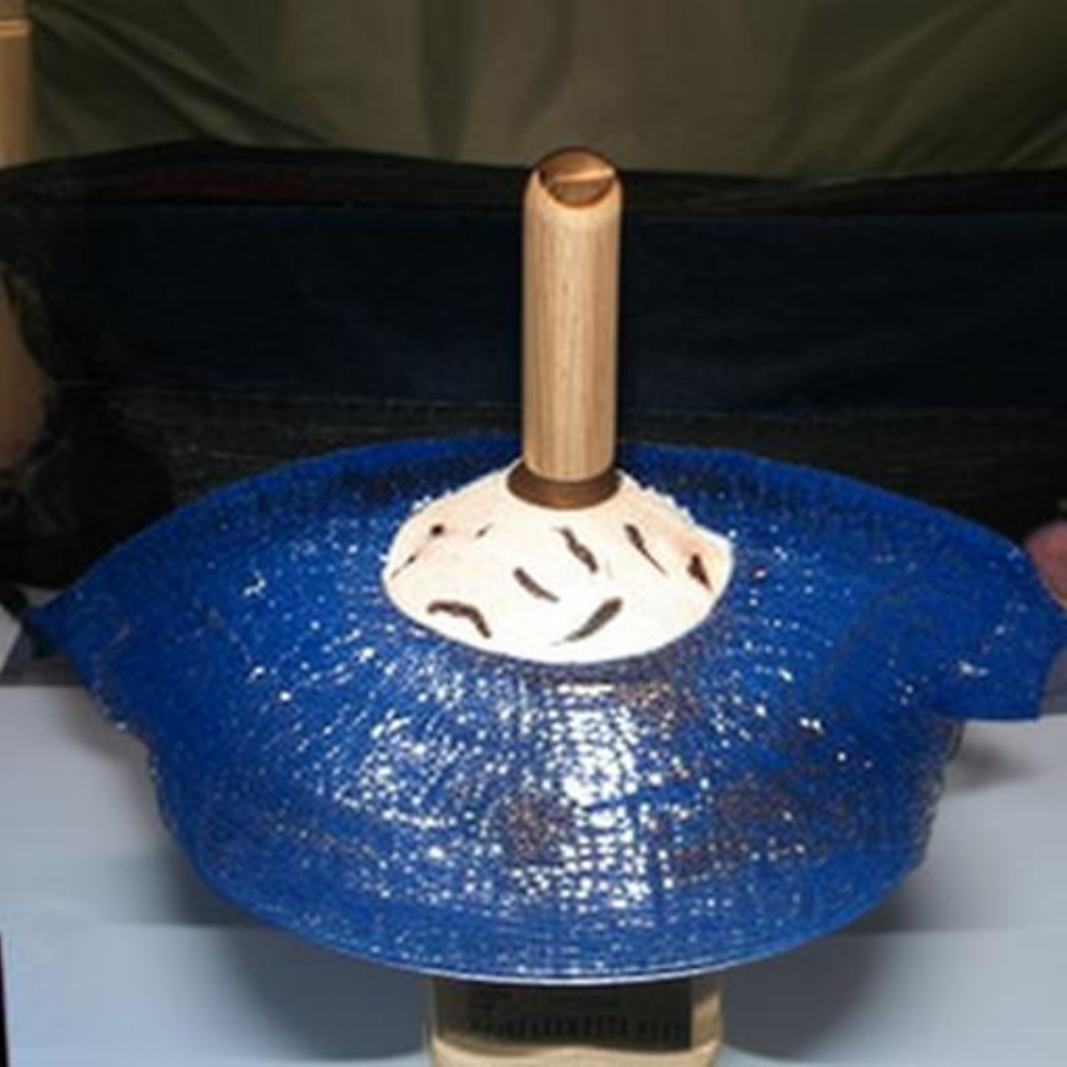}
\\
\includegraphics[width=0.15\columnwidth]{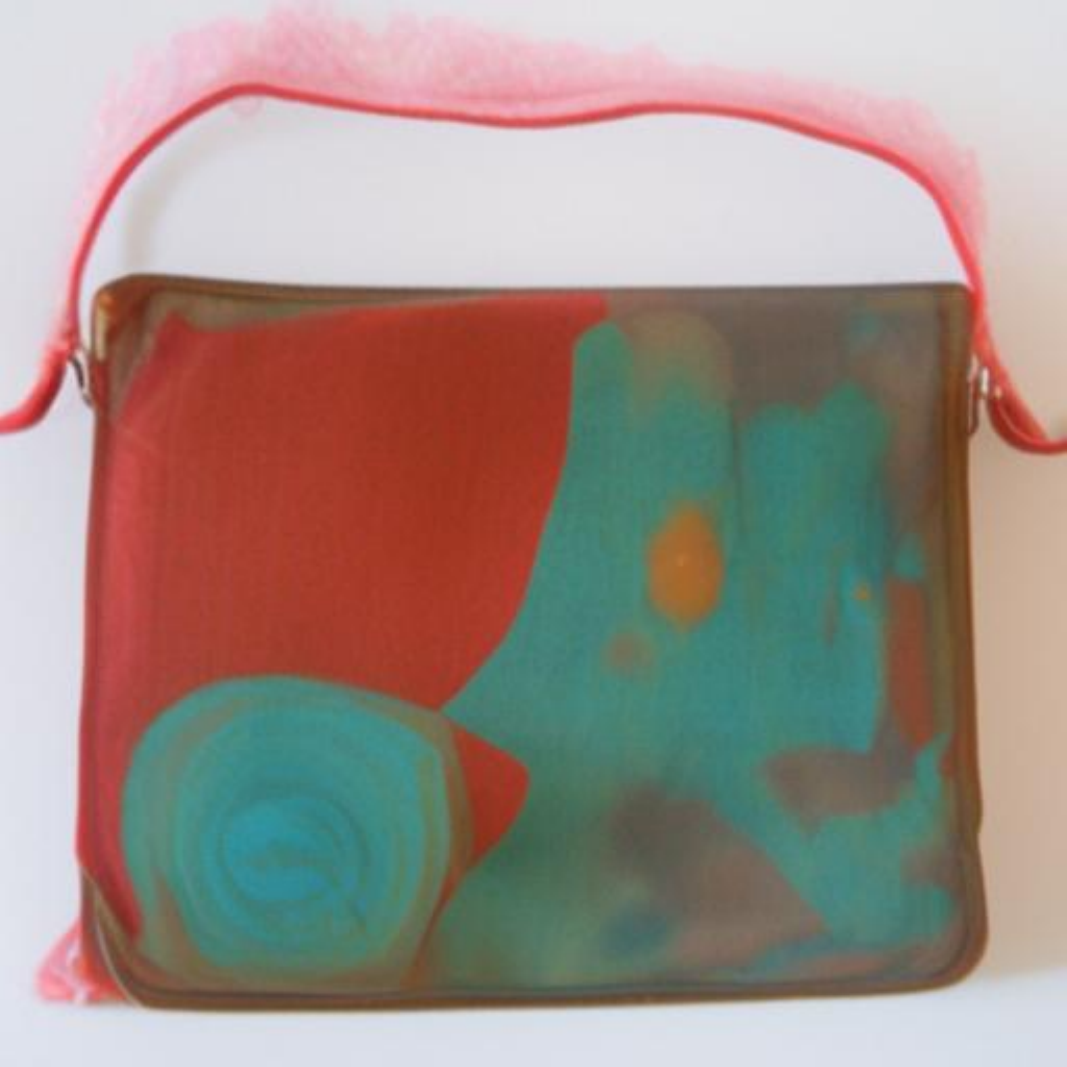}
\includegraphics[width=0.15\columnwidth]{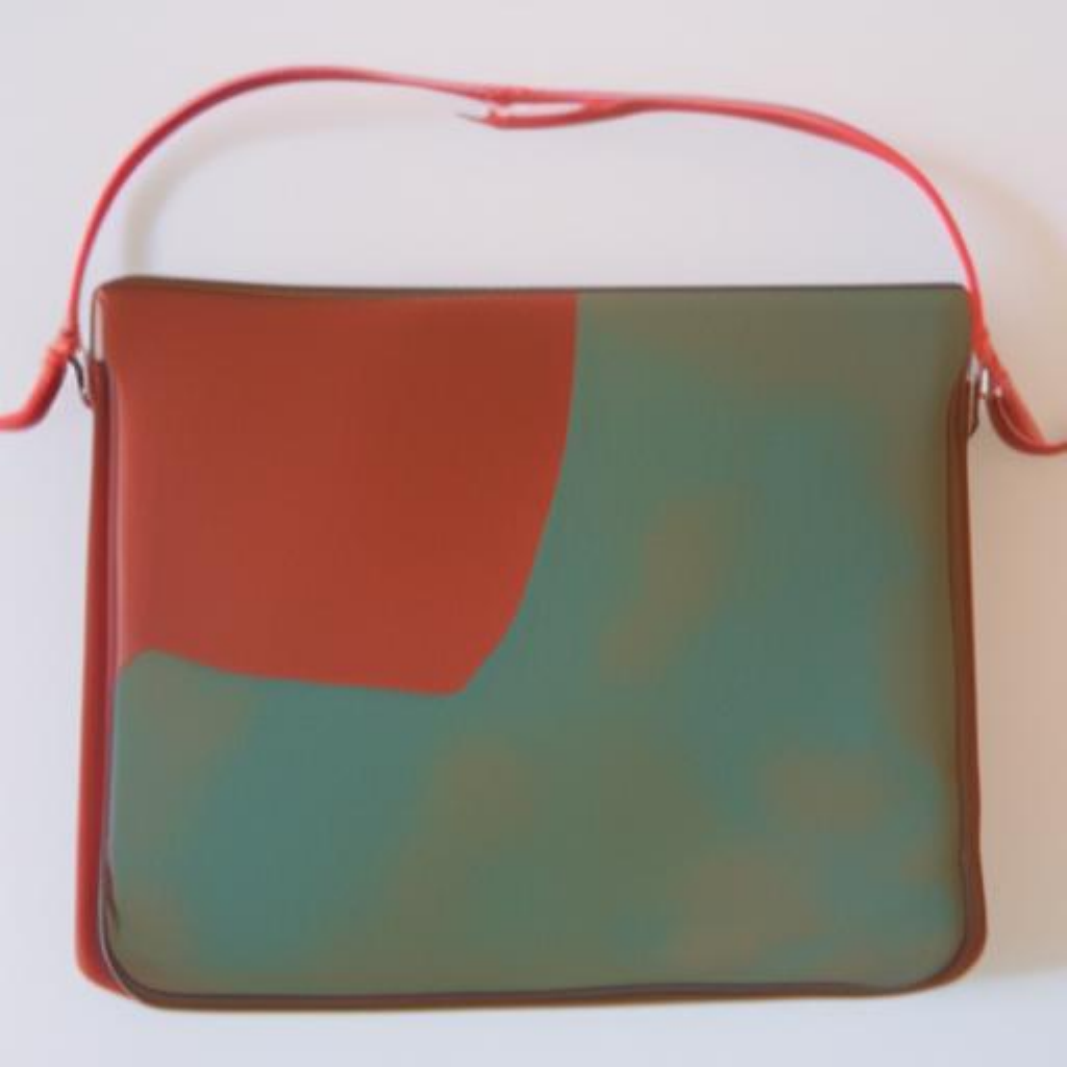}
\includegraphics[width=0.15\columnwidth]{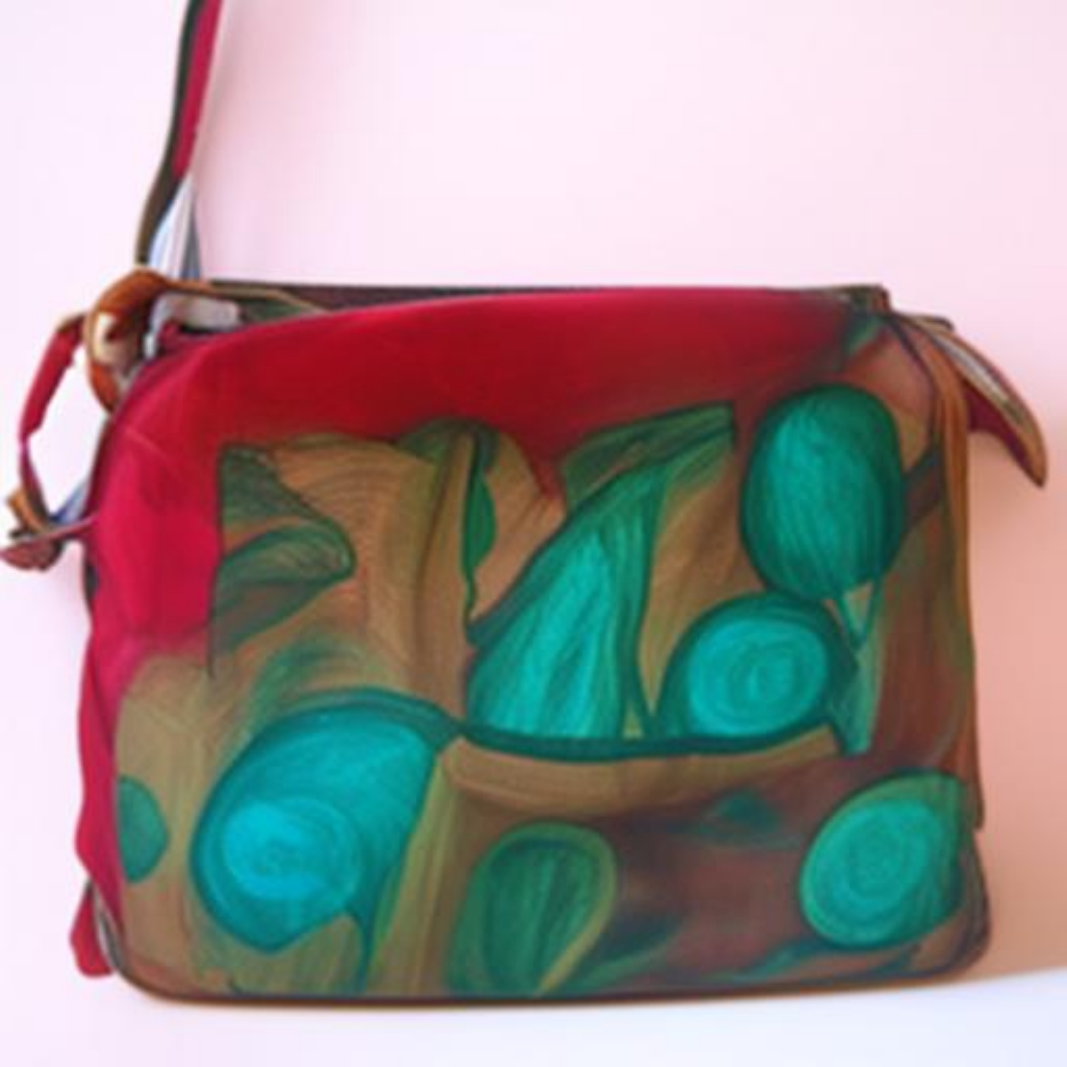}
\includegraphics[width=0.15\columnwidth]{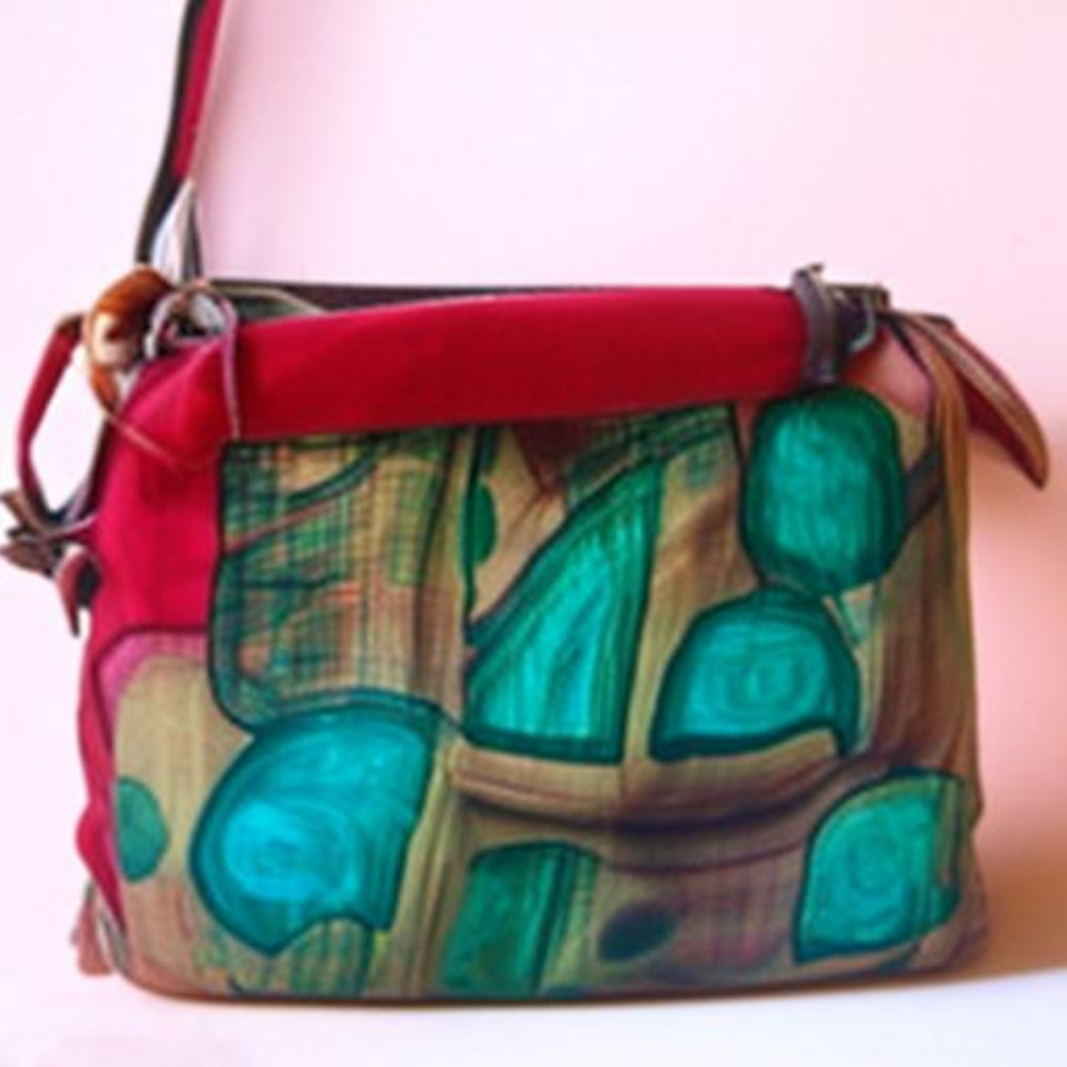}
\caption{\textbf{DiT with $40$ Steps.} For each row, from left to right: baseline, ablation, generated image with variance scaling $1.15$, and one with $1.35$.
%
The classes for each row are: \emph{clumber spaniel}, \emph{cottontail bunny}, \emph{tree frog}, \emph{miniature poodle}, \emph{whisky jug}, \emph{ambulance}, \emph{lampshade}, and \emph{mailbag}, respectively.
}
    \label{fig:dit-40}
    \end{figure}

\begin{figure}[t]
    \centering
\includegraphics[width=0.15\columnwidth]{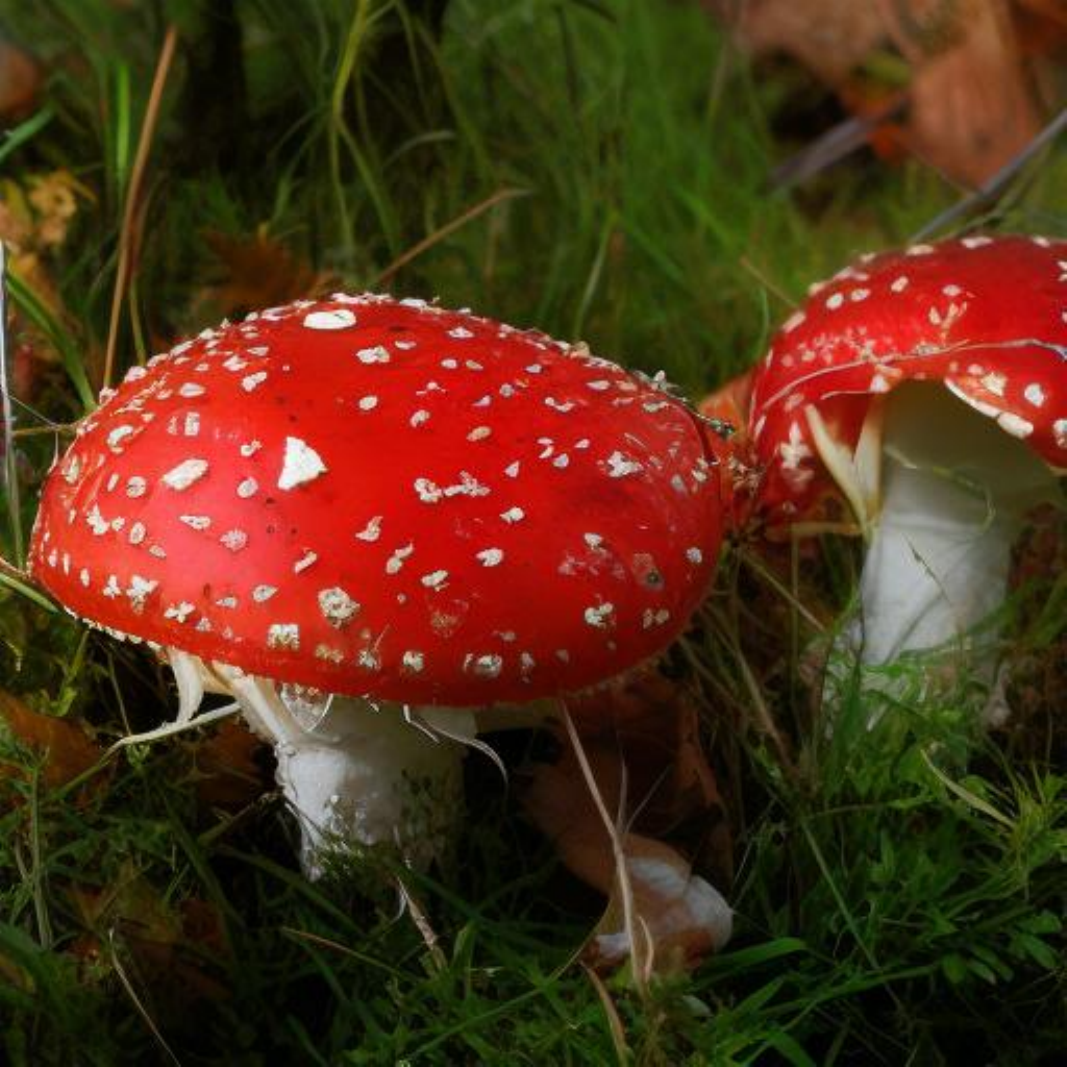}
\includegraphics[width=0.15\columnwidth]{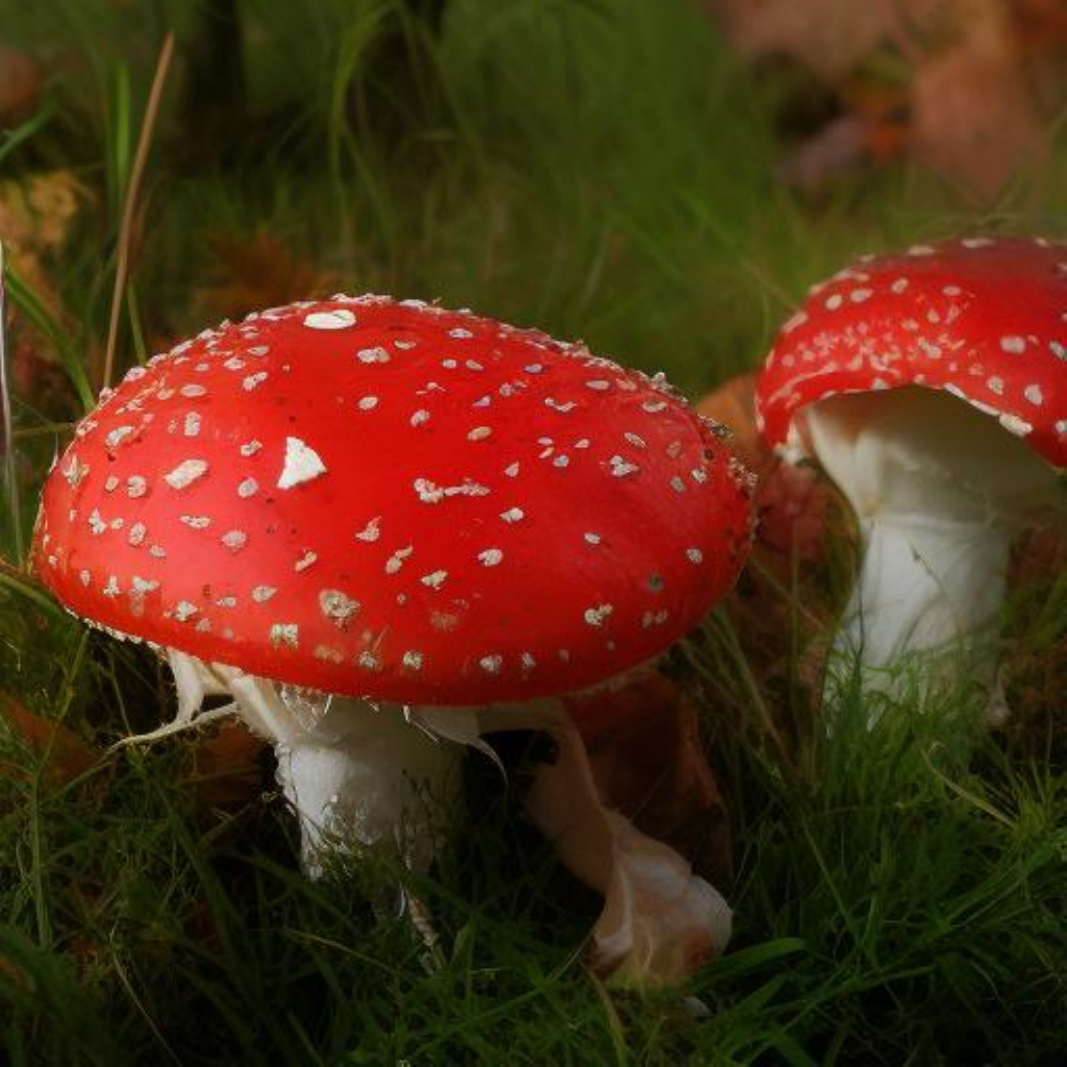}
\includegraphics[width=0.15\columnwidth]{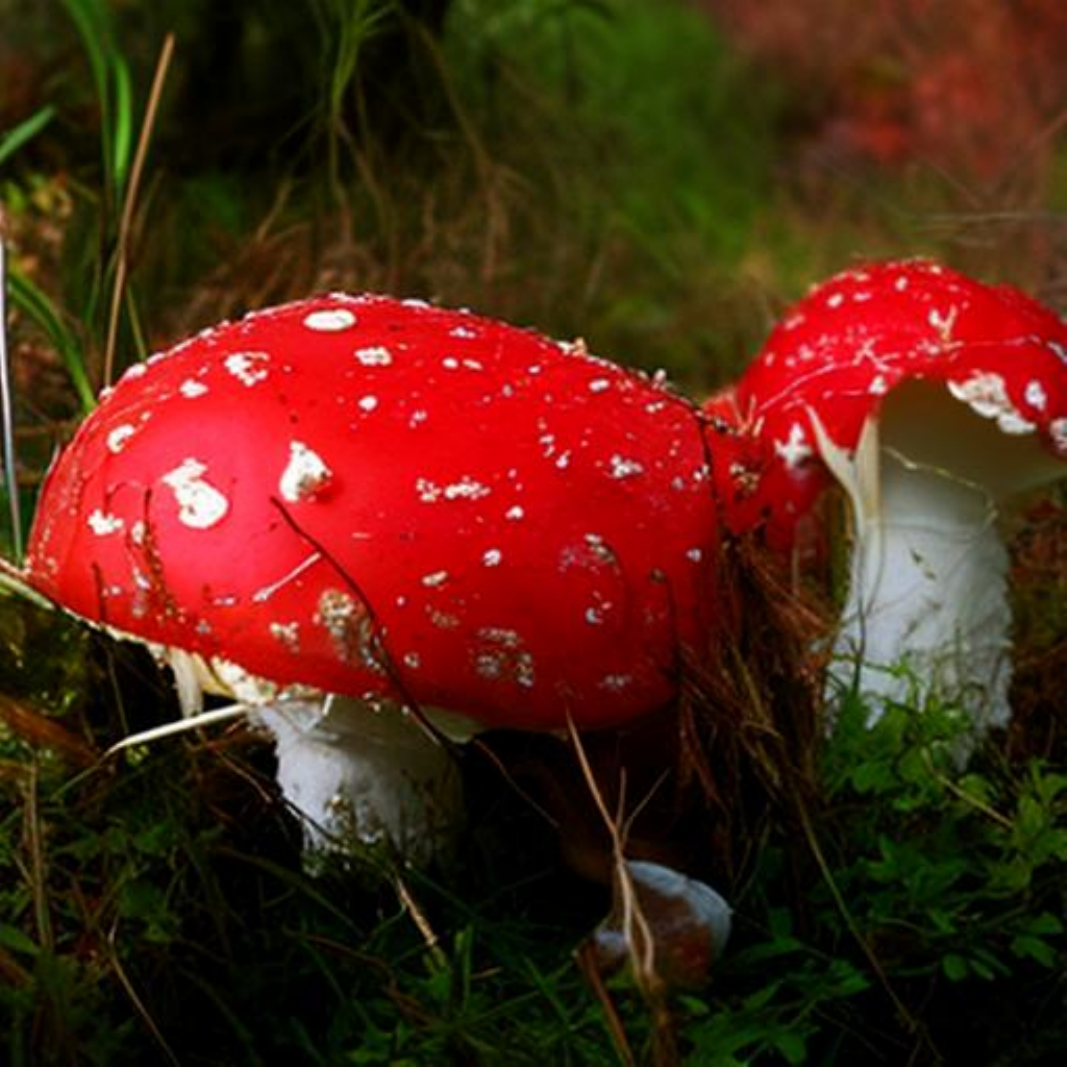}
\includegraphics[width=0.15\columnwidth]{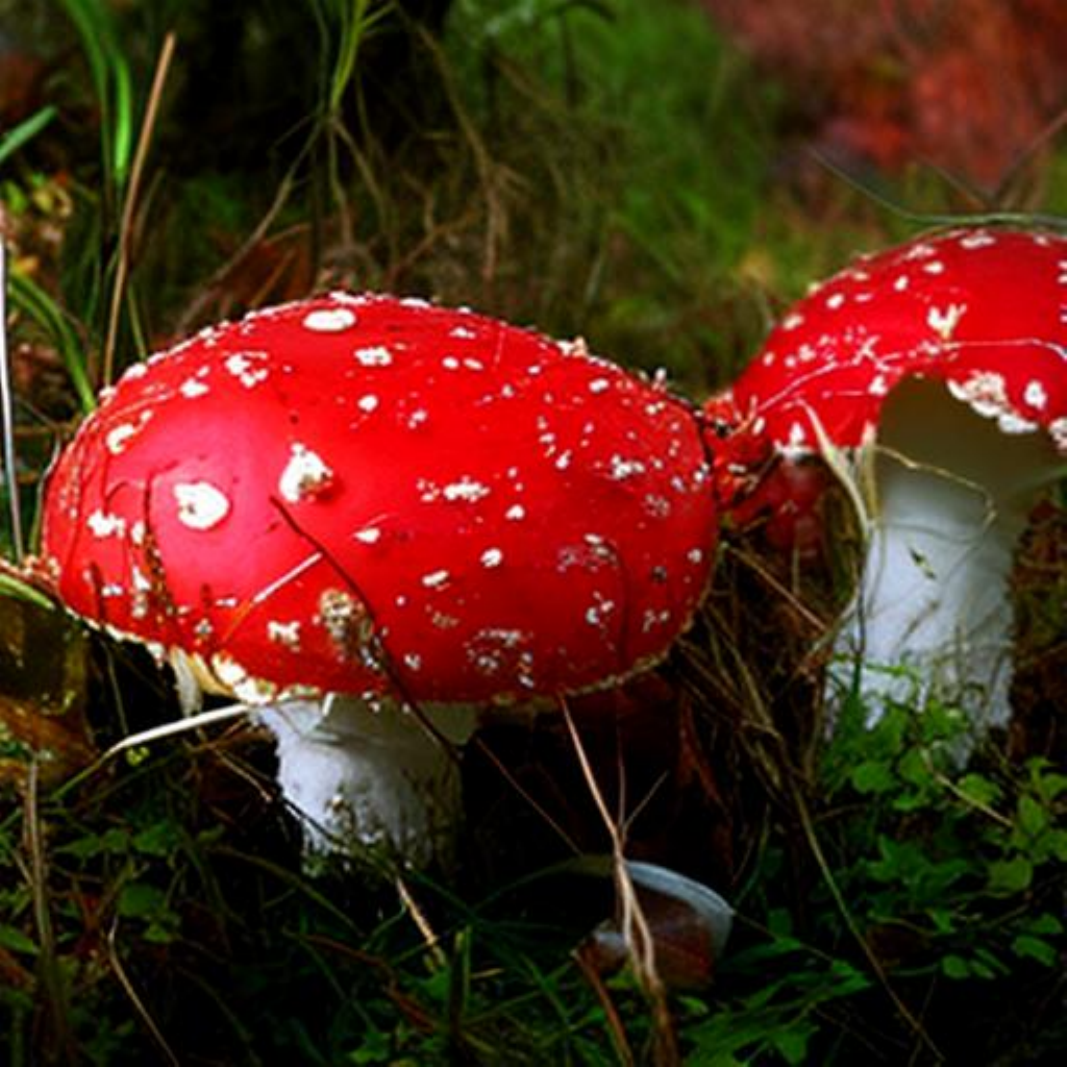}
\\
\includegraphics[width=0.15\columnwidth]{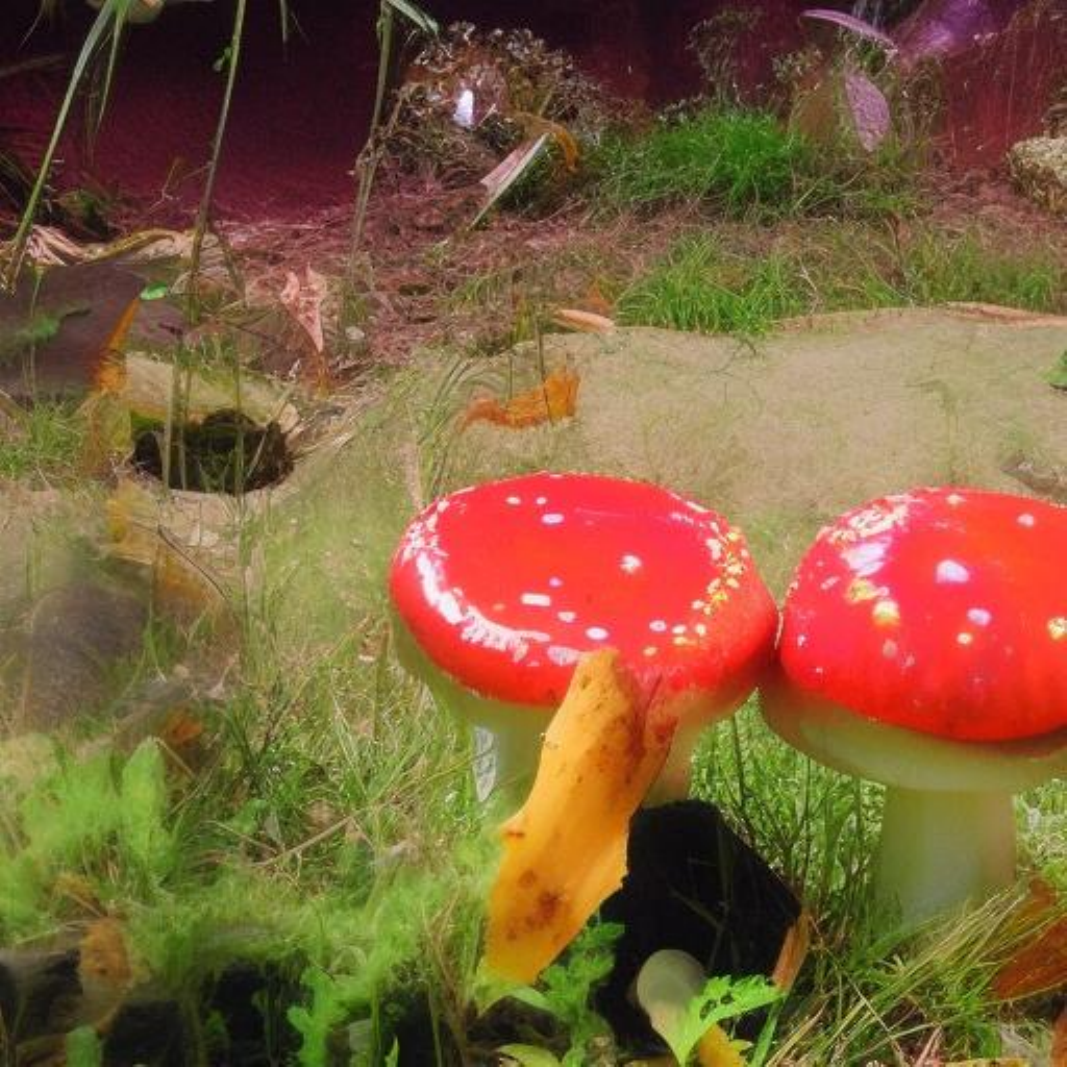}
\includegraphics[width=0.15\columnwidth]{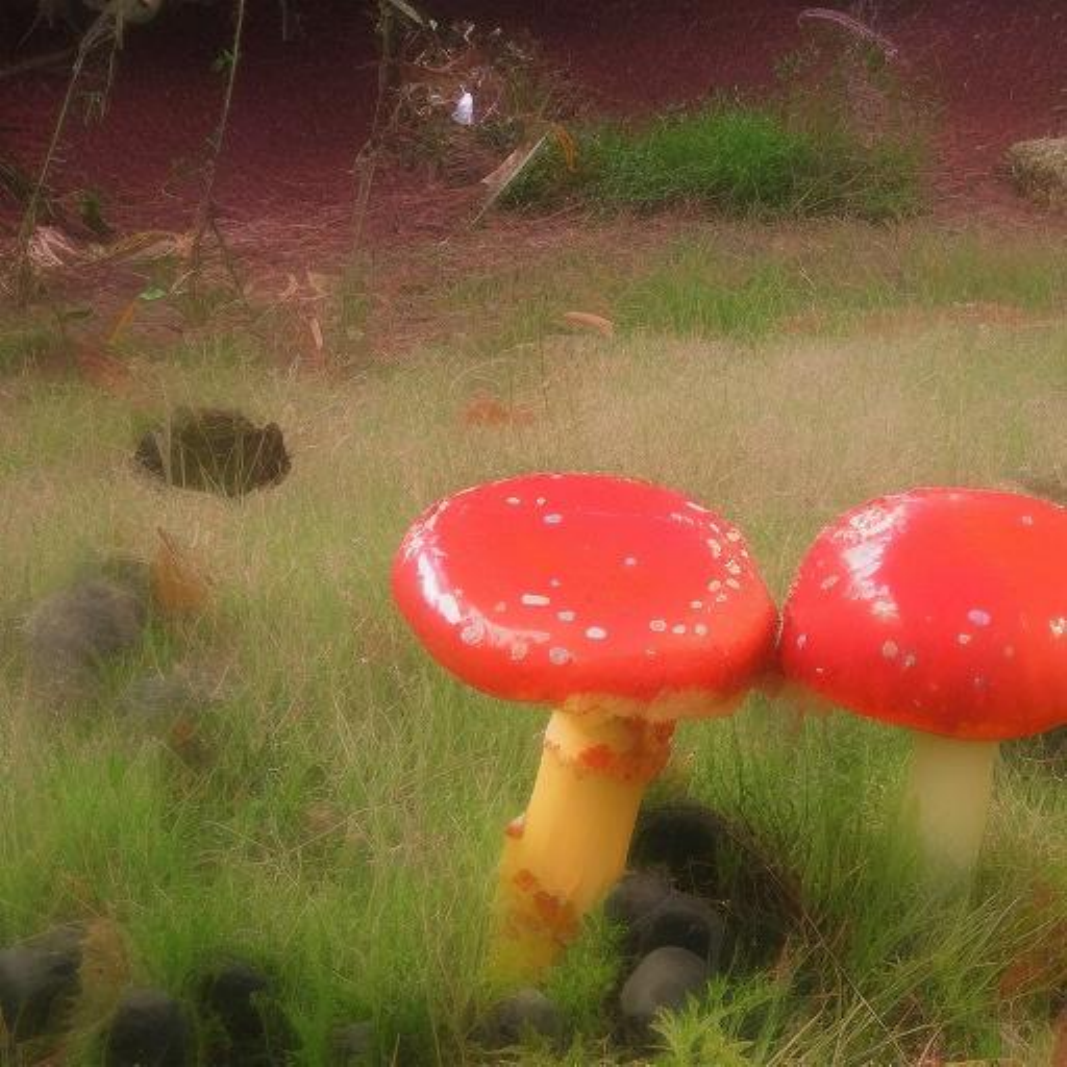}
\includegraphics[width=0.15\columnwidth]{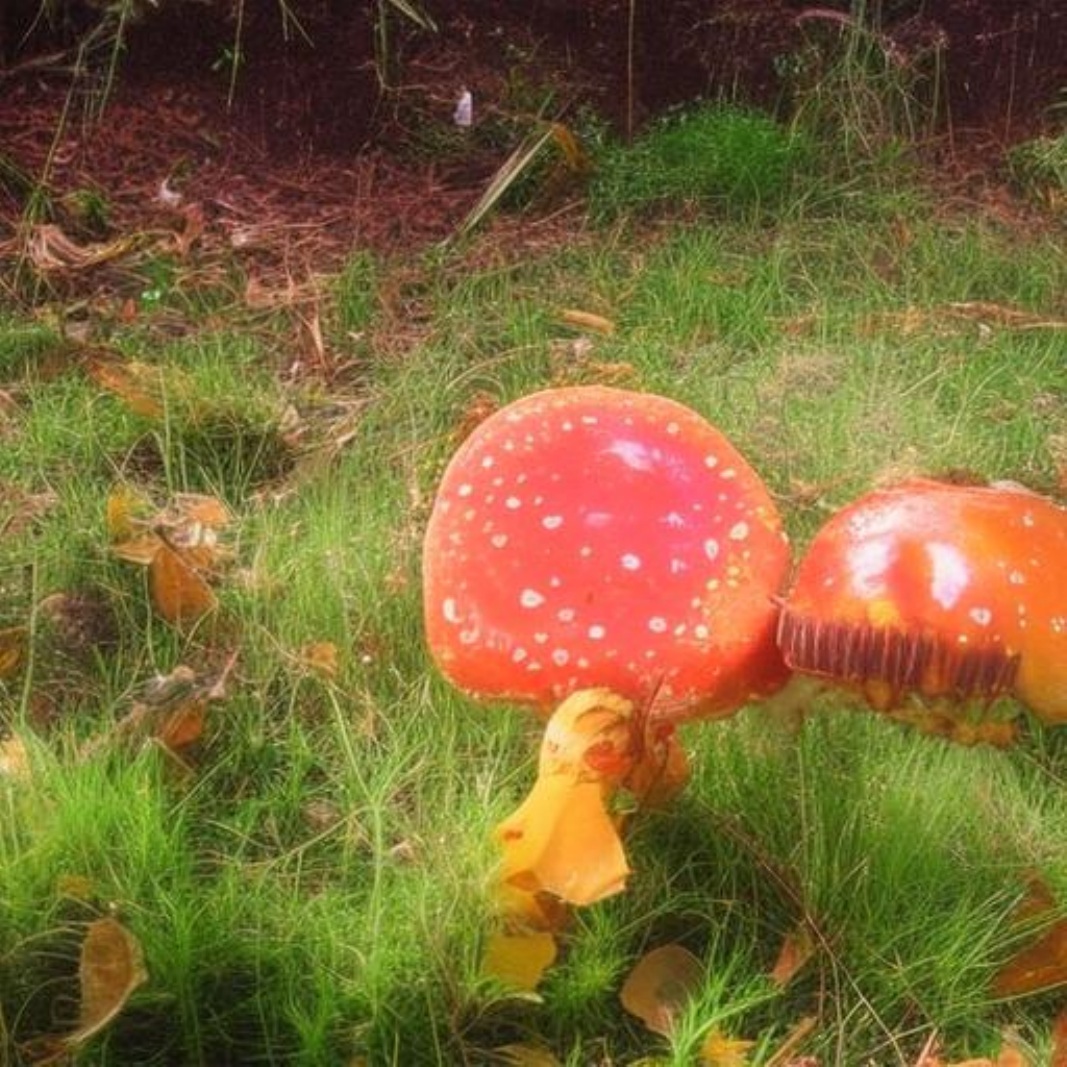}
\includegraphics[width=0.15\columnwidth]{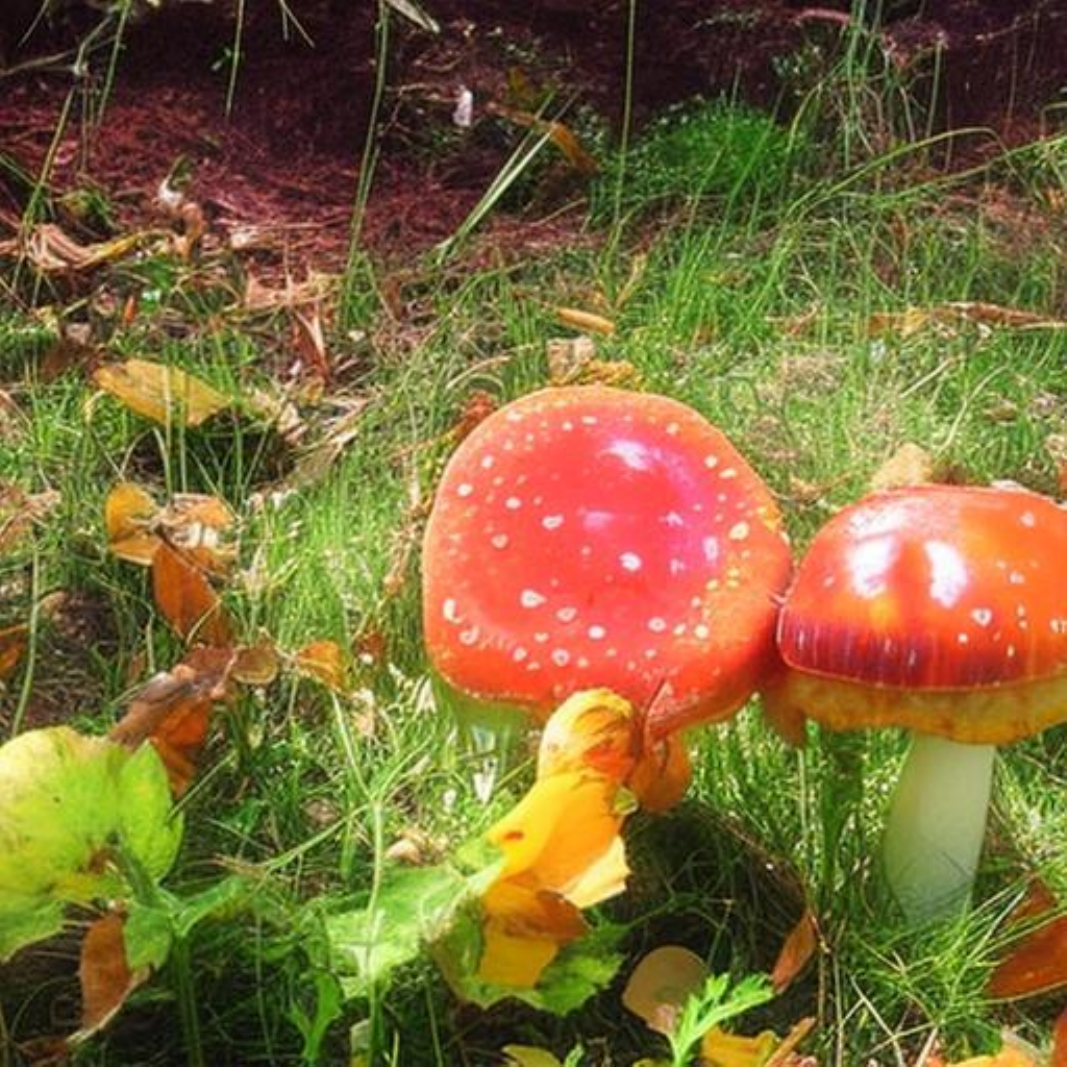}
\\
\includegraphics[width=0.15\columnwidth]{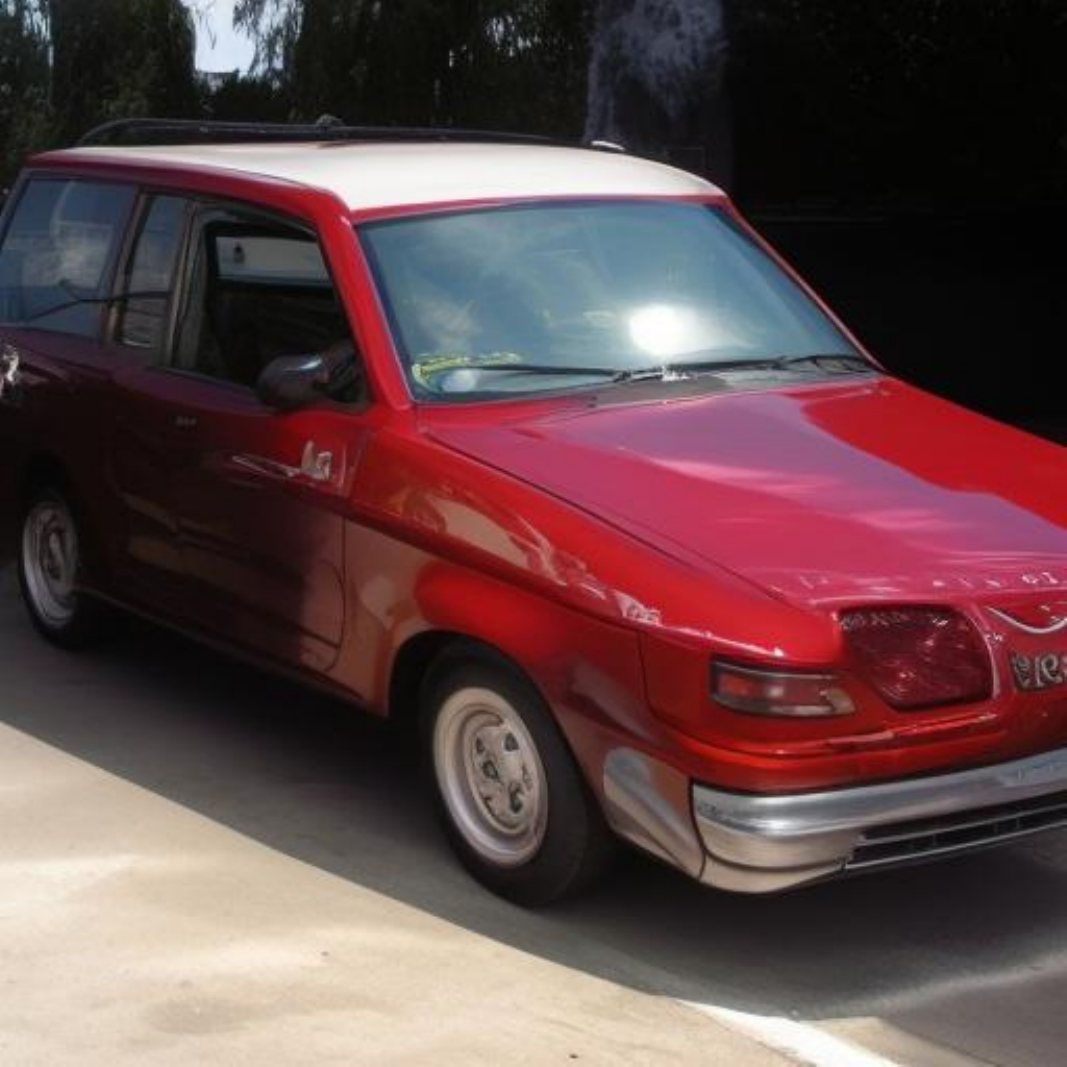}
\includegraphics[width=0.15\columnwidth]{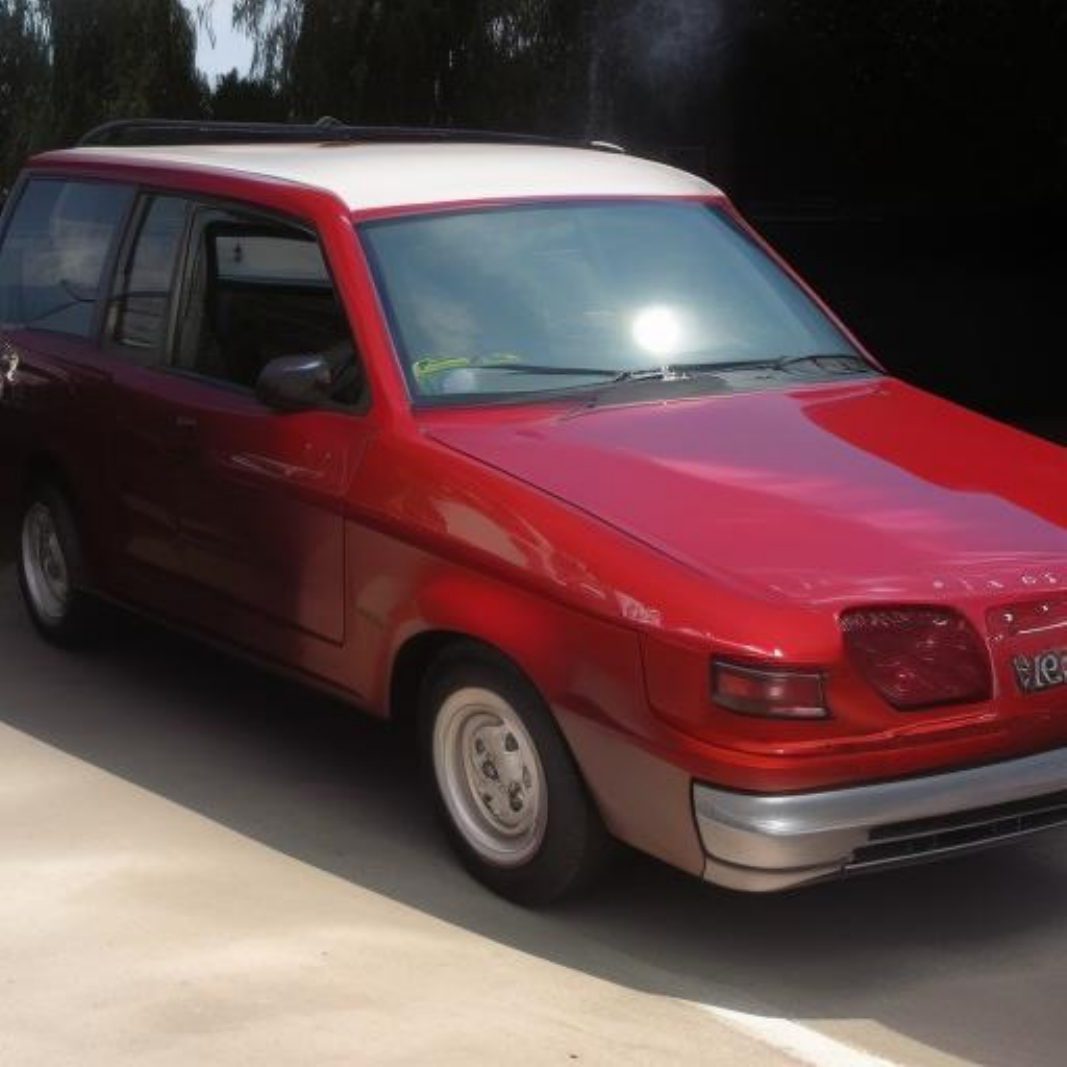}
\includegraphics[width=0.15\columnwidth]{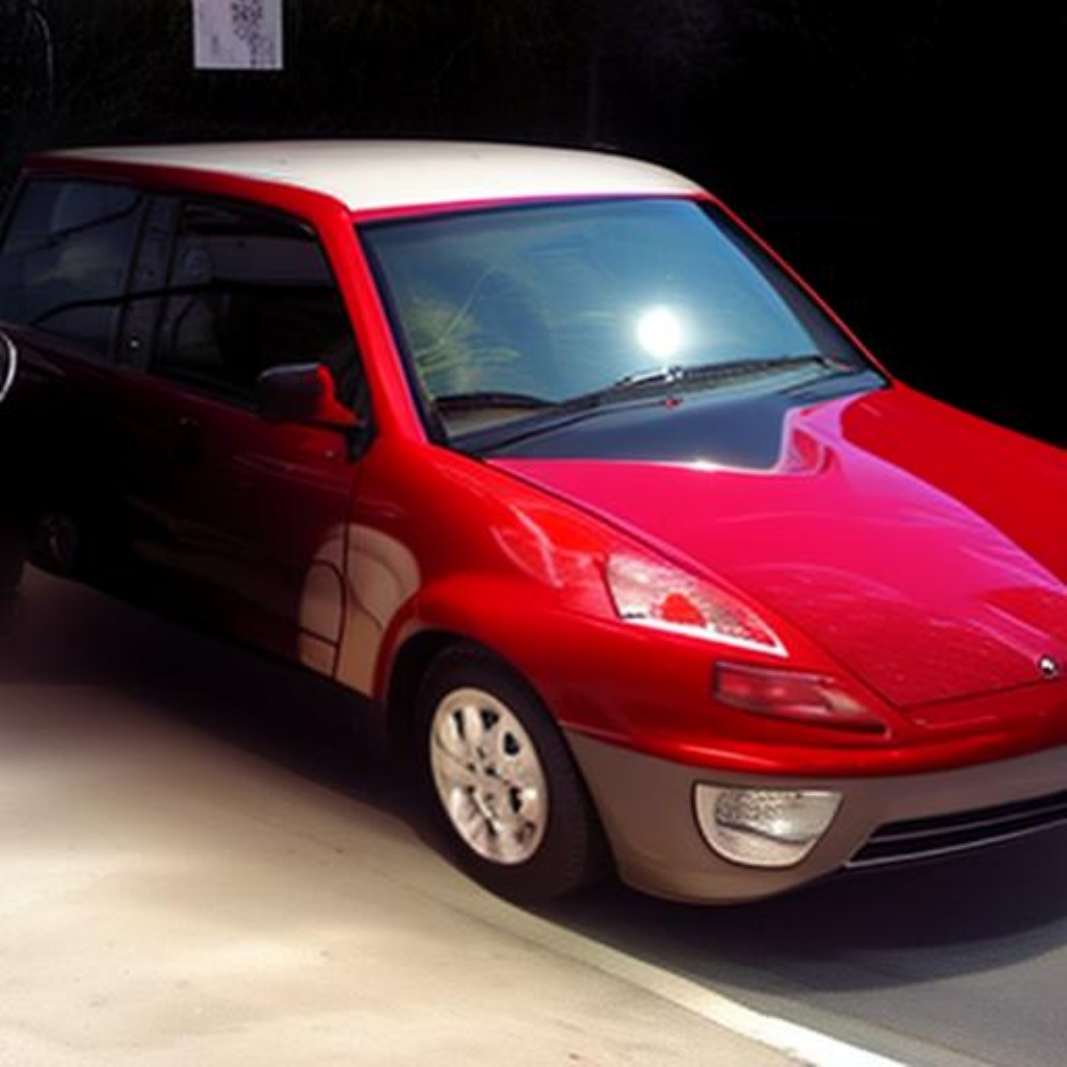}
\includegraphics[width=0.15\columnwidth]{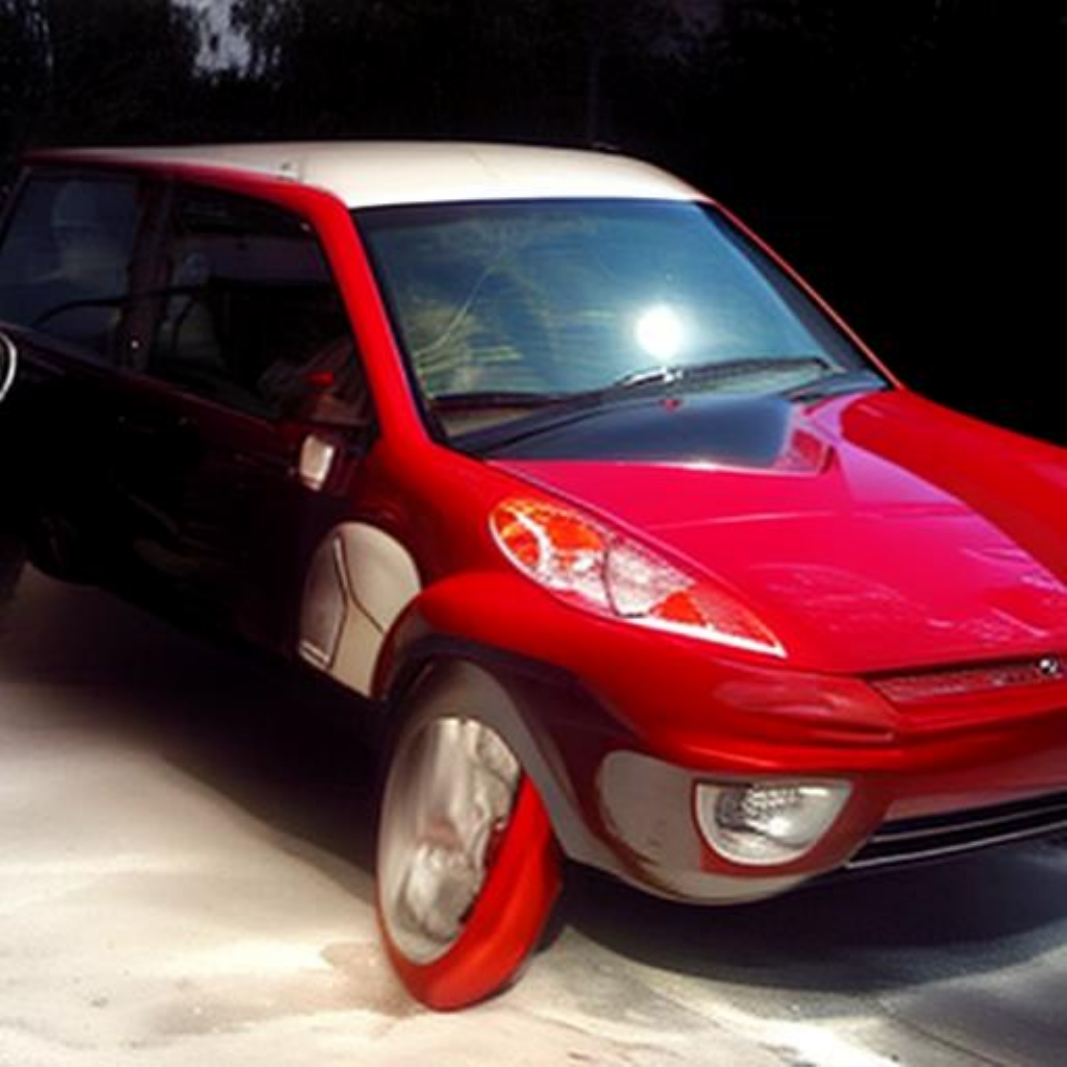}
\\
\includegraphics[width=0.15\columnwidth]{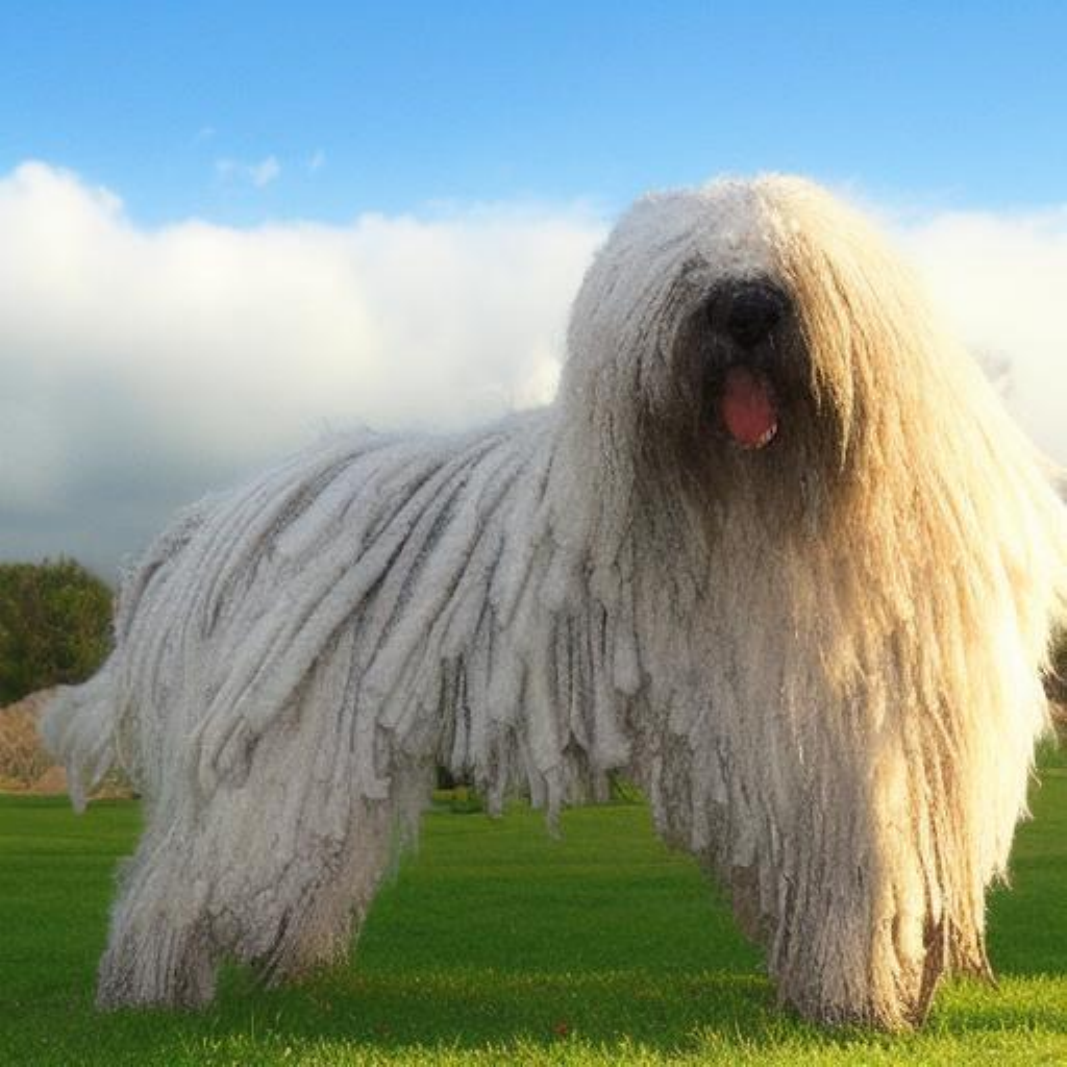}
\includegraphics[width=0.15\columnwidth]{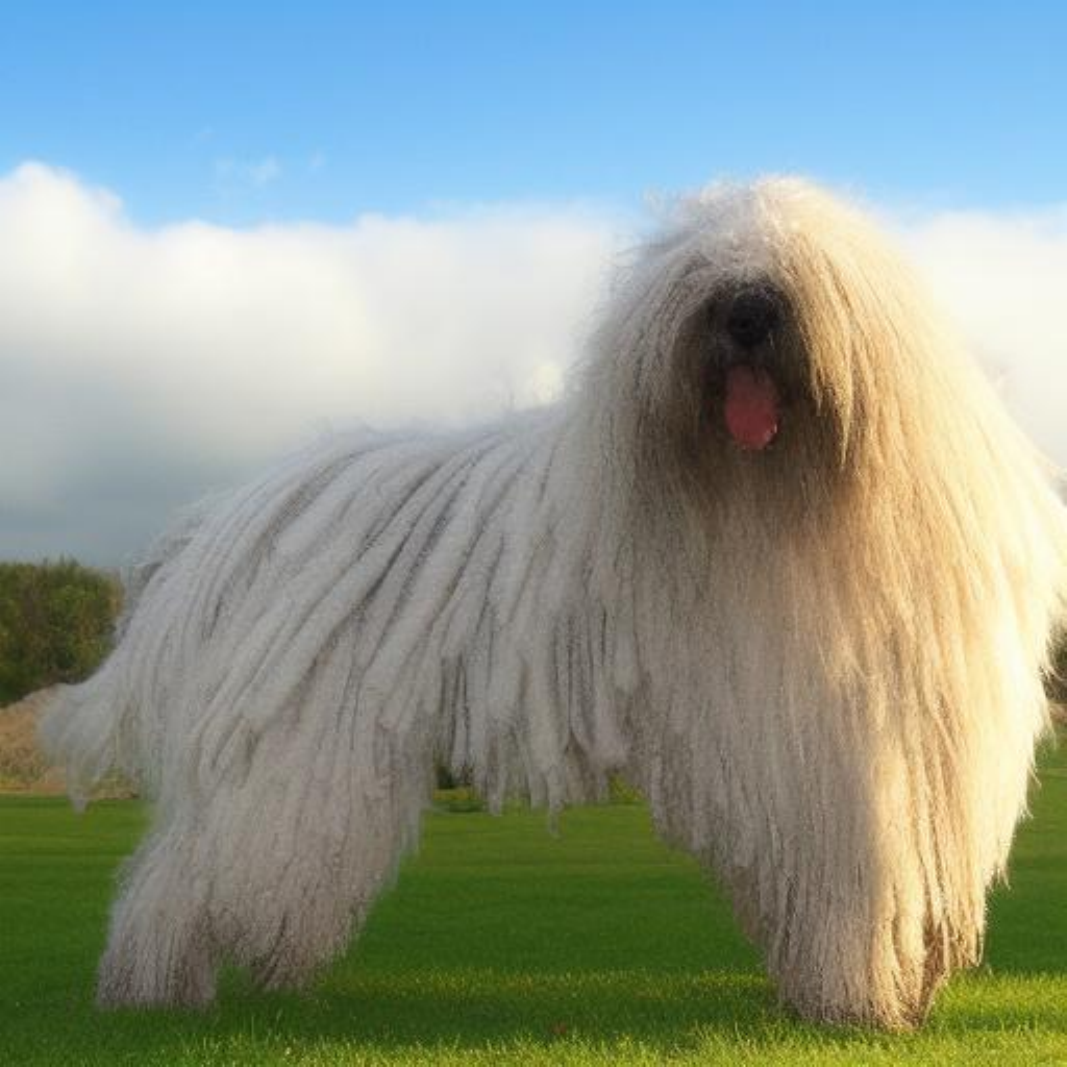}
\includegraphics[width=0.15\columnwidth]{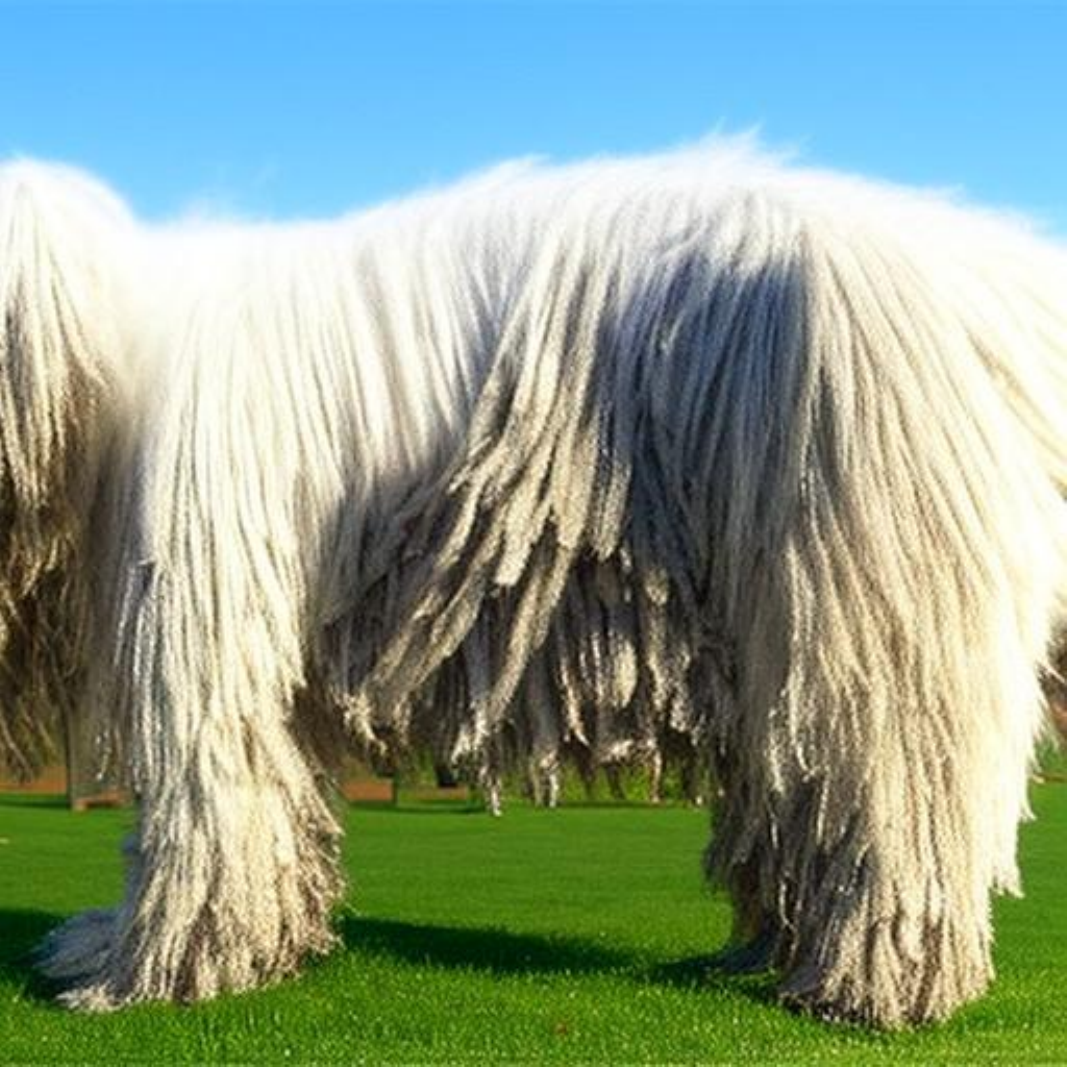}
\includegraphics[width=0.15\columnwidth]{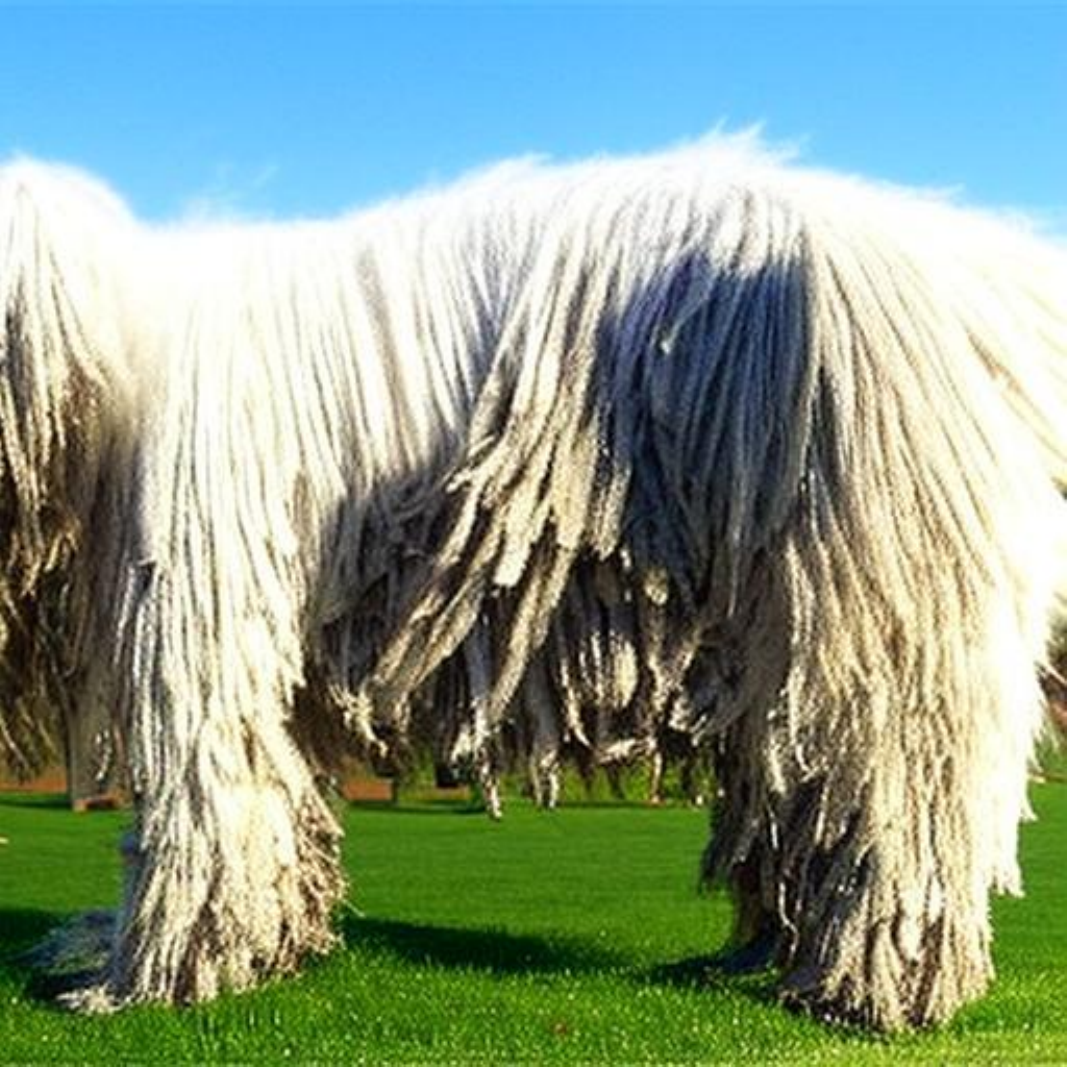}
\caption{\textbf{Phenomena for DiT with $40$ Steps.} \textbf{Rows 1-2:} For the same class of pictures (\emph{agaric}), SE2P can both increase and decrease intensity. \textbf{Rows 2-3:} Examples of shortcomings (\emph{wagon} and \emph{komondor}, respectively).
}
    \label{fig:dit-40-things}
    \end{figure}

\begin{figure}[t]
    \centering
\includegraphics[width=0.15\columnwidth]{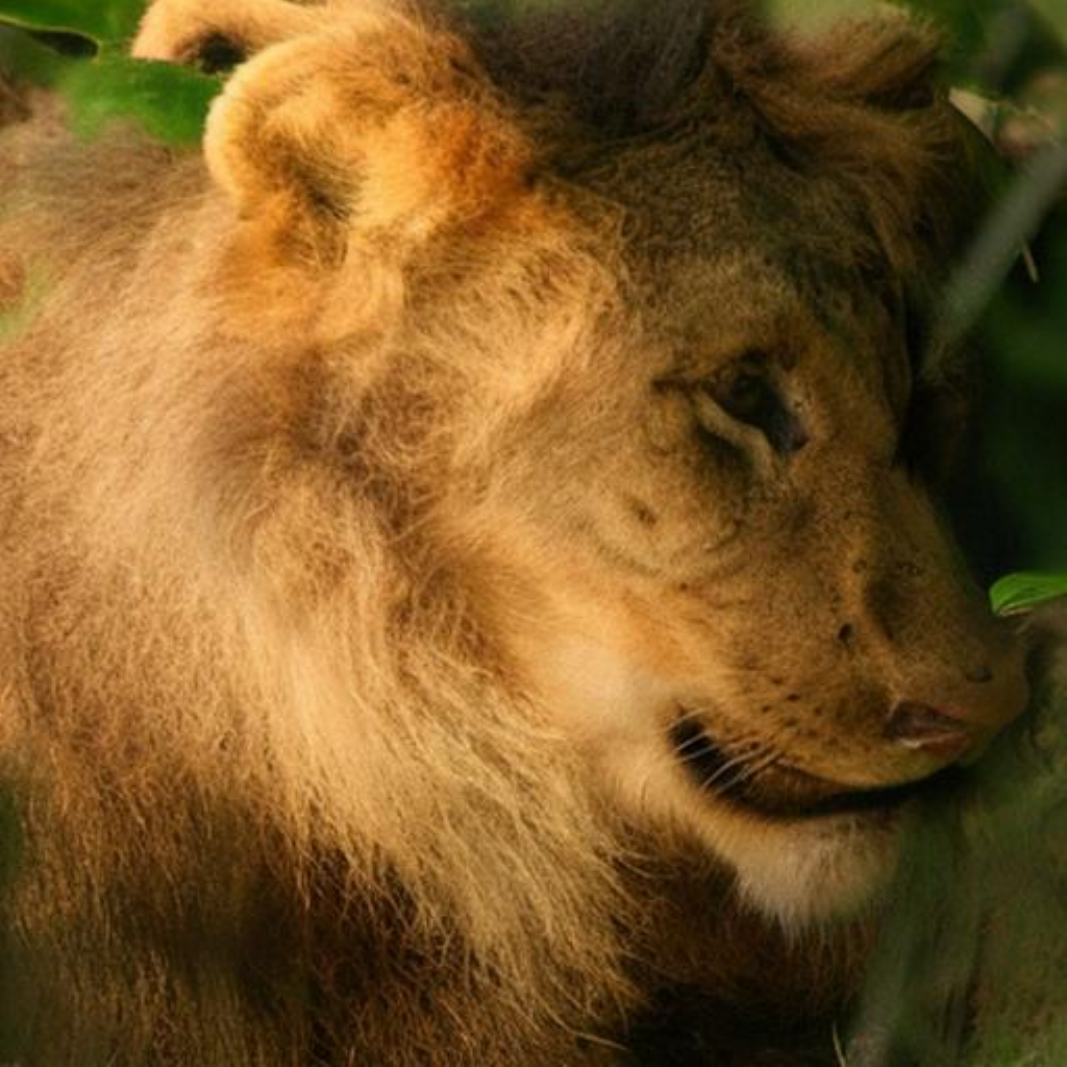}
\includegraphics[width=0.15\columnwidth]{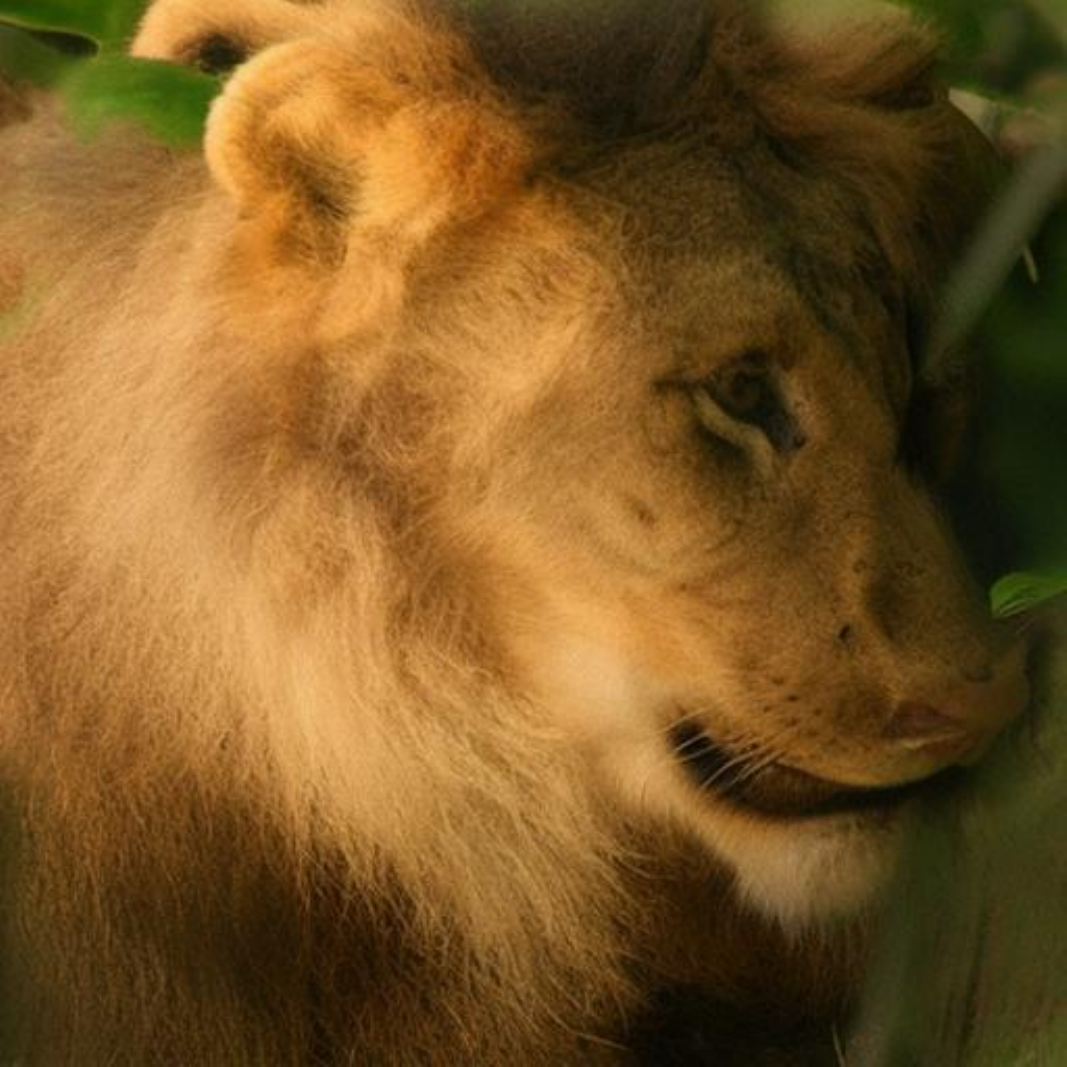}
\includegraphics[width=0.15\columnwidth]{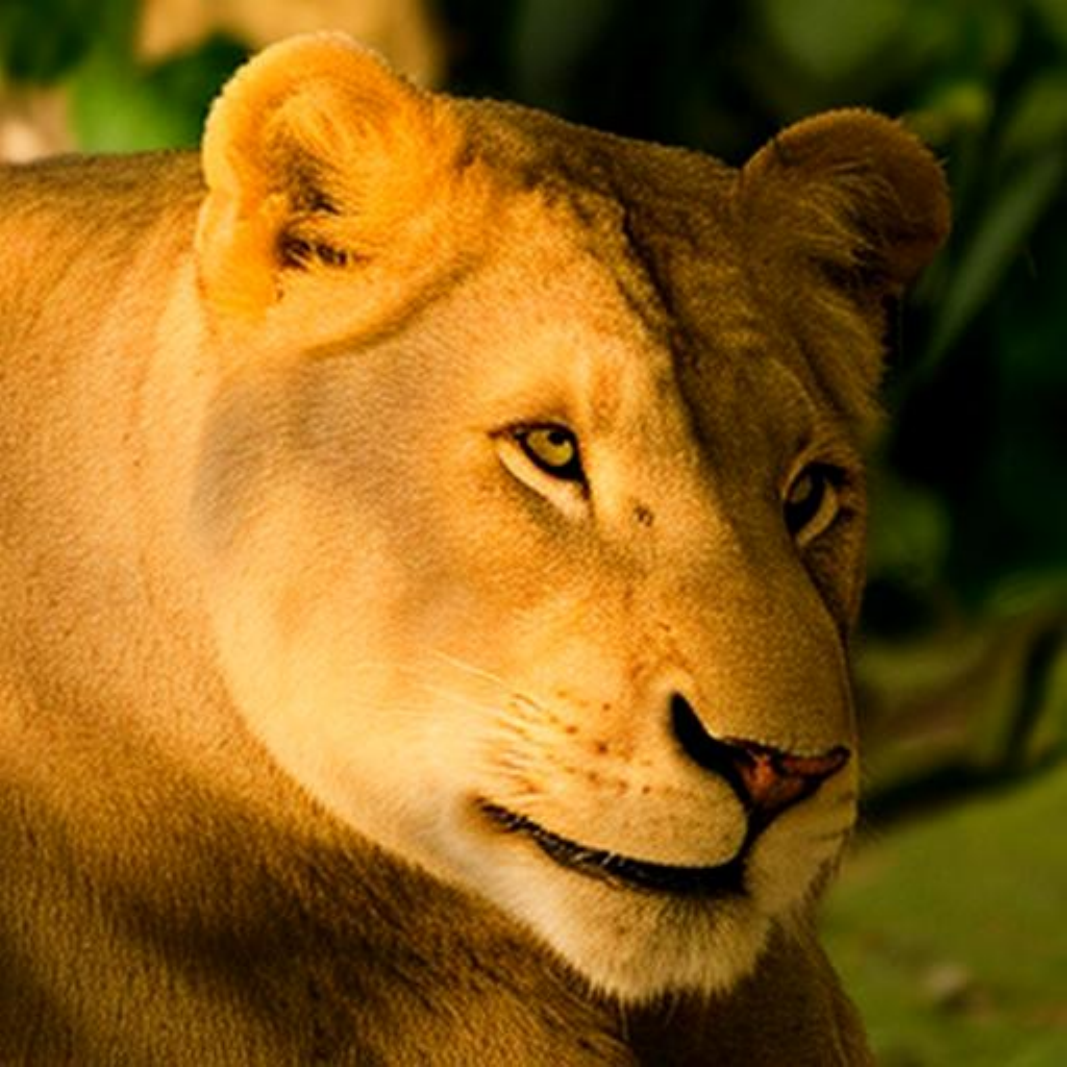}
\includegraphics[width=0.15\columnwidth]{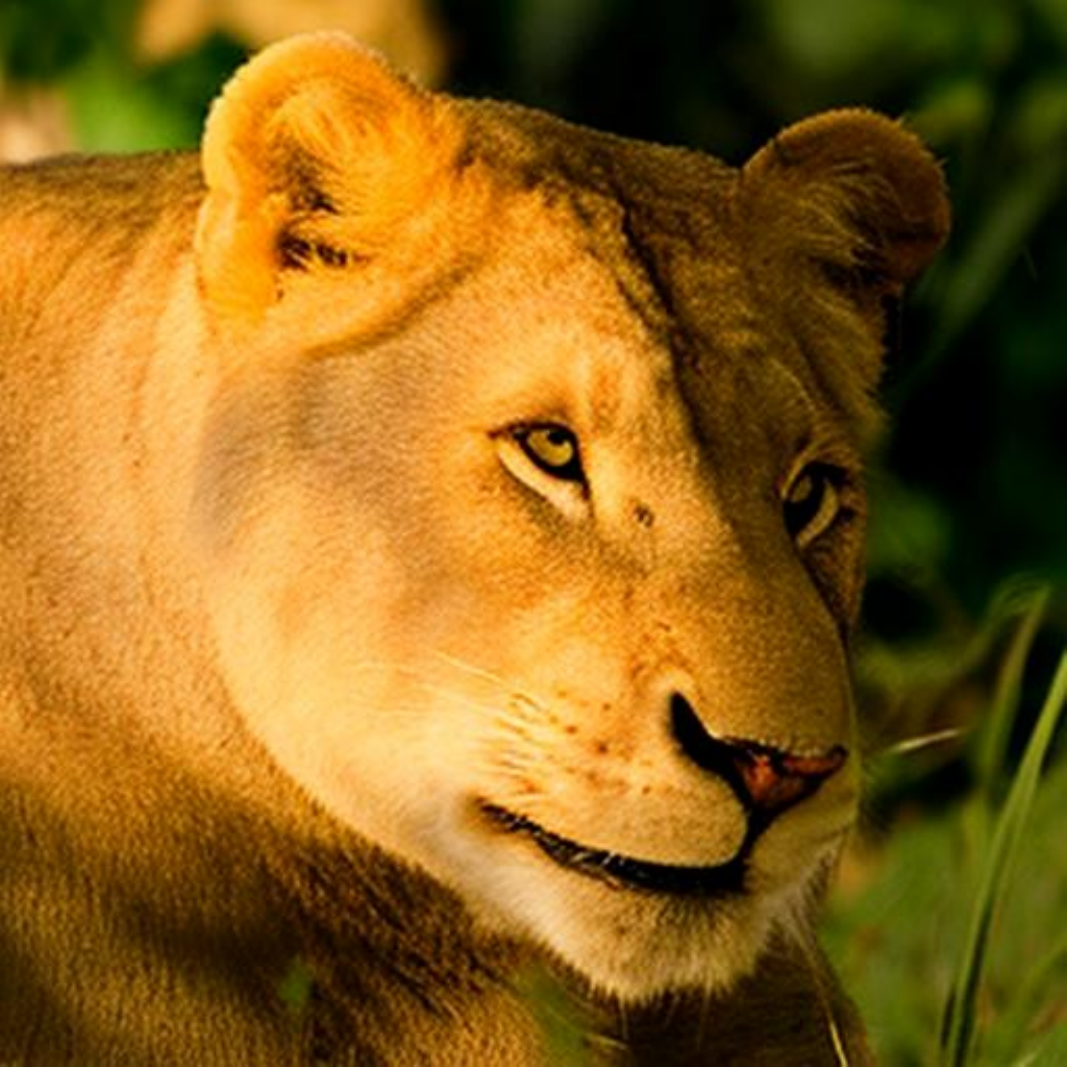}
\\
\includegraphics[width=0.15\columnwidth]{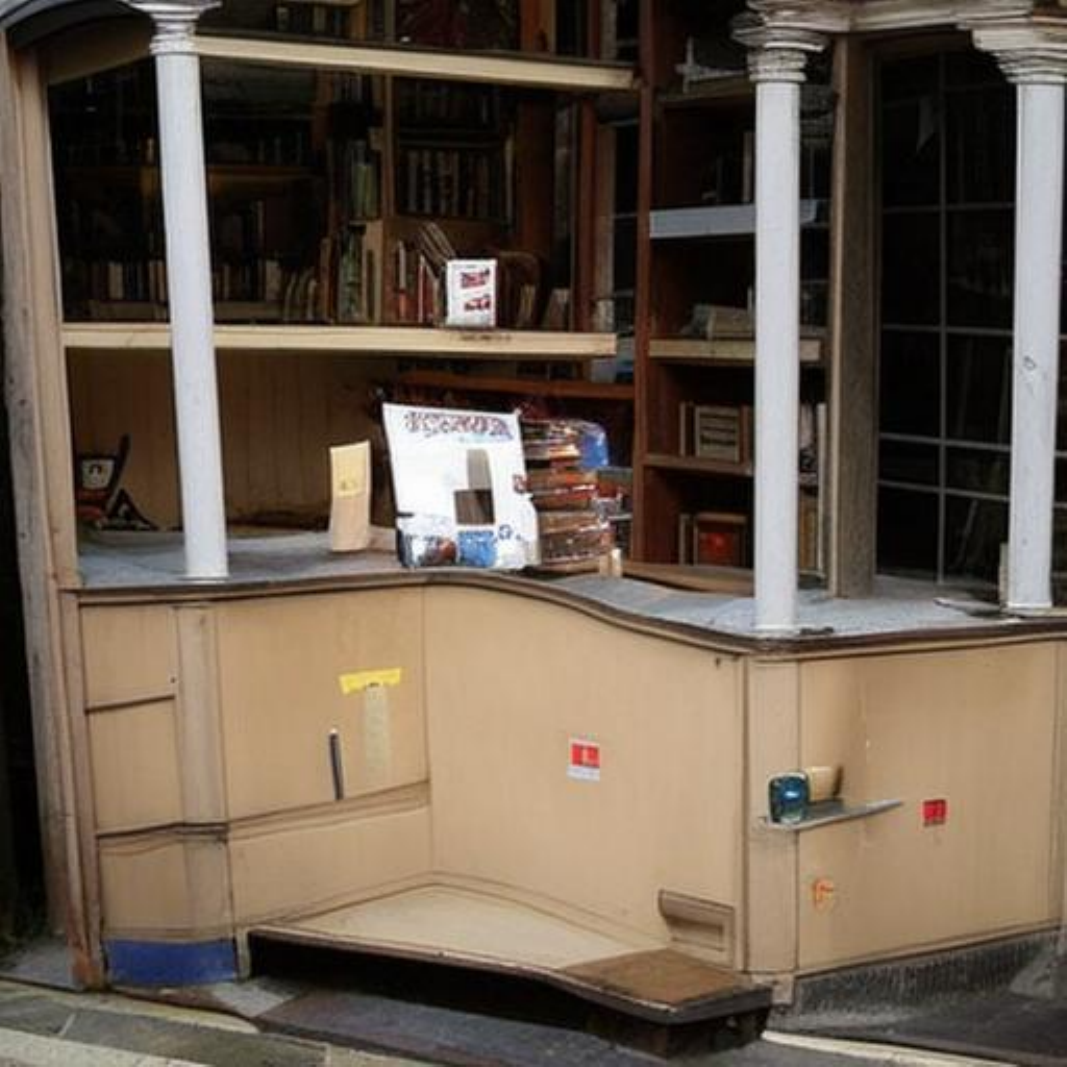}
\includegraphics[width=0.15\columnwidth]{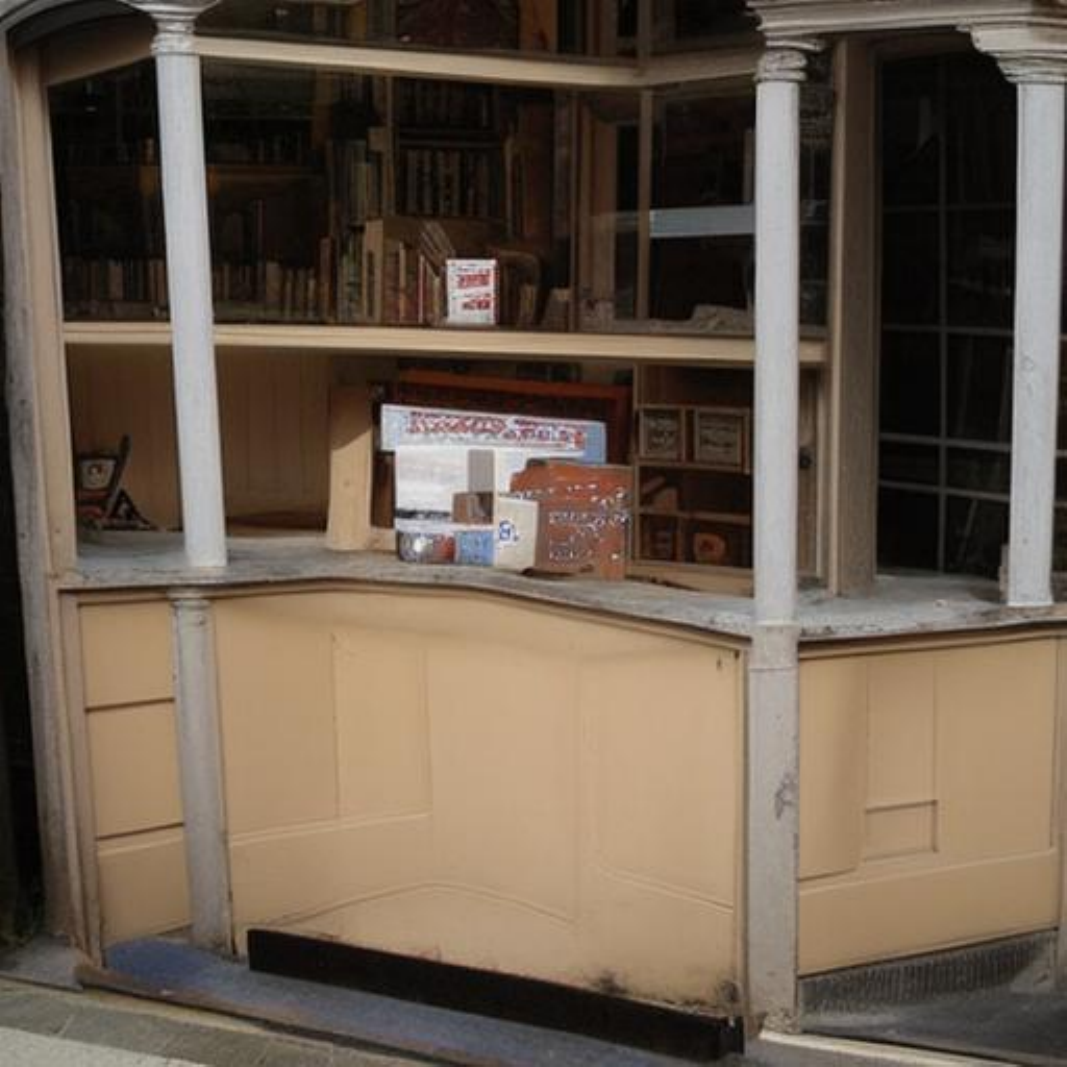}
\includegraphics[width=0.15\columnwidth]{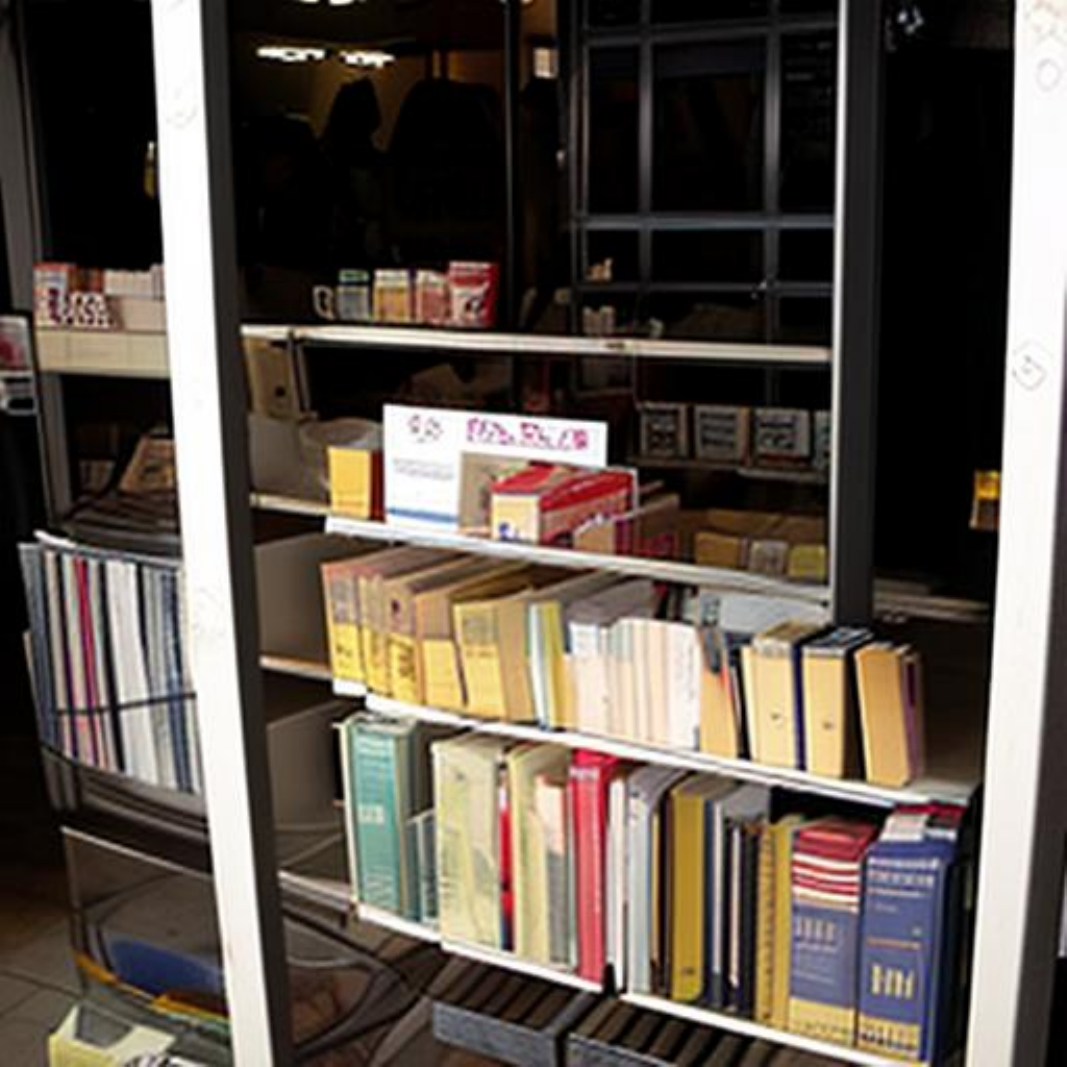}
\includegraphics[width=0.15\columnwidth]{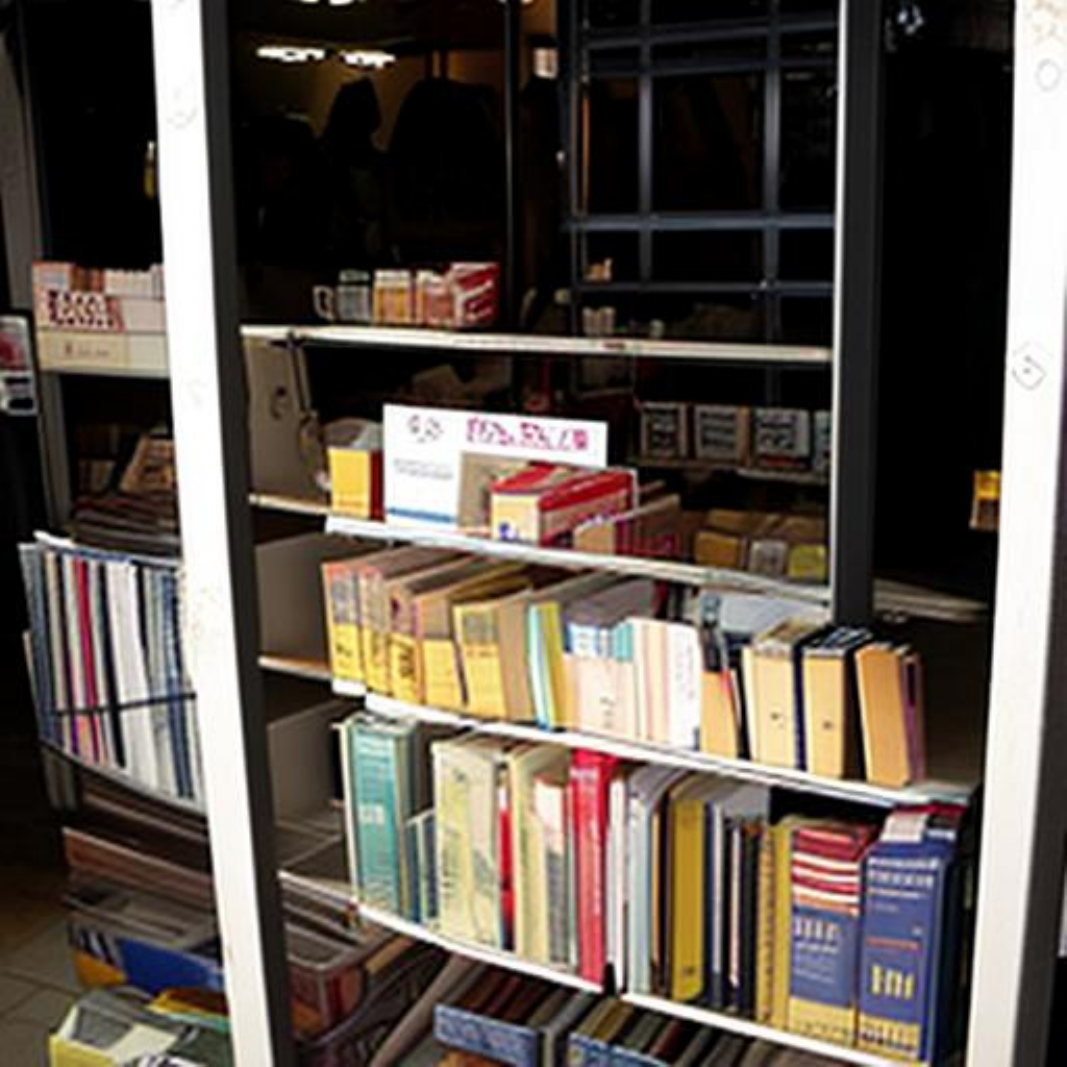}
\\
\includegraphics[width=0.15\columnwidth]{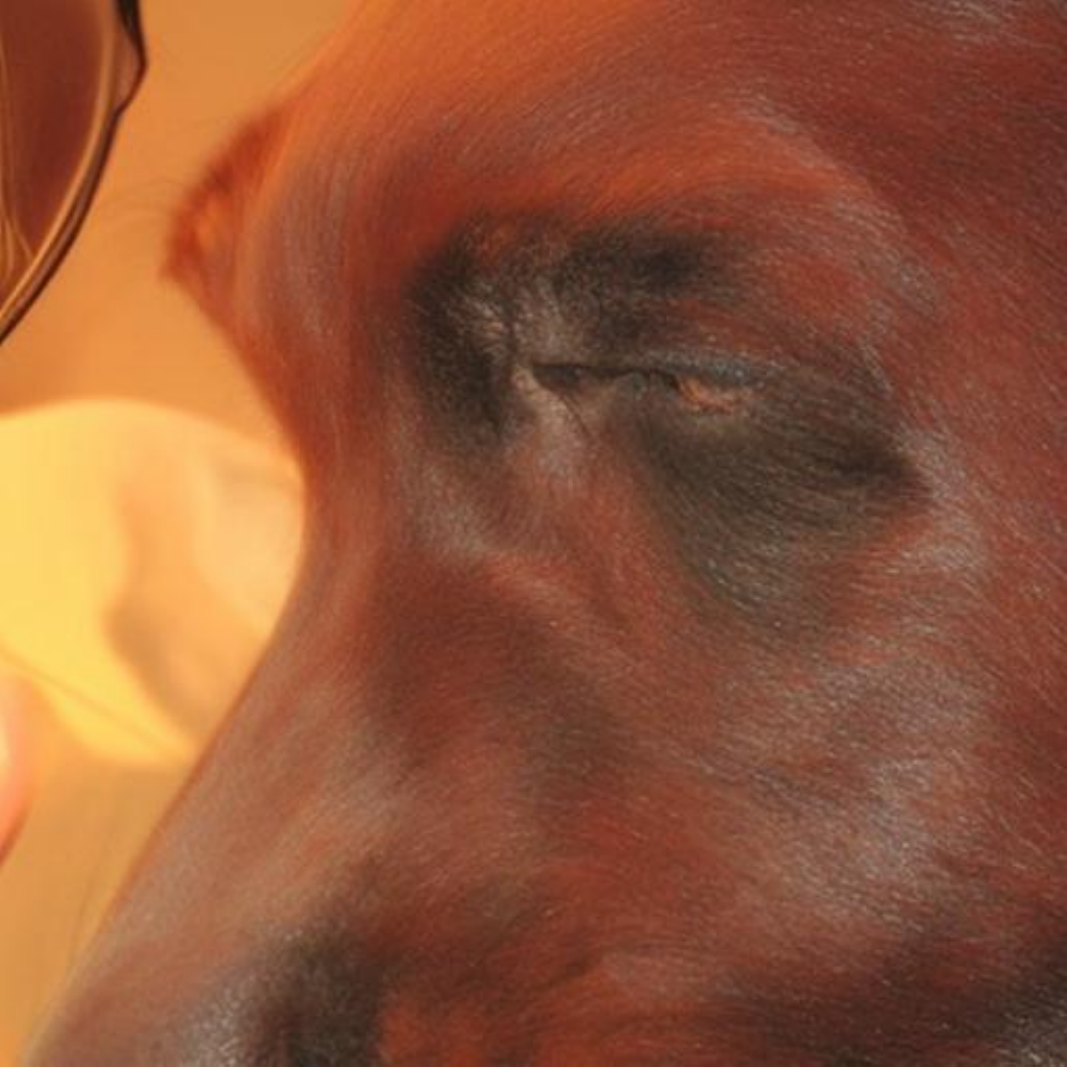}
\includegraphics[width=0.15\columnwidth]{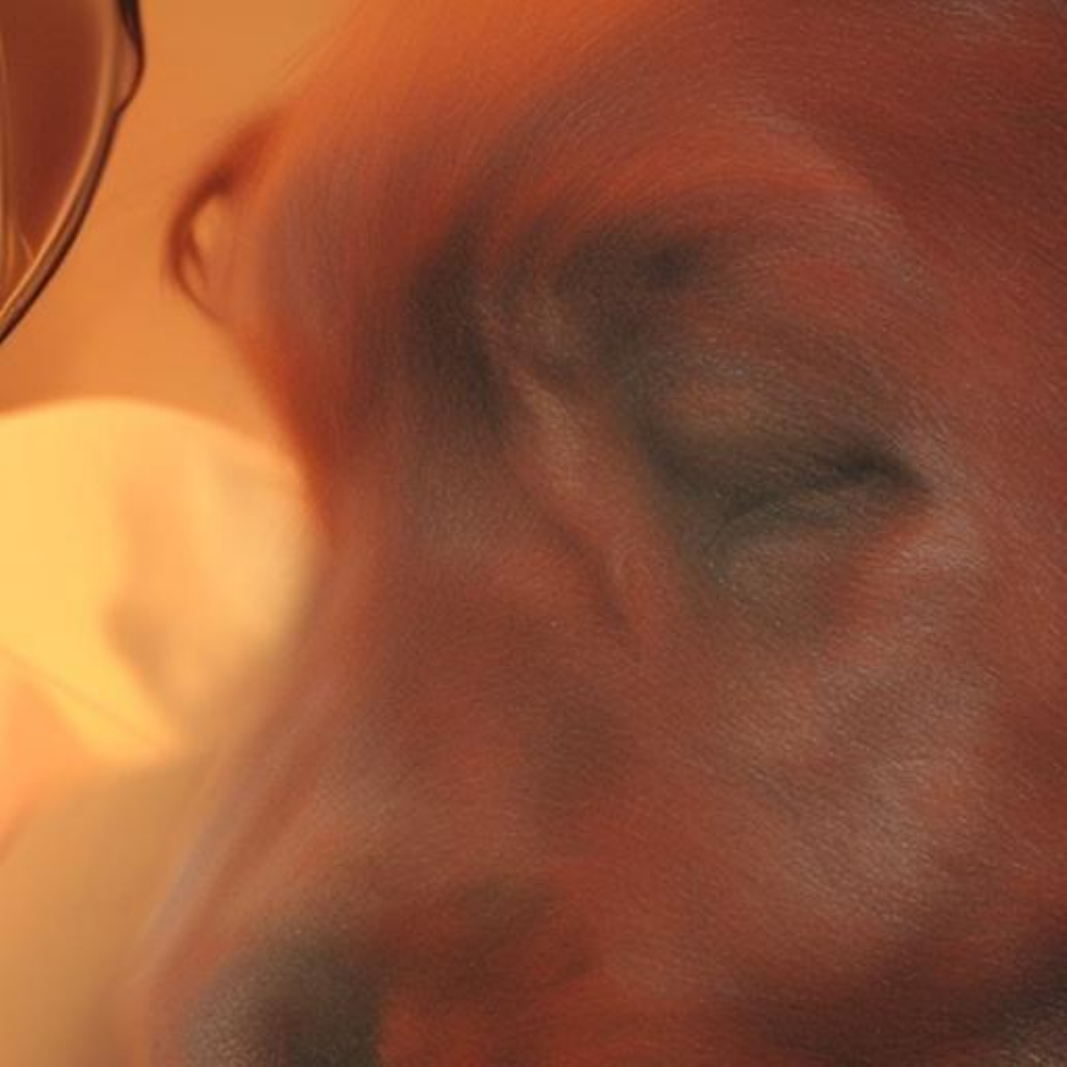}
\includegraphics[width=0.15\columnwidth]{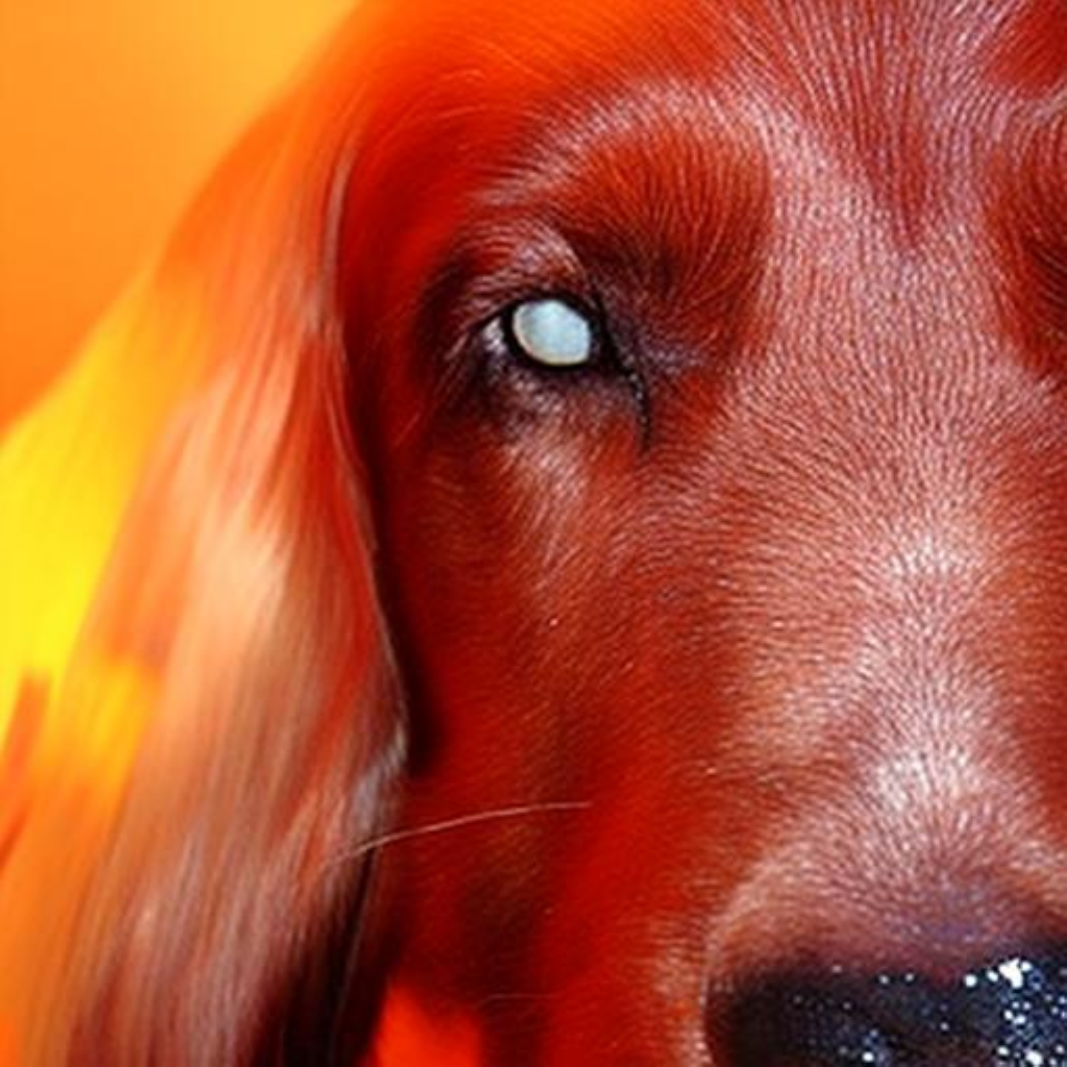}
\includegraphics[width=0.15\columnwidth]{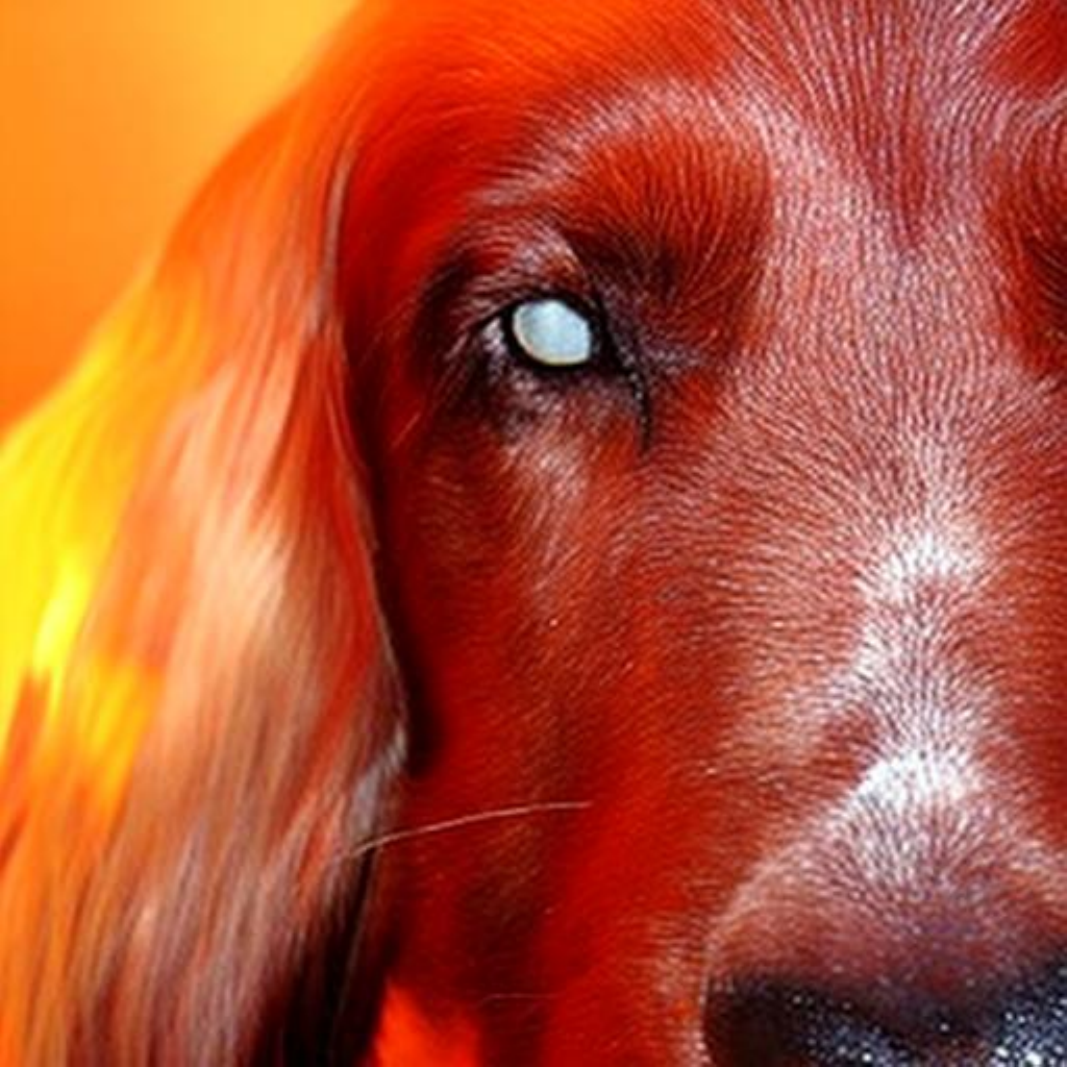}
\\
\includegraphics[width=0.15\columnwidth]{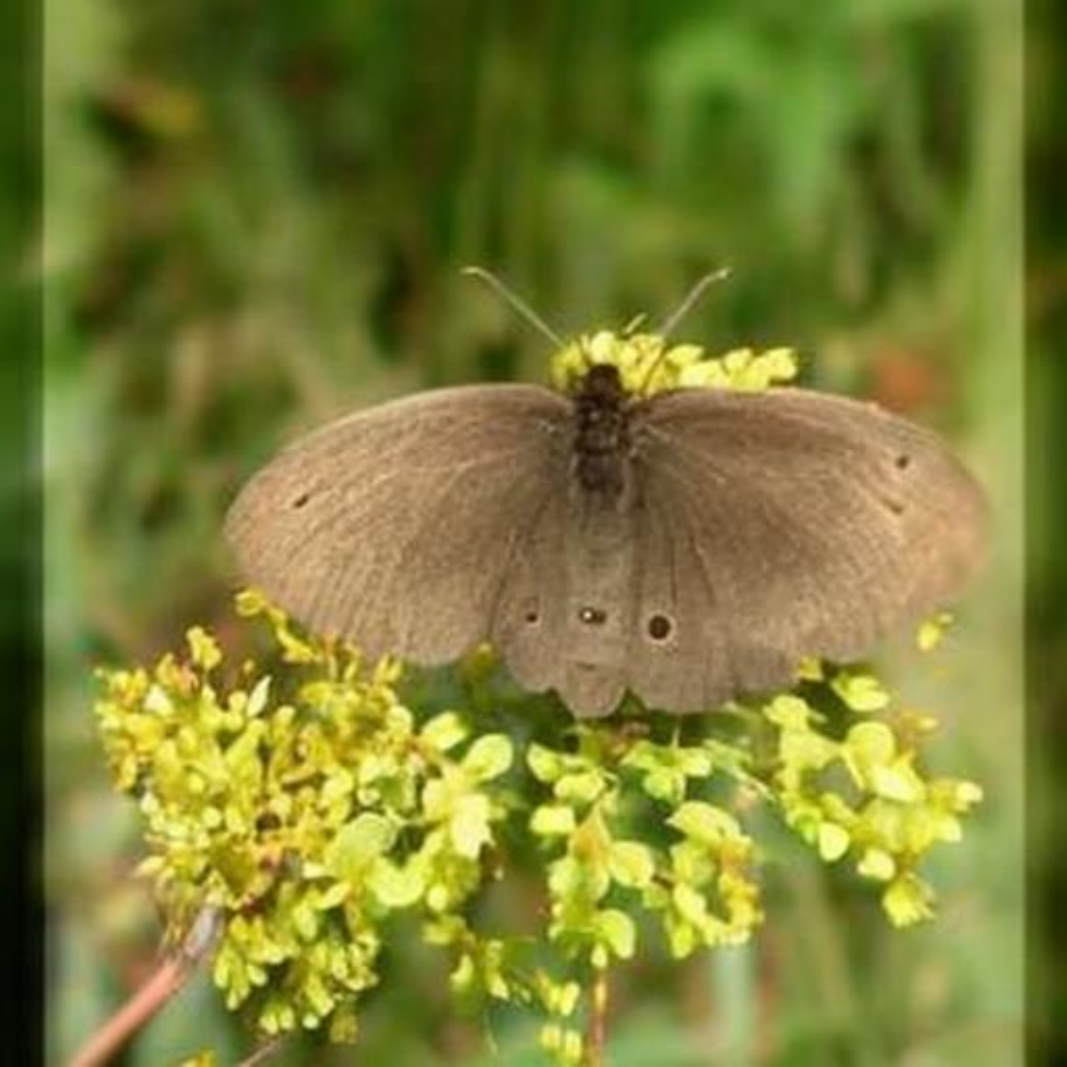}
\includegraphics[width=0.15\columnwidth]{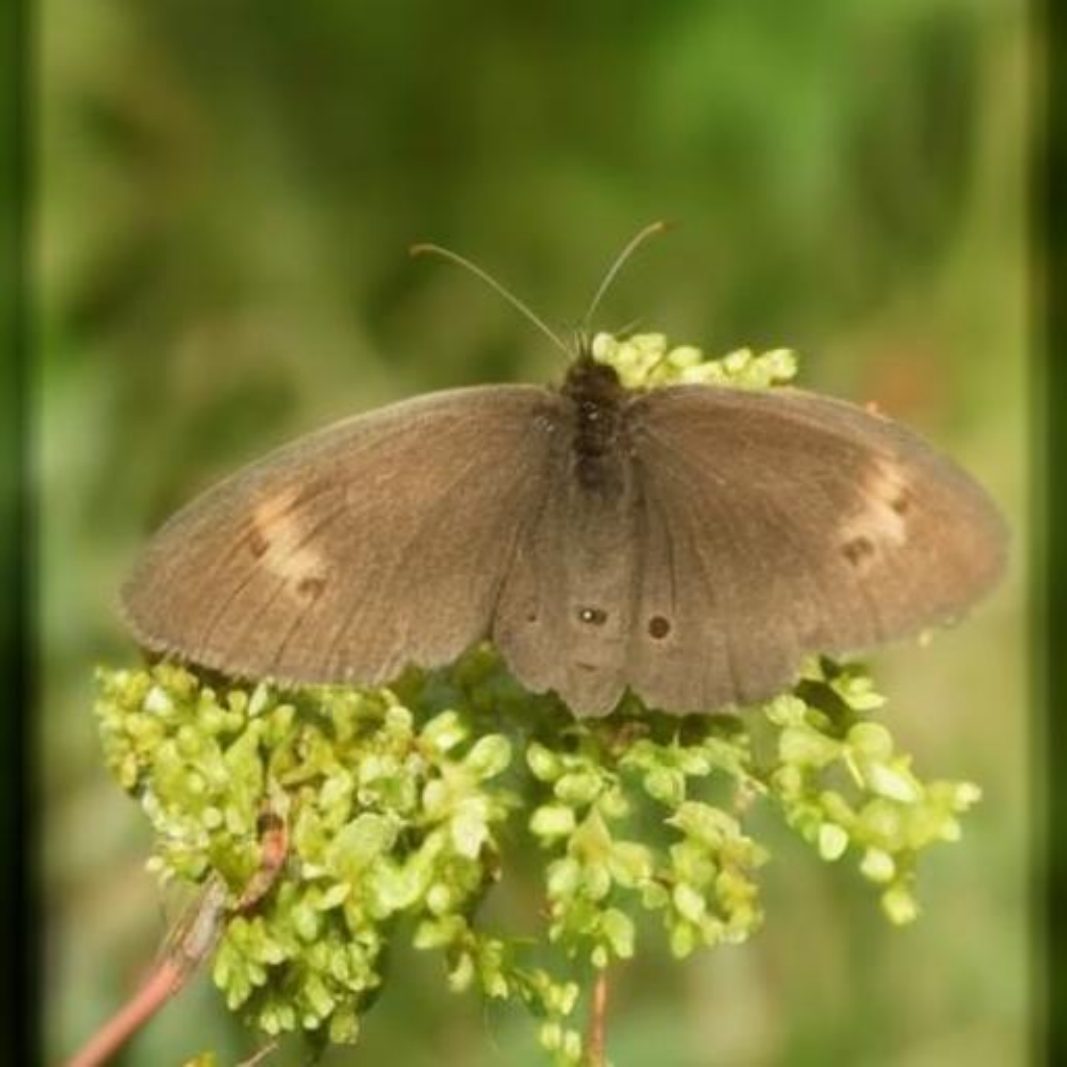}
\includegraphics[width=0.15\columnwidth]{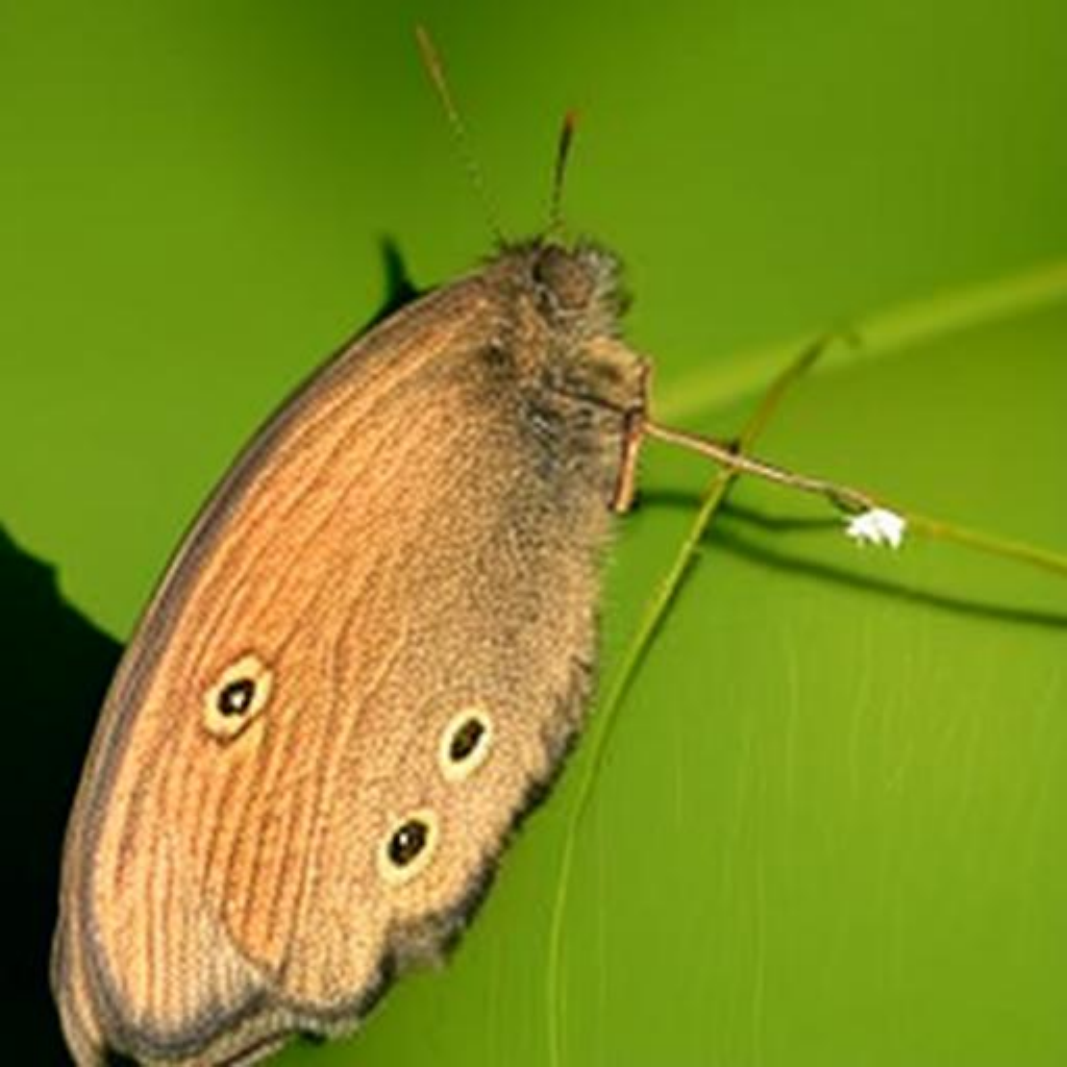}
\includegraphics[width=0.15\columnwidth]{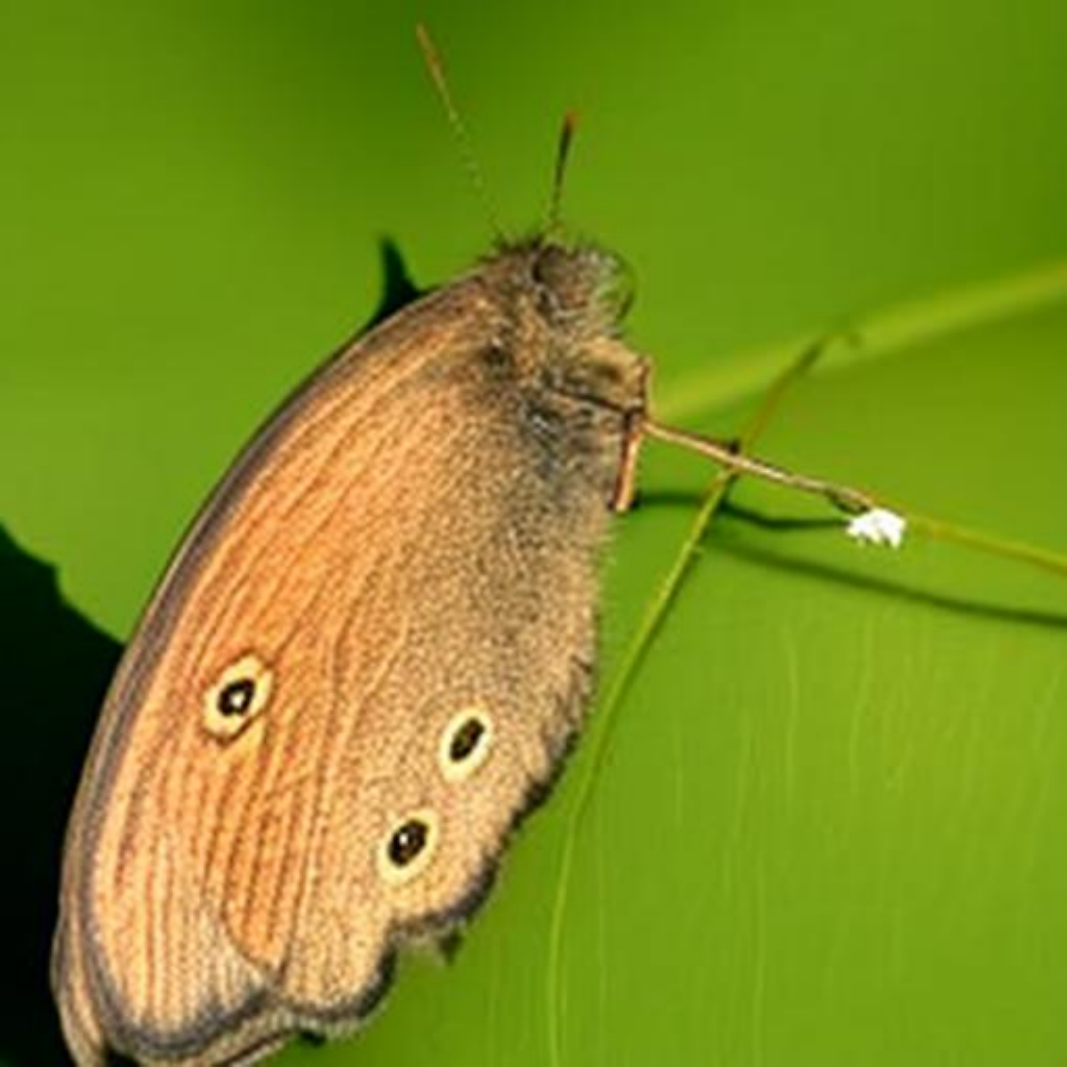}
\\
\includegraphics[width=0.15\columnwidth]{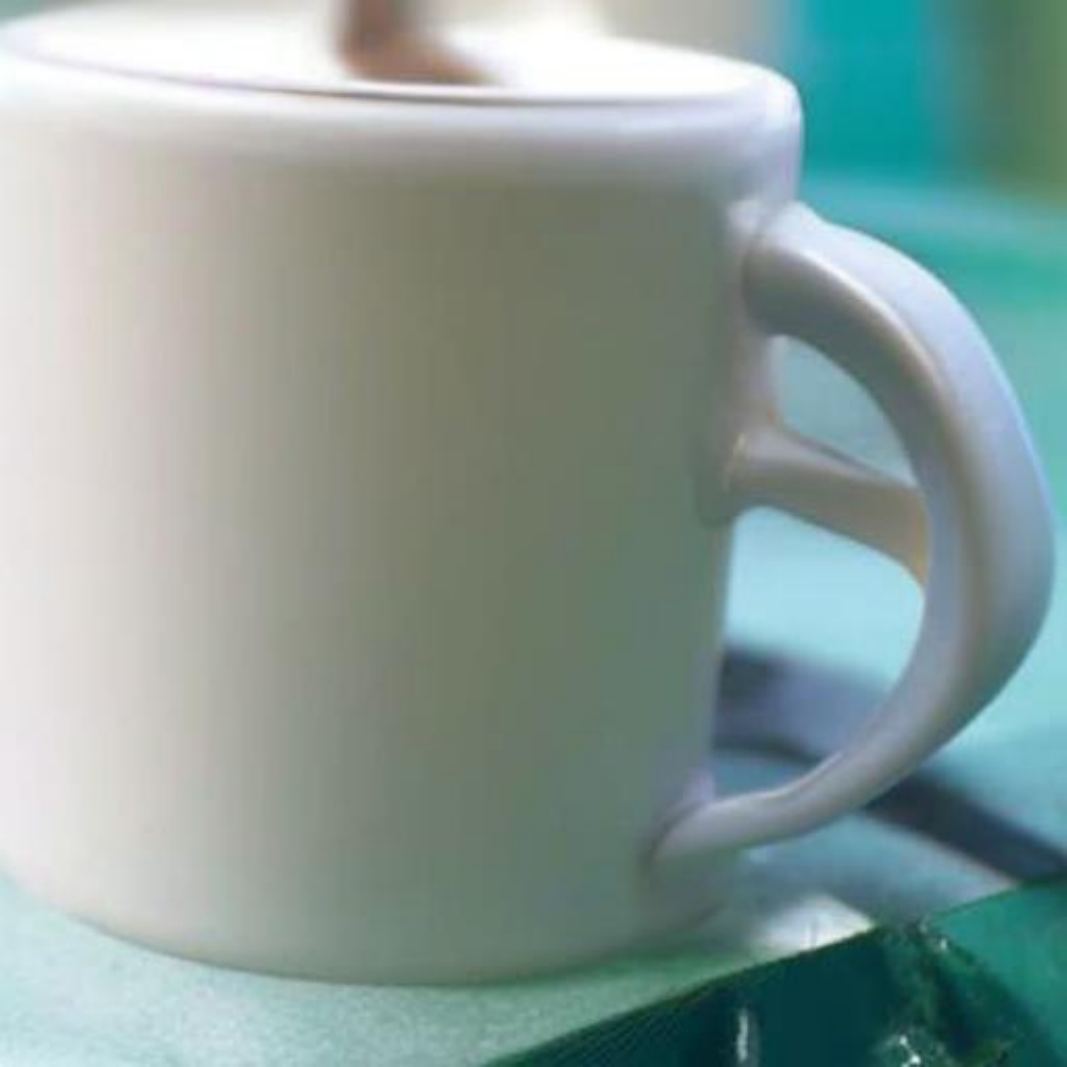}
\includegraphics[width=0.15\columnwidth]{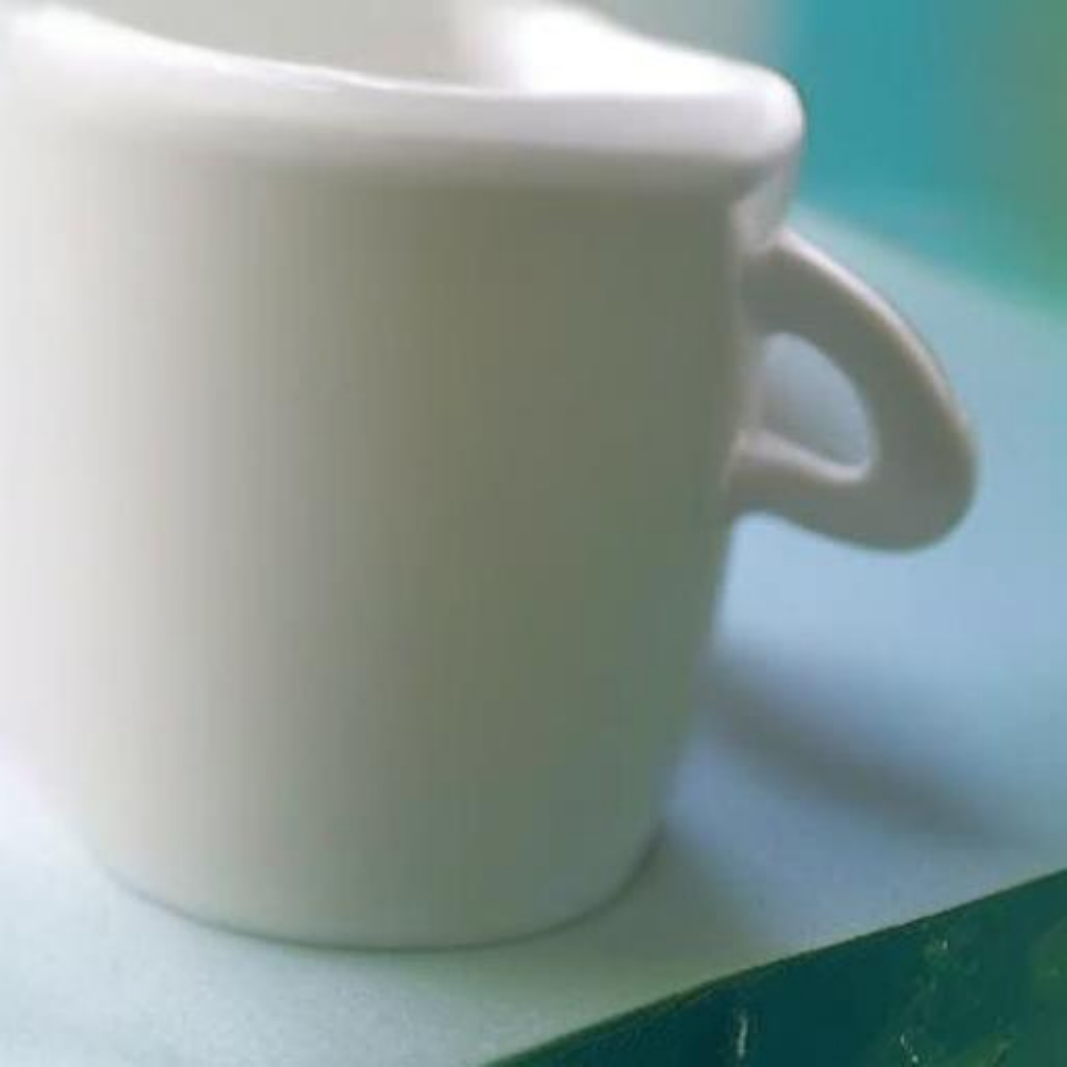}
\includegraphics[width=0.15\columnwidth]{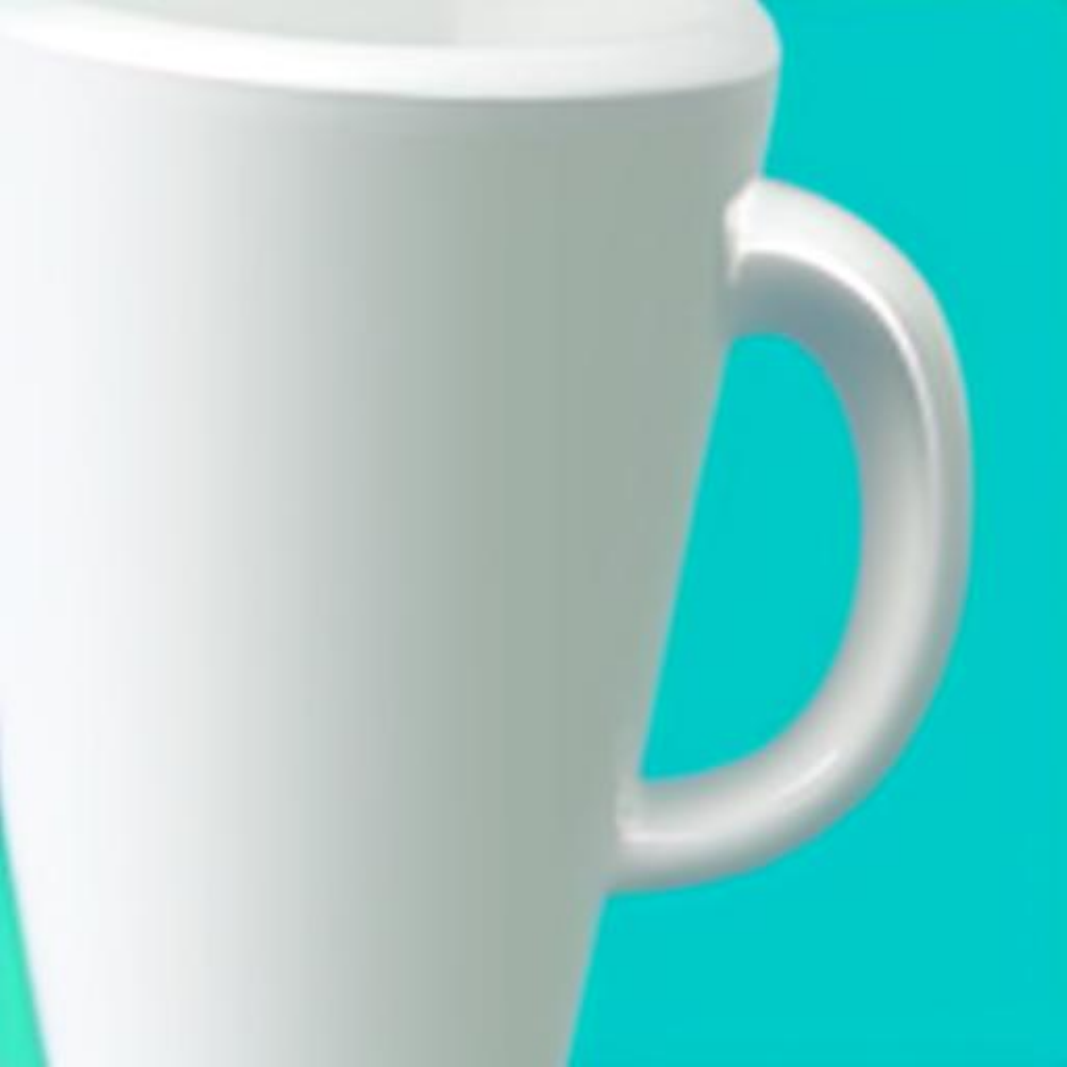}
\includegraphics[width=0.15\columnwidth]{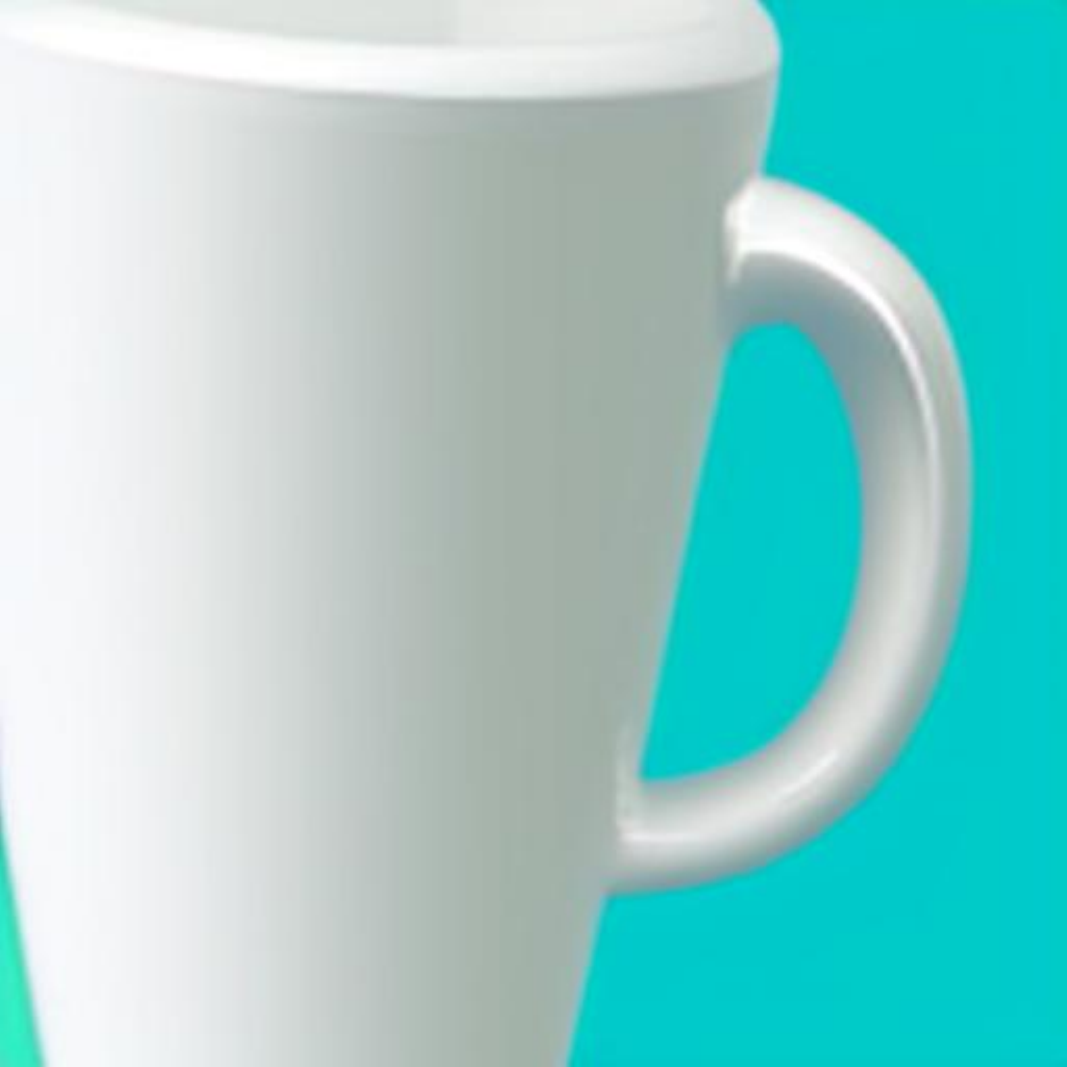}
\\
\includegraphics[width=0.15\columnwidth]{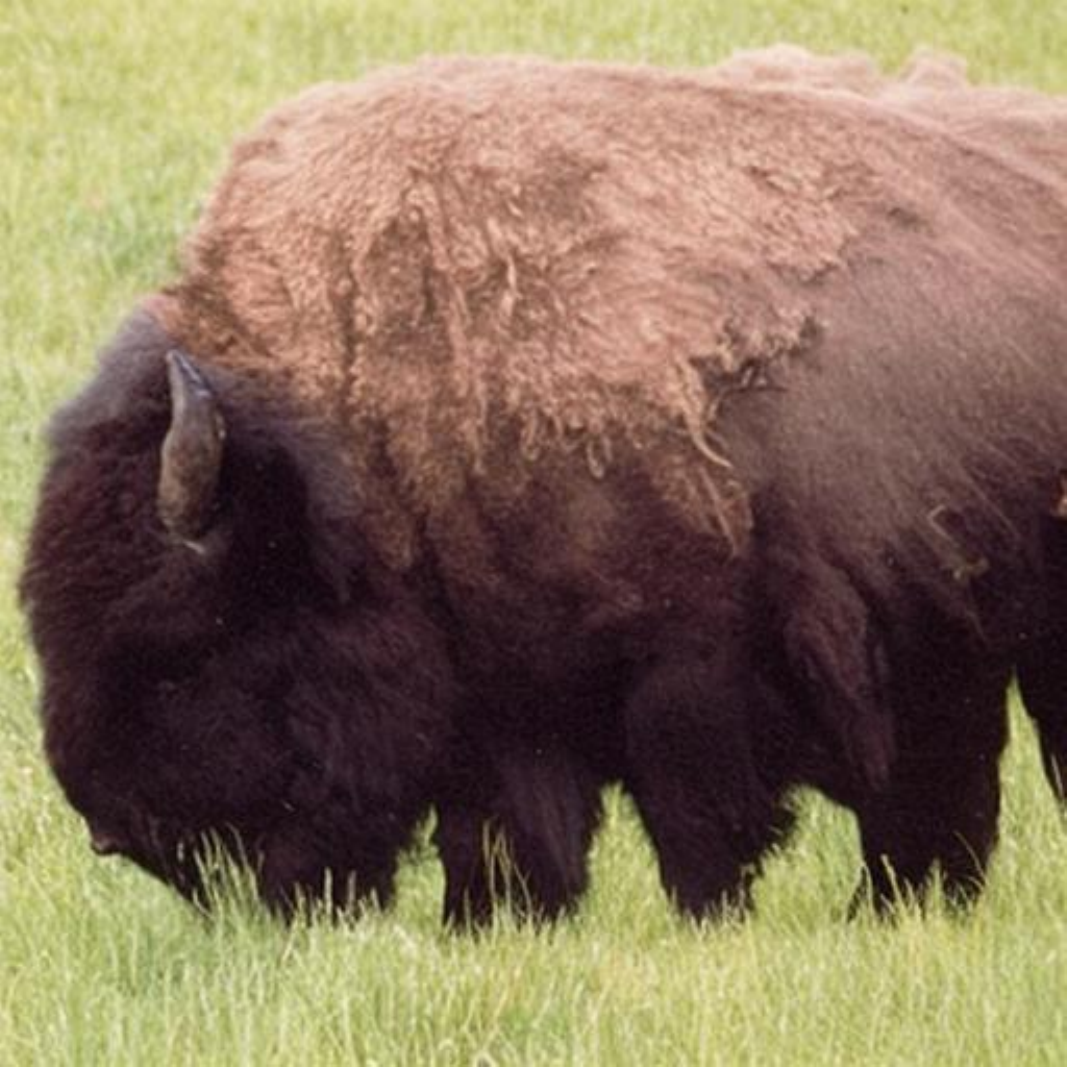}
\includegraphics[width=0.15\columnwidth]{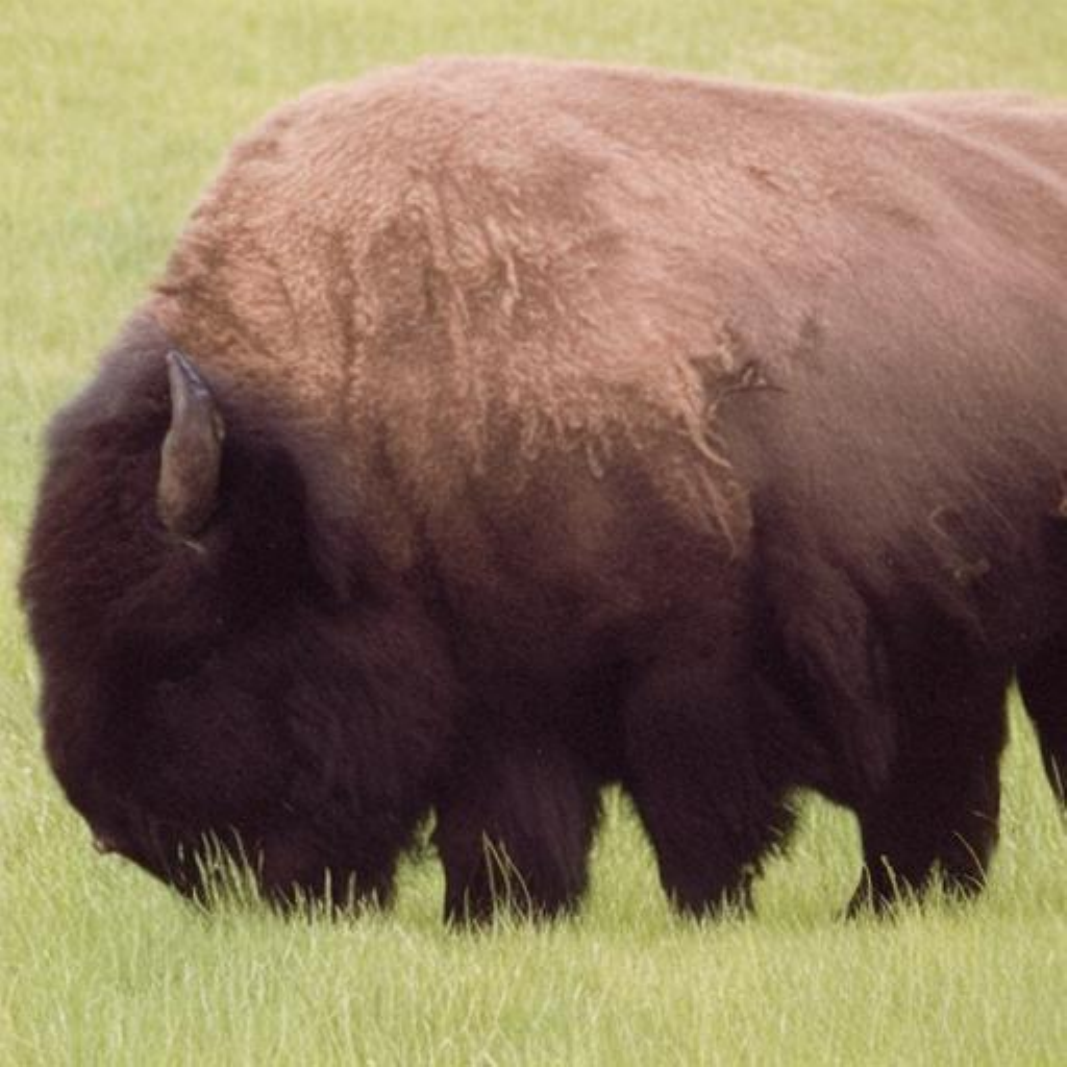}
\includegraphics[width=0.15\columnwidth]{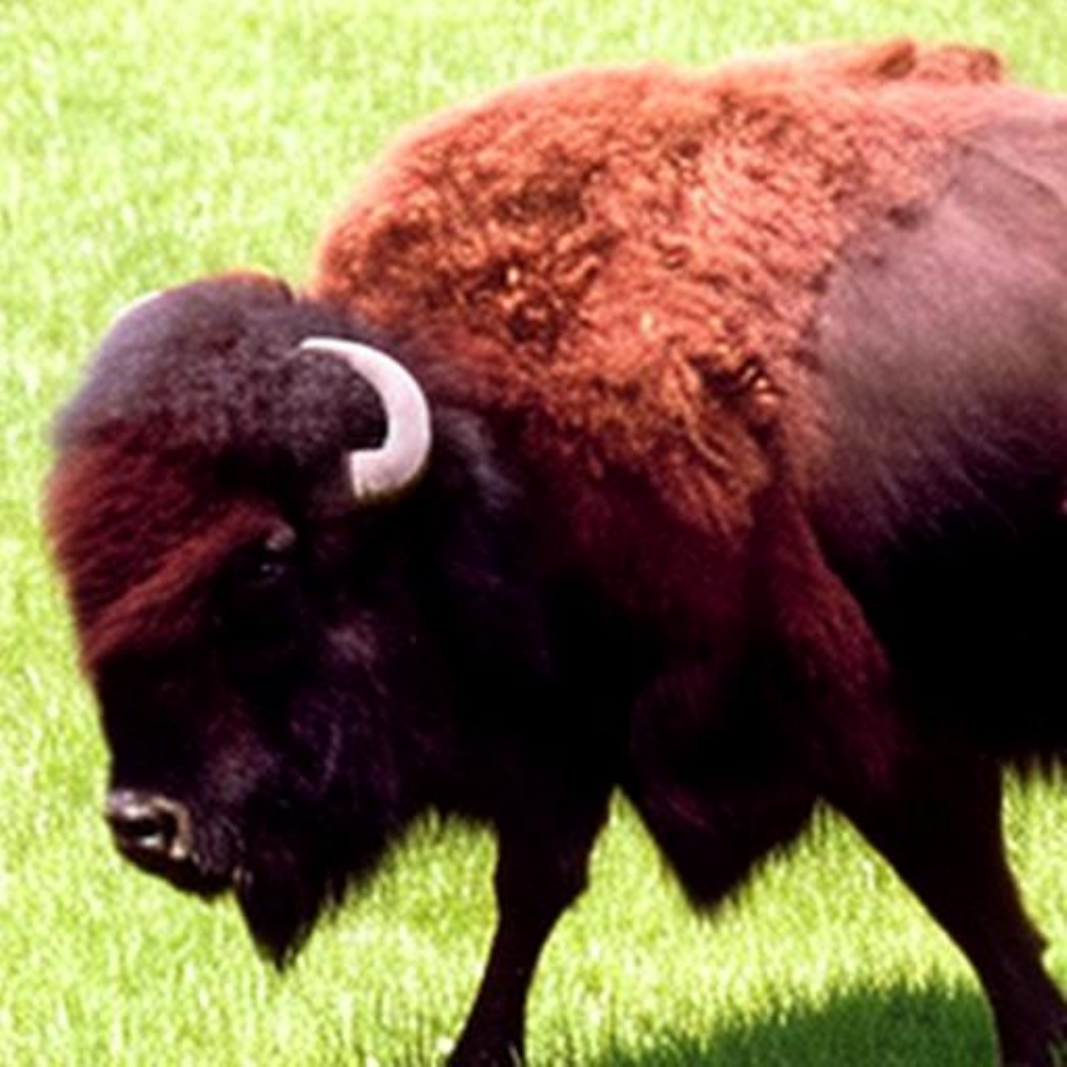}
\includegraphics[width=0.15\columnwidth]{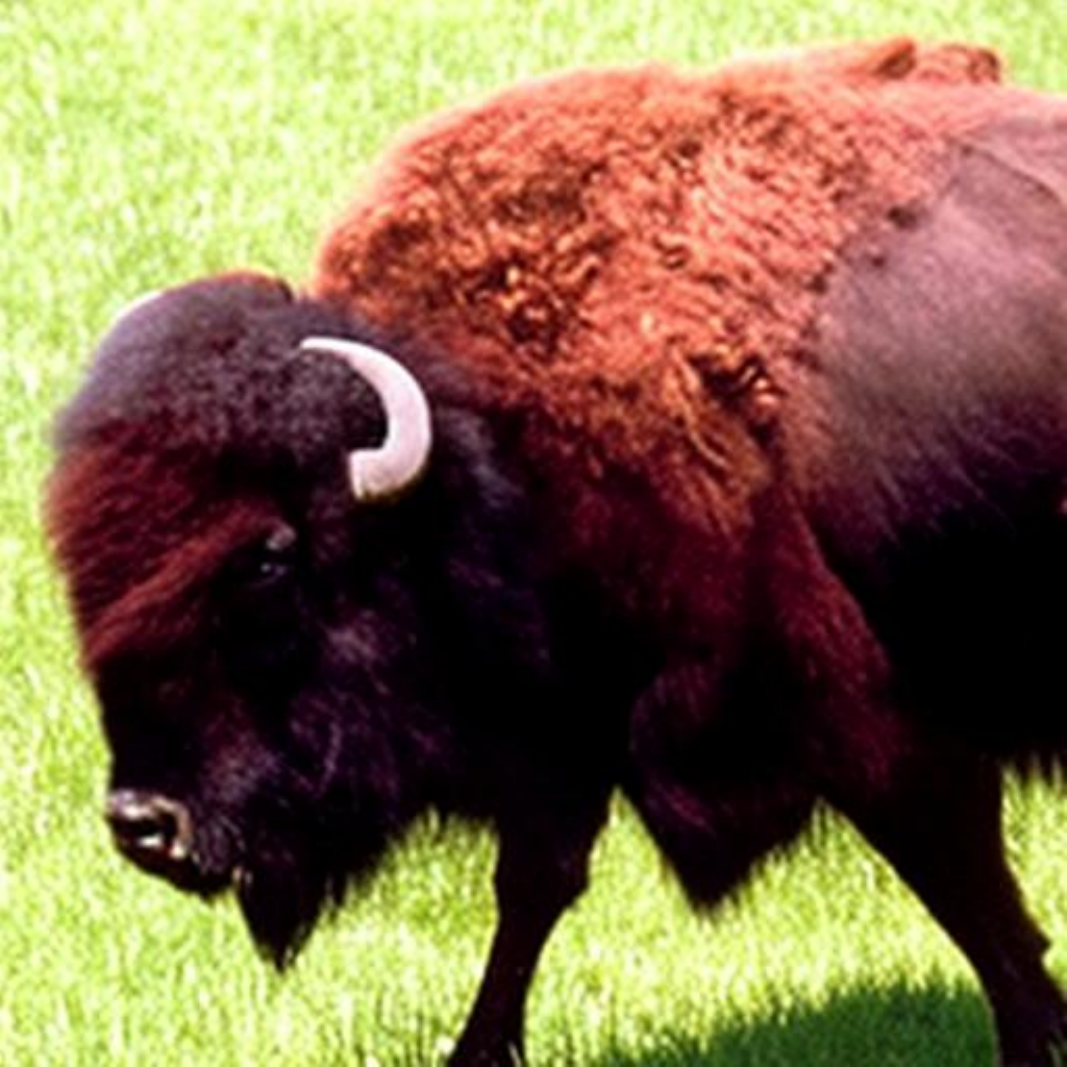}
\\
\includegraphics[width=0.15\columnwidth]{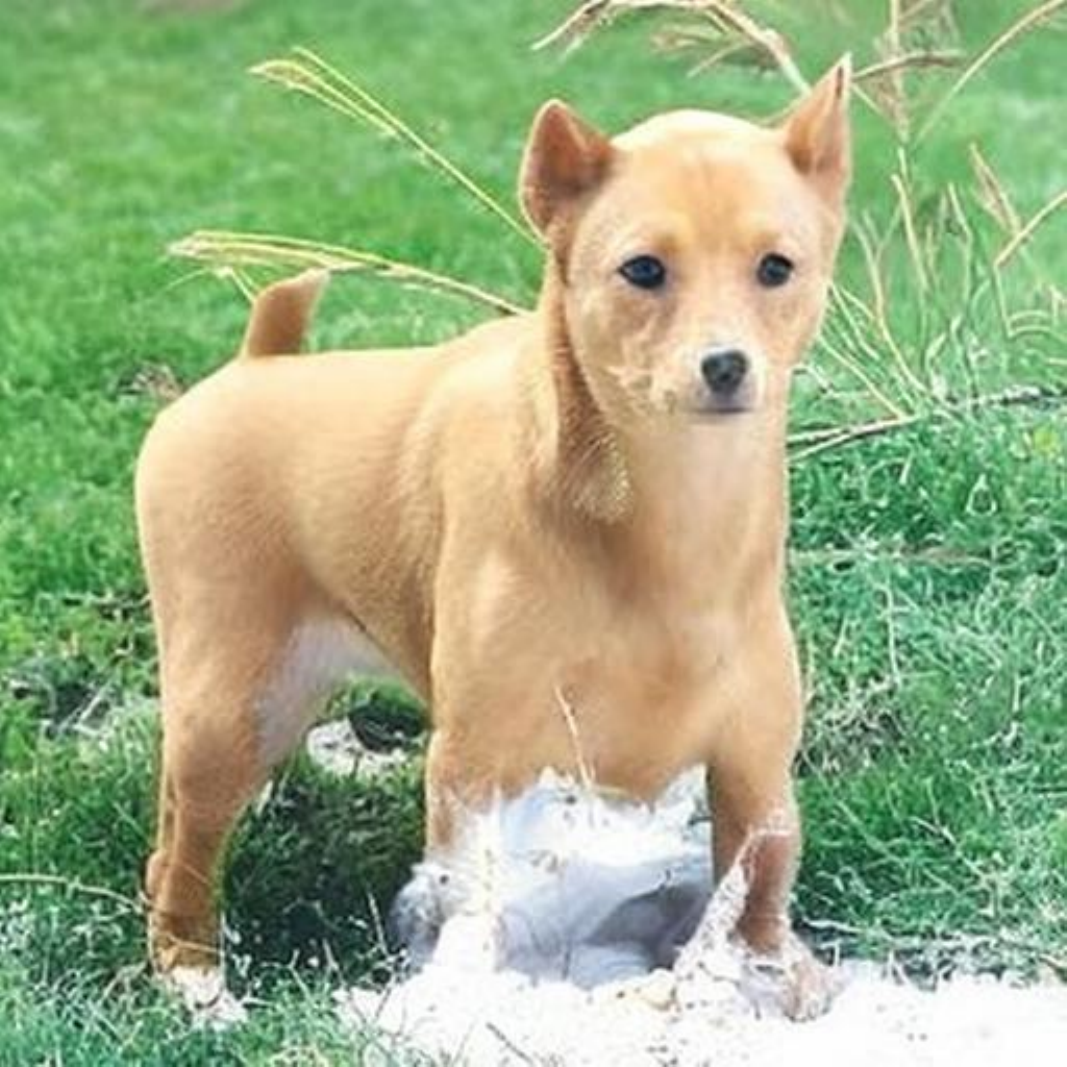}
\includegraphics[width=0.15\columnwidth]{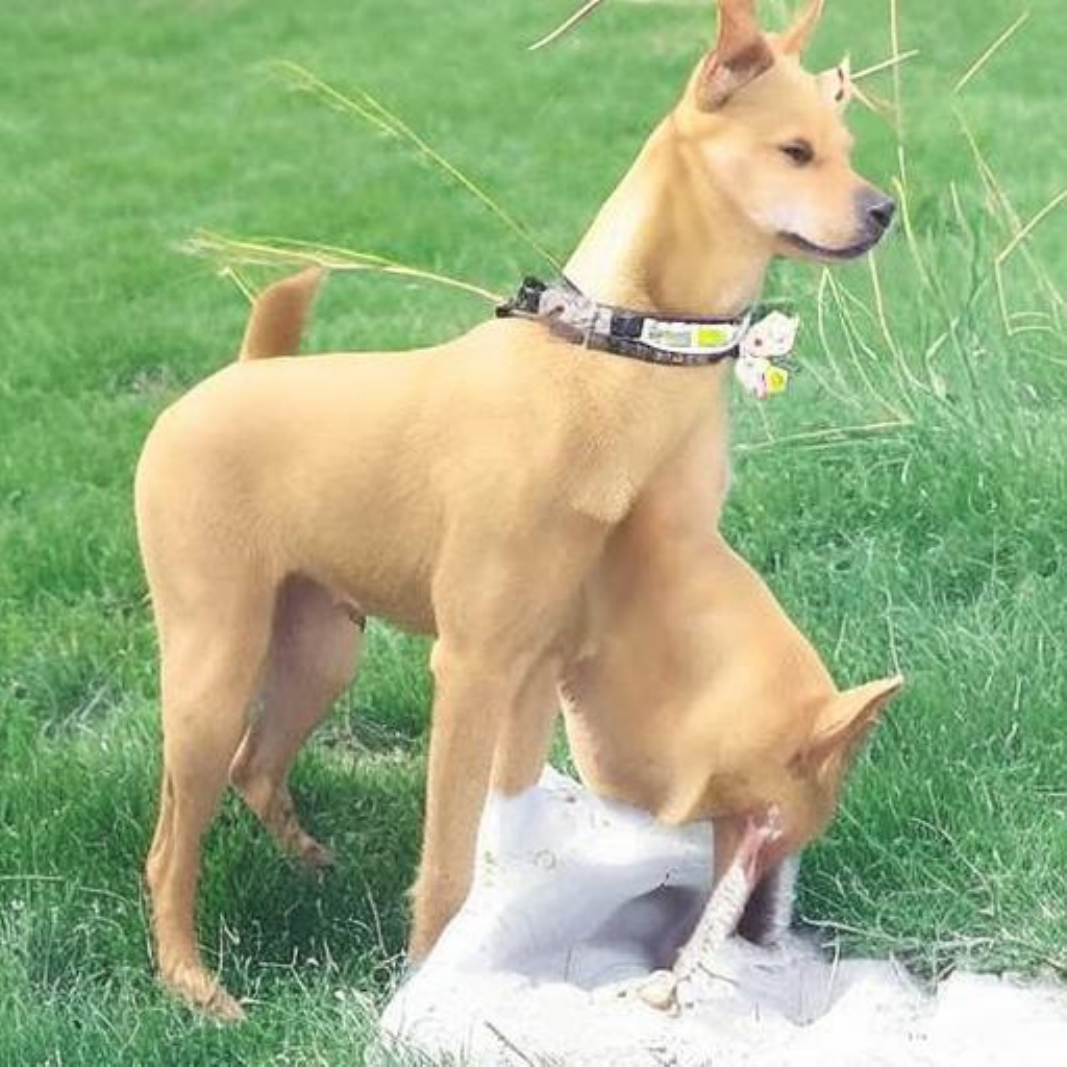}
\includegraphics[width=0.15\columnwidth]{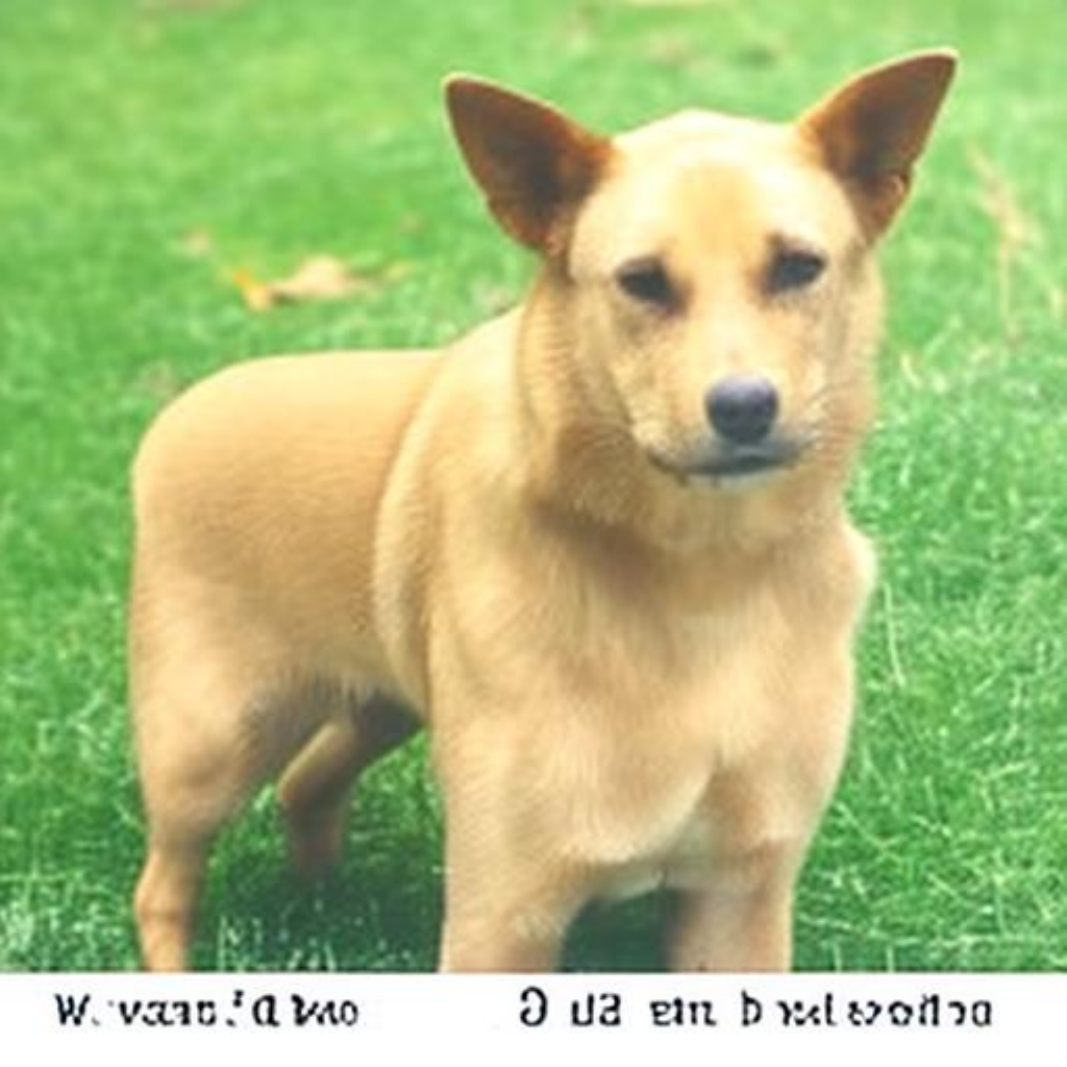}
\includegraphics[width=0.15\columnwidth]{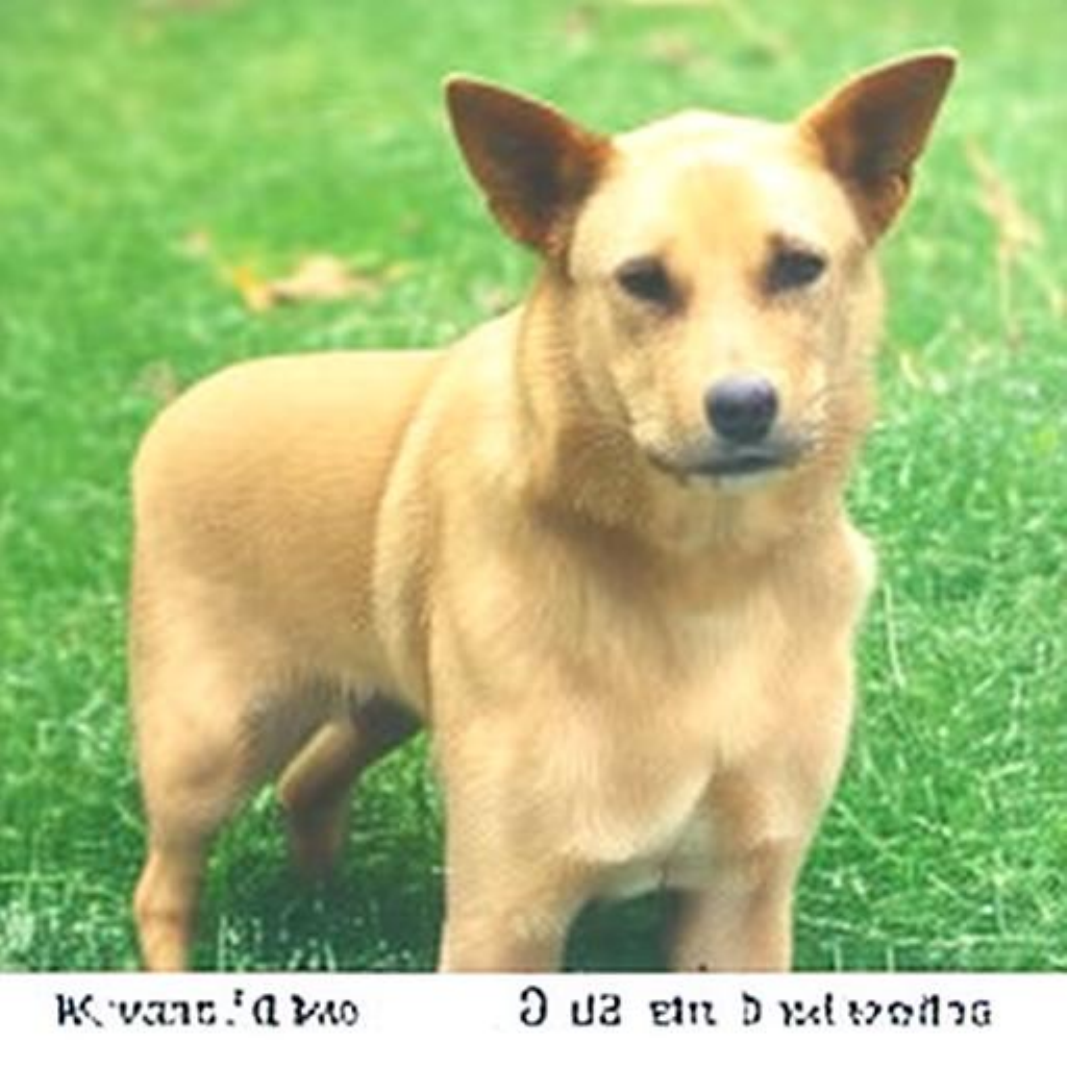}
\\
\includegraphics[width=0.15\columnwidth]{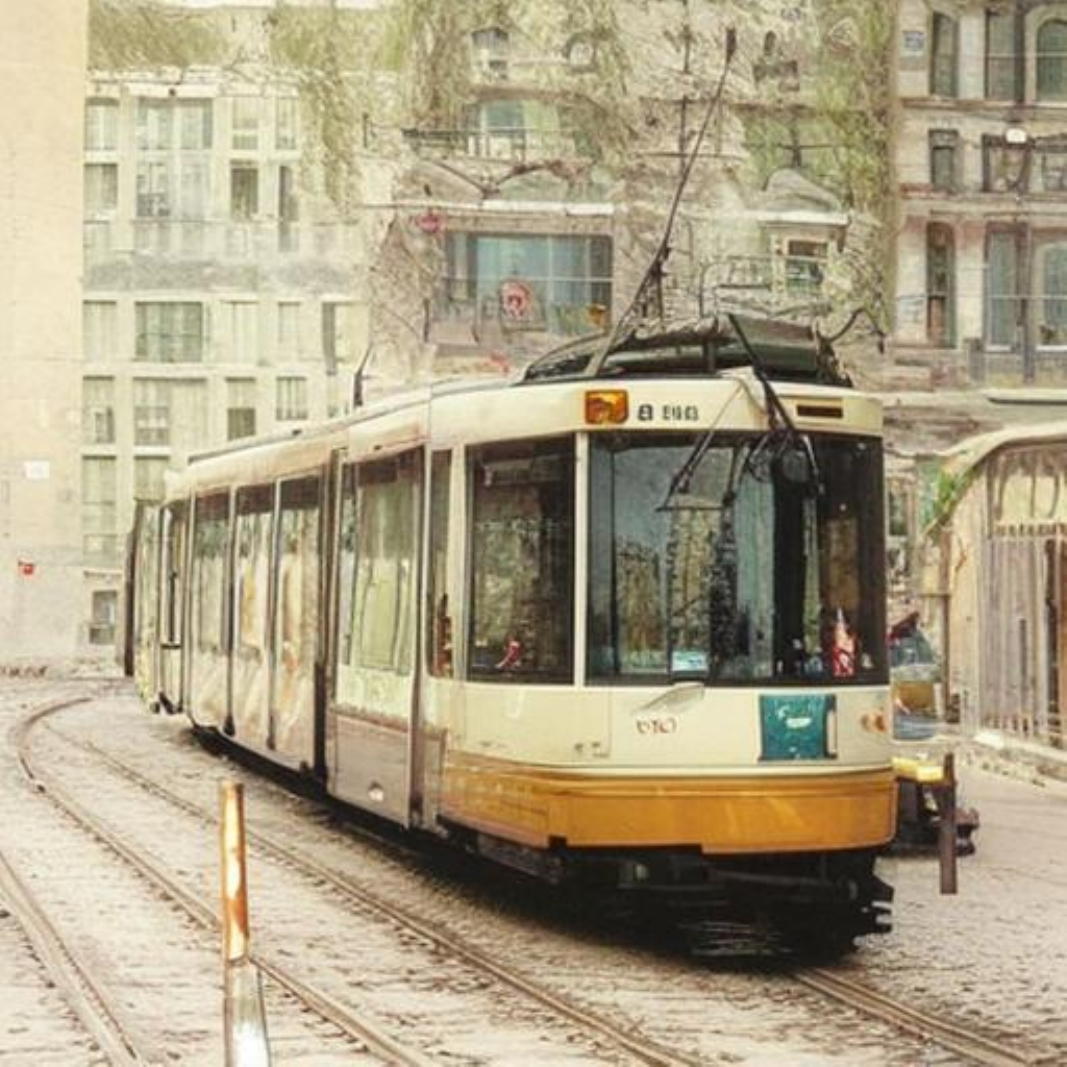}
\includegraphics[width=0.15\columnwidth]{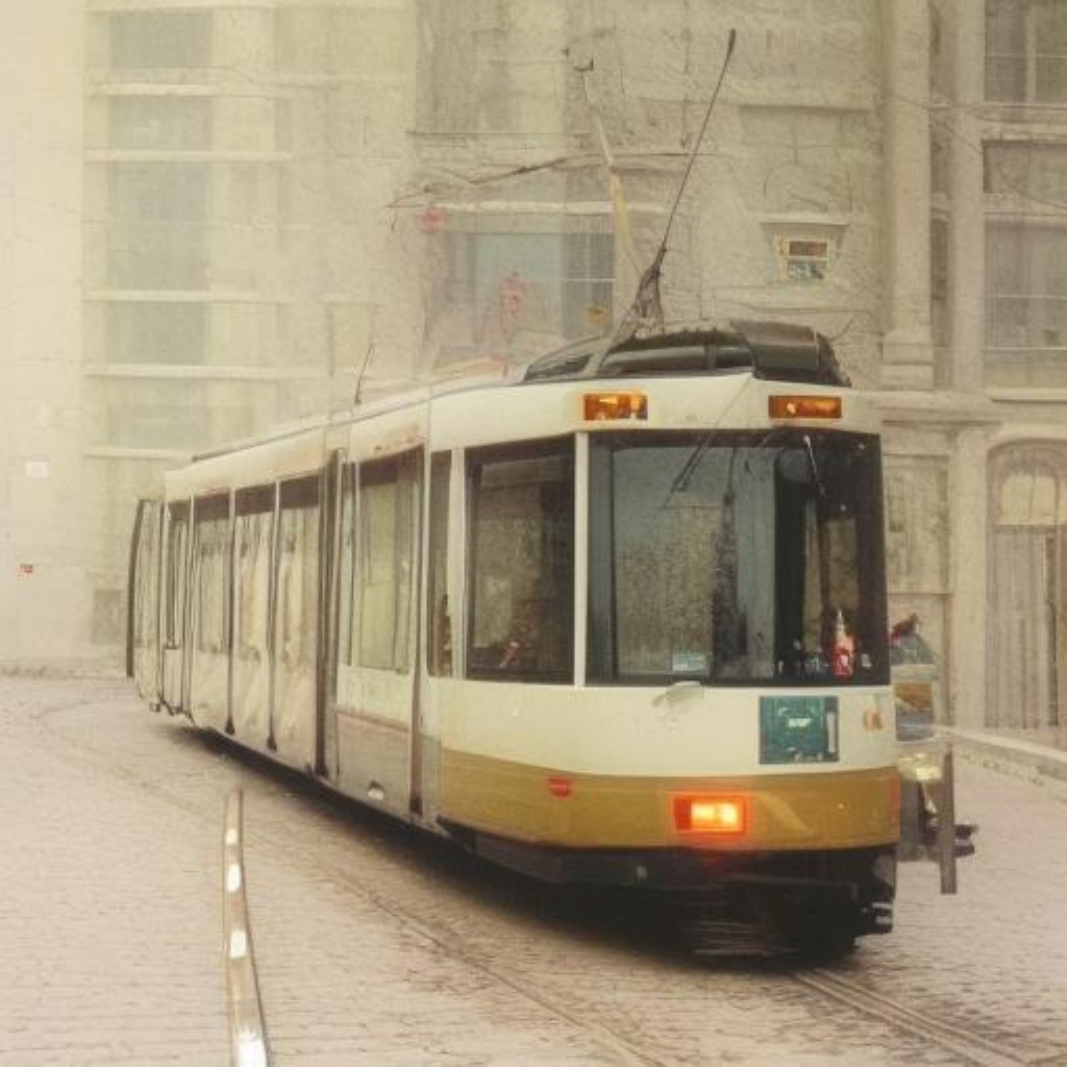}
\includegraphics[width=0.15\columnwidth]{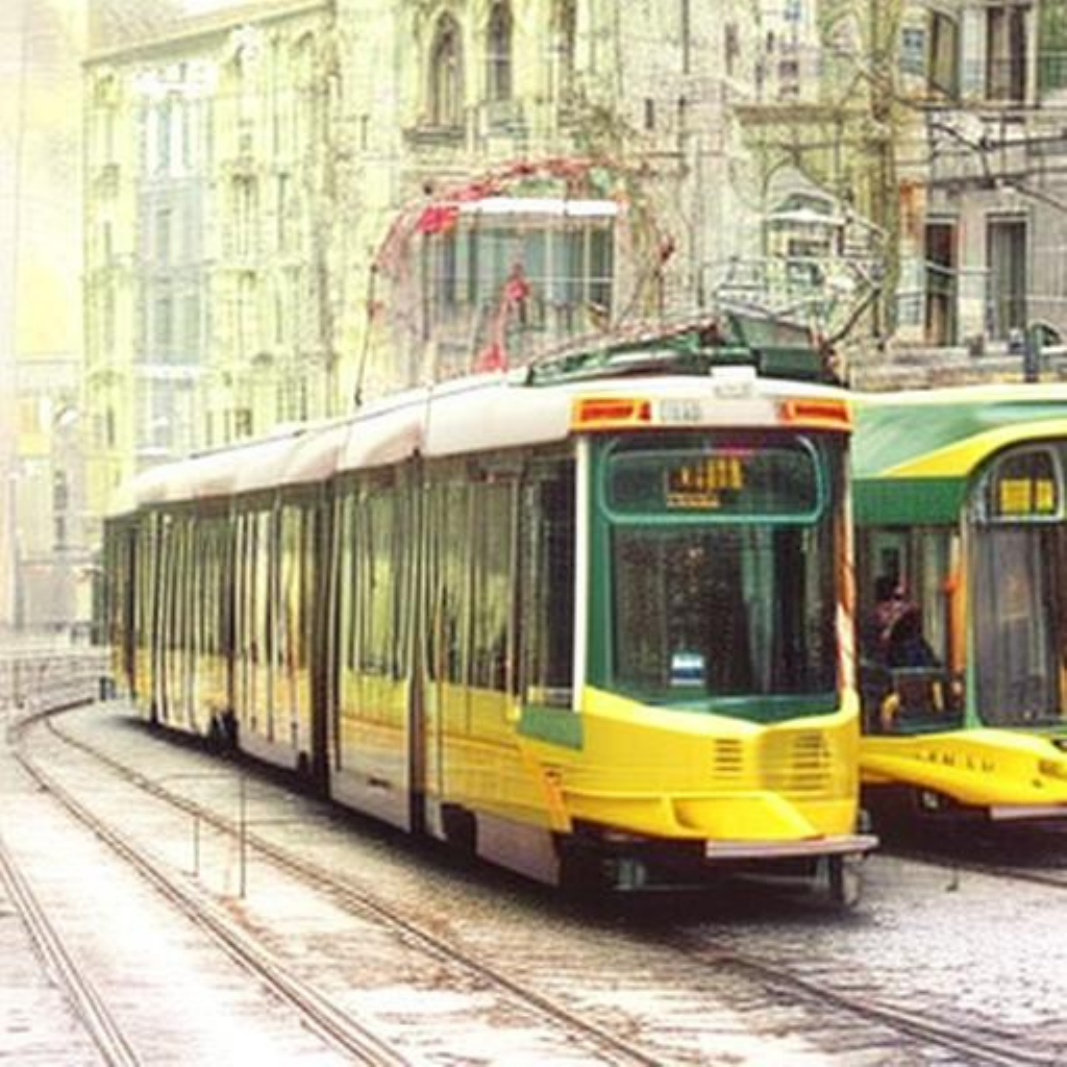}
\includegraphics[width=0.15\columnwidth]{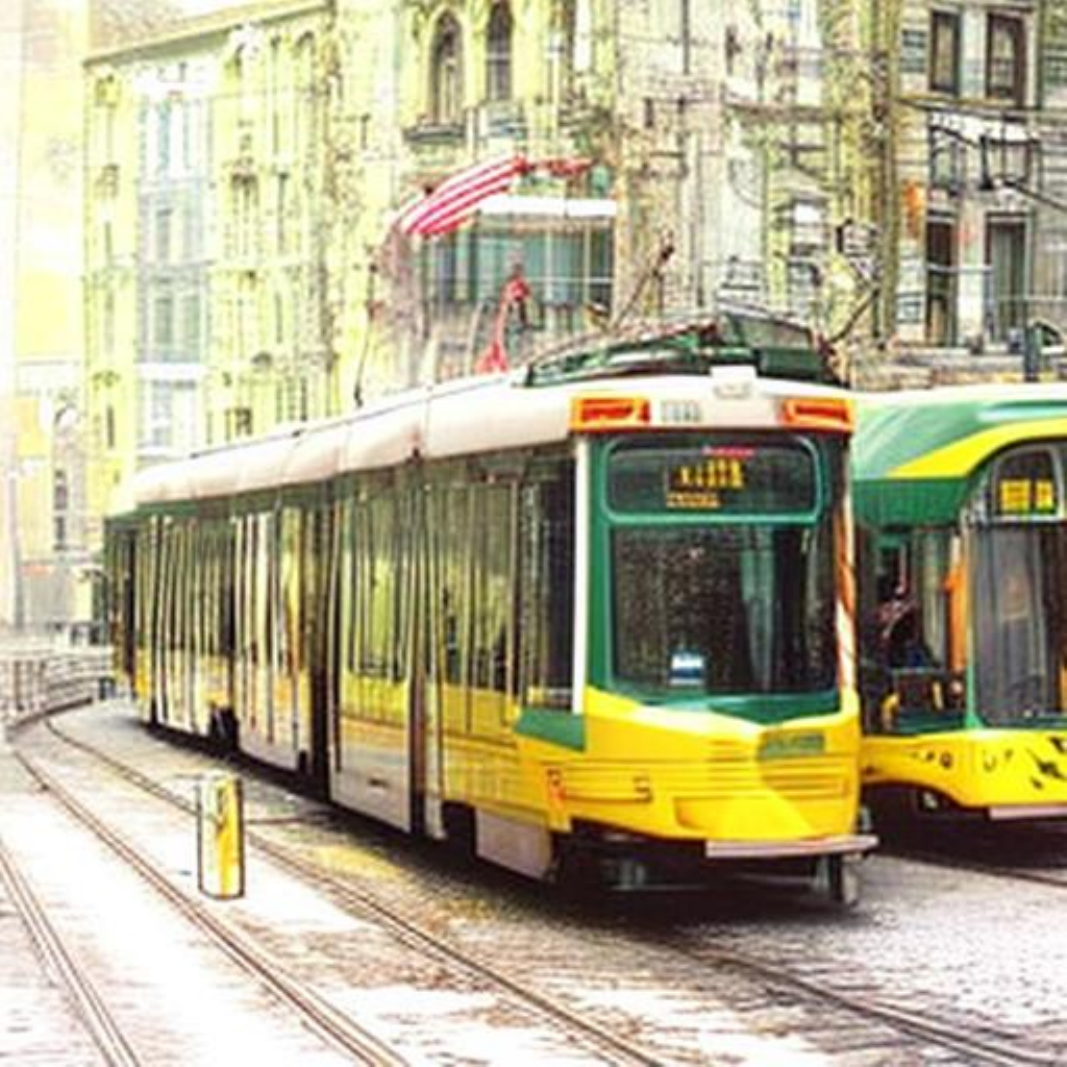}
\caption{\textbf{DiT with $80$ Steps.} For each row, from left to right: baseline, ablation, generated image with variance scaling $1.15$, and one with $1.20$. 
The classes for each row are: \emph{lion}, \emph{bookshop}, \emph{red setter}, \emph{ringlet butterfly}, \emph{coffee mug}, \emph{bison}, \emph{kelpie}, and \emph{trolley}, respectively.
}
    \label{fig:dit-80}
    \end{figure}

\begin{figure}[t]
    \centering
    \includegraphics[width=0.15\columnwidth]{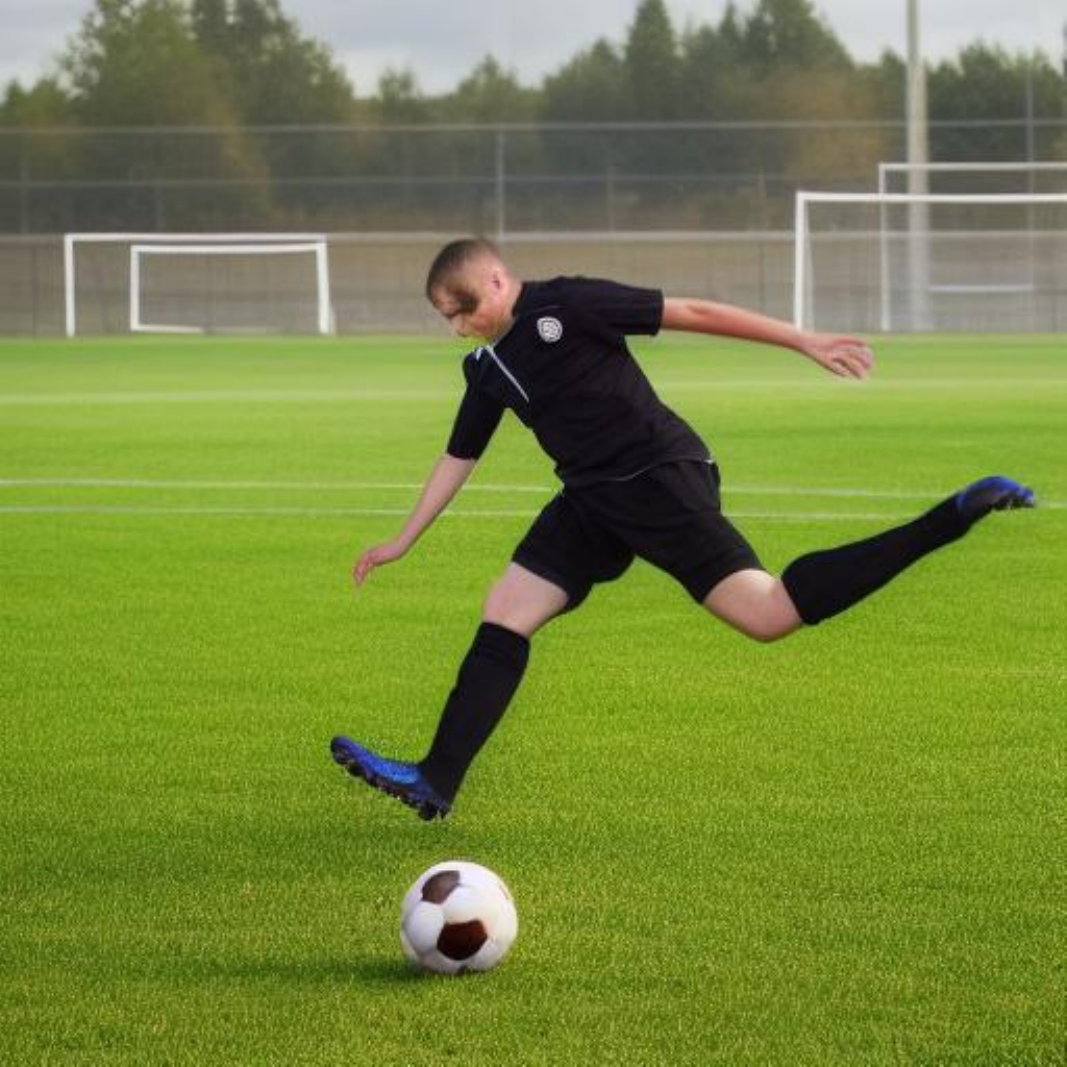}
\includegraphics[width=0.15\columnwidth]{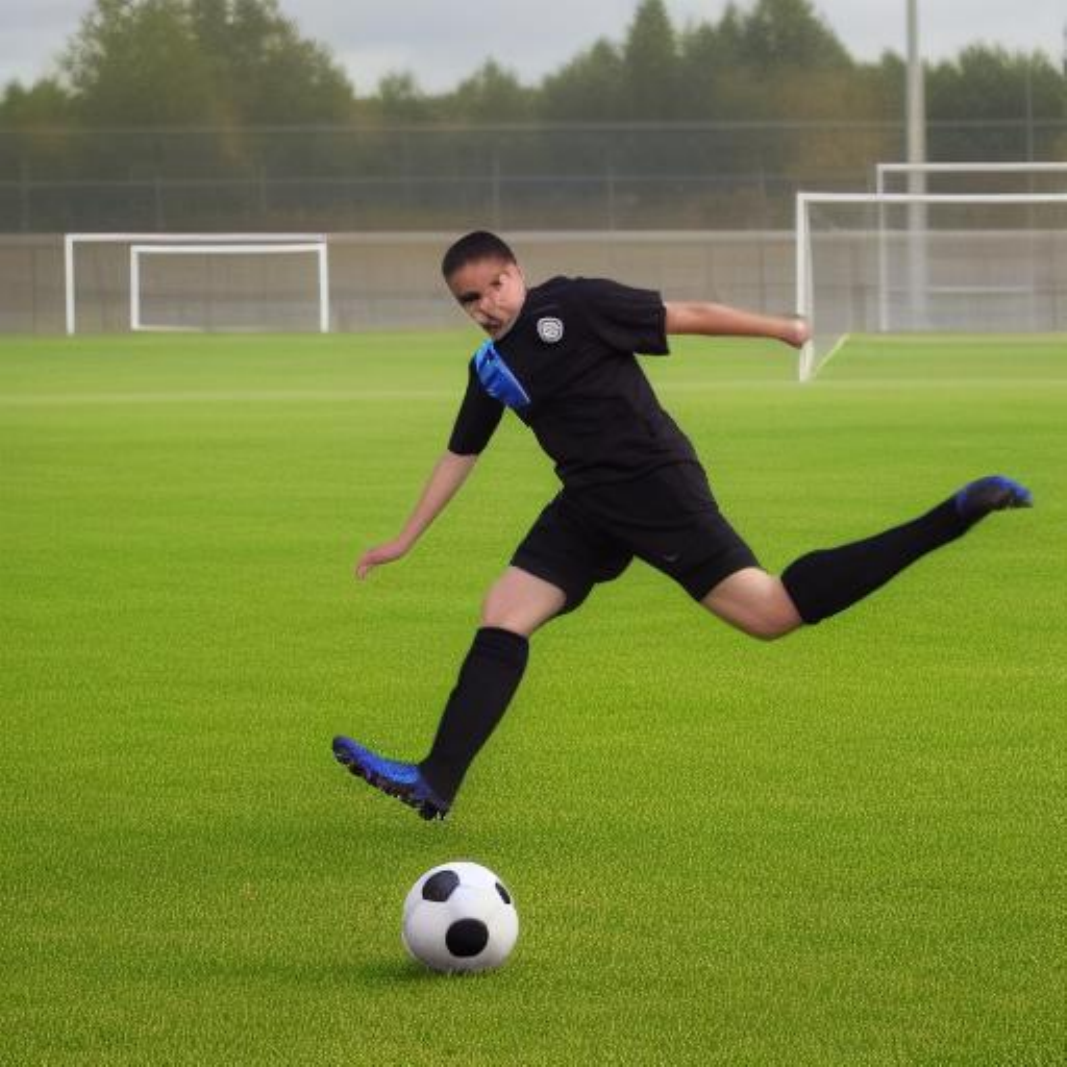}
\includegraphics[width=0.15\columnwidth]{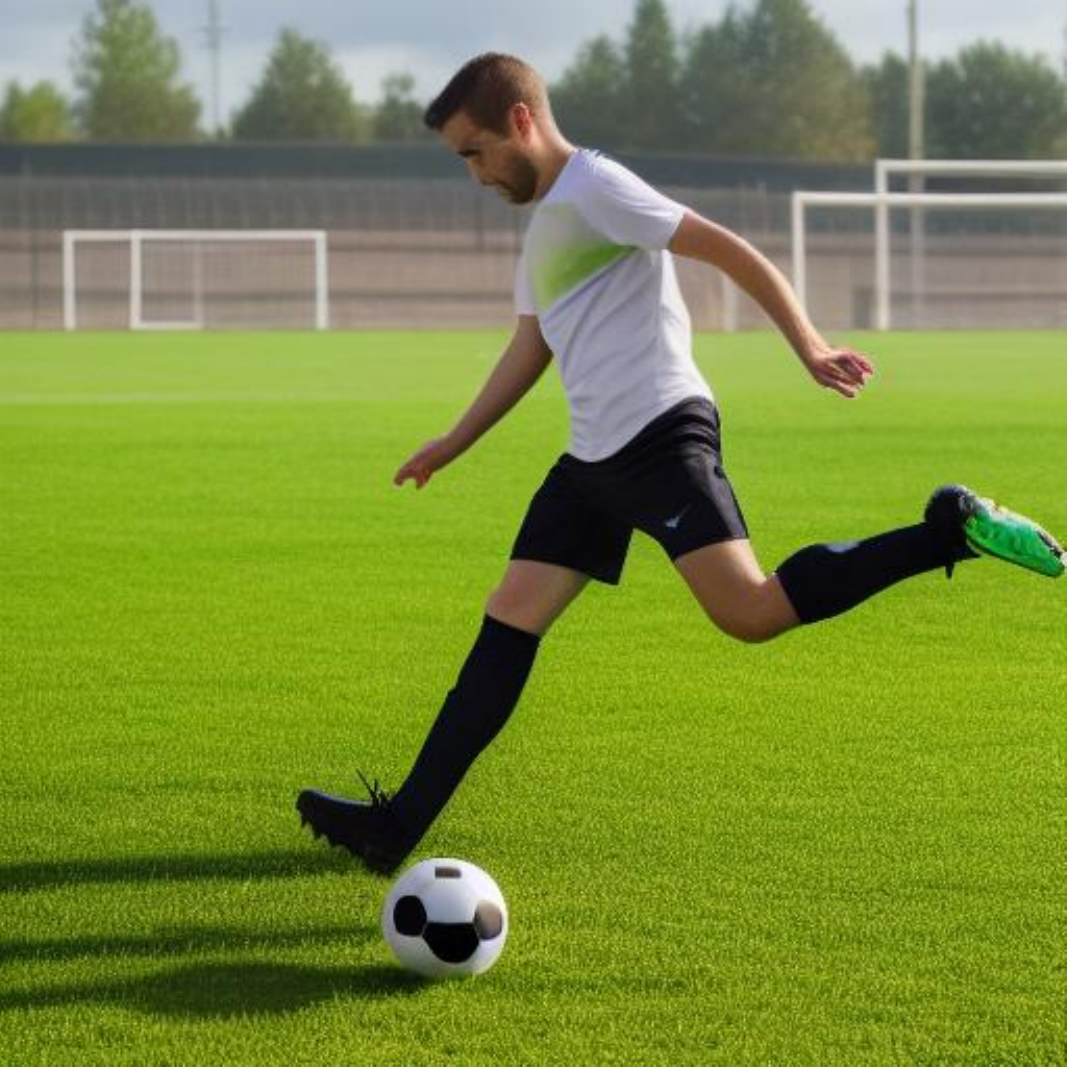}
\includegraphics[width=0.15\columnwidth]{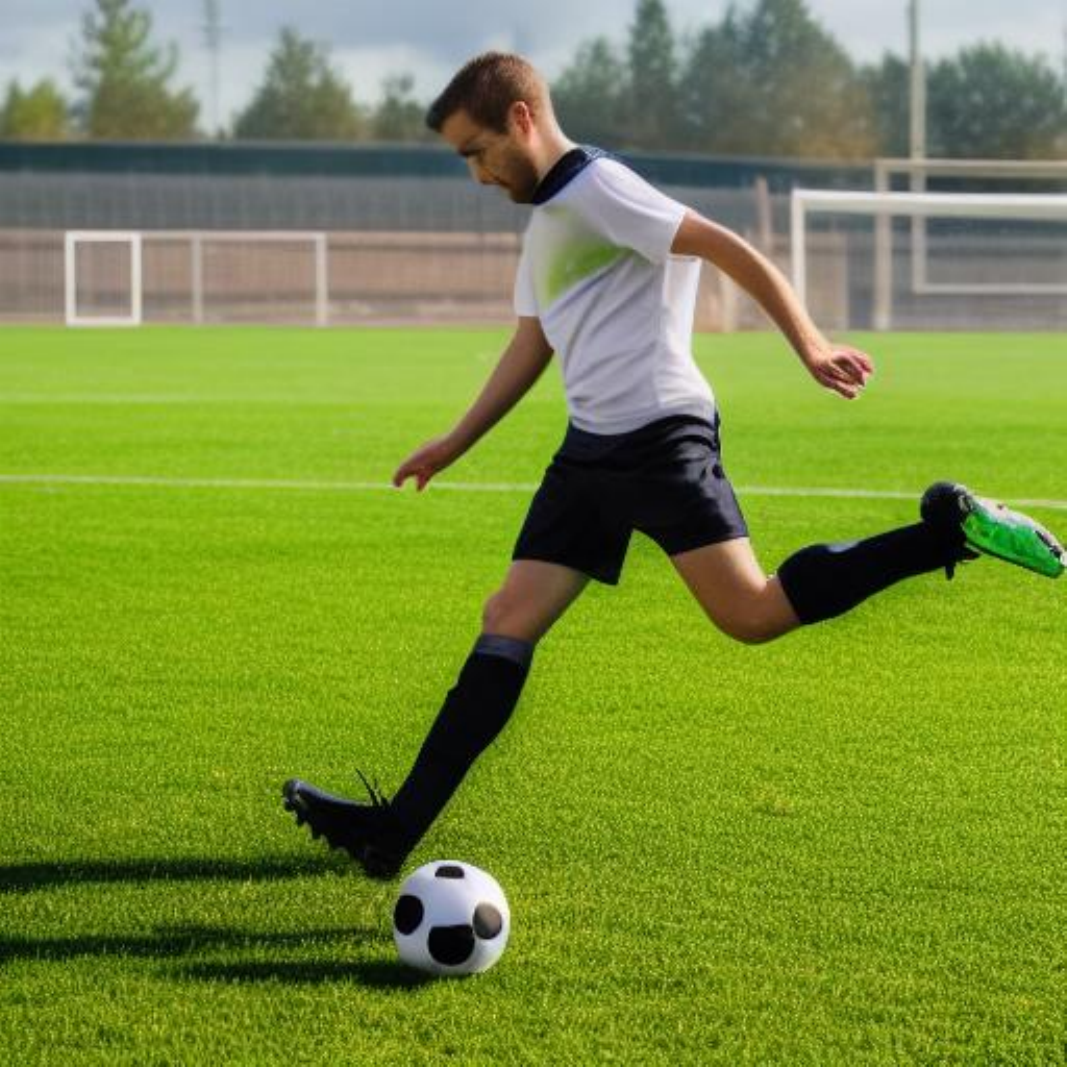}
\\
\includegraphics[width=0.15\columnwidth]{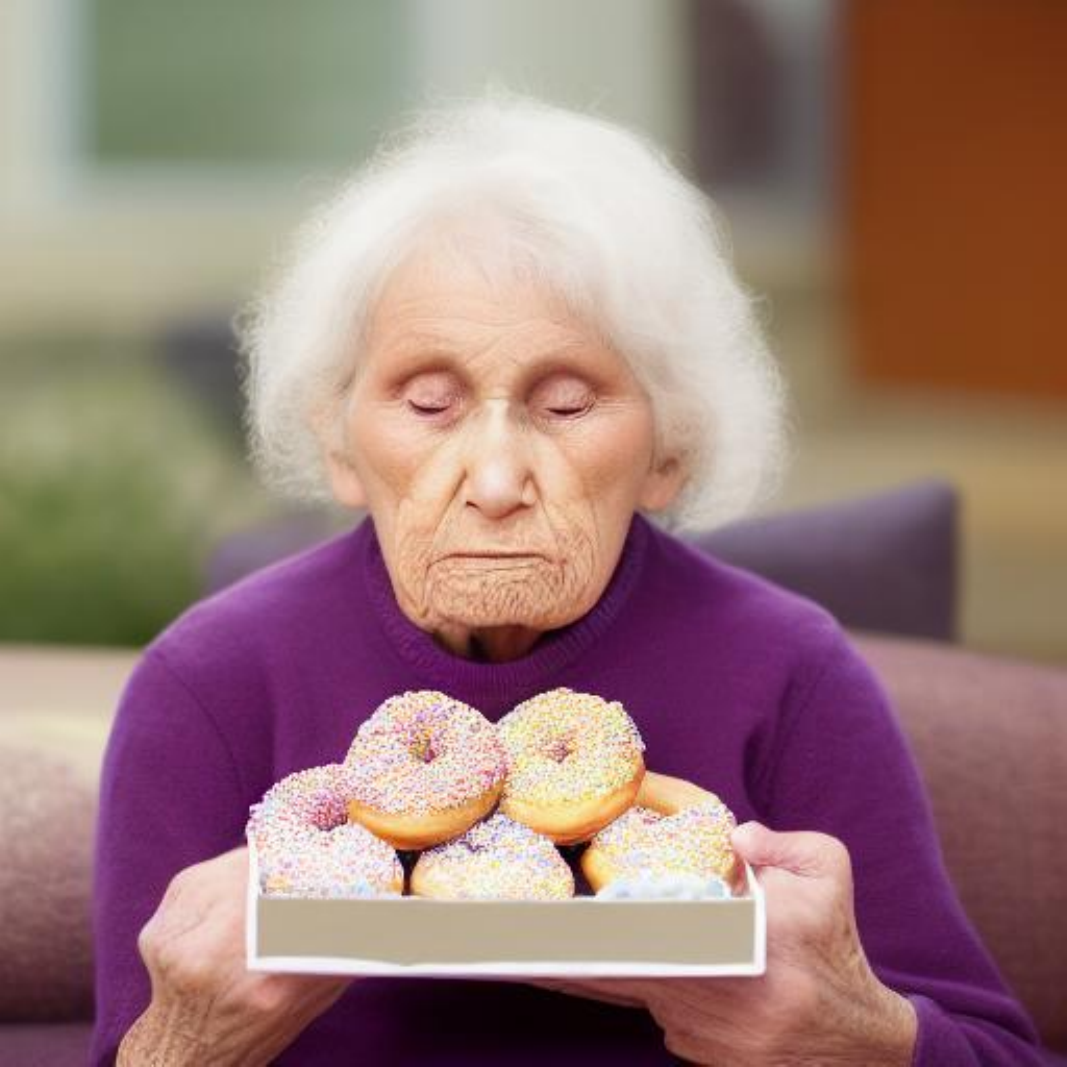}
\includegraphics[width=0.15\columnwidth]{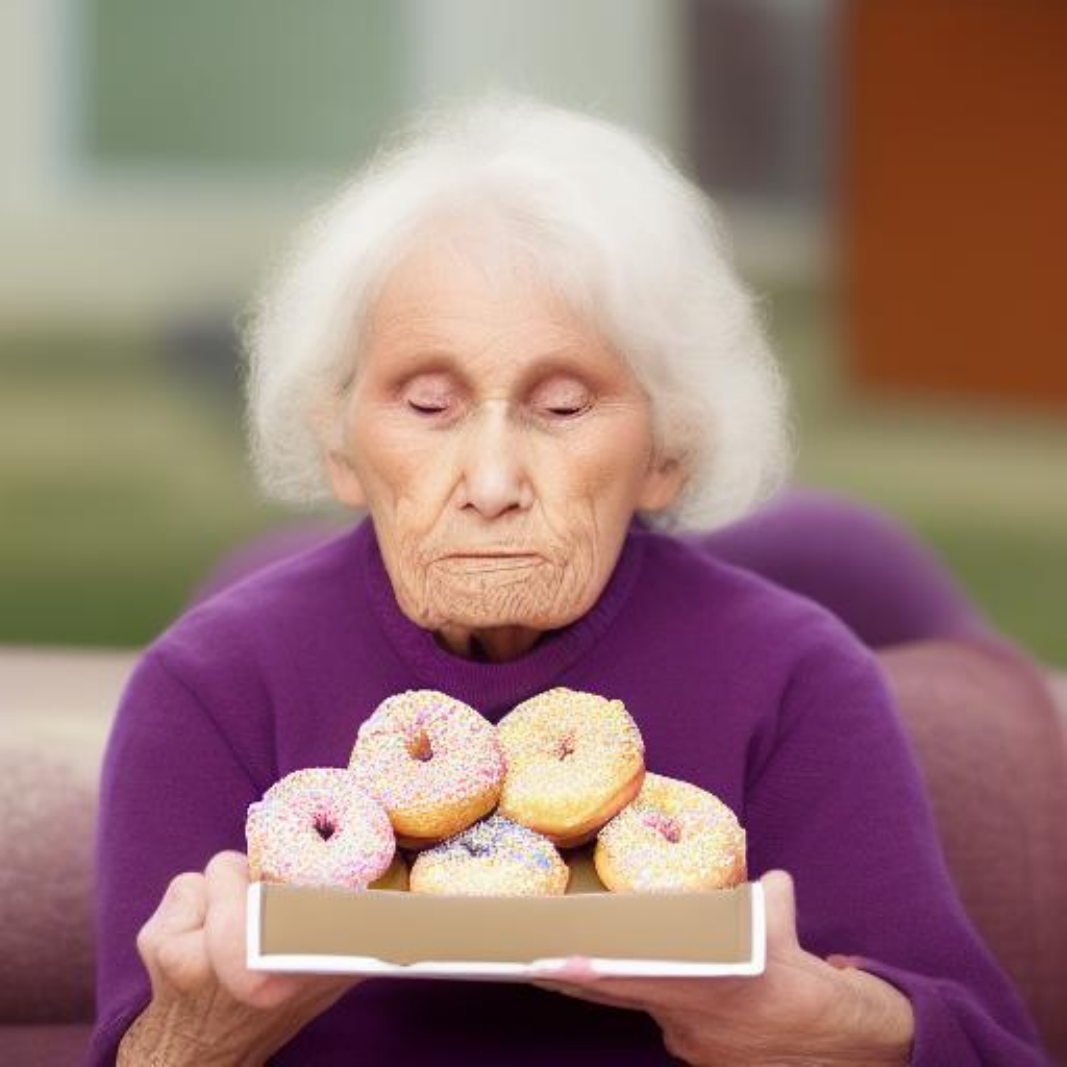}
\includegraphics[width=0.15\columnwidth]{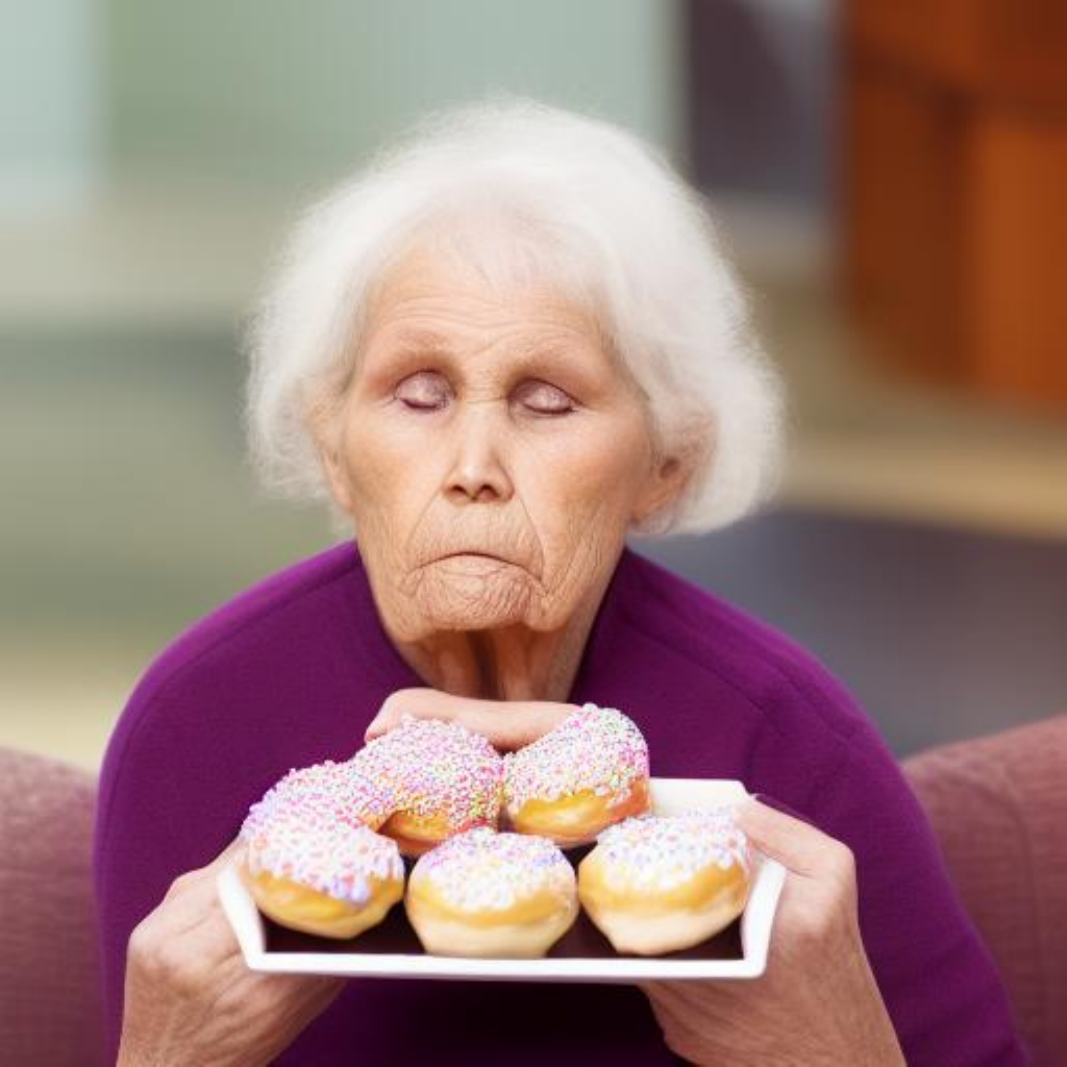}
\includegraphics[width=0.15\columnwidth]{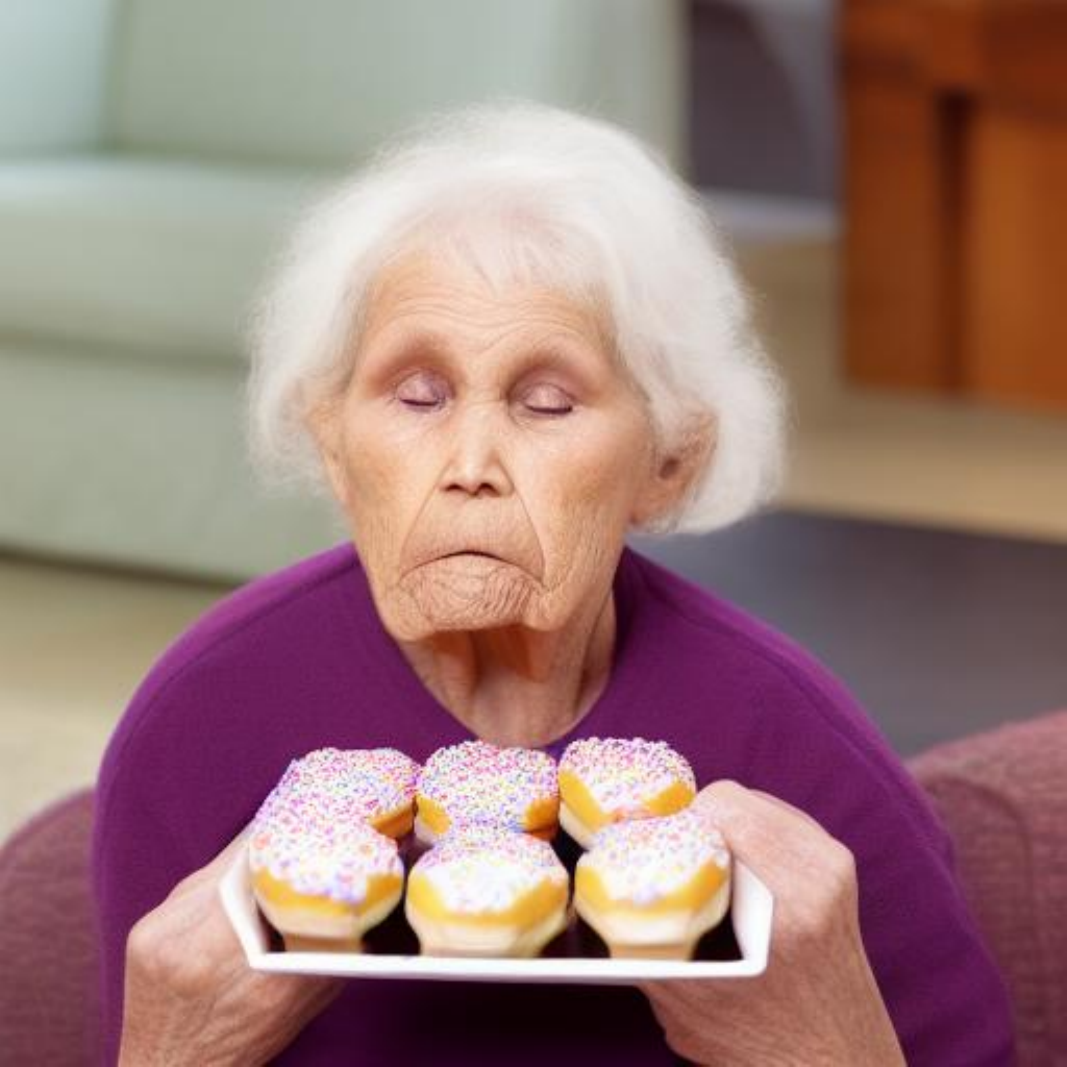}
\\
\includegraphics[width=0.15\columnwidth]{img_dif/SD_20_mix/b_5.pdf}
\includegraphics[width=0.15\columnwidth]{img_dif/SD_20_mix/a_5.pdf}
\includegraphics[width=0.15\columnwidth]{img_dif/SD_20_mix/m_125_5.pdf}
\includegraphics[width=0.15\columnwidth]{img_dif/SD_20_mix/m_155_5.pdf}
\\
\includegraphics[width=0.15\columnwidth]{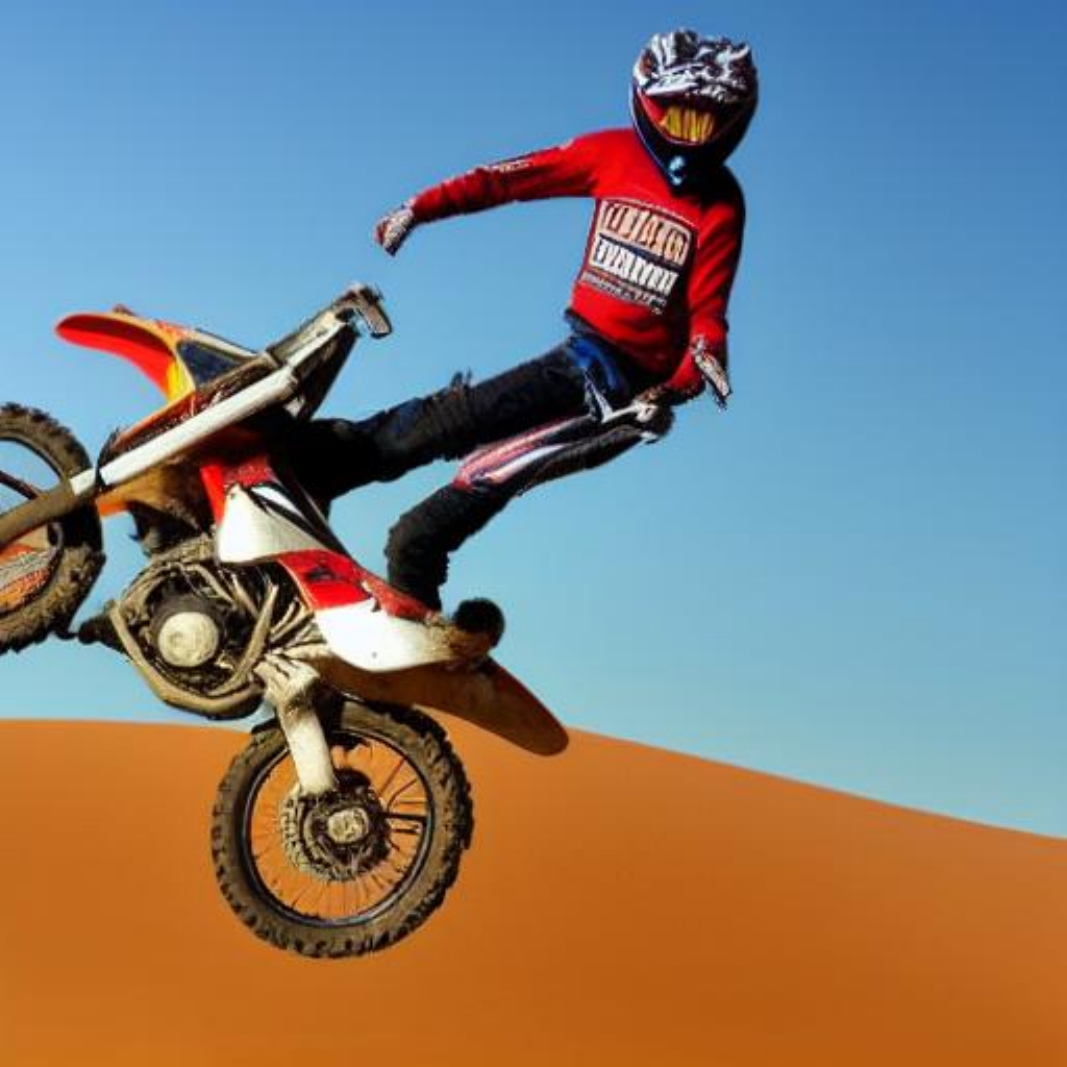}
\includegraphics[width=0.15\columnwidth]{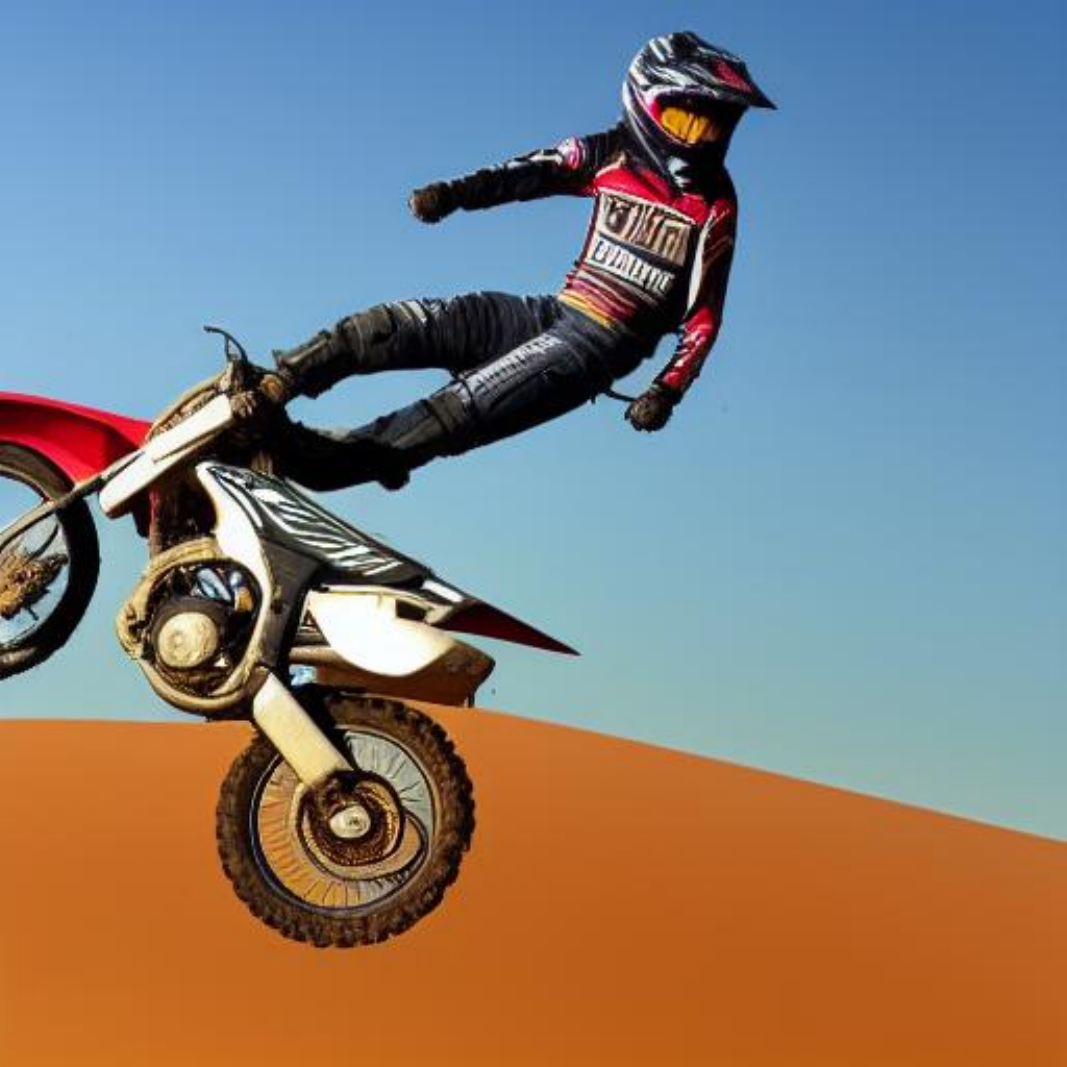}
\includegraphics[width=0.15\columnwidth]{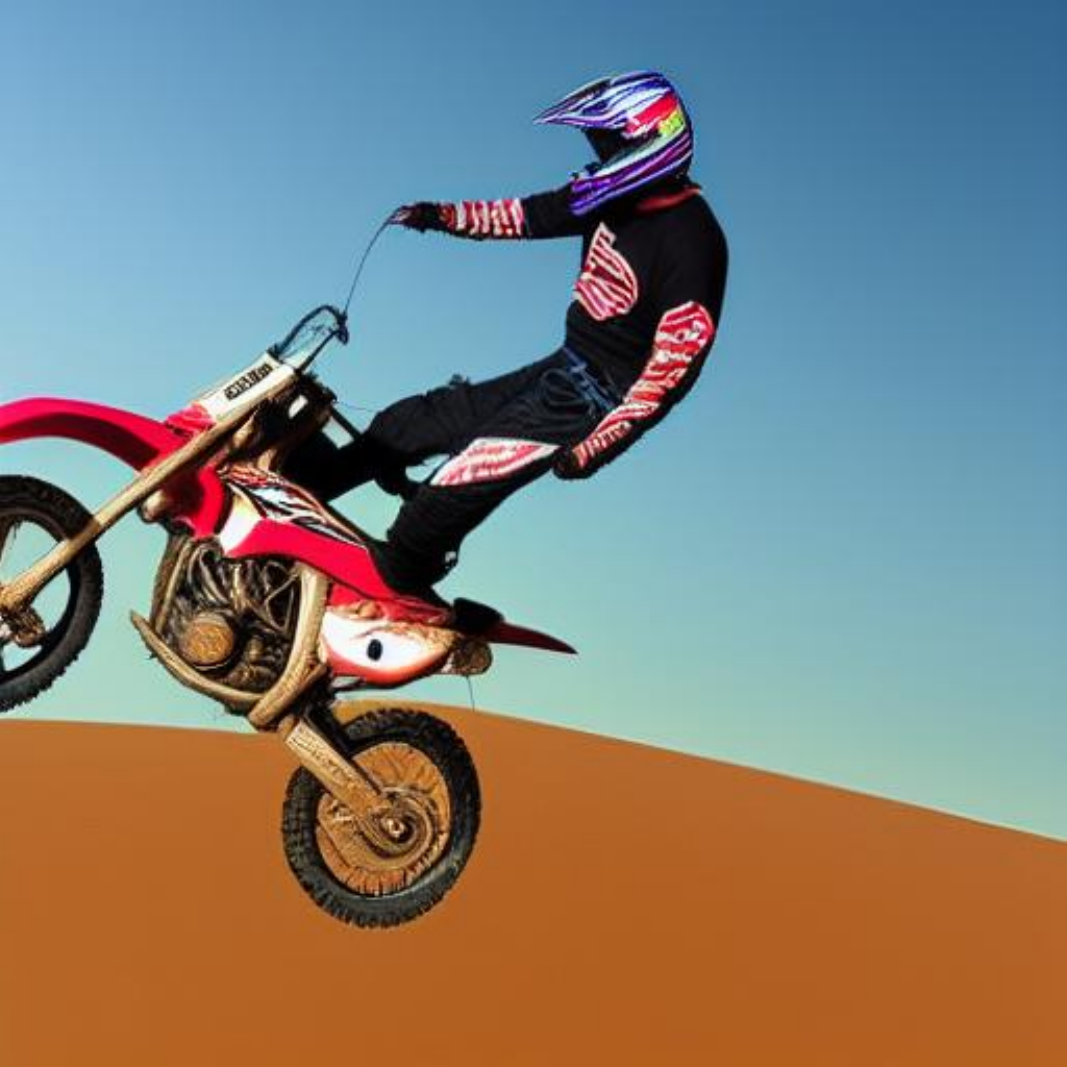}
\includegraphics[width=0.15\columnwidth]{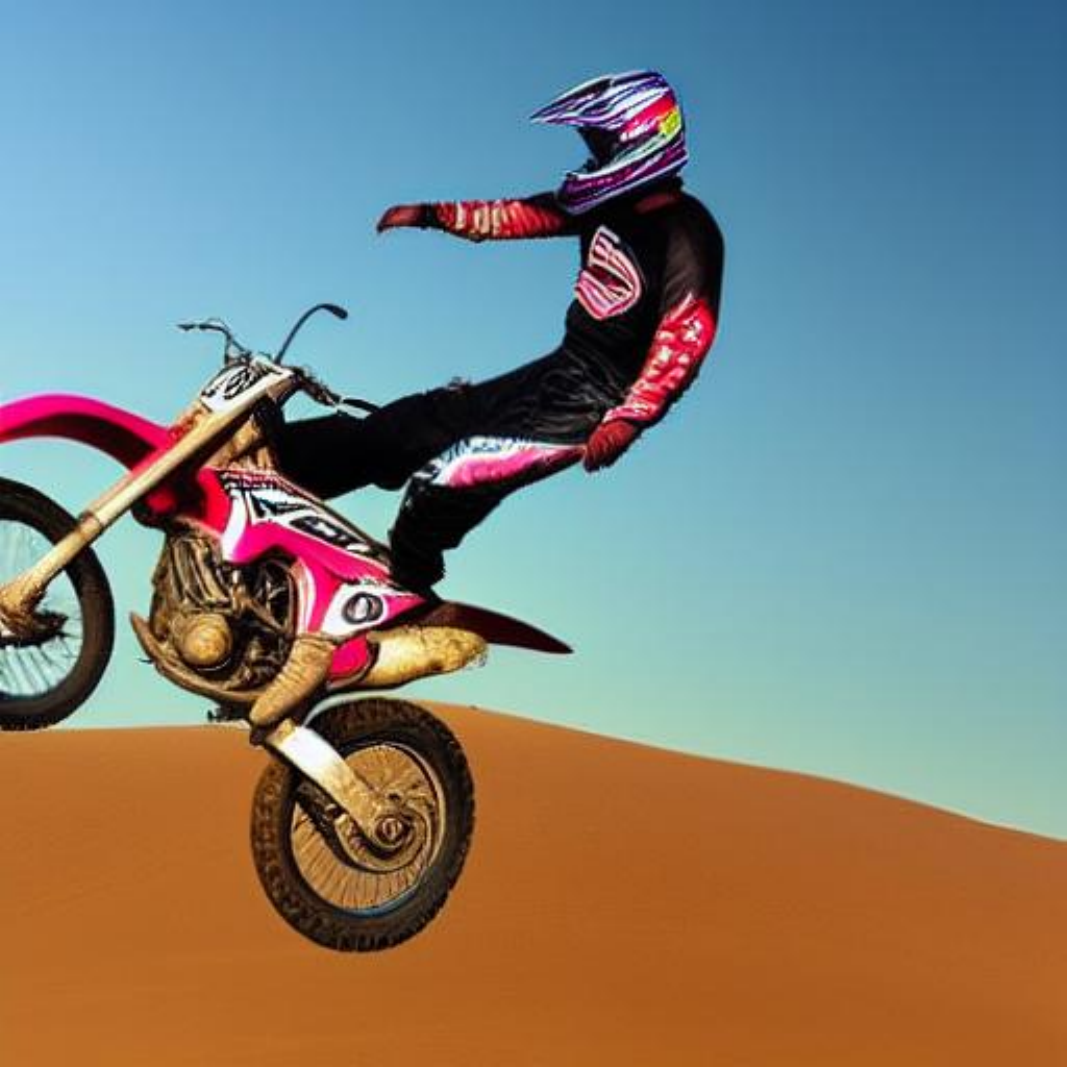}
\\
\includegraphics[width=0.15\columnwidth]{img_dif/SD_20_mix/b_13.pdf}
\includegraphics[width=0.15\columnwidth]{img_dif/SD_20_mix/a_13.pdf}
\includegraphics[width=0.15\columnwidth]{img_dif/SD_20_mix/m_125_13.pdf}
\includegraphics[width=0.15\columnwidth]{img_dif/SD_20_mix/m_155_13.pdf}
\\
\includegraphics[width=0.15\columnwidth]{img_dif/SD_20_mix/b_71.pdf}
\includegraphics[width=0.15\columnwidth]{img_dif/SD_20_mix/a_71.pdf}
\includegraphics[width=0.15\columnwidth]{img_dif/SD_20_mix/m_125_71.pdf}
\includegraphics[width=0.15\columnwidth]{img_dif/SD_20_mix/m_155_71.pdf}
\\
\includegraphics[width=0.15\columnwidth]{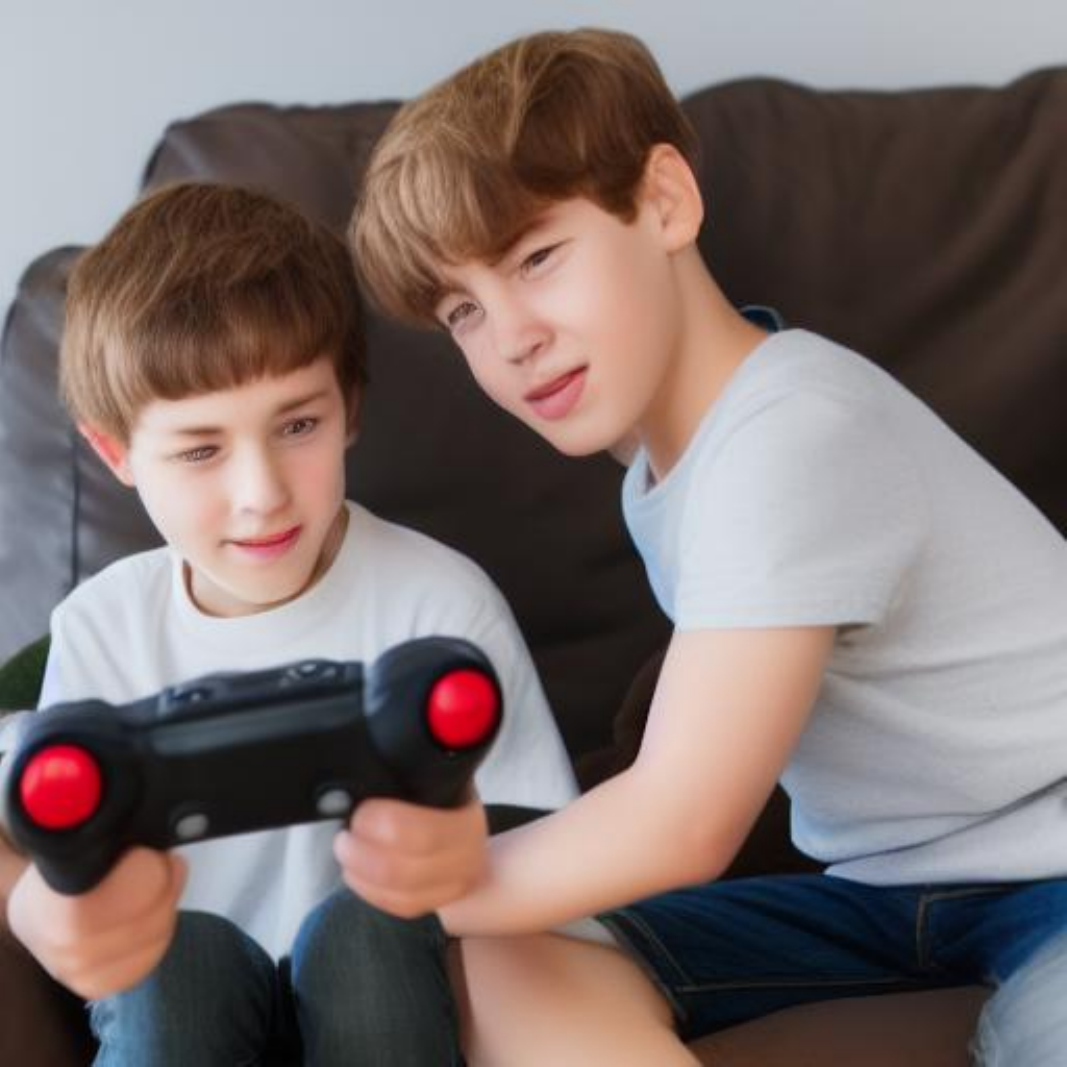}
\includegraphics[width=0.15\columnwidth]{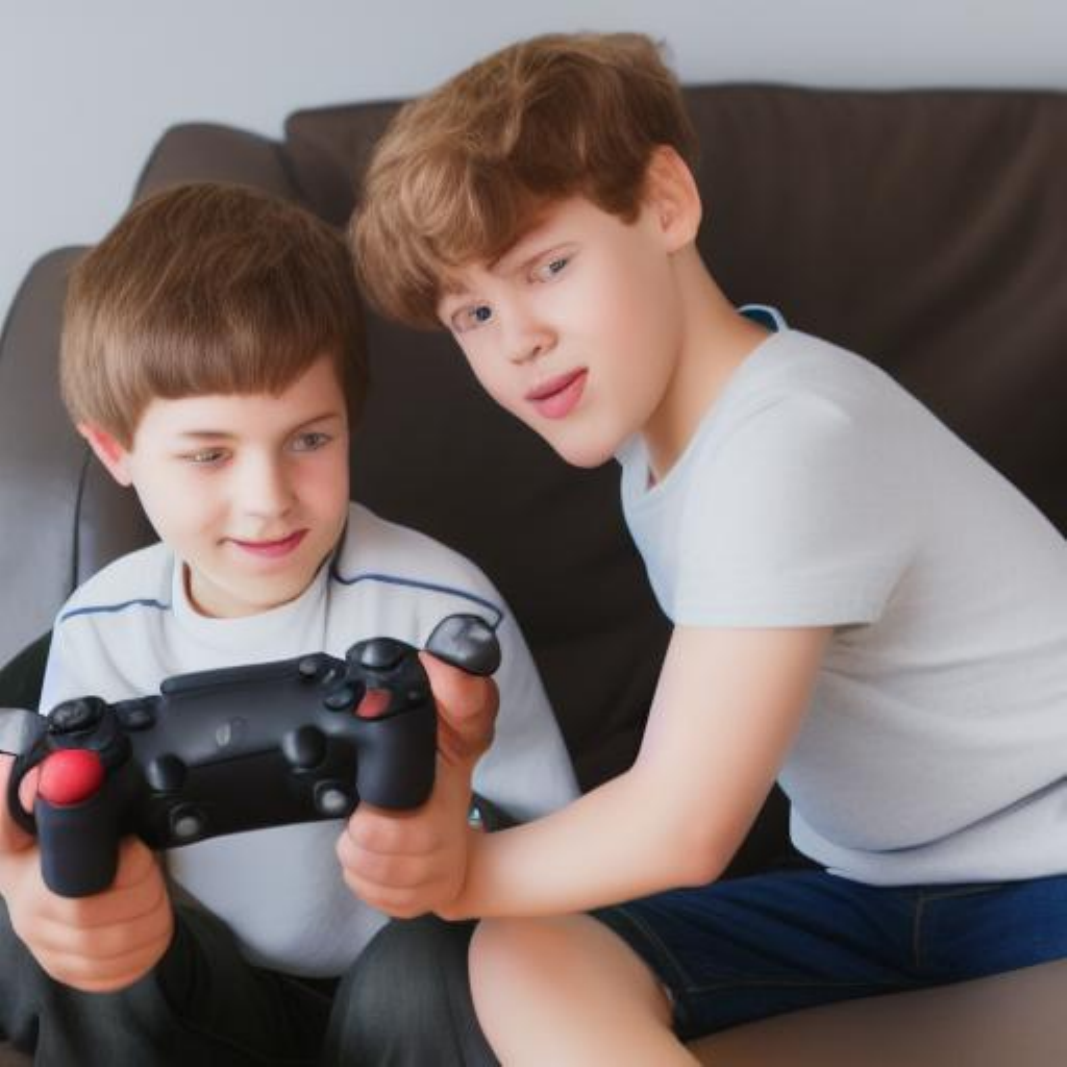}
\includegraphics[width=0.15\columnwidth]{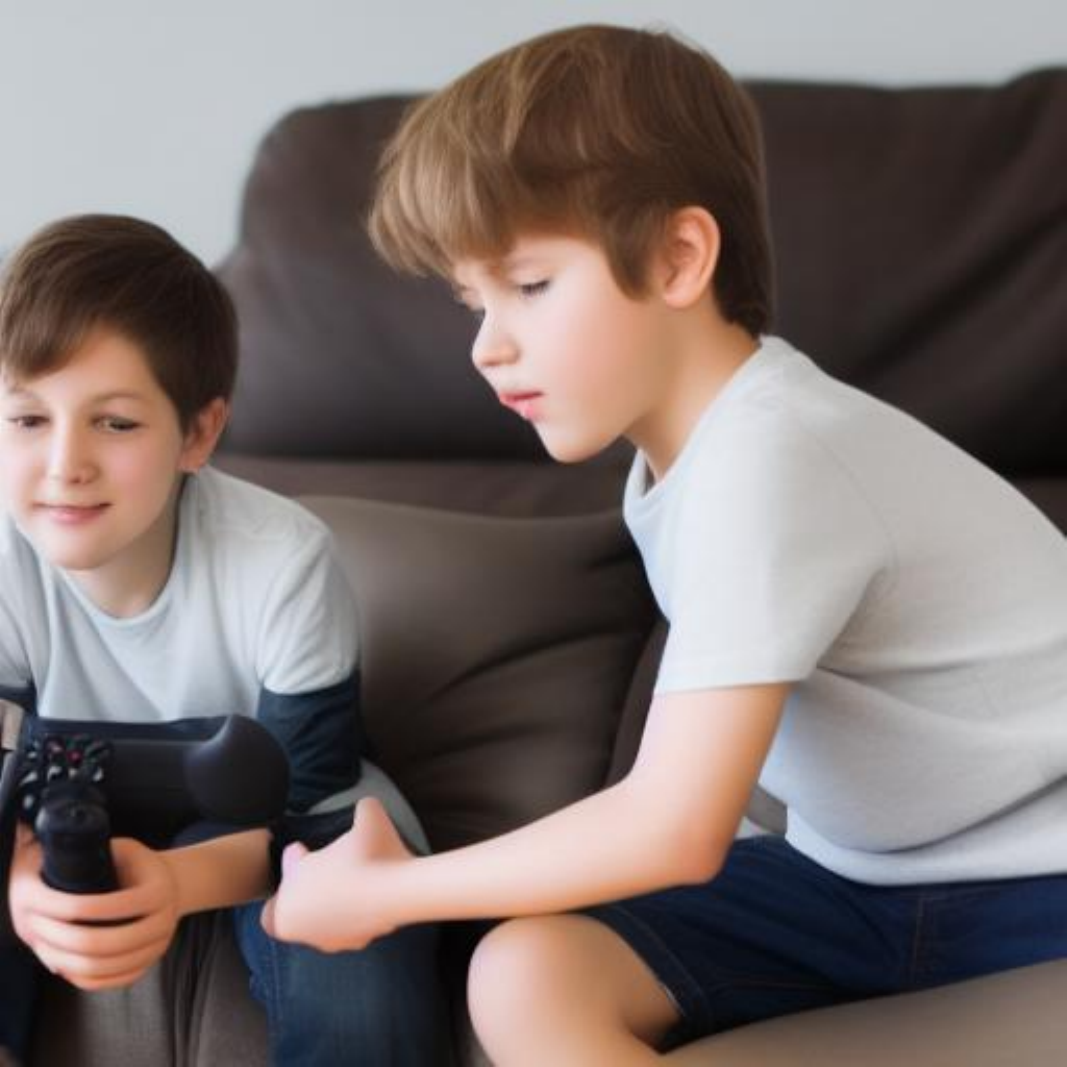}
\includegraphics[width=0.15\columnwidth]{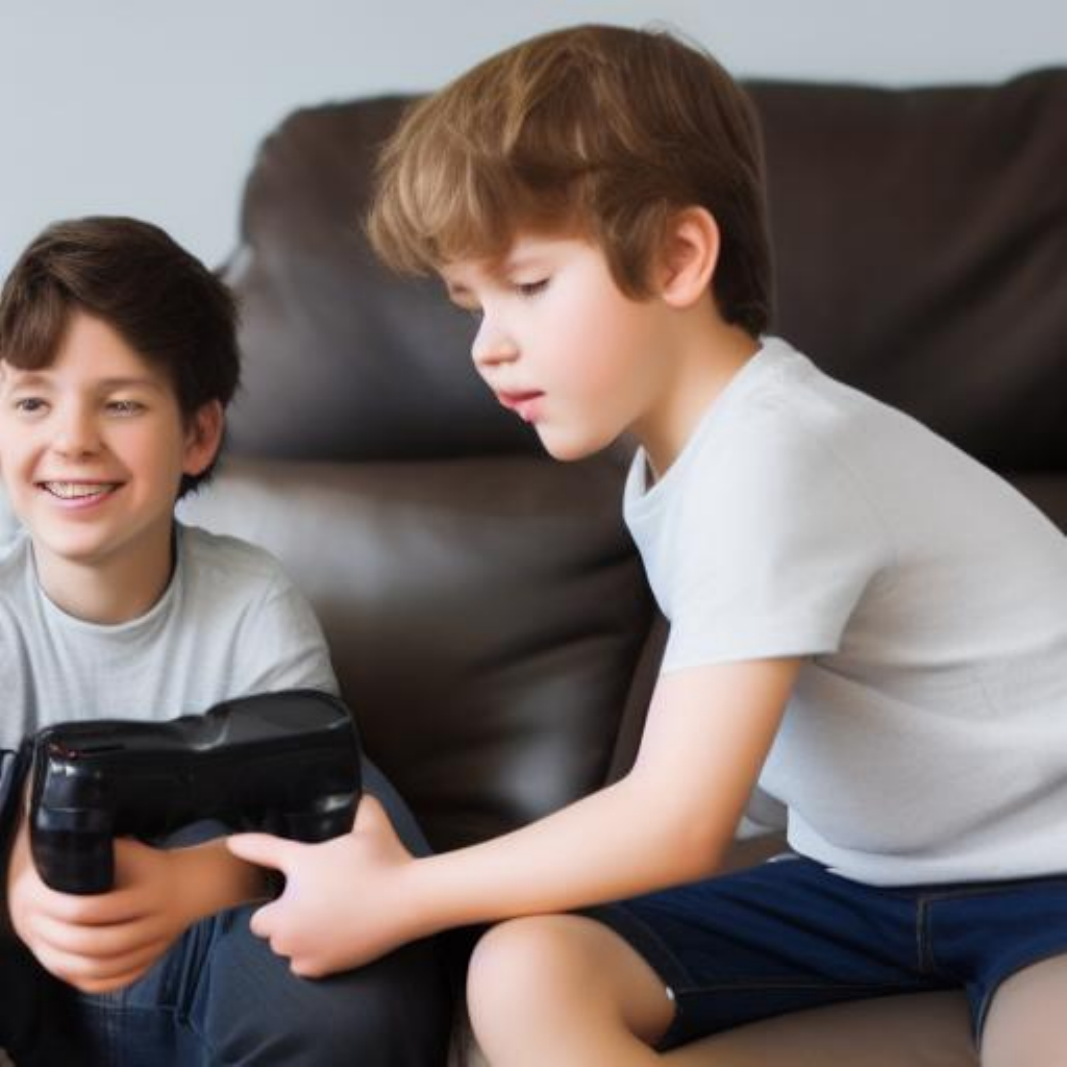}
\\
\includegraphics[width=0.15\columnwidth]{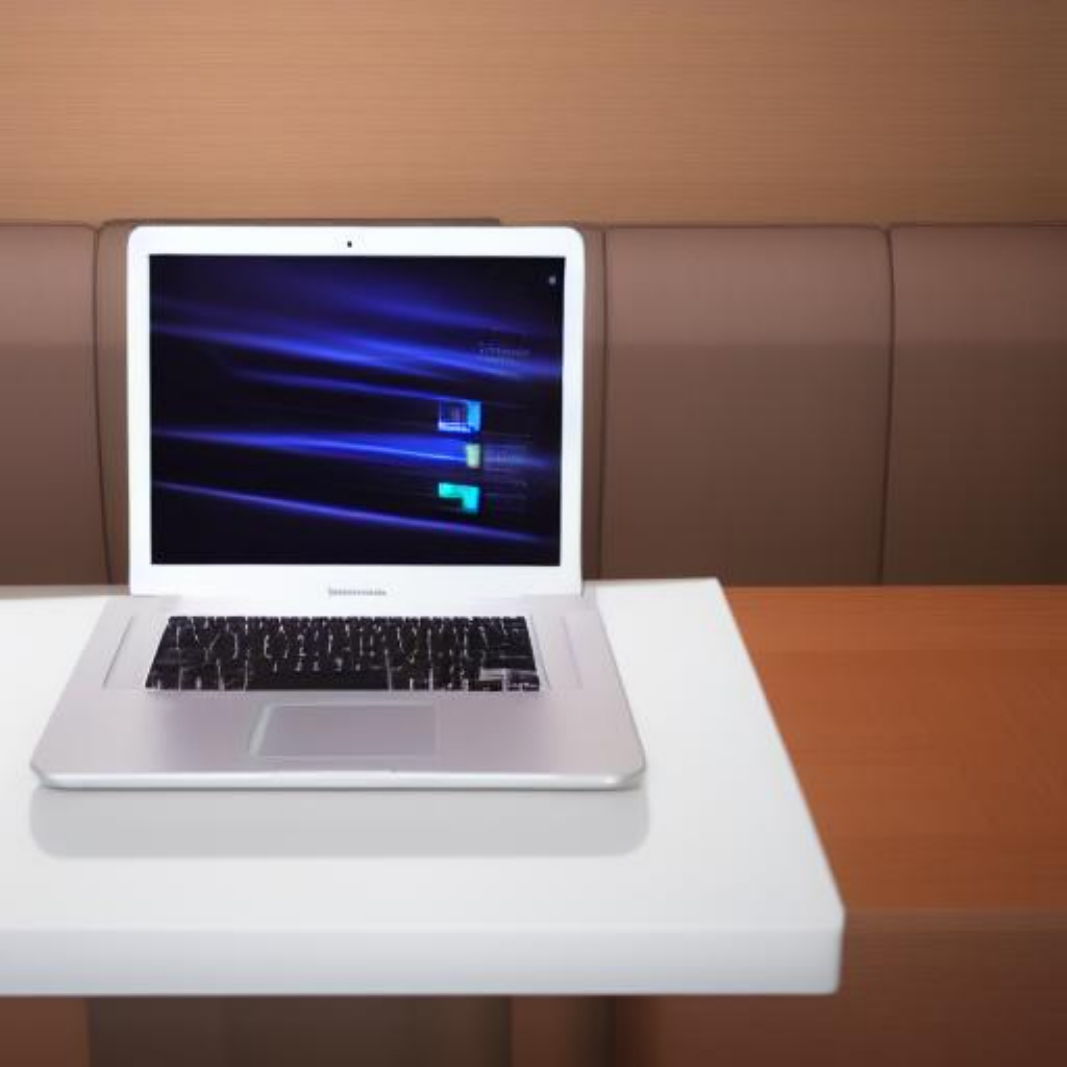}
\includegraphics[width=0.15\columnwidth]{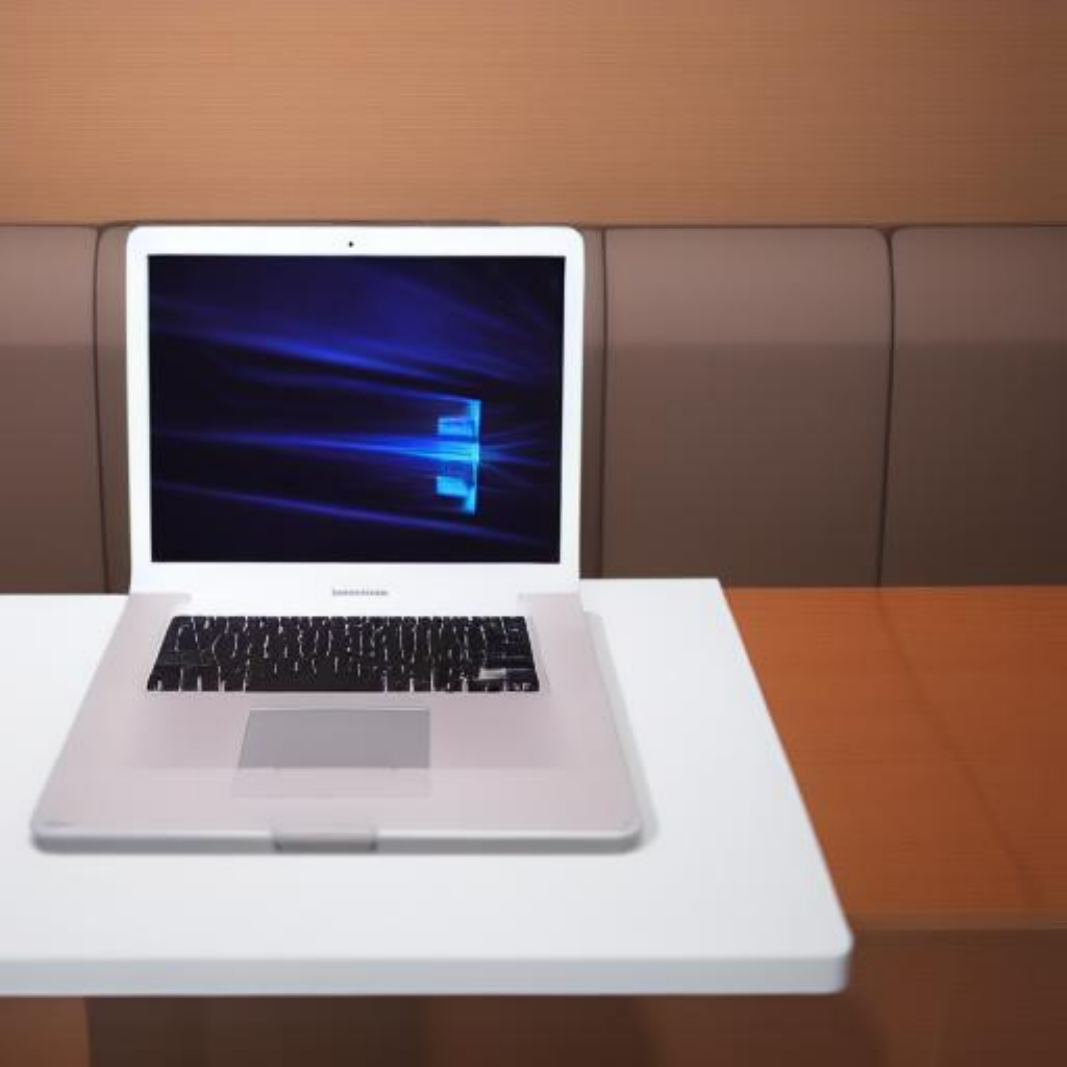}
\includegraphics[width=0.15\columnwidth]{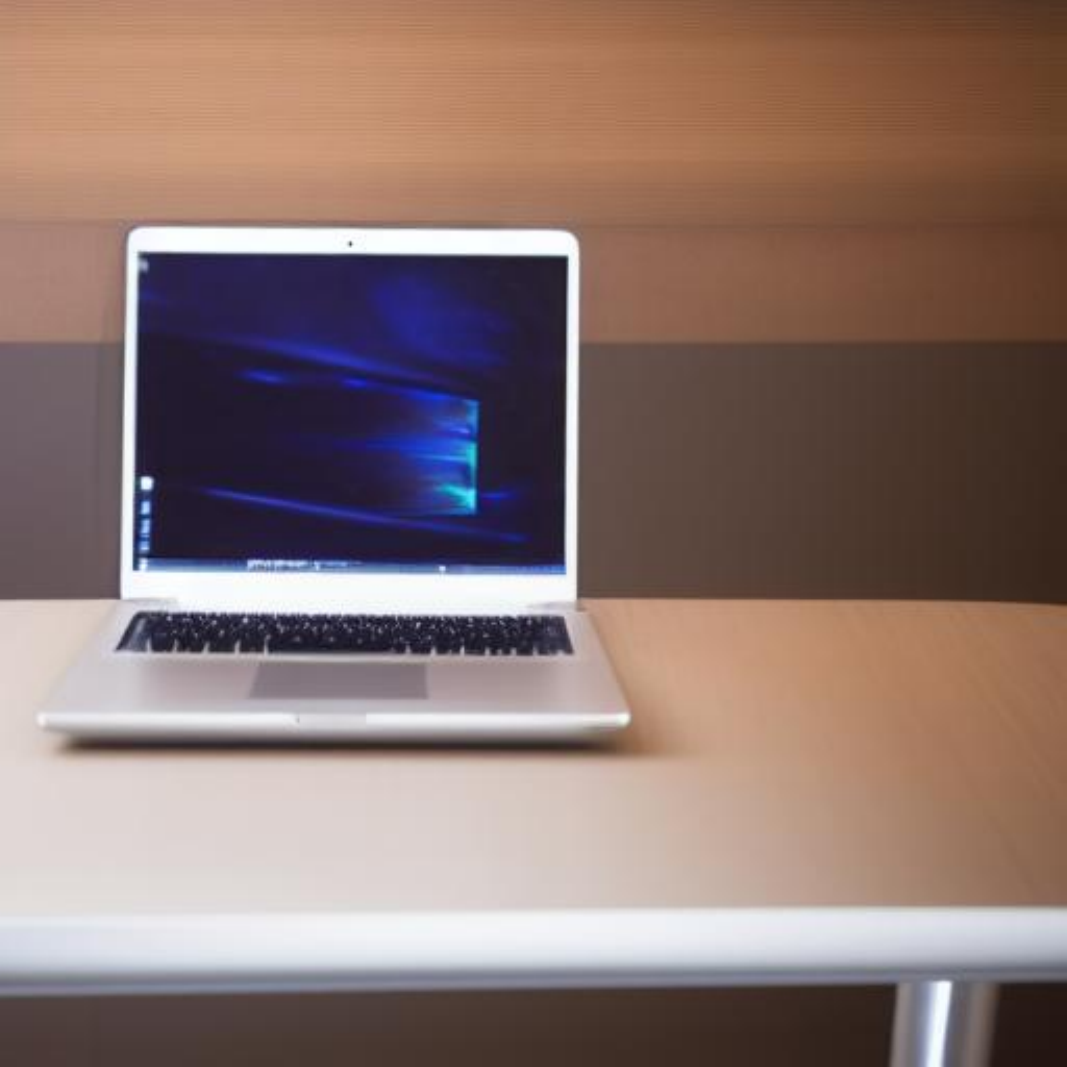}
\includegraphics[width=0.15\columnwidth]{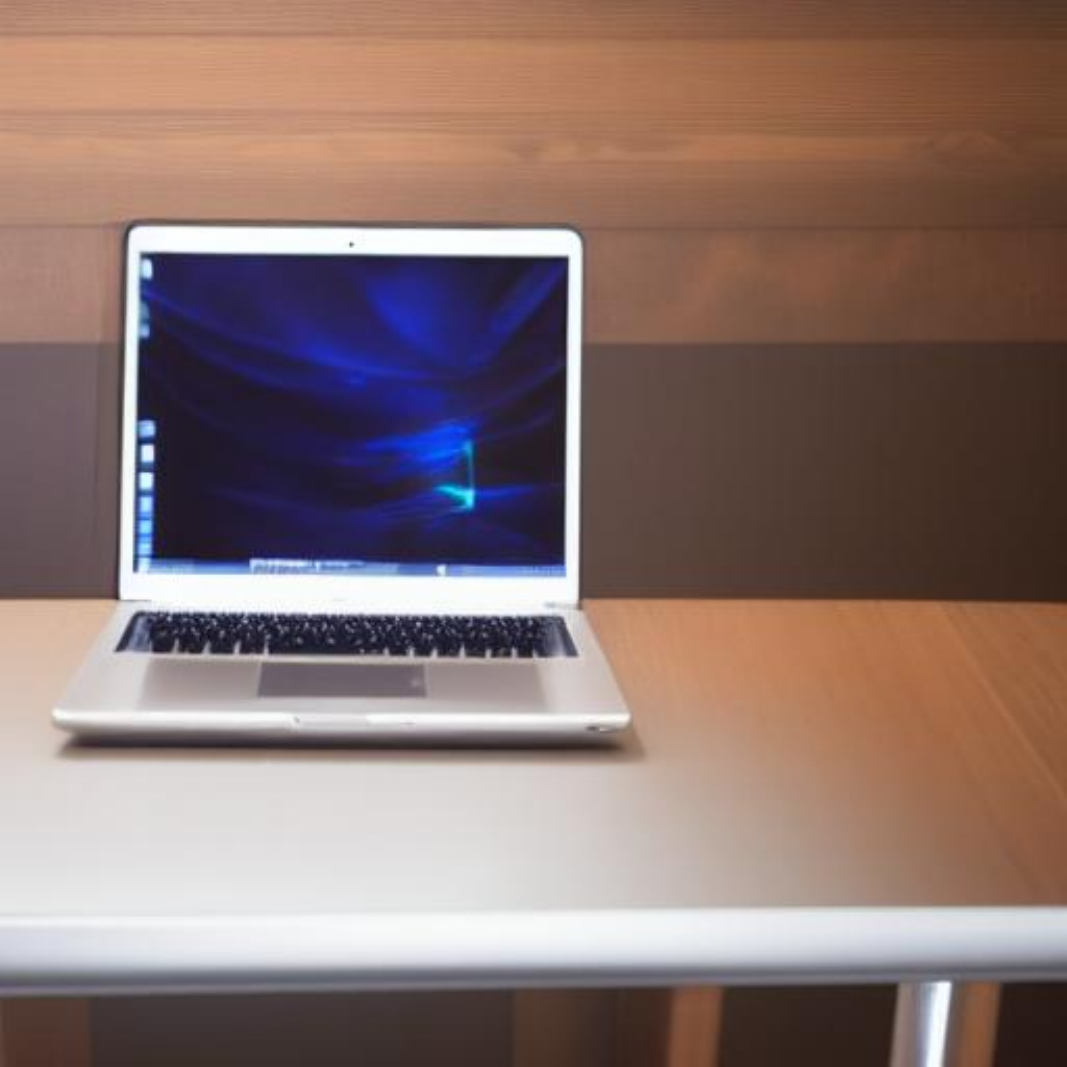}
\caption{\textbf{SD with $20$ Steps.} For each row, from left to right: baseline, ablation, generated image with variance scaling $1.25$, and one with $1.55$. 
%
%
}
    \label{fig:sd-20}
    \end{figure}

\begin{figure}[t]
    \centering
\includegraphics[width=0.15\columnwidth]{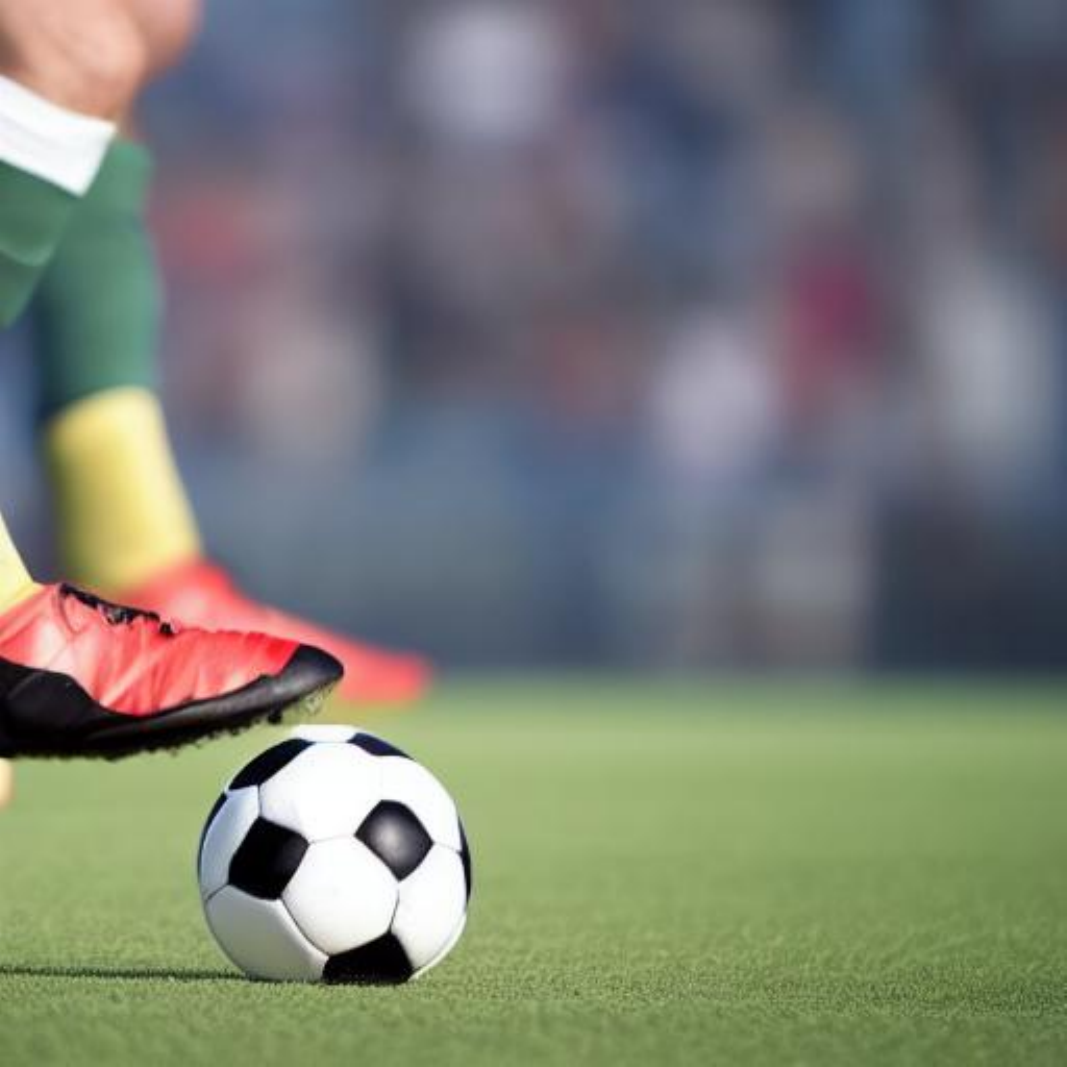}
\includegraphics[width=0.15\columnwidth]{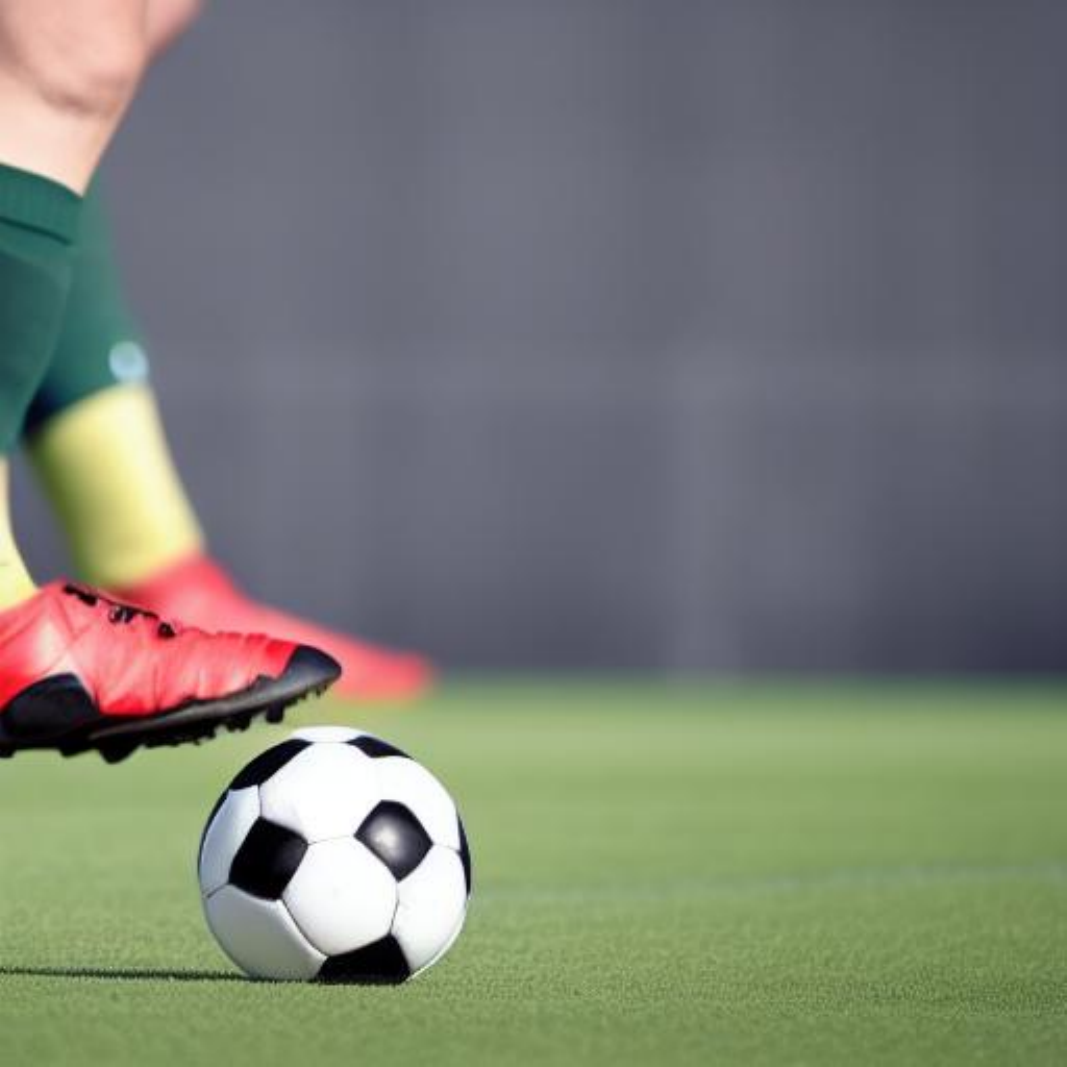}
\includegraphics[width=0.15\columnwidth]{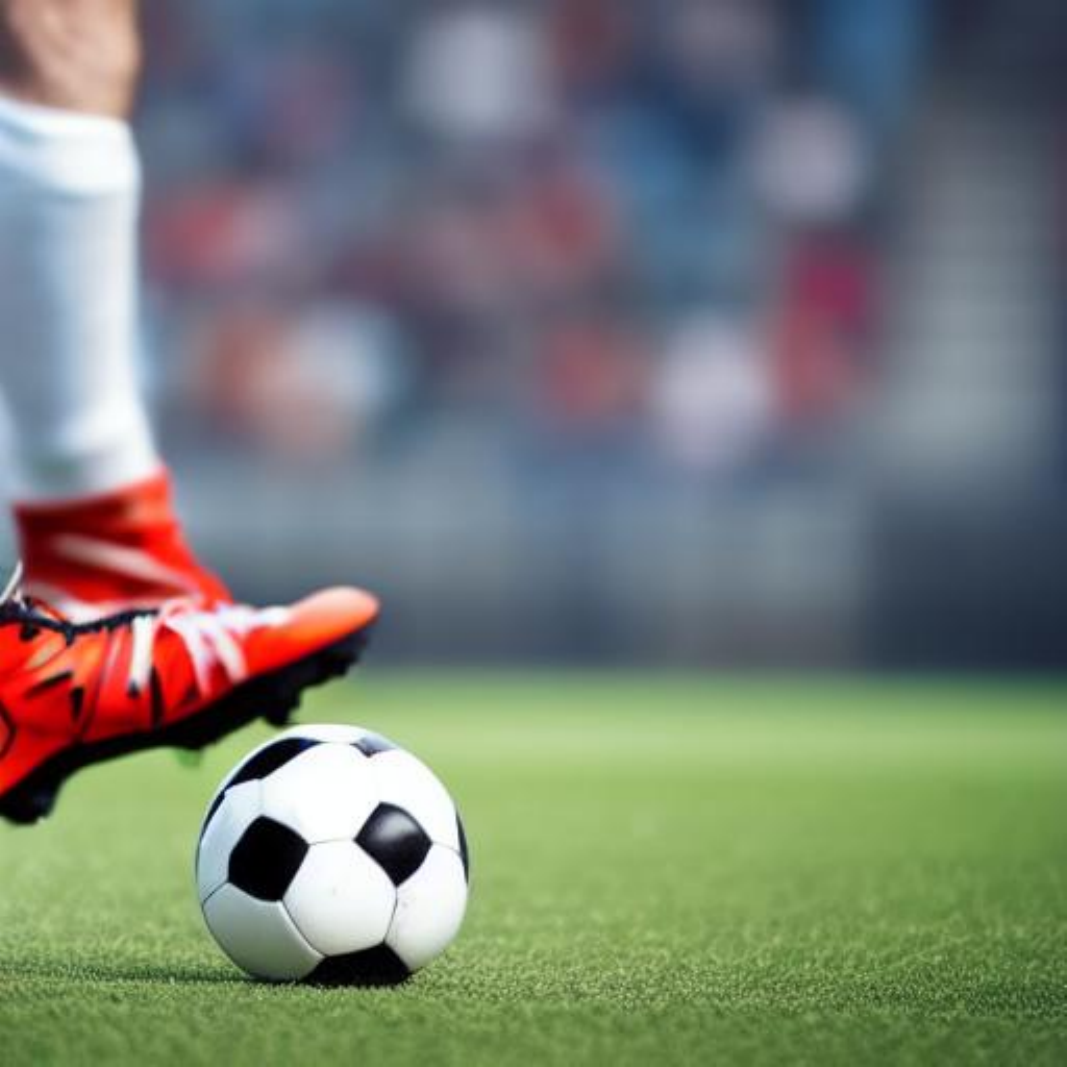}
\includegraphics[width=0.15\columnwidth]{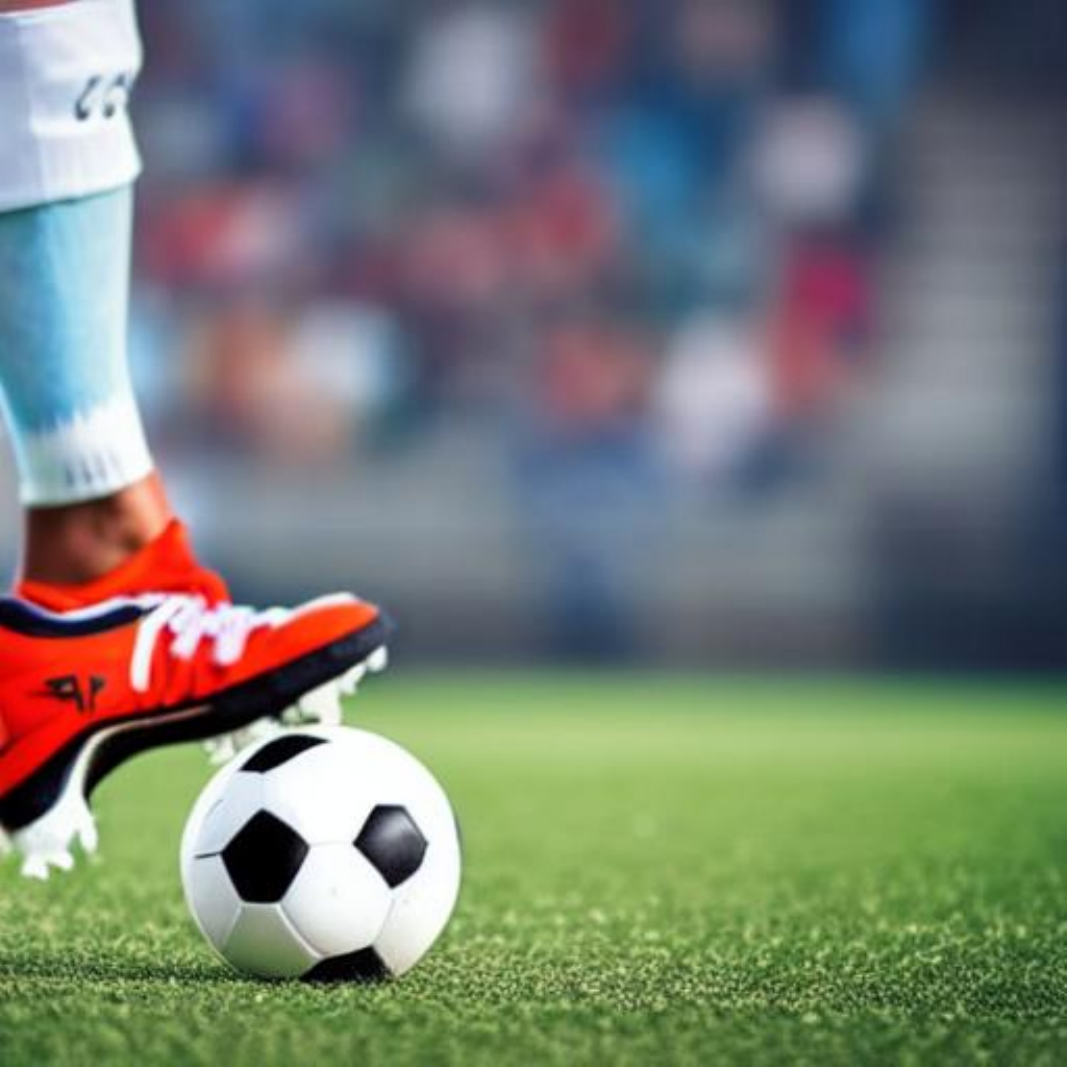}
\\
\includegraphics[width=0.15\columnwidth]{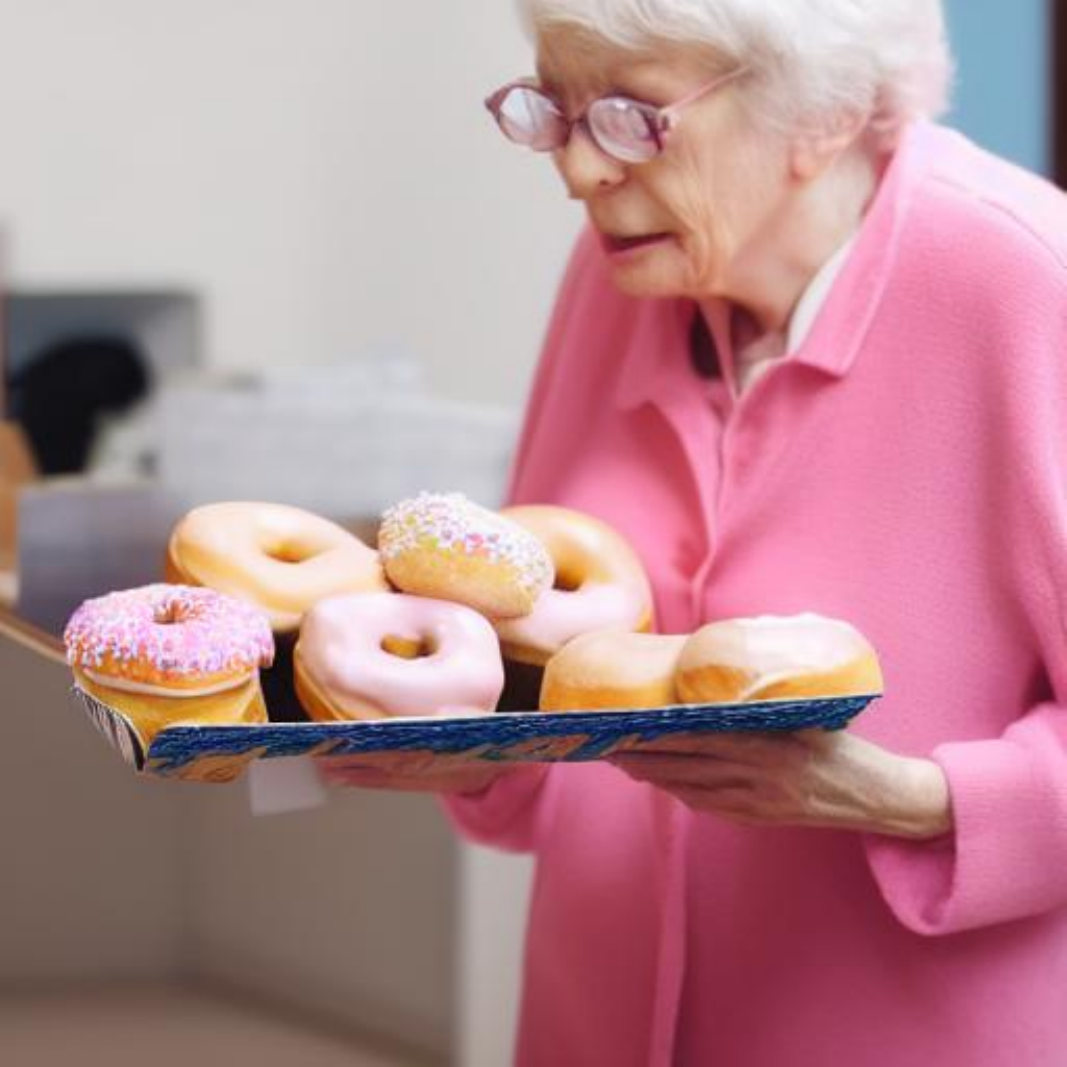}
\includegraphics[width=0.15\columnwidth]{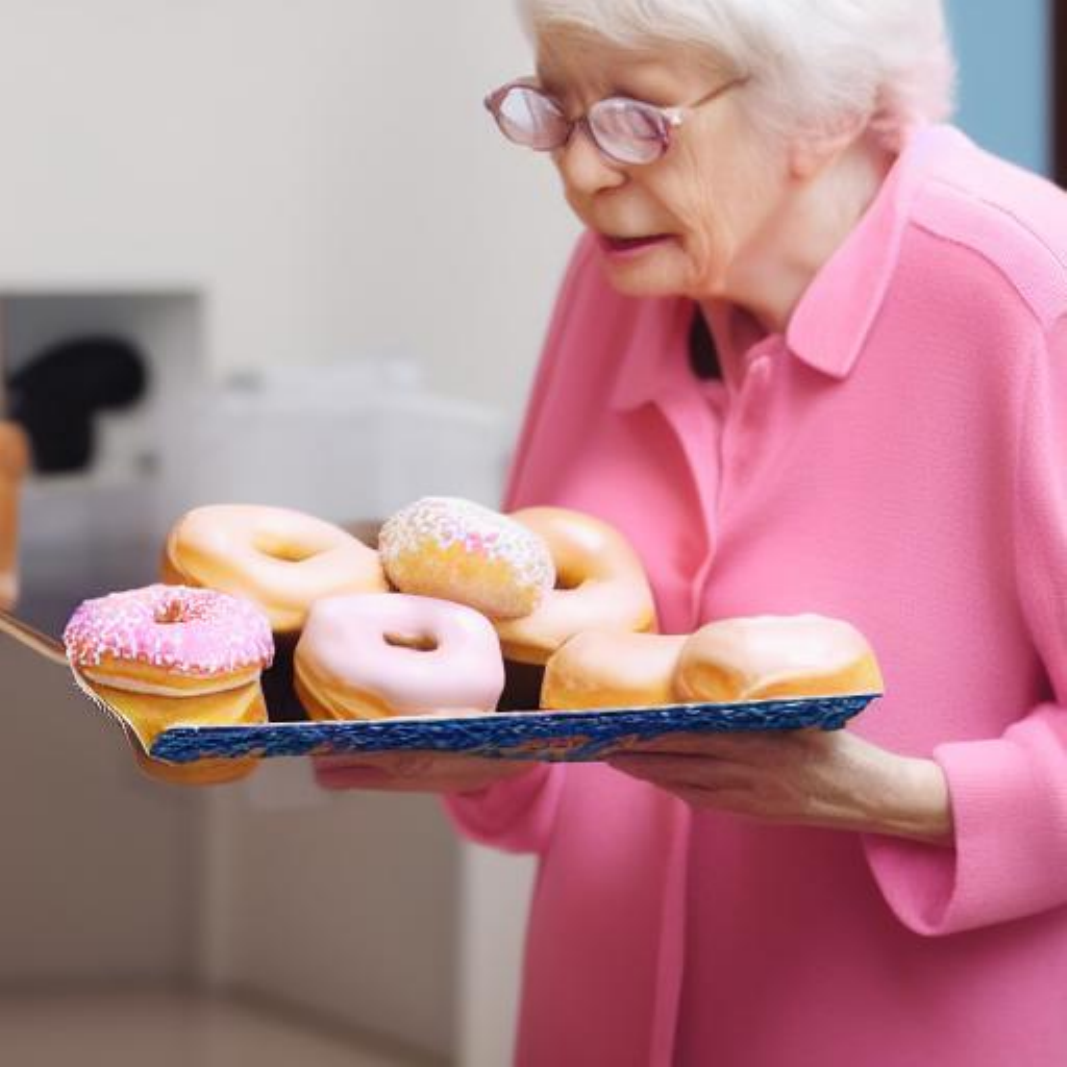}
\includegraphics[width=0.15\columnwidth]{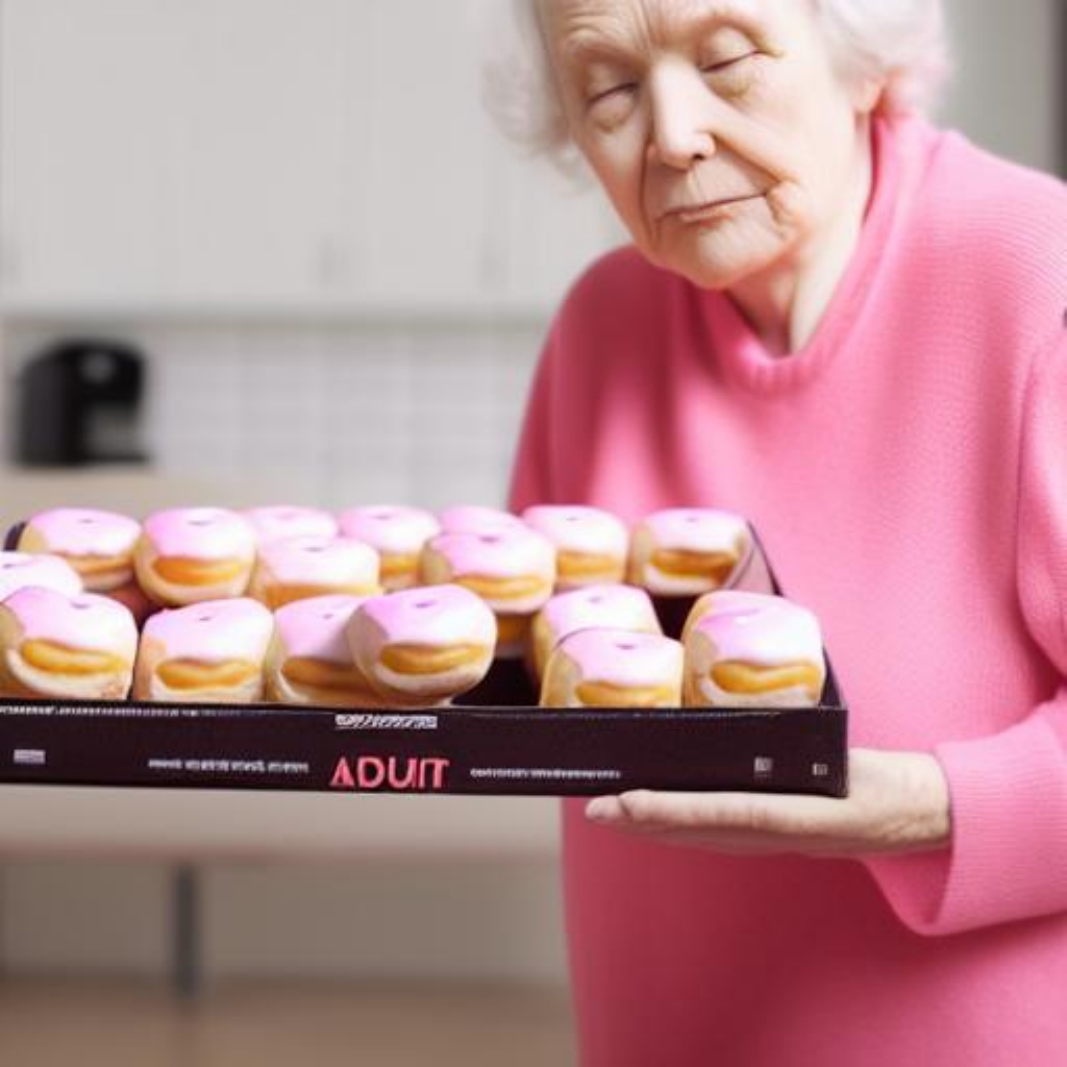}
\includegraphics[width=0.15\columnwidth]{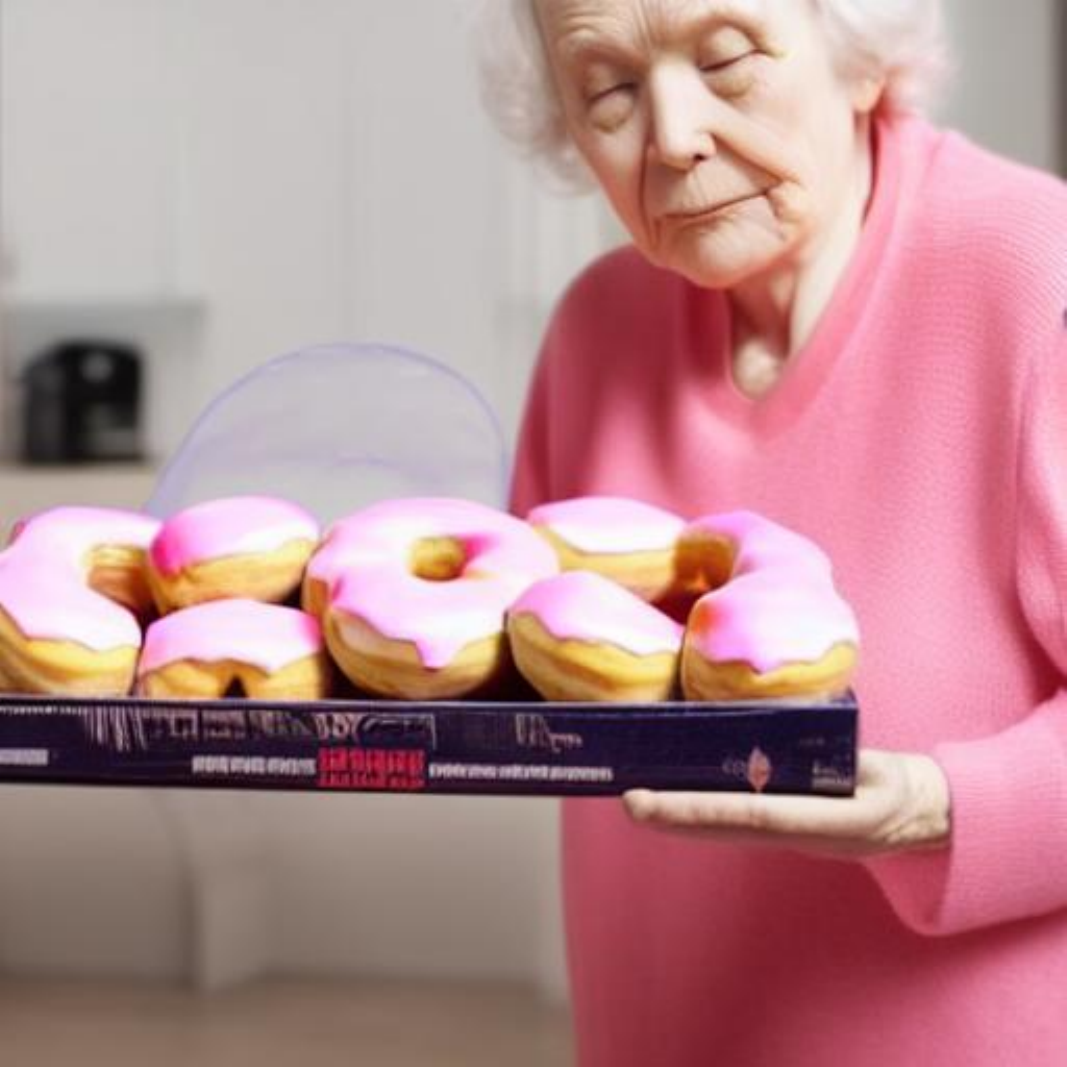}
\\
\includegraphics[width=0.15\columnwidth]{img_dif/SD_40_mix/b_5.pdf}
\includegraphics[width=0.15\columnwidth]{img_dif/SD_40_mix/a_5.pdf}
\includegraphics[width=0.15\columnwidth]{img_dif/SD_40_mix/m_125_5.pdf}
\includegraphics[width=0.15\columnwidth]{img_dif/SD_40_mix/m_155_5.pdf}
\\
\includegraphics[width=0.15\columnwidth]{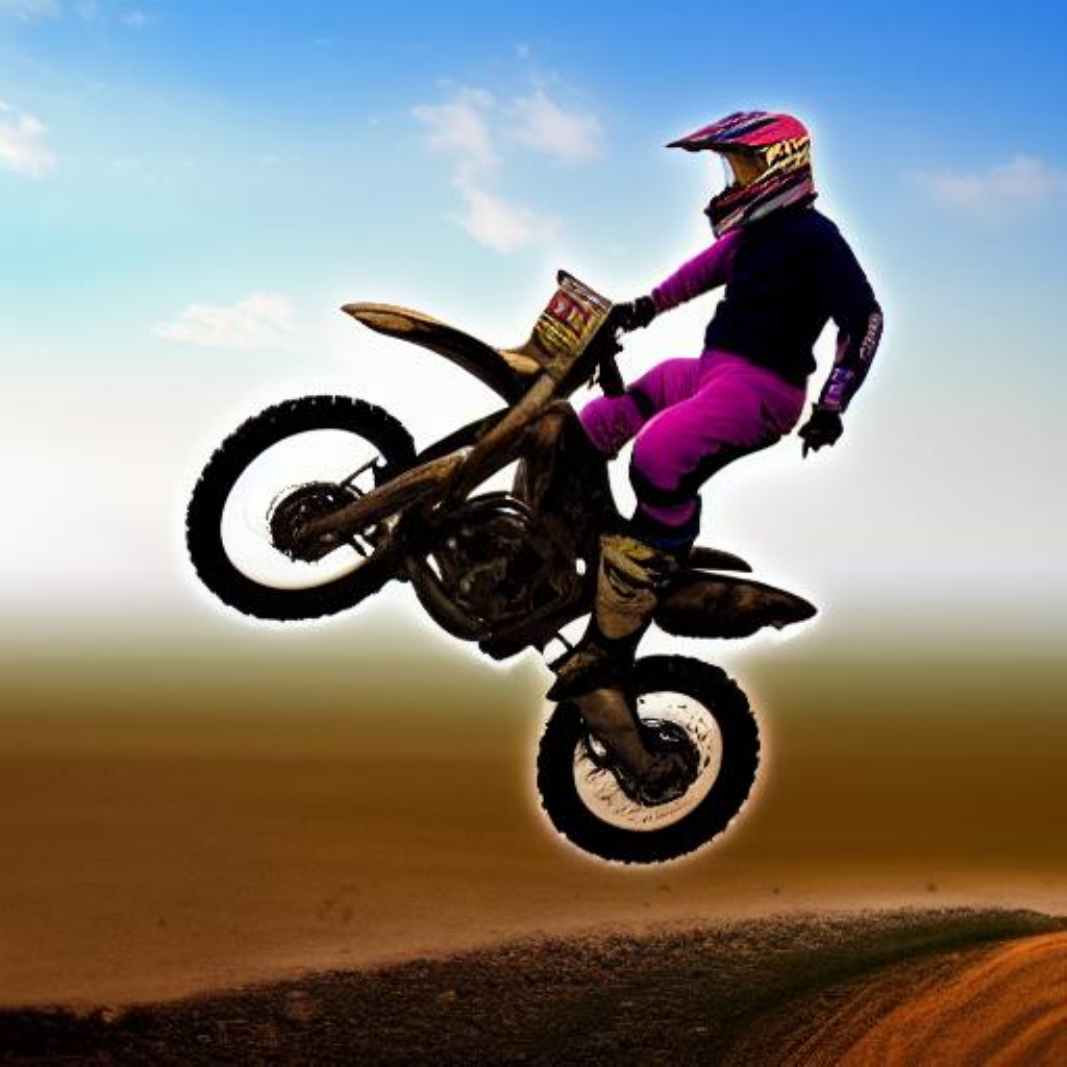}
\includegraphics[width=0.15\columnwidth]{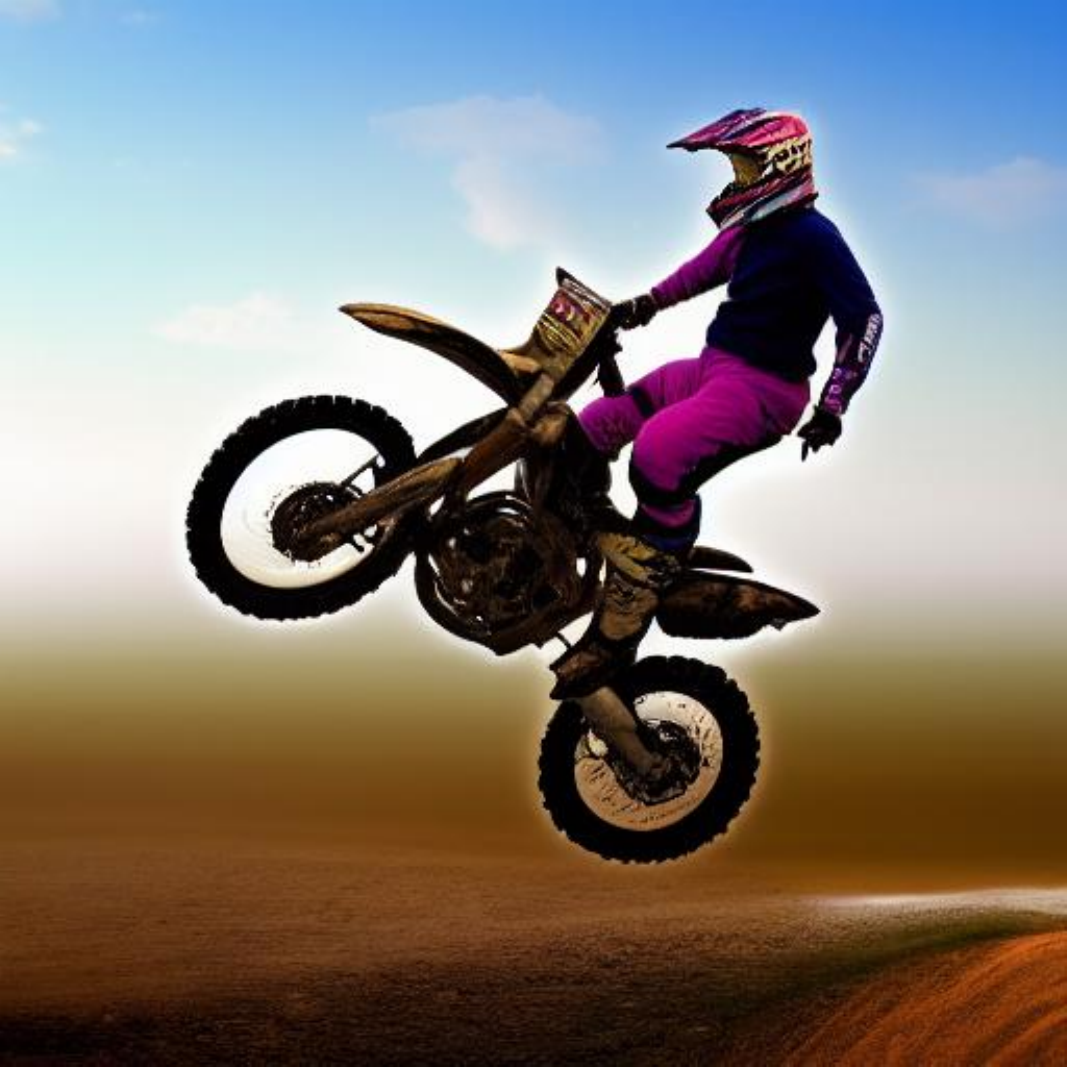}
\includegraphics[width=0.15\columnwidth]{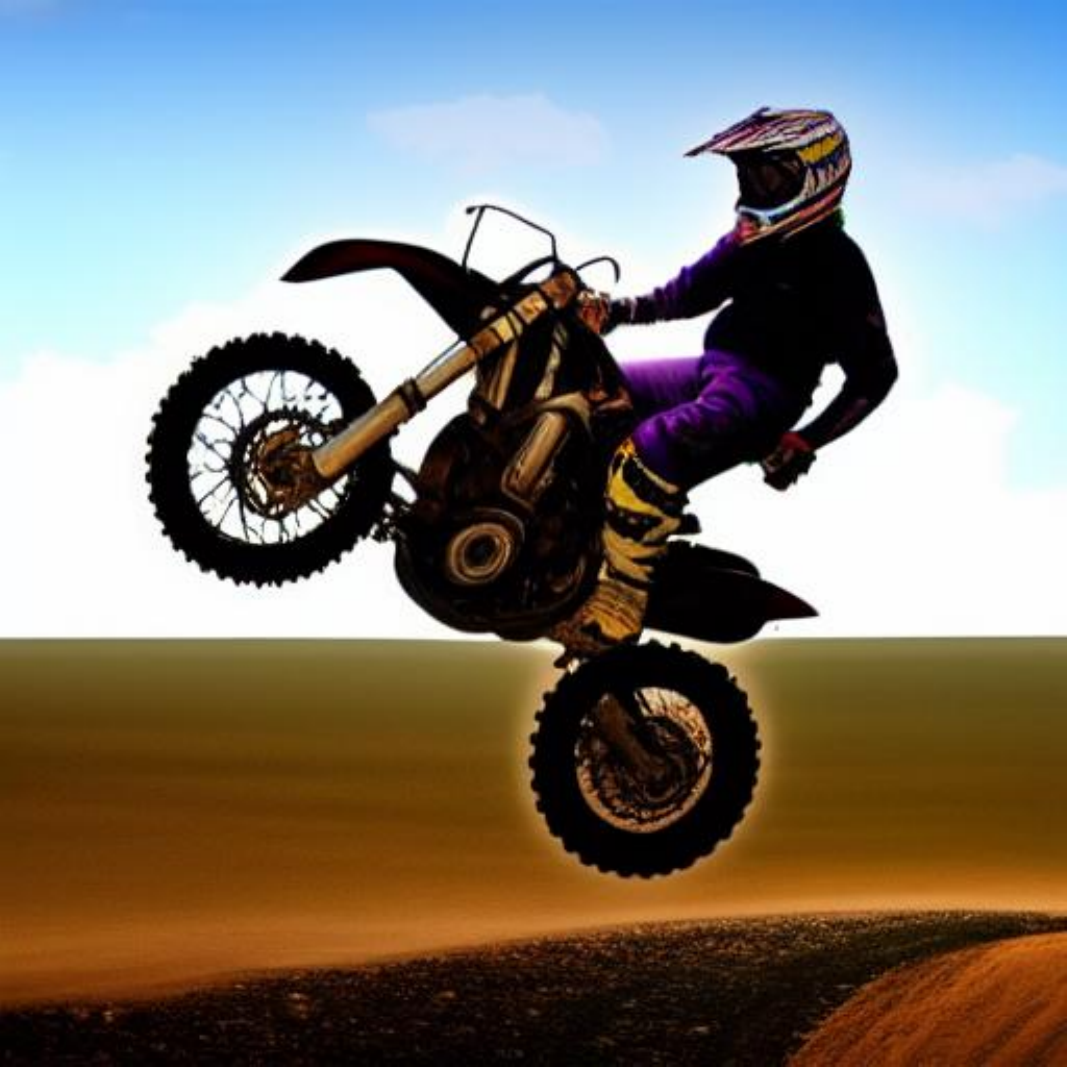}
\includegraphics[width=0.15\columnwidth]{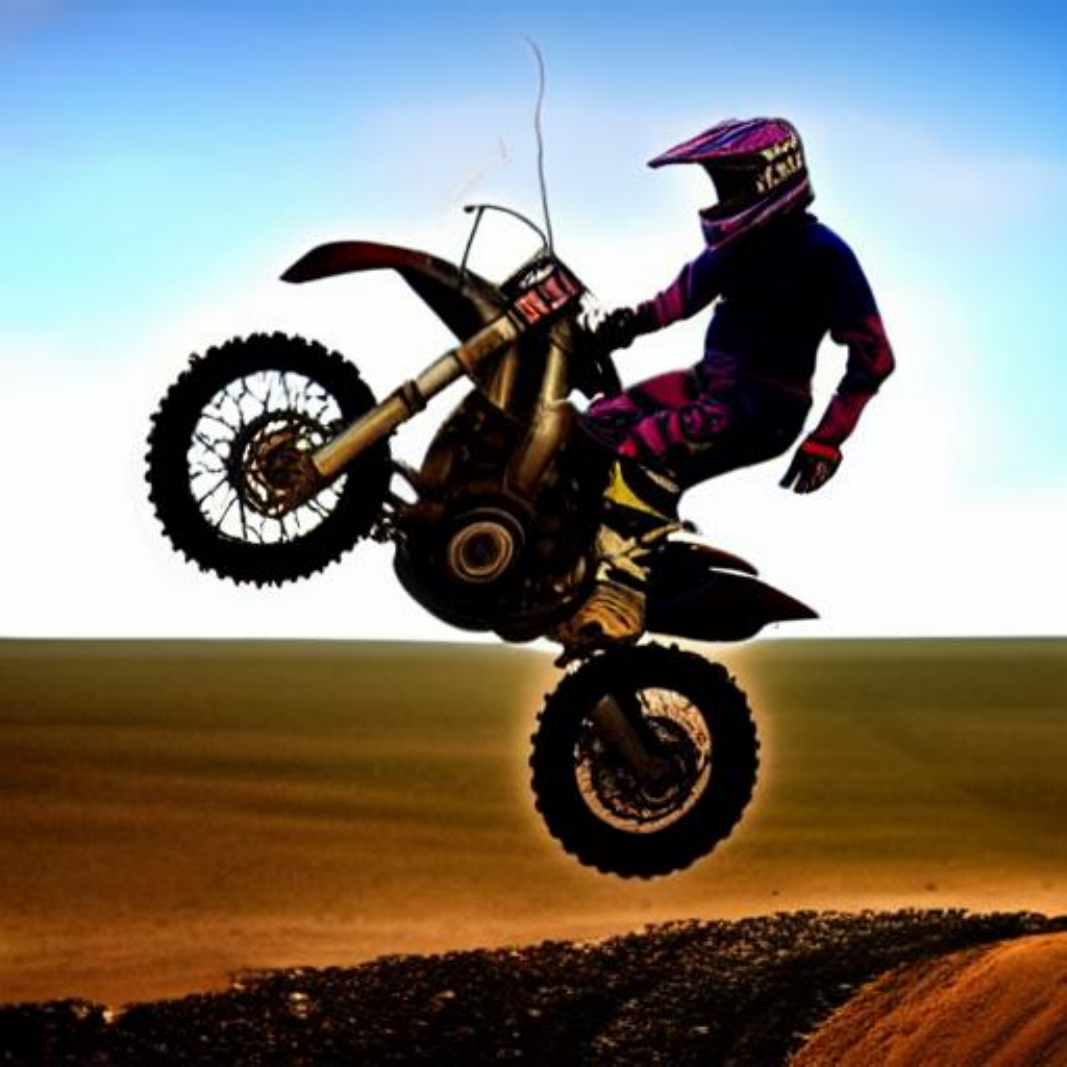}
\\
\includegraphics[width=0.15\columnwidth]{img_dif/SD_40_mix/b_13.pdf}
\includegraphics[width=0.15\columnwidth]{img_dif/SD_40_mix/a_13.pdf}
\includegraphics[width=0.15\columnwidth]{img_dif/SD_40_mix/m_125_13.pdf}
\includegraphics[width=0.15\columnwidth]{img_dif/SD_40_mix/m_155_13.pdf}
\\
\includegraphics[width=0.15\columnwidth]{img_dif/SD_40_mix/b_71.pdf}
\includegraphics[width=0.15\columnwidth]{img_dif/SD_40_mix/a_71.pdf}
\includegraphics[width=0.15\columnwidth]{img_dif/SD_40_mix/m_125_71.pdf}
\includegraphics[width=0.15\columnwidth]{img_dif/SD_40_mix/m_155_71.pdf}
\\
\includegraphics[width=0.15\columnwidth]{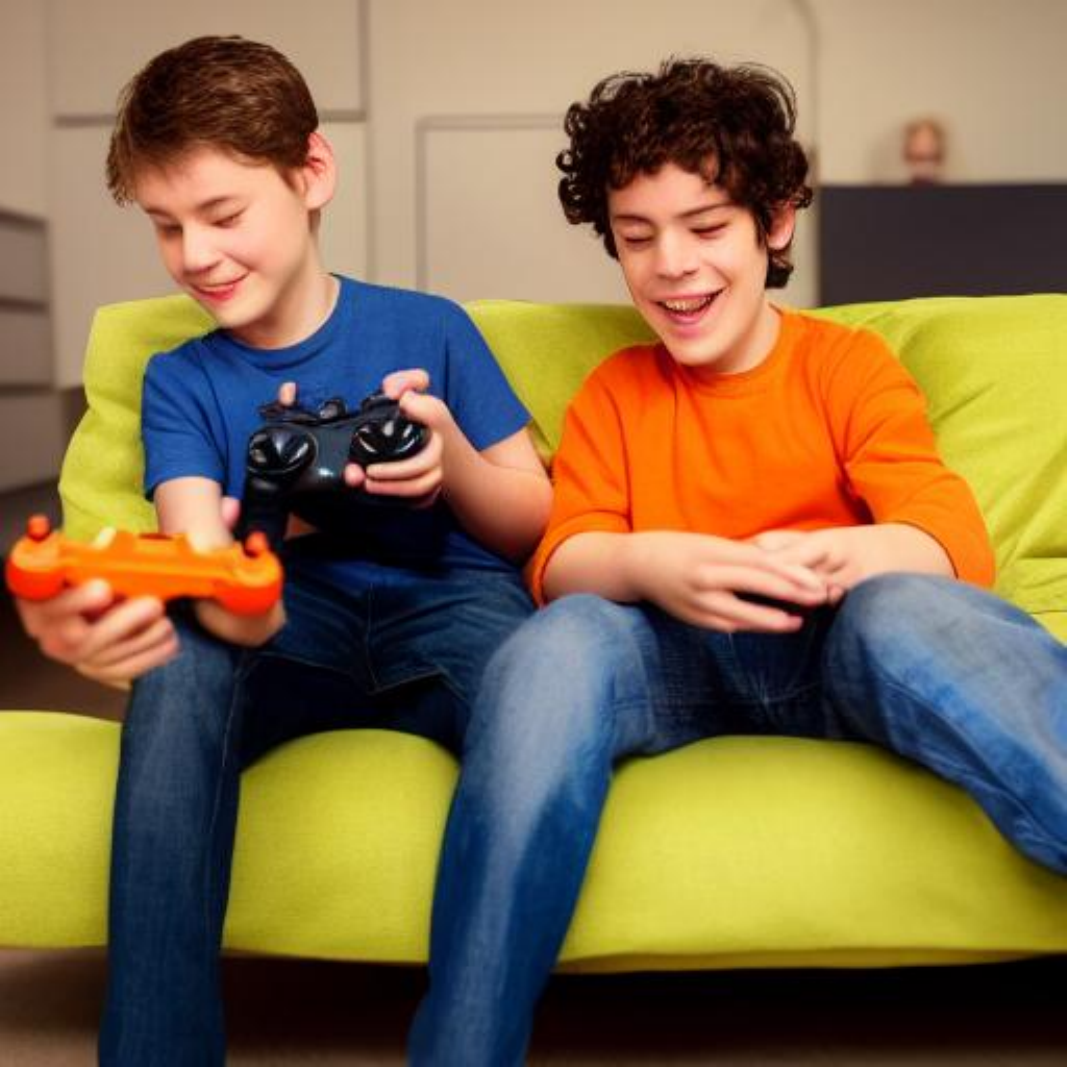}
\includegraphics[width=0.15\columnwidth]{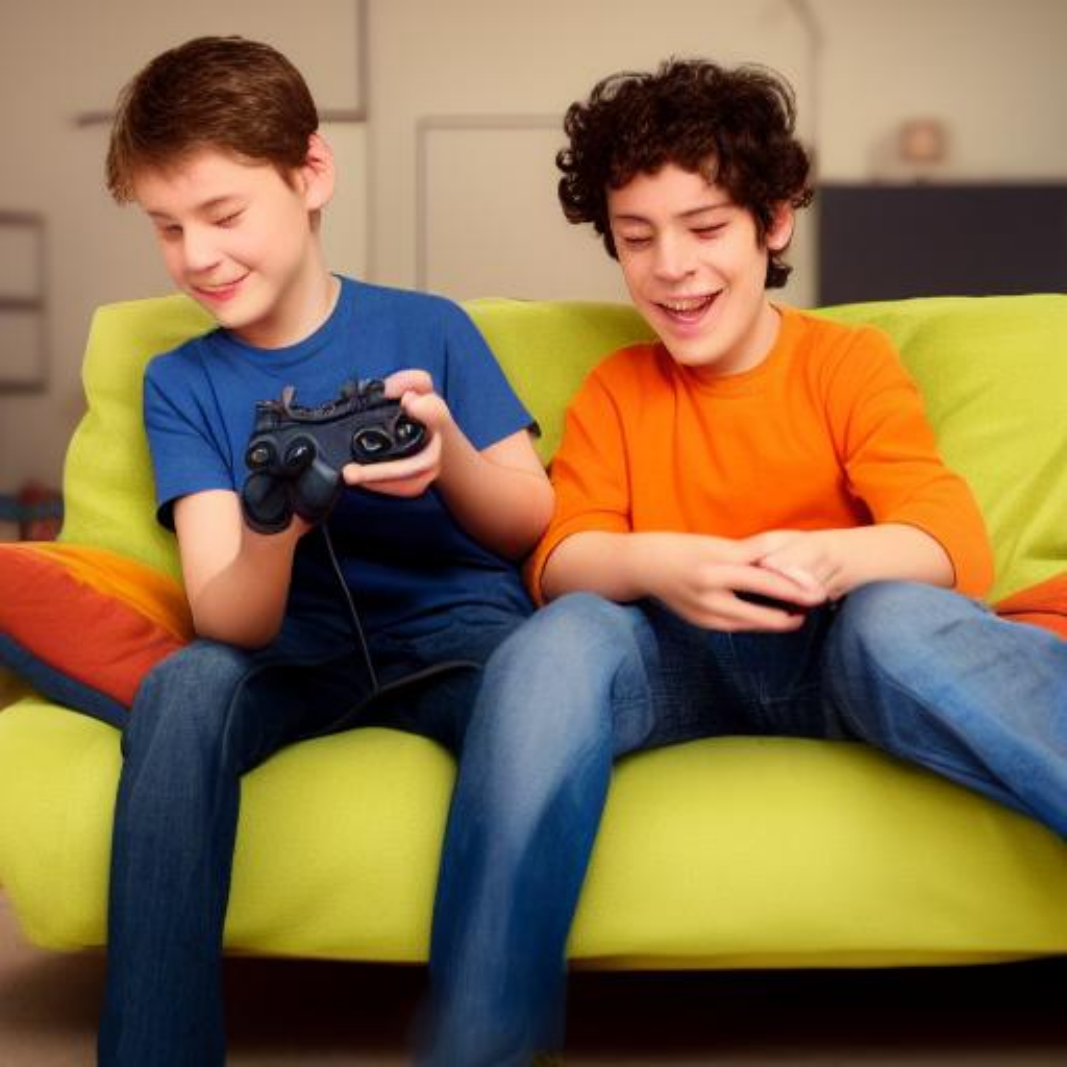}
\includegraphics[width=0.15\columnwidth]{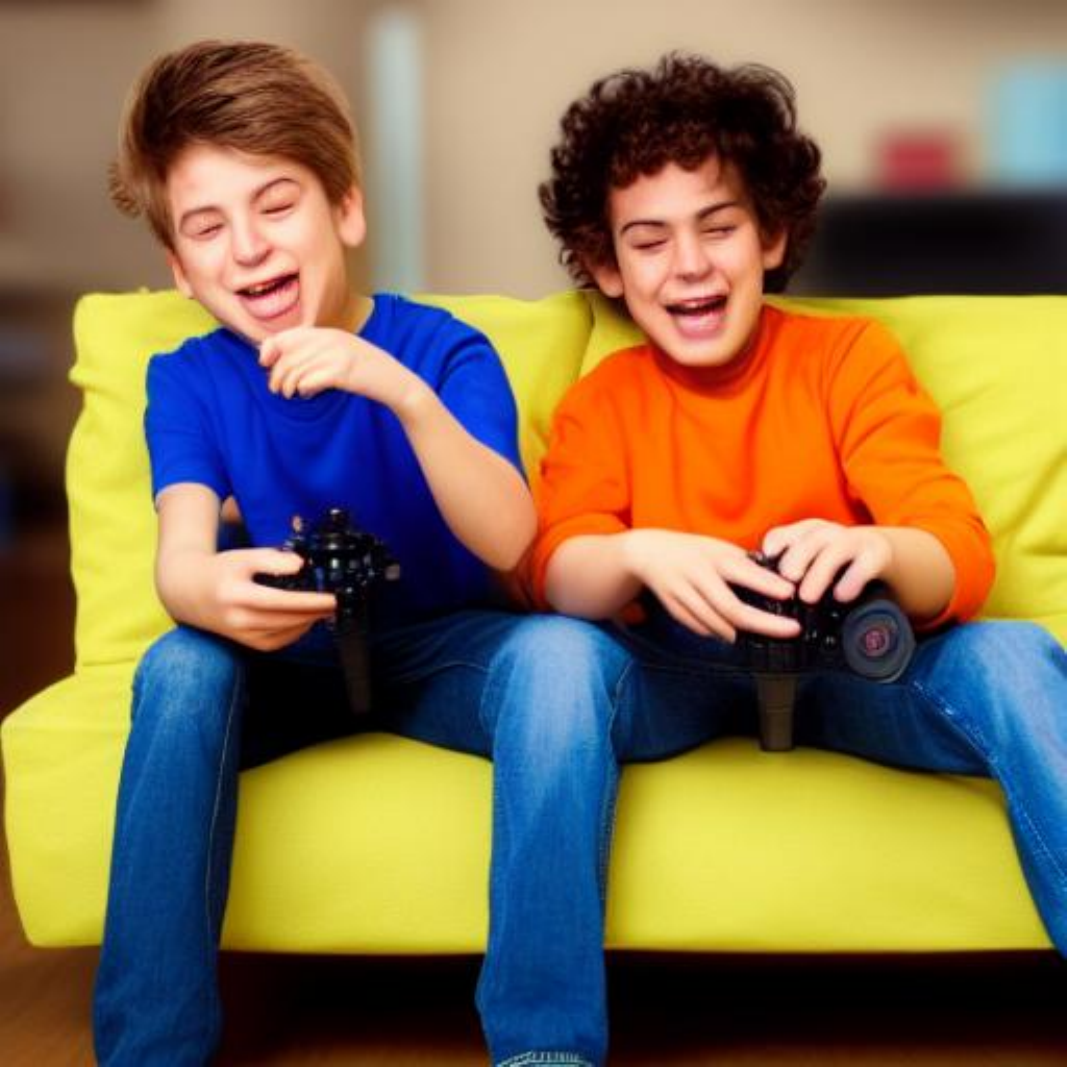}
\includegraphics[width=0.15\columnwidth]{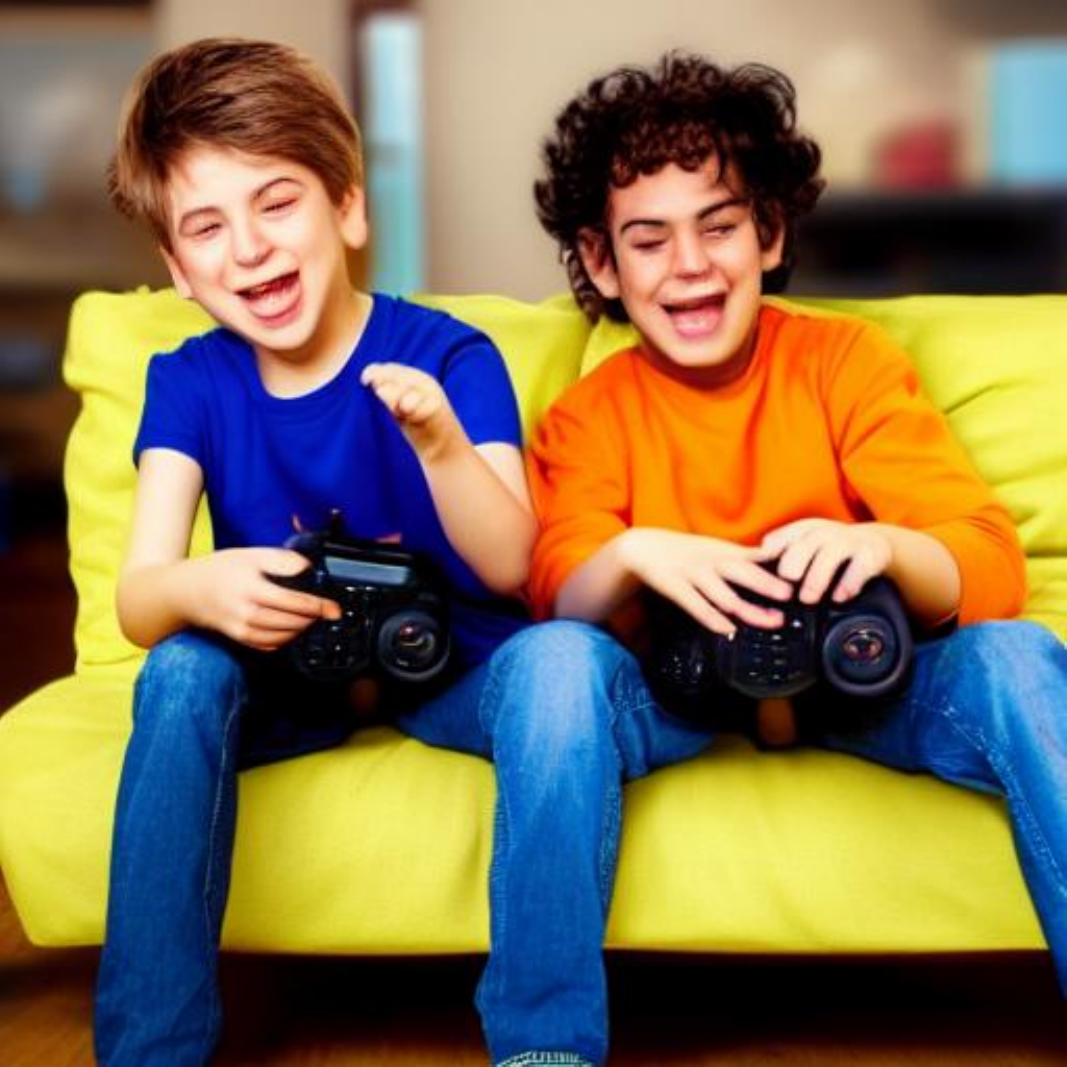}
\\
\includegraphics[width=0.15\columnwidth]{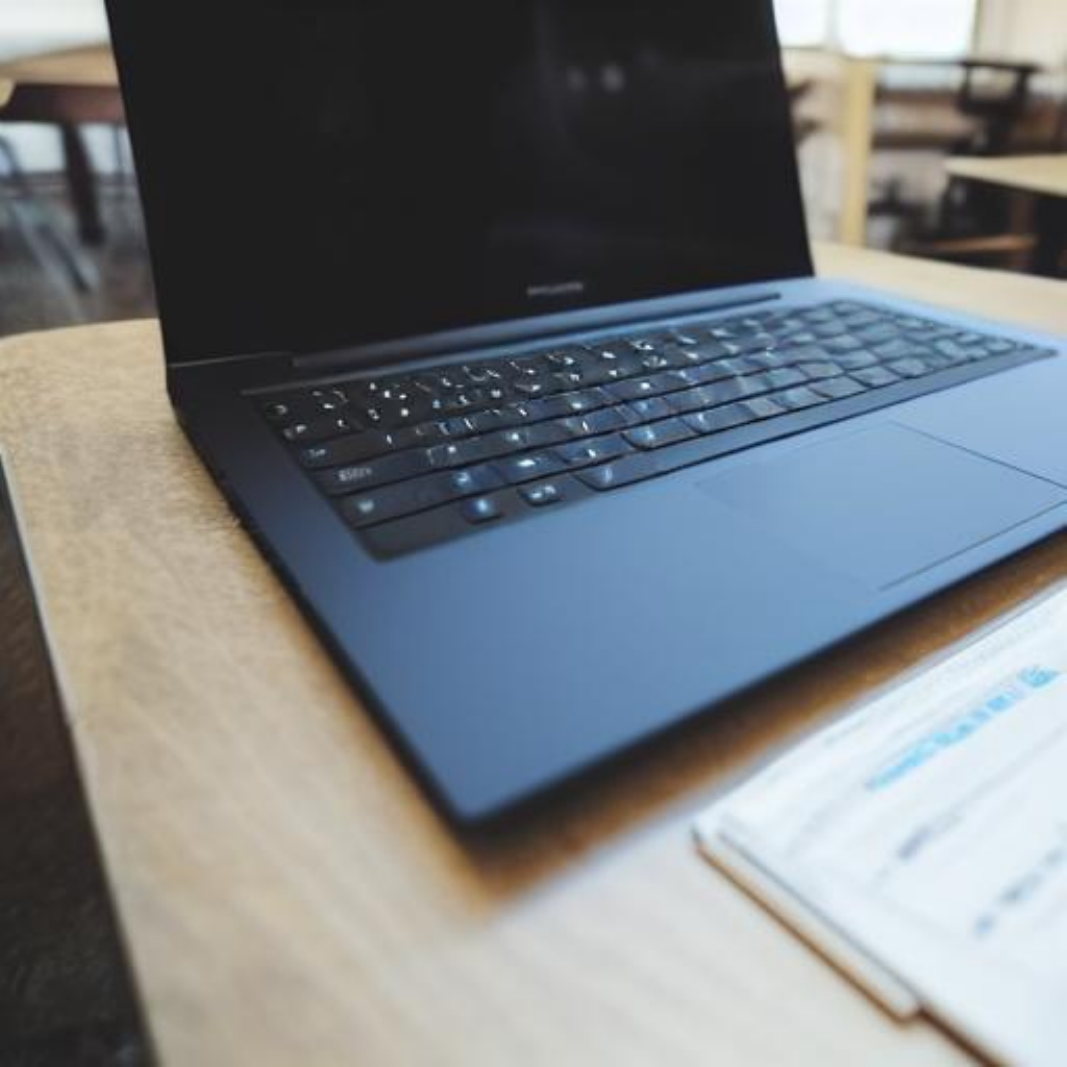}
\includegraphics[width=0.15\columnwidth]{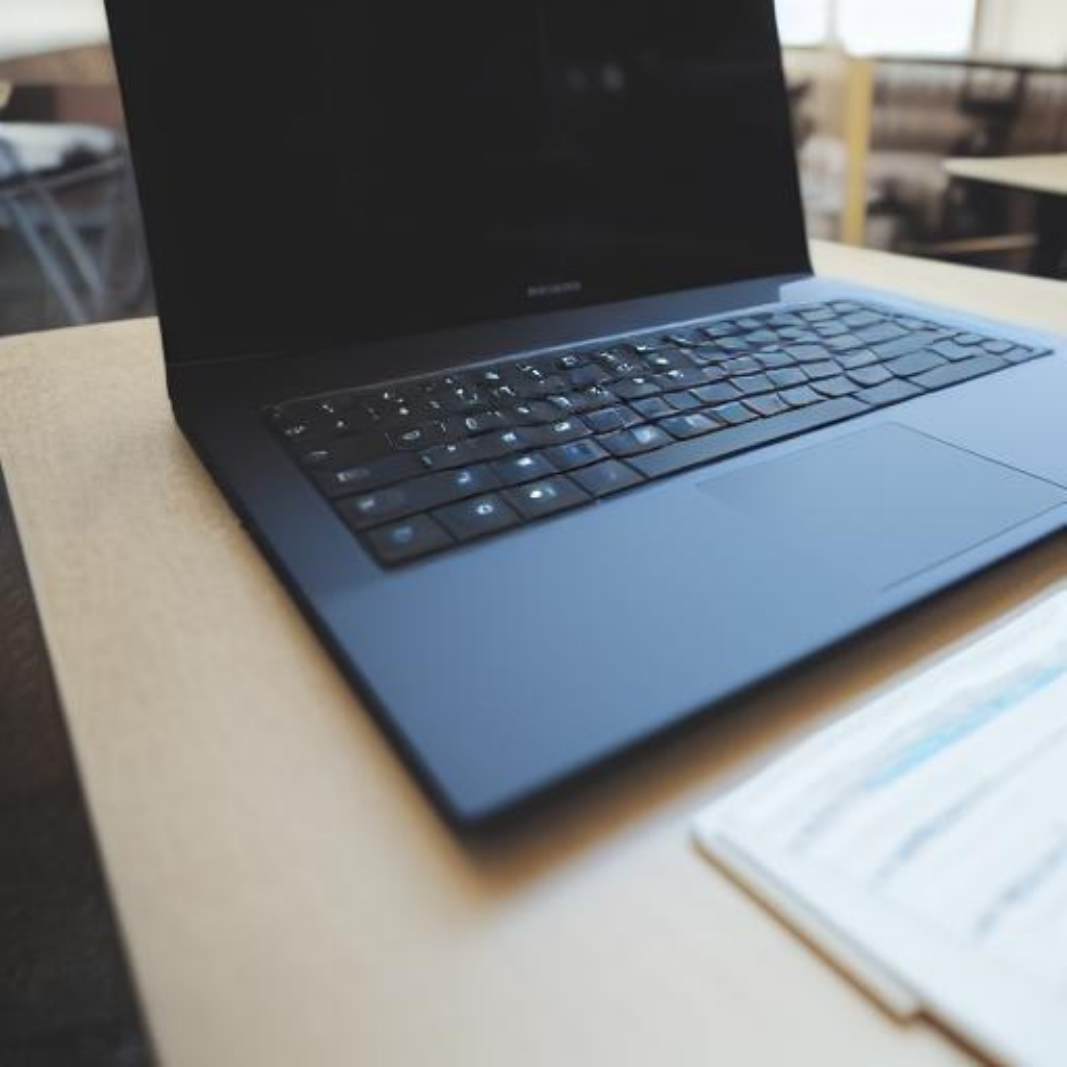}
\includegraphics[width=0.15\columnwidth]{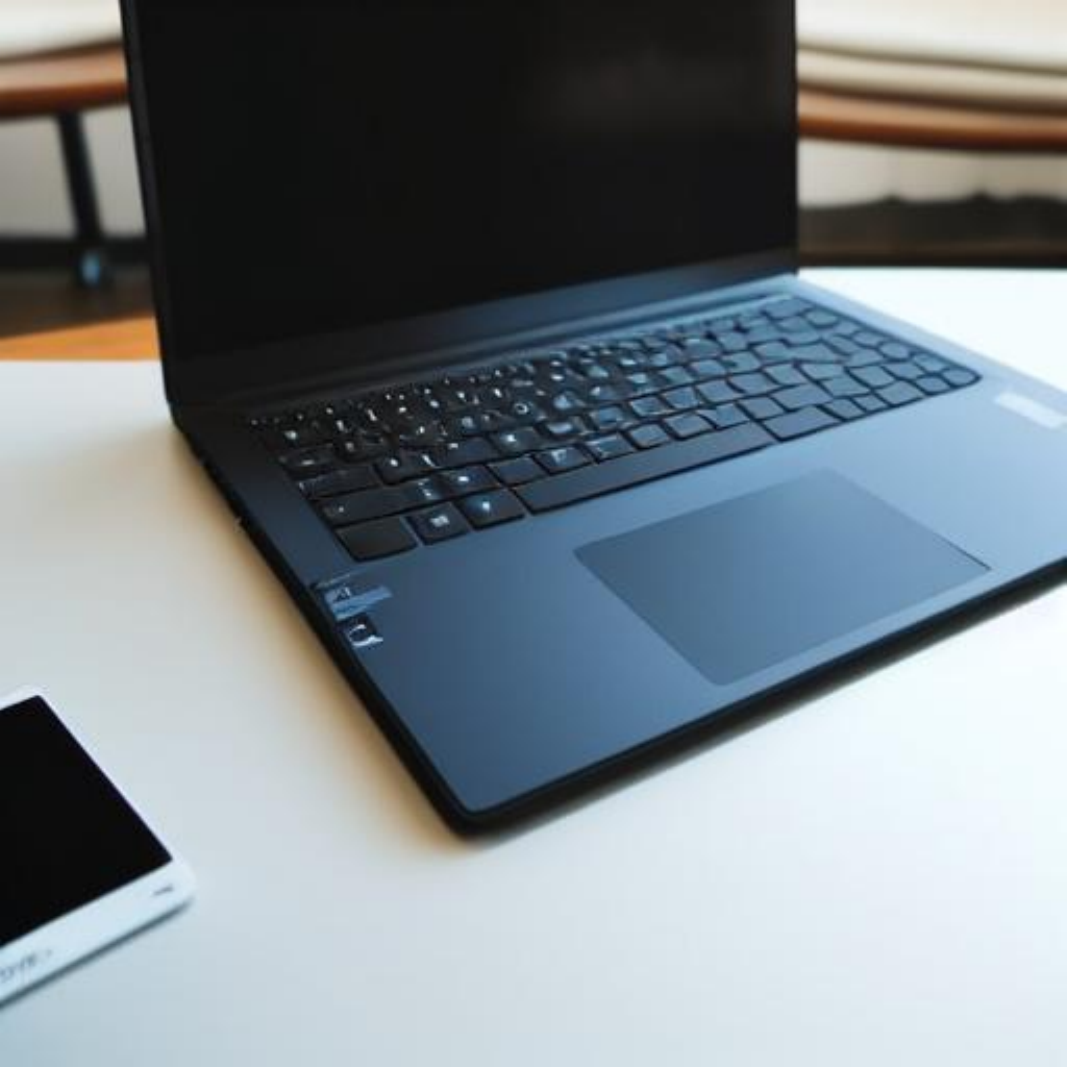}
\includegraphics[width=0.15\columnwidth]{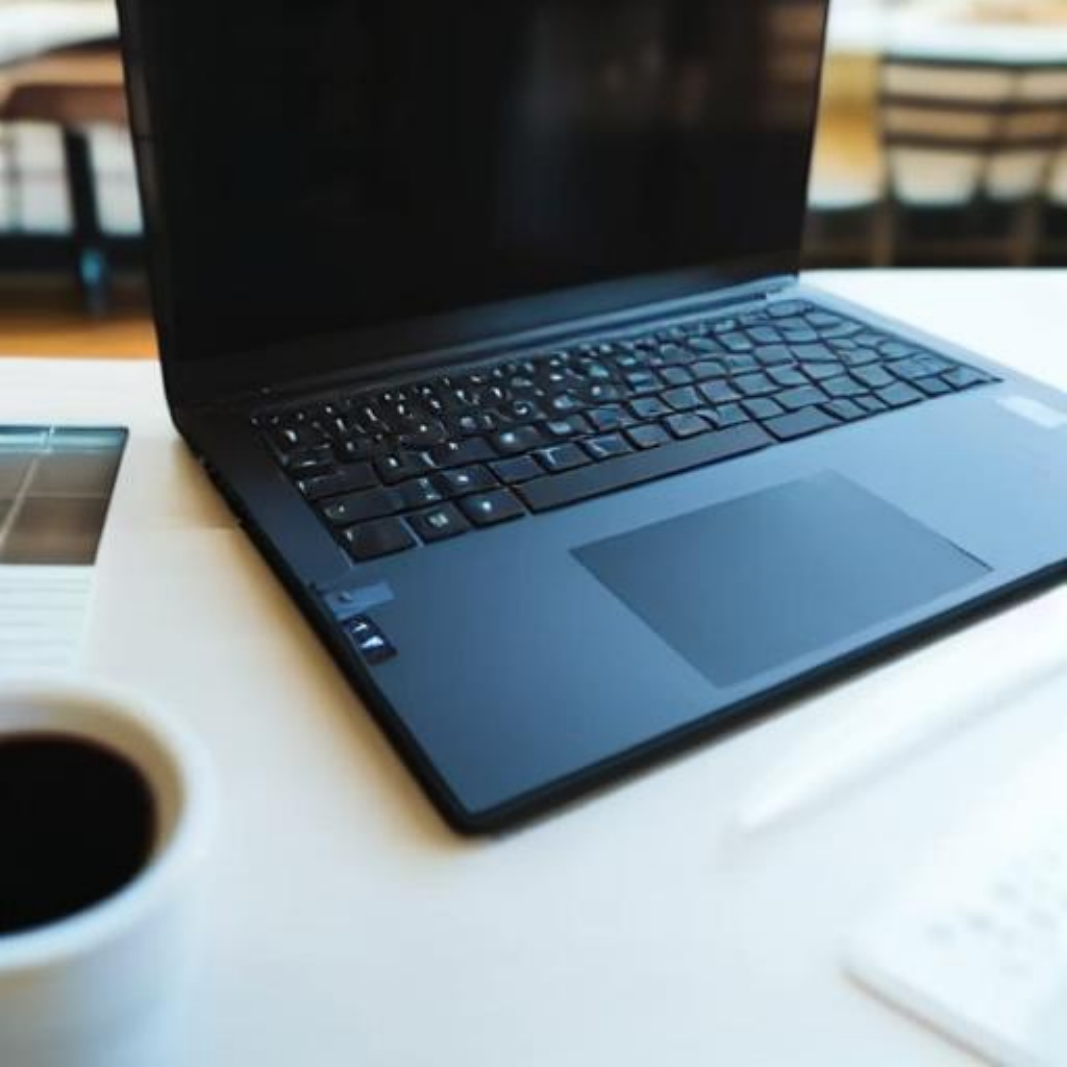}
\caption{\textbf{SD with $40$ Steps.} For each row, from left to right: baseline, ablation, generated image with variance scaling $1.25$, and one with $1.55$. 
%
%
}
    \label{fig:sd-40}
    \end{figure}

\begin{figure}[t]
    \centering
\includegraphics[width=0.15\columnwidth]{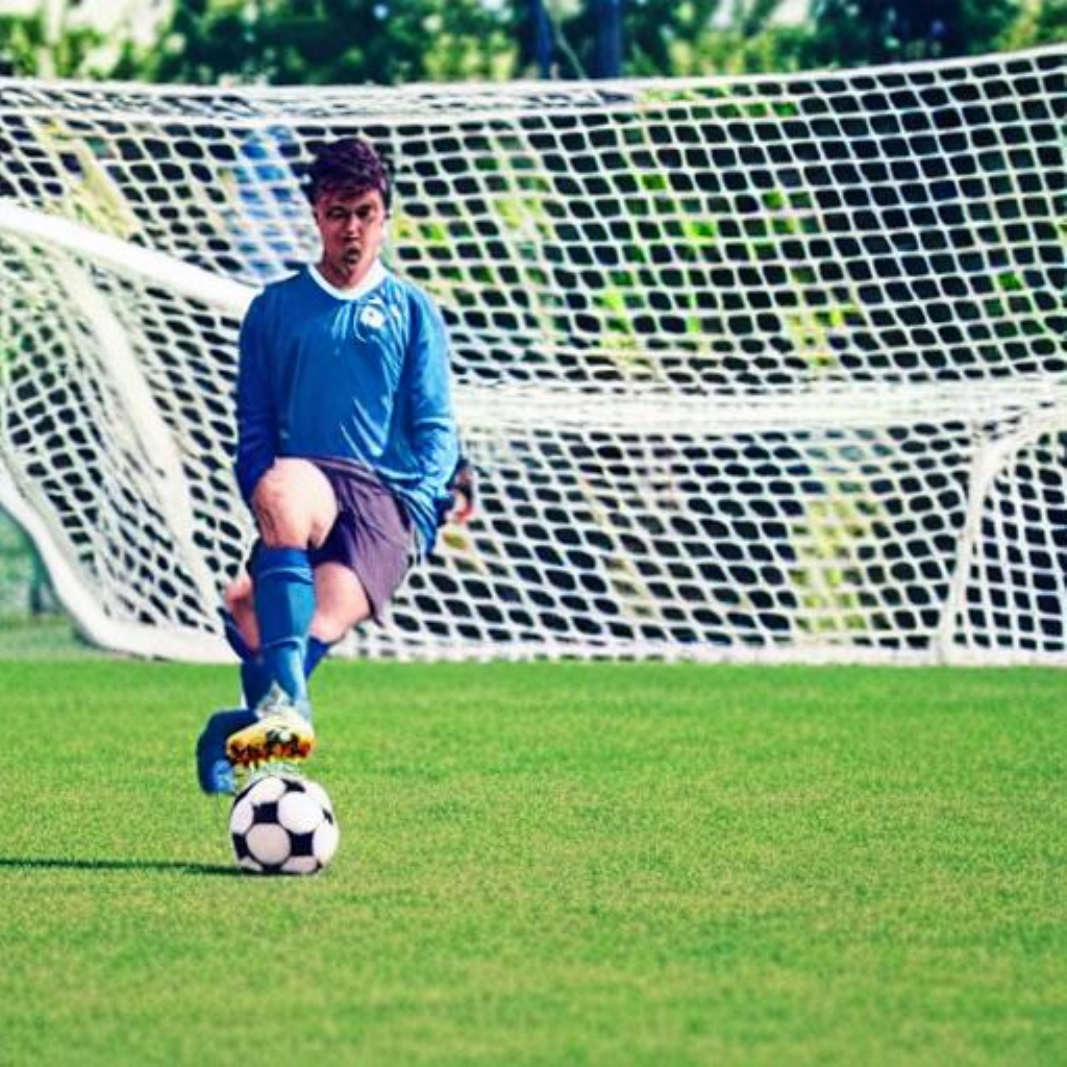}
\includegraphics[width=0.15\columnwidth]{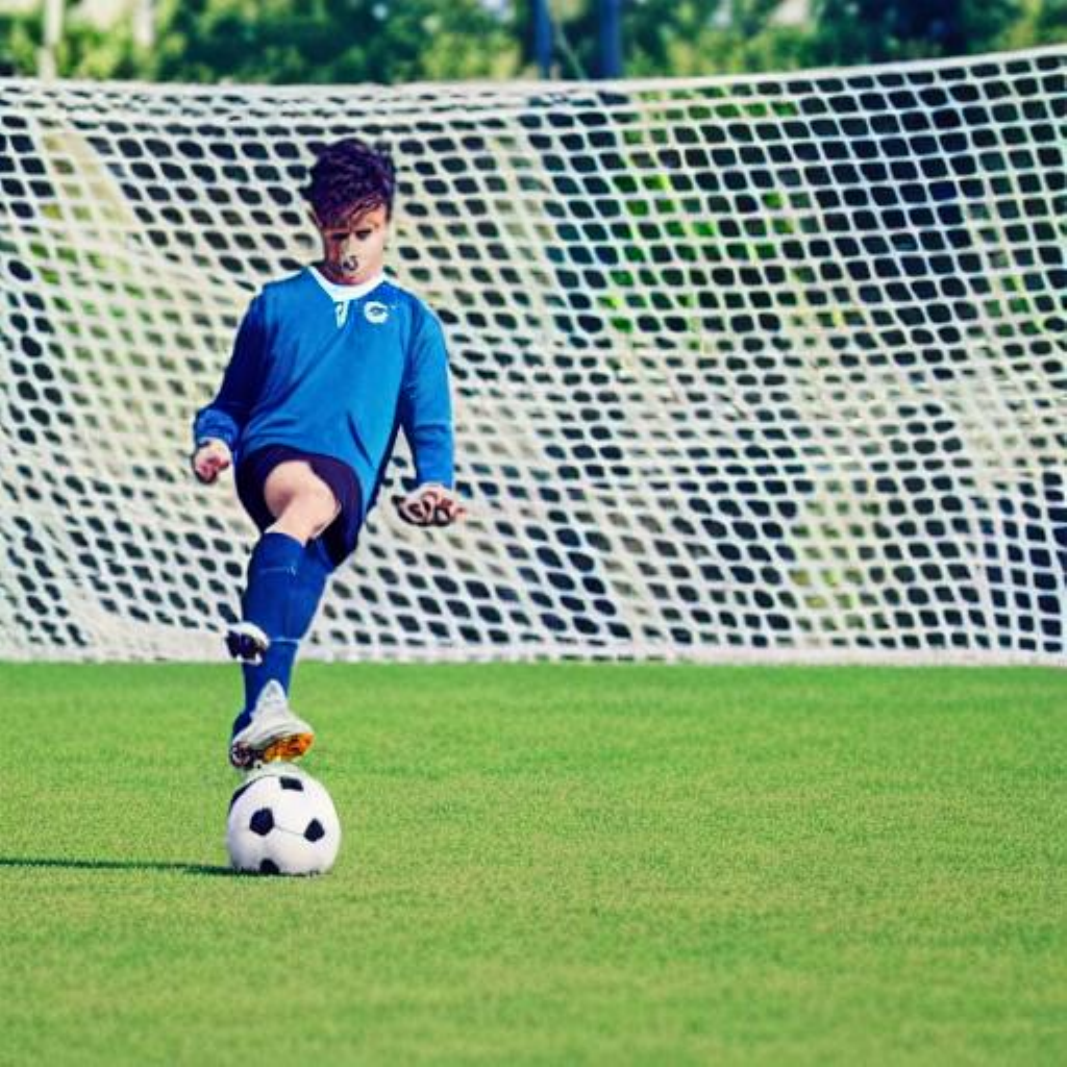}
\includegraphics[width=0.15\columnwidth]{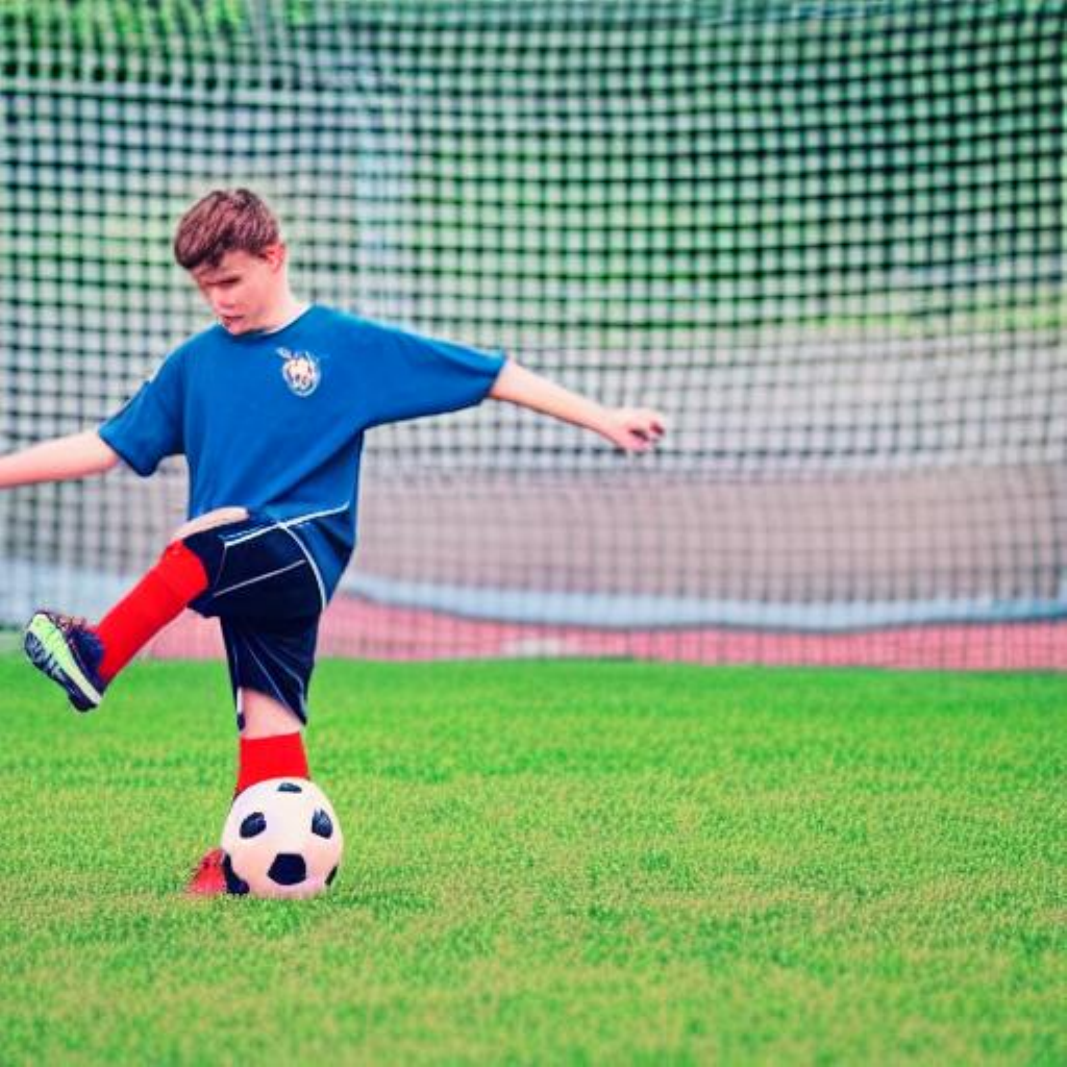}
\includegraphics[width=0.15\columnwidth]{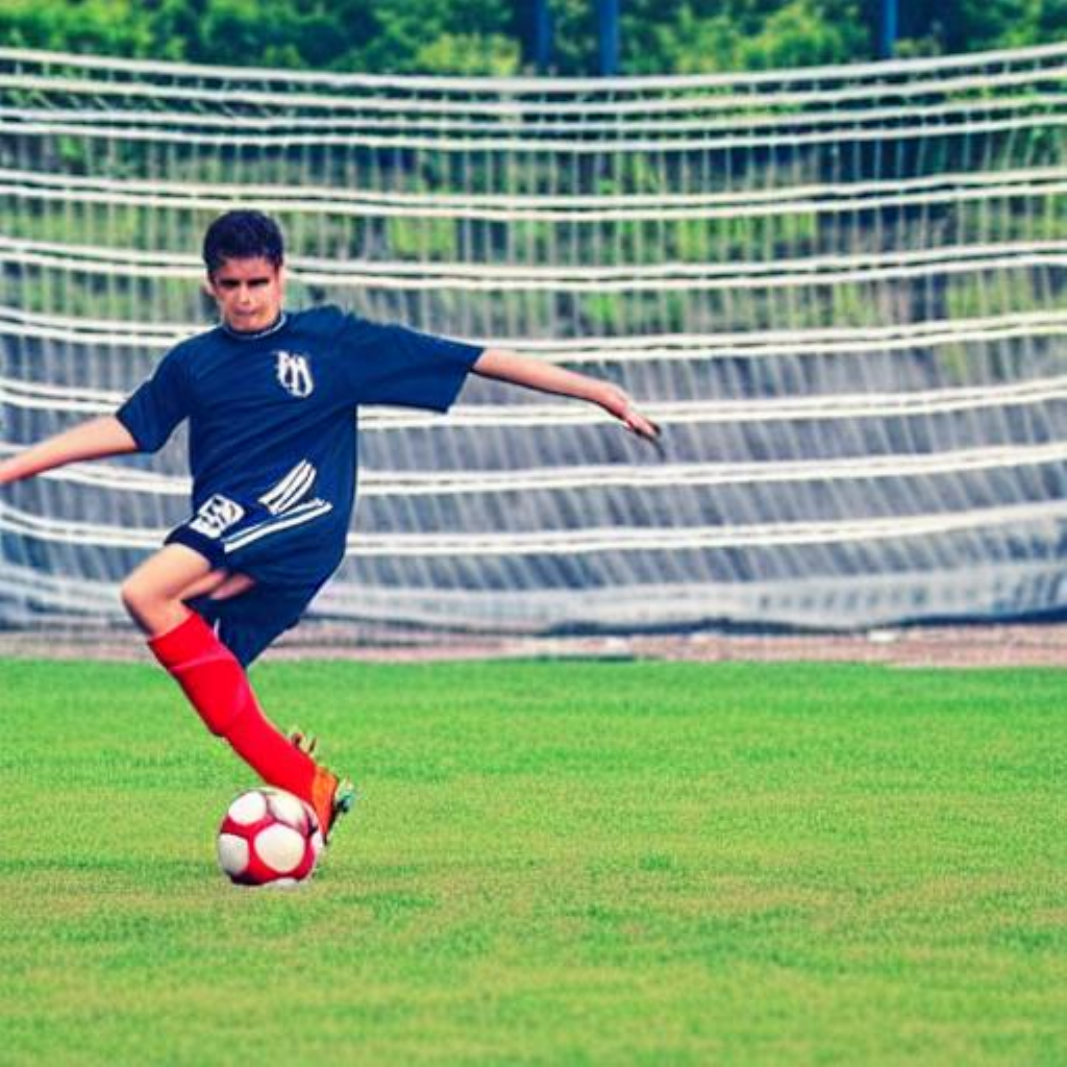}
\\
\includegraphics[width=0.15\columnwidth]{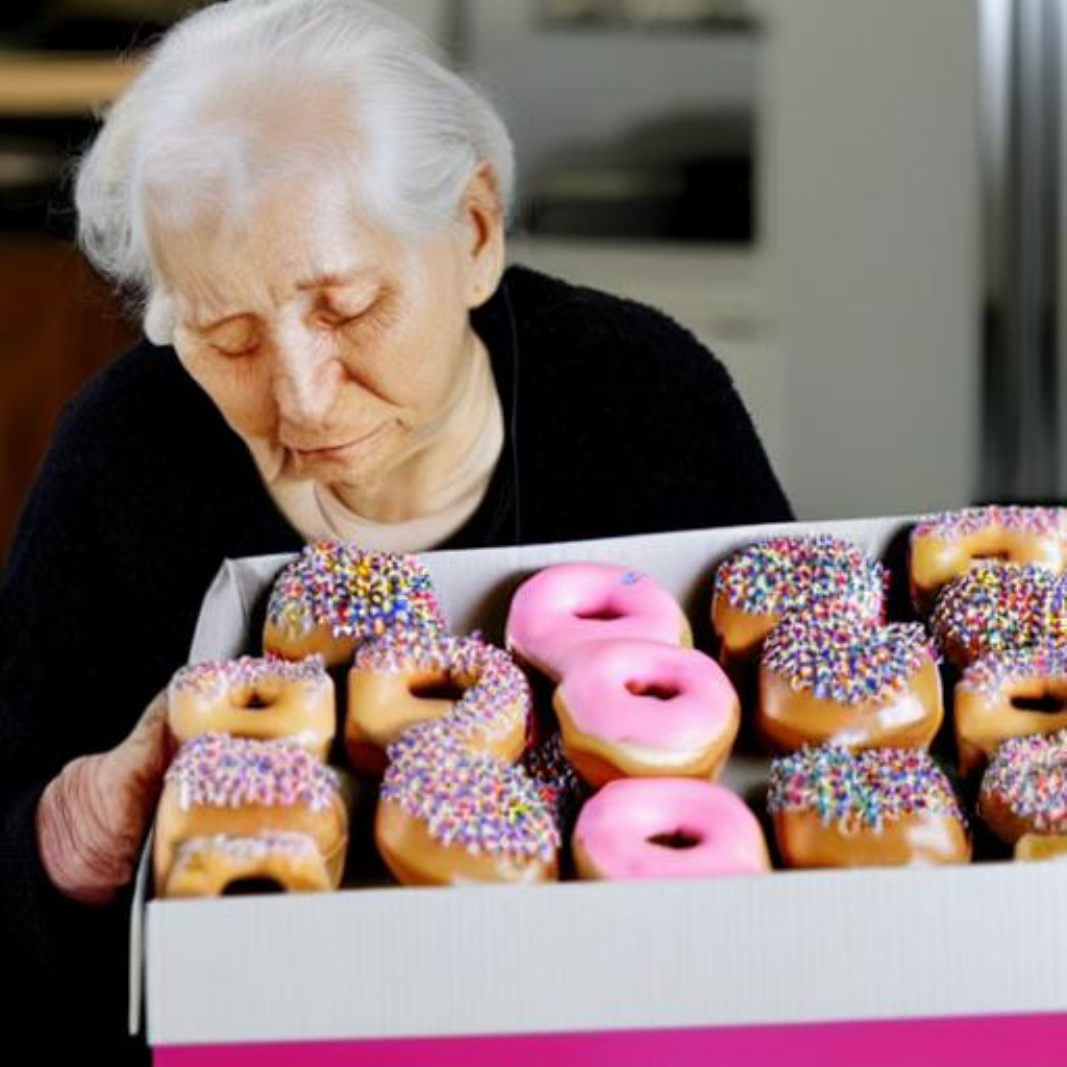}
\includegraphics[width=0.15\columnwidth]{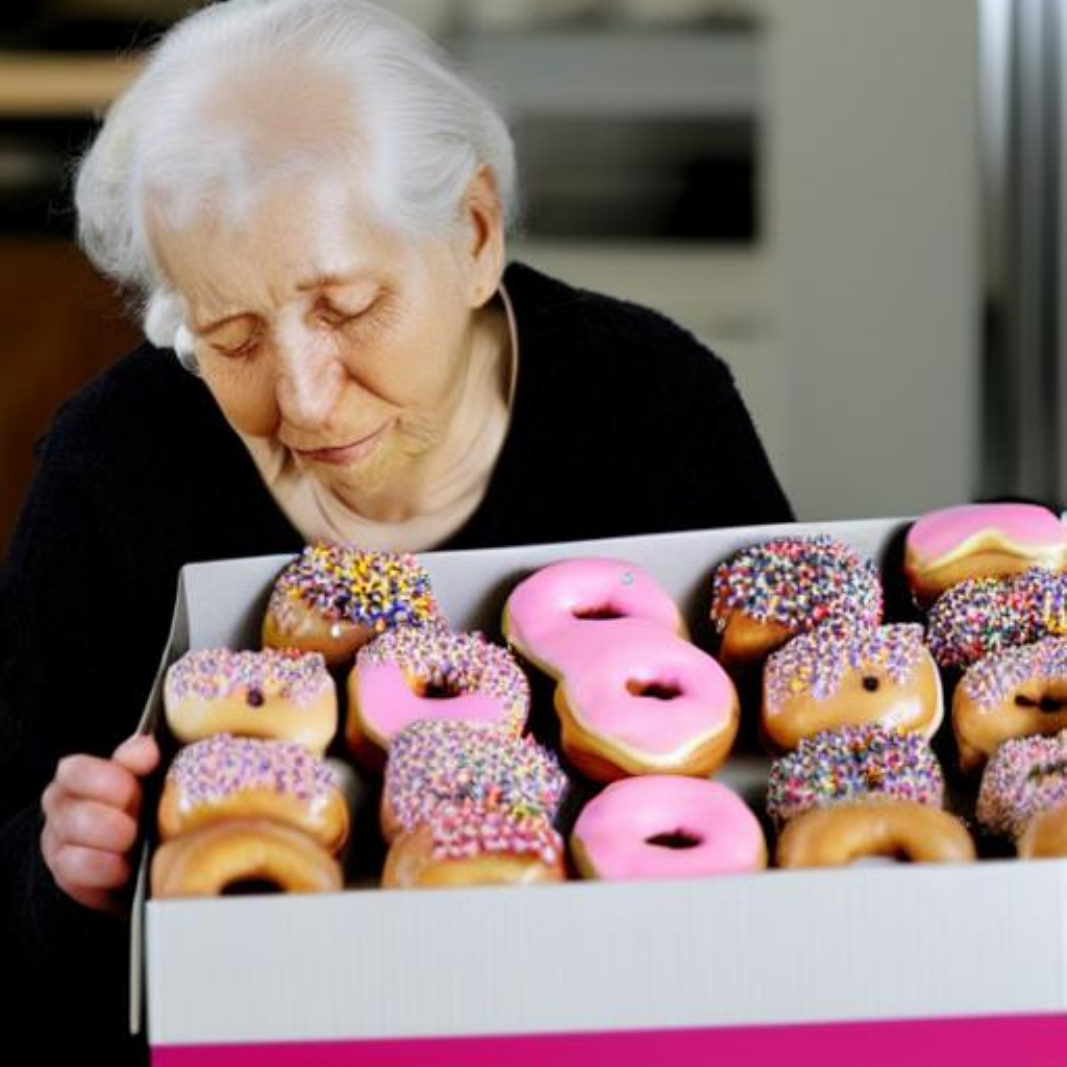}
\includegraphics[width=0.15\columnwidth]{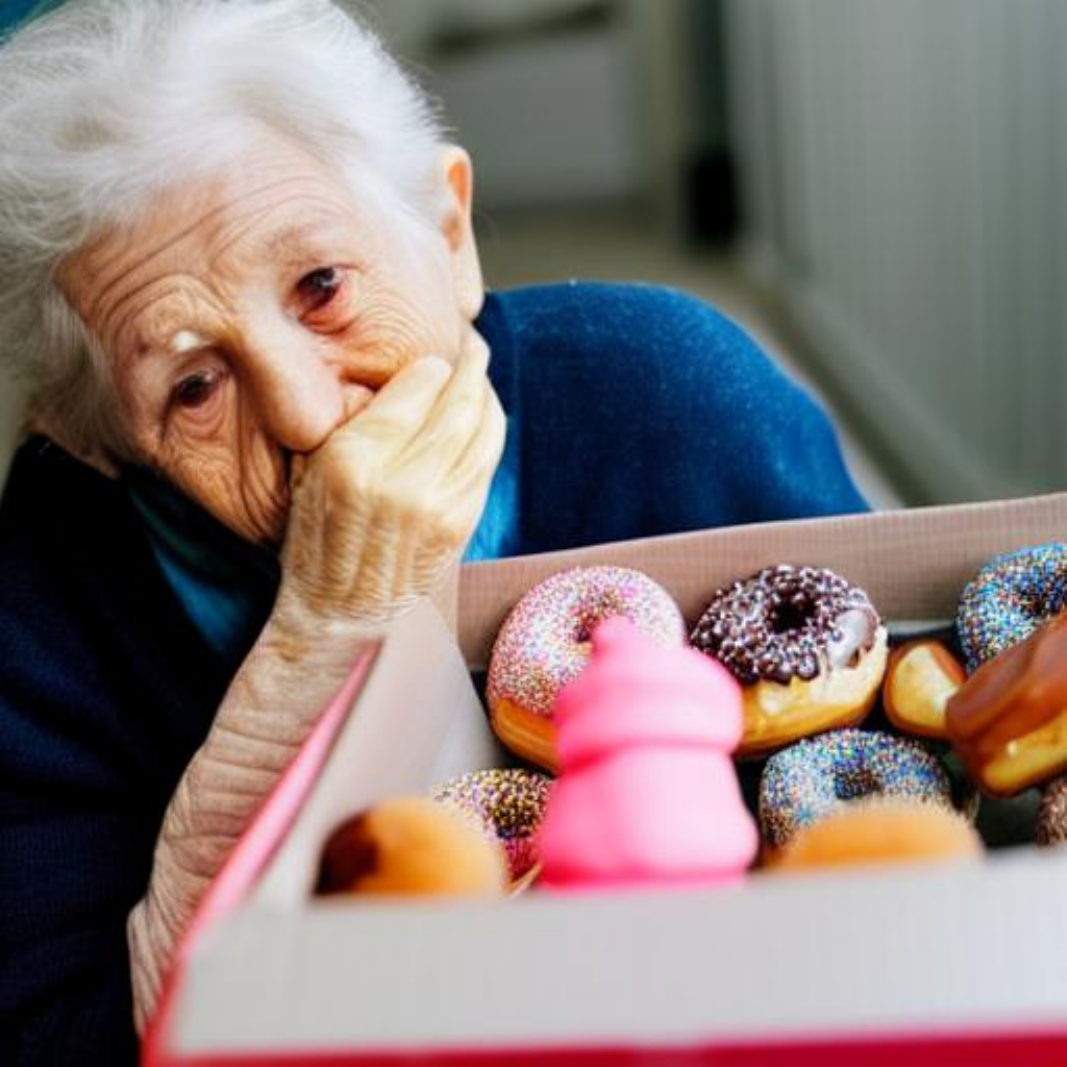}
\includegraphics[width=0.15\columnwidth]{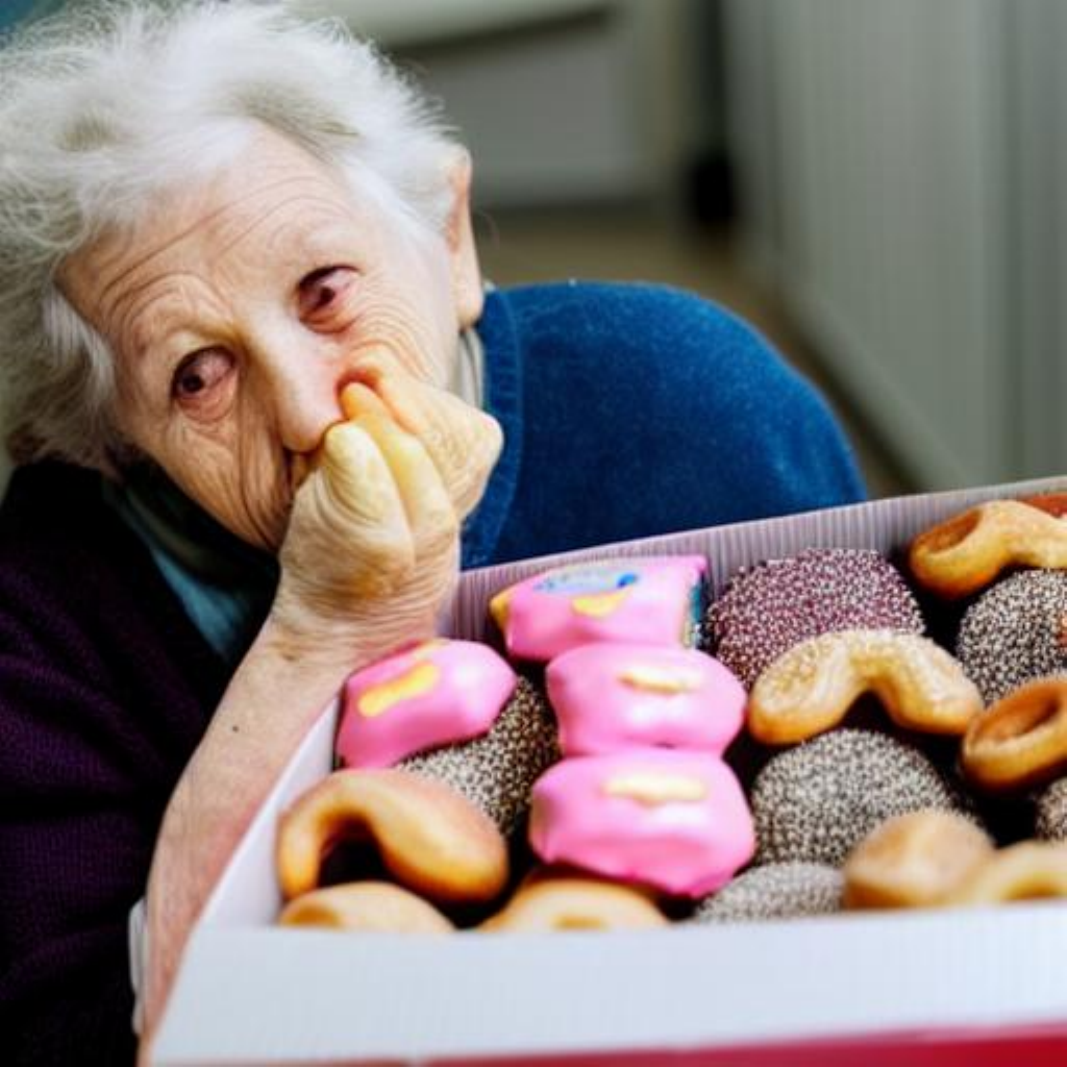}
\\
\includegraphics[width=0.15\columnwidth]{img_dif/SD_80_mix/b_5.pdf}
\includegraphics[width=0.15\columnwidth]{img_dif/SD_80_mix/a_5.pdf}
\includegraphics[width=0.15\columnwidth]{img_dif/SD_80_mix/m_110_5.pdf}
\includegraphics[width=0.15\columnwidth]{img_dif/SD_80_mix/m_120_5.pdf}
\\
\includegraphics[width=0.15\columnwidth]{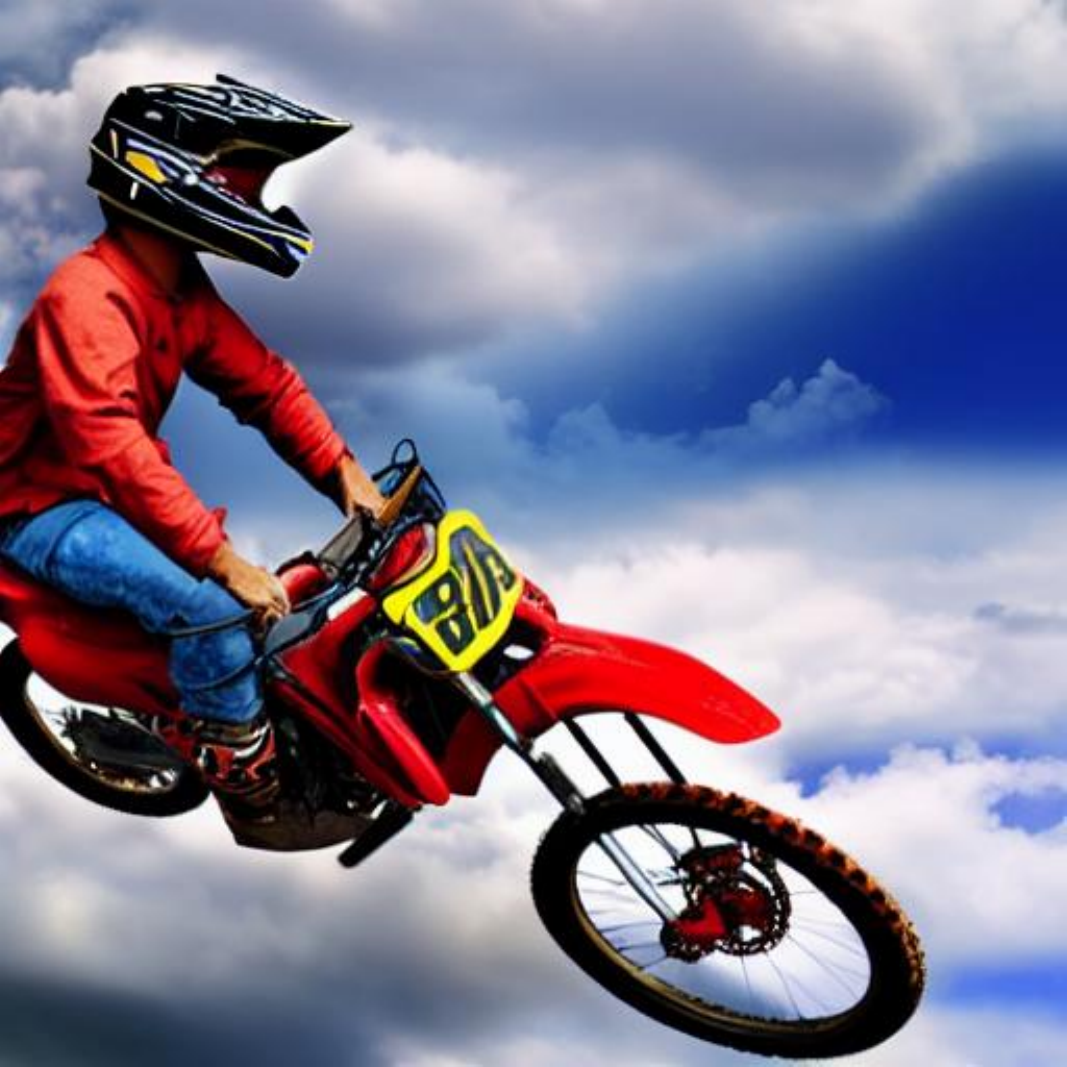}
\includegraphics[width=0.15\columnwidth]{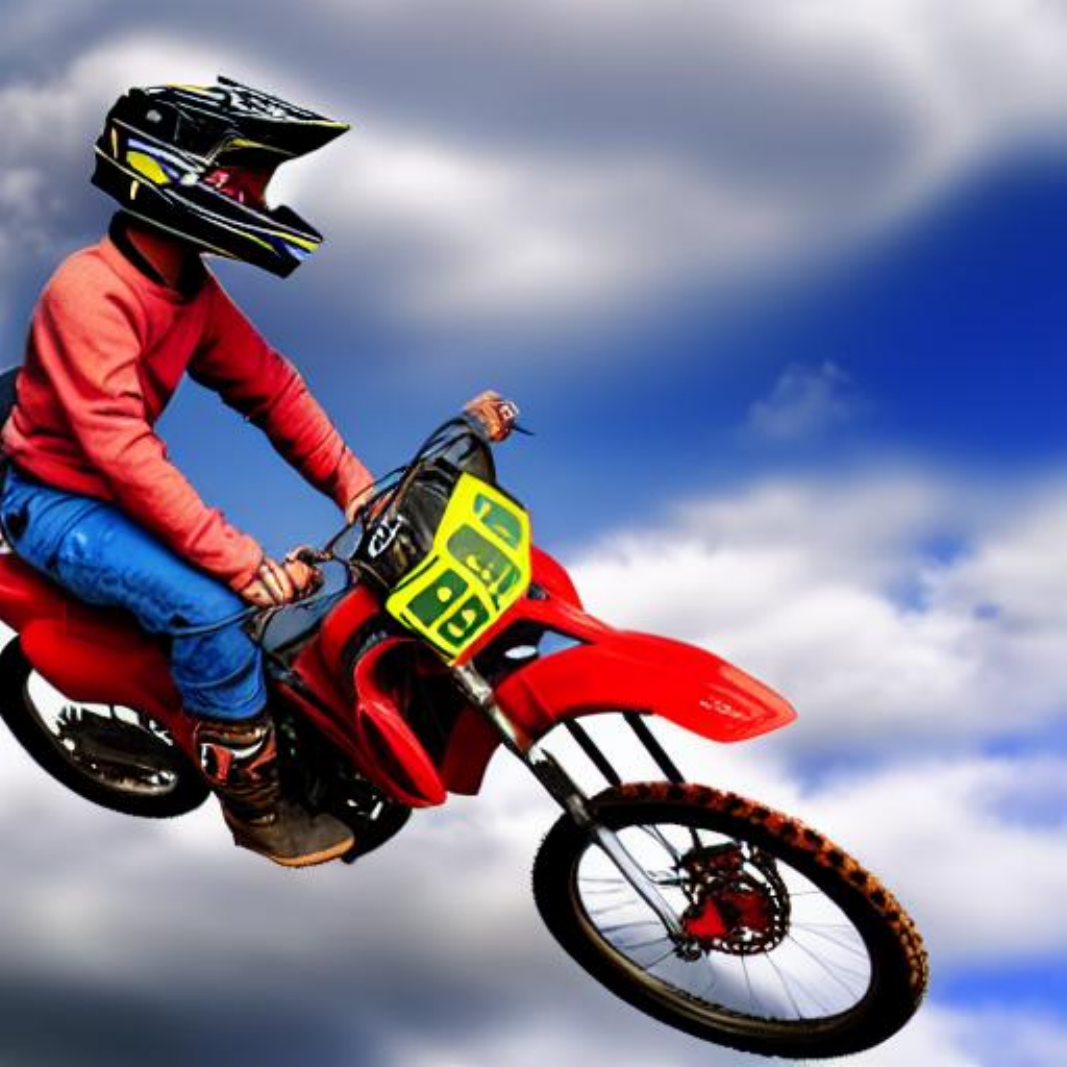}
\includegraphics[width=0.15\columnwidth]{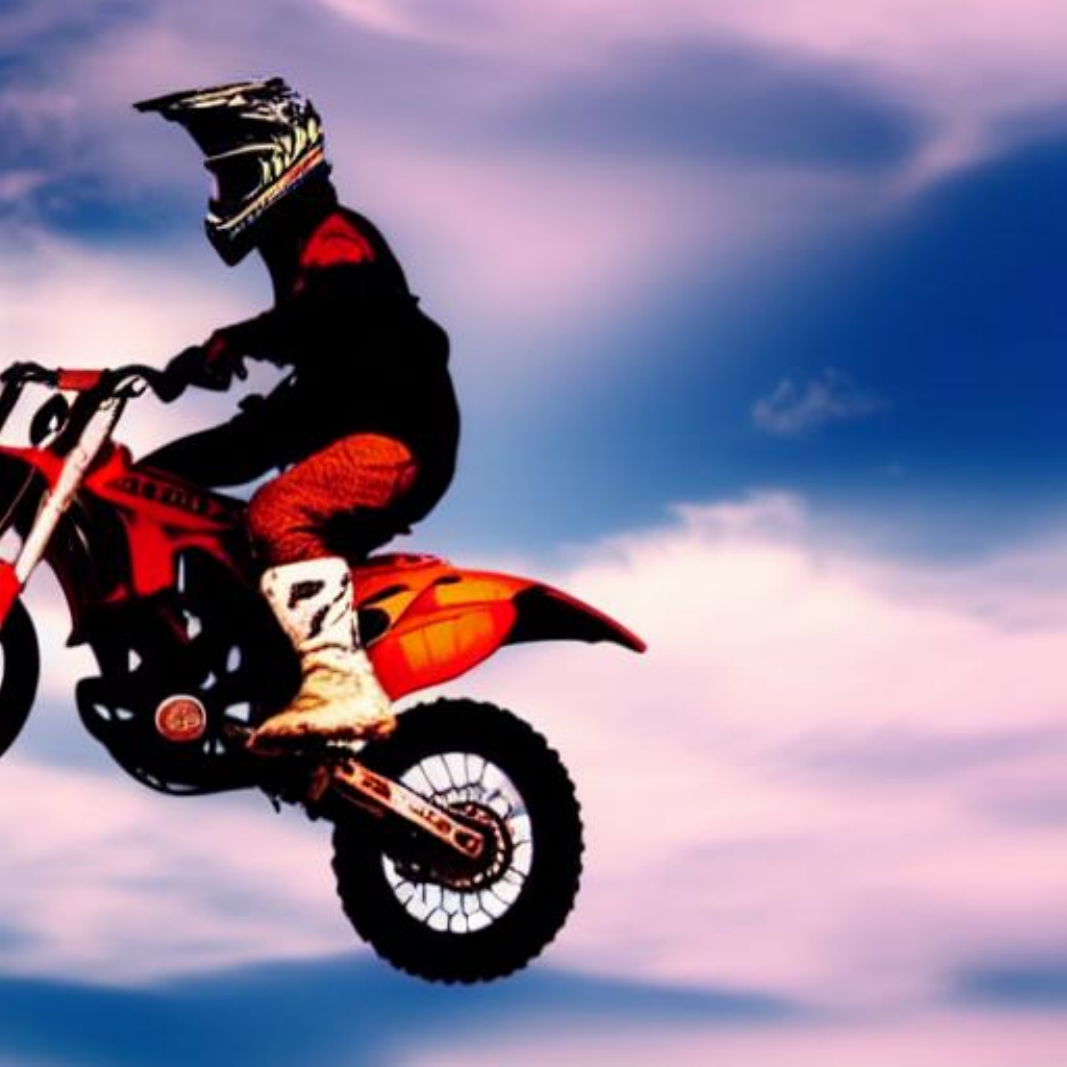}
\includegraphics[width=0.15\columnwidth]{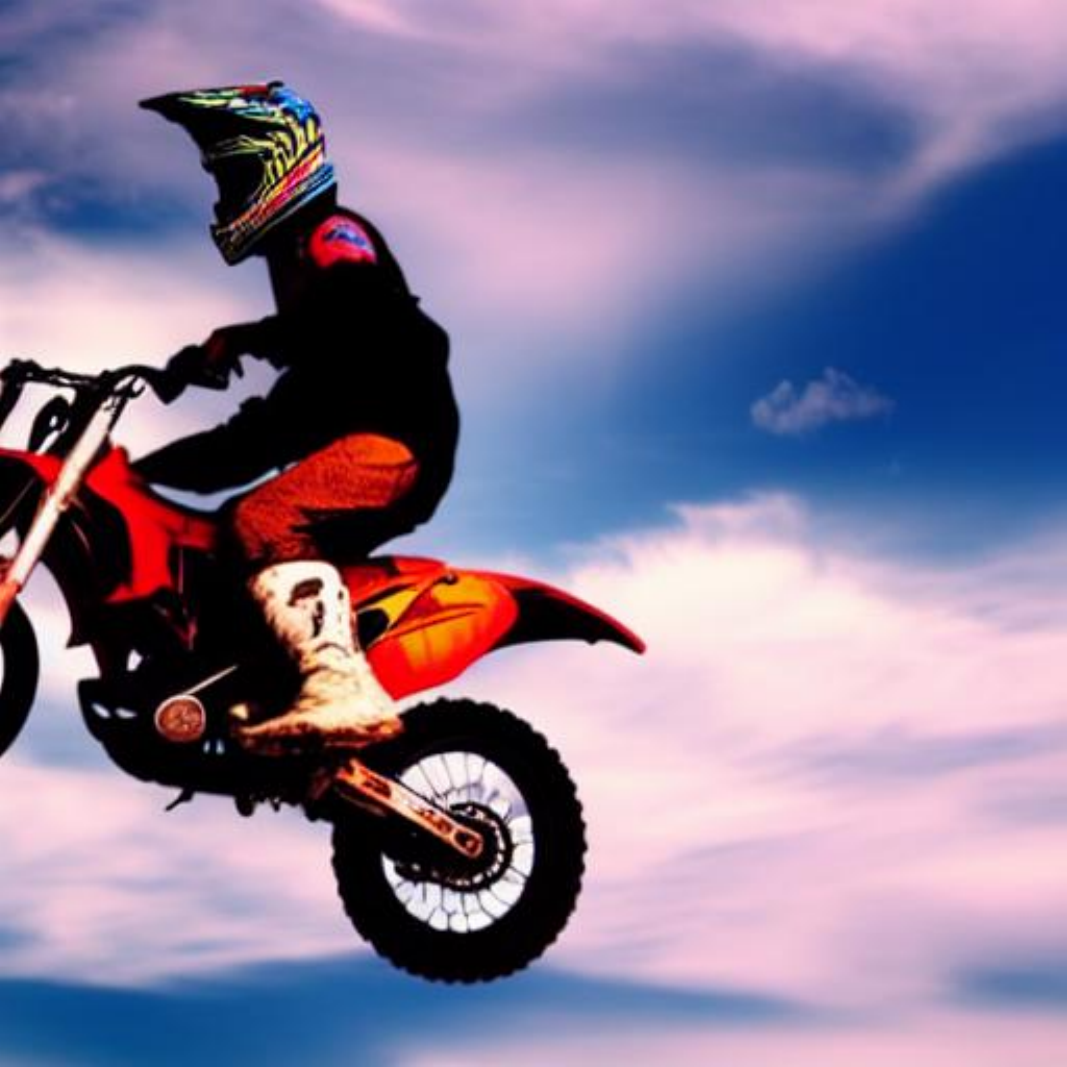}
\\
\includegraphics[width=0.15\columnwidth]{img_dif/SD_80_mix/b_13.pdf}
\includegraphics[width=0.15\columnwidth]{img_dif/SD_80_mix/a_13.pdf}
\includegraphics[width=0.15\columnwidth]{img_dif/SD_80_mix/m_110_13.pdf}
\includegraphics[width=0.15\columnwidth]{img_dif/SD_80_mix/m_120_13.pdf}
\\
\includegraphics[width=0.15\columnwidth]{img_dif/SD_80_mix/b_71.pdf}
\includegraphics[width=0.15\columnwidth]{img_dif/SD_80_mix/a_71.pdf}
\includegraphics[width=0.15\columnwidth]{img_dif/SD_80_mix/m_110_71.pdf}
\includegraphics[width=0.15\columnwidth]{img_dif/SD_80_mix/m_120_71.pdf}
\\
\includegraphics[width=0.15\columnwidth]{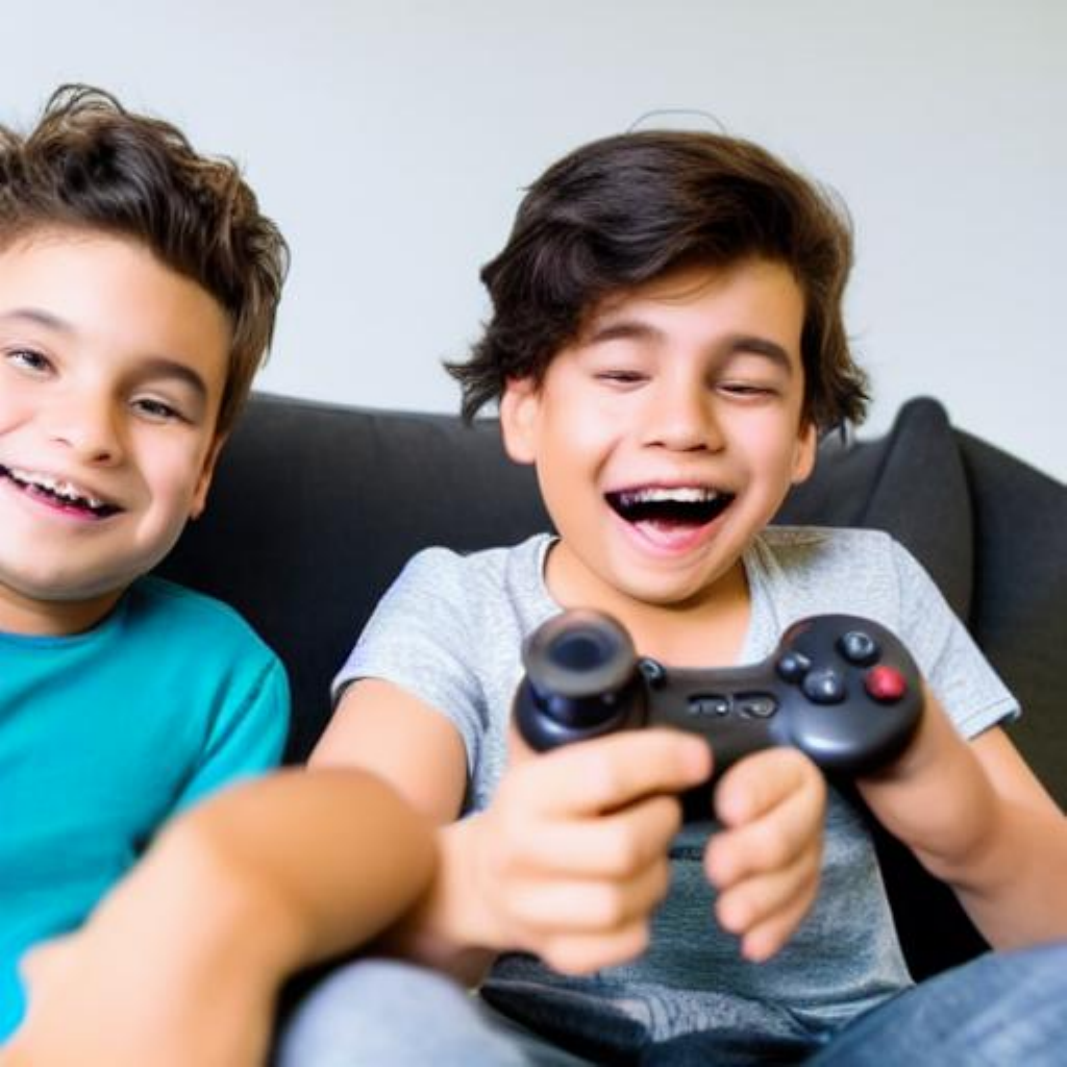}
\includegraphics[width=0.15\columnwidth]{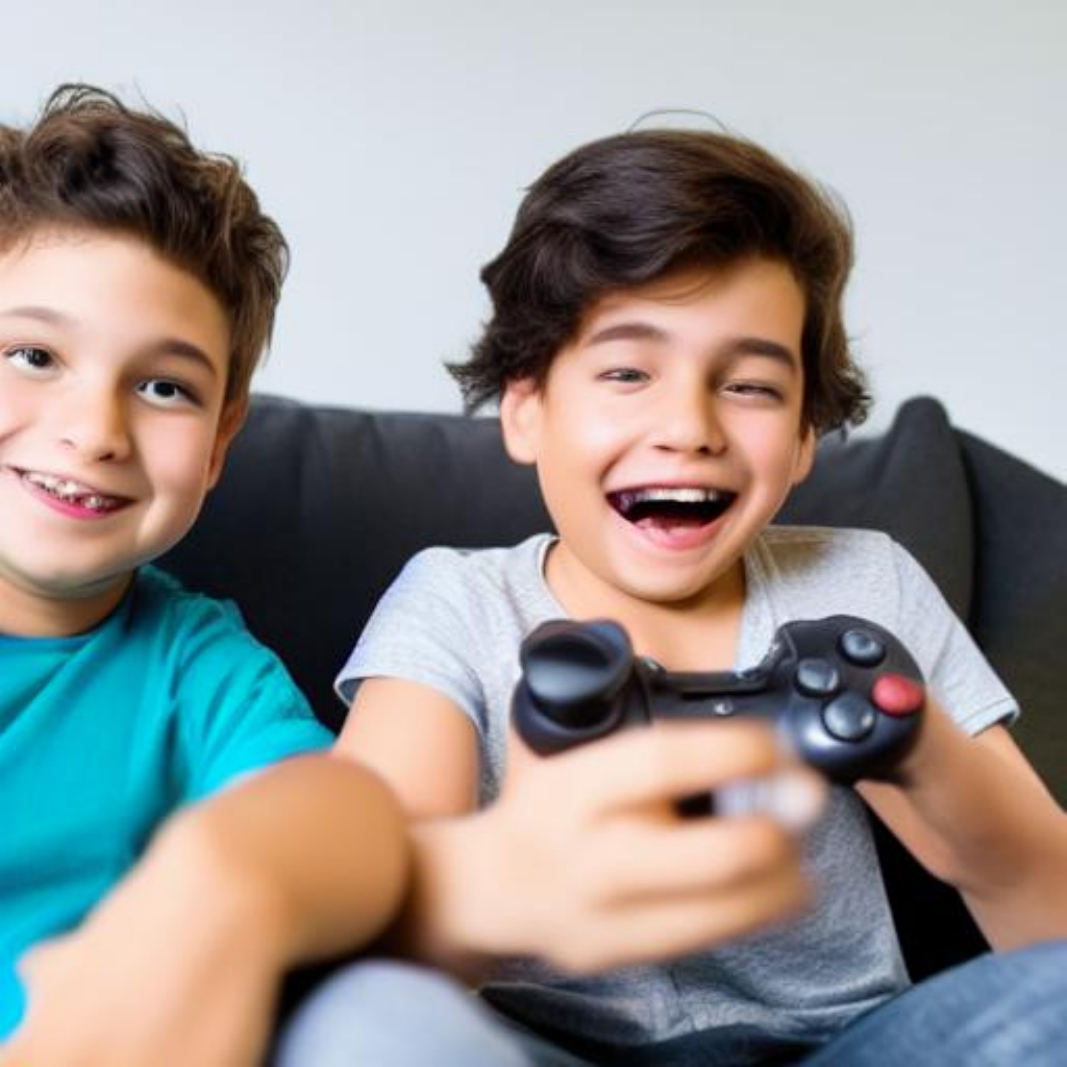}
\includegraphics[width=0.15\columnwidth]{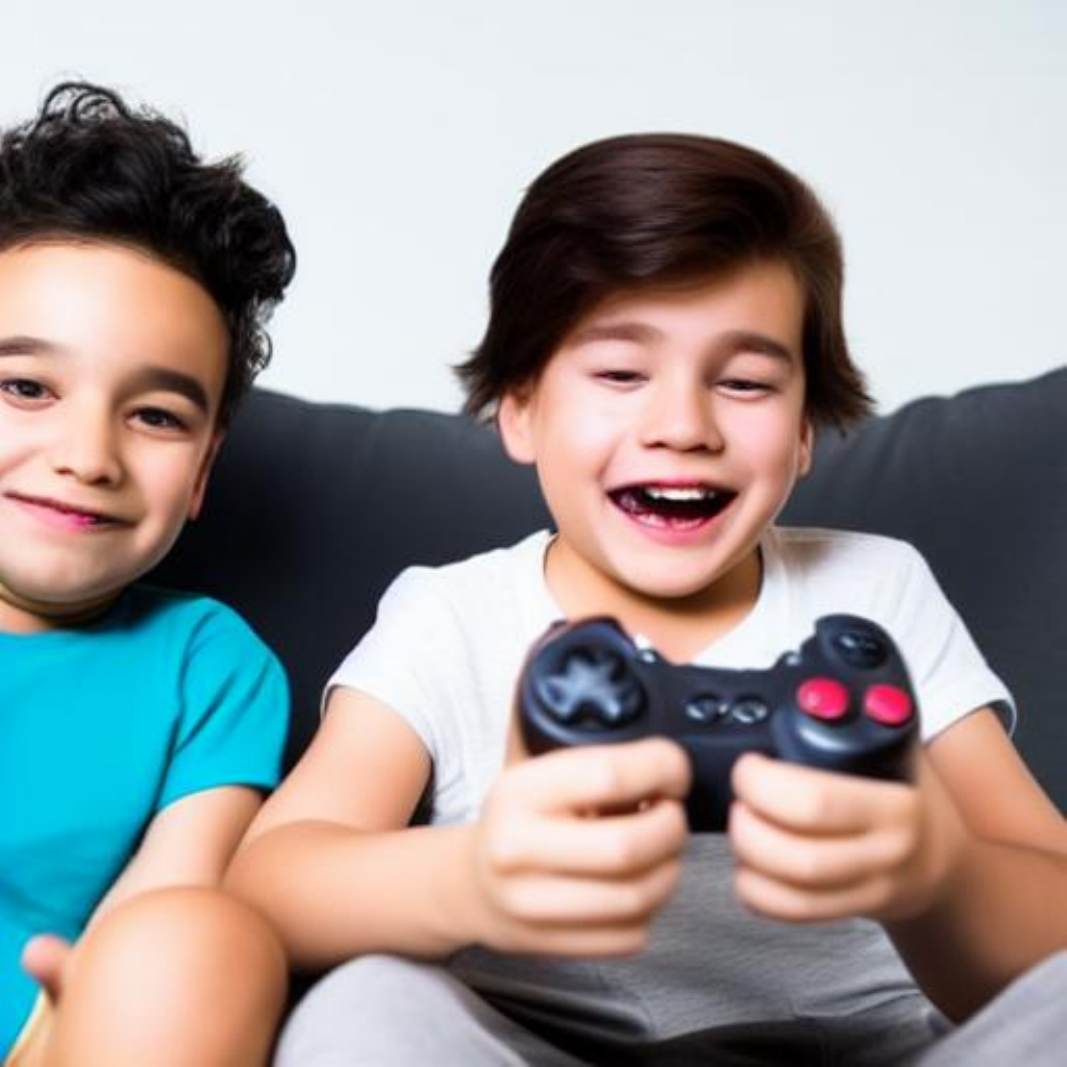}
\includegraphics[width=0.15\columnwidth]{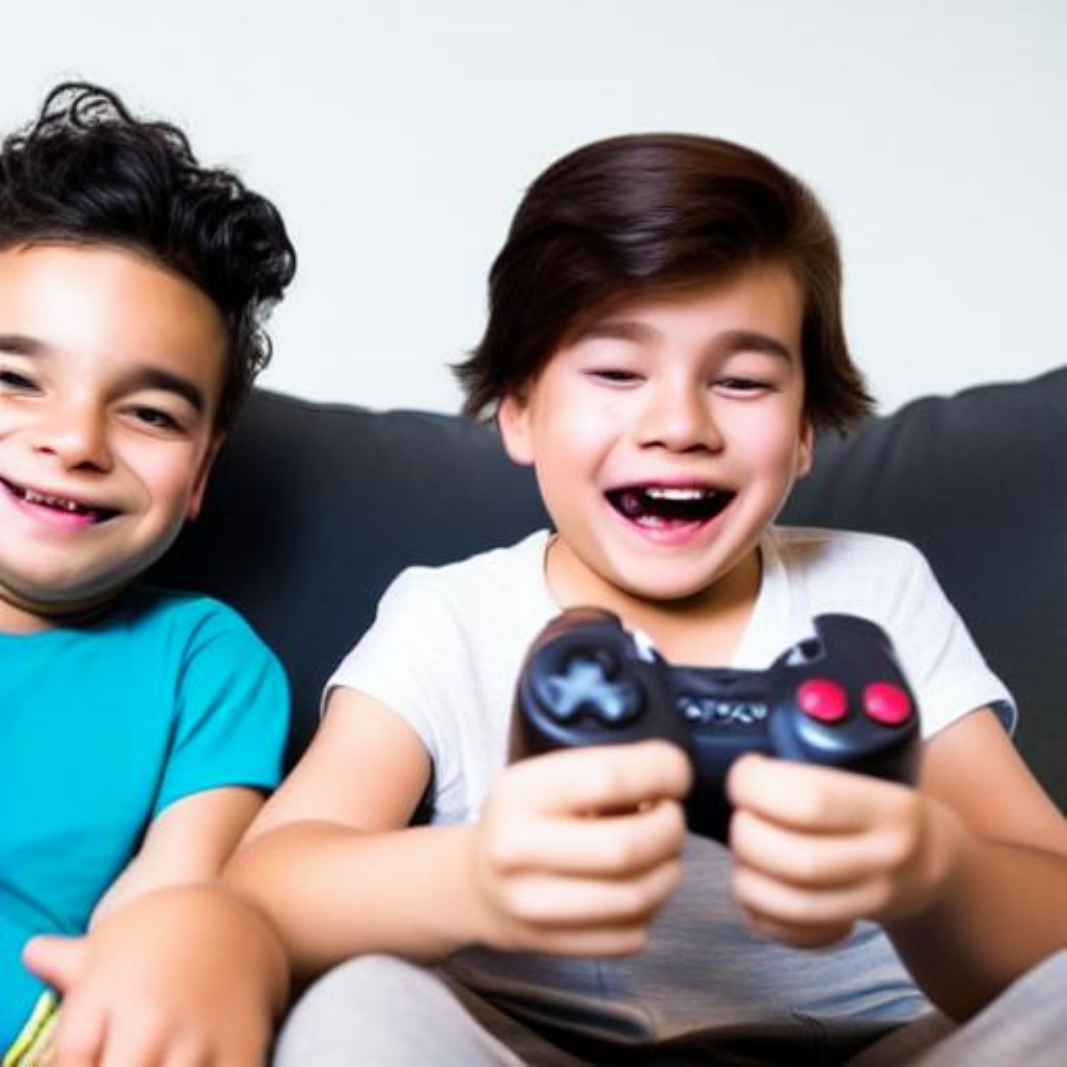}
\\
\includegraphics[width=0.15\columnwidth]{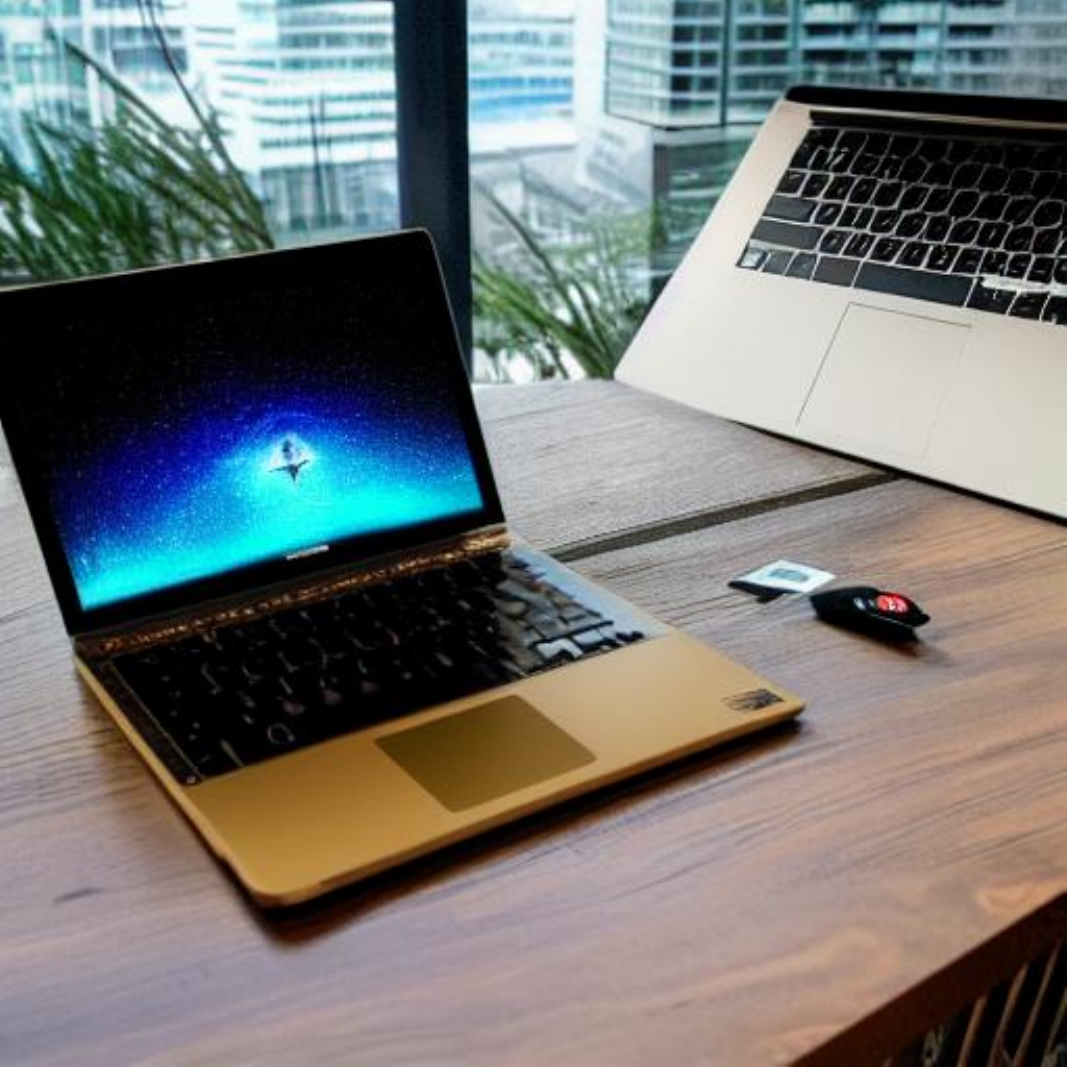}
\includegraphics[width=0.15\columnwidth]{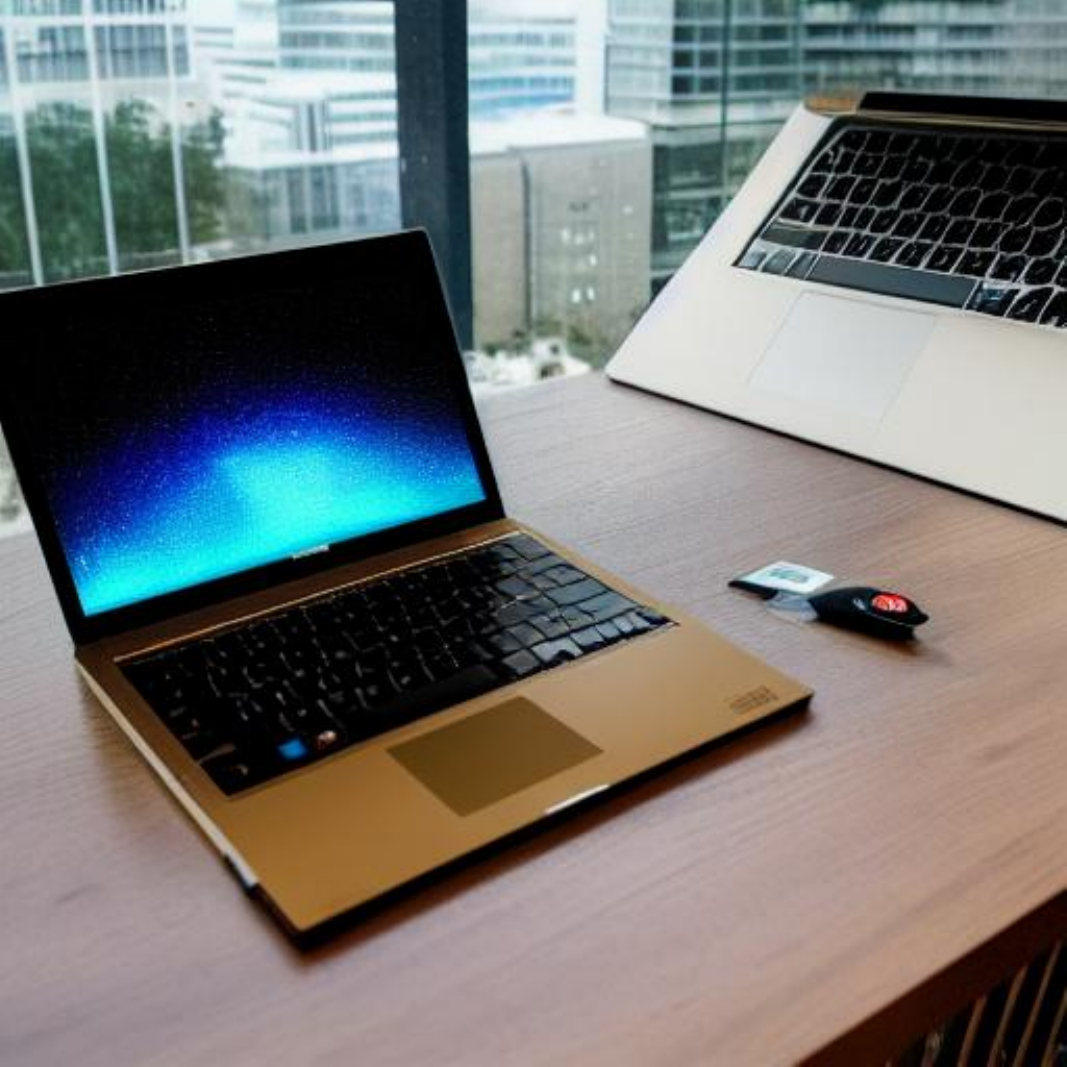}
\includegraphics[width=0.15\columnwidth]{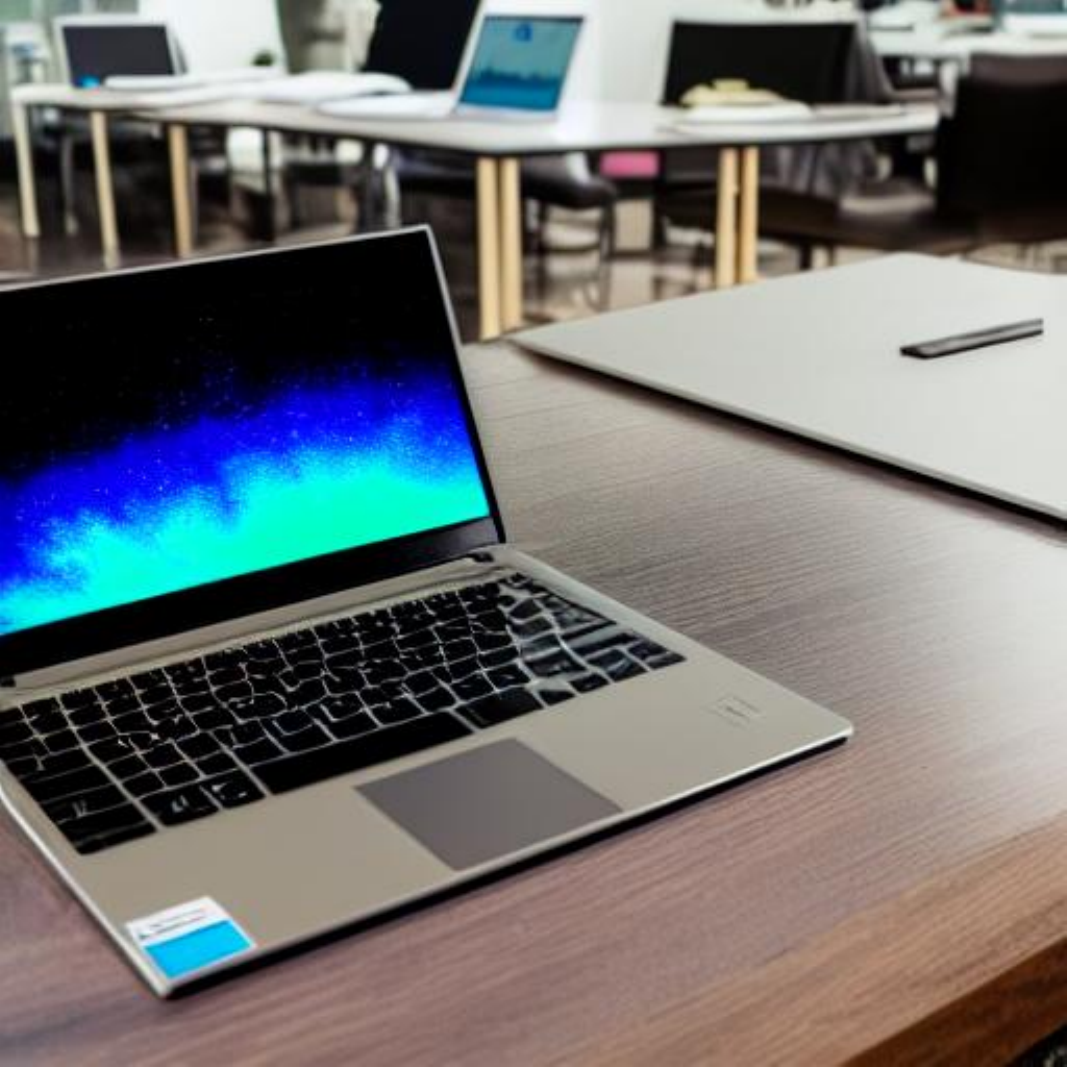}
\includegraphics[width=0.15\columnwidth]{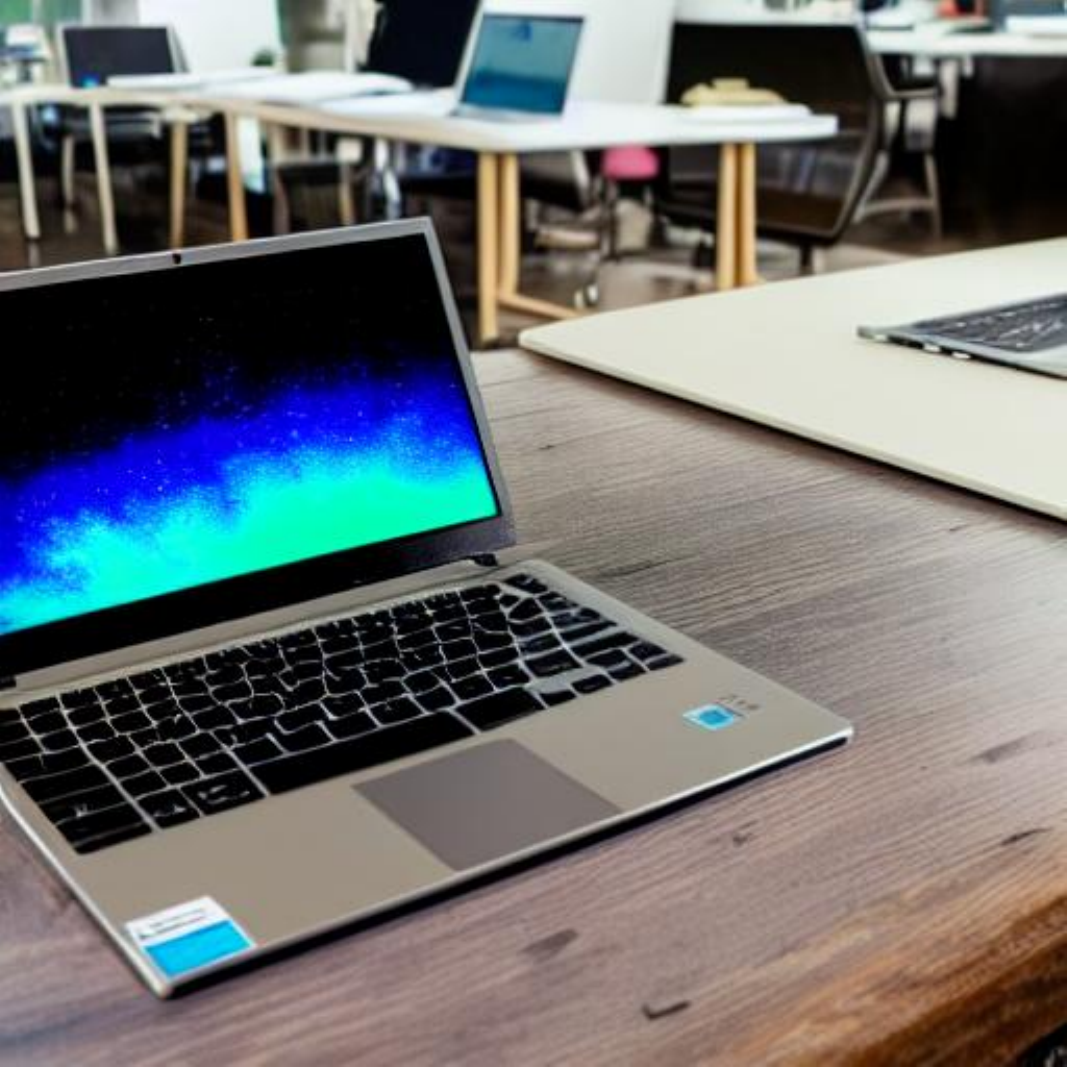}
\caption{\textbf{SD with $80$ Steps.} For each row, from left to right: baseline, ablation, generated image with variance scaling $1.10$, and one with $1.20$.  
}
    \label{fig:sd-80}
    \end{figure}

We present the figures whose references are in Section~\ref{sec:qualitative} of the main paper:
\begin{itemize}
    \item DDPM for $10$ steps in Fig.~\ref{fig:ddpm-10}, for $20$ steps in Fig.~\ref{fig:ddpm-20}, for $100$ steps in Fig.~\ref{fig:DDPM-100-semantic}. The shortcoming of DDPM is illustrated in Fig.~\ref{fig:shortcoming}. Different phenomena occurring with DDPM with $100$ steps are shown in Fig.~\ref{fig:DDPM-100-contrast}.    
    \item LD for $20$ steps in Fig.~\ref{fig:ld-20}, for $40$ steps in Fig.~\ref{fig:ld-40}, for $80$ steps in Fig.~\ref{fig:ld-80}. The shortcoming of LD is illustrated in Fig.~\ref{fig:shortcoming}.
    Different phenomena occurring with LD with $80$ steps are shown in Fig.~\ref{fig:LD-80-contrast}.    
    \item DiT for $40$ steps in Fig.~\ref{fig:dit-40}, for $80$ steps in Fig.~\ref{fig:dit-80}. Different phenomena occurring with DiT with $40$ steps are shown in Fig.~\ref{fig:dit-40-things}.
    \item SD for $20$ steps in Fig.~\ref{fig:sd-20}, for $40$ steps in Fig.~\ref{fig:sd-40}, for $80$ steps in Fig.~\ref{fig:sd-80}. 
    The prompts describing the images in the order they appear in each row of these figures: 
(i)~\emph{``A soccer playing getting ready to kick a ball during a game.''}, 
(ii)~\emph{``An old woman is smelling a box of donuts.''},  
(iii)~\emph{``Two big brown bears playing with one another''},  
(iv)~\emph{``a person riding a dirt bike in the air with a sky background''}, 
(v)~\emph{``A red and black motorcycle with shiny chrome.''}, 
(vi)~\emph{``A baby sitting on a bed comparing toy and real laptop computers.''}, 
(vii)~\emph{``A boy and a man playing video games sitting on a couch.''}, 
(viii)~\emph{``A laptop that is sitting on a table.''}.
\end{itemize}


\subsection{Automated Evaluation: Specifics and More Results}
\label{app:automated-eval}

For DDPM pretrained on the CelebA-HQ dataset, LD, and DiT, the evaluations are over $30000$ sampled images. For SD and DDPM pretrained 
on 
the LSUN dataset, the evaluations are over $10000$ samples images. For the automated evaluation of DiT and SD, the images were downsampled to $256\times 256$ (i.e., half dimension). Additional results for other low numbers of steps are found in Table~\ref{tab:DDPM-auto} for DDPM (pretrained on CelebA-HQ dataset, as in Section~\ref{sec:qualitative}), in Table~\ref{tab:LD-auto} for LD, and in Table~\ref{tab:SD-auto} for SD.

\begin{table*}[t!]
    \caption{
    \textbf{IQA Scores -- DDPM with CelebA-HQ.} }
    \label{tab:DDPM-auto}
    \centering
    \def\arraystretch{1}
    
     \small
    \resizebox{0.9\textwidth}{!}{
    \begin{tabularx}{\textwidth}{c*{7}{>{\centering\arraybackslash}X}}
    \toprule
    & \textbf{MUSIQ-koniq} & \textbf{MUSIQ-ava} & \textbf{MUSIQ-paq2piq} & \textbf{CLIP-IQA} & \textbf{CLIP-IQA$^+$} & \textbf{NIMA-inceptv2} & \textbf{NIMA-vgg16} \\
    \cmidrule{2-8}

\textbf{DDPM--10--B}  & \underline{45.376} & 4.322 & 74.062 & 0.523 & 0.559 & 4.161 & 4.524 \\
\textbf{DDPM--10--Abl}  & 44.303 & 4.353 & 73.876 & 0.519 & 0.554 & 4.184 & 4.498 \\
\textbf{DDPM--10--1.35}  & 45.129 & \textbf{4.393} & \underline{74.262} & \underline{0.531} & \underline{0.567} & \textbf{4.302} & \underline{4.710} \\
\textbf{DDPM--10--1.55}  & \textbf{45.529} & \underline{4.381} & \textbf{74.342} & \textbf{0.533} & \textbf{0.569} & \underline{4.295} & \textbf{4.722} \\
\midrule
%
\textbf{DDPM--20--B}  & \underline{50.586} & 4.362 & 75.832 & 0.563 & \underline{0.599} & 4.226 & 4.723 \\
\textbf{DDPM--20--Abl}  & 49.306 & \underline{4.391} & 75.542 & 0.551 & 0.596 & 4.219 & 4.681 \\
\textbf{DDPM--20--1.35}  & 49.864 & \textbf{4.420} & \underline{75.979} & \underline{0.569} & 0.598 & \textbf{4.395} & \textbf{4.850} \\
\textbf{DDPM--20--1.55}  & \textbf{50.658} & 4.387 & \textbf{76.130} & \textbf{0.576} & \textbf{0.600} & \underline{4.363} & \underline{4.839}\\
    \bottomrule
    \end{tabularx}}
    %
    %
\end{table*}

\begin{table*}[t!]
    \caption{
    \textbf{IQA Scores -- LD.} }
    \label{tab:LD-auto}
    \centering
    \def\arraystretch{1}
    
     \small
    \resizebox{0.9\textwidth}{!}{
    \begin{tabularx}{\textwidth}{c*{7}{>{\centering\arraybackslash}X}}
    \toprule
    & \textbf{MUSIQ-koniq} & \textbf{MUSIQ-ava} & \textbf{MUSIQ-paq2piq} & \textbf{CLIP-IQA} & \textbf{CLIP-IQA$^+$} & \textbf{NIMA-inceptv2} & \textbf{NIMA-vgg16}\\

    \cmidrule{2-8}

\textbf{LD--20--B}  & 54.939 & 4.327 & 77.007 & 0.605 & 0.634 & 4.114 & 4.653 \\
\textbf{LD--20--Abl}  & 51.785 & \underline{4.362} & 76.395 & 0.571 & 0.614 & 4.107 & 4.604\\
\textbf{LD--20--1.35}  & \underline{55.877} & \textbf{4.381} & \underline{77.270} & \underline{0.614} & \underline{0.646} & \underline{4.216} & \underline{4.738} \\
\textbf{LD--20--1.55}  & \textbf{57.948} & 4.350 & \textbf{77.596} & \textbf{0.635} & \textbf{0.655} & \textbf{4.223} & \textbf{4.767}\\
\midrule
%
\textbf{LD--40--B}  & \underline{60.843} & 4.316 & 78.058 & 0.647 & 0.670 & 4.194 & 4.775 \\
\textbf{LD--40--Abl}  & 57.572 & 4.358 & 77.547 & 0.614 & 0.657 & 4.164 & 4.716 \\
\textbf{LD--40--1.20}  & 61.305 & \underline{4.403} & \underline{78.237} & \underline{0.655} & \underline{0.683} & \underline{4.356} & \underline{4.893} \\
\textbf{LD--40--1.35}  & \textbf{63.631} & \textbf{4.362} & \textbf{78.524} & \textbf{0.677} & \textbf{0.685} &\textbf{ 4.360} & \textbf{4.911} \\
    \bottomrule
    \end{tabularx}}
    %
    %
\end{table*}

\begin{table*}[t!]
    \caption{
    \textbf{IQA Scores -- SD.} }
    \label{tab:SD-auto}
    \centering
    \def\arraystretch{1}
    
     \small
    \resizebox{0.9\textwidth}{!}{
    \begin{tabularx}{\textwidth}{c*{7}{>{\centering\arraybackslash}X}}
    \toprule
    & \textbf{MUSIQ-koniq} & \textbf{MUSIQ-ava} & \textbf{MUSIQ-paq2piq} & \textbf{CLIP-IQA} & \textbf{CLIP-IQA$^+$} & \textbf{NIMA-inceptv2} & \textbf{NIMA-vgg16}\\

    \cmidrule{2-8}

\textbf{SD--20--B}  & 57.096 & 4.327 & \underline{77.072} & \underline{0.604} & \underline{0.679} & 4.456 & 4.587 \\
\textbf{SD--20--Abl}  & 56.767 & \textbf{4.358} & 76.776 & 0.599 & 0.673 & \underline{4.494} & 4.593 \\
\textbf{SD--20--1.25}  & \underline{56.928} & \underline{4.356} & 77.029 & \underline{0.604} & 0.677 & \textbf{4.503} & \textbf{4.617} \\
\textbf{SD--20--1.55}  & \textbf{57.214} & 4.326 & \textbf{77.270} & \textbf{0.607} & \textbf{0.681} & 4.466 & \underline{4.609} \\
\midrule
\textbf{SD--40--B}  & \textbf{58.180} & 4.323 & 77.742 & \underline{0.622} & \textbf{0.686} & 4.439 & 4.607 \\
\textbf{SD--40--Abl}  & 57.970 & \textbf{4.362} & 77.496 & 0.621 & 0.684 & \textbf{4.487} & \underline{4.624} \\
\textbf{SD--40--1.25}  & 57.777 & \underline{4.343} & \underline{77.780} & \textbf{0.623} & \underline{0.685} & \underline{4.475} & \textbf{4.633} \\
\textbf{SD--40--1.55}  & \underline{58.021} & 4.269 & \textbf{78.136} & \textbf{0.623} & 0.684 & 4.383 & 4.589 \\
    \bottomrule
    \end{tabularx}}
    %
    %
\end{table*}

\begin{figure}[t]
    \centering
\includegraphics[width=0.15\columnwidth]{img_dif/DDPM_20_churches/b_25.pdf}
\includegraphics[width=0.15\columnwidth]{img_dif/DDPM_20_churches/a_25.pdf}
\includegraphics[width=0.15\columnwidth]{img_dif/DDPM_20_churches/m_135_25.pdf}
\includegraphics[width=0.15\columnwidth]{img_dif/DDPM_20_churches/m_155_25.pdf}
\\
\includegraphics[width=0.15\columnwidth]{img_dif/DDPM_20_churches/b_61.pdf}
\includegraphics[width=0.15\columnwidth]{img_dif/DDPM_20_churches/a_61.pdf}
\includegraphics[width=0.15\columnwidth]{img_dif/DDPM_20_churches/m_135_61.pdf}
\includegraphics[width=0.15\columnwidth]{img_dif/DDPM_20_churches/m_155_61.pdf}
\\
\includegraphics[width=0.15\columnwidth]{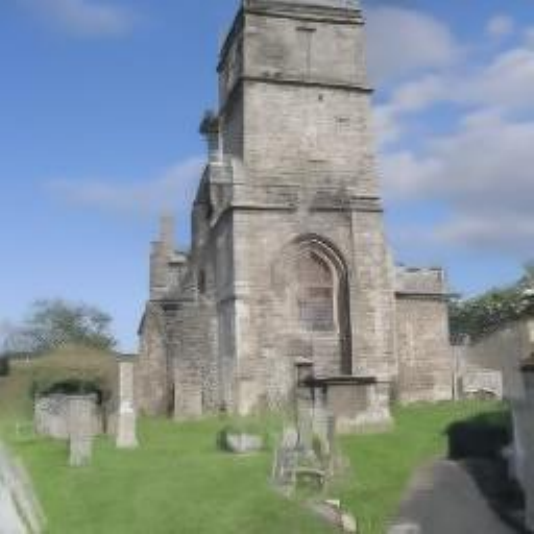}
\includegraphics[width=0.15\columnwidth]{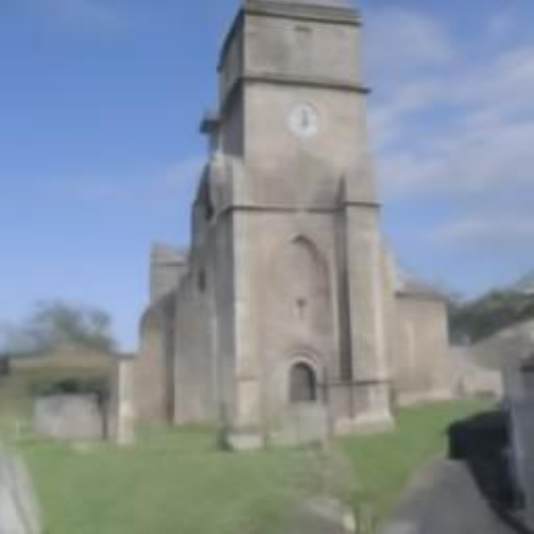}
\includegraphics[width=0.15\columnwidth]{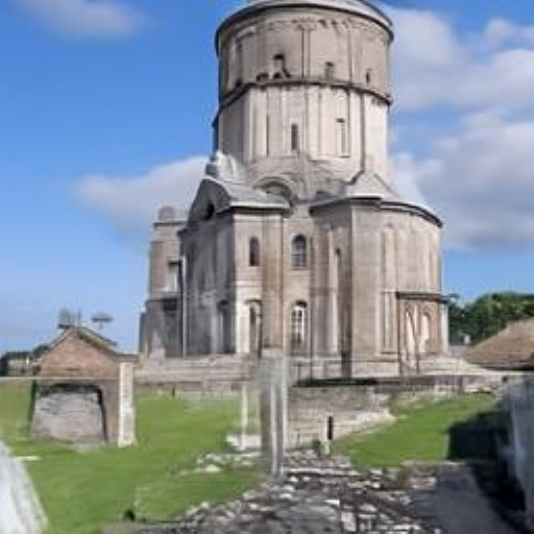}
\includegraphics[width=0.15\columnwidth]{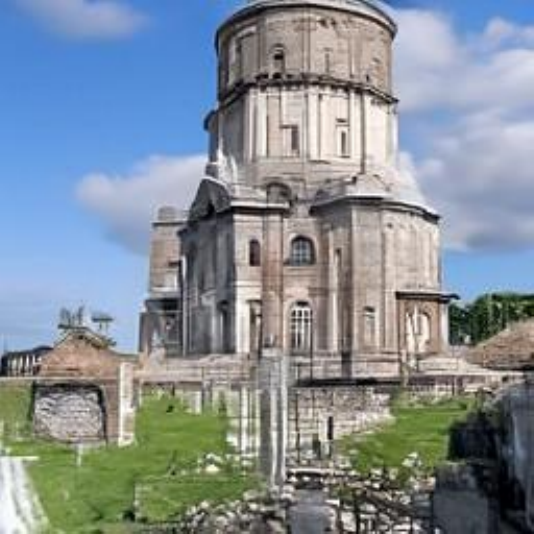}
\\
\includegraphics[width=0.15\columnwidth]{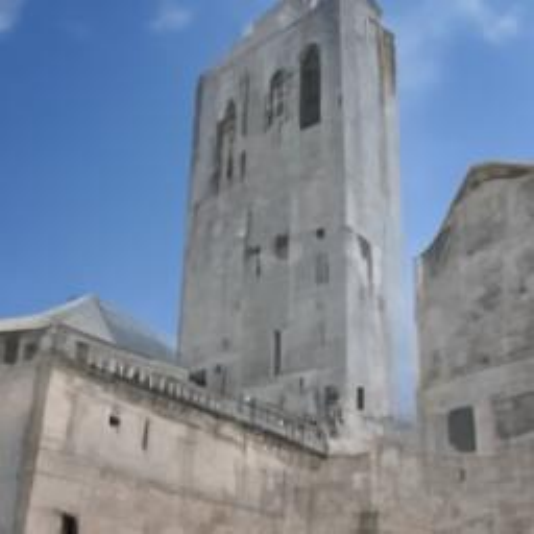}
\includegraphics[width=0.15\columnwidth]{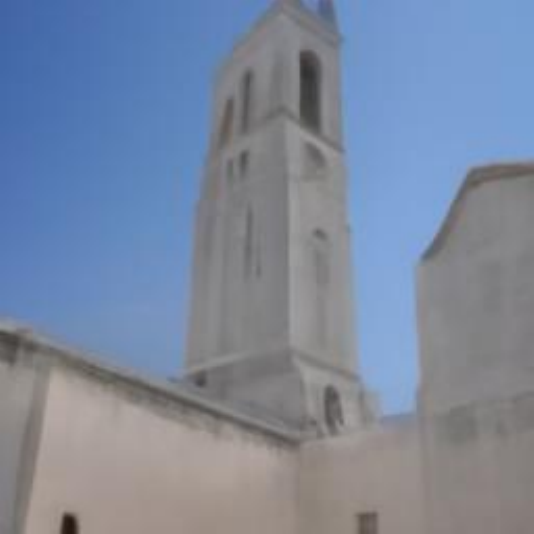}
\includegraphics[width=0.15\columnwidth]{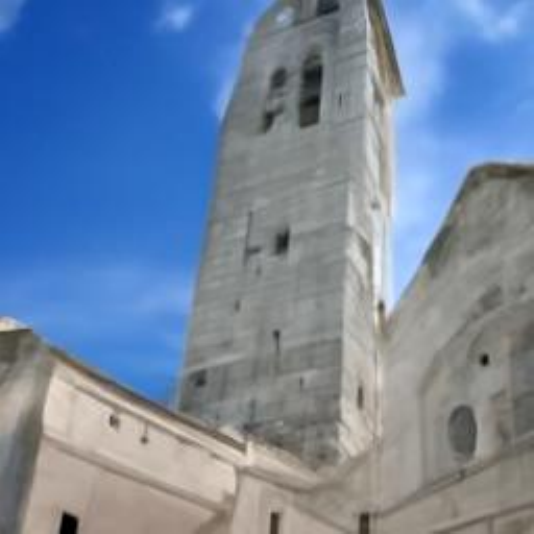}
\includegraphics[width=0.15\columnwidth]{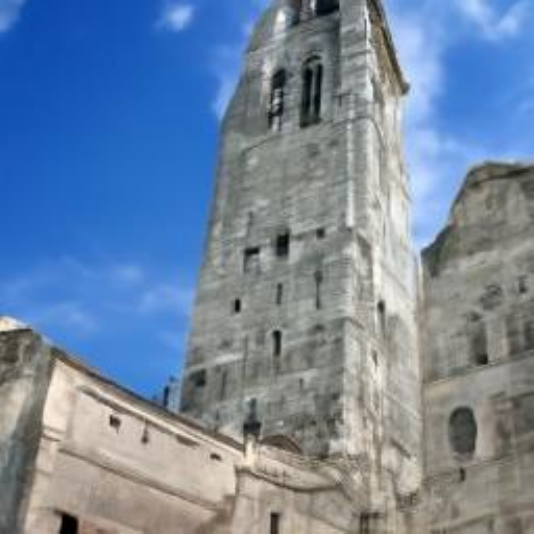}
\\
\includegraphics[width=0.15\columnwidth]{img_dif/DDPM_20_churches/b_99.pdf}
\includegraphics[width=0.15\columnwidth]{img_dif/DDPM_20_churches/a_99.pdf}
\includegraphics[width=0.15\columnwidth]{img_dif/DDPM_20_churches/m_135_99.pdf}
\includegraphics[width=0.15\columnwidth]{img_dif/DDPM_20_churches/m_155_99.pdf}
\\
\includegraphics[width=0.15\columnwidth]{img_dif/DDPM_20_churches/b_182.pdf}
\includegraphics[width=0.15\columnwidth]{img_dif/DDPM_20_churches/a_182.pdf}
\includegraphics[width=0.15\columnwidth]{img_dif/DDPM_20_churches/m_135_182.pdf}
\includegraphics[width=0.15\columnwidth]{img_dif/DDPM_20_churches/m_155_182.pdf}
\\
\includegraphics[width=0.15\columnwidth]{img_dif/DDPM_20_churches/b_203.pdf}
\includegraphics[width=0.15\columnwidth]{img_dif/DDPM_20_churches/a_203.pdf}
\includegraphics[width=0.15\columnwidth]{img_dif/DDPM_20_churches/m_135_203.pdf}
\includegraphics[width=0.15\columnwidth]{img_dif/DDPM_20_churches/m_155_203.pdf}
\\
\includegraphics[width=0.15\columnwidth]{img_dif/DDPM_20_churches/b_235.pdf}
\includegraphics[width=0.15\columnwidth]{img_dif/DDPM_20_churches/a_235.pdf}
\includegraphics[width=0.15\columnwidth]{img_dif/DDPM_20_churches/m_135_235.pdf}
\includegraphics[width=0.15\columnwidth]{img_dif/DDPM_20_churches/m_155_235.pdf}
\\
\caption{\textbf{DDPM with $20$ Steps for Church Outdoor Category of the LSUN Dataset.} 
For each row, from left to right: baseline, ablation, generated image with variance scaling $1.35$, and one with $1.55$. 
%
%
}
    \label{fig:ddpm-20-churches}
    \end{figure}

\begin{figure}[t]
    \centering
\includegraphics[width=0.15\columnwidth]{img_dif/DDPM_20_bedrooms/b_68.pdf}
\includegraphics[width=0.15\columnwidth]{img_dif/DDPM_20_bedrooms/a_68.pdf}
\includegraphics[width=0.15\columnwidth]{img_dif/DDPM_20_bedrooms/m_135_68.pdf}
\includegraphics[width=0.15\columnwidth]{img_dif/DDPM_20_bedrooms/m_155_68.pdf}
\\
\includegraphics[width=0.15\columnwidth]{img_dif/DDPM_20_bedrooms/b_93.pdf}
\includegraphics[width=0.15\columnwidth]{img_dif/DDPM_20_bedrooms/a_93.pdf}
\includegraphics[width=0.15\columnwidth]{img_dif/DDPM_20_bedrooms/m_135_93.pdf}
\includegraphics[width=0.15\columnwidth]{img_dif/DDPM_20_bedrooms/m_155_93.pdf}
\\
\includegraphics[width=0.15\columnwidth]{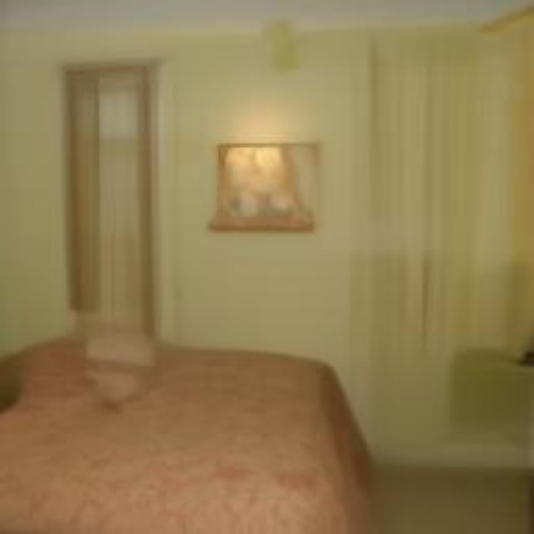}
\includegraphics[width=0.15\columnwidth]{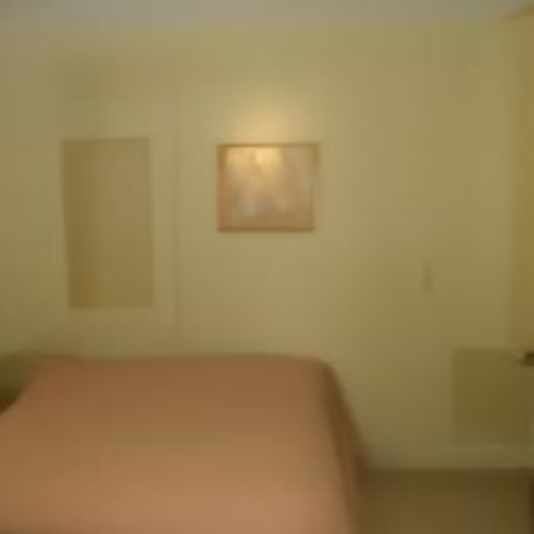}
\includegraphics[width=0.15\columnwidth]{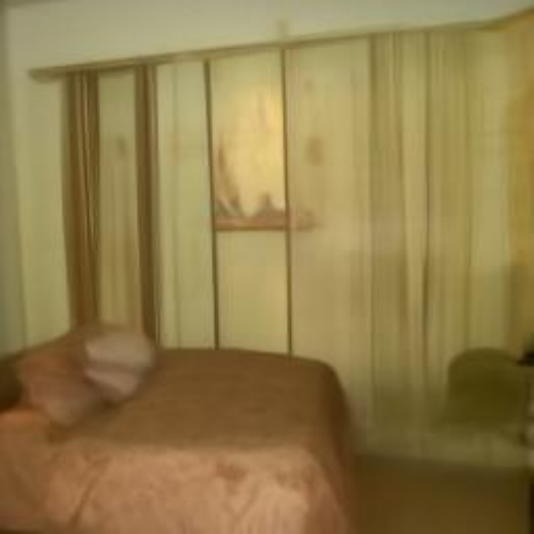}
\includegraphics[width=0.15\columnwidth]{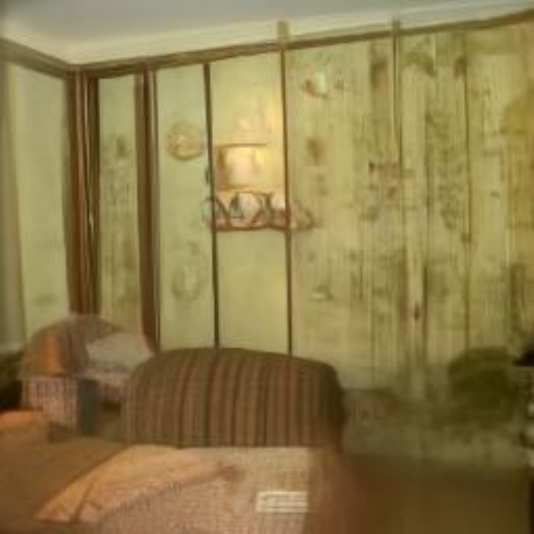}
\\
\includegraphics[width=0.15\columnwidth]{img_dif/DDPM_20_bedrooms/b_210.pdf}
\includegraphics[width=0.15\columnwidth]{img_dif/DDPM_20_bedrooms/a_210.pdf}
\includegraphics[width=0.15\columnwidth]{img_dif/DDPM_20_bedrooms/m_135_210.pdf}
\includegraphics[width=0.15\columnwidth]{img_dif/DDPM_20_bedrooms/m_155_210.pdf}
\\
\includegraphics[width=0.15\columnwidth]{img_dif/DDPM_20_bedrooms/b_236.pdf}
\includegraphics[width=0.15\columnwidth]{img_dif/DDPM_20_bedrooms/a_236.pdf}
\includegraphics[width=0.15\columnwidth]{img_dif/DDPM_20_bedrooms/m_135_236.pdf}
\includegraphics[width=0.15\columnwidth]{img_dif/DDPM_20_bedrooms/m_155_236.pdf}
\\
\includegraphics[width=0.15\columnwidth]{img_dif/DDPM_20_bedrooms/b_242.pdf}
\includegraphics[width=0.15\columnwidth]{img_dif/DDPM_20_bedrooms/a_242.pdf}
\includegraphics[width=0.15\columnwidth]{img_dif/DDPM_20_bedrooms/m_135_242.pdf}
\includegraphics[width=0.15\columnwidth]{img_dif/DDPM_20_bedrooms/m_155_242.pdf}
\\
\includegraphics[width=0.15\columnwidth]{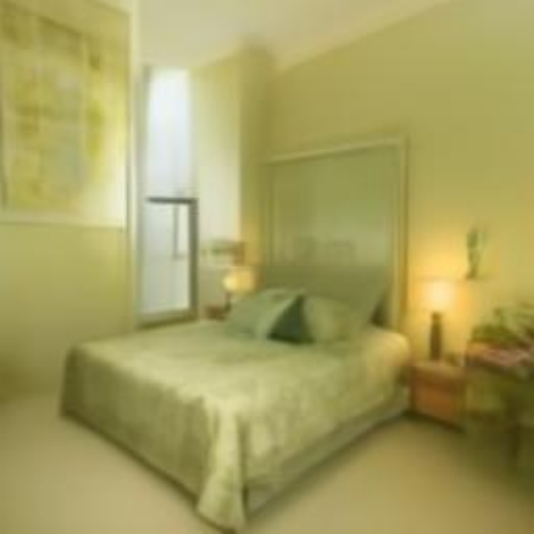}
\includegraphics[width=0.15\columnwidth]{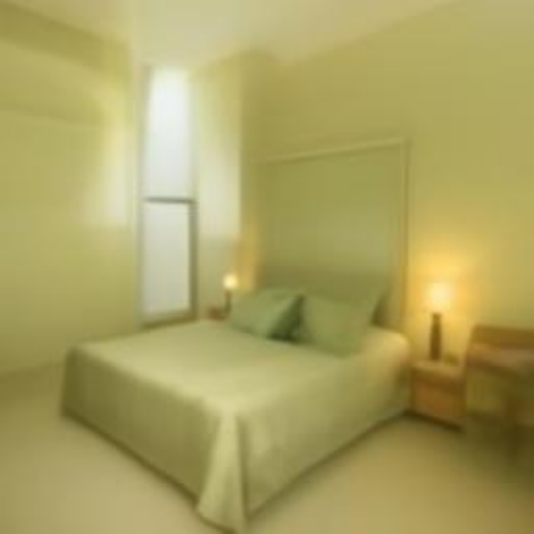}
\includegraphics[width=0.15\columnwidth]{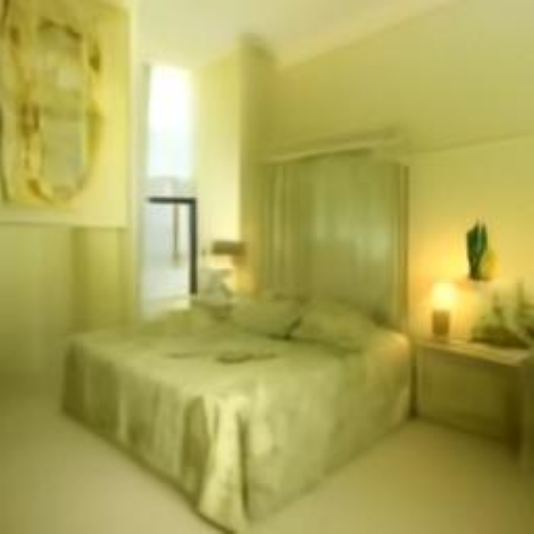}
\includegraphics[width=0.15\columnwidth]{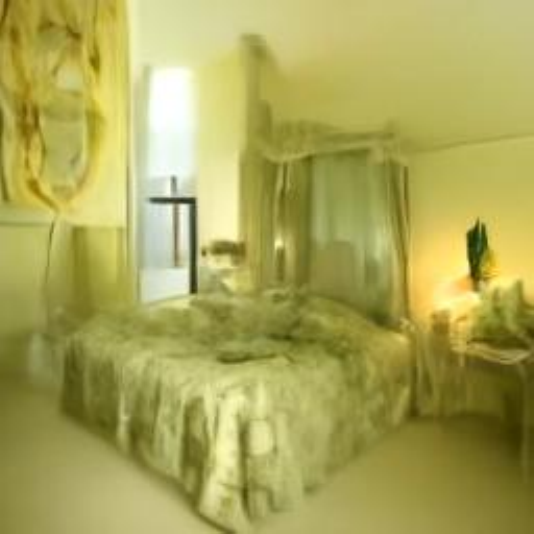}
\\
\includegraphics[width=0.15\columnwidth]{img_dif/DDPM_20_bedrooms/b_409.pdf}
\includegraphics[width=0.15\columnwidth]{img_dif/DDPM_20_bedrooms/a_409.pdf}
\includegraphics[width=0.15\columnwidth]{img_dif/DDPM_20_bedrooms/m_135_409.pdf}
\includegraphics[width=0.15\columnwidth]{img_dif/DDPM_20_bedrooms/m_155_409.pdf}
\\
\caption{\textbf{DDPM with $20$ Steps for Bedroom Category of the LSUN Dataset.} 
For each row, from left to right: baseline, ablation, generated image with variance scaling $1.35$, and one with $1.55$. 
%
%
}
    \label{fig:ddpm-20-bed}
    \end{figure}

The figures showing more sampled images corresponding to the pretrained DDPMs on the LSUN dataset are in Fig.~\ref{fig:ddpm-20-churches} for the ``outdoor church'' category and in Fig.~\ref{fig:ddpm-20-bed} for the ``bedroom'' category.

\section{About the Human Evaluation Study}
\label{app:humaneval}

The use of 
2-alternative force-choice (2FAC) testing in our human evaluation study is motivated by its prior use in the literature~\citep{zhang-2012-colorfulimagecolorization,LD_Rombach_2022_CVPR,saharia-2023-superresolution} 
and its known reliability for image quality assessment
~\citep{mantiuk-2012-comparison}. 

For the images generated by S2EP, we set the mixing parameter to $0.015$ and variance scaling to $1.35$. The DDPM model has $10$ steps, LD has $20$ steps, and DiT has $40$ steps. These settings are found in both our qualitative (Section~\ref{sec:qualitative}) and automated evaluation (Section~\ref{sec:metric}) studies.

\subsection{Further Details on the Evaluation}

We now specify the instructions given to the evaluators. For DDPM and LD, since they are unconditional models that generate faces, the instructions start with:
``\emph{You will be shown 25 pairs of images. For each pair, please select which image ``Option A'' (left) or ``Option B'' (right) you think is more visually appealing and has a better quality (for example, sharp edges, clear object details/shapes/proportions, etc.), regardless of the picture subject content.}'' Then, two lines below we have:
``\emph{Please, evaluate each pair of images independently of the other pairs.}''

%
%

In the case of DiT, each pair of images are from the same class subject, but the classes are different across pairs. Thus, to help the human evaluator, we provided the \emph{class name} of each pair. 
Now, since DiT is a class conditioned model, there \emph{has to be} a correspondence between the sampled image and the class subject, \emph{unless} the quality of the image is extremely poor that it is hard to see what it is truly depicting.  
%
For this reason, the instructions start with: ``\emph{You will be shown 25 pairs of images. For each pair, please select which image ``Option A'' (left) or ``Option B'' (right) you think better matches the subject (the subject is indicated above each pair of images). If both images equally matches the subject, consider the image that is more visually appealing and has a better quality (for example, sharp edges, clear object details/shapes/proportions, etc.) for the picture subject content.}'' Then, two lines below we have: ``\emph{Please, evaluate each pair of images independently of the other pairs.}''

\subsection{Human Evaluations in the Literature}

We recall that our human evaluation study consists of three studies: one per diffusion model. Each study has $33$ participants performing $25$ pairwise comparisons. This gives a total of $2475$ pairwise comparisons or responses.

For the sake of having a perspective on how human evaluations are reported in the literature of diffusion models, we mention a few specific examples. These works in the literature evaluate 
diverse criteria: while some evaluate image quality or aesthetics, others evaluate text-to-image correspondence, photorealism, or resolution. If there is any conclusion to be made, is that there is no \emph{standard} on the number of human evaluators---very often is not even mentioned---and on how much work each evaluator does. Another conclusion is that \emph{pairwise comparisons} are a commonly used type of evaluation.
\begin{itemize}
    \item \citep{nichol-2022-glide} does not report the number of human evaluators. It performs three tests with $1000$, $1000$, and $500$ pairwise comparisons, respectively---a total of $2500$ responses. It is unclear how all these tests are distributed across the human participants. Choices are not 2FAC: the human judge can choose a third option which is that ``neither image is significantly better than the other''. 
    \item \citep{zhang-2023-addingcontroltexttoimage} reports $12$ human evaluators who rank twenty groups of five images, ranking each result from $1$ to $5$ (lower is worse)---so no pairwise comparisons. There are a total of $240$ responses. 
    \item \citep{xue-2023-raphael} does not report the number of human evaluators. It uses a benchmark described in~\citep{feng-2023-ernie} for human pairwise comparison of sets of images. 
Thus, it performs four experiments with two comparison rubrics each, totaling eight studies. 
Although the benchmark by~\citep{feng-2023-ernie} contains $300$ tests for evaluation, 
it is unclear to us whether all of them are used in each of the studies by \citep{xue-2023-raphael} (and even by~\citep{feng-2023-ernie} itself). Thus, we are not sure about the total number of responses. Moreover, while \citep{feng-2023-ernie} reports $5$ subjects in its study, \citep{xue-2023-raphael} does not report any number.
%
Choices are not 2FAC: the human judge can choose a third option which is that ``there is no measurable difference" between the two images.
%
    %
    \item \citep{LD_Rombach_2022_CVPR} indicates it uses the same human evaluation protocol as~\citep{saharia-2023-superresolution}---though the latter mentions that $50$ human evaluators participated with each one comparing $50$ pairs of images, the former does not provide the number of evaluators nor the number of comparisons per evaluator. 
    The evaluation setting is 2FAC. 
    \citep{LD_Rombach_2022_CVPR} has a total of eight studies (a super-resolution and an inpainting setting, each one doing two comparisons between models across two tasks). Since we do not know the number of evaluators and pairwise comparisons, and how the evaluators were divided among the studies, we cannot calculate the total amount of responses.
    \citep{saharia-2023-superresolution} has a total of twelve studies (in one setting: four comparisons between models across two different tasks; in the other setting: two comparisons between models across two different configurations). Again, it is unclear to us how the evaluators were divided across the studies---one possibility is that each study had $50$ pairs of images done by the $50$ different participants, giving a total of $30000$ responses. 
\end{itemize}


\section{About Parameter Changes}
\label{app:param-changes}

\subsection{Figures}

\begin{figure}[t!]
    \centering
\includegraphics[width=0.1\columnwidth]{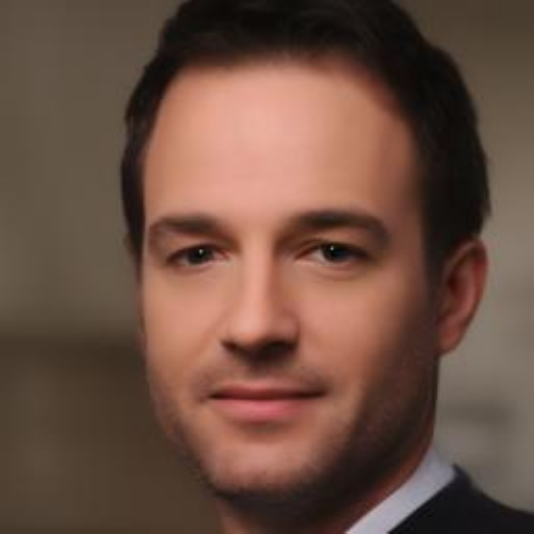}
\includegraphics[width=0.1\columnwidth]{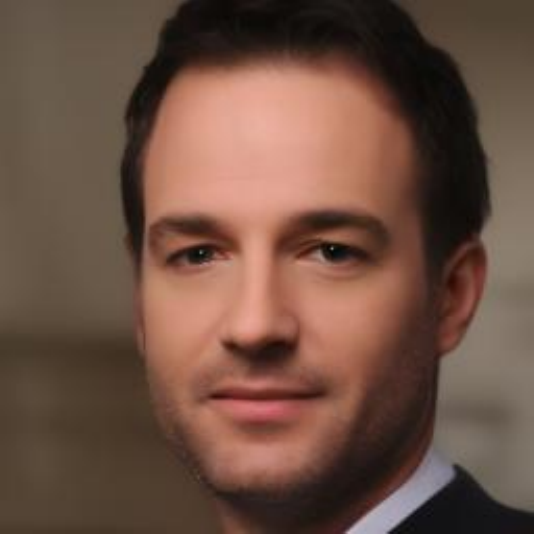}
\includegraphics[width=0.1\columnwidth]{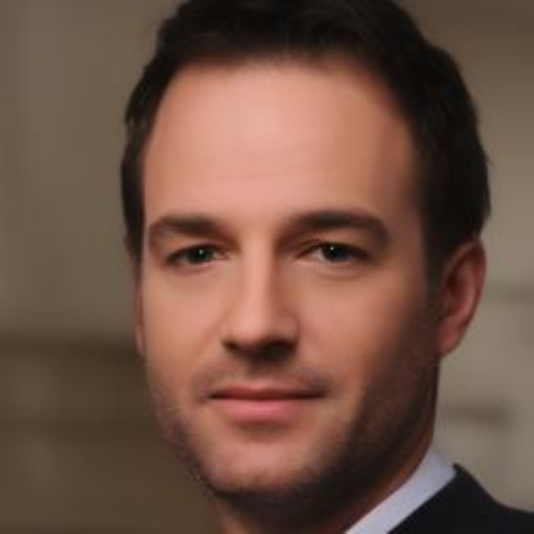}
\includegraphics[width=0.1\columnwidth]{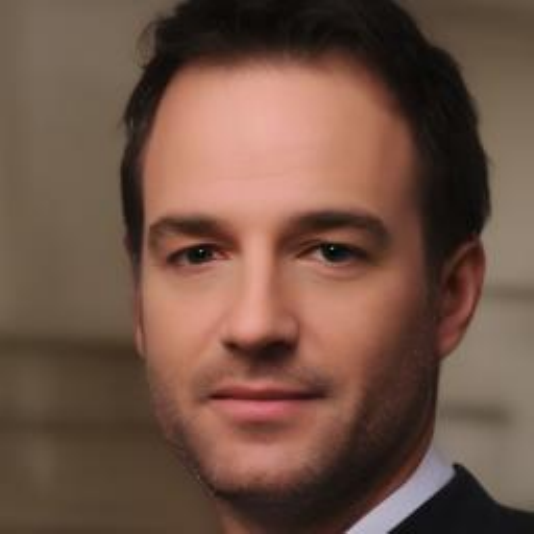}
\includegraphics[width=0.1\columnwidth]{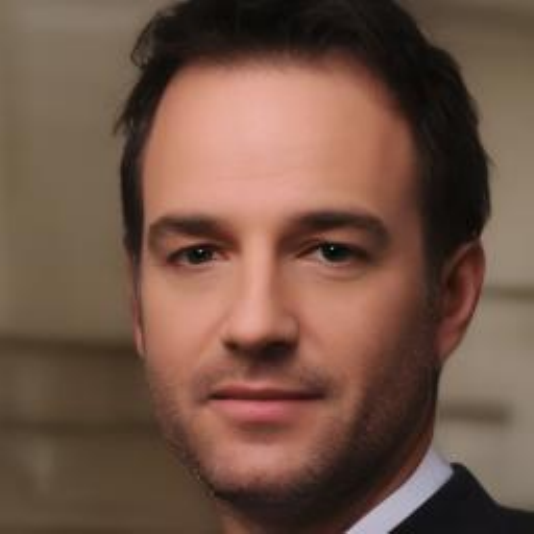}
\includegraphics[width=0.1\columnwidth]{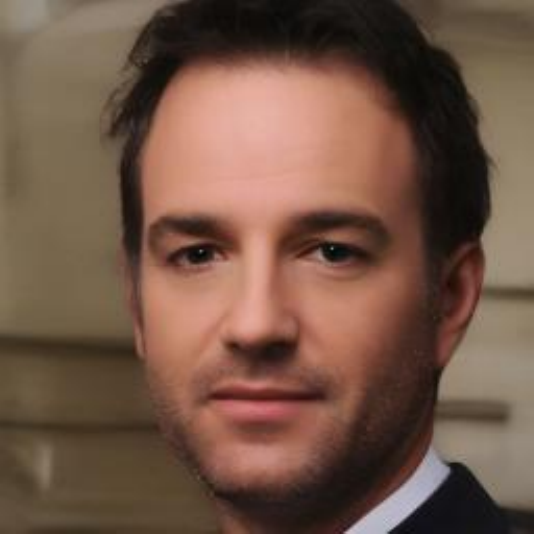}
\includegraphics[width=0.1\columnwidth]{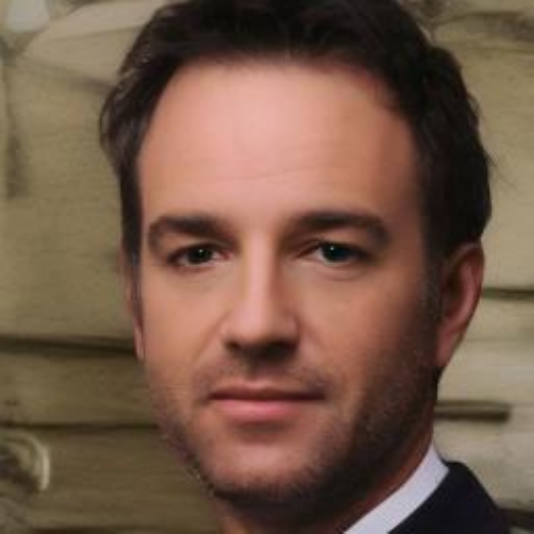}
\includegraphics[width=0.1\columnwidth]{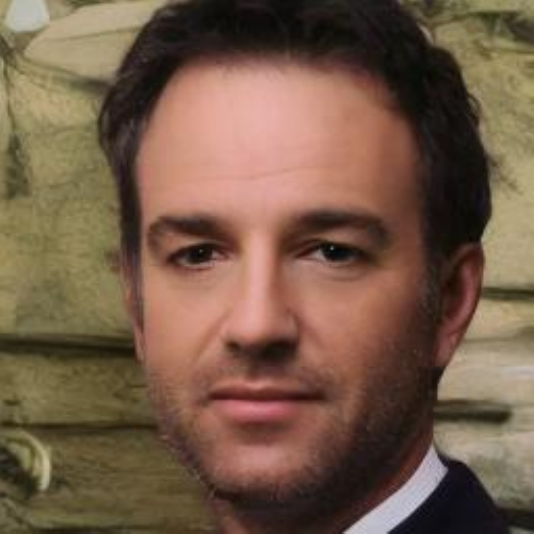}
\includegraphics[width=0.1\columnwidth]{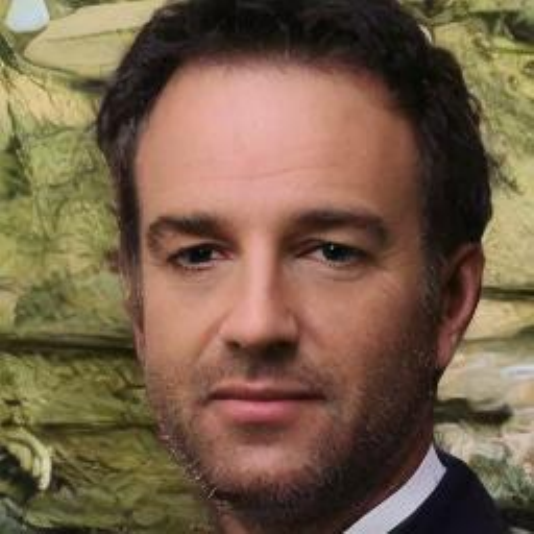}
\\
\includegraphics[width=0.1\columnwidth]{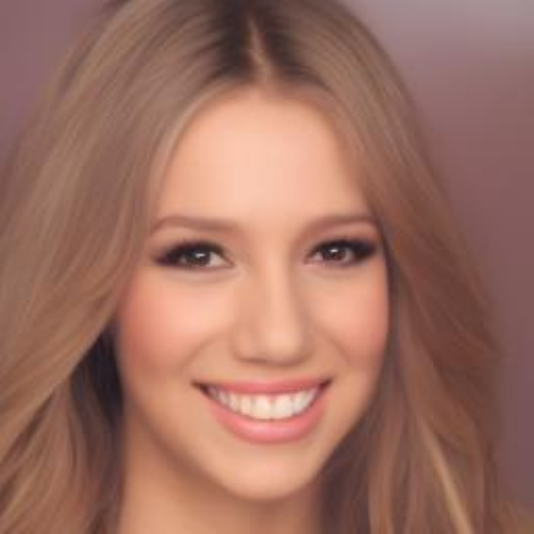}
\includegraphics[width=0.1\columnwidth]{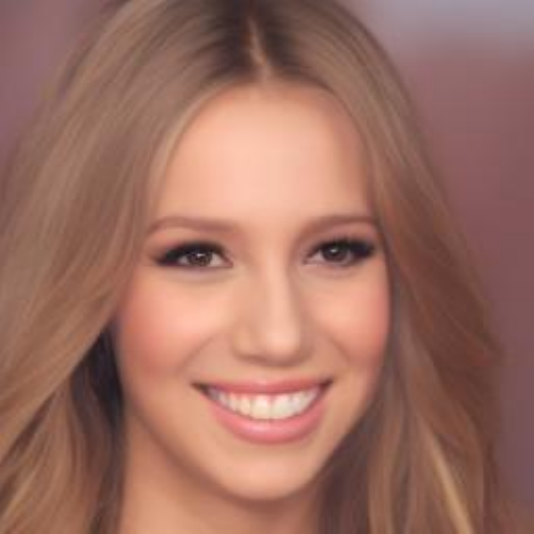}
\includegraphics[width=0.1\columnwidth]{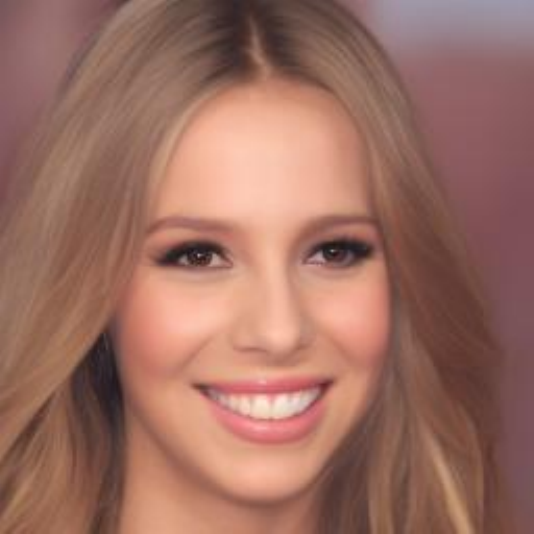}
\includegraphics[width=0.1\columnwidth]{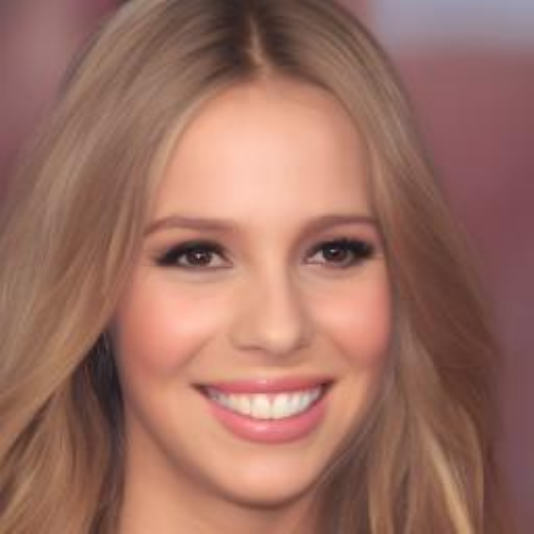}
\includegraphics[width=0.1\columnwidth]{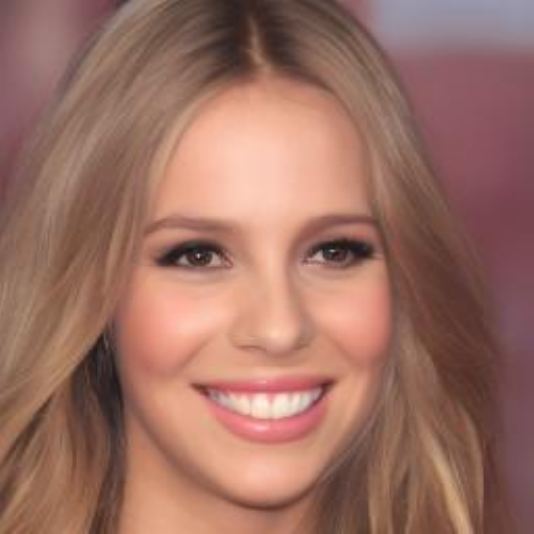}
\includegraphics[width=0.1\columnwidth]{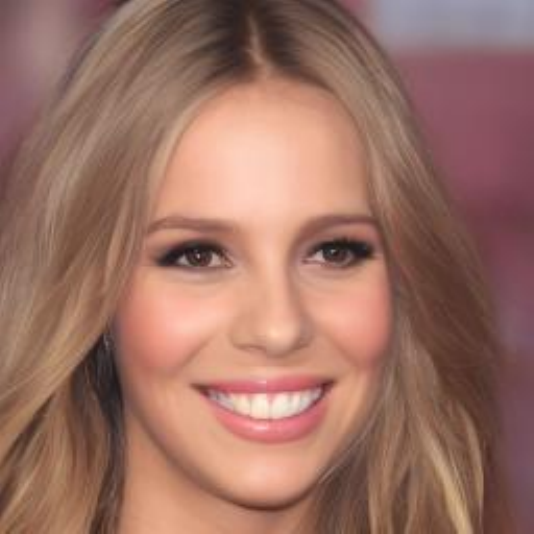}
\includegraphics[width=0.1\columnwidth]{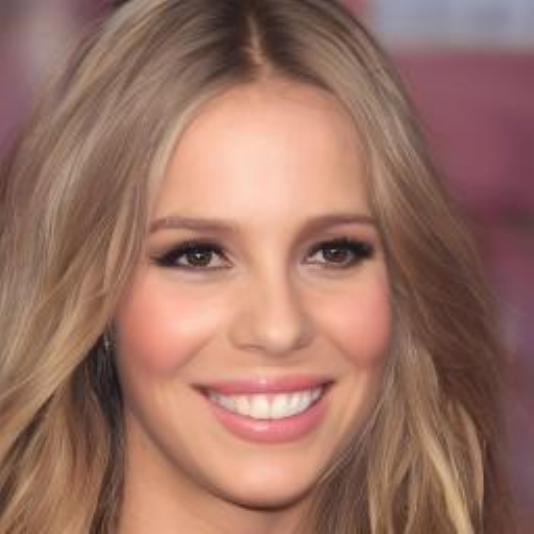}
\includegraphics[width=0.1\columnwidth]{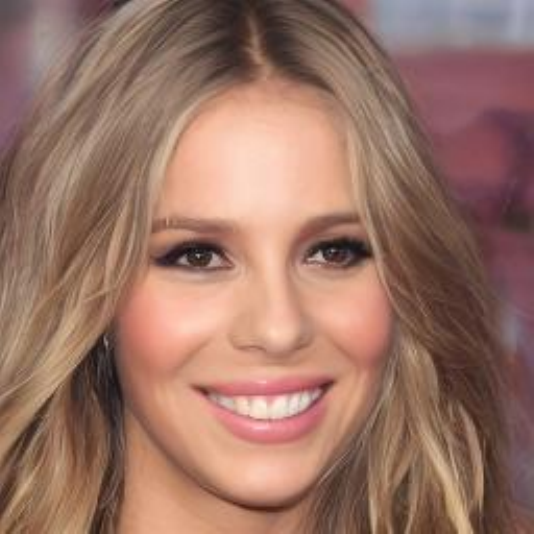}
\includegraphics[width=0.1\columnwidth]{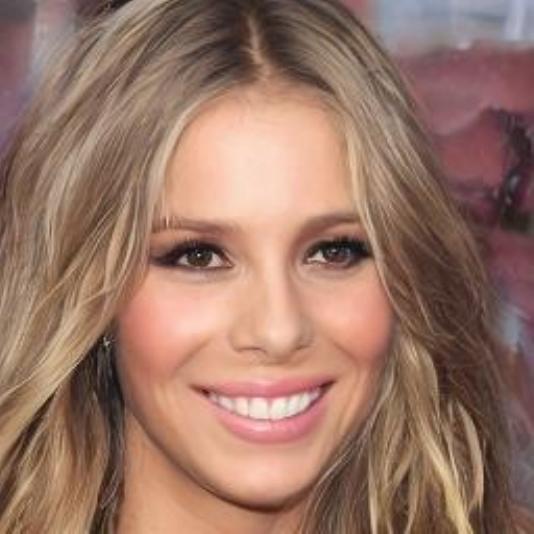}
\\
\includegraphics[width=0.1\columnwidth]{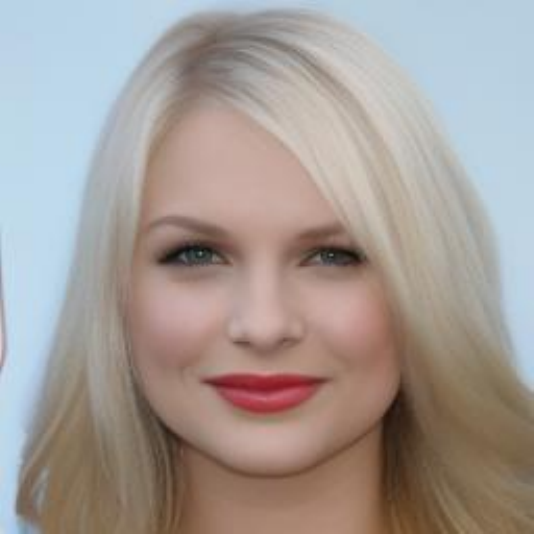}
\includegraphics[width=0.1\columnwidth]{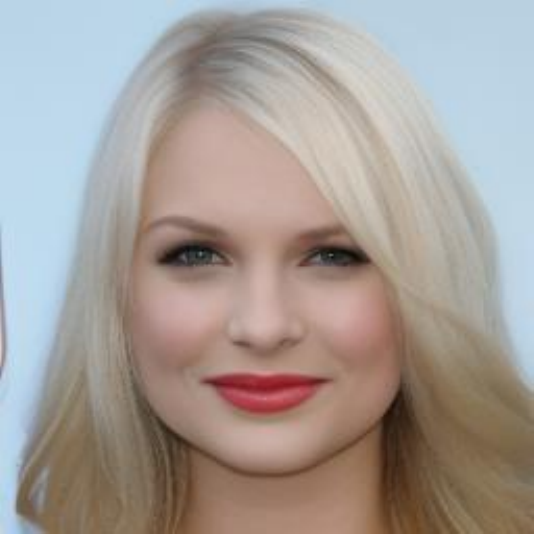}
\includegraphics[width=0.1\columnwidth]{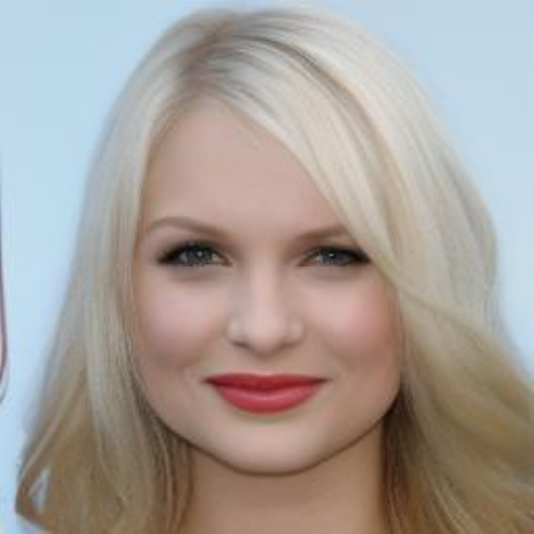}
\includegraphics[width=0.1\columnwidth]{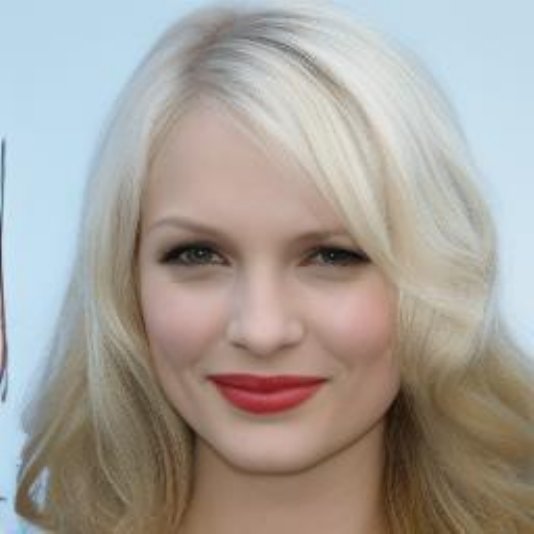}
\includegraphics[width=0.1\columnwidth]{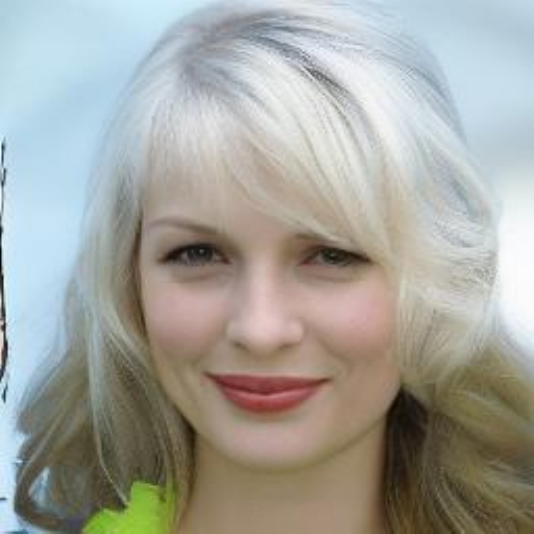}
\includegraphics[width=0.1\columnwidth]{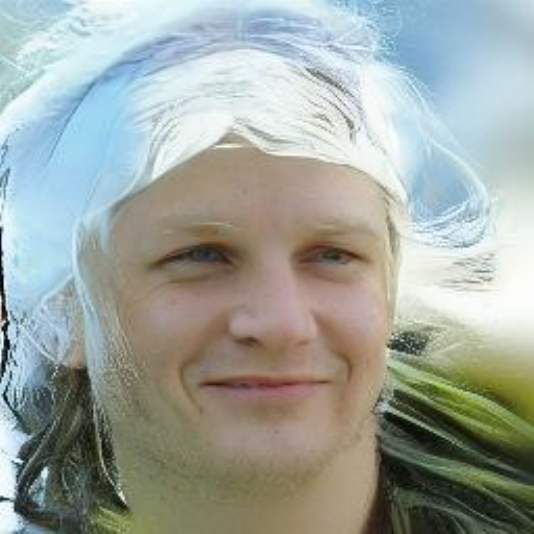}
\includegraphics[width=0.1\columnwidth]{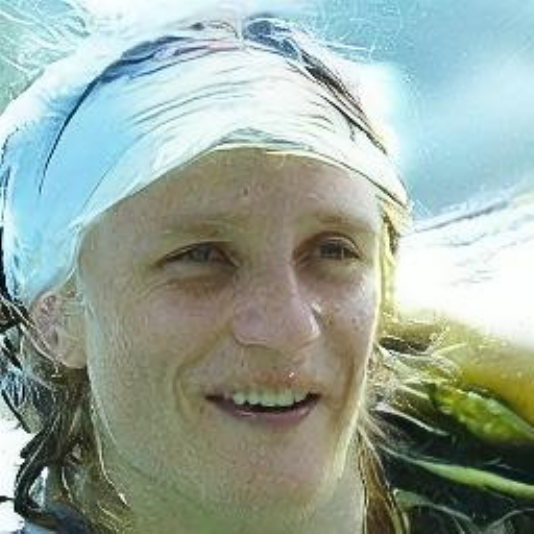}
\includegraphics[width=0.1\columnwidth]{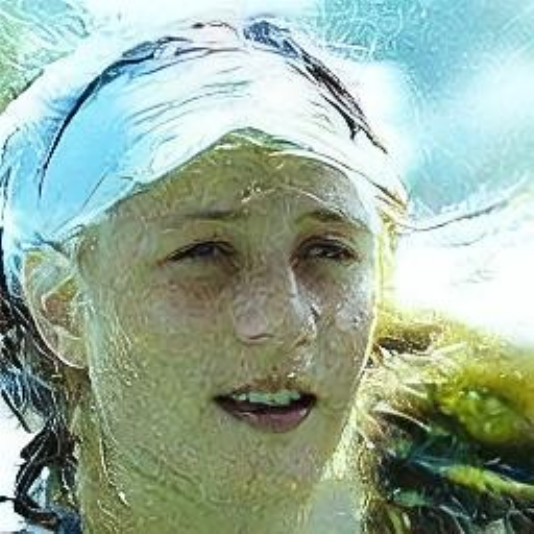}
\includegraphics[width=0.1\columnwidth]{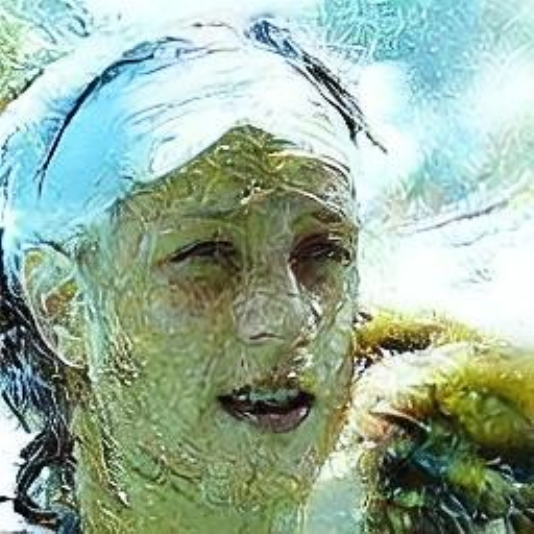}
\\
\includegraphics[width=0.1\columnwidth]{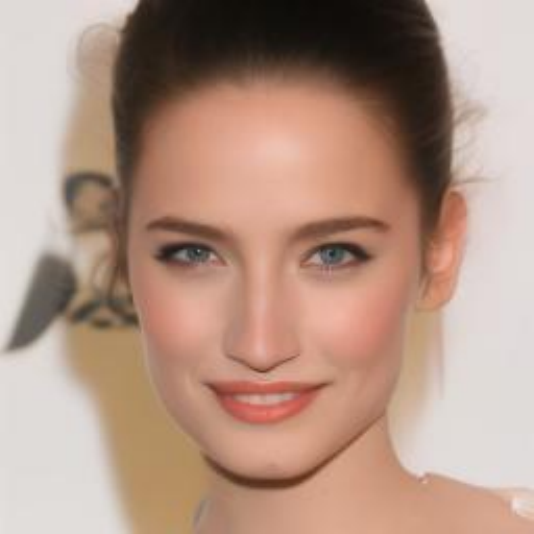}
\includegraphics[width=0.1\columnwidth]{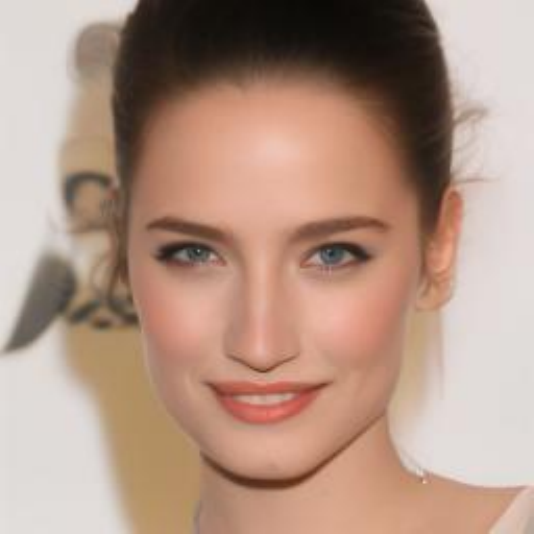}
\includegraphics[width=0.1\columnwidth]{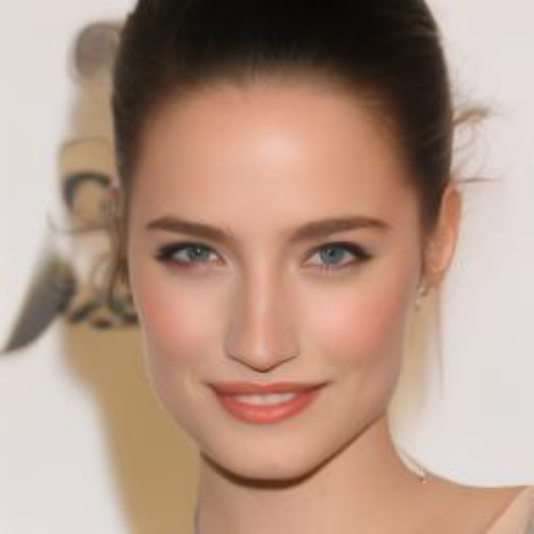}
\includegraphics[width=0.1\columnwidth]{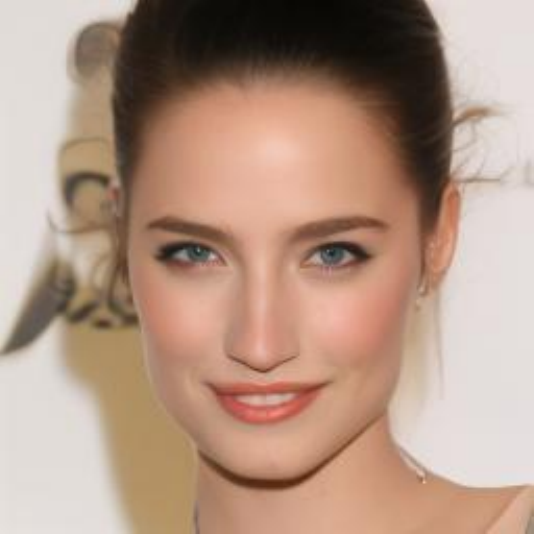}
\includegraphics[width=0.1\columnwidth]{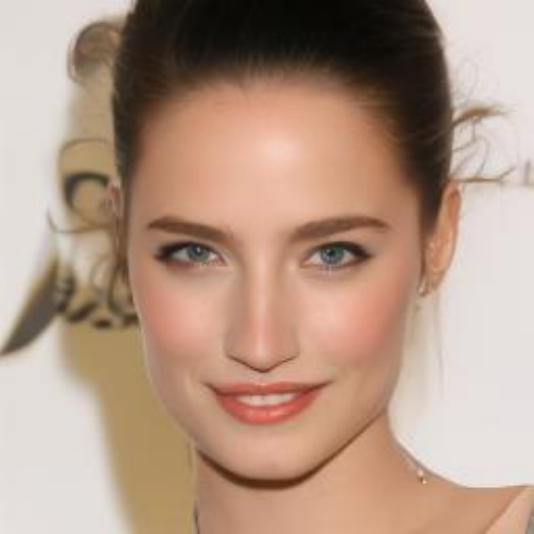}
\includegraphics[width=0.1\columnwidth]{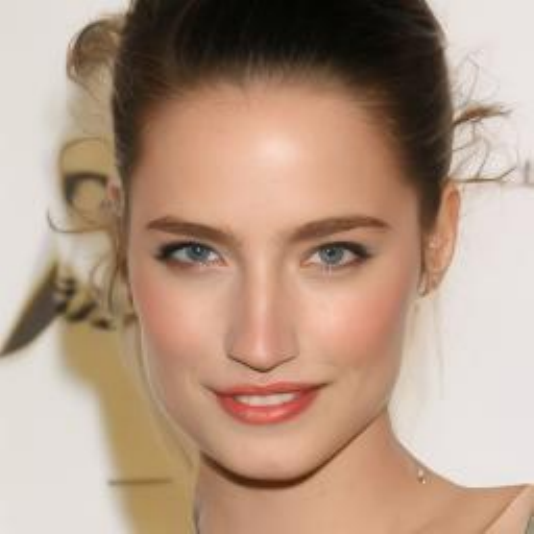}
\includegraphics[width=0.1\columnwidth]{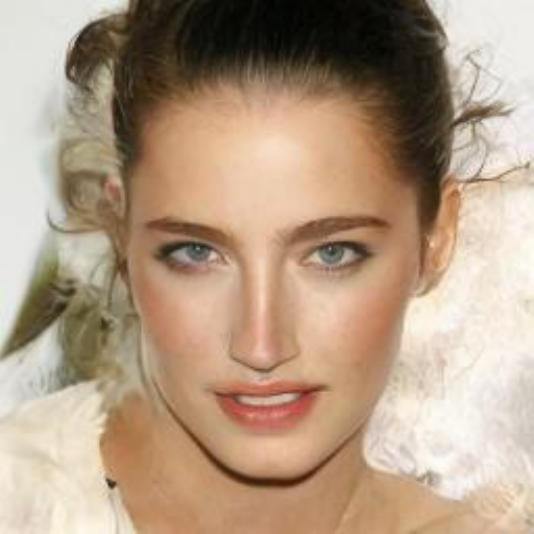}
\includegraphics[width=0.1\columnwidth]{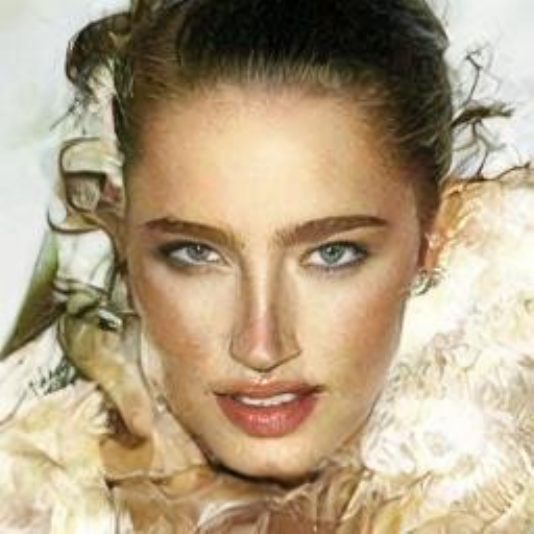}
\includegraphics[width=0.1\columnwidth]{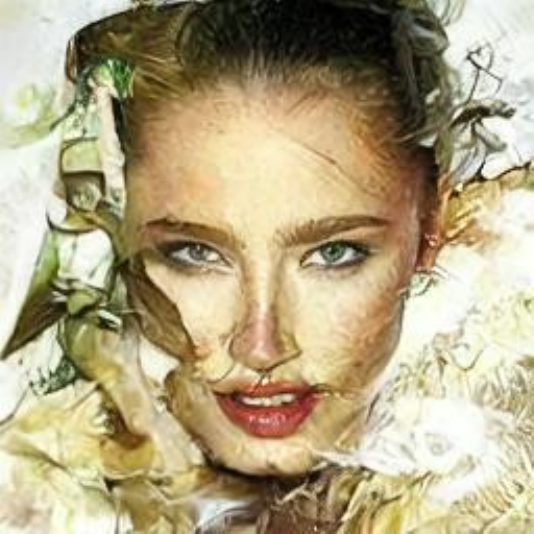}
\caption{\textbf{Varying the Variance Scaling for Low Numbers of Steps.
} For each row, the variance scaling changes from $0.40$ to $2.00$ in intervals of $0.20$.  \textbf{Rows 1-2:} LD model with $20$ steps. \textbf{Rows 3-4:} LD model with $40$ steps. 
%
%
}
    \label{fig:sweep-var}
    \end{figure}

\begin{figure}[t]
    \centering
\includegraphics[width=0.15\columnwidth]{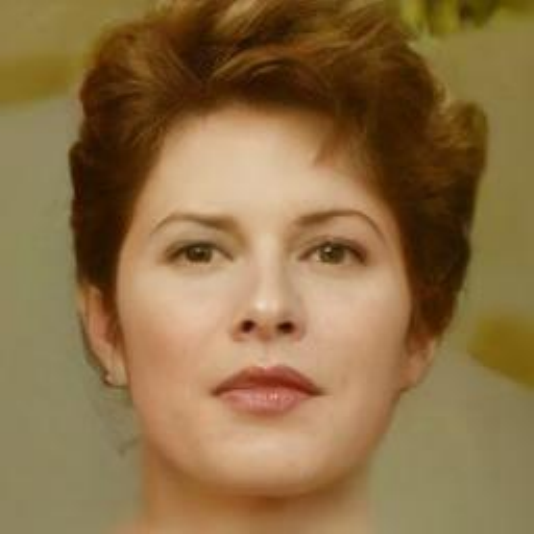}
\includegraphics[width=0.15\columnwidth]{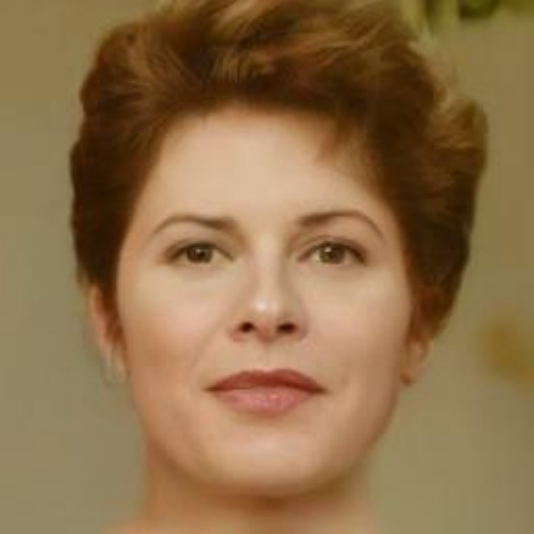}
\includegraphics[width=0.15\columnwidth]{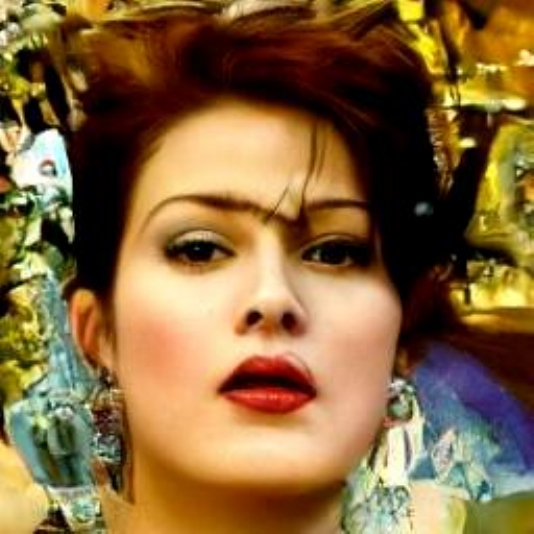}
\includegraphics[width=0.15\columnwidth]{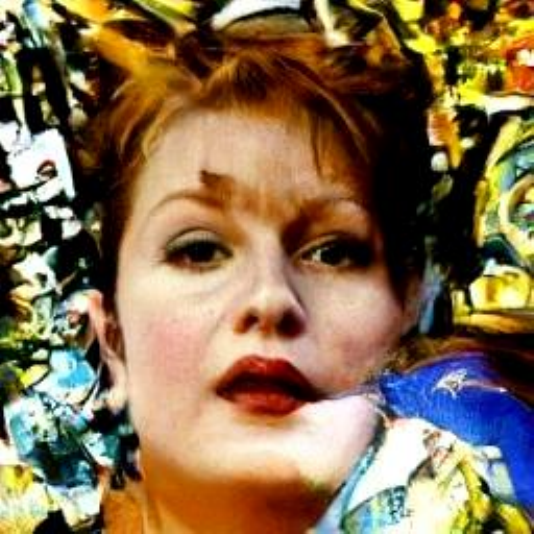}
\\
\includegraphics[width=0.15\columnwidth]{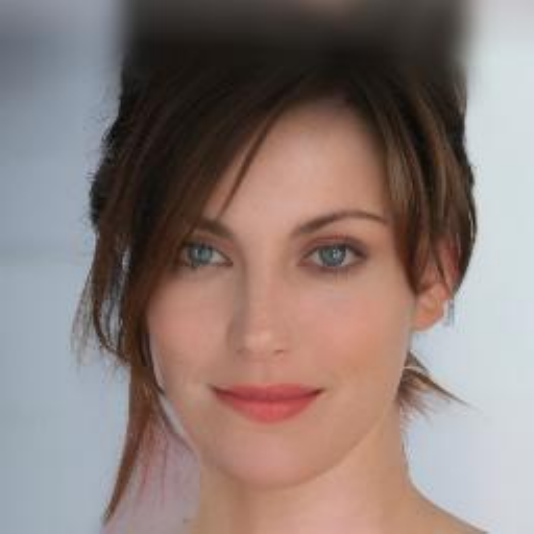}
\includegraphics[width=0.15\columnwidth]{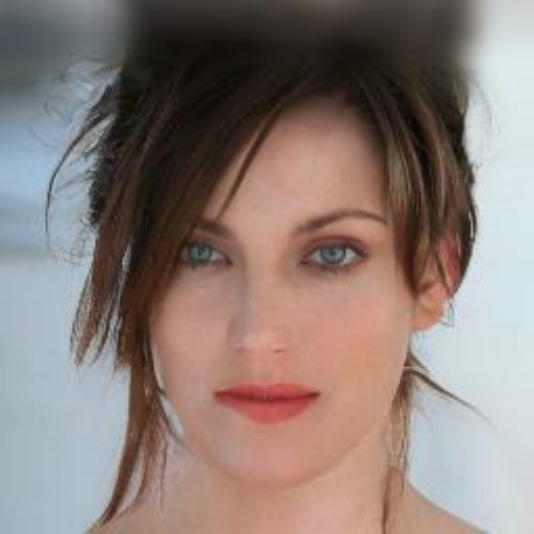}
\includegraphics[width=0.15\columnwidth]{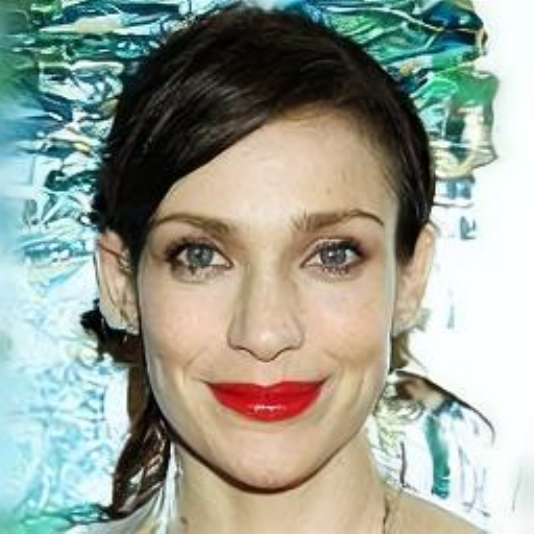}
\includegraphics[width=0.15\columnwidth]{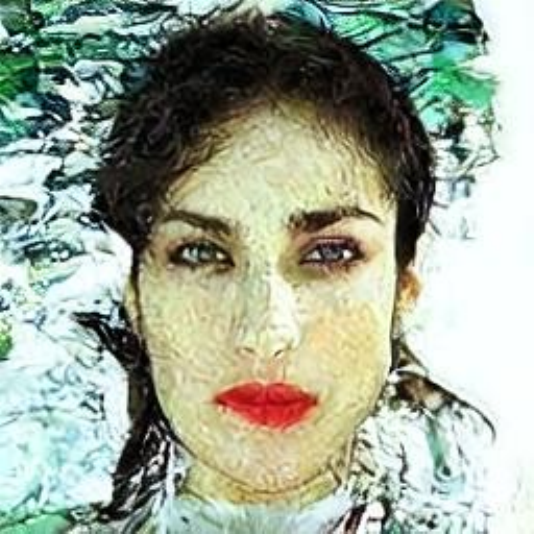}
\\
\includegraphics[width=0.15\columnwidth]{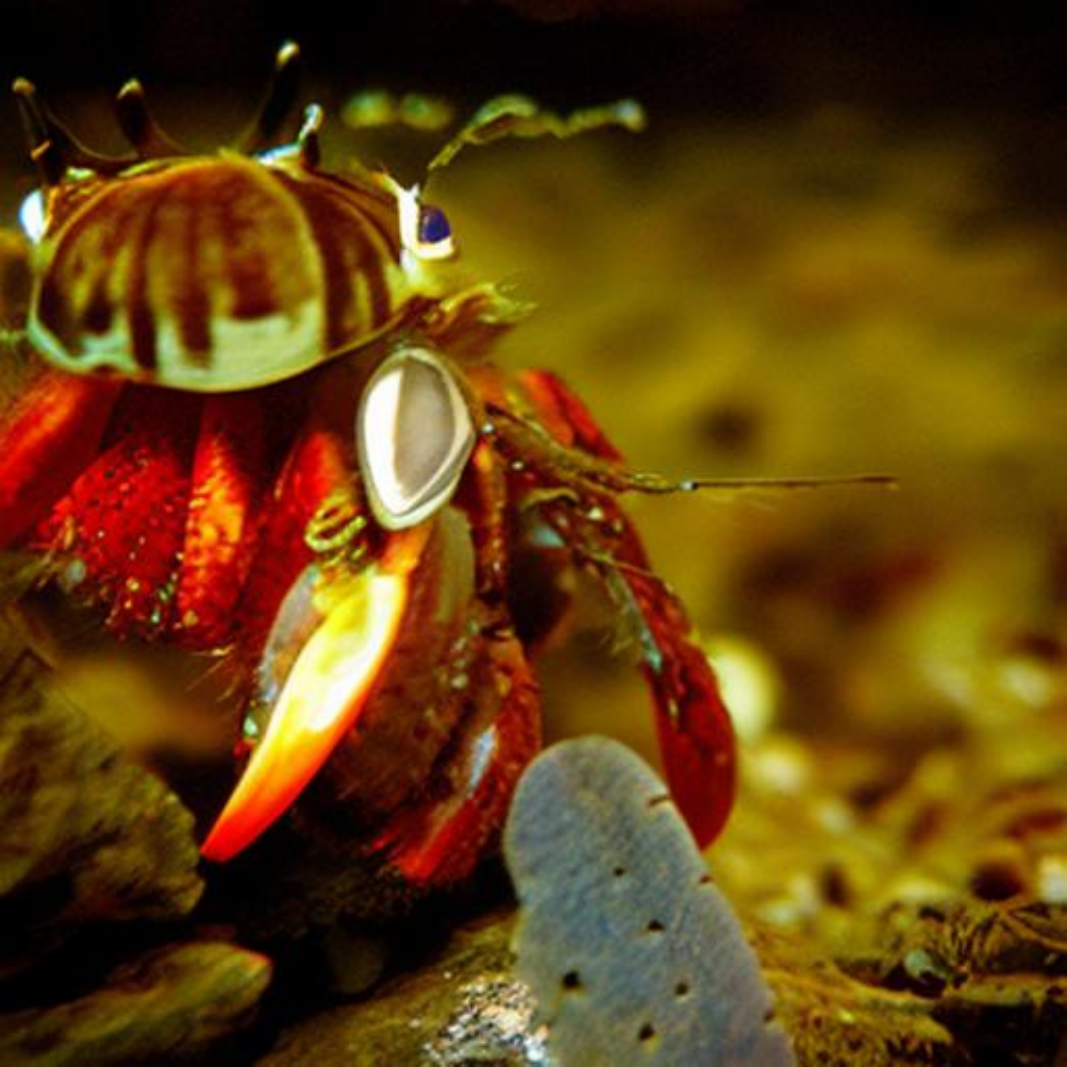}
\includegraphics[width=0.15\columnwidth]{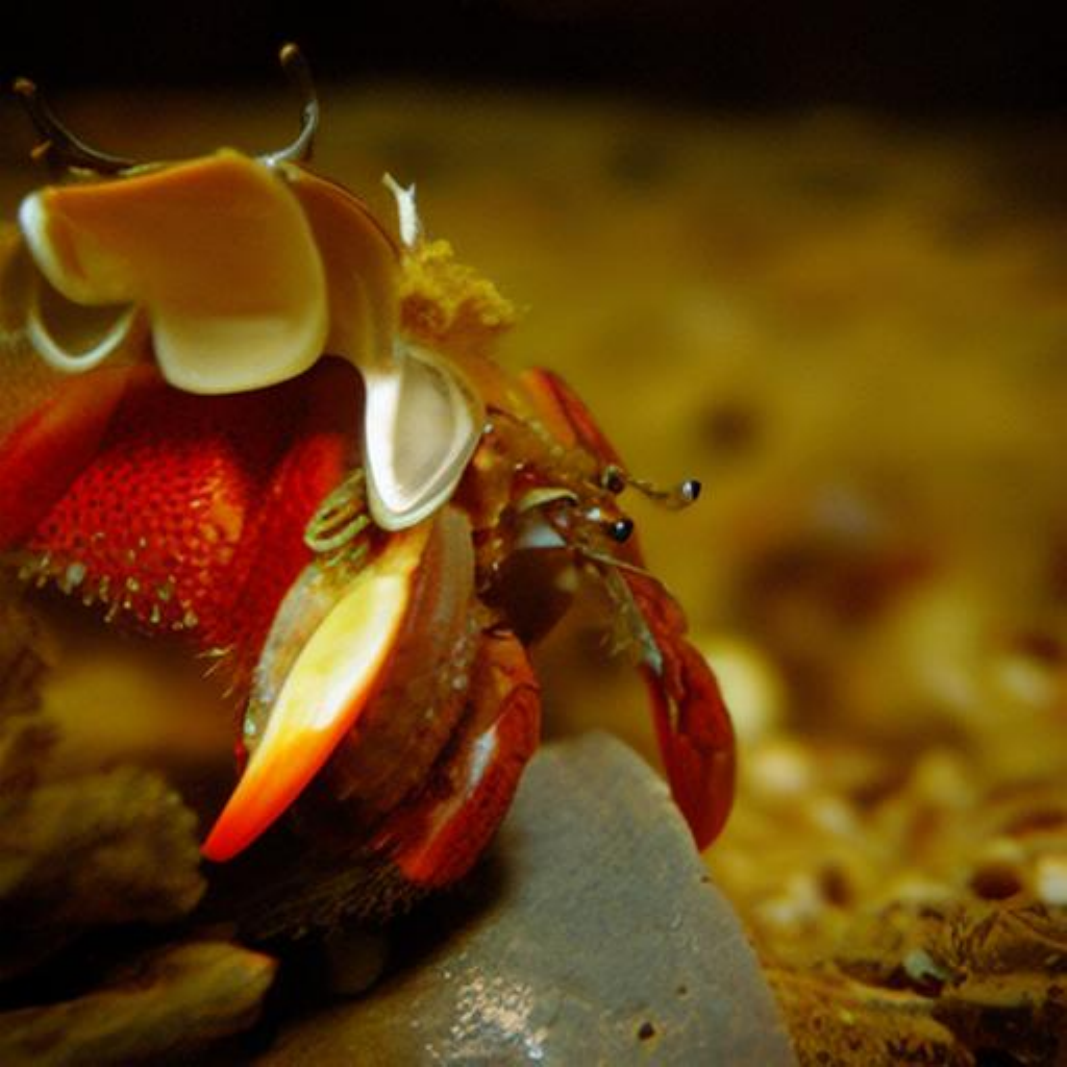}
\includegraphics[width=0.15\columnwidth]{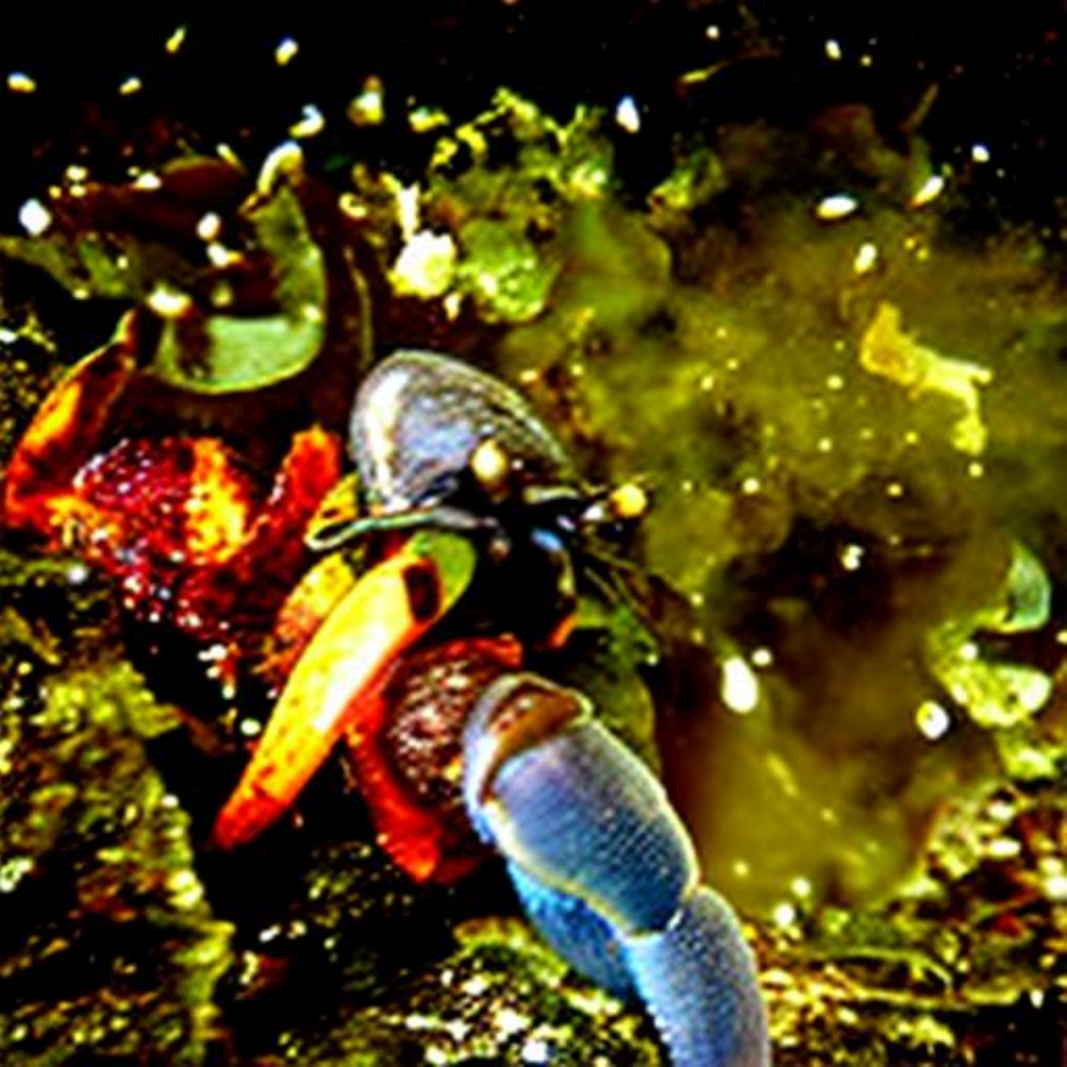}
\includegraphics[width=0.15\columnwidth]{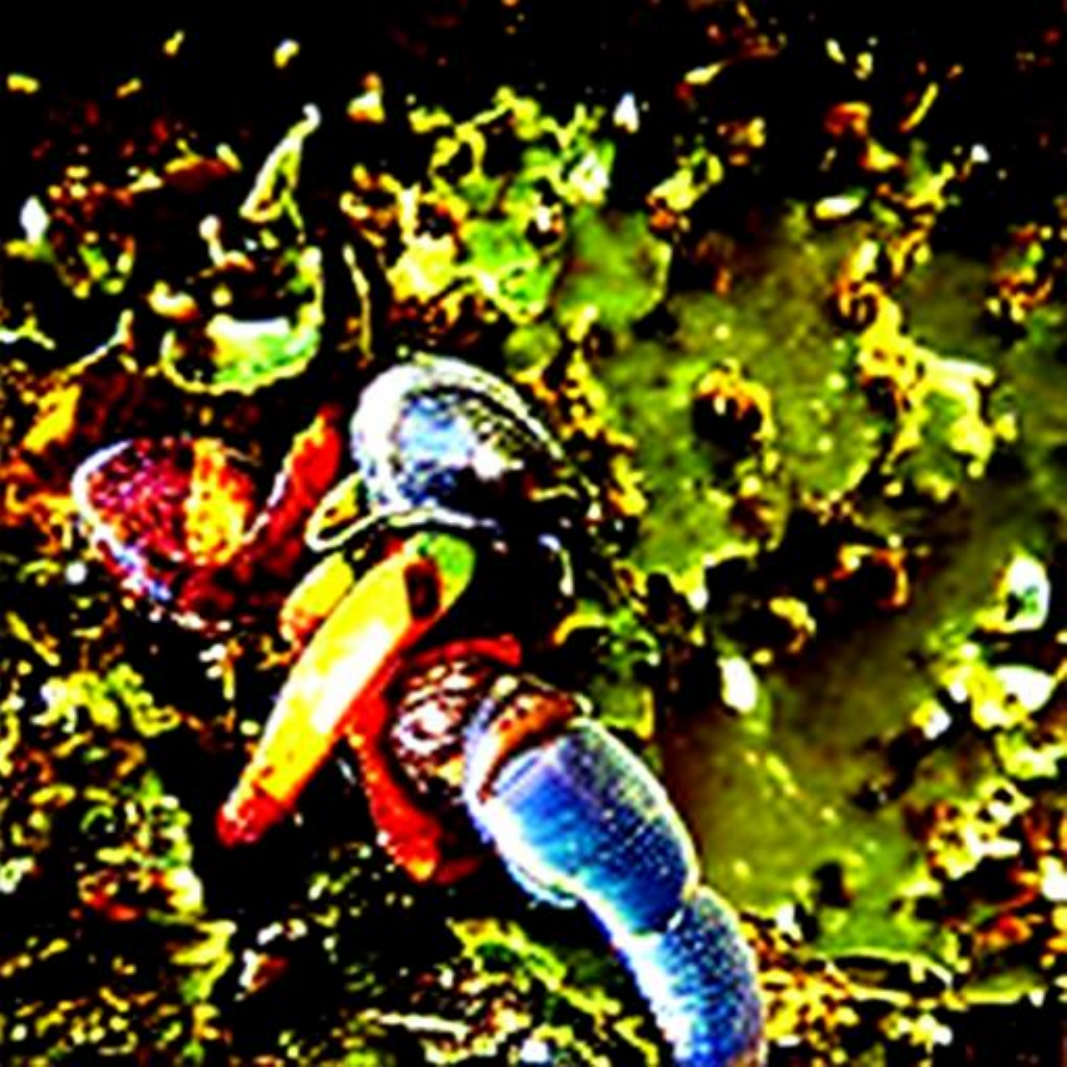}
\\
\includegraphics[width=0.15\columnwidth]{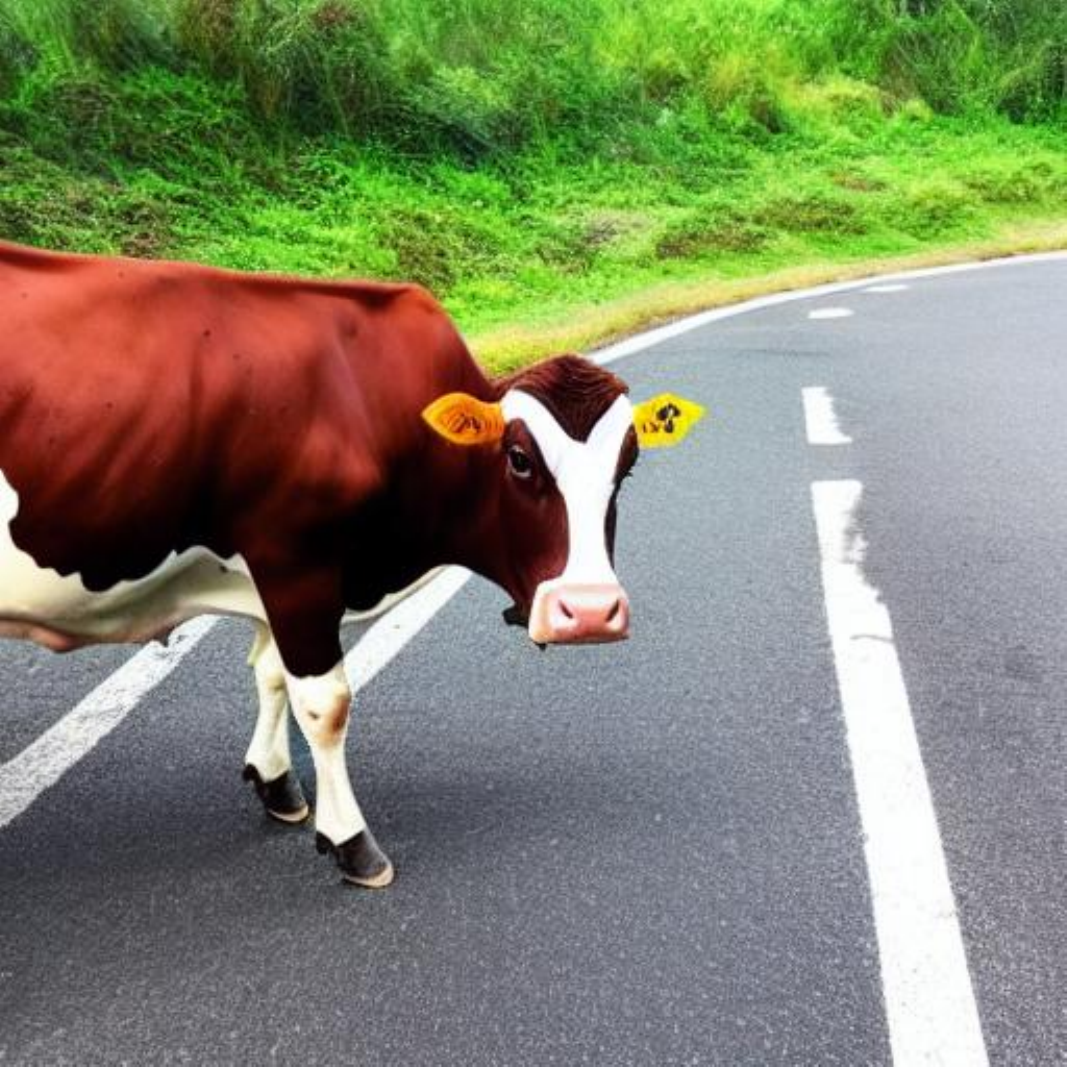}
\includegraphics[width=0.15\columnwidth]{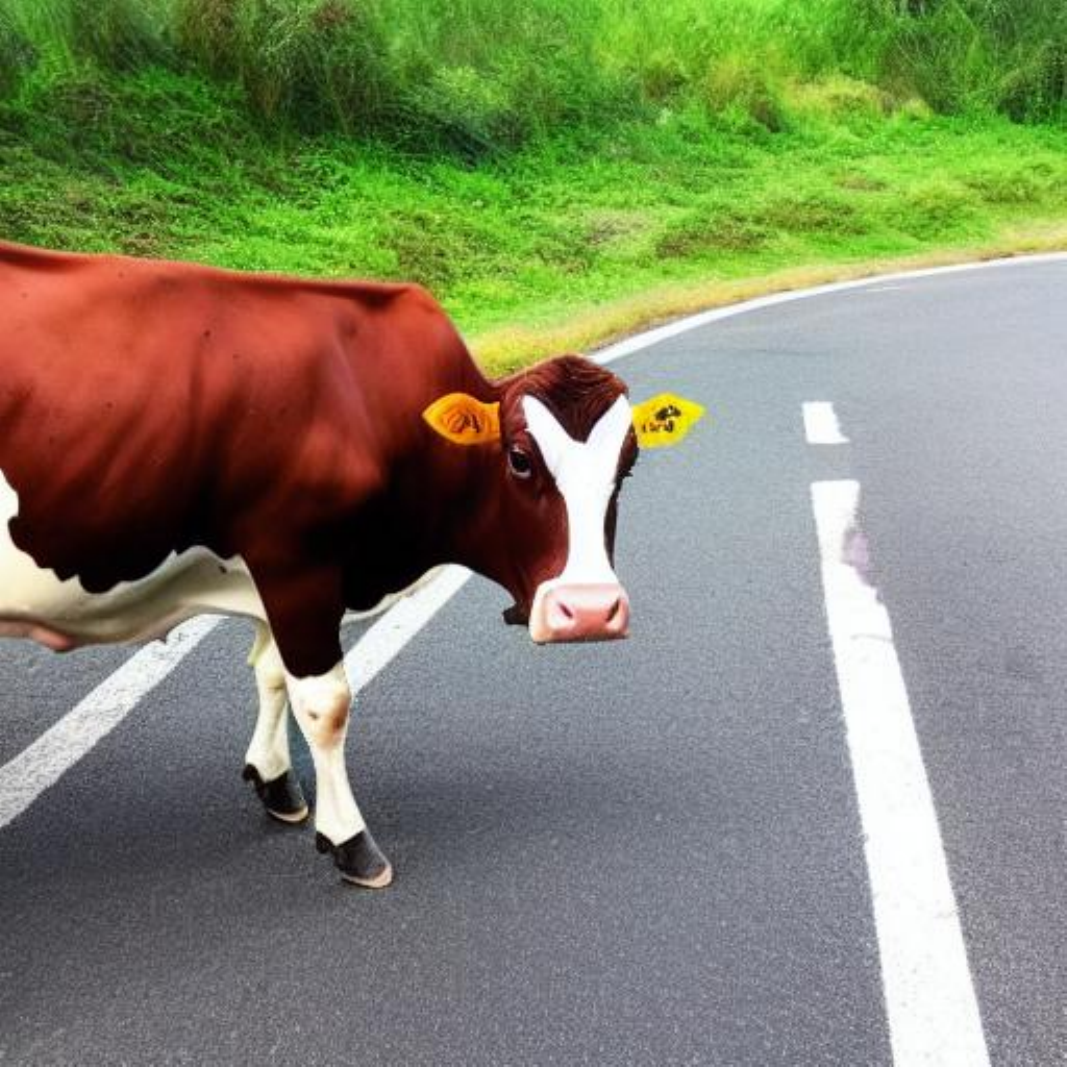}
\includegraphics[width=0.15\columnwidth]{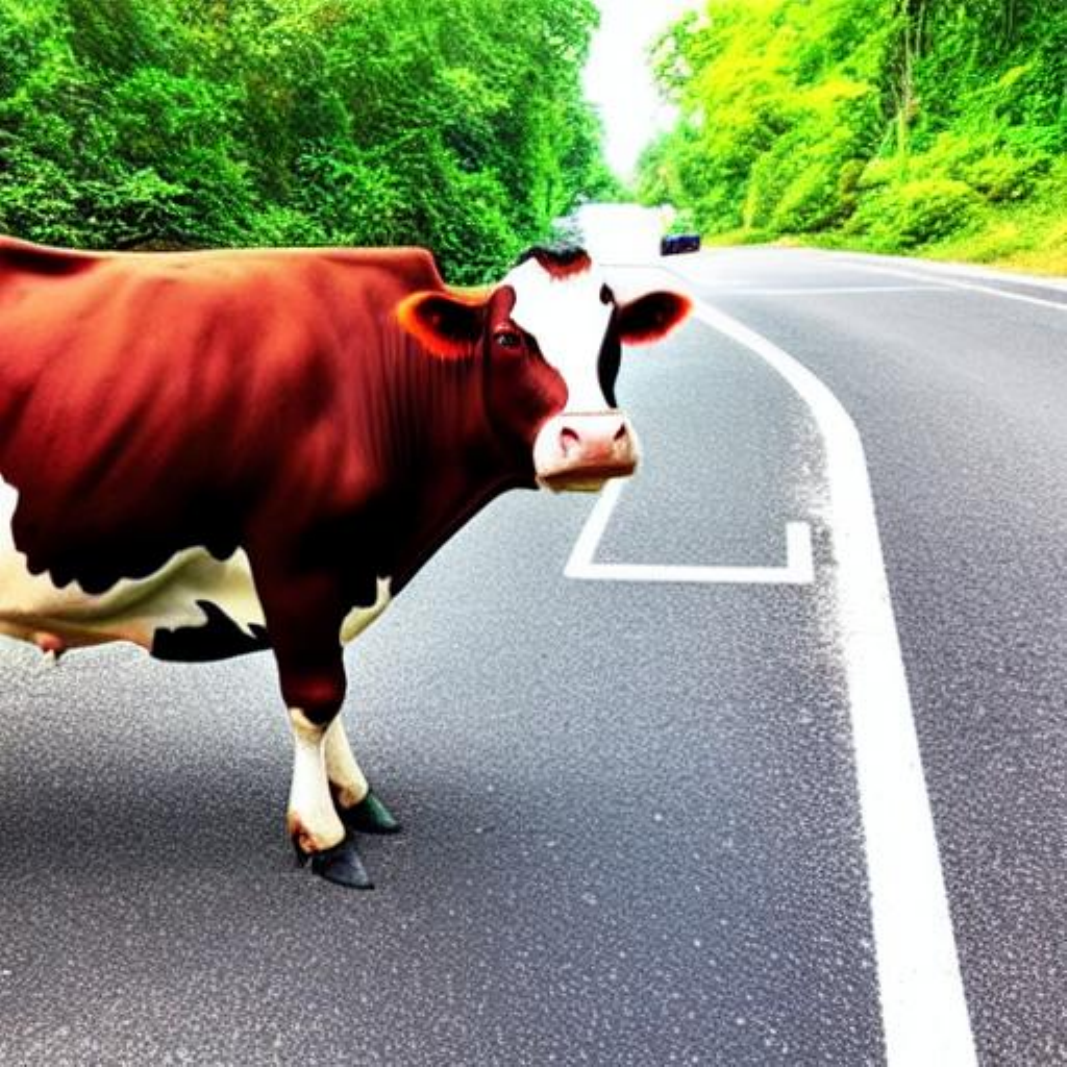}
\includegraphics[width=0.15\columnwidth]{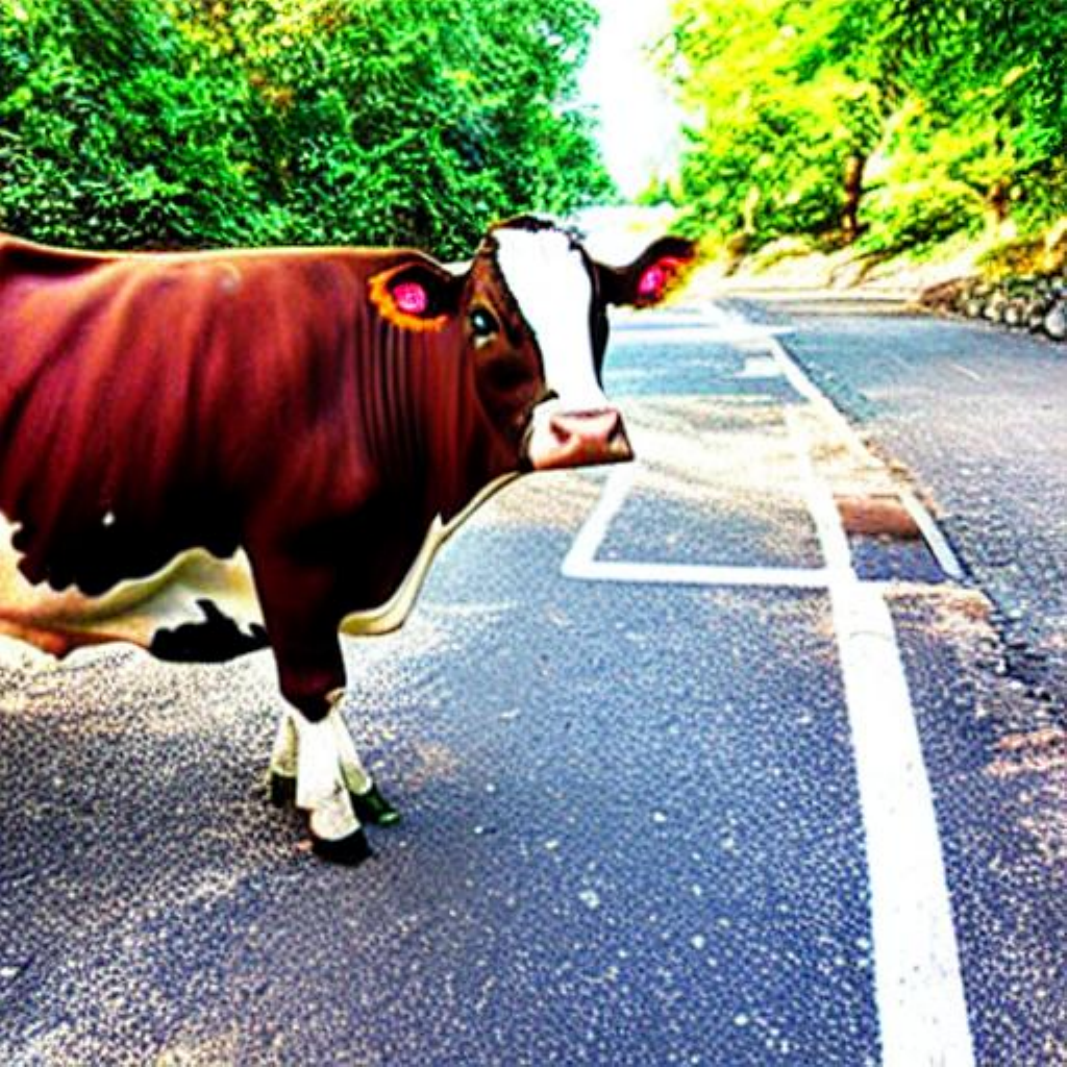}
%
%
\caption{\textbf{Large Variance Scaling Leads to Image Quality Loss in a Large Number of Steps.} We consider $100$ steps. For each row, from left to right: baseline, ablation, generated image with variance scaling $\rho_1$, and one with variance scaling $\rho_2$. 
\textbf{Rows 1-3:} DDPM, LD, and DiT (image class is \emph{hermit crab}), respectively, with $\rho_1=1.35$ and $\rho_2=1.55$. 
\textbf{Row 4:} SD with $\rho_1=1.25$ and $\rho_2=1.55$. 
Notice that these values of variance scaling are used to improve sampled image quality in lower number of steps.
%
%
%
}
    \label{fig:sc-var-wr}
    \end{figure}

\begin{figure}[ht!]
    \centering
\includegraphics[width=0.13\columnwidth]{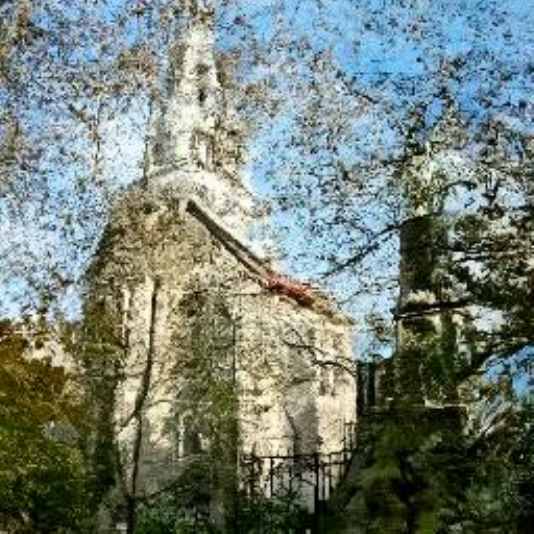}
\includegraphics[width=0.13\columnwidth]{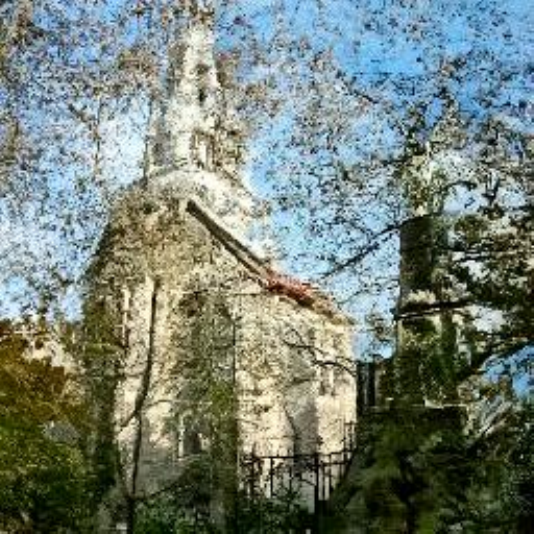}
\includegraphics[width=0.13\columnwidth]{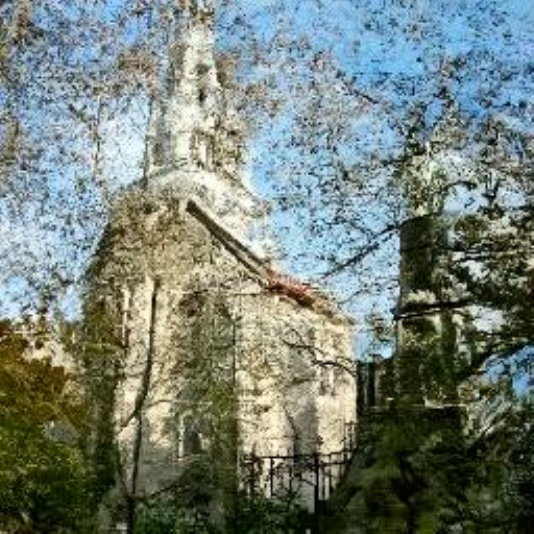}
\includegraphics[width=0.13\columnwidth]{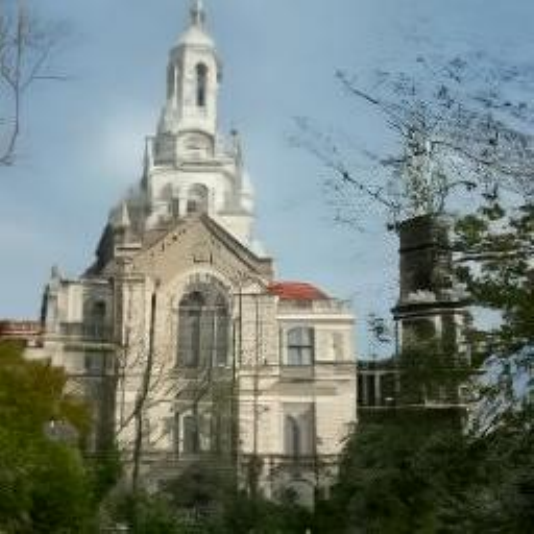}
\includegraphics[width=0.13\columnwidth]{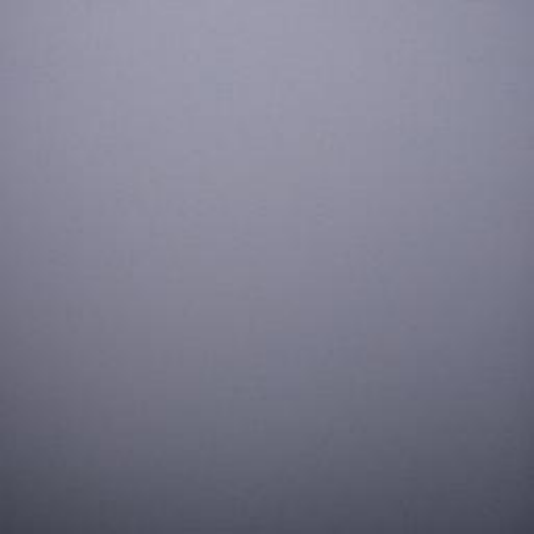}
\includegraphics[width=0.13\columnwidth]{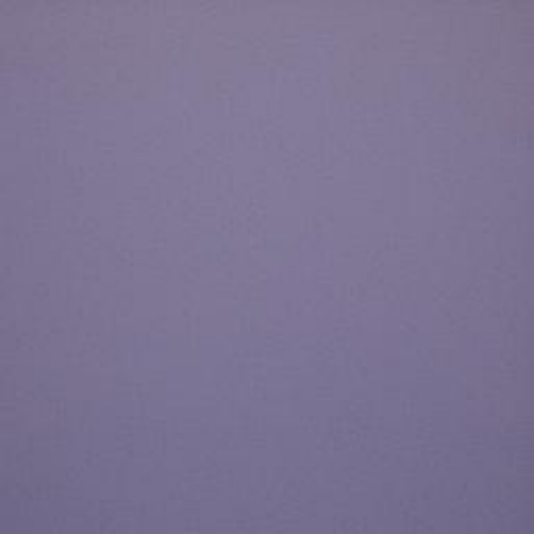}
\includegraphics[width=0.13\columnwidth]{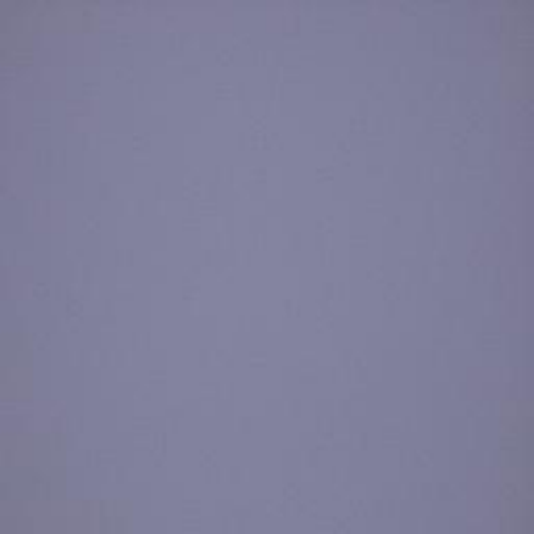}
\\
\includegraphics[width=0.13\columnwidth]{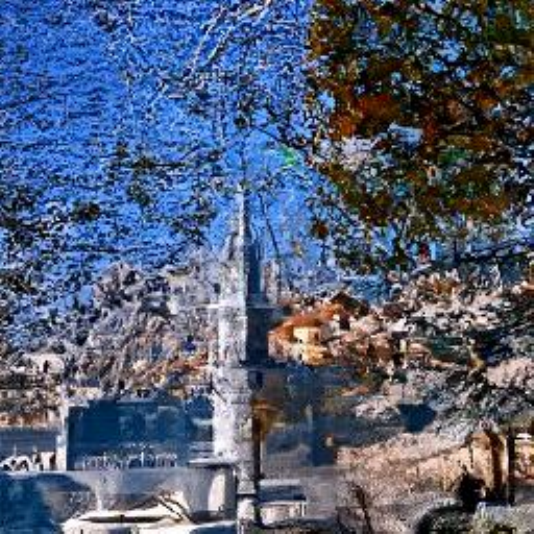}
\includegraphics[width=0.13\columnwidth]{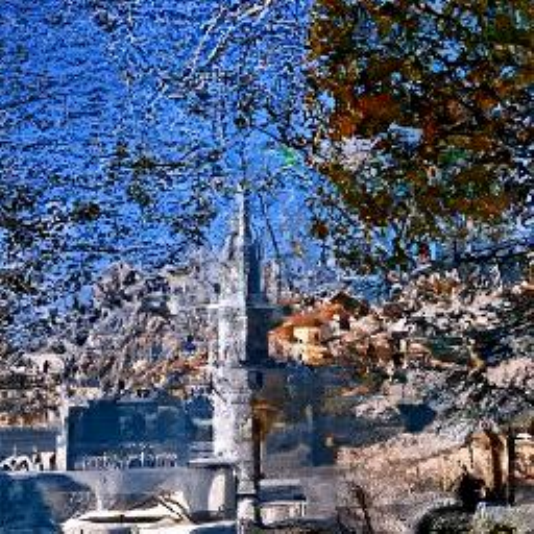}
\includegraphics[width=0.13\columnwidth]{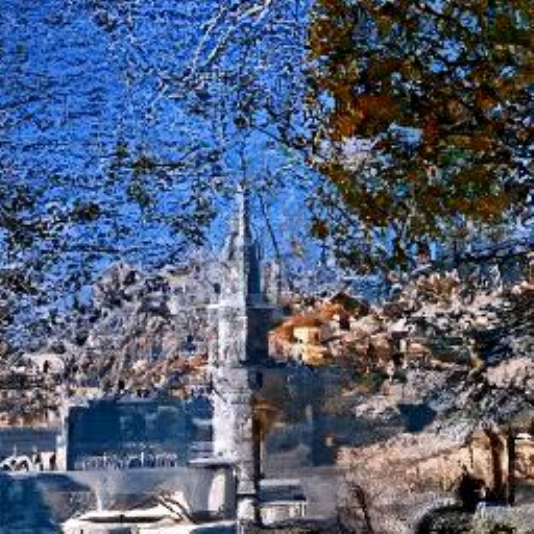}
\includegraphics[width=0.13\columnwidth]{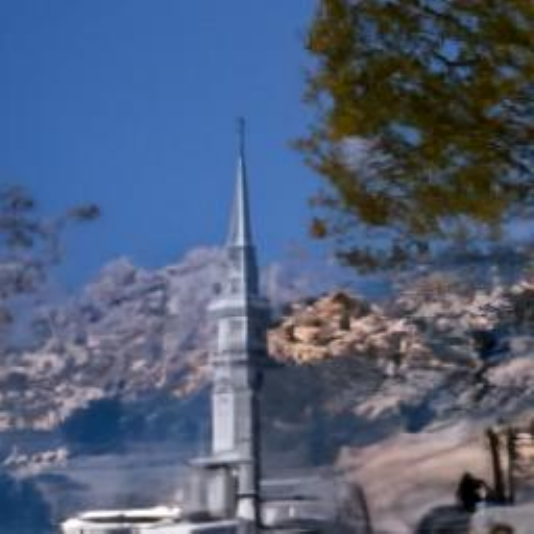}
\includegraphics[width=0.13\columnwidth]{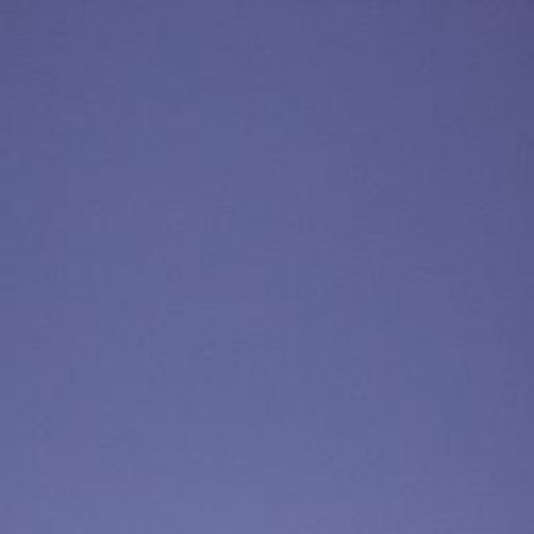}
\includegraphics[width=0.13\columnwidth]{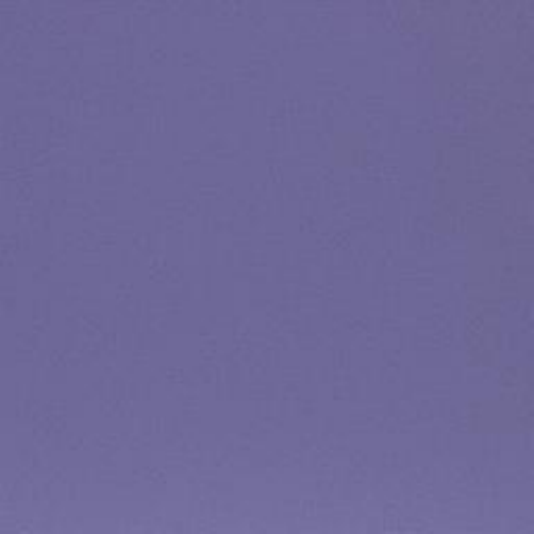}
\includegraphics[width=0.13\columnwidth]{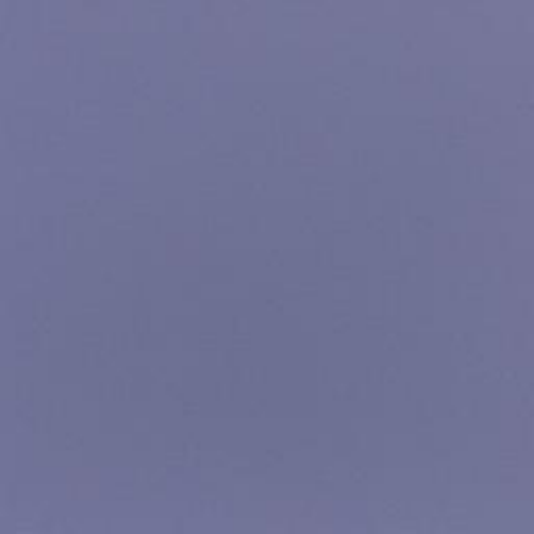}
\\
\includegraphics[width=0.13\columnwidth]{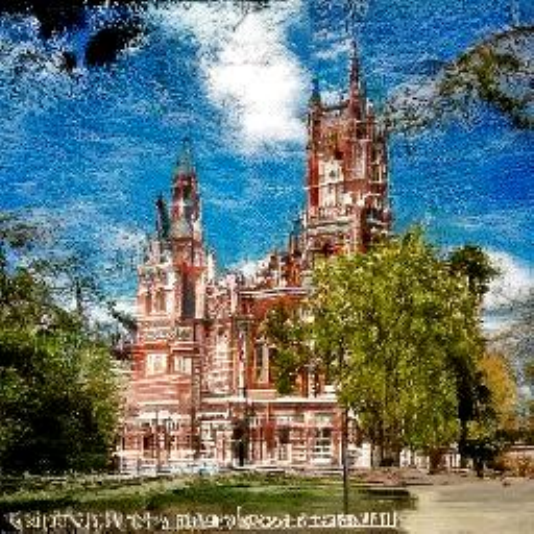}
\includegraphics[width=0.13\columnwidth]{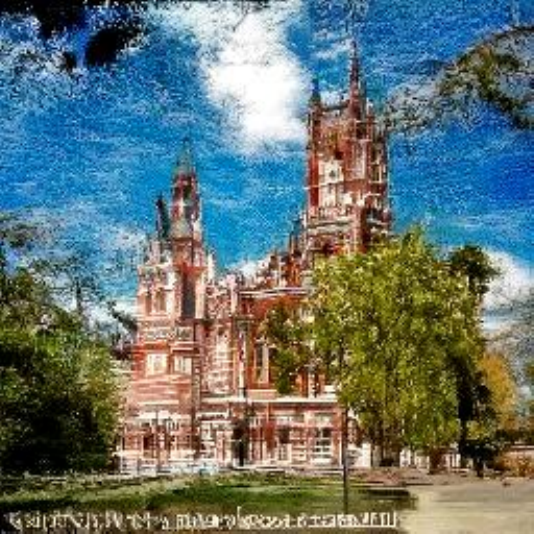}
\includegraphics[width=0.13\columnwidth]{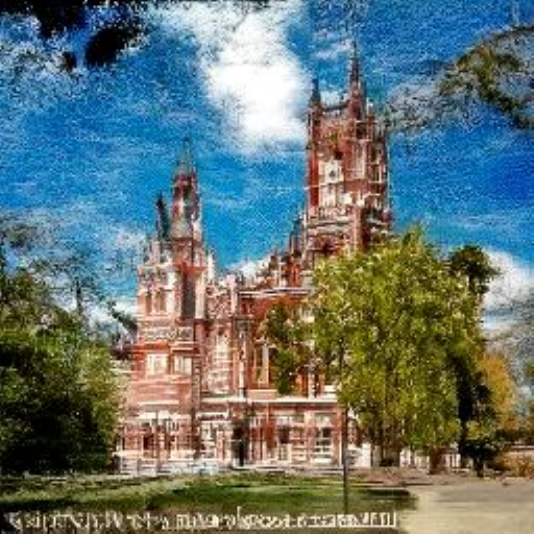}
\includegraphics[width=0.13\columnwidth]{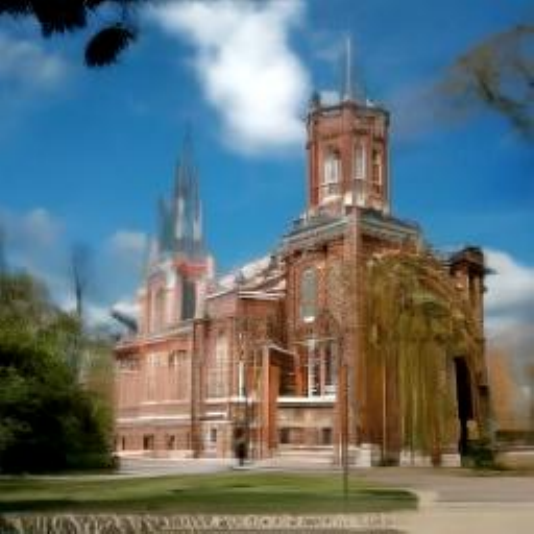}
\includegraphics[width=0.13\columnwidth]{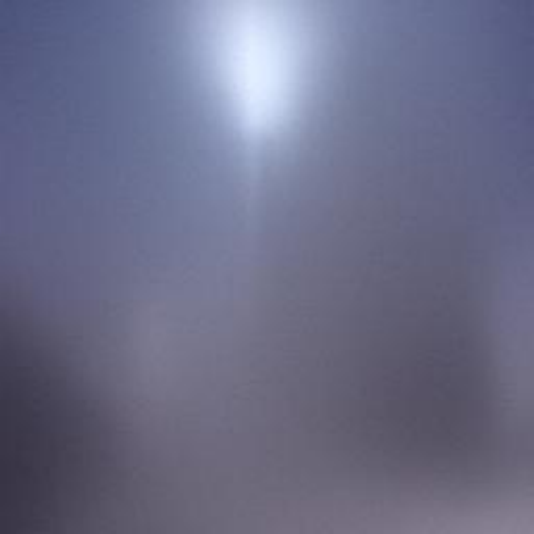}
\includegraphics[width=0.13\columnwidth]{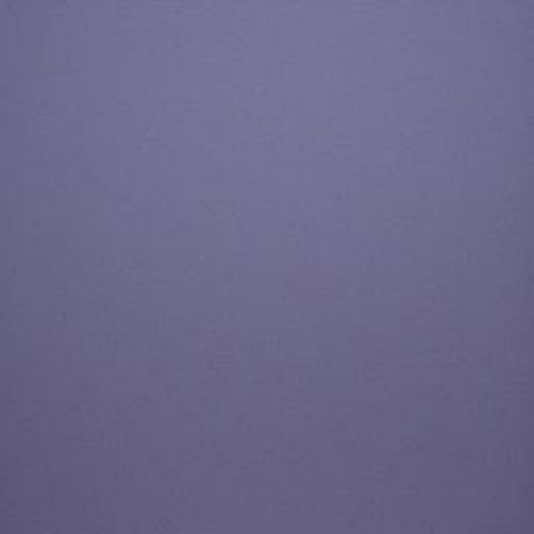}
\includegraphics[width=0.13\columnwidth]{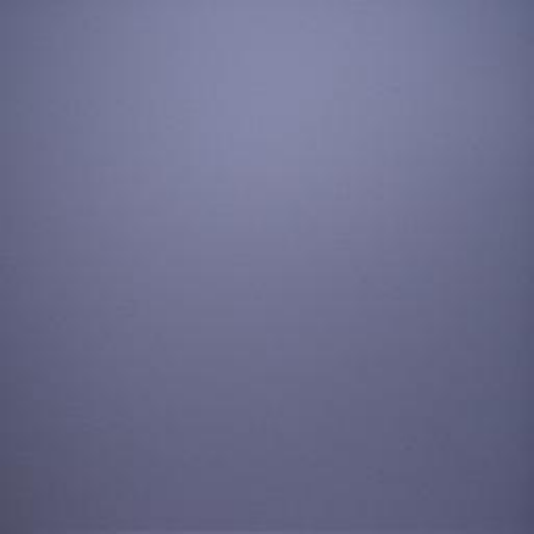}
\\
\includegraphics[width=0.13\columnwidth]{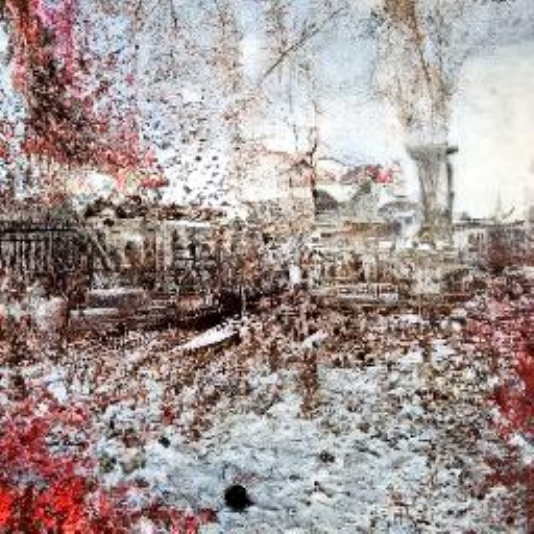}
\includegraphics[width=0.13\columnwidth]{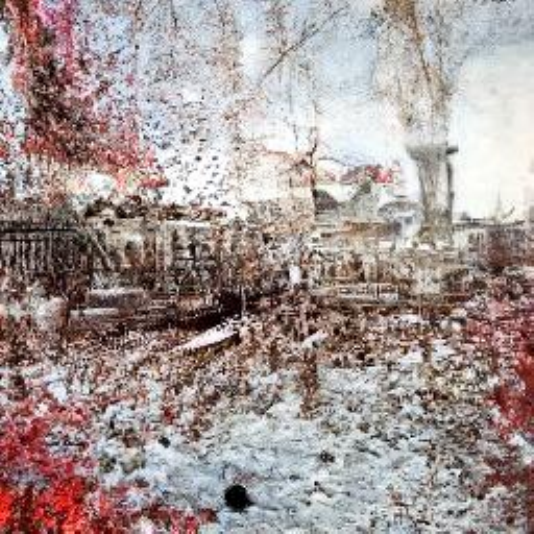}
\includegraphics[width=0.13\columnwidth]{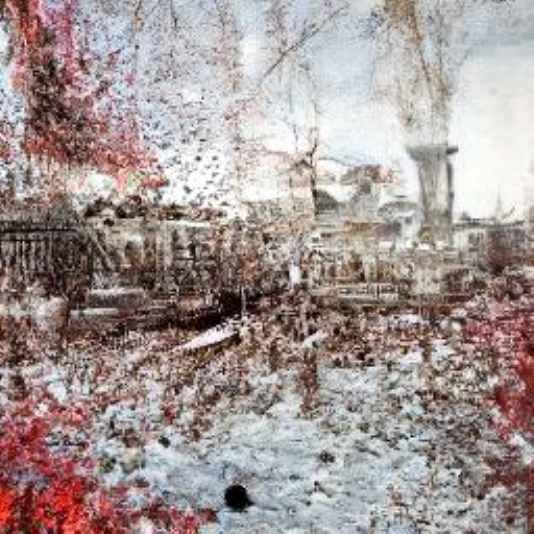}
\includegraphics[width=0.13\columnwidth]{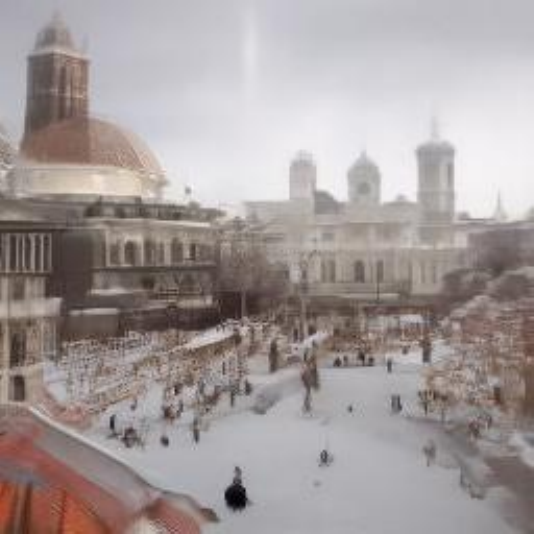}
\includegraphics[width=0.13\columnwidth]{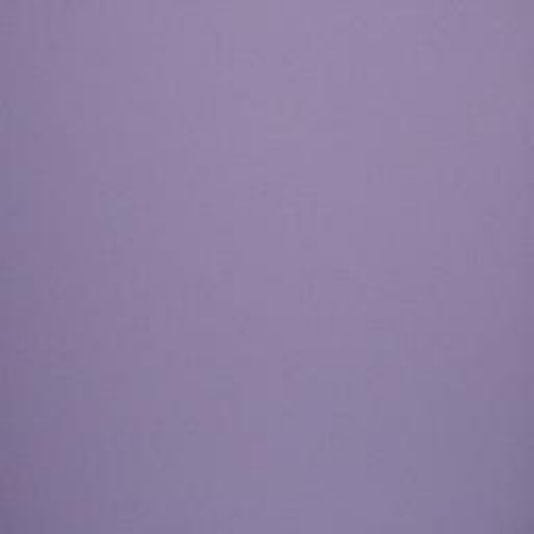}
\includegraphics[width=0.13\columnwidth]{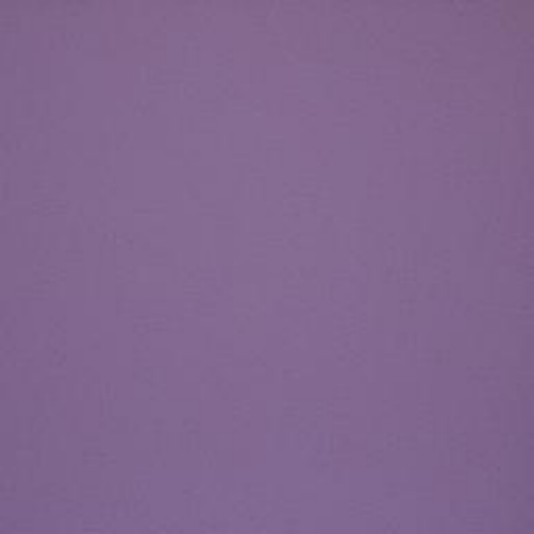}
\includegraphics[width=0.13\columnwidth]{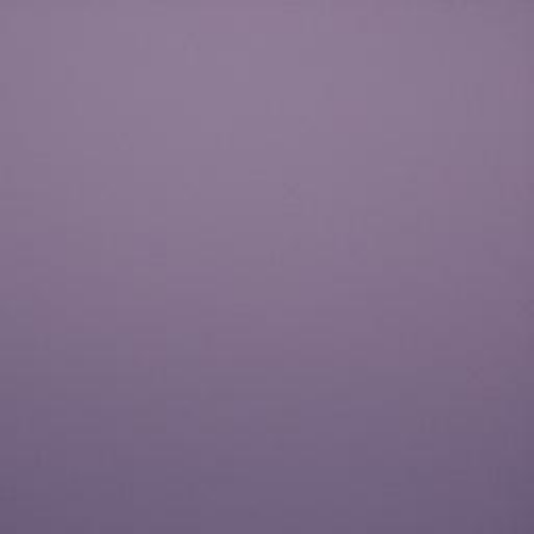}
\\
\includegraphics[width=0.13\columnwidth]{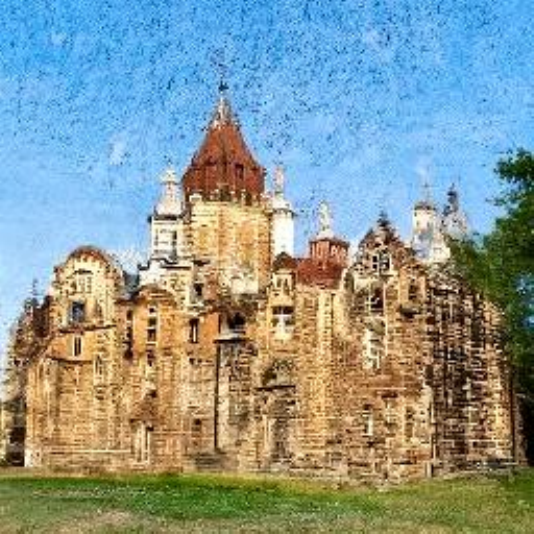}
\includegraphics[width=0.13\columnwidth]{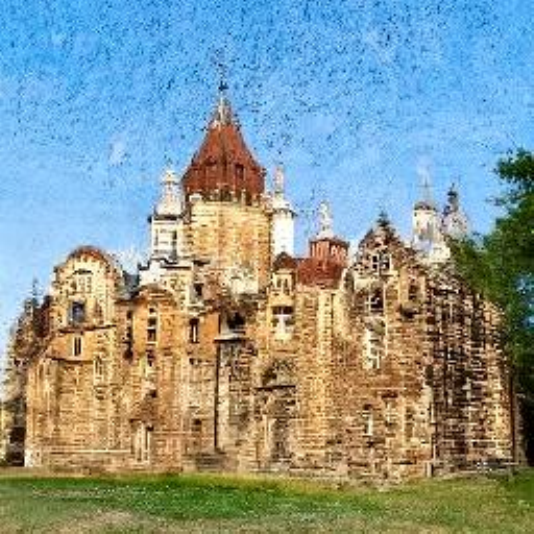}
\includegraphics[width=0.13\columnwidth]{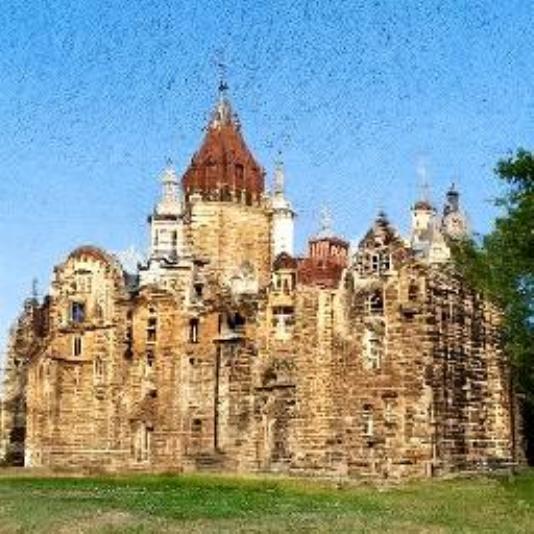}
\includegraphics[width=0.13\columnwidth]{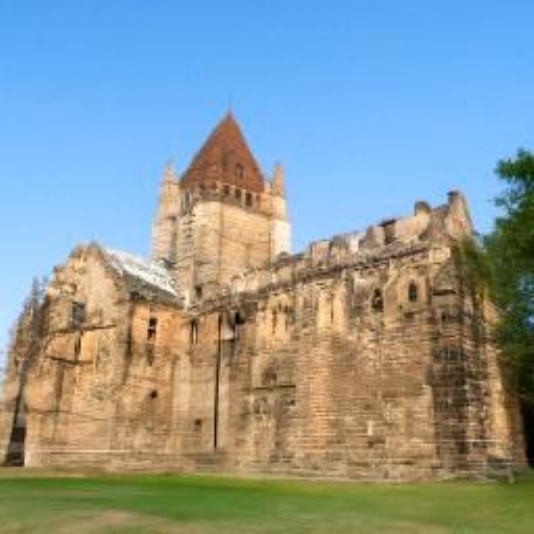}
\includegraphics[width=0.13\columnwidth]{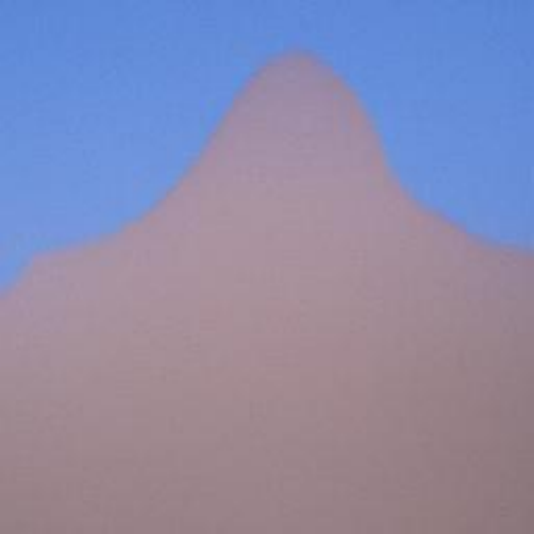}
\includegraphics[width=0.13\columnwidth]{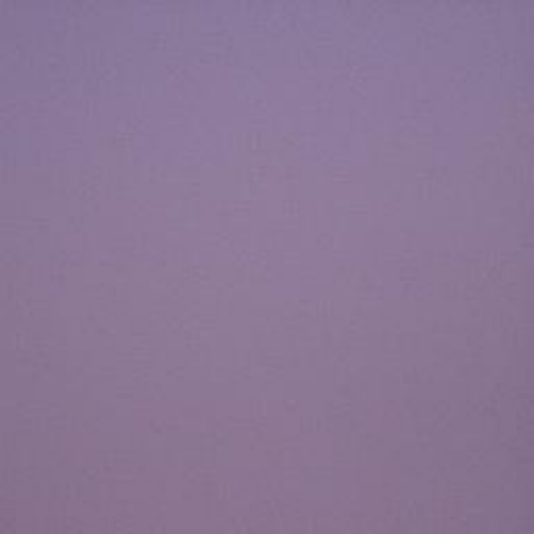}
\includegraphics[width=0.13\columnwidth]{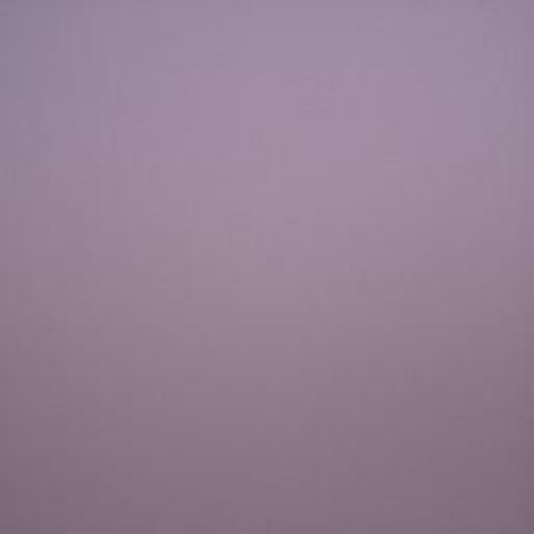}
\\
\includegraphics[width=0.13\columnwidth]{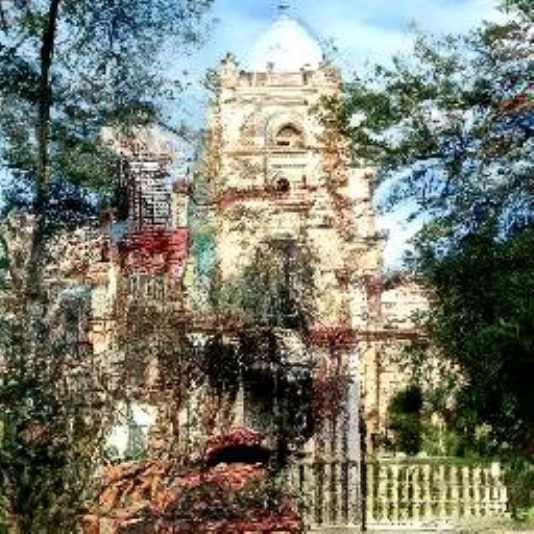}
\includegraphics[width=0.13\columnwidth]{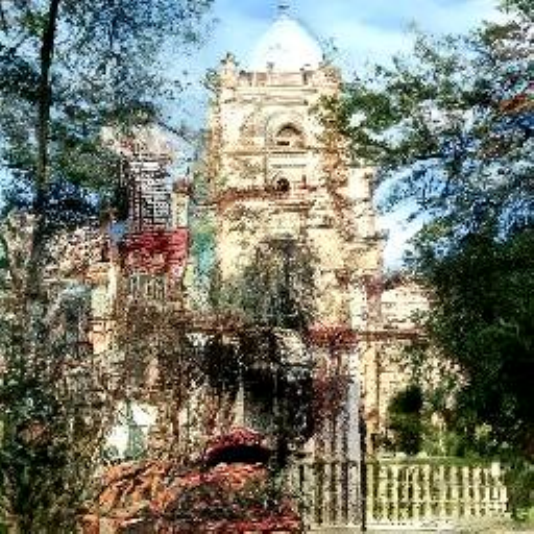}
\includegraphics[width=0.13\columnwidth]{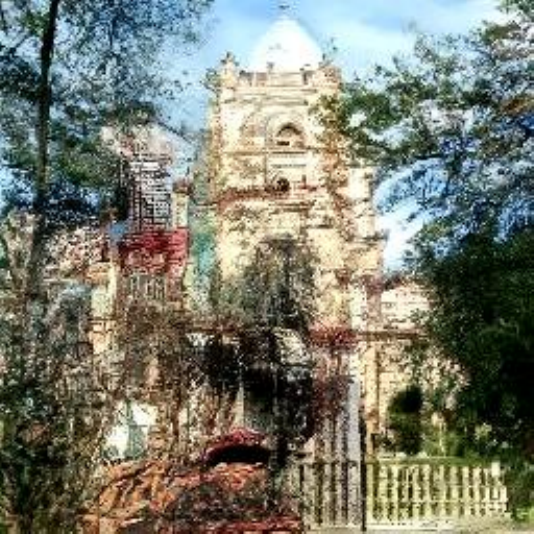}
\includegraphics[width=0.13\columnwidth]{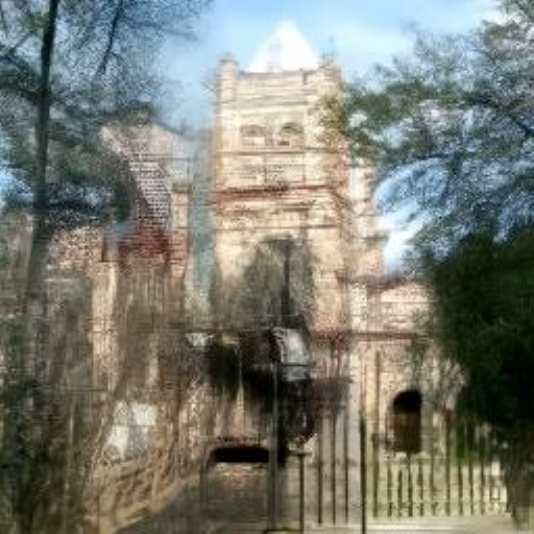}
\includegraphics[width=0.13\columnwidth]{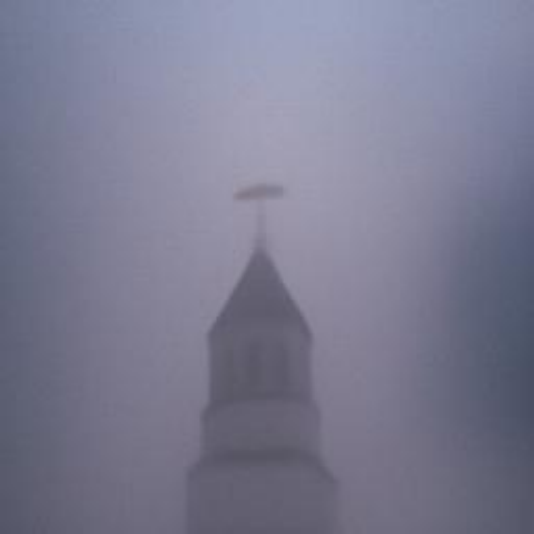}
\includegraphics[width=0.13\columnwidth]{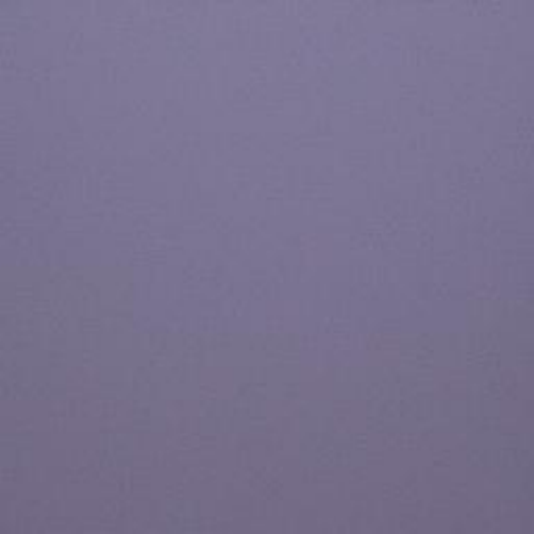}
\includegraphics[width=0.13\columnwidth]{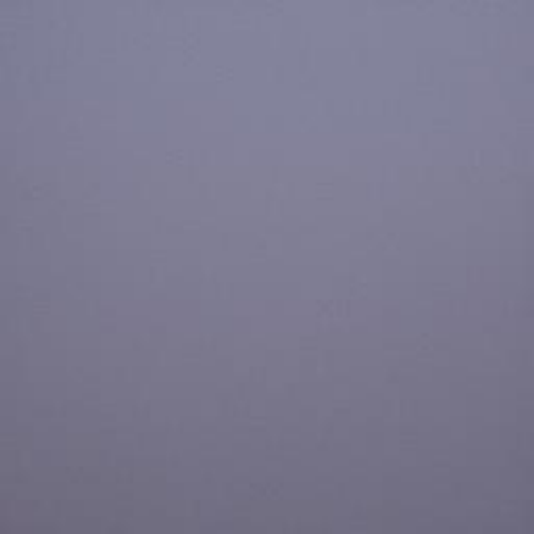}
\\
\includegraphics[width=0.13\columnwidth]{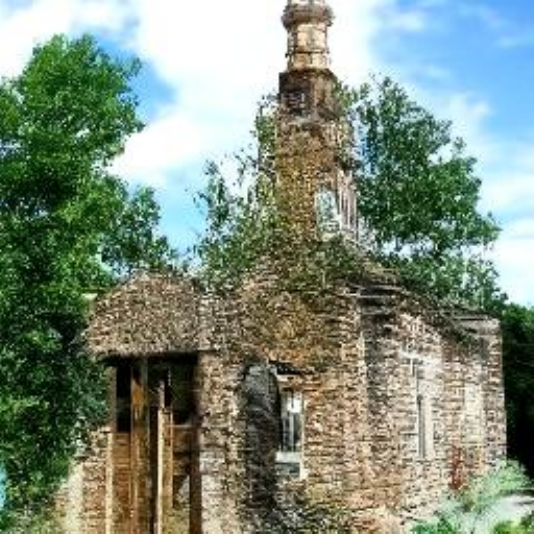}
\includegraphics[width=0.13\columnwidth]{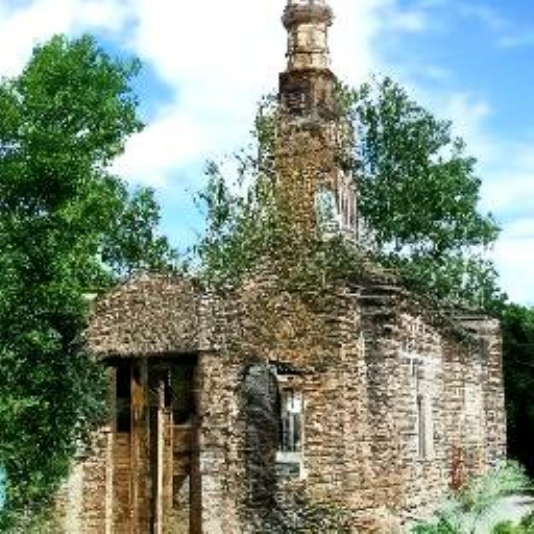}
\includegraphics[width=0.13\columnwidth]{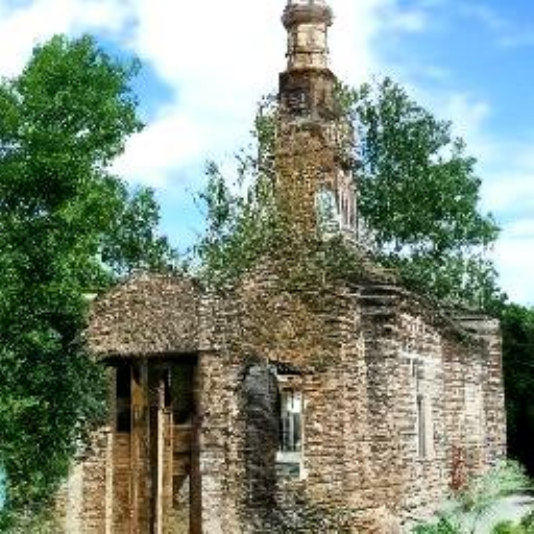}
\includegraphics[width=0.13\columnwidth]{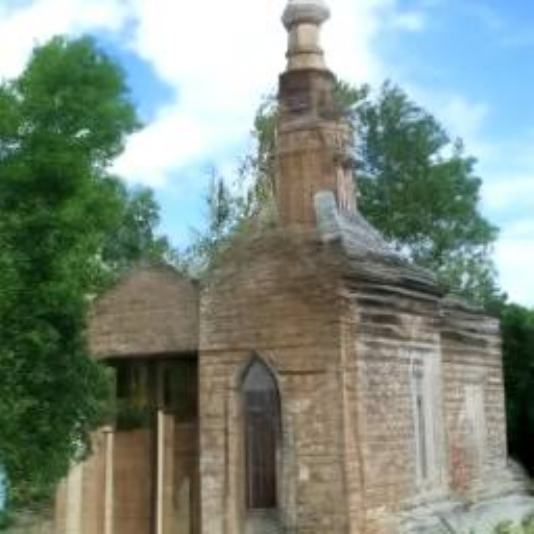}
\includegraphics[width=0.13\columnwidth]{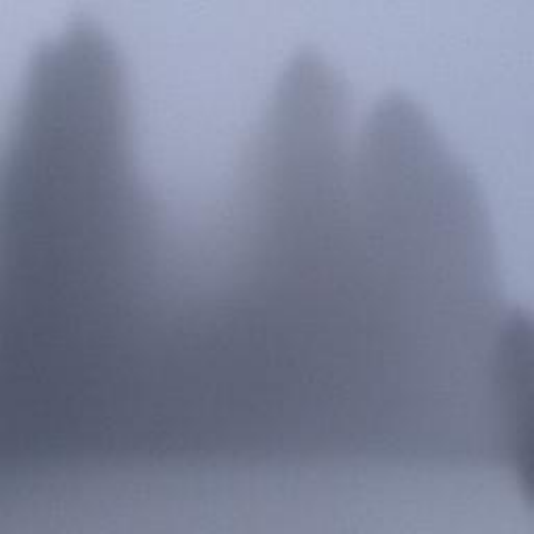}
\includegraphics[width=0.13\columnwidth]{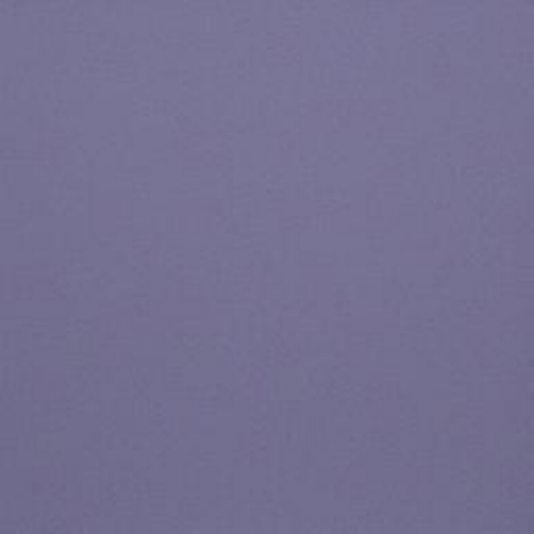}
\includegraphics[width=0.13\columnwidth]{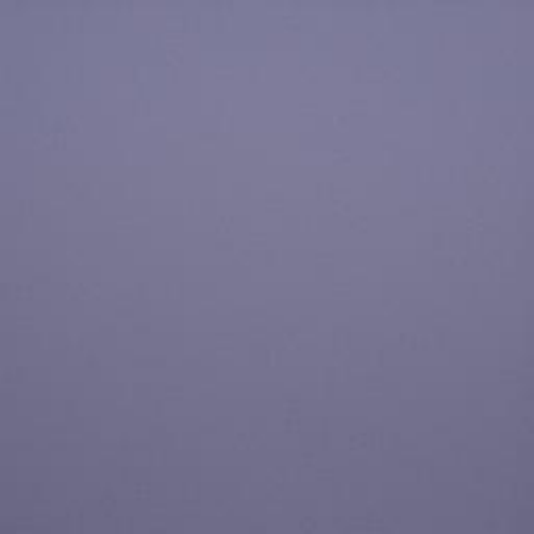}
\\
\includegraphics[width=0.13\columnwidth]{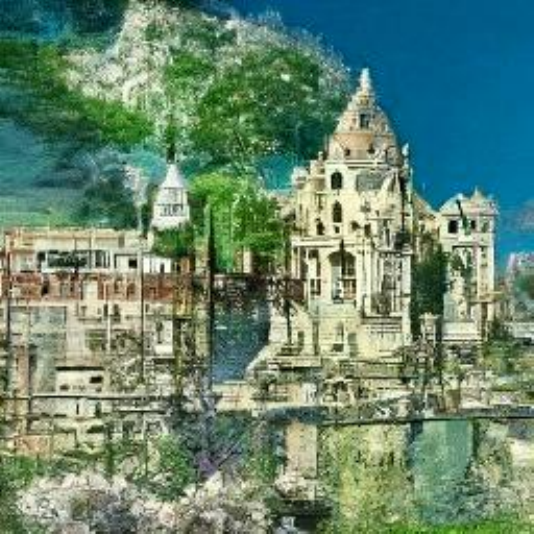}
\includegraphics[width=0.13\columnwidth]{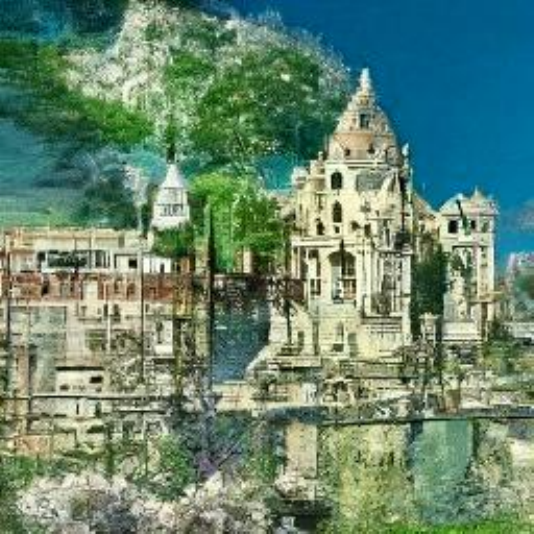}
\includegraphics[width=0.13\columnwidth]{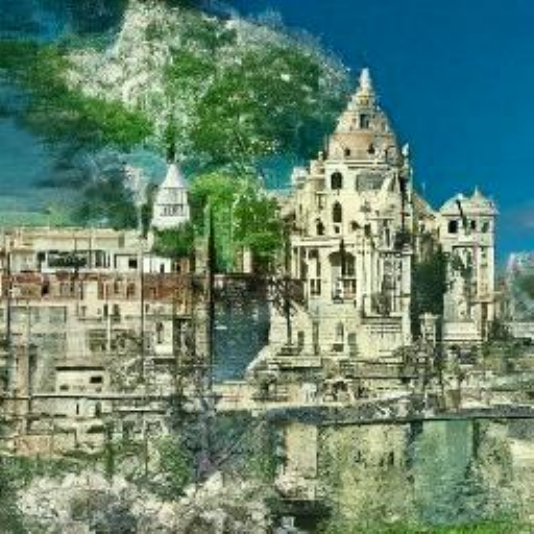}
\includegraphics[width=0.13\columnwidth]{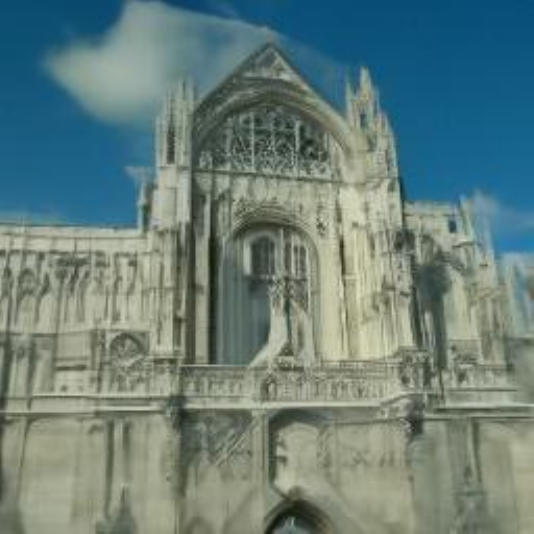}
\includegraphics[width=0.13\columnwidth]{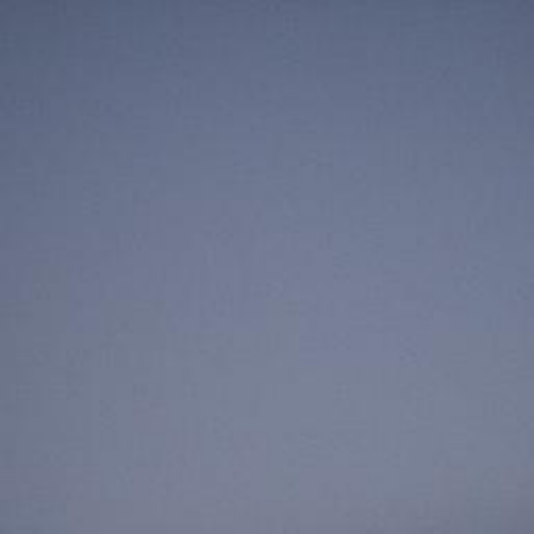}
\includegraphics[width=0.13\columnwidth]{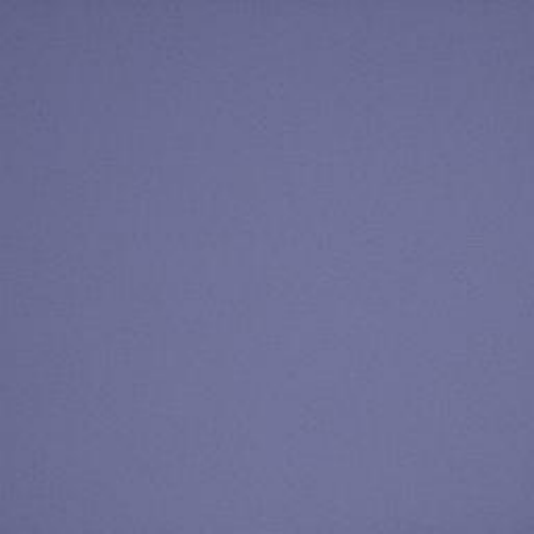}
\includegraphics[width=0.13\columnwidth]{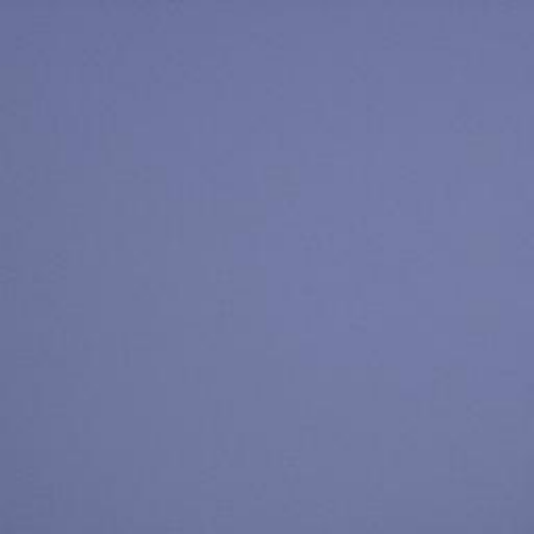}
\caption{\textbf{Varying the Mixing Parameter.
} We consider DDPM with $20$ steps. We fix the variance scaling at $1.55$, and each column corresponds to mixing parameters $0$, $0.00015$, $0.0015$, $0.015$, $0.15$, $0.5$, and $0.75$. Image degradation occurs for low values of the mixing parameter, and image loss for large ones. We point out that $0.015$ is the value of the mixing parameter used for both Sections~\ref{sec:qualitative} and \ref{sec:quantitative}. 
%
%
}
    \label{fig:sweep-mix}
    \end{figure}

\begin{figure}[t]
    \centering
\includegraphics[width=0.15\columnwidth]{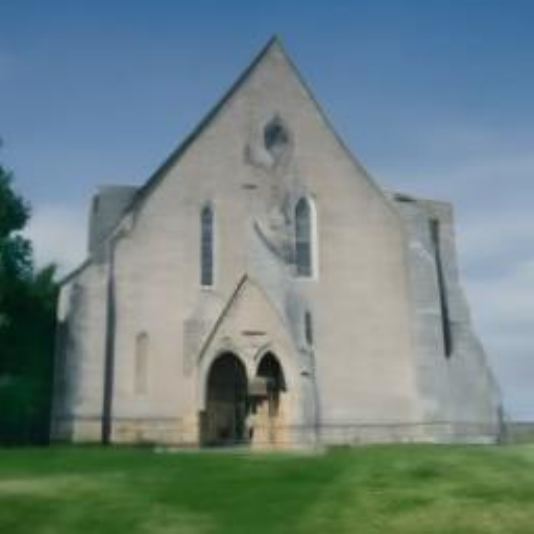}
\includegraphics[width=0.15\columnwidth]{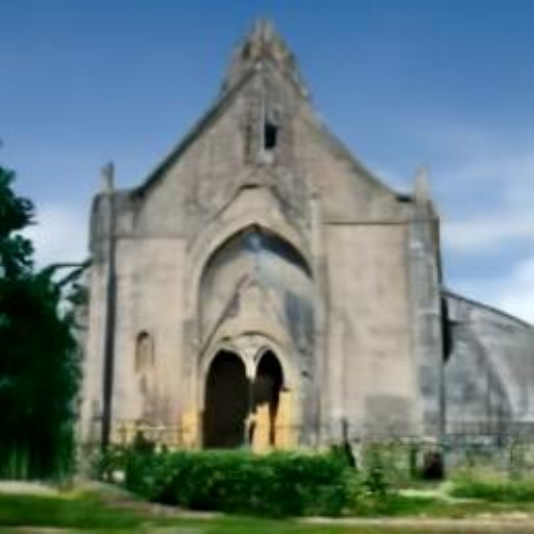}
\includegraphics[width=0.15\columnwidth]{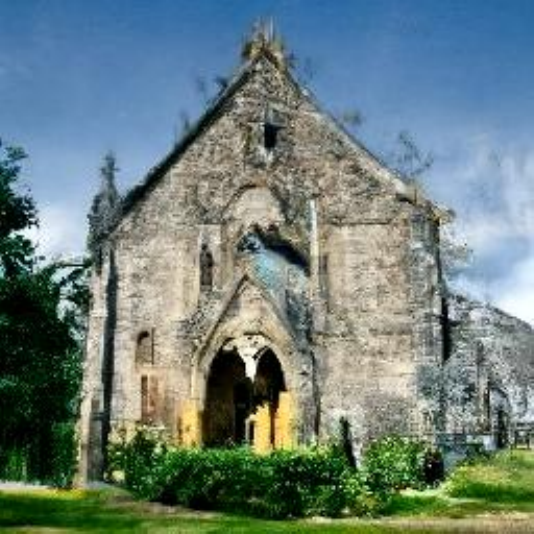}
\\
\includegraphics[width=0.15\columnwidth]{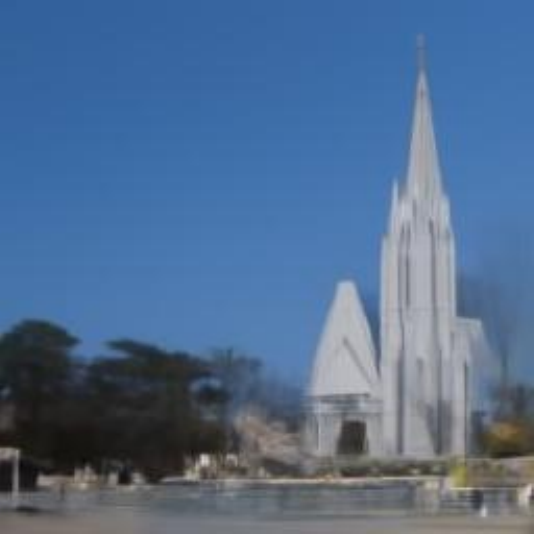}
\includegraphics[width=0.15\columnwidth]{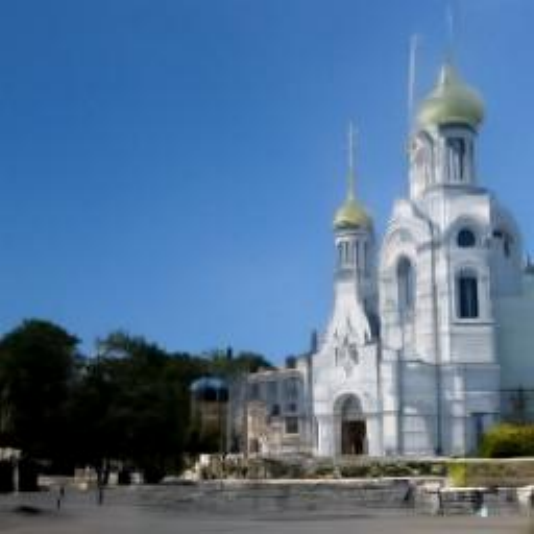}
\includegraphics[width=0.15\columnwidth]{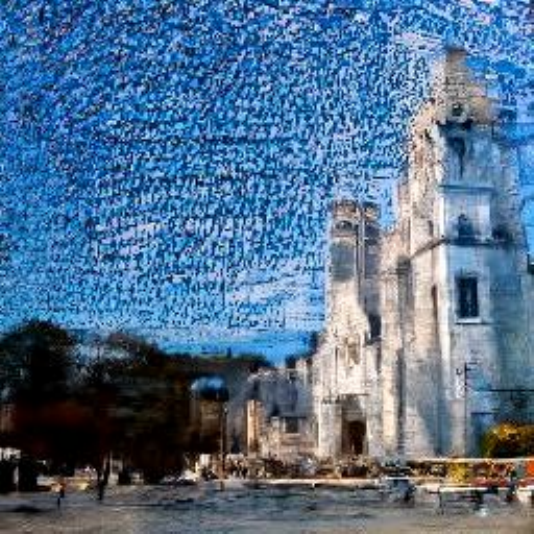}
\\
\includegraphics[width=0.15\columnwidth]{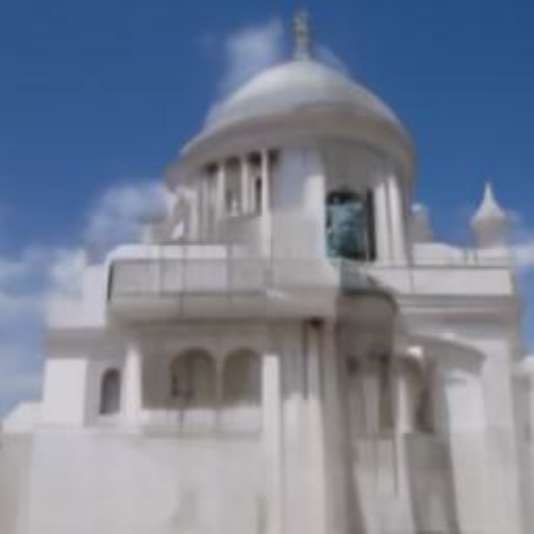}
\includegraphics[width=0.15\columnwidth]{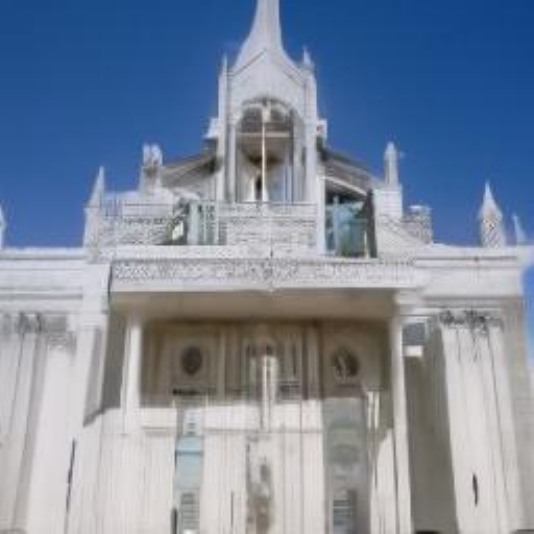}
\includegraphics[width=0.15\columnwidth]{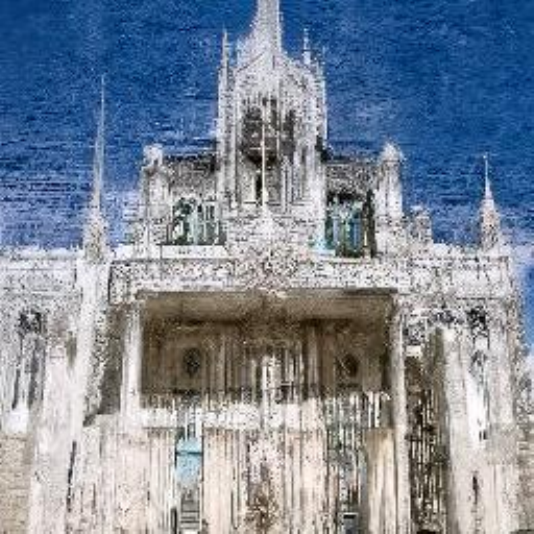}
\\
\includegraphics[width=0.15\columnwidth]{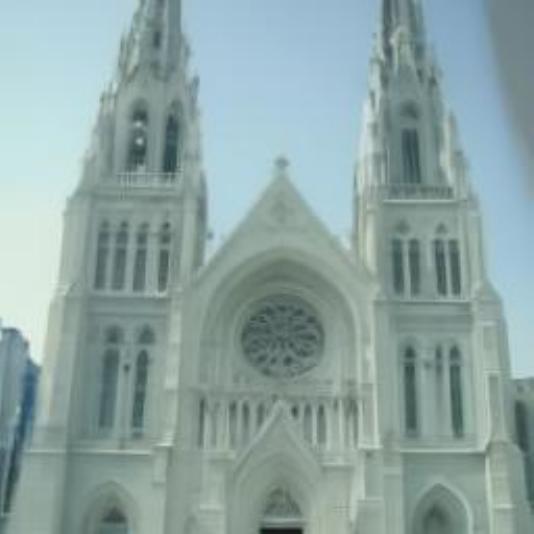}
\includegraphics[width=0.15\columnwidth]{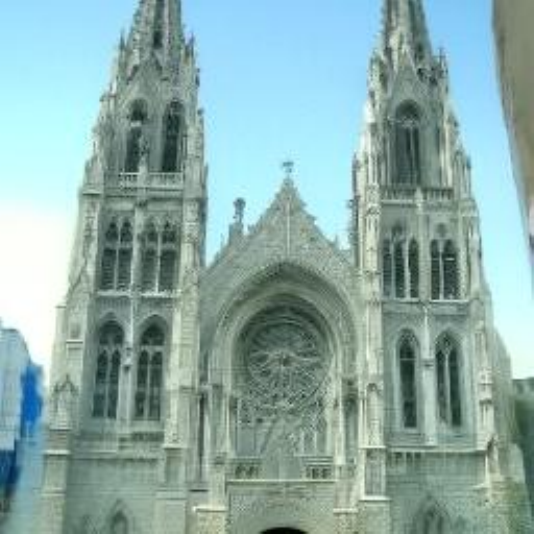}
\includegraphics[width=0.15\columnwidth]{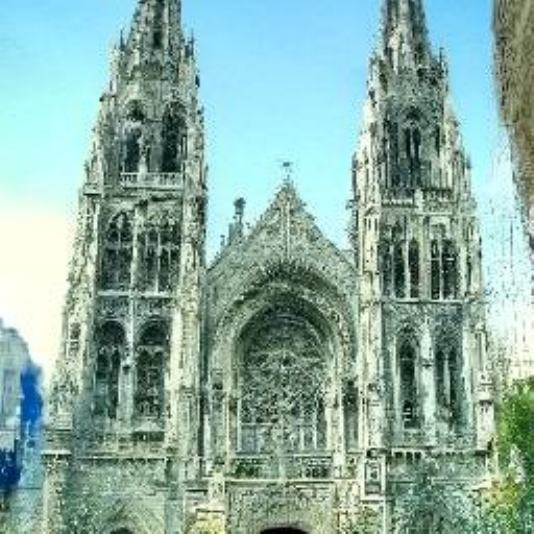}
\\
\includegraphics[width=0.15\columnwidth]{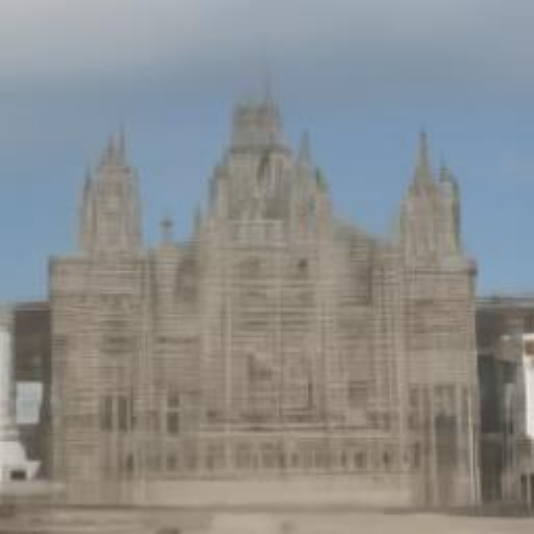}
\includegraphics[width=0.15\columnwidth]{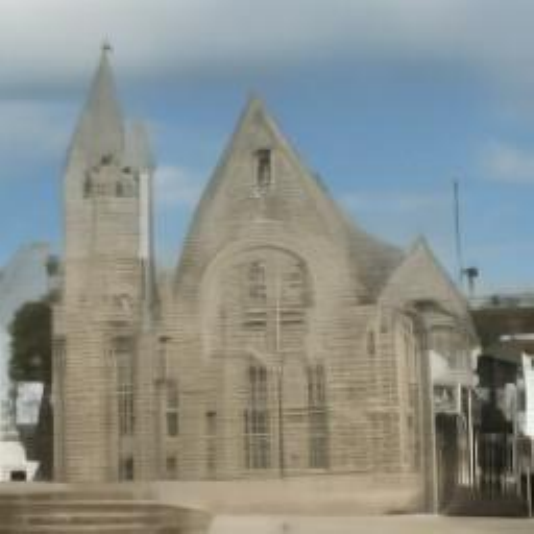}
\includegraphics[width=0.15\columnwidth]{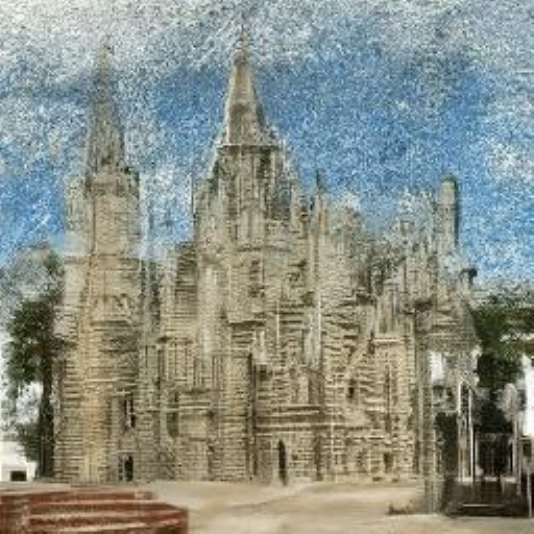}
\\
\includegraphics[width=0.15\columnwidth]{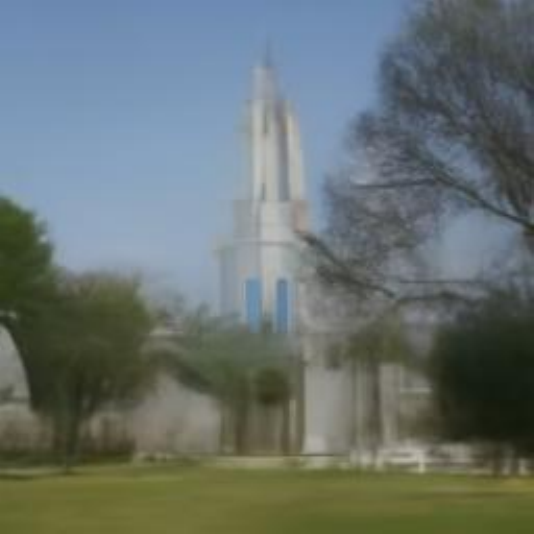}
\includegraphics[width=0.15\columnwidth]{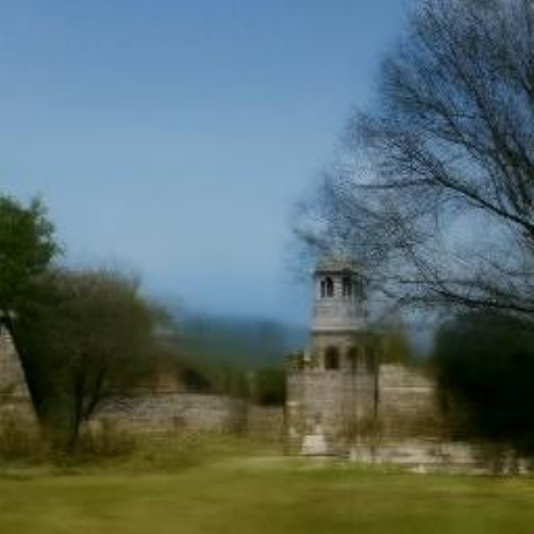}
\includegraphics[width=0.15\columnwidth]{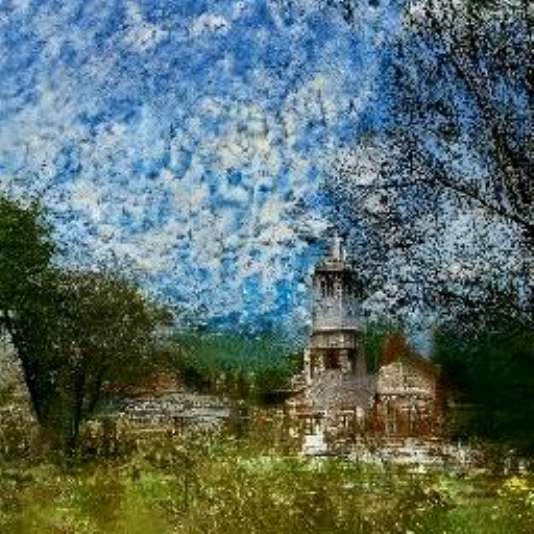}
\\
\includegraphics[width=0.15\columnwidth]{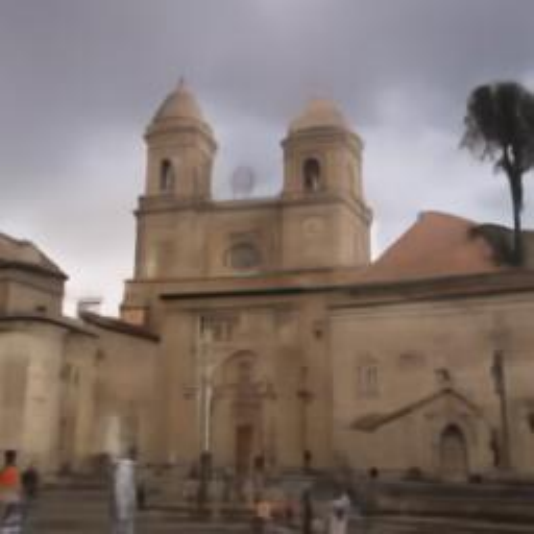}
\includegraphics[width=0.15\columnwidth]{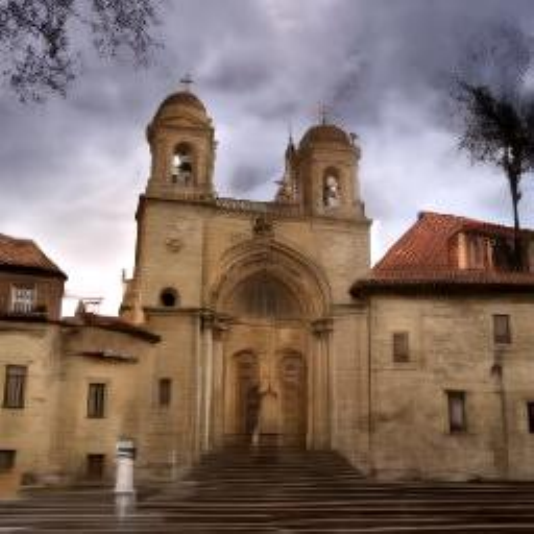}
\includegraphics[width=0.15\columnwidth]{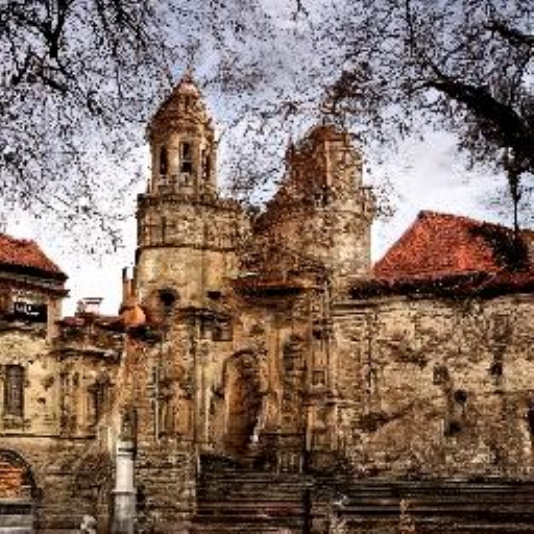}
\\
\includegraphics[width=0.15\columnwidth]{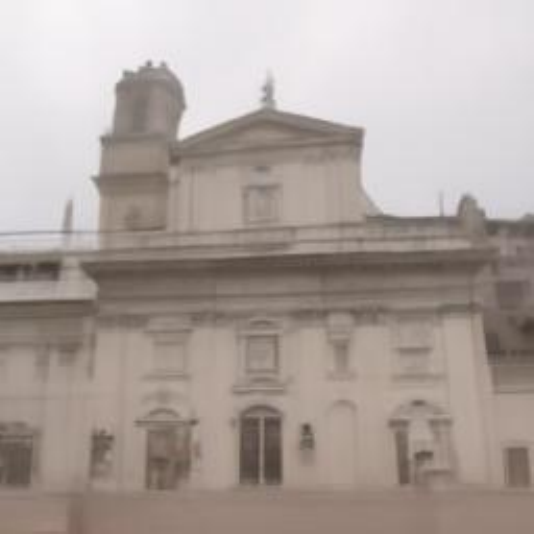}
\includegraphics[width=0.15\columnwidth]{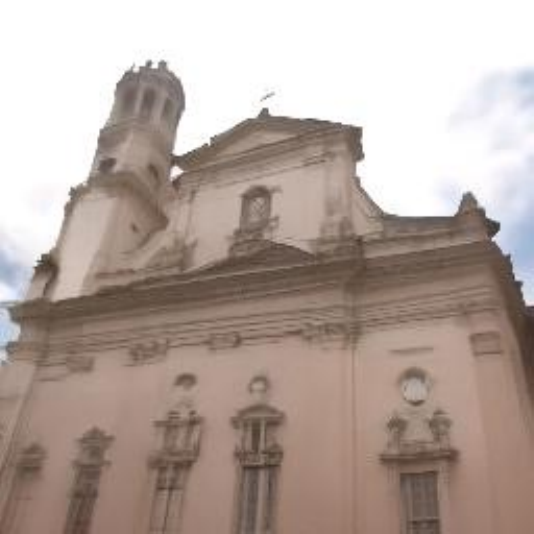}
\includegraphics[width=0.15\columnwidth]{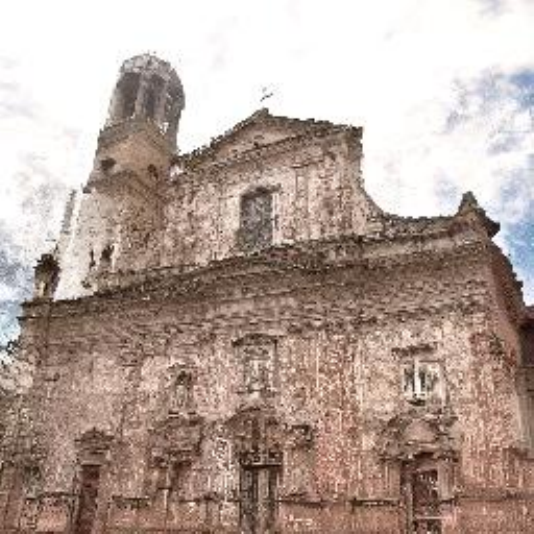}
\caption{\textbf{Image Degradation when Integrating the Predictor with the Latent at Step $t_k+1$.} We consider DDPM with $20$ steps. For each row, from left to right: baseline and two generated images with variance scaling $1.55$ and the usual mixing parameter $0.015$, one according to SE2P and the other one to SE2P with Algorithm~\ref{alg:main_SE2P}'s line $9$ replaced by $\x^{(1)}_k=\gamma\cdot \x_k^{(0)}+(1-\gamma)\cdot\hat{\x}_{\operatorname{pred}}$.
}
    \label{fig:back-int}
    \end{figure}

\begin{figure}[t]
    \centering
\includegraphics[width=0.1\columnwidth]{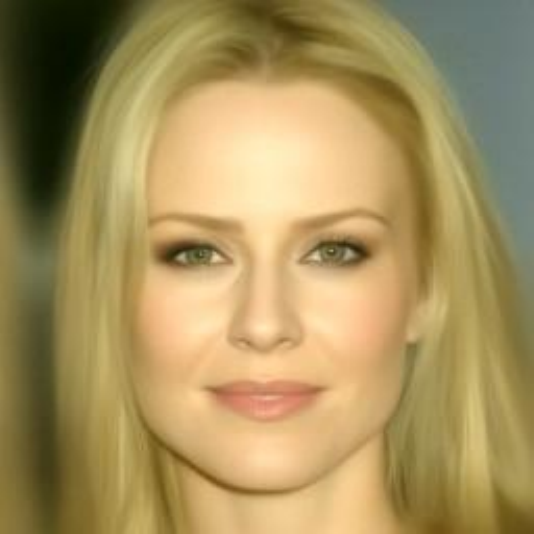}
\includegraphics[width=0.1\columnwidth]{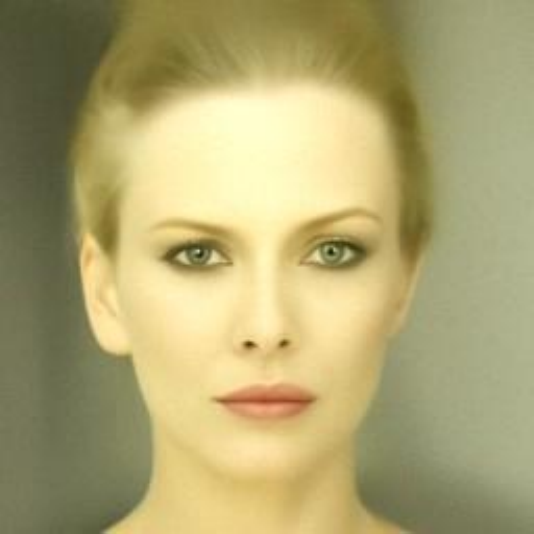}
\includegraphics[width=0.1\columnwidth]{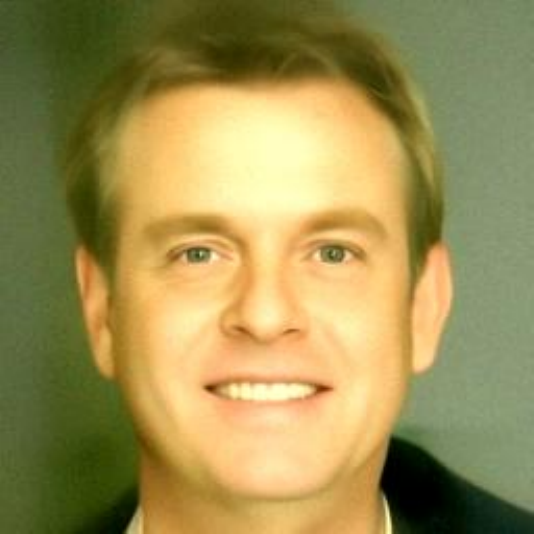}
\includegraphics[width=0.1\columnwidth]{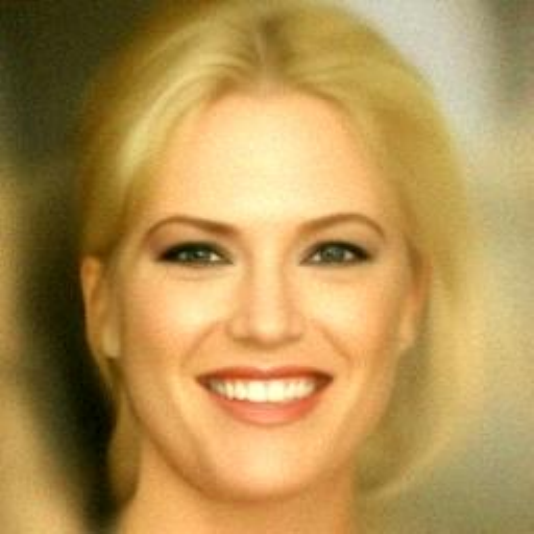}
\includegraphics[width=0.1\columnwidth]{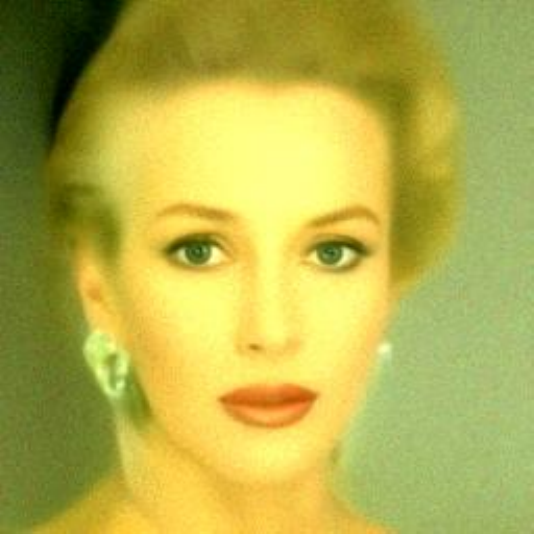}
\includegraphics[width=0.1\columnwidth]{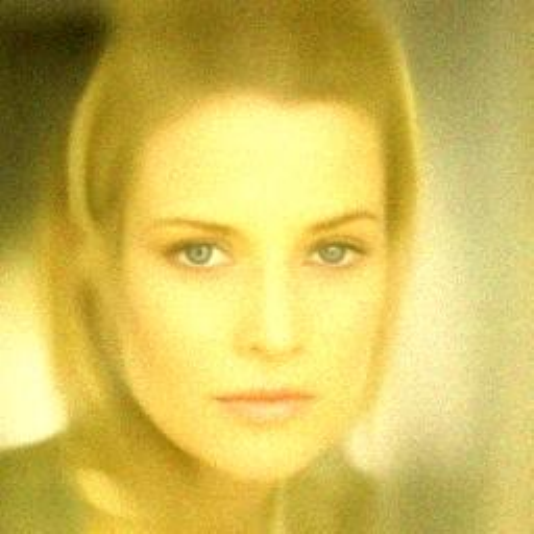}
\includegraphics[width=0.1\columnwidth]{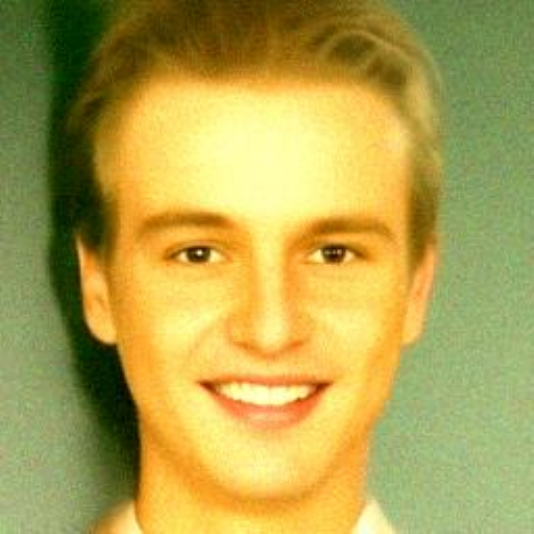}
\includegraphics[width=0.1\columnwidth]{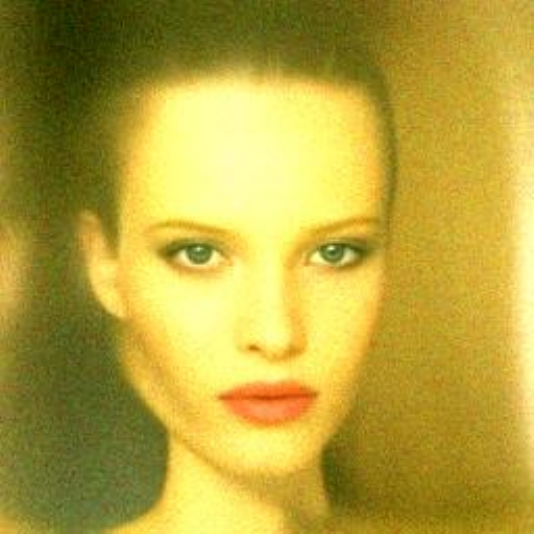}
\includegraphics[width=0.1\columnwidth]{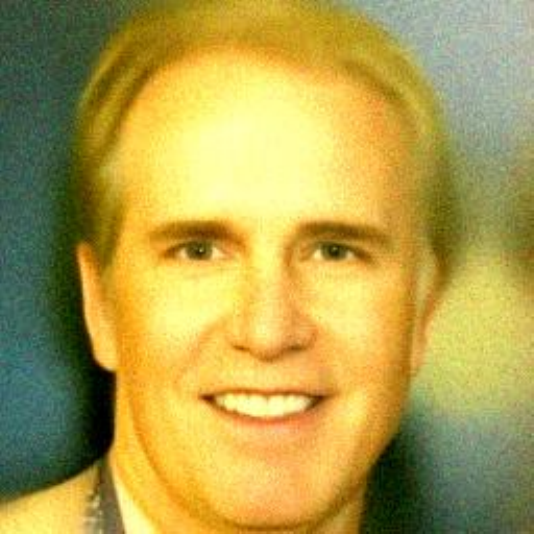}
\\
\includegraphics[width=0.1\columnwidth]{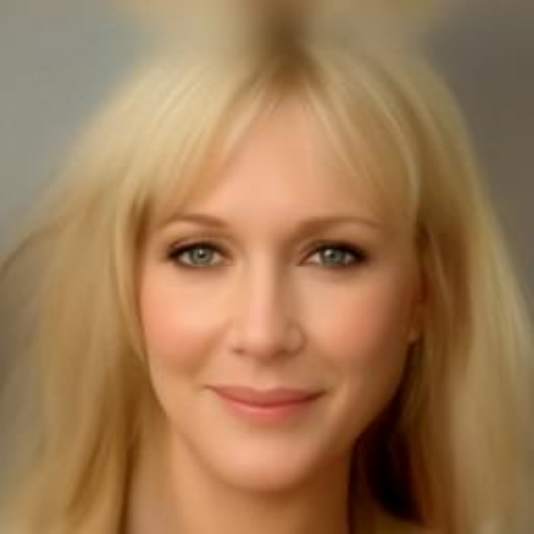}
\includegraphics[width=0.1\columnwidth]{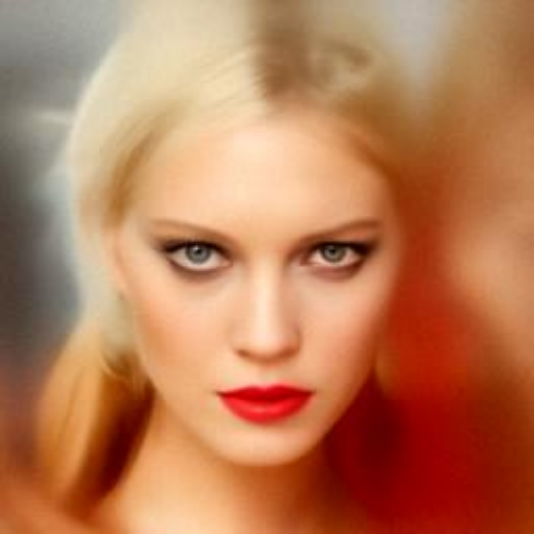}
\includegraphics[width=0.1\columnwidth]{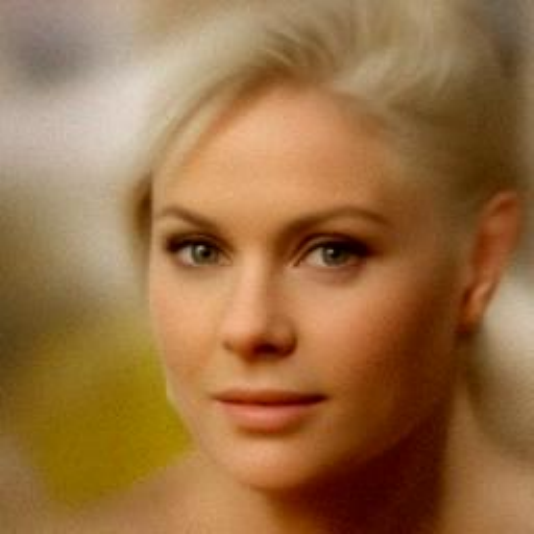}
\includegraphics[width=0.1\columnwidth]{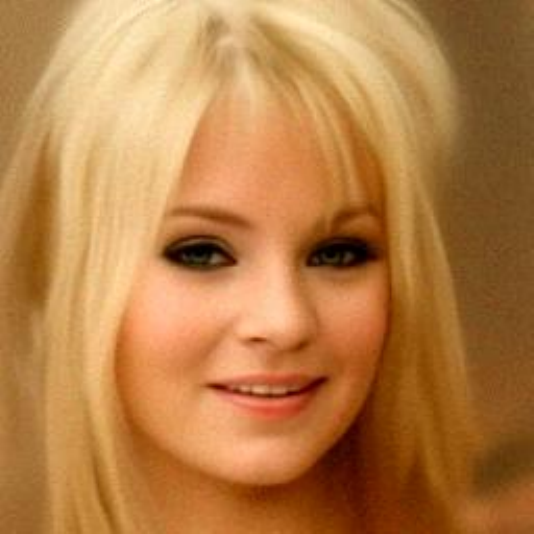}
\includegraphics[width=0.1\columnwidth]{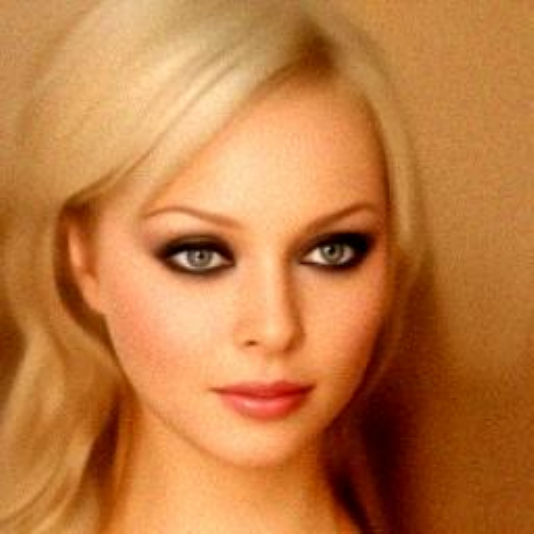}
\includegraphics[width=0.1\columnwidth]{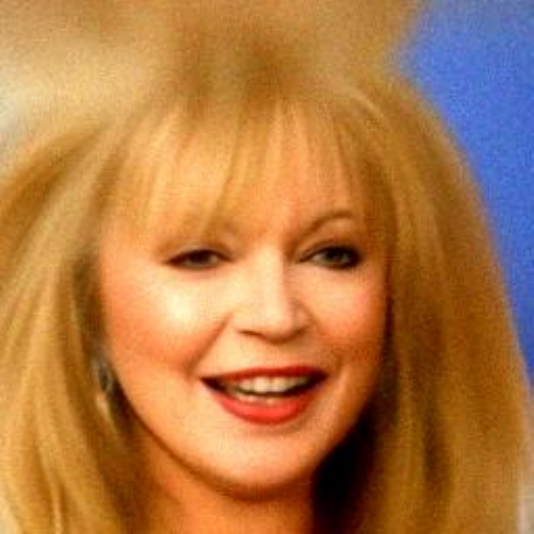}
\includegraphics[width=0.1\columnwidth]{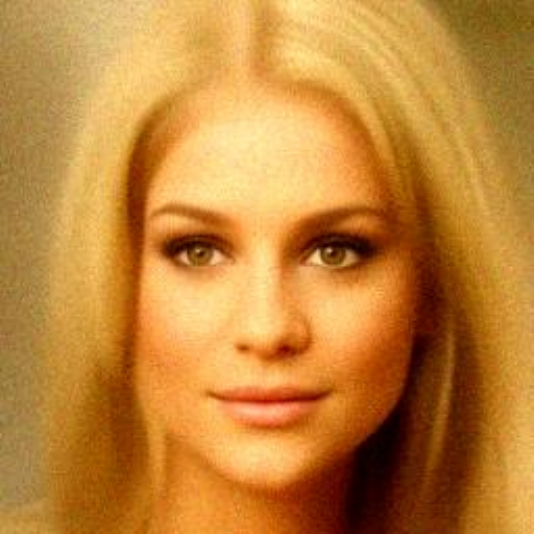}
\includegraphics[width=0.1\columnwidth]{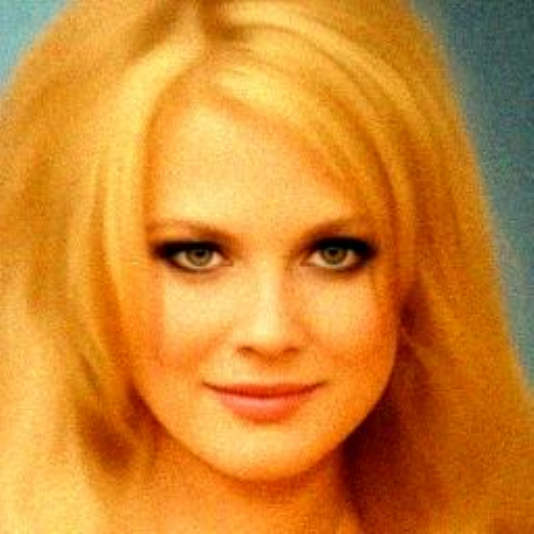}
\includegraphics[width=0.1\columnwidth]{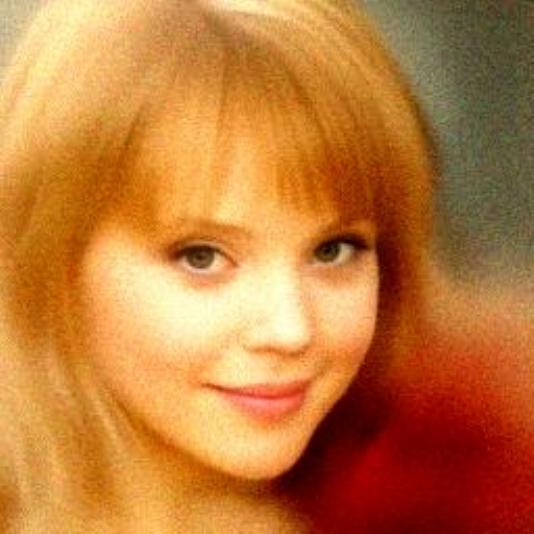}
\\
\includegraphics[width=0.1\columnwidth]{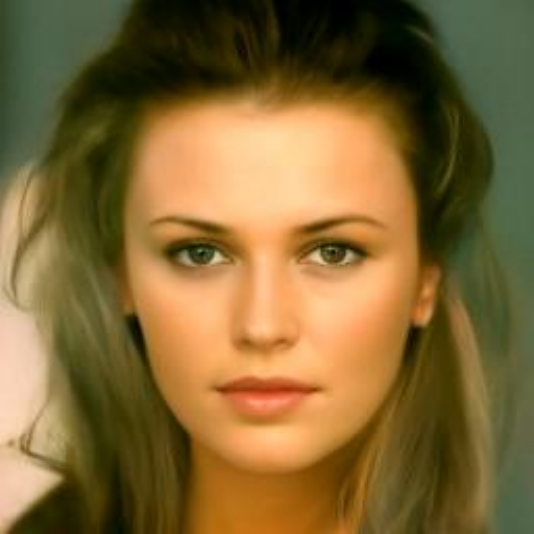}
\includegraphics[width=0.1\columnwidth]{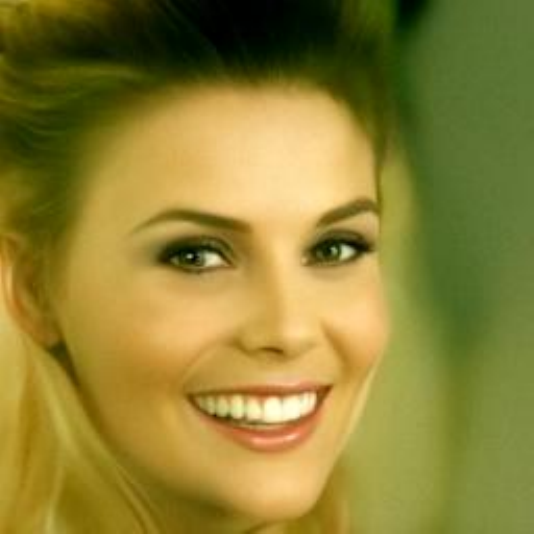}
\includegraphics[width=0.1\columnwidth]{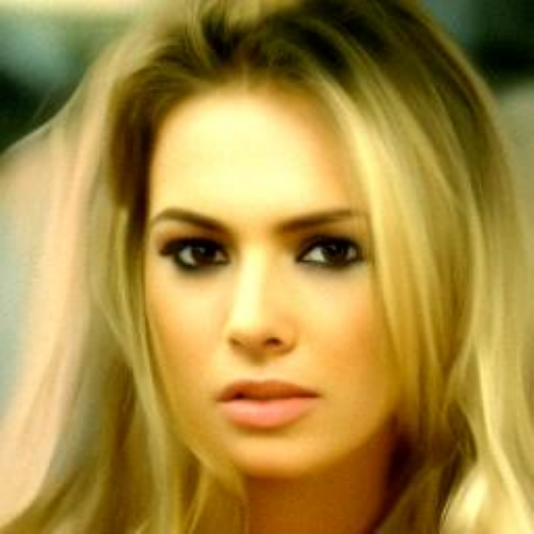}
\includegraphics[width=0.1\columnwidth]{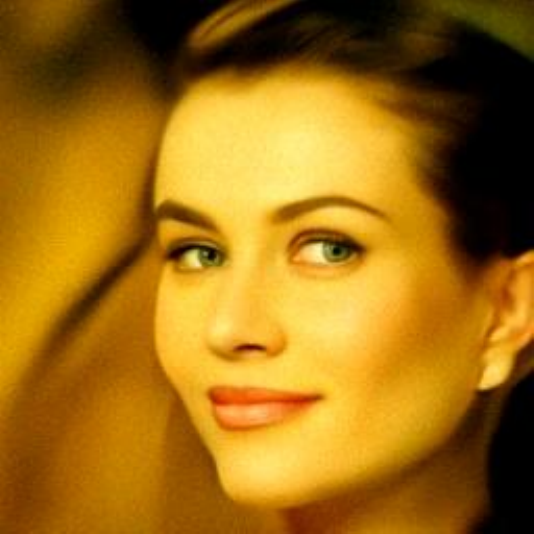}
\includegraphics[width=0.1\columnwidth]{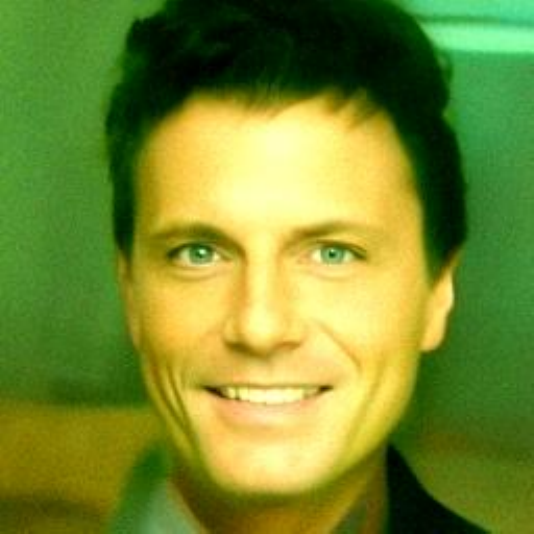}
\includegraphics[width=0.1\columnwidth]{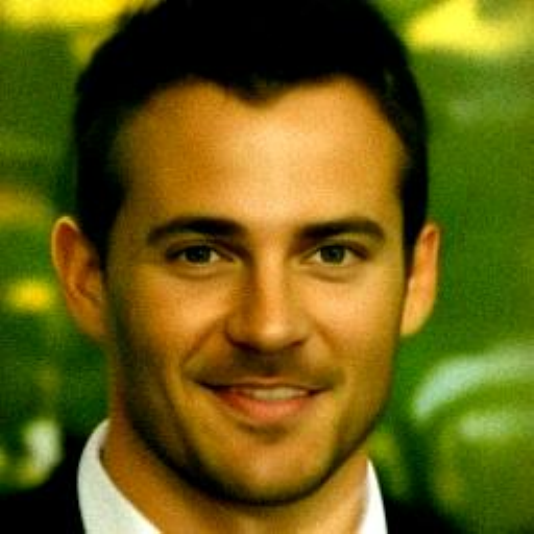}
\includegraphics[width=0.1\columnwidth]{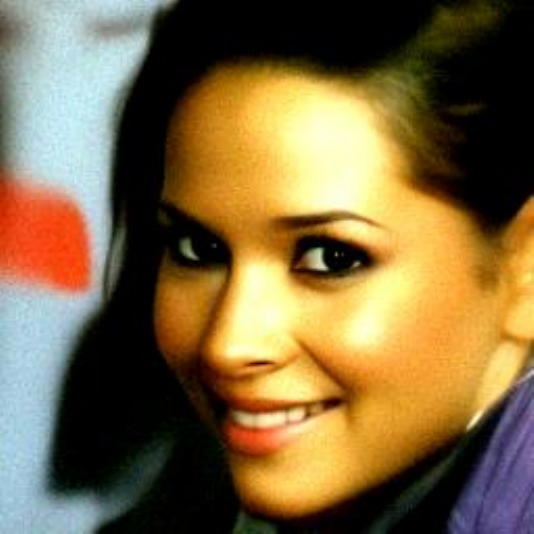}
\includegraphics[width=0.1\columnwidth]{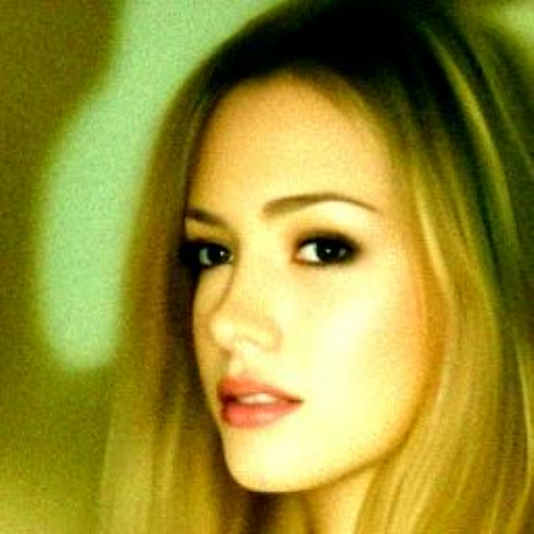}
\includegraphics[width=0.1\columnwidth]{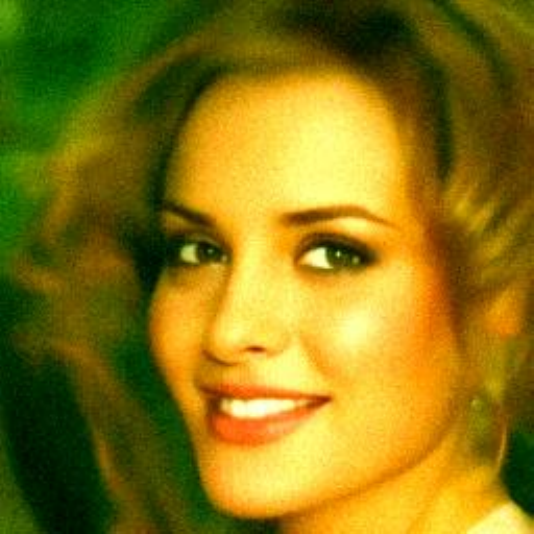}
\\
\includegraphics[width=0.1\columnwidth]{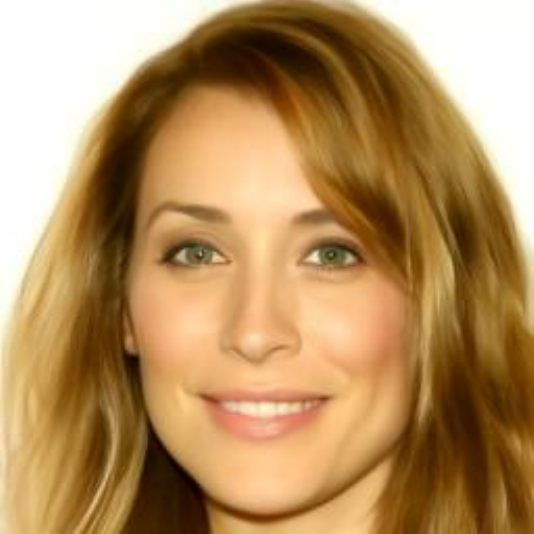}
\includegraphics[width=0.1\columnwidth]{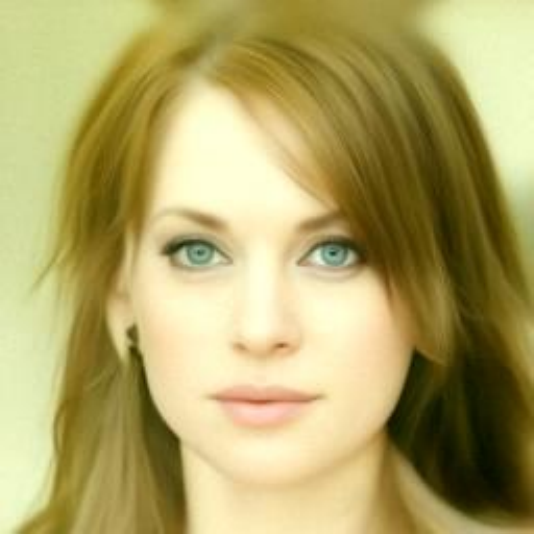}
\includegraphics[width=0.1\columnwidth]{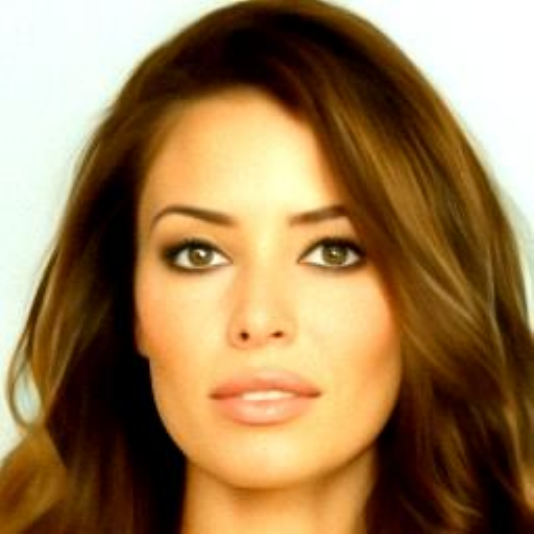}
\includegraphics[width=0.1\columnwidth]{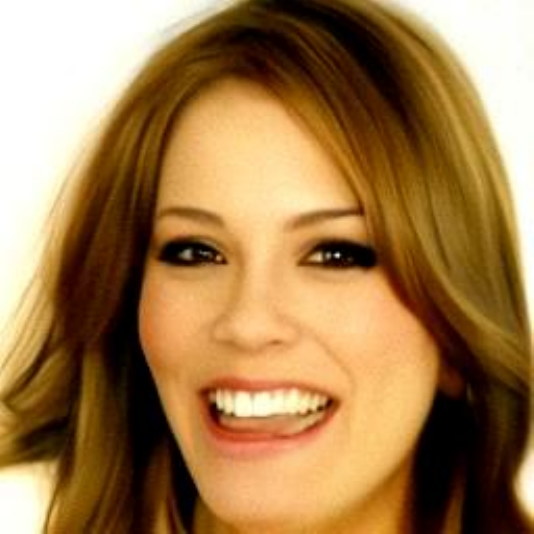}
\includegraphics[width=0.1\columnwidth]{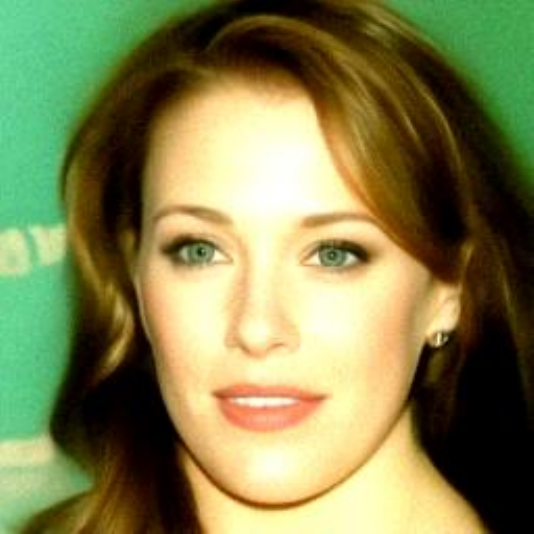}
\includegraphics[width=0.1\columnwidth]{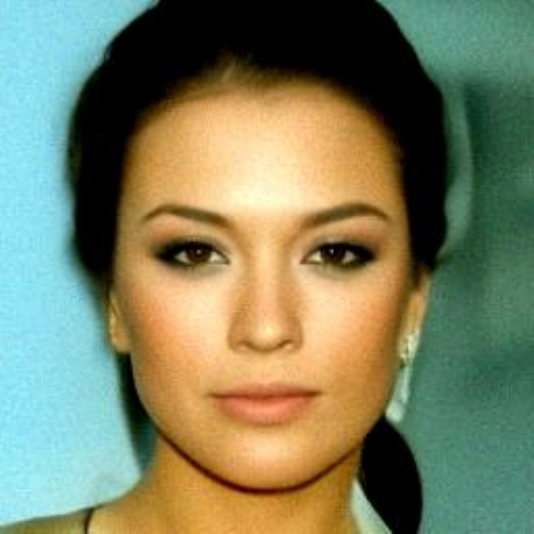}
\includegraphics[width=0.1\columnwidth]{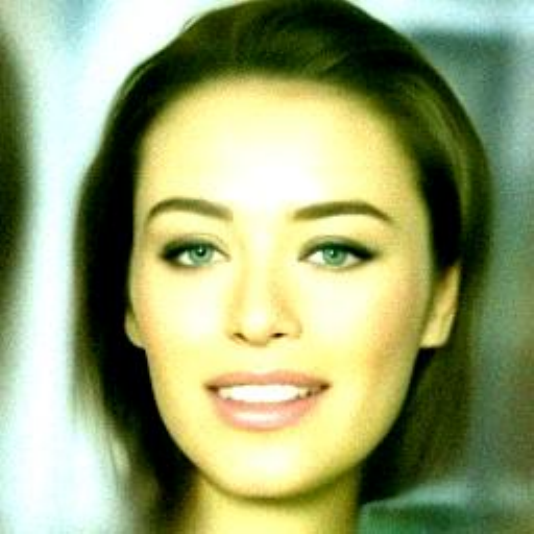}
\includegraphics[width=0.1\columnwidth]{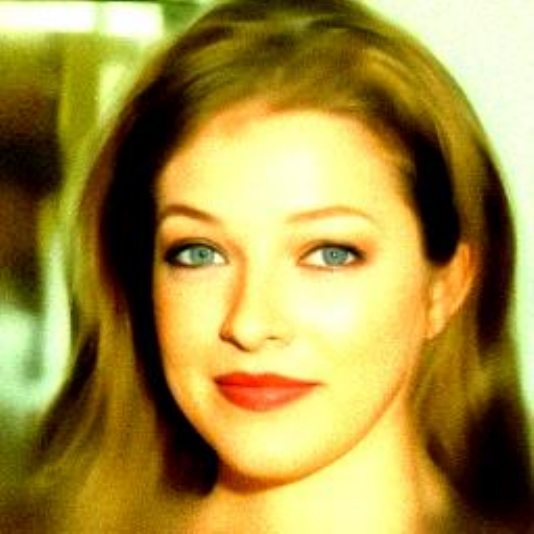}
\includegraphics[width=0.1\columnwidth]{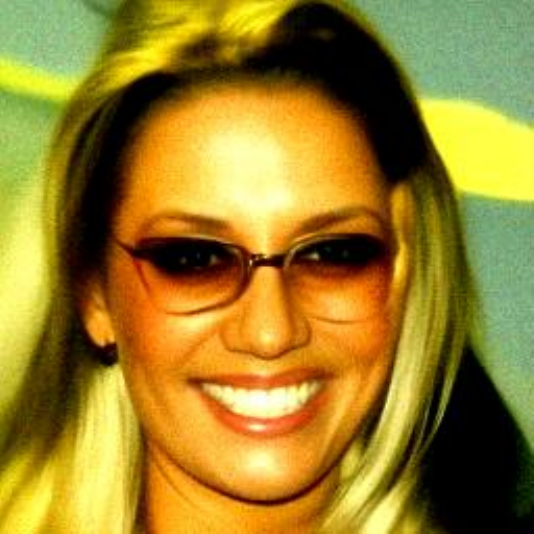}
\\
\includegraphics[width=0.1\columnwidth]{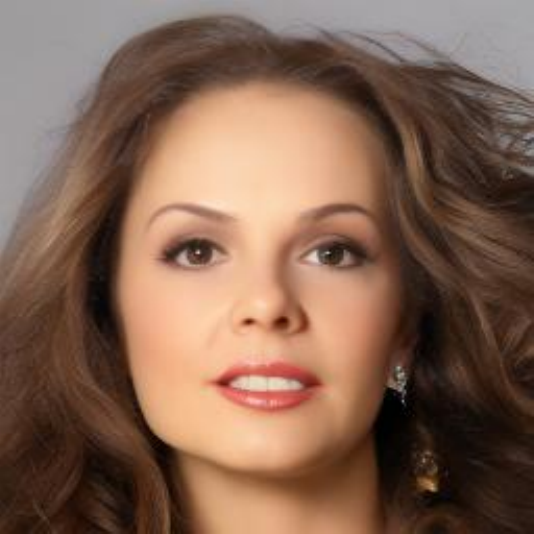}
\includegraphics[width=0.1\columnwidth]{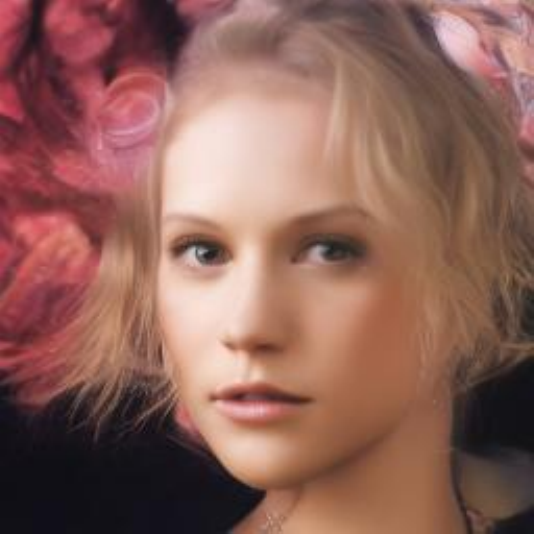}
\includegraphics[width=0.1\columnwidth]{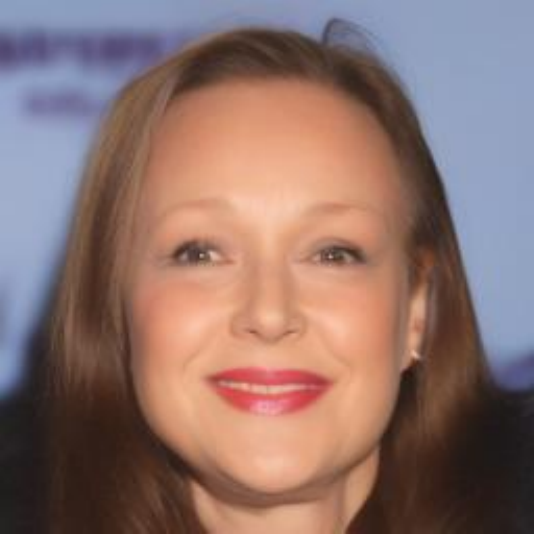}
\includegraphics[width=0.1\columnwidth]{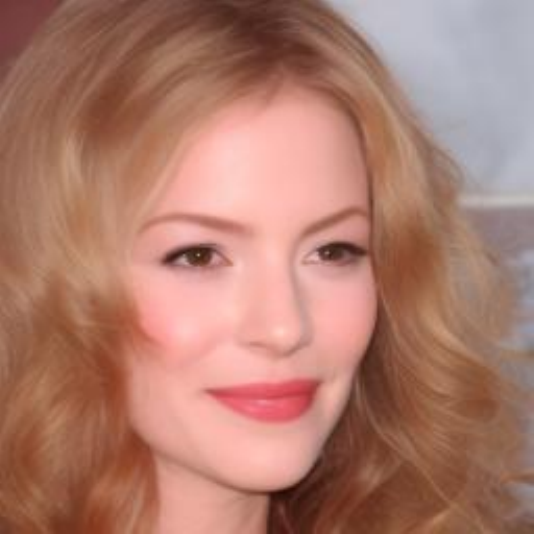}
\includegraphics[width=0.1\columnwidth]{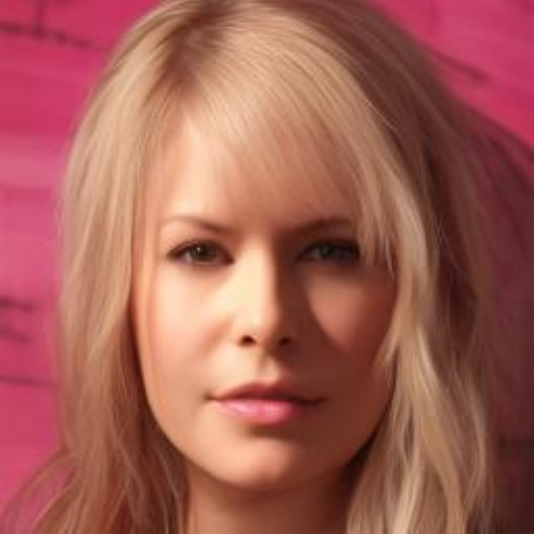}
\includegraphics[width=0.1\columnwidth]{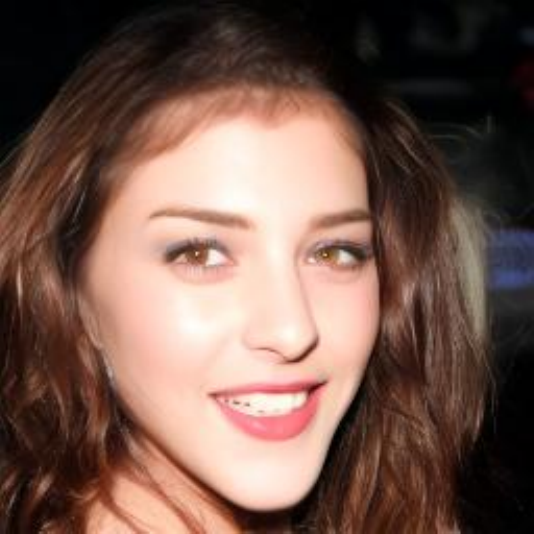}
\includegraphics[width=0.1\columnwidth]{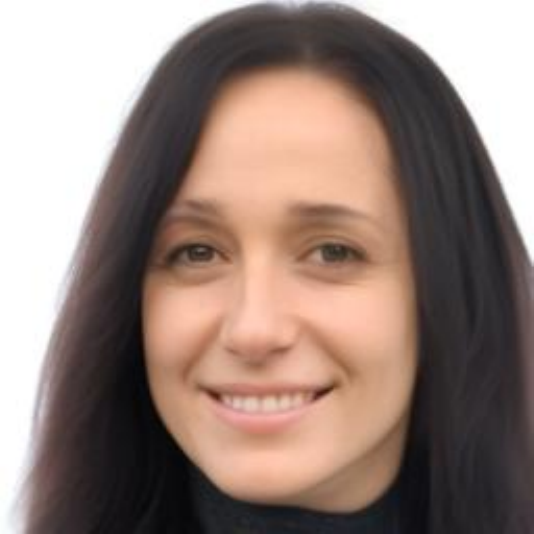}
\includegraphics[width=0.1\columnwidth]{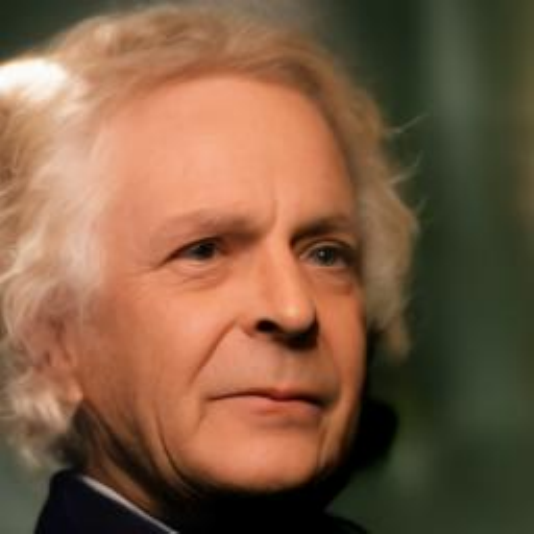}
\includegraphics[width=0.1\columnwidth]{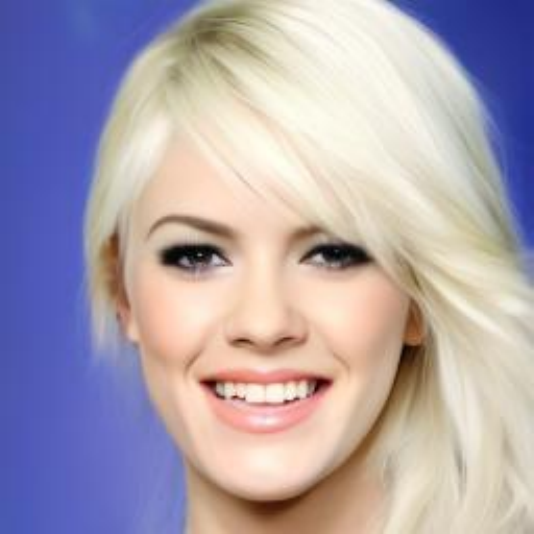}
\\
\includegraphics[width=0.1\columnwidth]{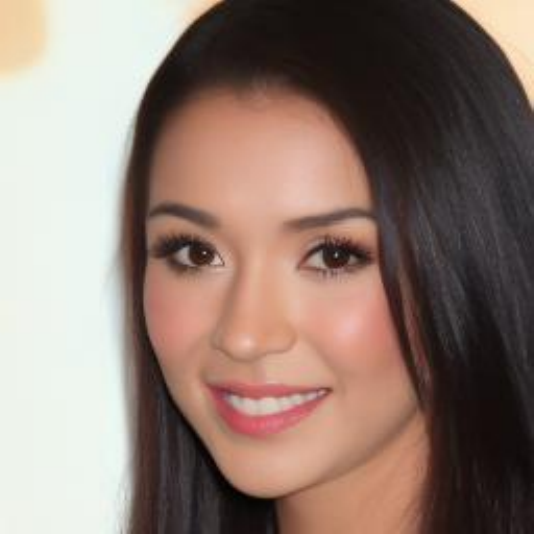}
\includegraphics[width=0.1\columnwidth]{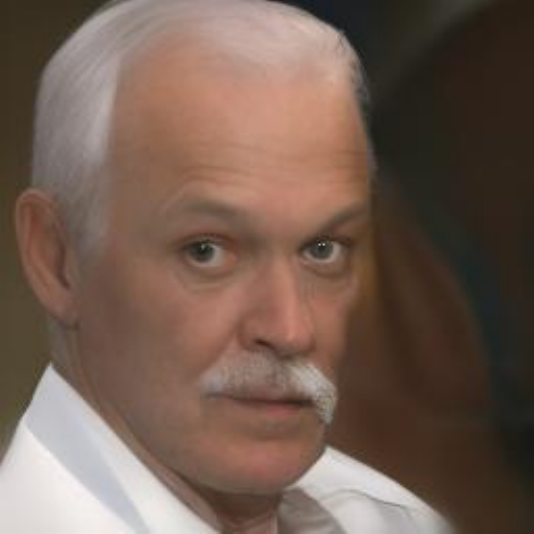}
\includegraphics[width=0.1\columnwidth]{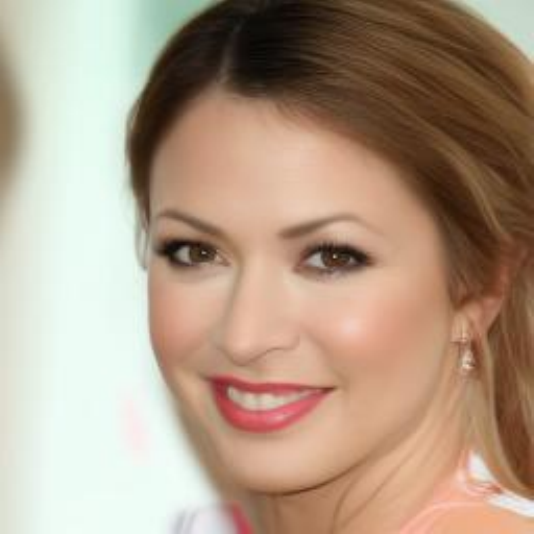}
\includegraphics[width=0.1\columnwidth]{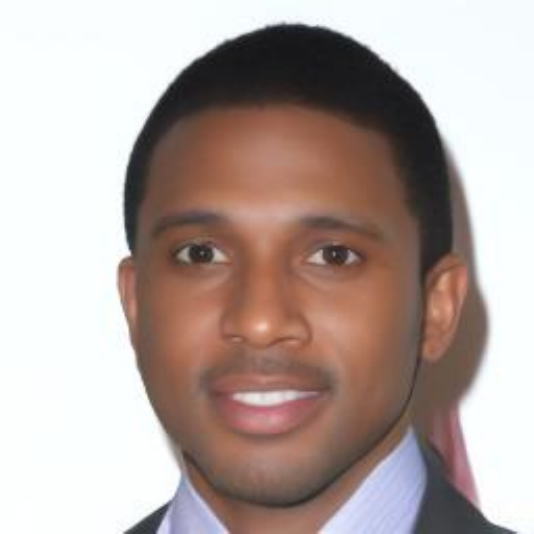}
\includegraphics[width=0.1\columnwidth]{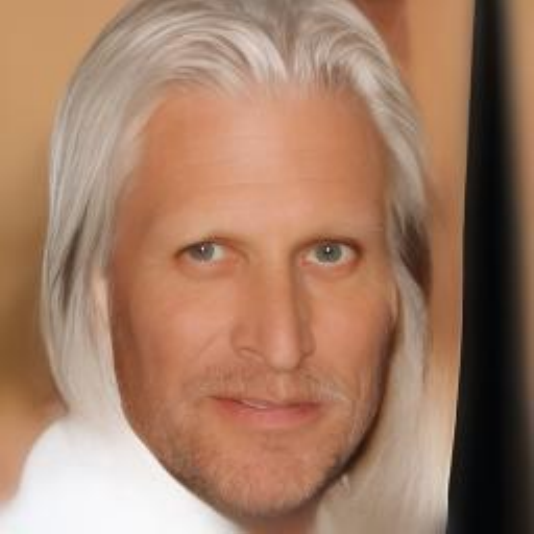}
\includegraphics[width=0.1\columnwidth]{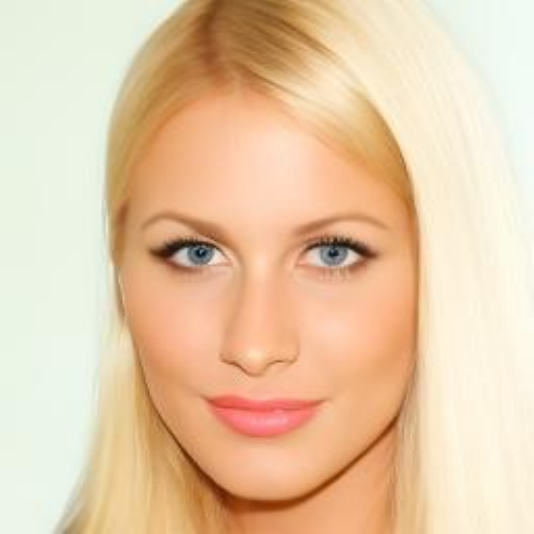}
\includegraphics[width=0.1\columnwidth]{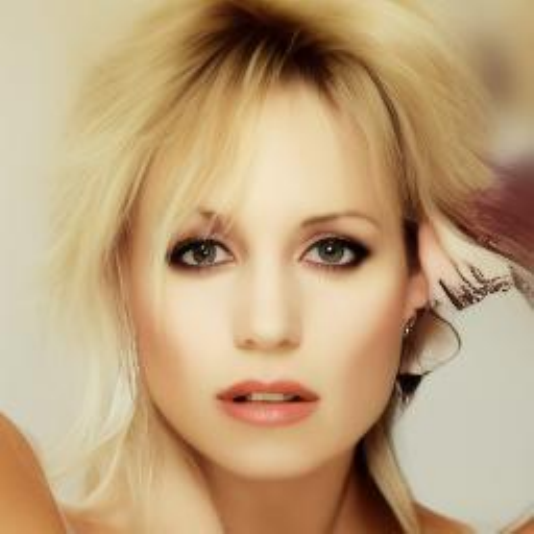}
\includegraphics[width=0.1\columnwidth]{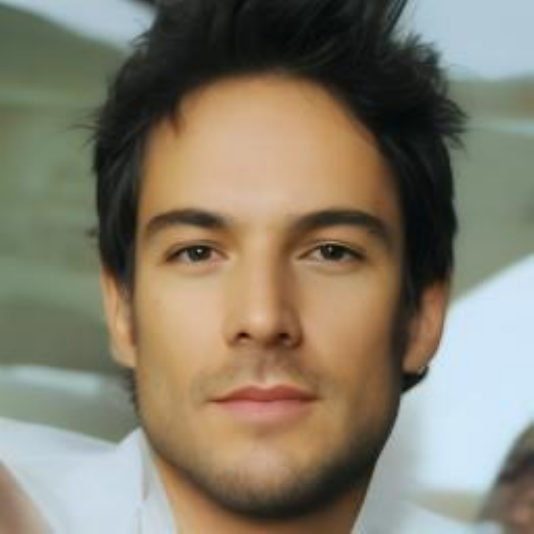}
\includegraphics[width=0.1\columnwidth]{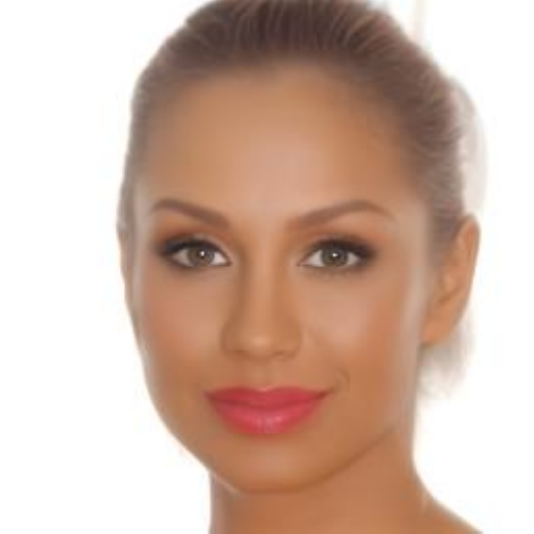}
\\
\includegraphics[width=0.1\columnwidth]{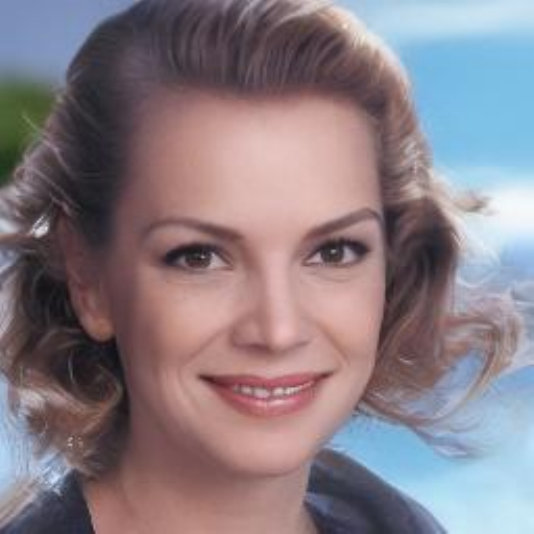}
\includegraphics[width=0.1\columnwidth]{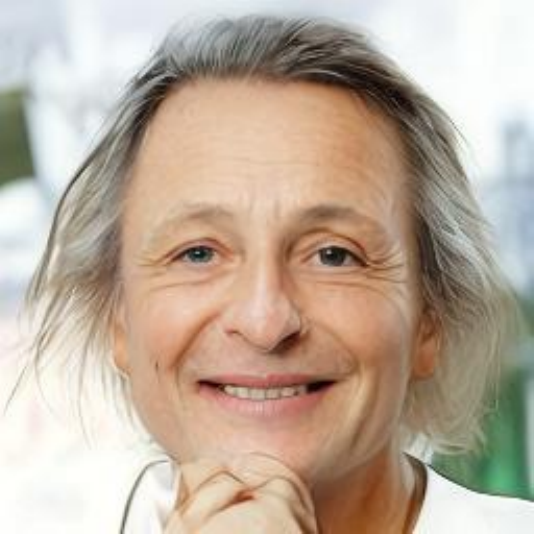}
\includegraphics[width=0.1\columnwidth]{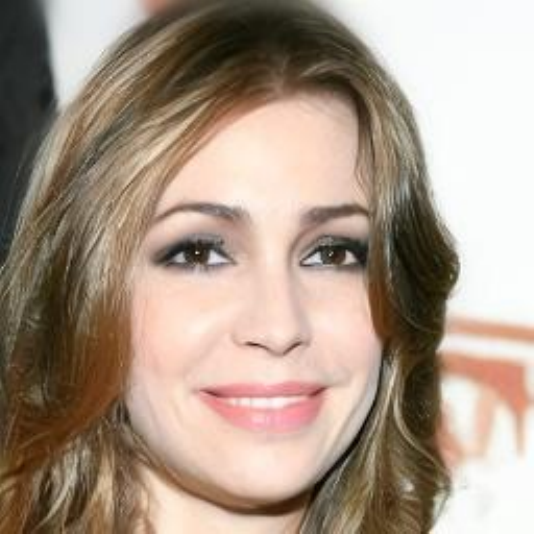}
\includegraphics[width=0.1\columnwidth]{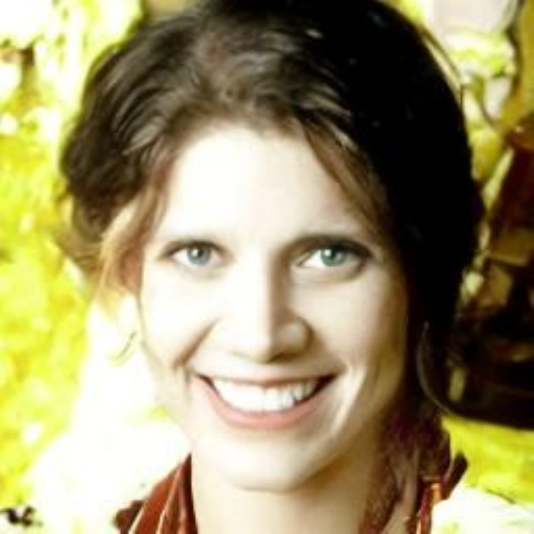}
\includegraphics[width=0.1\columnwidth]{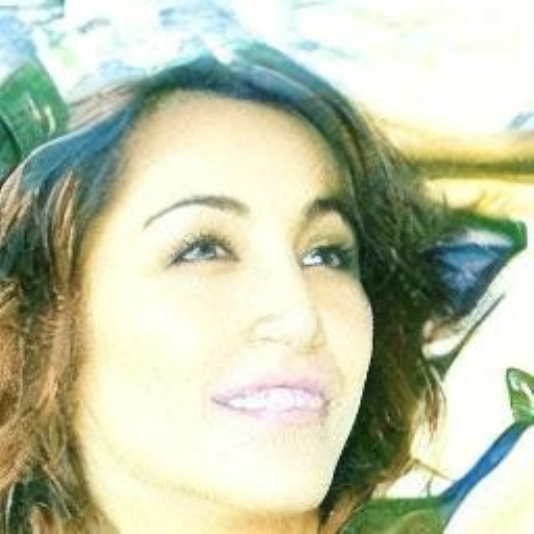}
\includegraphics[width=0.1\columnwidth]{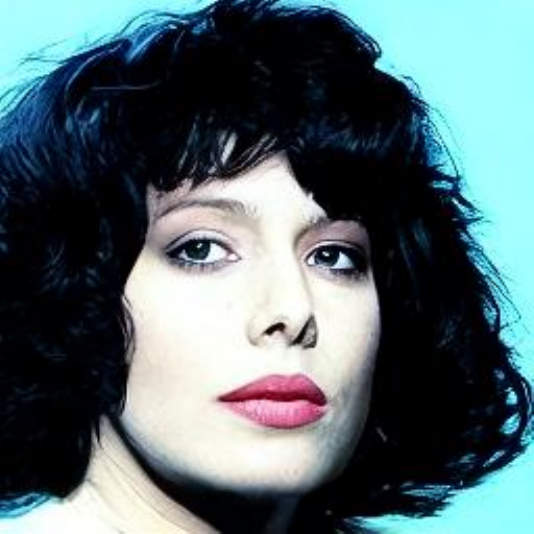}
\includegraphics[width=0.1\columnwidth]{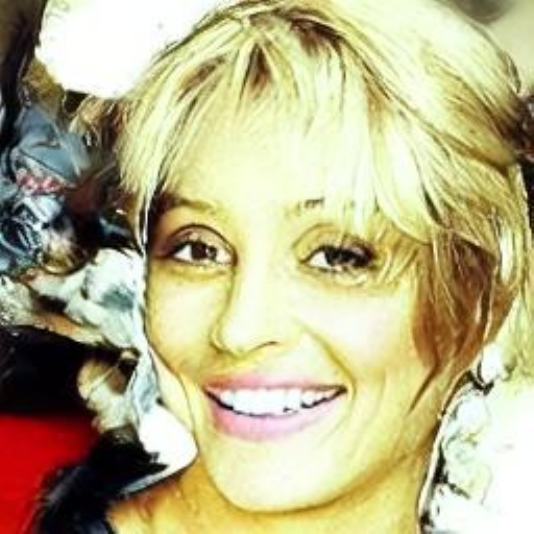}
\includegraphics[width=0.1\columnwidth]{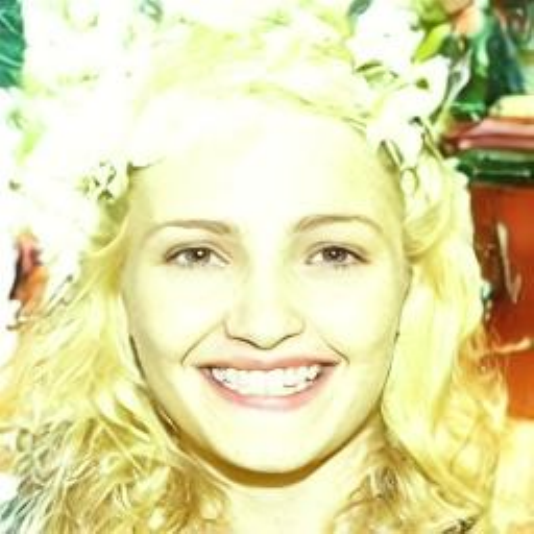}
\includegraphics[width=0.1\columnwidth]{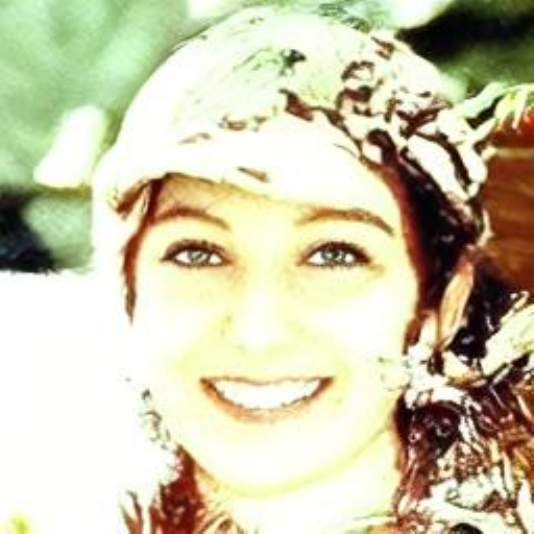}
\\
\includegraphics[width=0.1\columnwidth]{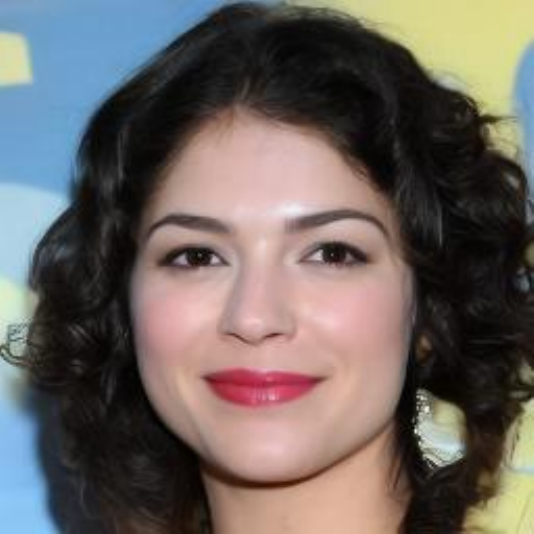}
\includegraphics[width=0.1\columnwidth]{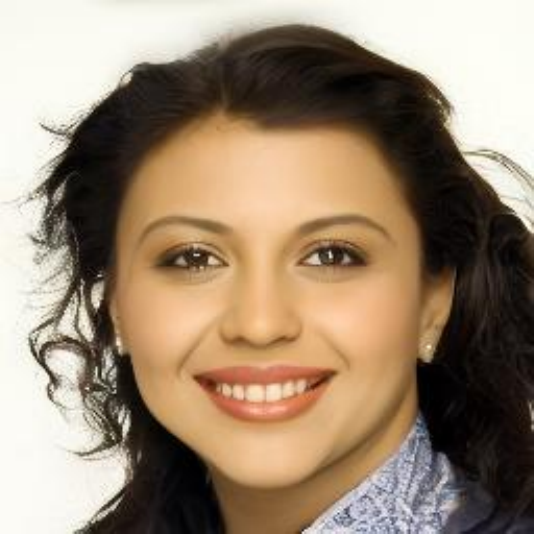}
\includegraphics[width=0.1\columnwidth]{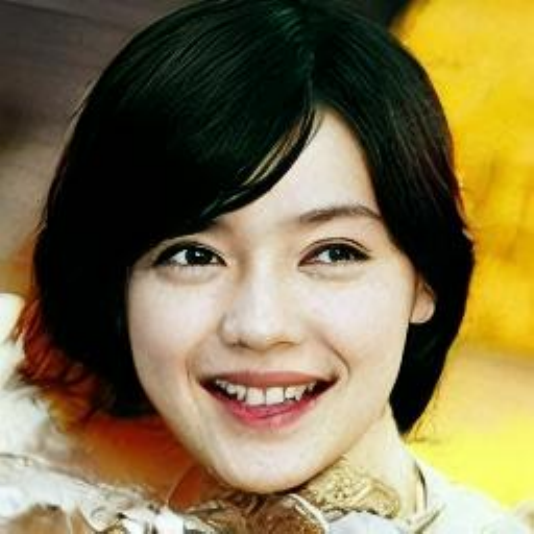}
\includegraphics[width=0.1\columnwidth]{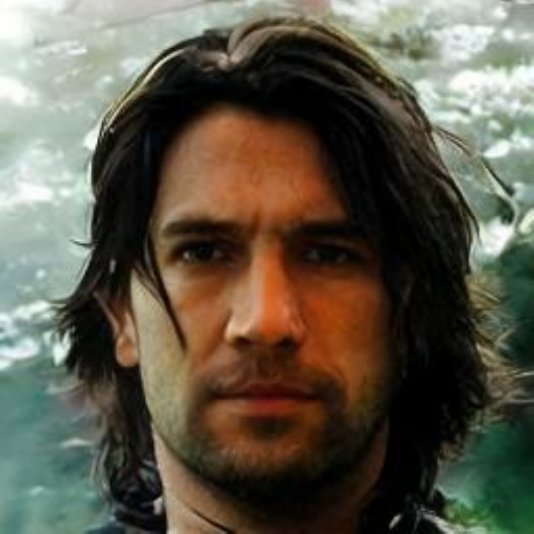}
\includegraphics[width=0.1\columnwidth]{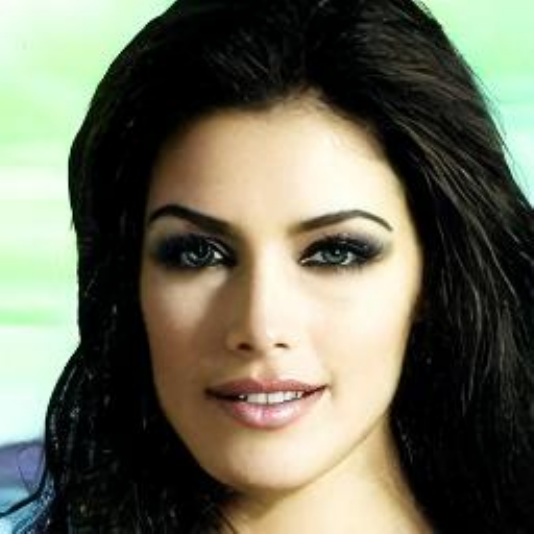}
\includegraphics[width=0.1\columnwidth]{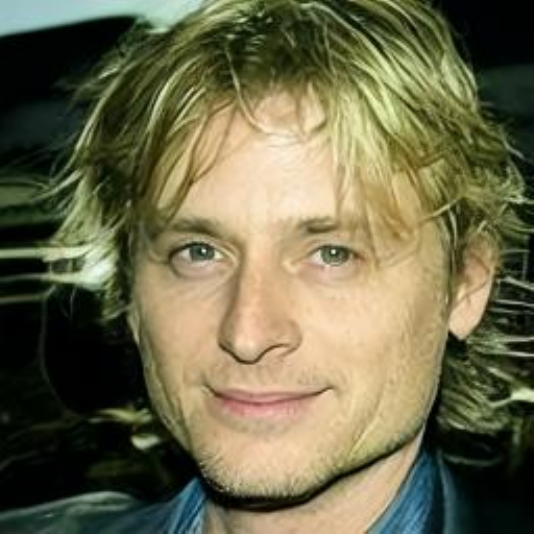}
\includegraphics[width=0.1\columnwidth]{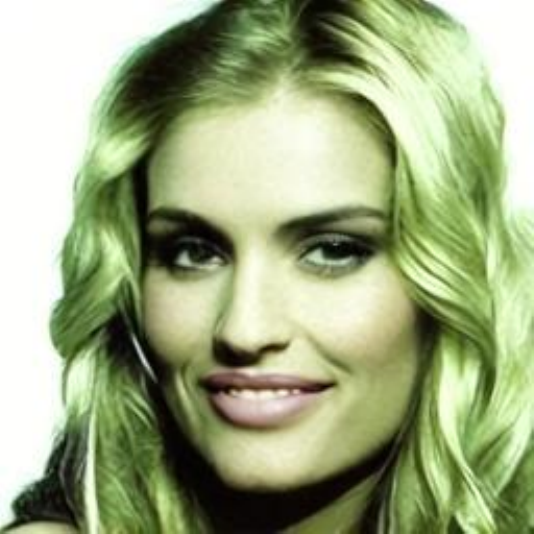}
\includegraphics[width=0.1\columnwidth]{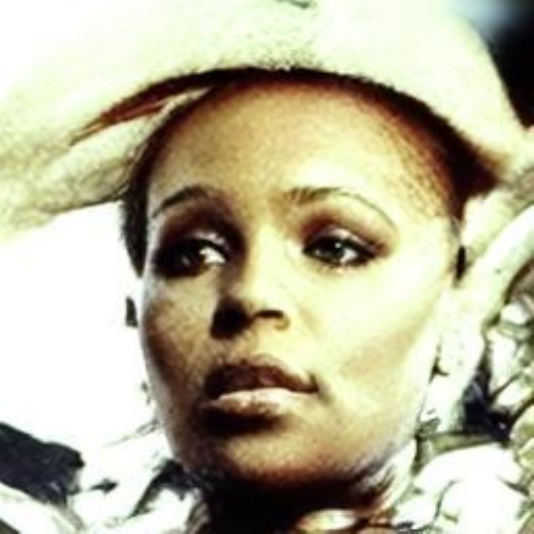}
\includegraphics[width=0.1\columnwidth]{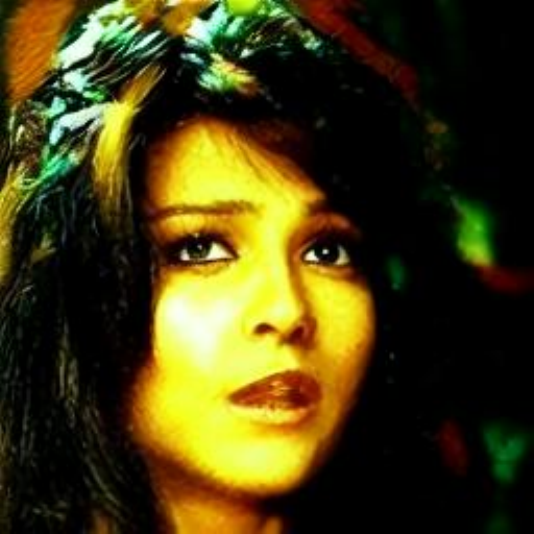}
%
\caption{\textbf{Varying the Number of Parallel Processors for Low Numbers of Steps.
} We fix the mixing parameter at $0.015$ and variance scaling at $1.35$. From left to right, each column corresponds to the number of parallel processors: $2$ (our case) to $10$. All images from the same row share the same initial random seed. \textbf{Rows 1-2:} DDPM with $10$ steps. \textbf{Rows 3-4:} DDPM with $20$ steps. \textbf{Rows 5-6:} LD with $20$ steps.  \textbf{Rows 7-8:} LD with $40$ steps. 
%
%
}
    \label{fig:num-proc}
    \end{figure}

In reference to Section~\ref{sec:param-change}, we illustrate the effects of changing: (i) the variance scaling in Figs.~\ref{fig:sweep-var} and~\ref{fig:sc-var-wr}; (ii) the mixing parameter in Fig.~\ref{fig:sweep-mix}; and (iii) the number of parallel processors in Fig.~\ref{fig:num-proc}. The image degradation by integrating the predictive value with the latent at step $t_k+1$ (instead of the latent at step $t_k$) is shown in Fig.~\ref{fig:back-int}.

\subsection{Extending SE2P to Multiple Processors}
\label{app:extend-SE2p}

Our extension of the SE2P algorithm (Algorithm~\ref{alg:main_SE2P}) to a number $P$ of parallel processors is in Algorithm~\ref{alg:main_SE2P_more}. Note that for two processors, the denoising is done in parallel at the consecutive evaluation steps $t_k$ and $t_k+1$ (see lines 13 to 17 in Algorithm~\ref{alg:main_SE2P}), whereas in Algorithm~\ref{alg:main_SE2P_more}, it is done in parallel at the consecutive evaluation steps $t_k,t_k+1,\cdots,t_k+P-1$. 
The technical notes in Appendix~\ref{app:technicalnotesS2EP} are easily extended to multiple parallel processors---for example, we highlight that every processor $p$ that computes its predictive value $\hat{\x}_{\operatorname{pred}}$ (line 9) does not need an extra evaluation of the diffusion model 
because of 
the previously stored variable $\mathbf{v}^{(p)}$ (line 7) computed during the parallel denoising process (line 17). 


\begin{algorithm}[t!]
  \caption{Extension of SE2P to $P$ Parallel Processors}
  \label{alg:main_SE2P_more}
\begin{algorithmic}[1]
    \STATE {\bfseries Input:} $(t_{N-1},\dots,t_{0})$, $(\beta_t)_{t=0}^T$, pre-trained model $\epsilon_\theta$, number of processors $P$, parameters $\rho>0$, $\gamma\in(0,1)$ 
    \STATE $\x^{(0)}_{N-1}\sim\cN(\zeros,I)$, $\x^{(p)}_{N-1}=\x^{(0)}_{N-1}$, $p=1,\dots,P-1$ 
%
%
    \FOR{$k=N-1,\cdots,0$}
        \IF{$k\neq N-1$}

        \FOR{$p=1,\dots,P-1$}

            \STATE $\hat{t}_{k} = t_k + P - p$ \# Proc $p-1$
            \STATE $\hat{\x}_{p-1} = (\x_{k}^{(p-1)}-\sqrt{1-\bar{\alpha}_{\hat{t}_{k}}}\mathbf{v}^{(p-1)})/\sqrt{\bar{\alpha}_{\hat{t}_{k}}}$ \# Proc $p-1$ 
            \STATE $\tilde{\mu}_{\operatorname{pred}}=(\sqrt{\bar{\alpha}_{\hat{t}_{k}-1}}\beta_{\hat{t}_{k}}\hat{\x}_{p-1}+\sqrt{\alpha_{\hat{t}_{k}}}(1-\bar{\alpha}_{{\hat{t}_{k}}-1})\x_{k}^{(p-1)})/\sqrt{1-\bar{\alpha}_{\hat{t}_{k}}}$ \# Proc $p-1$
            \STATE $\hat{\x}_{\operatorname{pred}}=\tilde{\mu}_{\operatorname{pred}}+\rho\cdot\sqrt{\tilde{\beta}_{\hat{t}_{k}}}\epsilon,~\epsilon\sim\cN(\zeros_n,I_n)$ \# Proc $p-1$
            \STATE $\x^{(p)}_{k} = \gamma\cdot\x^{(p)}_k + (1-\gamma)\cdot\hat{\x}_{\operatorname{pred}}$ \# Integration of information. Proc $p-1$ sends $\hat{\x}_{\operatorname{pred}}$ to Proc $p$. Proc $p$ 
            \STATE $\x^{(p-1)}_{k} = \x^{(p)}_{k}$ \# Proc $p$ sends $\hat{\x}^{(p)}_k$ to Proc $p-1$. Proc $p-1$
        \ENDFOR
        \ENDIF
        \STATE Proc $p$, $p=0,1,\dots,P-1$, fix the same random seed.
        \STATE {\bfseries PARALLEL} denoising for each Proc $p$, $p=0,1,\dots,P-1$
        %
        \begin{ALC@g}
        \STATE $t^{(p)}_{k} = t_k + P - 1 - p$ 
        \STATE $\mathbf{v}^{(p)}=\epsilon_{\theta}(\x^{(p)}_k,t^{(p)}_{k})$
        \STATE $\x^{(p)}_{k-1} = \operatorname{denoise}(t^{(p)}_{k},\x^{(p)}_k,\mathbf{v}^{(p)})$ \# Using previously defined seed 
        \end{ALC@g}
        \STATE {\bfseries end PARALLEL}
        %
        %
    \ENDFOR
    \STATE {\bfseries Return:} $\x_{0}^{(0)}$
%
\end{algorithmic}
\end{algorithm}

\end{document}